\documentclass[12pt,oneside,letterpaper]{report}

\newcommand{\draftfinal}[2]{\ifdefined\draftversion#1\else#2\fi}

\newcommand{\finalonly}[1]{\draftfinal{}{#1}}

\newcommand{\thesistitle}{Fundamental Novel Consistency Theory:\\ $\sH$-Consistency Bounds}
\newcommand{\thesisauthor}{Yutao Zhong}
\newcommand{\thesisadvisor}{Professor Mehryar Mohri}

\newcommand{\thesisdept}{Mathematics}
\newcommand{\gradmonth}{January}
\newcommand{\gradyear}{2025}
\newcommand{\thesisdedication}{}

\RequirePackage[margin=1in, includefoot, letterpaper]{geometry}

\RequirePackage[prologue]{xcolor}
\definecolor[named]{ThesisBlue}{cmyk}{1,0.1,0,0.1}
\definecolor[named]{ThesisYellow}{cmyk}{0,0.16,1,0}
\definecolor[named]{ThesisOrange}{cmyk}{0,0.42,1,0.01}
\definecolor[named]{ThesisRed}{cmyk}{0,0.90,0.86,0}
\definecolor[named]{ThesisLightBlue}{cmyk}{0.49,0.01,0,0}
\definecolor[named]{ThesisGreen}{cmyk}{0.20,0,1,0.19}
\definecolor[named]{ThesisPurple}{cmyk}{0.55,1,0,0.15}
\definecolor[named]{ThesisDarkBlue}{cmyk}{1,0.58,0,0.21}

\definecolor{SchoolColor}{rgb}{0.3412, 0.0235, 0.5490} 
\definecolor{chaptergrey}{rgb}{0.2600, 0.0200, 0.4600} 
\definecolor{midgrey}{rgb}{0.4, 0.4, 0.4}

\usepackage{chemformula}
\usepackage{array}

\RequirePackage[labelfont={bf,sf,small,singlespacing},
                textfont={sf,small,singlespacing},
                margin=0pt,
                figurewithin=chapter,
                tablewithin=chapter]{caption}

\usepackage{fix-cm}
\RequirePackage[raggedright,sc]{titlesec}
\definecolor{gray75}{gray}{0.75}
\newcommand{\hsp}{\hspace{20pt}}

\titleformat{\chapter}[hang]
{\Huge\sc}
{\textcolor{SchoolColor}{\thechapter}\hsp\textcolor{gray75}{|}\hsp}
{0pt}{\Huge\sc\raggedright}

\usepackage{setspace}

\finalonly{
  \doublespacing 
}

\usepackage{lipsum}

\input{defs}

\usepackage{comment}

\usepackage{hyperref}
\hypersetup{colorlinks,
  linkcolor=ThesisDarkBlue,
  citecolor=ThesisPurple,
  urlcolor=ThesisDarkBlue,
  filecolor=ThesisDarkBlue}
 
\usepackage[
    backend=biber,
    style=authoryear-comp,
    natbib=true,
    url=false, 
    doi=true,
    eprint=false
]{biblatex}
\newtheorem{lemma}[theorem]{Lemma}
 
 \usepackage{lscape}
 
\begin{document}

\pagenumbering{roman}
%
\thispagestyle{empty}
%

\vspace*{25pt}
\begin{center}

  {\Large
    \begin{doublespace}
      {\textcolor{SchoolColor}{\textsc{\thesistitle}}}
    \end{doublespace}
  }
  \vspace{.7in}

  by
  \vspace{.7in}

  \thesisauthor
  \vfill

  \begin{doublespace}
    \textsc{
    A dissertation submitted in partial fulfillment\\
    of the requirements for the degree of\\
    Doctor of Philosophy\\
    Department of \thesisdept\\
    New York University\\
    \gradmonth, \gradyear}
  \end{doublespace}
\end{center}
\vfill

\noindent\makebox[\textwidth]{\hfill\makebox[2.5in]{\hrulefill}}\\
\makebox[\textwidth]{\hfill\makebox[2.5in]{\hfill\thesisadvisor}}

\newpage

\thispagestyle{empty}
\vspace*{25pt}
\begin{center}
  \scshape \noindent \small \copyright \  \small  \thesisauthor \\
  All rights reserved, \gradyear
\end{center}
\vspace*{0in}
\newpage

\cleardoublepage
\phantomsection
\chapter*{Dedication}
\addcontentsline{toc}{chapter}{Dedication}
\vspace*{\fill}
\begin{center}
  \thesisdedication
   To my family.
\end{center}
\vfill
\newpage

\chapter*{Acknowledgements}
\addcontentsline{toc}{chapter}{Acknowledgements}

 First and foremost, I would like to express my deepest gratitude to my advisor, Prof. Mehryar Mohri, for
his unwavering guidance and support during my Ph.D. journey. Prof. Mohri is among the most
brilliant minds I have ever encountered, and every discussion with him has been truly
enlightening.

He taught us to be creative and ambitious, provided patient guidance through hands-on training,
and cultivated our sense of taste by steering us toward high-impact research. I deeply admire
his penetrating insights into machine learning, impeccable demeanor, sense of humor, and,
most importantly, his remarkable leadership, which serves as a lifelong inspiration. His example
has set a standard that I will strive to follow throughout my career.

I feel especially indebted to Prof. Mohri for being a role model in all aspects of research and life,
and for offering unwavering support at every step. It has been an immense honor to be his
student. To be honest, I could not have dreamed about having a better Ph.D. advisor.

Completing my Ph.D. and looking forward to a promising future would not have been possible
without his guidance. Thank you, Prof. Mohri!

I would like to express my heartfelt gratitude to Corinna Cortes for her invaluable guidance and encouragement throughout my academic journey. Corinna’s profound and comprehensive insights into machine learning have been a tremendous source of inspiration. I have greatly benefited from our collaboration and learned extensively from her about the art of conducting experiments and presenting experimental results effectively.

I am also deeply grateful to Pranjal Awasthi, my other primary collaborator during my academic journey. Our fruitful discussions on adversarial robustness have been immensely enriching, and I have truly enjoyed working with Pranjal during my internships at Google Research.

Additionally, I extend my sincere thanks to Prof. Esteban Tabak and Prof. Yanjun Han for their time and dedication as members of my thesis committee. I am especially thankful to Prof. Tabak for his exceptional support in his role as Director of Graduate Studies and for fostering our academic endeavors at the Courant Institute.

I also sincerely thank my peers and friends at the Courant Institute for their support and friendship throughout my Ph.D. journey. I am also deeply grateful to Michelle Shin and Gehan Abreu De Colon for their invaluable assistance and support as part of the administrative staff.

Finally, I would like to convey my heartfelt thanks to Prof. Georg Stadler for introducing me to the Courant Institute and to Prof. Jonathan Weare for serving as my faculty mentor during my first year. I am especially grateful to Prof. Stadler for his invaluable guidance and mentorship during my undergraduate research projects. I am also profoundly appreciative of the support and encouragement both Prof. Stadler and Prof. Weare provided in helping me prepare for my qualifying exams.


\chapter*{Abstract}
\addcontentsline{toc}{chapter}{Abstract}

In machine learning, the loss functions optimized during training
often differ from the target loss that defines task performance. This
is typically due to computational intractability or a lack of
desirable properties like differentiability or smoothness in the
target loss. But what guarantees can we
rely on for the target loss estimation error when resorting to a surrogate loss?

We present an in-depth study of the target loss estimation error of a
predictor with respect to its surrogate loss estimation error. Our
analysis leads to \emph{$\sH$-consistency bounds}, which are
guarantees accounting for the hypothesis set $\sH$ adopted. These
bounds offer stronger guarantees than previously established measures
such as Bayes-consistency, $\sH$-calibration, or standard
$\sH$-consistency, and are more informative than \emph{excess error
bounds} for unrestricted hypothesis sets.

We begin with a detailed analysis of $\sH$-consistency bounds in
binary classification, establishing both distribution-dependent and
distribution-independent results. Our theorems demonstrate tight
bounds, assuming convexity, and generalize earlier excess error
bounds. We provide explicit bounds for the zero-one loss using
different surrogate losses and hypothesis sets, including linear
models and single-layer neural networks. We also analyze the
adversarial setting, deriving bounds for surrogates like
$\rho$-margin and sigmoid loss, with enhanced results under
natural distributional assumptions. Simulation results confirm the
tightness of our theoretical guarantees.

Extending to multi-class classification, we present the first
$\sH$-consistency bounds for a variety of surrogate losses, covering
both non-adversarial and adversarial scenarios. These include max
losses, sum losses, and constrained losses with diverse auxiliary
functions. Our results demonstrate that in some cases, non-trivial
$\sH$-consistency bounds are unattainable. Our novel proof techniques
may have broader applications in similar analyses.

We also investigate a broad class of loss functions, \emph{comp-sum
losses}, including cross-entropy, generalized cross-entropy, and mean
absolute error. We derive the first $\sH$-consistency bounds for these
losses, proving their tightness and clarifying their dependence on
minimizability gaps. For comp-sum losses, we provide a detailed gap
analysis and introduce smooth adversarial variants, showing their
effectiveness in adversarial contexts and leading to new robust
learning algorithms.

Beyond theoretical contributions, we present empirical evaluations,
showing that our adversarial methods outperform the
state-of-the-art while maintaining high non-adversarial accuracy.

The specific proofs required for the previously mentioned results
highlight the need for more general techniques. To address this, we
develop a comprehensive framework for deriving $\sH$-consistency
bounds across various surrogate losses, with extensions to multi-class
classification. This framework introduces new characterizations for
constrained and comp-sum losses, relaxing completeness assumptions and
incorporating more realistic bounded hypothesis sets. Our approach is
grounded in explicit error transformations, which we apply to several
concrete examples. Additionally, we identify limitations in recent
multi-class $\sH$-consistency bounds for cross-entropy and provide
stronger, more reliable guarantees.

Finally, we examine the \emph{growth rates} of $\sH$-consistency and
excess error bounds across various surrogate losses, that is how
quickly the function of the surrogate estimation loss upper-bounding
the target estimation loss increases near zero.  In binary
classification, we establish a \emph{universal} square-root growth
rate near zero for smooth margin-based surrogates, generalizing to
excess error bounds with novel upper bounds. This extends to
multi-class classification, where we prove a similar universal rate
for smooth comp-sum and constrained losses, guiding the choice of
surrogates based on class number and minimizability gaps. Our analysis
concludes with a comprehensive examination of minimizability gaps,
highlighting their role in distinguishing between surrogate losses and
guiding practical selection.

\ignore{
We present a detailed study of the
  estimation error with respect to the loss tailored to a task, in
terms of surrogate loss estimation errors.  We refer to such
guarantees as \emph{$\sH$-consistency bounds}, since
they account for the hypothesis set $\sH$ adopted. These guarantees
are significantly stronger than Bayes-consistency, $\sH$-calibration or
$\sH$-consistency presented in previous work. They are also
more informative than similar \emph{excess error bounds} derived in
the literature, that is the special case where $\sH$ is
the family of all measurable functions.

We first present a comprehensive study of $\sH$-consistency bounds for
 binary classification. We prove general theorems providing such guarantees, for both the
distribution-dependent and distribution-independent settings.  We show
that our bounds are tight, modulo a convexity assumption. We also show
that previous excess error bounds can be recovered as special cases of
our general results.
We then present a series of explicit bounds in the case of the
zero-one loss, with multiple choices of the surrogate loss and for
both the family of linear functions and neural networks with one
hidden-layer. We further prove more favorable distribution-dependent
guarantees in that case. We also present a series of explicit bounds
in the case of the adversarial loss, with surrogate losses based on
the supremum of the $\rho$-margin, hinge or sigmoid loss and for the
same two general hypothesis sets. Here too, we prove several
enhancements of these guarantees under natural distributional
assumptions.  Finally, we report the results of simulations
illustrating our bounds and their tightness.

We also present an extensive study of $\sH$-consistency bounds for
multi-class classification. We give a series of new
$\sH$-consistency bounds for surrogate multi-class losses, including
max losses, sum losses, and constrained losses, both in the
non-adversarial and adversarial cases, and for different
differentiable or convex auxiliary functions used. We also prove that no
non-trivial $\sH$-consistency bound can be given in some cases.  To
our knowledge, these are the first $\sH$-consistency bounds proven for
the multi-class setting. Our proof techniques are also novel and
likely to be useful in the analysis of other such guarantees. 

We further present a theoretical
analysis of a broad family of loss functions, comp-sum losses,
that includes cross-entropy (or logistic loss), generalized
cross-entropy, the mean absolute error and other 
cross-entropy-like loss functions.
We give the first $\sH$-consistency bounds for these loss functions. We further show that our
bounds are tight. These bounds depend on quantities called
minimizability gaps. To make them more explicit, we give a
specific analysis of these gaps for comp-sum losses.
We also introduce a new family of loss functions, smooth
adversarial comp-sum losses, that are derived from their comp-sum
counterparts by adding in a related smooth term.  We show that these
loss functions are beneficial in the adversarial setting by proving
that they admit $\sH$-consistency bounds.  This leads to new
adversarial robustness algorithms that consist of minimizing a
regularized smooth adversarial comp-sum loss.
While our main purpose is a theoretical analysis, we also present an
extensive empirical analysis comparing comp-sum losses. We further
report the results of a series of experiments demonstrating that our
adversarial robustness algorithms outperform the current
state-of-the-art, while also achieving a superior non-adversarial
accuracy.

For all the results presented above, determining if $\sH$-consistency bounds hold
and deriving these bounds have required a specific proof and analysis
for each surrogate loss. Can we derive more general tools and
characterizations? Next, we provide both a general characterization
and an extension of $\sH$-consistency bounds for multi-class
classification.
We present new and tight $\sH$-consistency bounds for both the family
of constrained losses and that of comp-sum losses, which covers the
familiar cross-entropy, or logistic loss applied to the outputs of a
neural network. We further extend our analysis beyond the completeness
assumptions adopted in previous studies and cover more realistic
bounded hypothesis sets.  Our characterizations are based on error
transformations, which are explicitly defined for each formulation. We
illustrate the application of our general results through several
special examples. A by-product of our analysis is the observation that
a recently derived multi-class $\sH$-consistency bound for
cross-entropy reduces to an excess bound and is not
significant. Instead, we prove a much stronger and more significant
guarantee.

Finally, we present a comprehensive analysis of the growth rate of
$\sH$-consistency bounds (and excess error bounds) for various
surrogate losses used in classification. We prove a square-root growth
rate near zero for smooth margin-based surrogate losses in binary
classification, providing both upper and lower bounds under mild
assumptions. This result also translates to excess error bounds.  Our
lower bound requires weaker conditions than those in previous work for
excess error bounds, and our upper bound is entirely novel.  Moreover,
we extend this analysis to multi-class classification with a series of
novel results, demonstrating a universal square-root growth rate for
smooth comp-sum and constrained losses, covering common
choices for training neural networks in multi-class classification.
Given this universal rate, we turn to the question of choosing among
different surrogate losses.  We first examine how $\sH$-consistency
bounds vary across surrogates based on the number of classes. Next,
ignoring constants and focusing on behavior near zero, we identify
minimizability gaps as the key differentiating factor in these
bounds. Thus, we thoroughly analyze these gaps, to guide surrogate
loss selection, covering: comparisons across different comp-sum
losses, conditions where gaps become zero, and general conditions
leading to small gaps.  Additionally, we demonstrate the key role of
minimizability gaps in comparing excess error bounds and
$\sH$-consistency bounds.
}

\newpage

\tableofcontents

\cleardoublepage
\phantomsection
\addcontentsline{toc}{chapter}{List of Figures}
\listoffigures
\newpage

\cleardoublepage
\phantomsection
\addcontentsline{toc}{chapter}{List of Tables}
\listoftables
\newpage

\cleardoublepage
\phantomsection
\addcontentsline{toc}{chapter}{List of Appendices}
\listofappendices
\newpage

\pagenumbering{arabic} 




\chapter*{Introduction} \label{ch0}
\addcontentsline{toc}{chapter}{Introduction}

Most learning algorithms rely on optimizing a surrogate loss function
distinct from the \emph{target loss function} tailored to the task
considered. This is typically because the target loss function is
computationally hard to optimize or because it does not admit
favorable properties, such as differentiability or smoothness, crucial
to the convergence of optimization algorithms. But, what guarantees
can we count on for the target loss estimation error, when minimizing
a surrogate loss estimation error?

A desirable property of a surrogate loss function, often referred to
in that context is \emph{Bayes-consistency}. It requires that
asymptotically, nearly optimal minimizers of the surrogate excess
error also nearly optimally minimize the target excess error
\citep{steinwart2007compare}. This property holds for a broad family
of convex surrogate losses of the standard binary and
multi-class classification losses
\citep{Zhang2003,bartlett2006convexity,zhang2004statistical,tewari2007consistency,steinwart2007compare,MohriRostamizadehTalwalkar2018}. Consistency studies the asymptotic relation between the surrogate excess error and the target excess error while excess error bounds study the quantitative relation between them and thus is stronger. They both consider the  hypothesis set of all measurable functions. \citet{Zhang2003}, \citet{bartlett2006convexity}, and \citet{steinwart2007compare} studied consistency via the lens of calibration 
and showed that calibration and consistency are equivalent in the standard binary classification when considering the hypothesis set of all measurable functions.

\citet{Zhang2003} studied the closeness to the optimal excess error of the zero-one loss minimizers of convex surrogates.
\citet{bartlett2006convexity} extended the results of \citet{Zhang2003} and developed a general methodology for coming up with
quantitative bounds between the excess error corresponding to the zero-one loss and that of margin-based surrogate loss functions for all distributions.
In a more recent work, \citet{MohriRostamizadehTalwalkar2018} simplified these results and
provided different proofs for the excess error bounds of various
loss functions widely used in practice. Calibration and consistency analysis have also been extended to the multi-class settings \citep{zhang2004statistical,tewari2007consistency}, to ranking problems \citep{uematsu2011theoretically,gao2015consistency}, and to the context of regression \citep{Caponnetto2005,ChristmannSteinwart2007,steinwart2007compare}

But, Bayes-consistency is not relevant when learning with a hypothesis
set $\sH$ distinct from the family of all measurable
functions. Therefore, a new hypothesis set-dependent notion namely, $\sH$-consistency, has been proposed and explored in the more recent literature \citep{long2013consistency,kuznetsov2014multi,zhang2020bayes}. In particular, \citet{long2013consistency} argued that $\sH$-consistency is a more useful notion than consistency by empirically showing that certain loss functions that are $\sH$-consistent but not Bayes consistent can perform significantly better than a loss function known to be Bayes consistent. The work of \citet{kuznetsov2014multi} extended the $\sH$-consistency results in \citep{long2013consistency} to the case of structured prediction and provided positive results for $\sH$-consistency of several multi-class ensemble algorithms.

In a recent work \citet{zhang2020bayes} investigated the empirical phenomenon in \citep{long2013consistency} and designed a class of piecewise linear scoring functions such that minimizing a surrogate that is not $\sH$-consistent over this larger class yields $\sH$-consistency of linear models. For linear predictors, more general margin-based properties of convex surrogate losses are also studied in \citep{long2011learning,ben2012minimizing}. Aiming for such margin-based error guarantees, \citet{ben2012minimizing} argued that the hinge loss is optimal among convex losses.

Most recently, the notion of $\sH$-consistency along with $\sH$-calibration have also been studied in the context of adversarially robust classification \citep{bao2020calibrated,awasthi2021calibration,awasthi2021finer}. In the adversarial scenario, in contrast to standard classification, the target loss is the adversarial zero-one loss \citep{goodfellow2014explaining,madry2017towards,carlini2017towards,tsipras2018robustness,shafahi2019adversarial,wong2020fast,AwasthiMaoMohriZhong2023,awasthi2023dc}. This corresponds to the worst zero-one loss incurred over an adversarial perturbation of $x$ within a $\gamma$-ball as measured in a norm, typically $\ell_p$ for $p\in[1,+\infty]$. The adversarial loss presents new challenges and makes the consistency analysis significantly more complex. 

The work of \citet{bao2020calibrated} initiated the study of $\sH$-calibration with respect to the adversarial zero-one loss for the linear models. They showed that convex surrogates are not calibrated and introduced a class of nonconvex margin-based surrogate losses. They then provided sufficient conditions for such nonconvex losses to be calibrated in the linear case. The work of \citet{awasthi2021calibration} extended the results in \citep{bao2020calibrated} to the general nonlinear hypothesis sets and pointed out that although $\sH$-calibration is a necessary condition of $\sH$-consistency, it is not sufficient in the adversarial scenario. They then proposed sufficient conditions which guarantee calibrated losses to be consistent in the setting of adversarially robust classification. 

Nevertheless,
Bayes-consistency and $\sH$-consistency are both asymptotic properties and 
thus do not provide any guarantee for approximate minimizers learned
from finite samples. Instead, we will consider upper bounds on the target estimation error
expressed in terms of the surrogate estimation error, which we refer
to as \emph{$\sH$-consistency bounds}, since they
account for the hypothesis set $\sH$ adopted. These guarantees are
significantly stronger than $\sH$-calibration or $\sH$-consistency or
some margin-based properties of convex surrogate losses for linear
predictors studied by \citet{ben2012minimizing} and
\citet{long2011learning}. They are also more informative than similar
\emph{excess error bounds} derived in the literature, which correspond
to the special case where $\sH$ is the family of all measurable
functions \citep{Zhang2003,bartlett2006convexity} (see also
\cite{MohriRostamizadehTalwalkar2018}[section~4.7]).

The rest of this work is organized as follows. 

In Chapter~\ref{ch2}, we present an exhaustive study of $\sH$-consistency
bounds for binary classification. We prove general theorems providing such guarantees, which could be
used in both distribution-dependent and distribution-independent
settings.  We show that our bounds are
tight, modulo a convexity assumption. We also
show that previous excess error bounds can be recovered as special
cases of our general results. We then present a series of explicit bounds in the case of the $0/1$
loss, with multiple choices of the
surrogate loss and for both the family of linear functions and that of neural networks with one
hidden-layer. We further prove more
favorable distribution-dependent guarantees in that case. We also present a detailed analysis of the \emph{adversarial loss}. We show that there can be no non-trivial
adversarial $\sH$-consistency bound for
supremum-based convex loss functions and supremum-based sigmoid loss
function, under mild assumptions that hold for most hypothesis sets
used in practice. These results imply
that the loss functions commonly used in practice for optimizing the
adversarial loss cannot benefit from any useful $\sH$-consistency
bound guarantee! These are novel results that go beyond the
negative ones given for convex surrogates by
\citet{awasthi2021calibration}. We also present new $\sH$-consistency bounds for the
adversarial loss with surrogate losses based on the supremum of the
$\rho$-margin loss, for linear hypothesis sets and the family of neural networks with one
hidden-layer. Here too, we prove several
enhancements of these guarantees under some natural distributional
assumptions. Our results help compare different surrogate loss functions of the zero-one loss or adversarial loss, given the specific hypothesis set
used, based on the functional form of their $\sH$-consistency bounds. These results, combined with approximation error
properties of surrogate losses, can help select the most suitable
surrogate loss in practice. In addition to several general theorems, our study required a careful
inspection of the properties of various surrogate loss functions and
hypothesis sets. We report the results of simulations
illustrating our bounds and their tightness.

In Chapter~\ref{ch3}, we presented a comprehensive study of $\sH$-consistency bounds for
multi-class classification. We show that, in general, no
non-trivial $\sH$-consistency bounds can be derived for multi-class
\emph{max losses} such as those of \citet{crammer2001algorithmic},
when used with a convex loss auxiliary function such as the hinge
loss.  On the positive side, we prove multi-class $\sH$-consistency
bounds for max losses under a realizability assumption and give
multi-class $\sH$-consistency bounds using as an auxiliary function
the $\rho$-margin loss, without requiring a realizability assumption. For \emph{sum losses}, that is multi-class losses such as that of
\citet{weston1998multi}, we give a series of results, including a
negative result when using as auxiliary function the hinge-loss, and
$\sH$-consistency bounds when using the exponential loss, the squared
hinge-loss, and the $\rho$-margin loss. We also present a series of results for the so-called
\emph{constrained losses}, such as the loss function adopted by
\citet{lee2004multicategory} in the analysis of multi-class SVM. Here,
we prove multi-class $\sH$-consistency bounds when using as an
auxiliary function the hinge-loss, the squared hinge-loss, the
exponential loss, and the $\rho$-margin loss. We further give multi-class \emph{adversarial} $\sH$-consistency
bounds for all three of the general multi-class losses just mentioned
(max losses, sum losses and constrained losses). All of our results are novel,
including our proof techniques. Our results are given for the
hypothesis set $\sH$ being the family of all measurable functions, the
family of linear functions, or the family of one-hidden-layer ReLU
neural networks. The binary classification results in Chapter~\ref{ch2}
do not readily extend to the multi-class setting since the study of
calibration and conditional risk is more complex, the form of the
surrogate losses is more diverse, and in general the analysis is more
involved and requires entirely novel proof techniques in the
multi-class setting.

In Chapter~\ref{ch4}, we present the first $\sH$-consistency bounds for the
logistic loss, which can be used to derive directly guarantees for
current algorithms used in the machine learning community. More
generally, we will consider a broader family of loss functions that we
refer to as \emph{comp-sum losses}, that is loss functions obtained by
composition of a concave function, such as logarithm in the case of
the logistic loss, with a sum of functions of differences of score,
such as the negative exponential. We prove $\sH$-consistency bounds
for a wide family of comp-sum losses, which includes as special cases
the logistic loss
\citep{Verhulst1838,Verhulst1845,Berkson1944,Berkson1951}, the
\emph{generalized cross-entropy loss} \citep{zhang2018generalized},
and the \emph{mean absolute error loss} \citep{ghosh2017robust}.
We further show that our bounds are \emph{tight} and thus cannot be
improved. $\sH$-consistency bounds are expressed in terms of a quantity called
\emph{minimizability gap}, which only depends on the loss function and
the hypothesis set $\sH$ used. It is the difference of the best-in
class expected loss and the expected pointwise infimum of the loss.
For the loss functions we consider, the minimizability gap vanishes
when $\sH$ is the full family of measurable functions. However, in
general, the gap is non-zero and plays an important role, depending on
the property of the loss function and the hypothesis set. Thus, to
better understand $\sH$-consistency bounds for comp-sum losses, we
specifically analyze their minimizability gaps, which we
use to compare their guarantees. A recent challenge in the application of neural networks is their
robustness to imperceptible perturbations
\citep{szegedy2013intriguing}. While neural networks trained on large
datasets often achieve a remarkable performance
\citep{SutskeverVinyalsLe2014,KrizhevskySutskeverHinton2012}, their
accuracy remains substantially lower in the presence of such
perturbations. One key issue in this scenario is the definition of a
useful surrogate loss for the adversarial loss. To tackle this
problem, we introduce a family of loss functions designed for
adversarial robustness that we call \emph{smooth adversarial comp-sum
loss functions}. These are loss functions derived from their comp-sum
counterparts by augmenting them with a natural smooth term.  We show
that these loss functions are beneficial in the adversarial setting by
proving that they admit $\sH$-consistency bounds. This leads to a
family of algorithms for adversarial robustness that consist of
minimizing a regularized smooth adversarial comp-sum loss. While our main purpose is a theoretical analysis, we also present an
extensive empirical analysis. We compare the empirical performance of
comp-sum losses for different tasks and relate that to their
theoretical properties. We further report the results of experiments
with the CIFAR-10, CIFAR-100 and SVHN datasets comparing the
performance of our algorithms based on smooth adversarial comp-sum
losses with that of the state-of-the-art algorithm for this task
\textsc{trades} \citep{zhang2019theoretically}. The results show that
our adversarial algorithms outperform \textsc{trades} and also achieve
a substantially better non-adversarial (clean) accuracy.

In Chapter~\ref{ch5}, we provided both a general characterization
and an extension of $\sH$-consistency bounds to cover more realistic
bounded hypothesis sets for multi-class classification. Previous
approaches to deriving these bounds required the development of new
proofs for each specific case. In contrast, we introduce the general
concept of an \emph{error transformation function} that serves as a
very general tool for deriving such guarantees with tightness
guarantees. We show that deriving an $\sH$-consistency bound for
comp-sum losses and constrained losses for both complete and bounded
hypothesis sets can be reduced to the calculation of their
corresponding error transformation function. Our general tools and
tight bounds show several remarkable advantages: first, they improve
existing bounds for complete hypothesis sets previously proven in
\citep{awasthi2022multi}; second, they encompass all previously
comp-sum and constrained losses studied thus far as well as many new
ones \citep{awasthi2022Hconsistency,mao2023cross}; third, they extend
beyond the completeness assumption adopted in previous work; fourth,
they provide novel guarantees for bounded hypothesis sets; and, finally,
they help prove a much stronger and more significant guarantee for
logistic loss with linear hypothesis set than
\citep{zheng2023revisiting}.

In Chapter~\ref{ch6}, we presents a comprehensive analysis of the growth rate of
$\sH$-consistency bounds (and excess error bounds) for various
surrogate losses used in classification. We prove a square-root growth
rate near zero for smooth margin-based surrogate losses in binary
classification, providing both upper and lower bounds under mild
assumptions. This result also translates to excess error bounds.  Our
lower bound requires weaker conditions than those in previous work for
excess error bounds, and our upper bound is entirely novel.  Moreover,
we extend this analysis to multi-class classification with a series of
novel results, demonstrating a universal square-root growth rate for
smooth \emph{comp-sum} and \emph{constrained losses}, covering common
choices for training neural networks in multi-class classification. Given this universal rate, we turn to the question of choosing among
different surrogate losses.  We first examine how $\sH$-consistency
bounds vary across surrogates based on the number of classes. Next,
ignoring constants and focusing on behavior near zero, we identify
\emph{minimizability gaps} as the key differentiating factor in these
bounds. Thus, we thoroughly analyze these gaps, to guide surrogate
loss selection, covering: comparisons across different comp-sum
losses, conditions where gaps become zero, and general conditions
leading to small gaps.  Additionally, we demonstrate the key role of
minimizability gaps in comparing excess error bounds and
$\sH$-consistency bounds.

This thesis is based on \citet*{awasthi2022Hconsistency,awasthi2022multi,mao2023cross,MaoMohriZhong2023characterization,mao2024universal}.


\chapter{Binary Classification} \label{ch2}
In this chapter, we present an exhaustive study of $\sH$-consistency bounds for binary classification.
We prove general theorems providing such guarantees, which could be
used in both distribution-dependent and distribution-independent
settings (Section~\ref{sec:general}).  We show that our bounds are
tight, modulo a convexity assumption
(Section~\ref{sec:non-adv-general} and \ref{sec:adv-general}). We also
show that previous excess error bounds can be recovered as special
cases of our general results (Section~\ref{sec:all}).

We then present a series of explicit bounds in the case of the $0/1$
loss (Section~\ref{sec:non-adv}), with multiple choices of the
surrogate loss and for both the family of linear functions
(Section~\ref{sec:non-adv-lin}) and that of neural networks with one
hidden-layer (Section~\ref{sec:non-adv-NN}). We further prove more
favorable distribution-dependent guarantees in that case
(Section~\ref{sec:noise-non-adv}).

We also present a detailed analysis of the \emph{adversarial loss}
(Section~\ref{sec:adv}). We show that there can be no non-trivial
adversarial $\sH$-consistency bound for
supremum-based convex loss functions and supremum-based sigmoid loss
function, under mild assumptions that hold for most hypothesis sets
used in practice (Section~\ref{sec:negative}). These results imply
that the loss functions commonly used in practice for optimizing the
adversarial loss cannot benefit from any useful $\sH$-consistency
bound guarantee! These are novel results that go beyond the
negative ones given for convex surrogates by
\citet{awasthi2021calibration}.

We present new $\sH$-consistency bounds for the
adversarial loss with surrogate losses based on the supremum of the
$\rho$-margin loss, for linear hypothesis sets
(Section~\ref{sec:adv-lin}) and the family of neural networks with one
hidden-layer (Section~\ref{sec:adv-NN}). Here too, we prove several
enhancements of these guarantees under some natural distributional
assumptions (Section~\ref{sec:noise-adv}).

Our results help compare different surrogate loss functions of the zero-one loss or adversarial loss, given the specific hypothesis set
used, based on the functional form of their $\sH$-consistency bounds. These results, combined with approximation error
properties of surrogate losses, can help select the most suitable
surrogate loss in practice.
In addition to several general theorems, our study required a careful
inspection of the properties of various surrogate loss functions and
hypothesis sets. Our proofs and techniques could be adopted for the
analysis of many other surrogate loss functions and hypothesis sets.

In Section~\ref{sec:simulations}, we report the results of simulations
illustrating our bounds and their tightness. We start
with some preliminary definitions and notation.

The presentation in this chapter is based on \citep{awasthi2022Hconsistency}.

\section{Preliminaries}

Let $\sX$ denote the input space and $\sY = \curl*{-1,+1}$ the binary
label space.
We will denote by $\sD$ a distribution over $\sX \times \sY$, by $\sP$
a set of such distributions and by $\sH$ a hypothesis set of functions
mapping from $\sX$ to $\Rset$. The \emph{generalization error} and
\emph{minimal generalization error} for a loss function $\ell(h,x,y)$
are defined as $\sR_{\ell}(h)=\E_{(x,y)\sim
  \sD}\bracket*{\ell(h,x,y)}$ and $\sR_{\ell}^*(\sH) = \inf_{h\in
  \sH}\sR_{\ell}(h)$.  Let $\sH_{\mathrm{all}}$ denote the hypothesis
set of all measurable functions.  The \emph{excess error} of a
hypothesis $h$ is defined as the difference
$\sR_{\ell}(h)-\sR_{\ell}^*\paren*{\sH_{\mathrm{all}}}$, which can be
decomposed into the sum of two terms, the \emph{estimation error} $\paren*{\sR_{\ell}(h)-\sR_{\ell}^*(\sH)}$ and
\emph{approximation error} $\paren*{\sR_{\ell}^*(\sH) - \sR_{\ell}^*\paren*{\sH_{\mathrm{all}}}}$:
\begin{equation}
\label{eq:excess-split}
    \sR_{\ell}(h)-\sR_{\ell}^*\paren*{\sH_{\mathrm{all}}}
    = \paren*{\sR_{\ell}(h)-\sR_{\ell}^*(\sH)}+ \paren*{\sR_{\ell}^*(\sH) - \sR_{\ell}^*\paren*{\sH_{\mathrm{all}}}}.
\end{equation}
Given two loss functions $\ell_1$ and $\ell_2$, a fundamental question is whether $\ell_1$ is \emph{consistent} with respect to $\ell_2$ for a hypothesis set $\sH$ and a set of distributions $\sP$ \citep{bartlett2006convexity,steinwart2007compare,long2013consistency,bao2020calibrated,awasthi2021calibration}.
\begin{definition}[\textbf{$(\sP,\sH)$-consistency}]
\ignore{
Let $\sP$ be a set of distributions over $\sX\times\sY$. 
}
We say that $\ell_1$ is \emph{$(\sP,\sH)$-consistent} with respect to $\ell_2$, if, for all distributions $\sD\in \sP$ and sequences $\{h_n\}_{n\in \Nset}\subset \sH$, we have 
\begin{equation}
\label{eq:consistency}
    \lim_{n \to +\infty}
    \sR_{\ell_1}(h_n)-\sR_{\ell_1}^*(\sH) = 0
    \Rightarrow 
    \lim_{n \to +\infty}
    \sR_{\ell_2}(h_n)-\sR_{\ell_2}^*(\sH) = 0.
\end{equation}
\end{definition}
\ignore{ $\sP$ can be chosen to be any set of interest, such as the
  set of distributions that are
  \emph{realizable}~\citep{long2013consistency,zhang2020bayes,awasthi2021calibration},
  or more generally, the set of distributions that verify certain low
  noise conditions. When $\sP$ consists of a set of distributions all
  supported on a singleton, $(\sP,\sH)$-Consistency reduces to
  \emph{calibration}~\citep{bartlett2006convexity,steinwart2007compare,awasthi2021calibration}.} 
\vspace{-0.3pt}
We will denote by $\Phi$ a margin-based loss if a loss function $\ell$ can be represented as $\ell(h,x,y) = \Phi\paren*{yh(x)}$ and by
$\wt{\Phi}\colon=\sup_{x'\colon \|x-x'\|_p\leq \gamma}\Phi\paren*{y
  h(x')}$, $p\in [1,+\infty]$, the supremum-based counterpart.  In
the standard binary classification, $\ell_2$ is the $0/1$ loss
$\ell_{0-1}\colon=\mathds{1}_{\sign(h(x))\neq y}$, where
$\sign(\alpha) = \mathds{1}_{\alpha \geq 0} - \mathds{1}_{\alpha < 0}$
and $\ell_1$ is the margin-based loss for some function $\Phi\colon
\Rset \to \Rset_{+}$, typically convex. In the adversarial binary
classification, $\ell_2$ is the adversarial $0/1$ loss
$\ell_{\gamma}\colon=\sup_{x'\colon \|x-x'\|_p\leq
  \gamma}\mathds{1}_{y h(x') \leq 0}$, for some $\gamma \in (0, 1)$
and $\ell_1$ is the supremum-based margin loss $\wt{\Phi}$.

Let $B_p^d(r)$ denote the $d$-dimensional $\ell_p$-ball with radius
$r$: $B_p^d(r) = \big\{z \in \Rset^d \mid \norm*{z}_p\leq r\big\}$.
\ignore{where $\norm*{z}_p\colon =\bracket*{\sum_{i=1}^d
    \abs*{z_i}^p}^{\frac{1}{p}}$, $p\in [1, +\infty]$.}
Without loss of generality, we consider $\sX=B_p^d(1)$\ignore{ and the perturbation size $\gamma \in (0,1)$ throughout the paper}. 
Let $p, q \in[1, +\infty]$ be conjugate numbers, that is
$\frac{1}{p} + \frac{1}{q} = 1$. We will specifically study the family
of linear hypotheses $\sH_{\mathrm{lin}}=\big\{x\mapsto w \cdot x + b
\mid \norm*{w}_q\leq W,\abs*{b}\leq B\big\}$ and one-hidden-layer
ReLU networks $\sH_{\mathrm{NN}} = \big\{x\mapsto \sum_{j =
  1}^n u_j(w_j \cdot x+b)_{+} \mid \|u \|_{1}\leq
\Lambda,\|w_j\|_q\leq W, \abs*{b}\leq B\big\}$, where $(\cdot)_+ = \max(\cdot, 0)$, with $W$, $\Lambda$, and $B$ as positive constants.
Finally, for any $\e > 0$, we will denote by $\tri*{t}_{\e}$ the
$\e$-truncation of $t \in \Rset$ defined by $t\mathds{1}_{t>\e}$.

\section{\texorpdfstring{$\sH$}{H}-consistency bound definitions}

$(\sP,\sH)$-Consistency is an asymptotic relation between two loss
functions. However, we are interested in a more quantitative relation
in many applications. This motivates the study of
\emph{$\sH$-consistency bound}.

\begin{definition}[\textbf{$\sH$-consistency bound}]
\ignore{Let $\sP$ be a set of distributions over $\sX\times\sY$.}  If
for some non-decreasing function $f\colon \Rset_{+}\to \Rset_{+}$, a bound of the
following form holds for all $h\in \sH$ and $\sD\in \sP$:
\begin{align}
\label{eq:est-bound}
    \sR_{\ell_2}(h) - \sR_{\ell_2}^*(\sH)
    \leq f\paren*{\sR_{\ell_1}(h)-\sR_{\ell_1}^*(\sH)},
\end{align}
then, we call it an \emph{$\sH$-consistency bound}. Furthermore, if $\sP$ consists of all distributions over
$\sX\times\sY$, we say that the bound is
\emph{distribution-independent}.
\end{definition}
When $\sH=\sH_{\mathrm{all}}$ and $\sP$ is the set of all
distributions, a bound of the form \eqref{eq:est-bound} is also called
a \emph{consistency excess error bound}. Note when $f(0)= 0$ and $f$
is continuous at $0$, the $\sH$-consistency bound \eqref{eq:est-bound}
implies $\sH$-consistency \eqref{eq:consistency}. Thus,
$\sH$-consistency bounds provide stronger quantitative results
than consistency and calibration. Furthermore, there is a fundamental
reason to study such bounds from the statistical learning point of
view: they can be turned into more favorable generalization bounds for
the target loss $\ell_2$ than the excess error bound. For example,
when $\sP$ is the set of all distributions, by
\eqref{eq:excess-split}, relation \eqref{eq:est-bound} implies that, for all $h\in \sH$, the following inequality holds:
\begin{equation}
\label{eq:est-gen}
    \sR_{\ell_2}
    (h)
    -
    \sR_{\ell_2}^*\paren*{\sH_{\mathrm{all}}}
    \leq
    f\paren*{\sR_{\ell_1}
    (h)
    -
    \sR_{\ell_1}^*(\sH)}
    +
    \sR_{\ell_2}^*(\sH)
    -
    \sR_{\ell_2}^*\paren*{\sH_{\mathrm{all}}}.
\end{equation}
Similarly, the excess error bound can be
written as follows:
\begin{equation}
\label{eq:gen}
    \sR_{\ell_2}
    (h)
    -
    \sR_{\ell_2}^*\paren*{\sH_{\mathrm{all}}}
    \leq
     f\paren*{\sR_{\ell_1}
     (h)
     -
     \sR_{\ell_1}^*(\sH)
     +
     \sR_{\ell_l}^*(\sH)
     -
     \sR_{\ell_l}^*\paren*{\sH_{\mathrm{all}}}}. 
\end{equation}
If we further bound the estimation error $[\sR_{\ell_1}(h) -
  \sR_{\ell_1}^*(\sH)]$ by the empirical error plus a complexity term,
\eqref{eq:est-gen} and \eqref{eq:gen} both turn into generalization
bounds. However, the generalization bound obtained by
\eqref{eq:est-gen} is linearly dependent on the approximation error of
target loss $\ell_2$, while the one obtained by \eqref{eq:gen} depends
on the approximation error of the surrogate loss $\ell_1$ and can
potentially be worse than linear dependence.
Moreover, \eqref{eq:est-gen} can be easily used to compare different
surrogates by directly comparing the corresponding mapping
$f$. However, only comparing the mapping $f$ for different surrogates
in \eqref{eq:gen} is not sufficient since the approximation errors of
surrogates may differ as well.

\ignore{ As an example,
  \citep{Zhang2003,MohriRostamizadehTalwalkar2018} show the
  consistency excess error bound for the logistic loss
  $\Phi_{\mathrm{log}}(t)\colon=\log_2(1+e^{-t})$ with respect to the
  $0/1$ loss is
\begin{align*}
  \forall h\in \sH_{\mathrm{all}}, \quad \sR_{\ell_{0-1}}(h)
  - \sR_{\ell_{0-1}}^*\paren*{\sH_{\mathrm{all}}}\leq  \sqrt{2}\,\bracket*{\sR_{\Phi_{\mathrm{log}}}(h)
    - \sR_{\Phi_{\mathrm{log}}}^*\paren*{\sH_{\mathrm{all}}}}^{\frac12}.
\end{align*}
Using the formulation~\eqref{eq:excess-split}, it can be rewritten as
follows
\begin{align}
\label{eq:gen-bound-old}
\forall h\in \sH_{\mathrm{lin}},\quad  \sR_{\ell_{0-1}}(h)
- \sR_{\ell_{0-1}}^*\paren*{\sH_{\mathrm{all}}}\leq  \sqrt{2}\,\bracket*{\sR_{\Phi_{\mathrm{log}}}(h)
  - \sR_{\Phi_{\mathrm{log}}}^*\paren*{\sH_{\mathrm{lin}}}+ \sR_{\Phi_{\mathrm{log}}}^*\paren*{\sH_{\mathrm{lin}}}
  - \sR_{\Phi_{\mathrm{log}}}^*\paren*{\sH_{\mathrm{all}}}}^{\frac12}.
\end{align}
As shown below, when $B\neq \infty$ and $\sR_{\Phi_{\mathrm{log}}}(h)
- \sR_{\Phi_{\mathrm{log}}}^*\paren*{\sH_{\mathrm{lin}}}$ is small,
$\sH_{\mathrm{lin}}$-consistency bound gives
\begin{align}
\label{eq:gen-bound-new}
\forall h\in \sH_{\mathrm{lin}},\quad
\sR_{\ell_{0-1}}(h)-\sR_{\ell_{0-1}}^*\paren*{\sH_{\mathrm{all}}}\leq
\sqrt{2}\,\paren*{\sR_{\Phi_{\mathrm{log}}}(h)-
  \sR_{\Phi_{\mathrm{log}}}^*\paren*{\sH_{\mathrm{lin}}}+\sM_{\Phi_{\mathrm{log}}}\paren*{\sH_{\mathrm{lin}}}}^{\frac12}
\end{align}
where
$\sM_{\Phi_{\mathrm{log}}}\paren*{\sH_{\mathrm{lin}}}\leq\sR_{\Phi_{\mathrm{log}}}^*\paren*{\sH_{\mathrm{lin}}}
- \sR_{\Phi_{\mathrm{log}}}^*\paren*{\sH_{\mathrm{all}}}$. Therefore, after
the term $\sR_{\Phi_{\mathrm{log}}}(h) -
\sR_{\Phi_{\mathrm{log}}}^*\paren*{\sH_{\mathrm{lin}}}$ is further bounded by
the empirical error and the complexity term, \eqref{eq:gen-bound-new}
gives the tighter generalization bound for the $0/1$ loss and the
linear hypothesis set $\sH_{\mathrm{lin}}$ than
\eqref{eq:gen-bound-old}.  }

\paragraph{Minimizability gap.} We will adopt the standard notation
for the conditional distribution of $Y$ given $X = x$: $\eta(x) =
\sD(Y = 1 \!\mid\! X = x)$ and will also use the shorthand $\Delta
\eta(x) = \eta(x) - \frac{1}{2}$. It is useful to write the
generalization error as
$\sR_{\ell}(h)=\mathbb{E}_{X}\bracket*{\sC_{\ell}(h,x)}$, where
$\sC_{\ell}(h,x)$ is the \emph{conditional $\ell$-risk} defined by
$\sC_{\ell}(h,x) = \eta(x)\ell(h, x, +1) + (1 - \eta(x))\ell(h, x,-1)$.  The \emph{minimal conditional $\ell$-risk} is denoted by
$\sC_{\ell}^*(\sH)(x) = \inf_{h\in \sH}\sC_{\ell}(h,x)$. We also use
the following shorthand for the gap $\Delta\sC_{\ell,\sH}(h,x) =
\sC_{\ell}(h,x)-\sC_{\ell}^*(\sH)(x)$. We call
$\tri*{\Delta\sC_{\ell,\sH}(h,x)}_{\e}=\Delta\sC_{\ell,\sH}(h,x)\mathds{1}_{\Delta\sC_{\ell,\sH}(h,x)>\e}$ the \emph{conditional
$\e$-regret} for $\ell$. To simplify the notation, we also define for
any $t\in [0,1]$, $\sC_{\ell}(h,x,t) = t\ell(h, x, +1) + (1 -
t)\ell(h, x, -1)$ and $\Delta\sC_{\ell,\sH}(h, x, t) = \sC_{\ell}(h, x, t) - \inf_{h \in \sH}\sC_{\ell}(h, x, t)$. Thus, we have
$\Delta\sC_{\ell,\sH}(h, x, \eta(x)) = \Delta\sC_{\ell, \sH}(h, x)$.

A key quantity that appears in our bounds is the \emph{$\paren*{\ell,
  \sH}$-minimizability gap} $\sM_{\ell}(\sH)$, which is the difference
of the best-in class error and the expectation of the minimal
conditional $\ell$-risk: 
\[ \sM_{\ell}(\sH)
 = \sR^*_{\ell}(\sH) - \mathbb{E}_{X} \bracket* {\sC^*_{\ell}(\sH, x)}.
 \]  
 This is an inherent property of the hypothesis set $\sH$ and
 distribution $\sD$ that we cannot hope to estimate or minimize. As an
 example, the minimizability gap for the $0/1$ loss and adversarial
 $0/1$ loss with $\sH_{\mathrm{all}}$ can be expressed as follows:
\begin{align*}
\sM_{\ell_{0-1}}\paren*{\sH_{\mathrm{all}}} &= \sR_{\ell_{0-1}}^*\paren*{\sH_{\mathrm{all}}}-\mathbb{E}_X\bracket*{\min\curl*{\eta(x),1-\eta(x)}}=0,\\ \sM_{\ell_{\gamma}}\paren*{\sH_{\mathrm{all}}} &= \sR_{\ell_{\gamma}}^*\paren*{\sH_{\mathrm{all}}}-\mathbb{E}_X\bracket*{\min\curl*{\eta(x),1-\eta(x)}}.
\end{align*}
\citet[Lemma~2.5]{steinwart2007compare} shows that the minimizability
gap vanishes when the loss $\ell$ is
\emph{minimizable}. \citet{awasthi2021calibration} point out that the
minimizability condition does not hold for adversarial loss functions,
and therefore that, in general,
$\sM_{\ell_{\gamma}}\paren*{\sH_{\mathrm{all}}}$ is strictly positive, thereby
presenting additional challenges for adversarial robust
classification. Thus, the minimizability gap is critical in the
study of adversarial surrogate loss functions. The minimizability gaps
for some common loss functions and hypothesis sets are given in
Table~\ref{tab:loss} in Section~\ref{sec:non-adv-general} for completeness.

\section{General theorems}
\label{sec:general} 

We first introduce two main theorems that provide a general
$\sH$-consistency bound between any target loss and
surrogate loss. These bounds are $\sH$-dependent, taking into
consideration the specific hypothesis set used by a learning
algorithm. To the best of our knowledge, no such guarantee has
appeared in the past. For both theoretical and practical computational
reasons, learning algorithms typically seek a good hypothesis within a
restricted subset of $\sH_{\mathrm{all}}$. Thus, in general,
$\sH$-dependent bounds can provide more relevant guarantees than
excess error bounds. Our proposed bounds are also more general in the
sense that $\sH_{\mathrm{all}}$ can be used as a special case.
\ignore { As shown in Section~\ref{sec:all}, when
  $\sH=\sH_{\mathrm{all}}$, $\ell_2=\ell_{0-1}$ and $\epsilon=0$,
  Theorem~\ref{Thm:excess_bounds_Psi} covers the excess error bounds
  in \citep{bartlett2006convexity,MohriRostamizadehTalwalkar2018}.  }
Theorems~\ref{Thm:excess_bounds_Psi} and \ref{Thm:excess_bounds_Gamma}
are counterparts of each other, while the latter may provide a more
explicit form of bounds as in \eqref{eq:est-bound}.

\begin{restatable}[\textbf{Distribution-dependent $\Psi$-bound}]
  {theorem}{ExcessBoundsPsi}
\label{Thm:excess_bounds_Psi}
 Assume that there exists a convex function $\Psi\colon
 \mathbb{R_{+}}\to \Rset$ with $\Psi(0)\geq0$ and $\epsilon\geq0$ such
 that the following holds for all $h\in \sH$ and $x\in \sX$:
\begin{equation}
\label{eq:cond_psi}
\Psi\paren*{\tri*{\Delta\sC_{\ell_2,\sH}(h,x)}_{\e}}\leq \Delta\sC_{\ell_1,\sH}(h,x).
\end{equation}
Then, the following inequality holds for any $h \in \sH$:
\begin{equation}
\label{eq:bound_Psi_general}
     \Psi\paren*{\sR_{\ell_2}(h)- \sR_{\ell_2}^*(\sH)+\sM_{\ell_2}(\sH)}
     \leq  \sR_{\ell_1}(h)-\sR_{\ell_1}^*(\sH) +\sM_{\ell_1}(\sH) +\max\curl*{\Psi(0),\Psi(\e)}.
\end{equation}
\end{restatable}

\begin{restatable}[\textbf{Distribution-dependent $\Gamma$-bound}]
  {theorem}{ExcessBoundsGamma}
\label{Thm:excess_bounds_Gamma}
Assume that there exists a concave function $\Gamma\colon
\mathbb{R_{+}}\to \Rset$ and $\epsilon\geq0$ such that the following
holds for all $h\in \sH$ and $x\in \sX$:
\begin{equation}
\label{eq:cond_gamma}
\tri*{\Delta\sC_{\ell_2,\sH}(h,x)}_{\e}\leq \Gamma \paren*{\Delta\sC_{\ell_1,\sH}(h,x)}.
\end{equation}
Then, the following inequality holds for any $h \in \sH$:
\begin{equation}
\label{eq:bound_Gamma_general}
     \sR_{\ell_2}
     (h)
     -
     \sR_{\ell_2}^*(\sH)
     \leq
     \Gamma\big(\sR_{\ell_1}
     (h)
     -
     \sR_{\ell_1}^*(\sH)
     +
     \sM_{\ell_1}(\sH)\big)
     -
     \sM_{\ell_2}(\sH)
     +
     \epsilon.
\end{equation}
\end{restatable}
The proofs of Theorems~\ref{Thm:excess_bounds_Psi} and
\ref{Thm:excess_bounds_Gamma} are included in
Appendix~\ref{app:excess_bounds}, where we make use of the convexity of $\Psi$ and concavity of $\Gamma$. Below, we will mainly focus on the
case where $\Psi(0) = 0$ and $\e = 0$. Note that if $\ell_2$ is upper
bounded by $\ell_1$ and
$\sR_{\ell_1}^*(\sH)-\sM_{\ell_1}(\sH) = \sR_{\ell_2}^*(\sH)-\sM_{\ell_2}(\sH)$,
then, the following inequality automatically holds for any $h\in \sH$:
\begin{align*}
  \sR_{\ell_2}(h)- \sR_{\ell_2}^*(\sH) + \sM_{\ell_2}(\sH)
  \leq  \sR_{\ell_1}(h)-\sR_{\ell_1}^*(\sH)+\sM_{\ell_1}(\sH).
\end{align*}
This is a special case of Theorems~\ref{Thm:excess_bounds_Psi} and
\ref{Thm:excess_bounds_Gamma}. Indeed, since
$\sR_{\ell_1}^*(\sH)-\sM_{\ell_1}(\sH) =
\sR_{\ell_2}^*(\sH)-\sM_{\ell_2}(\sH)$, we have $\sC_{\ell_2}^*(\sH)(x)
\equiv \sC_{\ell_1}^*(\sH)(x)$ and thus
$\Delta\sC_{\ell_2,\sH}(h,x)\leq
\Delta\sC_{\ell_1,\sH}(h,x)$. Therefore, $\Phi$ and $\Gamma$ can be
the identity function. We refer to such cases as ``trivial
cases''. They occur when $\sM_{\ell_1}(\sH)$ and $\sM_{\ell_2}(\sH)$
respectively coincide with the corresponding approximation errors and
$\sR_{\ell_1}^*\paren*{\sH_{\mathrm{all}}} =
\sR_{\ell_2}^*\paren*{\sH_{\mathrm{all}}}$. We will later see such cases for
specific loss functions and hypothesis sets (See
\eqref{eq:rho-lin-est-2} in Appendix~\ref{app:rho-lin} and
\eqref{eq:rho-lin-est-adv-3} in Appendix~\ref{app:rho-lin-adv}). Let
us point out, however, that the corresponding $\sH$-consistency
bounds are still valid and worth studying
since they can be shown to be the tightest 
(Theorems~\ref{Thm:tightness} and \ref{Thm:tightness-adv}).

Theorem~\ref{Thm:excess_bounds_Psi} is distribution-dependent, in the
sense that, for a fixed distribution, if we find a $\Psi$ that
satisfies condition~\eqref{eq:cond_psi}, then the bound
\eqref{eq:bound_Psi_general} only gives guarantee for that same
distribution. Since the distribution $\sD$ of interest is typically
unknown, to obtain guarantees for $\sD$, if the only information given
is that $\sD$ belongs to a set of distributions $\sP$, we need to find
a $\Psi$ that satisfies condition~\eqref{eq:cond_psi} for all the
distributions in $\sP$. The choice of $\Psi$ is critical, since it
determines the form of the bound obtained.

We say that $\Psi$ is
\emph{optimal} if any function that makes the bound
\eqref{eq:bound_Psi_general} hold for all distributions in $\sP$ is
everywhere no larger than $\Psi$. The optimal $\Psi$ leads to the
tightest $\sH$-consistency bound
\eqref{eq:bound_Psi_general} uniform over $\sP$. Specifically, when
$\sP$ consists of all distributions, we say that the bound is
distribution-independent. The above also applies to
Theorem~\ref{Thm:excess_bounds_Gamma}, except that $\Gamma$ is
\emph{optimal} if any function that makes the bound
\eqref{eq:bound_Gamma_general} hold for all distributions in $\sP$ is
everywhere no less than $\Gamma$.

When $\ell_2$ is the $0/1$ loss or the adversarial $0/1$ loss, the
conditional $\epsilon$-regret that appears in
condition~\eqref{eq:cond_psi} has explicit forms for common hypothesis
sets as characterized later in
Lemma~\ref{lemma:explicit_assumption_01} and
\ref{lemma:explicit_assumption_01_adv}, establishing the basis for
introducing non-adversarial and adversarial $\sH$-estimation error transformation in Section~\ref{sec:non-adv-general}
and \ref{sec:adv-general}. We will see later in these sections that
the transformations introduced are often the optimal $\Psi$ we are
seeking for, which respectively leads to tight non-adversarial and
adversarial distribution-independent guarantees.
In Section~\ref{sec:non-adv} and \ref{sec:adv}, we also apply our
general theorems and tools to loss functions and hypothesis sets
widely used in practice. Each case requires a careful analysis
that we present in detail.

\section{Guarantees for the zero-one loss \texorpdfstring{$\ell_2 = \ell_{0-1}$}{l2}}
\label{sec:non-adv}

In this section, we discuss guarantees in the non-adversarial scenario
where $\ell_2$ is the zero-one loss, $\ell_{0-1}$.  The lemma stated
next characterizes the minimal conditional $\ell_{0-1}$-risk and the
conditional $\epsilon$-regret, which will be helpful for introducing
the general tools in Section~\ref{sec:non-adv-general}. The proof is
given in Appendix~\ref{app:explicit_assumption}.  For convenience, we
will adopt the following notation: $\ov \sH(x) =
\curl*{h \in \sH \colon \sign(h(x)) \Delta \eta(x) \leq 0}$.

\begin{restatable}{lemma}{ExplicitAssumption}
\label{lemma:explicit_assumption_01}
Assume that $\sH$ satisfies the following condition for any $x\in \sX$:
\[\curl*{\sign(h(x)) \colon h\in \sH} = \curl*{-1, +1}.\]
Then, the minimal conditional $\ell_{0-1}$-risk is
\begin{align*}
\sC^*_{\ell_{0-1}}(\sH, x)=\sC^*_{\ell_{0-1}}(\sH_{\mathrm{all}}, x)=\min\curl*{\eta(x),1-\eta(x)}.
\end{align*}
The conditional $\epsilon$-regret for $\ell_{0-1}$ can be characterized as
\begin{align*}
  \tri*{\Delta\sC_{\ell_{0-1},\sH}(h,x)}_{\e}=\tri*{2 \abs*{\Delta \eta(x)}}_{\e}
  \mathds{1}_{h \in \ov \sH(x)}\,.
\end{align*}
\end{restatable}

\subsection{Hypothesis set of all measurable functions}
\label{sec:all}

Before introducing our general tools, we will consider the case where
$\sH = \sH_{\mathrm{all}}$ and will show that previous excess error
bounds can be recovered as special cases of our results.  As shown in
\citep{steinwart2007compare}, both
$\sM_{\ell_{0-1}}\paren*{\sH_{\mathrm{all}}}$ and
$\sM_{\Phi}\paren*{\sH_{\mathrm{all}}}$ vanish. Thus by
Lemma~\ref{lemma:explicit_assumption_01}, we obtain the following
corollary of Theorem~\ref{Thm:excess_bounds_Psi} by taking
$\epsilon=0$.

\begin{corollary}
\label{cor:excess_bounds_Psi_01_B}
Assume that there exists a convex function $\Psi\colon
\mathbb{R_{+}}\to \Rset$ with $\Psi(0)=0$ such that for any $x\in
\sX$, $\Psi\paren*{2 \abs*{\Delta \eta(x)}}\leq
\inf_{h\in\ov{\sH_{\mathrm{all}}}(x)}\Delta\sC_{\Phi,\sH_{\mathrm{all}}}(h,x).$
Then, for any hypothesis $h\in\sH_{\mathrm{all}}$,
the following inequality holds:
    \begin{align*}
      \Psi\paren*{\sR_{\ell_{0-1}}(h)
        - \sR_{\ell_{0-1}}^*\paren*{\sH_{\mathrm{all}}}}
      \leq  \sR_{\Phi}(h) - \sR_{\Phi}^*\paren*{\sH_{\mathrm{all}}}.
    \end{align*}
\end{corollary}
Furthermore, Corollary~\ref{cor:excess_bounds_Psi_01_M} follows from
Corollary~\ref{cor:excess_bounds_Psi_01_B} by taking the convex
function $\Psi(t)=\paren*{t/(2c)}^s$.
\begin{corollary}
\label{cor:excess_bounds_Psi_01_M}
Assume there exist $s\geq 1$ and $c>0$ such that for any $x\in \sX$,
\[\abs*{\Delta \eta(x)}\leq c~\inf_{h\in\ov{\sH_{\mathrm{all}}}(x)}\paren*{\Delta\sC_{\Phi,\sH_{\mathrm{all}}}(h,x)}^{\frac1s}.\]
Then, for any hypothesis $h\in\sH_{\mathrm{all}}$,
    \begin{align*}
     \sR_{\ell_{0-1}}(h)- \sR_{\ell_{0-1}}^*\paren*{\sH_{\mathrm{all}}}\leq 2c~\paren*{ \sR_{\Phi}(h)-\sR_{\Phi}^*\paren*{\sH_{\mathrm{all}}}}^{\frac1s}.
    \end{align*}
\end{corollary}
The excess error bound results in the literature are all covered by the above corollaries.  
As shown in Appendix~\ref{app:compare-all-measurable}, Theorem~4.7 in \citep{MohriRostamizadehTalwalkar2018} is a special case of Corollary~\ref{cor:excess_bounds_Psi_01_M} and Theorem 1.1 in \citep{bartlett2006convexity} is a special case of Corollary~\ref{cor:excess_bounds_Psi_01_B}.

\subsection{General hypothesis sets \texorpdfstring{$\sH$}{H}}
\label{sec:non-adv-general}

In this section, we provide general tools to study
$\sH$-consistency bounds when the target loss is the
$0/1$ loss. We will then apply them to study specific hypothesis sets
and surrogates in Section~\ref{sec:non-adv-lin} and
\ref{sec:non-adv-NN}. Lemma~\ref{lemma:explicit_assumption_01}
characterizes the conditional $\epsilon$-regret for $\ell_{0-1}$ with
common hypothesis sets. Thus, Theorems~\ref{Thm:excess_bounds_Psi} and \ref{Thm:excess_bounds_Gamma} can be instantiated as
Theorems~\ref{Thm:excess_bounds_Psi_01_general} and \ref{Thm:excess_bounds_Gamma_01_general} in these cases (see
Appendix~\ref{app:theorems}). They are powerful distribution-dependent
bounds and, as discussed in Section~\ref{sec:general}, the bounds
become distribution-independent if the corresponding conditions can be
verified for all the distributions with some $\Psi$, which is
equivalent to verifying the condition in the following theorem.

\begin{restatable}[\textbf{Distribution-independent $\Psi$-bound}]
  {theorem}{ExcessBoundsPsiUniform}
\label{Thm:excess_bounds_Psi_uniform}
Assume that $\sH$ satisfies the condition of
Lemma~\ref{lemma:explicit_assumption_01}. Assume that there exists a
convex function $\Psi\colon \mathbb{R_{+}} \to \Rset$ with $\Psi(0) =
0$ and $\epsilon\geq0$ such that for any $t\in\left[1/2,1\right]$,
\begin{align*}
  \Psi \paren*{\tri*{2t-1}_{\e}}
  \leq \inf_{x\in \sX,h\in\sH:h(x)<0}\Delta\sC_{\Phi,\sH}(h,x,t).
\end{align*}
Then, for any hypothesis $h\in\sH$ and any distribution,
\begin{equation}
\label{eq:bound_Psi_01}
     \Psi\paren*{\sR_{\ell_{0-1}}(h)- \sR_{\ell_{0-1}}^*(\sH)+\sM_{\ell_{0-1}}(\sH)}
     \leq  \sR_{\Phi}(h)-\sR_{\Phi}^*(\sH)+\sM_{\Phi}(\sH)+\max\curl*{0,\Psi(\e)}.
\end{equation}
\end{restatable}
The counterpart of Theorem~\ref{Thm:excess_bounds_Psi_uniform} is
Theorem~\ref{Thm:excess_bounds_Gamma_uniform}
(distribution-independent $\Gamma$-bound), deferred to
Appendix~\ref{app:theorems} due to space limitations. The proofs for
both theorems are included in
Appendix~\ref{app:uniform}. Theorem~\ref{Thm:excess_bounds_Psi_uniform}
provides the general tool to derive distribution-independent
$\sH$-consistency bounds. They are in fact
tight if we choose $\Psi$ to be the \emph{$\sH$-estimation error transformation} defined as follows.

\begin{figure}[t]
\begin{center}
\includegraphics[scale=0.5]{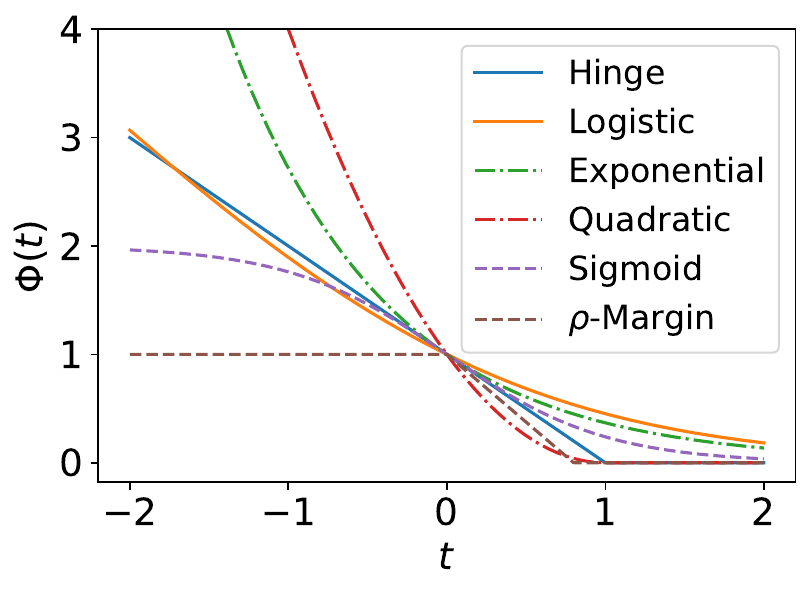}
\includegraphics[scale=0.5]{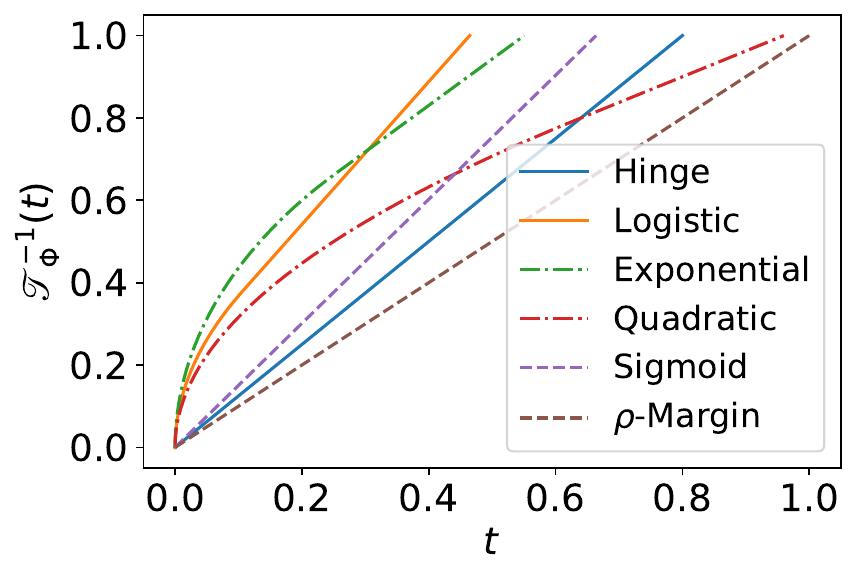}
\caption{Left: surrogates. Right: $\sH_{\mathrm{lin}}$-est. error trans. inv.}
\label{fig:surrogate}
\end{center}
\end{figure}
\begin{definition}[\textbf{$\sH$-estimation error transformation}]
\label{def:trans}
The \emph{$\sH$-estimation error transformation} of
$\Phi$ is defined on $t\in \left[0,1\right]$ by
$
\sT_{\Phi}\paren*{t}=\sT(t)\mathds{1}_{t\in
  \left[\epsilon,1\right]}+(\sT(\e)/\e)\,t\mathds{1}_{t\in
  \left[0,\epsilon\right)}
$,
where \[\sT(t):=\inf_{x\in \sX,h\in\sH:h(x)<
  0}\Delta\sC_{\Phi,\sH}\paren*{h,x,\frac{t+1}{2}}.\]
\end{definition}
When $\e=0$, $\sT_{\Phi}\paren*{t}$ coincides with $\sT\paren*{t}$. Observe that for any $t\in\bracket[big]{(1+\epsilon)/2,1}$, the following equality holds:
\begin{align*}
\sT_{\Phi}\paren*{2t-1}
  = \inf_{x\in \sX,h\in\sH:h(x)<0}\Delta\sC_{\Phi,\sH}(h,x,t).    
\end{align*}
Taking $\Psi=\sT_{\Phi}$ satisfies the condition in
Theorem~\ref{Thm:excess_bounds_Psi_uniform} if $\sT_{\Phi}$ is convex
with $\sT_{\Phi}(0)=0$. Moreover, as mentioned earlier, it actually
leads to the tightest $\sH$-consistency bound
\eqref{eq:bound_Psi_01} when $\e=0$.

\begin{restatable}[\textbf{Tightness}]{theorem}{Tightness}
\label{Thm:tightness}
Suppose that $\sH$ satisfies the condition of
Lemma~\ref{lemma:explicit_assumption_01} and that $\e=0$. If
$\sT_{\Phi}$ is convex with $\sT_{\Phi}(0)=0$, then, for any
$t\in[0,1]$ and $\delta>0$, there exist a distribution $\sD$ and a
hypothesis $h\in\sH$ such that $\sR_{\ell_{0-1}}(h)-
\sR_{\ell_{0-1}}^*(\sH)+\sM_{\ell_{0-1}}(\sH)=t$ and
$\sT_{\Phi}(t)\leq\sR_{\Phi}(h)-\sR_{\Phi}^*(\sH)+\sM_{\Phi}(\sH)\leq
\sT_{\Phi}(t) + \delta$.
\end{restatable}
The proof is included in Appendix~\ref{app:tightness}. In other words,
when $\e=0$, if $\sT_{\Phi}$ is convex with $\sT_{\Phi}(0)=0$, it is
optimal for the distribution-independent
bound~\eqref{eq:bound_Psi_01}. Moreover, if $\sT_{\Phi}$ is
additionally invertible and non-increasing, $\sT_{\Phi}^{-1}$ is the
optimal function for the distribution-independent bound in
Theorem~\ref{Thm:excess_bounds_Gamma_uniform}
(Appendix~\ref{app:theorems}) and the two bounds are equivalent.

In the following sections, we will see that all these assumptions hold
for common loss functions with linear and neural network hypothesis
sets. Next, we will apply Theorems~\ref{Thm:excess_bounds_Psi_uniform}
and \ref{Thm:tightness} to the linear models
(Section~\ref{sec:non-adv-lin}) and neural networks
(Section~\ref{sec:non-adv-NN}). Each case requires a detailed analysis
(See Appendix~\ref{app:derivation-lin} and \ref{app:derivation-NN}).

The loss functions considered below and their minimizability gaps are defined in
Table~\ref{tab:loss}. In some cases,
  the minimizability gap coincides with the approximation error. For
  example,
  \[\sM_{\Phi_{\mathrm{sig}}}\paren*{\sH_{\mathrm{lin}}}=
  \sR_{\Phi_{\mathrm{sig}}}^*\paren*{\sH_{\mathrm{lin}}}-\mathbb{E}_{X}\bracket[big]{1-\abs*{1-2\eta(x)}\tanh\paren*{k\paren[big]{W\norm*{x}_p+B}}}\]
  coincides with the
  $\paren*{\Phi_{\mathrm{sig}},\sH_{\mathrm{lin}}}$-approximation
  error
  $\sR_{\Phi_{\mathrm{sig}}}^*\paren*{\sH_{\mathrm{lin}}}-\mathbb{E}_{X}\bracket[big]{1-\abs*{1-2\eta(x)}}$
  for $B = + \infty$;
  \[\sM_{\Phi_{\mathrm{hinge}}}\paren*{\sH_{\mathrm{NN}}} =
  \sR_{\Phi_{\mathrm{hinge}}}^*\paren*{\sH_{\mathrm{NN}}}-\mathbb{E}_{X}\bracket*{1-\abs*{2\eta(x)-1}\min\curl*{\Lambda
      W\norm*{x}_p+\Lambda B,1}}\] coincides with the
  $\paren*{\Phi_{\mathrm{hinge}},\sH_{\mathrm{NN}}}$-approximation
  error
  $\sR_{\Phi_{\mathrm{hinge}}}^*\paren*{\sH_{\mathrm{NN}}}-\mathbb{E}_{X}\bracket[big]{1-\abs[big]{1-2\eta(x)}}$
  for $\Lambda B \geq 1$. The detailed derivation is included in
  Appendix~\ref{app:derivation-non-adv}, \ref{app:derivation-adv}.
  \begin{table}[t]
\caption{Loss functions and their minimizability gaps.}
    \label{tab:loss}
    \begin{center}
    \begin{tabular}{l|lll}
    \toprule
      Loss Functions & Definitions & $\sM_{\ell}\paren*{\sH_{\mathrm{lin}}}$ & $\sM_{\ell}\paren*{\sH_{\mathrm{NN}}}$\\
    \midrule
     Hinge & $\Phi_{\mathrm{hinge}}(t)=\max\curl*{0,1-t}$ & \eqref{eq:M-hinge-lin} & \eqref{eq:M-hinge-NN}\\
     Logistic & $\Phi_{\mathrm{log}}(t)=\log_2(1+e^{-t})$ & \eqref{eq:M-log-lin} & \eqref{eq:M-log-NN}\\
     Exponential & $\Phi_{\mathrm{exp}}(t)=e^{-t}$ & \eqref{eq:M-exp-lin} & \eqref{eq:M-exp-NN} \\ Quadratic  & $\Phi_{\mathrm{quad}}(t)=(1-t)^2\mathds{1}_{t\leq 1}$ & \eqref{eq:M-quad-lin} & \eqref{eq:M-quad-lin} \\
     Sigmoid & $\Phi_{\mathrm{sig}}(t)=1-\tanh(kt),~k>0$ & \eqref{eq:M-sig-lin} & \eqref{eq:M-sig-NN}\\
     $\rho$-Margin & $\Phi_{\rho}(t)=\min\curl*{1,\max\curl*{0,1-\frac{t}{\rho}}},~\rho>0$ & \eqref{eq:M-rho-lin} & \eqref{eq:M-hinge-NN}\\
     Sup-$\rho$-Margin & $\wt{\Phi}_{\rho}=\sup_{x'\colon \|x-x'\|_p\leq \gamma}\Phi_{\rho}(y h(x'))$ & \eqref{eq:M-rho-lin-adv} & \eqref{eq:M-rho-NN-adv}\\
     Zero-One & $\ell_{0-1}=\mathds{1}_{\sign(h(x))\neq y}$ & \eqref{eq:M-01-lin} & \eqref{eq:M-01-NN}\\
     Adversarial Zero-One & $\ell_{\gamma}=\sup_{x'\colon \|x-x'\|_p\leq
  \gamma}\mathds{1}_{y h(x') \leq 0}$ & \eqref{eq:M-01-lin-adv} & \eqref{eq:M-01-NN-adv}\\
    \bottomrule
    \end{tabular}
    \end{center}
\end{table}

\subsection{Linear hypotheses}
\label{sec:non-adv-lin}
\begin{table}[t]
\caption{$\sH_{\mathrm{lin}}$-estimation error
  transformation and $\sH_{\mathrm{lin}}$-consistency
  bounds with $\epsilon=0$.}
    \label{tab:compare}
\begin{center}
    \begin{tabular}{l|ll}
    \toprule
      Surrogates & $\sT_{\Phi}(t),\, t\in [0,1]$   &  Bound \\
    \midrule
      Hinge & $\min \curl*{B, 1} \, t $  & \eqref{eq:hinge-lin-est} \\
      Logistic & $\begin{cases}
\frac{t+1}{2}\log_2(t+1)+\frac{1-t}{2}\log_2(1-t),\quad &  t\leq \frac{e^B-1}{e^B+1},\\
1-\frac{t+1}{2}\log_2(1+e^{-B})-\frac{1-t}{2}\log_2(1+e^B),\quad & t> \frac{e^B-1}{e^B+1}.
\end{cases}$ & \eqref{eq:log-lin-est}\\
      Exponential & $\begin{cases}
1-\sqrt{1-t^2}, & t\leq \frac{e^{2B}-1}{e^{2B}+1},\\
1-\frac{t+1}{2}e^{-B}-\frac{1-t}{2}e^B, & t> \frac{e^{2B}-1}{e^{2B}+1}.
\end{cases}$ & \eqref{eq:exp-lin-est}\\
      Quadratic & $\begin{cases}
t^2, & t\leq B,\\
2B \,t-B^2, & t> B.
\end{cases}$ & \eqref{eq:quad-lin-est}\\
      Sigmoid & $\tanh(kB) \, t$ & \eqref{eq:sig-lin-est}\\
      $\rho$-Margin & $\frac{\min\curl*{B,\rho}}{\rho} \, t$ & \eqref{eq:rho-lin-est}\\
    \bottomrule
    \end{tabular}
    \end{center}
\end{table}
By applying Theorems~\ref{Thm:excess_bounds_Psi_uniform} and \ref{Thm:tightness}, we can derive $\sH_{\mathrm{lin}}$-consistency
bounds for common loss functions defined in
Table~\ref{tab:loss}. Table~\ref{tab:compare} supplies the $\sH_{\mathrm{lin}}$-estimation error transformation $\sT_{\Phi}$ and the corresponding bounds for those loss functions. The inverse $\sT_{\Phi}^{-1}$ is given in Table~\ref{tab:compare_inverse} of Appendix~\ref{app:table}.
Surrogates $\Phi$ and their corresponding $\sT_{\Phi}^{-1}$ ($B=0.8$) are visualized in Figure~\ref{fig:surrogate}. Theorems~\ref{Thm:excess_bounds_Psi_uniform} and \ref{Thm:tightness} apply to all these cases since
$\sT_{\Phi}$ is convex, increasing, invertible and satisfies that
$\sT_{\Phi}(0)=0$. More precisely, taking $\Psi=\sT_{\Phi}$ and $\e=0$
in \eqref{eq:bound_Psi_01} and using the inverse function
$\sT_{\Phi}^{-1}$ directly give the tightest bound. As an example, for
the sigmoid loss,
$\sT_{\Phi_{\mathrm{sig}}}^{-1}(t)=\frac{t}{\tanh(kB)}$. Then the
bound~\eqref{eq:bound_Psi_01} becomes $\sR_{\ell_{0-1}}(h)-
\sR_{\ell_{0-1}}^*\paren*{\sH_{\mathrm{lin}}}\leq
\paren{\sR_{\Phi_{\mathrm{sig}}}(h)-
  \sR_{\Phi_{\mathrm{sig}}}^*\paren*{\sH_{\mathrm{lin}}}
+\sM_{\Phi_{\mathrm{sig}}}\paren*{\sH_{\mathrm{lin}}}}/\tanh(kB)
- \sM_{\ell_{0-1}}\paren*{\sH_{\mathrm{lin}}}$, which is
\eqref{eq:sig-lin-est} in Table~\ref{tab:compare}. Furthermore, after
plugging in the minimizability gaps concluded in Table~\ref{tab:loss},
we will obtain the novel bound $\sR_{\ell_{0-1}}(h)-
\sR_{\ell_{0-1}}^*\paren*{\sH_{\mathrm{all}}}\leq
\paren{\sR_{\Phi_{\mathrm{sig}}}(h)-
  \mathbb{E}_{X}\bracket*{1-\abs*{1-2\eta(x)}
    \tanh\paren*{k\paren*{W\norm*{x}_p+B}}}
}/\tanh(kB)$ (\eqref{eq:sig-lin-est-2} in
Appendix~\ref{app:sig-lin}). The bounds for other surrogates are
similarly derived in Appendix~\ref{app:derivation-lin}. For the
logistic loss and exponential loss, to simplify the expression, the
bounds are obtained by plugging in an upper bound of
$\sT_{\Phi}^{-1}$.

Let us emphasize that these $\sH$-consistency bounds
are novel in the sense that they are all hypothesis set-dependent and, to our knowledge,
no such guarantee has been presented before. More precisely, the
bounds of Table~\ref{tab:compare} depend directly on the parameter $B$
in the linear models and parameters of the loss function (e.g., $k$ in
sigmoid loss). Thus, for a fixed hypothesis $h\in \sH_{\mathrm{lin}}$,
we may give the tightest bound by choosing the best parameter $B$. As
an example, Appendix~\ref{app:sig-lin} shows that the bound
\eqref{eq:sig-lin-est-2} with $B = + \infty$ coincides with the
excess error bound known for the sigmoid loss
\citep{bartlett2006convexity}. However, for a fixed hypothesis $h$, by
varying $B$ (hypothesis set) and $k$ (loss function), we may obtain a
finer bound! Thus studying hypothesis set-dependent bounds can guide
us to select the most suitable hypothesis set and loss
function. Moreover, as shown by Theorem~\ref{Thm:tightness}, all the
bounds obtained by directly using $\sT_{\Phi}^{-1}$ are tight and
cannot be further improved.

\subsection{One-hidden-layer ReLU neural networks}
\label{sec:non-adv-NN}
\begin{table}[t]
    \caption{$\sH_{\mathrm{NN}}$-estimation error transformation and $\sH_{\mathrm{NN}}$-consistency bounds with $\epsilon=0$.}
    \label{tab:compare-NN}
    \begin{center}
    \begin{tabular}{l|ll}
    \toprule
      Surrogates & $\sT_{\Phi}(t),\, t\in [0,1]$  & Bound \\
    \midrule
      Hinge & $\min \curl*{\Lambda B, 1} \, t $  & \eqref{eq:hinge-NN-est} \\
      Logistic & $\begin{cases}
\frac{t+1}{2}\log_2(t+1)+\frac{1-t}{2}\log_2(1-t),\quad & t\leq \frac{e^{\Lambda B}-1}{e^{\Lambda B}+1},\\
1-\frac{t+1}{2}\log_2(1+e^{-\Lambda B})-\frac{1-t}{2}\log_2(1+e^{\Lambda B}),\quad & t> \frac{e^{\Lambda B}-1}{e^{\Lambda B}+1}.
\end{cases}$ & \eqref{eq:log-NN-est}\\
      Exponential & $\begin{cases}
1-\sqrt{1-t^2}, & t\leq \frac{e^{2\Lambda B}-1}{e^{2\Lambda B}+1},\\
1-\frac{t+1}{2}e^{-\Lambda B}-\frac{1-t}{2}e^{\Lambda B}, & t> \frac{e^{2\Lambda B}-1}{e^{2\Lambda B}+1}.
\end{cases}$ & \eqref{eq:exp-NN-est}\\
      Quadratic & $\begin{cases}
t^2,~t\leq \Lambda B,\\
2\Lambda B t-(\Lambda B)^2,~t> \Lambda B.
\end{cases}$ & \eqref{eq:quad-NN-est}\\
      Sigmoid & $\tanh(k\Lambda B) \, t$ & \eqref{eq:sig-NN-est}\\
      $\rho$-Margin & $\frac{\min\curl*{\Lambda B,\rho}}{\rho} \, t$ & \eqref{eq:rho-NN-est}\\
    \bottomrule
    \end{tabular}
    \end{center}
\end{table}
In this section, we give $\sH$-consistency bounds for
one-hidden-layer ReLU neural networks
$\sH_{\mathrm{NN}}$. Table~\ref{tab:compare-NN} is the counterpart of Table~\ref{tab:compare}
for $\sH_{\mathrm{NN}}$. Different from the bounds in the linear case,
all the bounds in Table~\ref{tab:compare-NN} not only depend on $B$,
but also depend on $\Lambda$, which is a new parameter in
$\sH_{\mathrm{NN}}$. This further illustrates that our bounds are
hypothesis set-dependent and that, as with the linear case, adequately
choosing the parameters $\Lambda$ and $B$ in $\sH_{\mathrm{NN}}$ would
give us better hypothesis set-dependent guarantees than standard
excess error bounds. The inverse $\sT_{\Phi}^{-1}$ is given in Table~\ref{tab:compare_inverse-NN} of Appendix~\ref{app:table}. Our proofs and techniques could also be adopted
for the analysis of multi-layer neural networks.

\subsection{Guarantees under Massart's noise condition}
\label{sec:noise-non-adv}

The distribution-independent $\sH$-consistency bound
\eqref{eq:bound_Psi_01} cannot be improved, since they are tight as
shown in Theorem~\ref{Thm:tightness}. However, the bounds can be
further improved in the distribution-dependent setting. Indeed, we
will study how $\sH$-consistency bounds can be
improved under low noise conditions, which impose the restrictions on
the conditional distribution $\eta(x)$. We consider Massart's noise
condition \citep{massart2006risk} which is defined as follows.

\begin{definition}[\textbf{Massart's noise}]
\label{def:massarts-noise}
The distribution $\sD$ over $\cX\times \cY$ satisfies Massart's noise condition if
$\abs*{\Delta \eta(x)} \geq \beta \text{ for almost all } x \in \cX$,
for some constant $\beta \in (0, 1/2]$.
\end{definition}
When it is known that the distribution $\sD$ satisfies Massart's noise condition
with $\beta$, in contrast with the distribution-independent bounds, we
can require the bounds \eqref{eq:bound_Psi_general} and
\eqref{eq:bound_Gamma_general} to hold uniformly only for such
distributions. With Massart's noise condition, we
introduce a modified $\sH$-estimation error
transformation in Proposition~\ref{prop:prop-noise}
(Appendix~\ref{app:derivation-all_noise}), which verifies
condition~\eqref{eq:condition_Psi_general} of
Theorem~\ref{Thm:excess_bounds_Psi_01_general} (the finer distribution
dependent guarantee mentioned before, deferred to
Appendix~\ref{app:theorems}) for all distributions under the noise
condition. Then, using this transformation, we can obtain more
favorable distribution-dependent bounds. As an example, we consider
the quadratic loss $\Phi_{\mathrm{quad}}$, the logistic loss
$\Phi_{\mathrm{log}}$ and the exponential loss $\Phi_{\mathrm{exp}}$
with $\sH_{\mathrm{all}}$. For all distributions and $h\in
\sH_{\mathrm{all}}$, as shown in
\citep{Zhang2003,bartlett2006convexity,MohriRostamizadehTalwalkar2018},
the following holds:
\begin{align*}
  \sR_{\ell_{0-1}}(h)- \sR_{\ell_{0-1}}^*\paren*{\sH_{\mathrm{all}}}&
  \leq \sqrt{2}\paren*{\sR_{\Phi}(h) - \sR_{\Phi}^*\paren*{\sH_{\mathrm{all}}}}^{1/2},
\end{align*}
when the surrogate loss $\Phi$ is $\Phi_{\mathrm{log}}$ or
$\Phi_{\mathrm{exp}}$. If $\Phi=\Phi_{\mathrm{quad}}$, then the
constant multiplier $\sqrt{2}$ can be removed. For distributions that
satisfy Massart's noise condition with $\beta$, as proven in
Appendix~\ref{app:derivation-all_noise}, for any $h\in
\sH_{\mathrm{all}}$ such that $\sR_{\Phi}(h) \leq
\sR_{\Phi}^*\paren*{\sH_{\mathrm{all}}}+\sT(2\beta)$, the consistency excess
error bound is improved from the square-root dependency to a linear
dependency:
\begin{align}
\label{eq:non-adv-noise}
    \sR_{\ell_{0-1}}(h)- \sR_{\ell_{0-1}}^*\paren*{\sH_{\mathrm{all}}}
    \leq
    2\beta\paren*{\sR_{\Phi}(h)- \sR_{\Phi}^*\paren*{\sH_{\mathrm{all}}}} / \sT(2\beta),
\end{align}
where $\sT(t)$ equals to $t^2$,
$\frac{t+1}{2}\log_2(t+1)+\frac{1-t}{2}\log_2(1-t)$ and
$1-\sqrt{1-t^2}$ for $\Phi_{\mathrm{quad}}$, $\Phi_{\mathrm{log}}$ and
$\Phi_{\mathrm{exp}}$ respectively. These linear dependent bounds are
tight, as illustrated in Section~\ref{sec:simulations}.

\section{Guarantees for the adversarial loss \texorpdfstring{$\ell_2 = \ell_{\gamma}$}{l2}}
\label{sec:adv}

In this section, we discuss the adversarial scenario where $\ell_2$ is
the adversarial $0/1$ loss $\ell_{\gamma}$. We consider
\emph{symmetric} hypothesis sets, which satisfy: $h\in\sH$ if and only
if $-h\in \sH$.
For convenience, we will adopt
the following definitions:
\begin{align*}
\uv h_\gamma(x)  =\inf_{x'\colon \|x - x'\|_p\leq\gamma} h(x') \qquad
\ov h_\gamma(x)  =\sup_{x'\colon \|x - x'\|_p\leq\gamma} h(x').
\end{align*}
We also define $\ov \sH_\gamma(x) = \curl*{h\in\sH:\uv h_\gamma(x)\leq 0
  \leq \ov h_\gamma(x)}$.
The following characterization of the minimal conditional
$\ell_{\gamma}$-risk and conditional $\epsilon$-regret is based on
\citep[Lemma~27]{awasthi2021calibration} and will be helpful in
introducing the general tools in Section~\ref{sec:adv-general}. The
proof is similar and is included in
Appendix~\ref{app:explicit_assumption} for completeness.

\begin{restatable}{lemma}{ExplicitAssumptionAdv}
\label{lemma:explicit_assumption_01_adv}
Assume that $\sH$ is symmetric. Then, the minimal conditional
$\ell_{\gamma}$-risk is
\begin{align*}
  \sC^*_{\ell_{\gamma}}(\sH, x)
  = \min\curl*{\eta(x), 1 - \eta(x)}\mathds{1}_{\ov \sH_\gamma(x)\neq \sH}
  + \mathds{1}_{\ov \sH_\gamma(x)=\sH}\,.
\end{align*}
The conditional $\e$-regret for $\ell_{\gamma}$ can be characterized
as
\begin{align*}
\tri*{\Delta\sC_{\ell_{\gamma},\sH}(h,x)}_{\e}
=
\begin{cases}
\tri*{\abs*{\Delta \eta(x)}+\frac12}_{\e} 
&h \in \ov \sH_\gamma(x)\subsetneqq \sH\\
\tri*{2\Delta \eta(x)}_{\e}
& \ov h_\gamma(x)<0\\
\tri*{-2\Delta \eta(x)}_{\e}
&\uv h_\gamma(x)>0 \\
0 
&\text{otherwise}
\end{cases}
\end{align*}
\end{restatable}

\subsection{General hypothesis sets \texorpdfstring{$\sH$}{H}}
\label{sec:adv-general}

As with the non-adversarial case, we begin by providing general
theoretical tools to study $\sH$-consistency bounds
when the target loss is the adversarial $0/1$
loss. Lemma~\ref{lemma:explicit_assumption_01_adv} characterizes the
conditional $\epsilon$-regret for $\ell_{\gamma}$ with symmetric
hypothesis sets. Thus, Theorems~\ref{Thm:excess_bounds_Psi} and \ref{Thm:excess_bounds_Gamma} can be instantiated as
Theorems~\ref{Thm:excess_bounds_Psi_01_general_adv} and
\ref{Thm:excess_bounds_Gamma_01_general_adv} (See
Appendix~\ref{app:theorems}) in these cases. These results are
distribution-dependent and can serve as general tools. For example, we
can use these tools to derive more favorable guarantees under noise
conditions (Section~\ref{sec:noise-adv}). As in the previous section,
we present their distribution-independent version in the following
theorem.

\begin{restatable}[\textbf{Adversarial distribution-independent $\Psi$-bound}]
  {theorem}{ExcessBoundsPsiUniformAdv}
\label{Thm:excess_bounds_Psi_uniform-adv}
Suppose that $\sH$ is symmetric. Assume there exist a convex function
$\Psi\colon \mathbb{R_{+}} \to \Rset$ with $\Psi(0)=0$ and
$\epsilon\geq0$ such that the following holds for any
$t\in\left[1/2,1\right]\colon$
\begin{align*}
  &\Psi\paren*{\tri*{t}_{\e}}
  \leq \inf_{x\in\sX,h\in \ov \sH_\gamma(x)\subsetneqq \sH}\Delta\sC_{\wt{\Phi},\sH}(h,x,t),\\
  &\Psi\paren*{\tri*{2t-1}_{\e}}
  \leq \inf_{x\in \sX,h\in\sH\colon  \ov h_\gamma(x)< 0}\Delta\sC_{\wt{\Phi},\sH}(h,x,t).
\end{align*}
Then, for any hypothesis $h\in\sH$ and any distribution,
\begin{equation}
\label{eq:bound_Psi_01_adv}
     \Psi\paren*{\sR_{\ell_{\gamma}}(h) - \sR_{\ell_{\gamma}}^*(\sH) + \sM_{\ell_{\gamma}}(\sH)}
     \leq  \sR_{\wt{\Phi}}(h)
     - \sR_{\wt{\Phi}}^*(\sH) + \sM_{\wt{\Phi}}(\sH)+\max\curl*{0,\Psi(\e)}.
\end{equation}
\end{restatable}
The counterpart of Theorem~\ref{Thm:excess_bounds_Psi_uniform-adv} is
Theorem~\ref{Thm:excess_bounds_Gamma_uniform-adv} (adversarial
distribution-independent $\Gamma$-bound), deferred to
Appendix~\ref{app:theorems} due to space limitations. The proofs for
both theorems are included in Appendix~\ref{app:uniform-adv}.  As with non-adversarial scenario, the tightest
distribution-independent $\sH$-consistency bounds
obtained by Theorem~\ref{Thm:excess_bounds_Psi_uniform-adv} can be
achieved by the optimal $\Psi$, which is the \emph{adversarial
$\sH$-estimation error transformation} defined as follows.
\begin{definition}[\textbf{Adversarial error transformation}]
\label{def:trans-adv}
The \emph{adversarial $\sH$-estimation error transformation} of
$\wt{\Phi}$ is defined on $t\in \left[0,1\right]$ by
$\sT_{\wt{\Phi}}\paren*{t}= \min\curl*{\sT_1(t),\sT_2(t)}$,
\begin{align*}
\text{where} \quad &\sT_1(t):=\h{\sT}_1(t)\mathds{1}_{t\in [1/2,1]}+ 2\,\h{\sT}_1(1/2)\, t\mathds{1}_{t\in [0,1/2)}, \\
    &\sT_2(t):=\h{\sT}_2(t)\mathds{1}_{t\in \left[\e,1\right]}+ \big(\h{\sT}_2(\e)/\epsilon\big)\,t\mathds{1}_{t\in \left[0,\e\right)}, \\
\text{with} \quad &\h{\sT}_1(t):= \inf_{x\in \sX,h\in \ov \sH_\gamma(x)\subsetneqq \sH}\Delta\sC_{\wt{\Phi},\sH}(h,x,t),\\[-.25cm]
    &\h{\sT}_2(t):= \inf_{x\in \sX,h\in\sH\colon \ov h_\gamma(x)< 0}\Delta\sC_{\wt{\Phi},\sH}\big(h,x,\frac{t+1}{2}\big).
\end{align*}
\end{definition}
It is clear that $\sT_{\wt{\Phi}}$ satisfies assumptions in
Theorem~\ref{Thm:excess_bounds_Psi_uniform-adv}. The next theorem
shows that it gives the tightest $\sH$-consistency bound \eqref{eq:bound_Psi_01_adv} under certain conditions.
\begin{restatable}[\textbf{Adversarial tightness}]{theorem}{TightnessAdv}
\label{Thm:tightness-adv}
Suppose that $\sH$ is symmetric and that $\e=0$. If
$\sT_{\wt{\Phi}}=\min\curl*{\sT_1,\sT_2}$ is convex with
$\sT_{\wt{\Phi}}(0)=0$ and $\sT_2\leq \sT_1$, then, for any
$t\in[0,1]$ and $\delta>0$, there exist a distribution $\sD$ and a
hypothesis $h\in\sH$ such that $\sR_{\ell_{\gamma}}(h)-
\sR_{\ell_{\gamma}}^*(\sH)+\sM_{\ell_{\gamma}}(\sH)=t$ and
$\sT_{\wt{\Phi}}(t)
\leq\sR_{\wt{\Phi}}(h)-\sR_{\wt{\Phi}}^*(\sH)+\sM_{\wt{\Phi}}(\sH)\leq
\sT_{\wt{\Phi}}(t) + \delta$.
\end{restatable}
The proof is included in Appendix~\ref{app:tightness}. In other words,
when $\e=0$, if $\sT_2\leq \sT_1$ and $\sT_{\wt{\Phi}}$ is convex with
$\sT_{\wt{\Phi}}(0)=0$, $\sT_{\wt{\Phi}}$ is the optimal function for
the distribution-independent
bound~\eqref{eq:bound_Psi_01_adv}. Moreover, if $\sT_{\wt{\Phi}}$ is
additionally invertible and non-increasing, $\sT_{\wt{\Phi}}^{-1}$ is
the optimal function for the distribution-independent bound in
Theorem~\ref{Thm:excess_bounds_Gamma_uniform-adv}
(Appendix~\ref{app:theorems}) and the two bounds will be equivalent.

We will see that all these assumptions hold for cases considered in
Section~\ref{sec:adv-lin} and \ref{sec:adv-NN}. Next, we will apply
Theorem~\ref{Thm:excess_bounds_Psi_uniform-adv} along with the
tightness guarantee Theorem~\ref{Thm:tightness-adv} to study specific
hypothesis sets and adversarial surrogate loss functions in
Section~\ref{sec:negative} for negative results and
Section~\ref{sec:adv-lin} and \ref{sec:adv-NN} for positive results. A
careful analysis is presented in each case (See
Appendix~\ref{app:derivation-adv}).

\subsection{Negative results for adversarial robustness}
\label{sec:negative}

\citet{awasthi2021calibration} show that supremum-based convex
loss functions of the type, \[\wt{\Phi} = \sup_{x' \colon \|x-x'\|_p\leq
  \gamma}\Phi(y h(x')),\] where $\Phi$ is convex and non-increasing,
are \emph{not $\sH$-calibrated with respect to $\ell_{\gamma}$} for
$\sH$ containing 0, that is \emph{regular for adversarial calibration},
e.g., $\sH_{\mathrm{lin}}$ and $\sH_{\mathrm{NN}}$. 
\begin{definition}[\textbf{Regularity for adversarial calibration}]\emph{[Definition~5 in \citep{awasthi2021calibration}]}
\label{def:regularity}
We say that a hypothesis set $\sH$ is \emph{regular for adversarial calibration}
if there exists a \emph{distinguishing $x$} in $\sX$, that is if there exist $f, g \in \sH$ 
such that $\inf_{\| x' - x \|_p \leq \gamma } f(x')>0$ and 
$\sup_{\| x' - x \|_p \leq \gamma }g(x')<0$.
\end{definition}
Similarly, we
show that there are no non-trivial adversarial $\sH$-consistency
bounds with respect to $\ell_{\gamma}$ for
supremum-based convex loss functions and supremum-based \emph{symmetric loss functions} (see Definition~\ref{def:sym} below) including sigmoid loss
with such hypothesis sets.
\begin{definition}[\textbf{Symmetric loss}]
\label{def:sym}
We say that a margin-based loss $\Phi$ is symmetric if there exists a constant $C\geq 0$ such that $\Phi(t)+\Phi(-t)=C$ for any $t\in \mathbb{R}$, and denote it by $\Phi_{\mathrm{sym}}$. We also define its supremum-based counterpart as $\wt{\Phi}_{\mathrm{sym}}:=\sup_{x' \colon \|x-x'\|_p\leq \gamma}\Phi_{\mathrm{sym}}(y h(x'))$ and call $\wt{\Phi}_{\mathrm{sym}}$ the supremum-based symmetric loss.
\end{definition}
For the sigmoid loss $\Phi_{\mathrm{sig}}(t)=1-\tanh(kt),~k>0$, we have $\Phi_{\mathrm{sig}}(t)+\Phi_{\mathrm{sig}}(-t)=2$, which implies that $\Phi_{\mathrm{sig}}$ is symmetric. Note that \citet{awasthi2021calibration} do
not study the sigmoid loss, which is non-convex. Thus, our results below go
beyond their results for convex adversarial surrogates.
\begin{restatable}[\textbf{Negative results for robustness}]{theorem}{NegativeConvexAdv}
\label{Thm:negative_convex_adv}
Suppose that $\sH$ contains $0$ and is regular for adversarial
calibration. Let $\ell_1$ be supremum-based convex loss or
supremum-based symmetric loss and $\ell_2=\ell_{\gamma}$. Then, $f(t)\geq
1/2$ for any $t\geq 0$ are the only non-decreasing functions $f$ such that
\eqref{eq:est-bound} holds.
\end{restatable}
The proof is given in
Appendix~\ref{app:negative_adv}. In other words, the function $f$ in bound \eqref{eq:est-bound} must be lower bounded by $1/2$ for such adversarial surrogates. Theorem~\ref{Thm:negative_convex_adv}
implies that the loss functions commonly used in practice for
optimizing the adversarial loss cannot benefit from any useful
$\sH$-consistency bound guarantee. Instead, we show in
Section~\ref{sec:adv-lin} and \ref{sec:adv-NN} that the supremum-based
$\rho$-margin loss $\wt{\Phi}_{\rho}=\sup_{x'\colon \|x-x'\|_p\leq\gamma}\Phi_{\mathrm{\rho}}(y h(x'))$ proposed by
\citet{awasthi2021calibration} admits favorable adversarial
$\sH$-consistency bounds. These bounds would also
imply significantly stronger results than the asymptotic
$\sH$-consistency guarantee in \citep{awasthi2021calibration}.


\subsection{Linear hypotheses}
\label{sec:adv-lin}
\begin{table}[t]
\caption{Adversarial $\sH$-consistency bounds. They are completely new consistency bounds in the adversarial setting and can turn into more significant $\e$-consistency results. The minimizability gaps appearing in the bounds for the surrogates are concluded in Table~\ref{tab:loss}. The detailed derivation is included in Appendix~\ref{app:derivation-adv}, \ref{app:derivation-adv_noise}.}
    \label{tab:compare-adv}
\begin{center}
    \begin{tabular}{l|lll}
    \toprule
      Surrogates & Bound ($\sH_{\mathrm{lin}}$) & Bound ($\sH_{\mathrm{NN}}$) & Distribution set\\
    \midrule
     $\wt{\Phi}_{\rho}$ & \eqref{eq:rho-lin-est-adv} & \eqref{eq:rho-NN-est-adv} & All distributions \\
     $\wt{\Phi}_{\mathrm{hinge}}$ & \eqref{eq:hinge-lin-est-adv} & \eqref{eq:hinge-NN-est-adv} & Massart’s noise\\
     $\wt{\Phi}_{\mathrm{sig}}$ & \eqref{eq:sig-lin-est-adv} & \eqref{eq:sig-NN-est-adv} & Massart’s noise\\
    \bottomrule
    \end{tabular}
    \end{center}
\end{table}
In this section, by applying
Theorems~\ref{Thm:excess_bounds_Psi_01_general_adv} and
\ref{Thm:excess_bounds_Gamma_01_general_adv}, we derive the
adversarial $\sH_{\mathrm{lin}}$-consistency
bound~\eqref{eq:rho-lin-est-adv} in Table~\ref{tab:compare-adv} for supremum-based $\rho$-margin loss. This is a completely new consistency bound in the adversarial setting. As with the non-adversarial
case, the bound is dependent on the parameter $B$ in linear hypothesis
set and $\rho$ in the loss function. This helps guide the choice of
loss functions once the hypothesis set is fixed. More precisely, if
$B>0$ is known, we can always choose $\rho<B$ such that the bound is
the tightest. Moreover, the bound can turn into more significant
$\e$-consistency results in adversarial setting than the
$\sH$-consistency result in \citep{awasthi2021calibration}.
\begin{corollary}
\label{cor:lin-adv-stonger}
 Let $\sD$ be a distribution over $\sX\times\sY$ such that
 $\sM_{\wt{\Phi}_{\rho}}\paren*{\sH_{\mathrm{lin}}}\leq\e$ for some $\e\geq
 0$.  Then, the following holds:
\begin{align*}
 \sR_{\ell_{\gamma}}
 (h)
 - 
 \sR_{\ell_{\gamma}}^*\paren*{\sH_{\mathrm{lin}}} 
  \leq 
  \rho\paren*{\sR_{\wt{\Phi}_{\rho}}
  (h)
  -
  \sR_{\wt{\Phi}_{\rho}}^*\paren*{\sH_{\mathrm{lin}}}
  +
 \e}/\min\curl*{B,\rho}.
\end{align*}
\end{corollary}
\citet{awasthi2021calibration} show that $\wt{\Phi}_{\rho}$ is
$\sH_{\mathrm{lin}}$-consistent with respect to $\ell_{\gamma}$ when
$\sM_{\wt{\Phi}_{\rho}}\paren*{\sH_{\mathrm{lin}}}=0$. This result can be
immediately implied by Corollary~\ref{cor:lin-adv-stonger}. Moreover,
Corollary~\ref{cor:lin-adv-stonger} provides guarantees for more
general cases where $\sM_{\wt{\Phi}_{\rho}}\paren*{\sH_{\mathrm{lin}}}$ can be
nonzero.

\subsection{One-hidden-layer ReLU neural networks}
\label{sec:adv-NN}

For the one-hidden-layer ReLU neural networks $\sH_{\mathrm{NN}}$ and
$\wt{\Phi}_{\rho}$, we have the $\sH_{\mathrm{NN}}$-consistency
bound \eqref{eq:rho-NN-est-adv} in Table~\ref{tab:compare-adv}.
Note $\inf_{x\in\sX}\sup_{h\in\sH_{\mathrm{NN}}}\uv h_\gamma(x)$ does
not have an explicit expression. However, \eqref{eq:rho-NN-est-adv}
can be further relaxed to be \eqref{eq:rho-NN-est-adv-2} in
Appendix~\ref{app:derivation-NN-adv}, which is identical to the bound
in the linear case modulo the replacement of $B$ by $\Lambda B$. As in
the linear case, the bound is new and also
implies stronger $\e$-consistency results as follows:
\begin{corollary}
Let $\sD$ be a distribution over $\sX\times\sY$ such that
$\sM_{\wt{\Phi}_{\rho}}\paren*{\sH_{\mathrm{NN}}}\leq \e$ for some $\e\geq 0$.
Then,
\begin{align*}
 \sR_{\ell_{\gamma}}
 (h)
 - 
 \sR_{\ell_{\gamma}}^*\paren*{\sH_{\mathrm{NN}}} 
  \leq 
  \rho\paren*{\sR_{\wt{\Phi}_{\rho}}
  (h)
  -
  \sR_{\wt{\Phi}_{\rho}}^*\paren*{\sH_{\mathrm{NN}}}
  +
 \e}
 /
 \min\curl*{\Lambda B,\rho}.
\end{align*}
\end{corollary}
Besides the bounds for $\wt{\Phi}_{\rho}$, Table~\ref{tab:compare-adv}
gives a series of results that are all new in the adversarial
setting. Like the bounds in Table~\ref{tab:compare} and
\ref{tab:compare-NN}, they are all hypothesis set dependent and very
useful. For example, the improved bounds for
$\wt{\Phi}_{\mathrm{hinge}}$ and $\wt{\Phi}_{\mathrm{sig}}$ under
noise conditions in the table can also turn into meaningful
consistency results under Massart's noise condition, as shown in
Section~\ref{sec:noise-adv}.

\subsection{Guarantees under Massart's noise condition}
\label{sec:noise-adv}

Section~\ref{sec:negative} shows that non-trivial
distribution-independent bounds for supremum-based hinge loss and
supremum-based sigmoid loss do not exist. However, under Massart's
noise condition (Definition~\ref{def:massarts-noise}), we will show
that there exist non-trivial adversarial $\sH$-consistency bounds for the two loss functions. Furthermore, we will see that
the bounds are linear dependent as those in
Section~\ref{sec:noise-non-adv}.

As with the non-adversarial scenario, we introduce a modified
adversarial $\sH$-estimation error transformation in
Proposition~\ref{prop-adv-noise}
(Appendix~\ref{app:derivation-adv_noise}). Using this tool, we derive
adversarial $\sH$-consistency bounds for
$\wt{\Phi}_{\mathrm{hinge}}$ and $\wt{\Phi}_{\mathrm{sig}}$ under
Massart's noise condition in Table~\ref{tab:compare-adv}.
From the bounds~\eqref{eq:hinge-lin-est-adv},
\eqref{eq:sig-lin-est-adv}, \eqref{eq:hinge-NN-est-adv}, and
\eqref{eq:sig-NN-est-adv}, we can also obtain novel $\e$-consistency
results for $\wt{\Phi}_{\mathrm{hinge}}$ and
$\wt{\Phi}_{\mathrm{sig}}$ with linear models and neural networks
under Massart's noise condition.
\begin{corollary}
Let $\sH$ be $\sH_{\mathrm{lin}}$ or $\sH_{\mathrm{NN}}$. Let $\sD$ be
a distribution over $\sX\times\sY$ which satisfies Massart's noise
condition with $\beta$ such that $\sM_{\wt{\Phi}}(\sH)\leq\e$ for some
$\e\geq 0$. Then,
\begin{align*}
\sR_{\ell_{\gamma}}(h)- \sR_{\ell_{\gamma}}^*(\sH) \leq 
     \frac{1+2\beta}{4\beta}\paren{\sR_{\wt{\Phi}}(h)-\sR_{\wt{\Phi}}^*(\sH)+\e}/\sT(B),
\end{align*}
where $\sT(t)$ equals to $\min\curl*{t,1}$ and $\tanh\paren*{k t}$ for
$\wt{\Phi}_{\mathrm{hinge}}$ and $\wt{\Phi}_{\mathrm{sig}}$
respectively, $B$ is replaced by $\Lambda B$ for
$\sH=\sH_{\mathrm{NN}}$.
\end{corollary}
In Section~\ref{sec:simulations}, we will further show that these
linear dependency bounds in adversarial setting are tight, along with
the non-adversarial bounds we discussed earlier in
Section~\ref{sec:noise-non-adv}.

\section{Simulations}
\label{sec:simulations}

\begin{figure}[t]
\begin{center}
\includegraphics[scale=0.5]{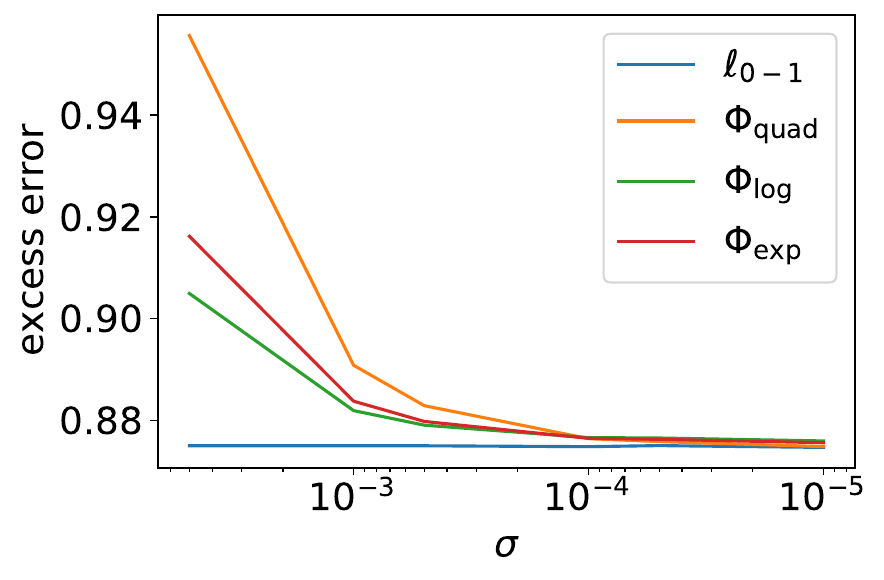}
\includegraphics[scale=0.5]{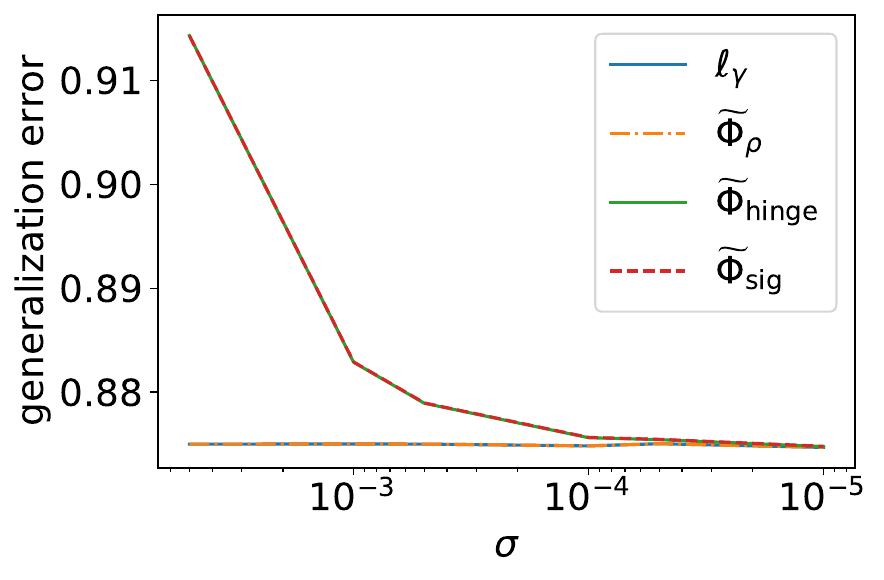}
\caption{Left: tightness of bound~\eqref{eq:non-adv-noise} in
  Section~\ref{sec:noise-non-adv}. Right: tightness of
  bounds~\eqref{eq:rho-lin-est-adv}, \eqref{eq:hinge-lin-est-adv} and
  \eqref{eq:sig-lin-est-adv} in Section~\ref{sec:adv-lin} and
  \ref{sec:noise-adv}.}
\label{fig:simulation}
\end{center}
\end{figure}
Here, we present experiments on simulated data to illustrate our
bounds and their tightness. We generate data points $x \in \mathbb{R}$
on $[-1, +1]$. All risks are approximated by their empirical
counterparts computed over $10^7$
i.i.d.\ samples.

\textbf{Non-adversarial.} To demonstrate the
tightness of our non-adversarial bounds, we consider a scenario where
the marginal distribution is symmetric about $x = 0$ with labels
flipped. With probability $\frac{1}{16}$, $(x, y)=(1, -1)$; with
probability $\frac{7}{16}$, the label is $+1$ and the data follows the
truncated normal distribution on $[\sigma, 1]$ with both mean and
standard deviation $\sigma$. We consider
$\Phi_{\mathrm{quad}}$, $\Phi_{\mathrm{log}}$ and
$\Phi_{\mathrm{exp}}$ defined in Table~\ref{tab:loss}.  The distribution considered satisfies
Massart's noise condition with $\beta = \frac{1}{2}$. Thus, our
bound \eqref{eq:non-adv-noise} in Section~\ref{sec:noise-non-adv}
becomes $\sR_{\ell_{0-1}}(h)- \sR_{\ell_{0-1}}^*\paren*{\sH_{\mathrm{all}}}
\leq \sR_{\Phi}(h)- \sR_{\Phi}^*\paren*{\sH_{\mathrm{all}}}$, for any $h\in
\sH_{\mathrm{all}}$ such that $\sR_{\Phi}(h) \leq
\sR_{\Phi}^*\paren*{\sH_{\mathrm{all}}}+1$. All the minimal generalization
errors vanish in this case. As shown in Figure~\ref{fig:simulation},
for $h(x) = -5x$, the bounds corresponding to $\Phi_{\mathrm{quad}}$,
$\Phi_{\mathrm{log}}$ and $\Phi_{\mathrm{exp}}$ are all tight as
$\sigma \to 0$.

\textbf{Adversarial}.  To demonstrate the tightness of
our adversarial bounds, the distribution is modified as follows: with
probability $\frac{1}{16}$, $(x,y)=(1,-1)$; with probability
$\frac{1}{16}$, $(x,y)=(-1,+1)$; with probability $\frac{7}{8}$, the
label is $-1$ and the data follows the truncated normal distribution
on $[-1, \gamma-\sigma]$ with mean $\gamma-\sigma$ and standard
deviation $\sigma$. We set $\gamma=0.1$ and consider
$\wt{\Phi}_{\rho}$ with $\rho=1$, $\wt{\Phi}_{\mathrm{hinge}}$ and
$\wt{\Phi}_{\mathrm{sig}}$ with $k=1$.  The distribution considered
satisfies Massart's noise condition with $\beta = \frac{1}{2}$. Thus, our
bounds \eqref{eq:rho-lin-est-adv}, \eqref{eq:hinge-lin-est-adv} and
\eqref{eq:sig-lin-est-adv} in Table~\ref{tab:compare-adv} become
$\sR_{\ell_{\gamma}}(h) \leq \sR_{ \ \wt{\Phi}}(h)$, for any $h\in
\sH_{\mathrm{lin}}$. As shown in Figure~\ref{fig:simulation}, for
$h(x) = -5x$, the bounds corresponding to $\wt{\Phi}_{\rho}$,
$\wt{\Phi}_{\mathrm{hinge}}$ and $\wt{\Phi}_{\mathrm{sig}}$ are all
tight as $\sigma \to 0$.

\section{Conclusion}

We presented an exhaustive study of $\sH$-consistency
bounds, including a series of new guarantees for both the
non-adversarial zero-one loss function and the adversarial zero-one
loss function. Our hypothesis-dependent guarantees are significantly
stronger than the consistency or calibration ones.  Our results
include a series of theoretical and conceptual tools helpful for the
analysis of other loss functions and other hypothesis sets, including
multi-class classification or ranking losses. They can be further extended to the analysis of non-i.i.d.\ settings such as that of drifting distributions \citep{HelmboldLong1994,Long1999,BarveLong1997,BartlettBenDavidKulkarni2000,
MohriMunozMedina2012,Gama2014} or, more generally, time series prediction \citep{Engle1982,Bollerslev1986,BrockwellDavis1986,BoxJenkins1990,
Hamilton1994,Meir2000,KuznetsovMohri2015,KuznetsovMohri2017,KuznetsovMohri2020}. Our results can also be extended to many other loss functions, using our general proof techniques or a similar analysis.

\chapter{Multi-Class Classification: Max Losses, Sum Losses and Constrained Losses} \label{ch3}
In this chapter, we present an extensive study of $\sH$-consistency bounds for
multi-class classification.
We show in Section~\ref{sec:max-losses} that, in general, no
non-trivial $\sH$-consistency bounds can be derived for multi-class
\emph{max losses} such as those of \citet{crammer2001algorithmic},
when used with a convex loss auxiliary function such as the hinge
loss.  On the positive side, we prove multi-class $\sH$-consistency
bounds for max losses under a realizability assumption and give
multi-class $\sH$-consistency bounds using as an auxiliary function
the $\rho$-margin loss, without requiring a realizability assumption.
For \emph{sum losses}, that is multi-class losses such as that of
\citet{weston1998multi}, we give a series of results, including a
negative result when using as auxiliary function the hinge-loss, and
$\sH$-consistency bounds when using the exponential loss, the squared
hinge-loss, and the $\rho$-margin loss (Section~\ref{sec:sum-losses}).
We also present a series of results for the so-called
\emph{constrained losses}, such as the loss function adopted by
\citet{lee2004multicategory} in the analysis of multi-class SVM. Here,
we prove multi-class $\sH$-consistency bounds when using as an
auxiliary function the hinge-loss, the squared hinge-loss, the
exponential loss, and the $\rho$-margin loss
(Section~\ref{sec:constrained-losses}).
We further give multi-class \emph{adversarial} $\sH$-consistency
bounds for all three of the general multi-class losses just mentioned
(max losses, sum losses and constrained losses) in
Section~\ref{sec:adv-mhcb}.

We are not aware of any prior $\sH$-consistency bound derived in the
multi-class setting, even in the special case of $\sH$ being the
family of all measurable functions, whether in the non-adversarial
or adversarial setting. All of our results are novel,
including our proof techniques.  Our results are given for the
hypothesis set $\sH$ being the family of all measurable functions, the
family of linear functions, or the family of one-hidden-layer ReLU
neural networks. 
The binary classification results of \citet{awasthi2022Hconsistency}
do not readily extend to the multi-class setting since the study of
calibration and conditional risk is more complex, the form of the
surrogate losses is more diverse, and in general the analysis is more
involved and requires entirely novel proof techniques in the
multi-class setting (see Section~\ref{sec:general-mhcb} for a more detailed
discussion of this point).

We start with the introduction of
several multi-class definitions, as well as key concepts and
definitions related to the study of $\sH$-consistency bounds
(Section~\ref{sec:preliminaries}).

The presentation in this chapter is based on \citep{awasthi2022multi}.

\section{Preliminaries}
\label{sec:preliminaries}

We consider the familiar multi-class classification scenario with $c
\geq 2$ classes.  We denote by $\sX$ the input space and by $\sY =
\curl*{1, \ldots, c}$ the set of classes or categories. Let $\sH$ be a
hypothesis set of functions mapping from $\sX \times \sY$ to
$\Rset$. The label $\hh(x)$ associated by a hypothesis $h \in \sH$ to
$x \in \sX$ is the one with the largest score: $\hh(x) = \argmax_{y
  \in \sY} h(x, y)$ with an arbitrary but fixed deterministic strategy
used for breaking ties. For simplicity, we fix that strategy to be
the one selecting the label with the highest index under the natural
ordering of labels. See Appendix~\ref{app:dicussion-01} for
a more detailed discussion of this choice.

The \emph{margin} $\rho_{h}(x, y)$ of a hypothesis $h\in \sH$ for a
labeled example $(x, y) \in \sX \times \sY$ is defined by
\begin{align*}
\rho_h(x, y) = h(x, y) - \max_{y' \neq y} h(x, y'),
\end{align*}
that is the difference between the score assigned to $(x, y)$ and that
of the runner-up.  Given a distribution $\sD$ over $\sX \times \sY$
and a loss function $\ell \colon \sH \times \sX \times \sY \to \Rset$,
the \emph{generalization error} of a hypothesis $h \in \sH$ and the
\emph{minimal generalization error} are defined as follows:
\begin{align*}
    \sR_{\ell}(h) = \E_{(x, y) \sim \sD}[\ell(h, x, y)] \quad\text{ and }\quad \sR_{\ell}^*(\sH)=\inf_{h \in \sH}\sR_{\ell}(h).
\end{align*}
The goal in multi-class classification is to select a hypothesis $h
\in \sH$ with small generalization error with respect to the
multi-class $0/1$ loss defined, for any $h \in \sH$, by $\ell_{0-1}(h,
x, y) = \mathds{1}_{\hh(x) \neq y}$.
In the adversarial scenario, the goal is to select a hypothesis $h \in
\sH$ with small \emph{adversarial generalization error} defined, for
any $\gamma \in (0, 1)$ and $p\in [1, +\infty]$, by
$\sR_{\ell_{\gamma}}(h) = \E_{(x, y)\sim \sD}[\ell_{\gamma}(h, x,
  y)]$, where
\begin{align*}
\ell_{\gamma}(h, x, y) = \sup_{x'\colon \|x - x'\|_p \leq
  \gamma}\mathds{1}_{\rho_h(x',y) \leq 0}=\mathds{1}_{\inf_{x'\colon
    \norm*{x - x'}_p \leq \gamma} \rho_h(x', y) \leq 0},
\end{align*}
is the adversarial multi-class $0/1$ loss. More generally, the
\emph{adversarial generalization error} and \emph{minimal adversarial
generalization error} for a loss function $\ell(h,x,y)$ are defined as
follows:
\begin{align*}
    \sR_{\wt{\ell}}(h) = \E_{(x, y) \sim \sD}\bracket*{\wt{\ell}(h, x, y)} \quad\text{ and }\quad \sR_{\wt{\ell}}^*(\sH)=\inf_{h\in\sH}\sR_{\wt{\ell}}(h),
\end{align*}
where $\wt{\ell}(h, x, y)=\sup_{x'\colon \|x-x'\|_p\leq\gamma}\ell(h,x',y)$ is the supremum-based counterpart of $\ell$.

For a distribution $\sD$ over $\sX \times \sY$, we define, for any $x
\in \sX$, $p(x)=(\sfp(1 \!\mid\! x), \ldots, \sfp(c \!\mid\! x))$, where $\sfp(y \!\mid\! x) = \sD(Y =
y \!\mid\! X = x)$ is the conditional probability of $Y = y$ given $X
= x$.  We can then write the generalization error as
$\sR_{\ell}(h)=\mathbb{E}_{X}\bracket*{\sC_{\ell}(h,x)}$, where
$\sC_{\ell}(h,x)$ is the \emph{conditional $\ell$-risk} defined by
$\sC_{\ell}(h,x) = \sum_{y\in \sY} \sfp(y \!\mid\! x) \ell(h,x,y)$.
We will denote by $\sP$ a set of distributions $\sD$ over $\sX \times
\sY$ and by $\sP_{\mathrm{all}}$ the set of all such
distributions. For convenience, we define $y_{\max}$ by $y_{\max} =
\argmax_{y \in \sY} \sfp(y \!\mid\! x)$. When there is a tie, we pick the label
with the highest index under the natural ordering of labels.

The \emph{minimal conditional $\ell$-risk} is denoted by
$\sC_{\ell}^*(\sH)(x) = \inf_{h\in \sH}\sC_{\ell}(h,x)$. We also use
the following shorthand for the gap $\Delta\sC_{\ell,\sH}(h,x) =
\sC_{\ell}(h,x)-\sC_{\ell}^*(\sH)(x)$ and call
$\Delta\sC_{\ell,\sH}(h,x)\mathds{1}_{\Delta\sC_{\ell,\sH}(h,x)>\epsilon}$
the \emph{conditional $\e$-regret} for $\ell$. For convenience, we
also define, for any vector $\tau = (\tau_1, \ldots, \tau_c)$ in the
probability simplex of $\Rset^c$, $\sC_{\ell}(h, x, \tau) = \sum_{y\in
  \sY} \tau_y \, \ell(h,x,y)$, $\sC_{\ell}^*(\sH)(x,\tau)=\inf_{h \in
  \sH}\sC_{\ell}(h, x, \tau)$ and $\Delta\sC_{\ell,\sH}(h, x, \tau) =
\sC_{\ell}(h, x, \tau) -\sC_{\ell}^*(\sH)(x,\tau)$. Thus, we have
$\Delta\sC_{\ell,\sH}(h, x, p(x)) = \Delta\sC_{\ell, \sH}(h, x)$.
For any $\e > 0$, we will denote by $\bracket*{t}_{\e}$ the
$\e$-truncation of $t \in \Rset$ defined by
$t\mathds{1}_{t>\e}$. Thus, the conditional $\e$-regret can be
rewritten as $\bracket*{\Delta\sC_{\ell,\sH}(h,x)}_{\e}$.

For a hypothesis set $\sH$ and distribution $\sD$, we also define the
\emph{$\paren*{\ell,\sH}$-minimizability gap} as $ \sM_{\ell}(\sH) =
\sR^*_{\ell}(\sH) - \mathbb{E}_{X} \bracket* {\sC^*_{\ell}(\sH, x)} $,
that is the difference between the best-in class error and the
expectation of the minimal conditional $\ell$-risk. This is a key
quantity appearing in our bounds that we cannot hope to estimate or
minimize. Its value only depends on the distribution $\sD$ and the
hypothesis set $\sH$. As an example, when $\sH$ is the family of all
measurable functions, then the minimizability gap for the multi-class
$0/1$ loss is zero for any distribution $\sD$.

\section{General theorems}
\label{sec:general-mhcb} 

The general form of the \emph{$\sH$-consistency bounds} that we are
seeking for a surrogate loss $\ell_1$ of a target loss $\ell_2$ is
$\sR_{\ell_2}(h)- \sR_{\ell_2}^*(\sH) \leq
f\paren{\sR_{\ell_1}(h)-\sR_{\ell_1}^*(\sH)}$ for all $h \in \sH$, for
some non-decreasing function $f$.
To derive such bounds for surrogate multi-class losses, we draw on the
following two general theorems, which show that, under some
conditions, the target loss estimation error can be bounded by some
functional form of the surrogate loss estimation error involving
minimizability gaps.

\begin{restatable}[\textbf{Distribution-dependent $\Psi$-bound}]
  {theorem}{ExcessBoundsPsiMhcb}
\label{Thm:excess_bounds_Psi-mhcb}
Assume that there exists a convex function $\Psi\colon
\mathbb{R_{+}}\to \Rset$ with $\Psi(0)\geq0$ and $\epsilon\geq0$
such that the following holds for all $h\in \sH$, $x\in \sX$ and
$\sD\in \sP$:
$\Psi\paren*{\bracket*{\Delta\sC_{\ell_2,\sH}(h,x)}_{\e}}\leq
\Delta\sC_{\ell_1,\sH}(h,x)$. Then, for any hypothesis $h\in\sH$ and
any distribution $\sD\in \sP$,
    \begin{align*}
     \Psi\paren*{\sR_{\ell_2}(h)- \sR_{\ell_2}^*(\sH)+\sM_{\ell_2}(\sH)}\leq  \sR_{\ell_1}(h)-\sR_{\ell_1}^*(\sH)+\sM_{\ell_1}(\sH)+\max\curl*{\Psi(0),\Psi(\e)}.
    \end{align*}
\end{restatable}

\begin{restatable}[\textbf{Distribution-dependent $\Gamma$-bound}]
  {theorem}{ExcessBoundsGammaMhcb}
\label{Thm:excess_bounds_Gamma-mhcb}
Assume that there exists a concave function $\Gamma\colon
\mathbb{R_{+}}\to \Rset$ and $\epsilon\geq0$ such that the
following holds for all $h\in \sH$, $x\in \sX$ and $\sD\in \sP$:
$\bracket*{\Delta\sC_{\ell_2,\sH}(h,x)}_{\e}\leq \Gamma
\paren*{\Delta\sC_{\ell_1,\sH}(h,x)}$.  Then, for any hypothesis
$h\in\sH$ and any distribution $\sD\in \sP$,
    \begin{align*}
     \sR_{\ell_2}(h)- \sR_{\ell_2}^*(\sH)\leq  \Gamma\paren*{\sR_{\ell_1}(h)-\sR_{\ell_1}^*(\sH)+\sM_{\ell_1}(\sH)}-\sM_{\ell_2}(\sH)+\epsilon.
    \end{align*}
\end{restatable}

The theorems show that, to derive such bounds for a specific
hypothesis set and a set of distributions, it suffices to verify that
for the same hypothesis set and set of distributions, the conditional
$\epsilon$-regret for the target loss can be upper-bounded with the
same functional form of the gap between the conditional risk and
minimal conditional risk of the surrogate loss.
These results are similar to their binary classification counterparts
due to \citet{awasthi2022Hconsistency}.  In particular, the
conditional $\ell$-risk $\sC_{\ell}(h,x)$ in our theorems is the
multi-class generalization of their binary definition. The proofs are
similar and are included in Appendix~\ref{app:deferred_proofs_general}
for completeness.

For a given hypothesis set $\sH$, the resulting bounds suggest three
key ingredients for the choice of a surrogate loss: (1) the functional
form of the $\sH$-consistency bound, which is specified by the
function $\Psi$ or $\Gamma$; (2) the smoothness of the loss and more
generally its optimization virtues, as needed for the minimization of
$\sR_{\ell_1}(h) - \sR_{\ell_1}^*(\sH)$; (3) and the approximation
properties of the surrogate loss function which determine the value of
the minimizability gap $\sM_{\ell_1}(\sH)$.  Our quantitative
$\sH$-consistency bounds can help select the most favorable surrogate
loss function among surrogate losses with good optimization merits and
comparable approximation properties.

In Section~\ref{sec:non-adv-mhcb} and Section~\ref{sec:adv-mhcb}, we will apply
Theorem~\ref{Thm:excess_bounds_Psi-mhcb} and
Theorem~\ref{Thm:excess_bounds_Gamma-mhcb} to the analysis of multi-class
loss functions and hypothesis sets widely used in practice. Here, we
wish to first comment on the novelty of our results and proof
techniques.  Let us emphasize that although the general tools of
Theorems~\ref{Thm:excess_bounds_Psi-mhcb} and \ref{Thm:excess_bounds_Gamma-mhcb}
are the multi-class generalization of that in
\citep{awasthi2022Hconsistency}, the binary classification results of
\citet{awasthi2022Hconsistency} do not readily extend to the
multi-class setting.  This is true, even in the classical study of
Bayes-consistency, where the multi-class setting
\citep{tewari2007consistency} does not readily follow the binary case
\citep{bartlett2006convexity} and required an alternative analysis and
new proofs. Note that, additionally, in the multi-class setting,
surrogate losses are more diverse: we will distinguish max losses, sum
losses, and constrained losses and present an analysis for each loss
family with various auxiliary functions for each (see
Section~\ref{sec:non-adv-mhcb}).

\textbf{Proof techniques}. More specifically, the need for novel proof
techniques stems from the following.
To use Theorem \ref{Thm:excess_bounds_Psi-mhcb} and Theorem
\ref{Thm:excess_bounds_Gamma-mhcb}, we need to find $\Psi$ and $\Gamma$
such that the inequality conditions in these theorems hold. This
requires us to characterize the conditional risk and the minimal
conditional risk of the multi-class zero-one loss function and the
corresponding ones for diverse surrogate loss functions in both the
non-adversarial and adversarial scenario. Unlike the binary case, such
a characterization in the multi-class setting is very difficult. For
example, for the constrained loss, solving the minimal conditional
risk given a hypothesis set is equivalent to solving a $c$-dimensional
constrained optimization problem, which does not admit an analytical
expression. In contrast, in the binary case, solving the minimal
conditional risk is equivalent to solving a minimization problem for a
univariate function and the needed function $\Psi$ can be
characterized explicitly by the $\sH$-estimation error transformation,
as shown in \citep{awasthi2022Hconsistency}. Unfortunately, such
binary classification transformation tools cannot be adapted to the
multi-class setting. Instead, in our proof for the multi-class
setting, we adopt a new idea that avoids directly characterizing the
explicit expression of the minimal conditional risk.

For example, for the constrained loss, we leverage the 
condition of \citep{lee2004multicategory} that the scores sum to zero,
and appropriately choose a hypothesis $\ov h$ that differs from $h$
only by its scores for $\hh(x)$ and $y_{\max}$ (see
Appendix~\ref{app:deferred_proofs_lee}). Then, we can upper-bound the
minimal conditional risk by the conditional risk of $\ov h$ without
having to derive the closed form expression of the minimal conditional
risk. Therefore, the conditional regret of the surrogate loss can be
lower bounded by that of the zero-one loss with an appropriate
function $\Psi$. To the best of our knowledge, this proof idea and
technique are entirely novel. We believe that they can be used for the
analysis of other multi-class surrogate losses. Furthermore, all of
our multi-class $\sH$-consistency results are new. Likewise, our
proofs of the $\sH$-consistency bounds for sum losses for the squared
hinge loss and exponential loss use similarly a new technique and
idea, and so does the proof for the $\rho$-margin loss. Furthermore, we
also present an analysis of the adversarial scenario (see
Section~\ref{sec:adv-mhcb}), for which the multi-class proofs are also
novel. Finally, our bounds in the multi-class setting are more general:
for $c = 2$, we recover the binary classification bounds of
\citep{awasthi2022Hconsistency}. Thus, our bounds benefit from the
same tightness guarantees shown by \citep{awasthi2022Hconsistency}. A
further analysis of the tightness of our guarantees in the multi-class
setting is left to future work.

\section{\texorpdfstring{$\sH$}{H}-consistency bounds}
\label{sec:non-adv-mhcb}

In this section, we discuss $\sH$-consistency bounds in the
non-adversarial scenario where the target loss $\ell_2$ is
$\ell_{0-1}$, the multi-class $0/1$ loss. The lemma stated next
characterizes the minimal conditional $\ell_{0-1}$-risk and the
corresponding conditional $\epsilon$-regret, which will be helpful
for instantiating Theorems~\ref{Thm:excess_bounds_Psi-mhcb} and
\ref{Thm:excess_bounds_Gamma-mhcb} in the non-adversarial scenario. 
For any $x \in \sX$, 
we will denote, by $\mathsf H(x)$
the set of labels generated by hypotheses in $\sH$:
$\mathsf H(x) =
\curl*{\hh(x) \colon h \in \sH}$.
\ignore{
For convenience, we will adopt the following notation:
\begin{align*}
\mathsf H(x) =
\curl*{\hh(x) \colon h \in \sH}.
\end{align*}
\ignore{
\begin{align*}
\mathsf H(x) =
\curl*{y\in\sY \colon \exists h\in \sH \text{ such that } \hh(x)=y}.
\end{align*}
}
For a fixed $x\in \sX$, $\mathsf H(x)$ consists of all the labels that
can be associated to $x$ by hypotheses in $\sH$.
}

\begin{restatable}{lemma}{ExplicitAssumptionMhcb}
\label{lemma:explicit_assumption_01-mhcb}
For any $x \in \sX$,
the minimal conditional $\ell_{0-1}$-risk and
the conditional $\epsilon$-regret for $\ell_{0-1}$ can be expressed as follows:
\begin{align*}
\sC^*_{\ell_{0-1}}(\sH, x) & = 1 - \max_{y\in \mathsf H(x)} \sfp(y \!\mid\! x)\\
\bracket*{\Delta\sC_{\ell_{0-1},\sH}(h,x)}_{\e} & = \bracket*{\max_{y\in \mathsf H(x)} \sfp(y \!\mid\! x) - \sfp(\hh(x) \!\mid\! x)}_{\e}.
\end{align*}
\end{restatable}
The proof of Lemma~\ref{lemma:explicit_assumption_01-mhcb} is given in
Appendix~\ref{app:deferred_proofs_cond}. By
Lemma~\ref{lemma:explicit_assumption_01-mhcb},
Theorems~\ref{Thm:excess_bounds_Psi-mhcb} and \ref{Thm:excess_bounds_Gamma-mhcb}
can be instantiated as Theorems~\ref{Thm:excess_bounds_Psi_01_general-mhcb}
and \ref{Thm:excess_bounds_Gamma_01_general-mhcb} in the non-adversarial
scenario as follows, where $\sH$-consistency bounds are provided
between the multi-class $0/1$ loss and a surrogate loss $\ell$.
\begin{theorem}[\textbf{Non-adversarial distribution-dependent $\Psi$-bound}]
\label{Thm:excess_bounds_Psi_01_general-mhcb}
Assume that there exists a convex function
$\Psi\colon \mathbb{R_{+}} \to \Rset$ with $\Psi(0)\geq 0$ and
$\epsilon\geq0$ such that the following holds for all $h\in \sH$, $x\in \sX$ and $\sD\in \sP$:
\begin{align}
\label{eq:condition_Psi_general-mhcb}
    \Psi\paren*{\bracket*{\max_{y\in \mathsf H(x)} \sfp(y \!\mid\! x) - \sfp(\hh(x) \!\mid\! x)}_{\e}}
    \leq \Delta\sC_{\ell}(\sH)(h,x).
\end{align}
Then, for any hypothesis $h \in \sH$ and any distribution $\sD\in \sP$, we have
\begin{align}
\label{eq:bound_Psi_01-mhcb}
     \Psi\paren*{\sR_{\ell_{0-1}}(h)- \sR_{\ell_{0-1}}^*(\sH)+\sM_{\ell_{0-1}}(\sH)}
     \leq  \sR_{\ell}(h)-\sR_{\ell}^*(\sH)+\sM_{\ell}(\sH)+\max\curl*{\Psi(0),\Psi(\e)}.
\end{align}
\end{theorem}

\begin{theorem}[\textbf{Non-adversarial distribution-dependent $\Gamma$-bound}]
\label{Thm:excess_bounds_Gamma_01_general-mhcb}
Assume that there exists a concave function $\Gamma\colon \mathbb{R_{+}}\to \Rset$
and $\epsilon\geq0$ such that the following holds for all $h\in \sH$, $x\in \sX$ and $\sD\in \sP$:
\begin{align}
\label{eq:condition_Gamma_general-mhcb}
\bracket*{\max_{y\in \mathsf H(x)} \sfp(y \!\mid\! x) - \sfp(\hh(x) \!\mid\! x)}_{\e} \leq \Gamma\paren*{\Delta\sC_{\ell,\sH}(h,x)}.
\end{align}
Then, for any hypothesis $h\in\sH$ and any distribution $\sD\in \sP$, we have
\begin{align}
\label{eq:bound_Gamma_01-mhcb}
     \sR_{\ell_{0-1}}(h)- \sR_{\ell_{0-1}}^*(\sH)
     \leq  \Gamma\paren*{\sR_{\ell}(h)-\sR_{\ell}^*(\sH)+\sM_{\ell}(\sH)}
     -\sM_{\ell_{0-1}}(\sH)+\epsilon.
\end{align}
\end{theorem}
In the following, we will apply
Theorems~\ref{Thm:excess_bounds_Psi_01_general-mhcb} and
\ref{Thm:excess_bounds_Gamma_01_general-mhcb} to study the
$\sH$-consistency bounds for different families of multi-class losses
parameterized by various auxiliary functions, for several general
hypothesis sets. It is worth emphasizing that the form of the
surrogate losses is more diverse in the multi-class setting and each
case requires a careful analysis and that the techniques used in the
binary case \citep{awasthi2022Hconsistency} do not apply and cannot
be readily extended to our case.

\textbf{Hypothesis sets}. Let $B_p^d(r) = \curl*{z \in \Rset^d \mid
  \norm*{z}_p\leq r}$ denote the $d$-dimensional
$\ell_p$-ball with radius $r$\ignore{, where $\norm*{z}_p\colon
=\bracket*{\sum_{i=1}^d \abs*{z_i}^p}^{\frac{1}{p}}$}, with $p\in [1,
  +\infty]$.  Without loss of generality, in the following, we choose
$\sX = B_p^d(1)$. Let $p, q \in[1, +\infty]$ be conjugate indices,
that is $\frac{1}{p} + \frac{1}{q} = 1$. In the following, we will
specifically study three families: the family of all measurable functions
$\sH_{\mathrm{all}}$, the family of linear hypotheses
\[
\sH_{\mathrm{lin}} = \curl*{(x,y) \mapsto w_y \cdot x + b_y \mid
  \norm*{w_y}_q\leq W,\abs*{b_y}\leq B},
\]
and that of one-hidden-layer ReLU networks defined by the
following, where $(\cdot)_+ = \max(\cdot, 0)$:
\[
\sH_{\mathrm{NN}} = \curl[\bigg]{(x,y)
\mapsto \sum_{j = 1}^n u_{y,j}(w_{y,j} \cdot x+b_{y,j})_{+} \mid \|u_y
\|_{1}\leq \Lambda,\|w_{y,j}\|_q\leq W, \abs*{b_{y,j}}\leq B}.
\]

\begin{figure}
    \centering
    \includegraphics[scale=0.5]{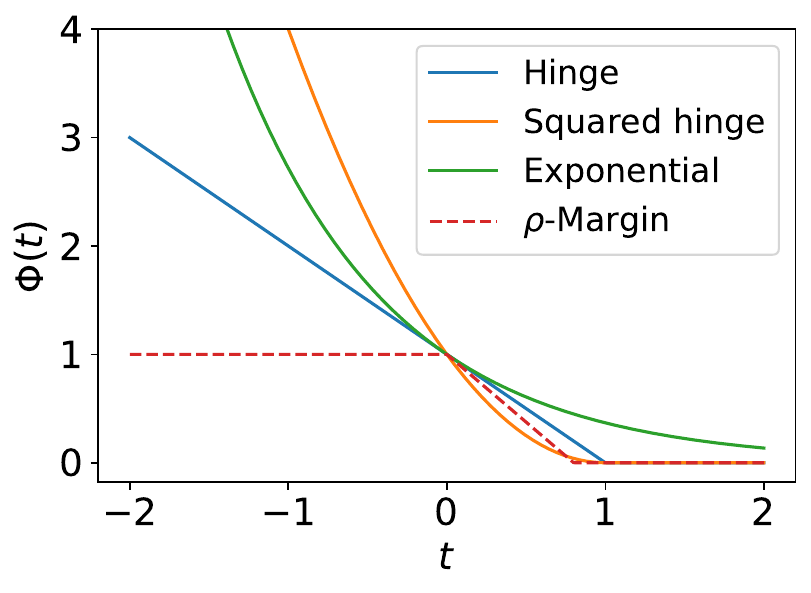}
    \includegraphics[scale=0.5]{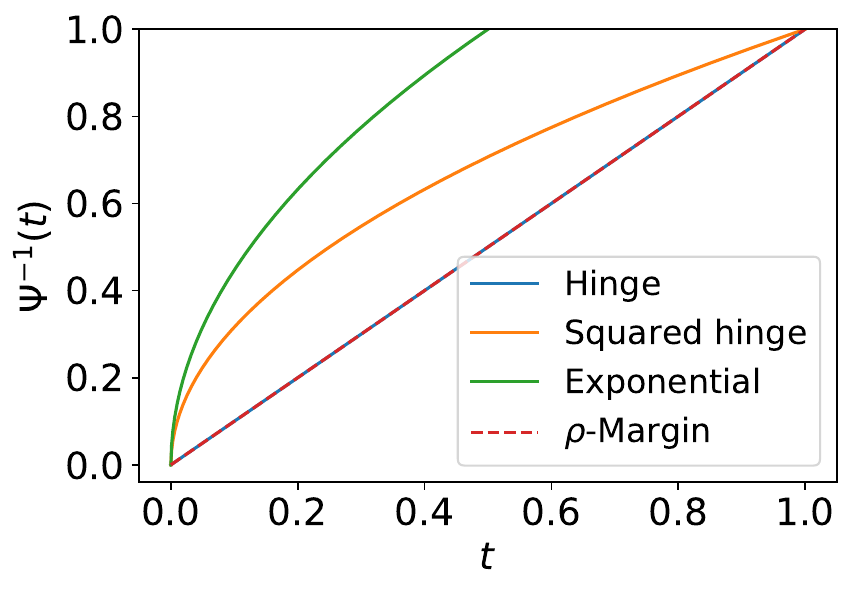}
    \caption{Left: auxiliary functions with $\rho = 0.8$. Right: $\sH$-consistency dependence between $\ell_{0-1}$ and $\Phi^{\mathrm{cstnd}}$ with $\rho = 0.8$.}
    \label{fig:auxiliary}
\end{figure}

\textbf{Multi-class loss families}. We will study three broad families
of multi-class loss functions: \emph{max losses}, \emph{sum losses}
and \emph{constrained losses}, each parameterized by an auxiliary
function $\Phi$ on $\Rset$, assumed to be non-increasing and
non-negative.
\ignore{We call a multi-class surrogate
loss $\ell$ \emph{max loss} if it is of the form in
\citep{crammer2001algorithmic}, \emph{sum loss} if it is of the form
in \citep{weston1998multi}, and \emph{constrained loss} if it is of
the form in \citep{lee2004multicategory} with rigorous definitions in
Sections~\ref{sec:max-losses}, \ref{sec:sum-losses} and
\ref{sec:constrained-losses} respectively. For each structure, we
denote by $\Phi$ a non-increasing and non-negative auxiliary function
on $\Rset$.}
In particular, we will consider the following common
auxiliary functions:
the hinge loss $\Phi_{\mathrm{hinge}}(t) = \max\curl*{0,1 - t}$, the
squared hinge loss $\Phi_{\mathrm{sq-hinge}}(t)=\max\curl*{0, 1 - t}^2$,
the exponential loss $\Phi_{\mathrm{exp}}(t)=e^{-t}$,
and the $\rho$-margin loss
$\Phi_{\rho}(t)=\min\curl*{\max\curl*{0,1 - t/\rho},1}$.
 Note that the first three auxiliary functions are convex, while the last one is not. Figure~\ref{fig:auxiliary} shows plots of these auxiliary functions.

We will say that a hypothesis set $\sH$ is \emph{symmetric} if
\ignore{individual scoring functions $h(\cdot,y)$ are independently drawn from
a family of real-valued functions, that is,} there exists a family
$\sF$ of functions $f$ mapping from $\sX$ to $\Rset$ such that
$\curl*{\bracket*{h(x,1),\ldots,h(x,c)}\colon h\in
  \sH} = \curl*{\bracket*{f_1(x),\ldots, f_c(x)}\colon f_1, \ldots, f_c\in
  \sF}$ and $\abs*{\curl*{f(x)\colon f\in \sF}}\geq 2$
for any $x\in \sX$. The hypothesis sets defined above
($\sH_{\mathrm{all}}$, $\sH_{\mathrm{lin}}$ and $\sH_{\mathrm{NN}}$)
are all symmetric. Note that for a symmetric hypothesis set $\sH$, we
have $\mathsf H(x)= \sY$.

We will say that a hypothesis set $\sH$ is \emph{complete} if the set
of scores it generates spans $\Rset$, that is, $\curl*{h(x,y)\colon
  h\in \sH} = \Rset$, for any $(x, y)\in \sX \times \sY$. The
hypothesis sets defined above, $\sH_{\mathrm{all}}$,
$\sH_{\mathrm{lin}}$ and $\sH_{\mathrm{NN}}$ with $B = +\infty$
are all complete.

\subsection{Max losses}
\label{sec:max-losses}

In this section, we discuss guarantees for \emph{max losses}, that is loss functions that can be defined by the application of an auxiliary function $\Phi$ to the margin
$\rho_h(x, y)$, as
in \citep{crammer2001algorithmic}:
\begin{align}
\label{eq:max_loss}
\forall (x, y) \in \sX \times \sY, \quad \Phi^{\mathrm{max}}(h,x,y)=\max_{y'\neq y}\Phi\paren*{h(x,y)-h(x,y')}=\Phi\paren*{\rho_h(x, y)}.
\end{align}

\textbf{i) Negative results}.  We first give negative results showing
that max losses $\Phi^{\mathrm{max}}(h,x, y)$ with convex and
non-increasing auxiliary functions $\Phi$ do not admit useful
$\sH$-consistency bounds for multi-class classification ($c > 2$). The
proof is given in Appendix~\ref{app:deferred_proofs_max}.
\begin{restatable}[\textbf{Negative results for convex $\Phi$}]{theorem}{NegativeMax}
\label{Thm:negative_max}
Assume that $c>2$. Suppose that $\Phi$ is convex and non-increasing, and $\sH$ satisfies there exist $x\in \sX$ and $h\in \sH$ such that $\abs*{\mathsf H(x)}\geq 2$ and $h(x,y)$ are equal for all $y\in \sY$. If for a non-decreasing function $f\colon\Rset_{+} \to \Rset_{+}$, the following $\sH$-consistency bound holds for any hypothesis $h\in\sH$ and any distribution $\sD$:
\begin{align}
\label{eq:bound_max_convex}
   \sR_{\ell_{0-1}}(h) - \sR_{\ell_{0-1}}^*(\sH)
    \leq f\paren*{\sR_{\Phi^{\mathrm{max}}}(h)-\sR_{\Phi^{\mathrm{max}}}^*(\sH)},
\end{align}
then, $f$ is lower bounded by $\frac12$.
 \end{restatable}
The condition on the hypothesis set in Theorem~\ref{Thm:negative_max} is very general and all symmetric hypothesis sets verify the condition, e.g. $\sH_{\mathrm{all}}$, $\sH_{\mathrm{lin}}$ and $\sH_{\mathrm{NN}}$. 
It is also worth pointing out that when $c=2$, that is, in binary classification, Theorem~\ref{Thm:negative_max} does not hold. Indeed, \citet{awasthi2022Hconsistency} present a series of results providing $\sH$-consistency bounds for convex $\Phi$ in the binary case.
In the proof, we make use of the assumption that $c>2$ and thus are able to take a probability vector $p(x)$ whose dimension is at least three, which is crucial for the proof.

\textbf{ii) Positive results without distributional assumptions}.  On the
positive side, the max loss with the non-convex auxiliary function
$\Phi = \Phi_{\rho}$ admits $\sH$-consistency bounds.

\begin{restatable}[\textbf{$\sH$-consistency bound of $\Phi_{\rho}^{\mathrm{max}}$}]
  {theorem}{BoundMaxRho}
\label{Thm:bound_max_rho}
Suppose that $\sH$ is symmetric. Then, for any hypothesis $h\in\sH$
and any distribution $\sD$,
\begin{align}
\label{eq:bound_max_rho}
     \sR_{\ell_{0-1}}(h)-\sR_{\ell_{0-1}}^*(\sH)\leq\frac{\sR_{\Phi_{\rho}^{\mathrm{max}}}(h)-\sR_{\Phi_{\rho}^{\mathrm{max}}}^*(\sH)+\sM_{\Phi_{\rho}^{\mathrm{max}}}(\sH)}{\min\curl*{1,\frac{\inf_{x\in \sX}\sup_{h\in\sH}\rho_h(x,\hh(x))}{\rho}}}-\sM_{\ell_{0-1}}(\sH).
\end{align}
\end{restatable}
See Appendix~\ref{app:deferred_proofs_max} for the
proof. Theorem~\ref{Thm:bound_max_rho} is very powerful since it only
requires $\sH$ to be symmetric. We can use it to derive
$\sH$-consistency bounds for $\Phi_{\rho}^{\mathrm{max}}$ with common
symmetric hypothesis sets such as $\sH_{\mathrm{all}}$,
$\sH_{\mathrm{lin}}$ and $\sH_{\mathrm{NN}}$, as summarized in
Table~\ref{table:bound-max-rho}.
The proofs with corresponding summarized
Corollaries~\ref{cor:bound_max_rho_all}, \ref{cor:bound_max_rho_lin}
and \ref{cor:bound_max_rho_NN} are included in
Appendix~\ref{app:bound_max_rho}.  In the proofs, we characterize the
term $\inf_{x\in \sX}\sup_{h\in\sH}\rho_h(x,\hh(x))$ for each
hypothesis set.

\begin{table}[t]
  \caption{$\sH$-consistency bounds for $\Phi_{\rho}^{\mathrm{max}}$ with common symmetric hypothesis sets.}
  \label{table:bound-max-rho}
  \centering
  \begin{tabular}{@{\hspace{0cm}}ll@{\hspace{0cm}}}
    \toprule
    Hypothesis set        & $\sH$-consistency bound of $\Phi_{\rho}^{\mathrm{max}}$ (Corollaries~\ref{cor:bound_max_rho_all}, \ref{cor:bound_max_rho_lin} and \ref{cor:bound_max_rho_NN}) \\
    \midrule
    $\sH_{\mathrm{all}}$  & $\sR_{\ell_{0-1}}(h)-\sR_{\ell_{0-1}}^*\paren*{\sH_{\mathrm{all}}}\leq\sR_{\Phi_{\rho}^{\mathrm{max}}}(h)-\sR_{\Phi_{\rho}^{\mathrm{max}}}^*\paren*{\sH_{\mathrm{all}}}$ \\
    $\sH_{\mathrm{lin}}$  & $\sR_{\ell_{0-1}}(h)-\sR_{\ell_{0-1}}^*\paren*{\sH_{\mathrm{lin}}}\leq\frac{\sR_{\Phi_{\rho}^{\mathrm{max}}}(h)-\sR_{\Phi_{\rho}^{\mathrm{max}}}^*\paren*{\sH_{\mathrm{lin}}}+\sM_{\Phi_{\rho}^{\mathrm{max}}}\paren*{\sH_{\mathrm{lin}}}}{\min\curl*{1,\frac{2B}{\rho}}}-\sM_{\ell_{0-1}}\paren*{\sH_{\mathrm{lin}}}$ \\
    $\sH_{\mathrm{NN}}$   & $\sR_{\ell_{0-1}}(h)-\sR_{\ell_{0-1}}^*\paren*{\sH_{\mathrm{NN}}}\leq\frac{\sR_{\Phi_{\rho}^{\mathrm{max}}}(h)-\sR_{\Phi_{\rho}^{\mathrm{max}}}^*\paren*{\sH_{\mathrm{NN}}}+\sM_{\Phi_{\rho}^{\mathrm{max}}}\paren*{\sH_{\mathrm{NN}}}}{\min\curl*{1,\frac{2\Lambda B}{\rho}}}-\sM_{\ell_{0-1}}\paren*{\sH_{\mathrm{NN}}}$ \\
    \bottomrule
  \end{tabular}
\end{table}

Note that by Theorem~\ref{Thm:negative_max}, there is no useful
$\sH$-consistency bound for the max loss with
$\Phi = \Phi_{\mathrm{hinge}}$, $\Phi_{\mathrm{sq-hinge}}$ or
$\Phi_{\mathrm{exp}}$ in these cases. However, under the realizability
assumption (Definition~\ref{def:rel}), we will show that such bounds
hold.

\textbf{iii) Positive results with realizable distributions}.  We
consider the $\sH$-realizability condition
\citep{long2013consistency,kuznetsov2014multi,
  cortes2016learning,cortes2016boosting,
  zhang2020bayes,awasthi2021calibration} which is defined as follows.
\begin{definition}[\textbf{$\sH$-realizability}]
\label{def:rel}
A distribution $\sD$ over $\sX\times\sY$ is $\sH$-realizable if it
labels points according to a deterministic model in $\sH$, i.e., if
$\exists h\in \sH$ such that $\mathbb{P}_{(x,y)\sim
  \sD}\paren*{\rho_h(x, y)>0}=1$.
\end{definition}
\begin{restatable}[\textbf{Realizable $\sH$-consistency bound of $\Phi^{\mathrm{max}}$}]
  {theorem}{BoundMaxRe}
\label{Thm:bound_max_re}
Suppose that $\sH$ is symmetric and complete, and $\Phi$ is
non-increasing and satisfies that $\lim_{t\to+\infty}\Phi(t) =
0$. Then, for any hypothesis $h\in\sH$ and any $\sH$-realizable
distribution $\sD$, we have
\begin{align}
\label{eq:bound_max_re}
\sR_{\ell_{0-1}}(h) - \sR_{\ell_{0-1}}^*(\sH)
\leq \sR_{\Phi^{\mathrm{max}}}(h) - \sR_{\Phi^{\mathrm{max}}}^*(\sH)+\sM_{\Phi^{\mathrm{max}}}(\sH).
\end{align}
\end{restatable}
See Appendix~\ref{app:deferred_proofs_max} for the
proof. \citet[Theorem~9]{long2013consistency} show that
$\Phi_{\mathrm{hinge}}^{\mathrm{max}}$ is realizable $\sH$-consistent
for any symmetric hypothesis set $\sH$ that is closed under
scaling. Since for any $\sH$-realizable distribution, the assumption
that $\sH$ is closed under scaling implies that $\sH$ is complete and
$\sM_{\Phi^{\mathrm{max}}}(\sH)=0$, Theorem~\ref{Thm:bound_max_re} also
yields a quantitative relationship in that case that is stronger
than the asymptotic consistency property of that previous work.

\subsection{Sum losses}
\label{sec:sum-losses}

In this section, we discuss guarantees for \emph{sum losses},
that is loss functions defined via a sum, as in
\citep{weston1998multi}:
\begin{align}
\label{eq:sum_loss}
\Phi^{\mathrm{sum}}(h,x,y)=\sum_{y'\neq y}\Phi\paren*{h(x,y)-h(x,y')}.
\end{align}

\textbf{i) Negative results}.  We first give a negative result showing
that when using as auxiliary function the hinge-loss, the sum loss
cannot benefit from any useful $\sH$-consistency guarantee. The proof
is deferred to Appendix~\ref{app:deferred_proofs_sum}.

\begin{restatable}[\textbf{Negative results for hinge loss}]{theorem}{NegativeSum}
\label{Thm:negative_sum}
Assume that $c>2$. Suppose that $\sH$ is symmetric and complete. If for a non-decreasing function $f\colon\Rset_{+} \to \Rset_{+}$, the following $\sH$-consistency bound holds for any hypothesis $h\in\sH$ and any distribution $\sD$:
\begin{align}
\label{eq:bound_sum_hinge}
    \sR_{\ell_{0-1}}(h) - \sR_{\ell_{0-1}}^*(\sH)
    \leq f\paren*{\sR_{\Phi_{\mathrm{hinge}}^{\mathrm{sum}}}(h)
      - \sR_{\Phi_{\mathrm{hinge}}^{\mathrm{sum}}}^*(\sH)},
\end{align}
then, $f$ is lower bounded by $\frac16$.
 \end{restatable}

\textbf{ii) Positive results}.  We then complement this negative
result with positive results when using the exponential loss, the
squared hinge-loss, and the $\rho$-margin loss, as summarized in
Table~\ref{table:bound-sum}.  The proofs with corresponding summarized
Theorems~\ref{Thm:bound_sum_sq-hinge}, \ref{Thm:bound_sum_exp} and
\ref{Thm:bound_sum_rho} are included in
Appendix~\ref{app:deferred_proofs_sum} for completeness. For
$\Phi_{\rho}^{\mathrm{sum}}$, the symmetry and completeness assumption
can be relaxed to symmetry and the condition that for any $x\in \sX$,
there exists a hypothesis $h \in \sH$ such that
$\abs*{h(x,i)-h(x,j)}\geq \rho$ for any $i\neq j \in \sY$, as shown in
Theorem~\ref{Thm:bound_sum_rho}. In the proof, we introduce an
auxiliary Lemma~\ref{lemma:sum_auxiliary} in
Appendix~\ref{app:deferred_proofs_sum_auxiliary}, which would be
helpful for lower bounding the conditional regret of
$\Phi_{\rho}^{\mathrm{sum}}$ with that of the multi-class $0/1$ loss.

\begin{table}[t]
  \caption{$\sH$-consistency bounds for sum losses with symmetric and complete hypothesis sets.}
  \label{table:bound-sum}
  \centering
  \begin{tabular}{@{\hspace{0cm}}l@{\hspace{.2cm}}l@{\hspace{0cm}}}
    \toprule
    Sum loss        & $\sH$-consistency bound (Theorems~\ref{Thm:bound_sum_sq-hinge}, \ref{Thm:bound_sum_exp} and \ref{Thm:bound_sum_rho}) \\
    \midrule
    $\Phi_{\mathrm{sq-hinge}}^{\mathrm{sum}}$   & $\sR_{\ell_{0-1}}(h)-\sR_{\ell_{0-1}}^*(\sH)\leq\paren*{\sR_{\Phi_{\mathrm{sq-hinge}}^{\mathrm{sum}}}(h)-\sR_{\Phi_{\mathrm{sq-hinge}}^{\mathrm{sum}}}^*(\sH)+\sM_{\Phi_{\mathrm{sq-hinge}}^{\mathrm{sum}}}(\sH)}^{\frac12}-\sM_{\ell_{0-1}}(\sH)$ \\
     $\Phi_{\mathrm{exp}}^{\mathrm{sum}}$  & $\sR_{\ell_{0-1}}(h)-\sR_{\ell_{0-1}}^*(\sH)\leq \sqrt{2}\paren*{\sR_{\Phi_{\mathrm{exp}}^{\mathrm{sum}}}(h)-\sR_{\Phi_{\mathrm{exp}}^{\mathrm{sum}}}^*(\sH)+\sM_{\Phi_{\mathrm{exp}}^{\mathrm{sum}}}(\sH)}^{\frac12}-\sM_{\ell_{0-1}}(\sH)$ \\
    $\Phi_{\rho}^{\mathrm{sum}}$   & $\sR_{\ell_{0-1}}(h)- \sR_{\ell_{0-1}}^*(\sH) \leq   \sR_{\Phi_{\rho}^{\mathrm{sum}}}(h)-\sR_{\Phi_{\rho}^{\mathrm{sum}}}^*(\sH)+\sM_{\Phi_{\rho}^{\mathrm{sum}}}(\sH)-\sM_{\ell_{0-1}}(\sH)$ \\
    \bottomrule
  \end{tabular}
\end{table}

\subsection{Constrained losses}
\label{sec:constrained-losses}

In this section, we discuss guarantees for \emph{constrained loss},
that is loss functions defined via a constraint, as in
\citep{lee2004multicategory}:
\begin{align}
\label{eq:lee_loss}
\Phi^{\mathrm{cstnd}}(h, x, y) = \sum_{y'\neq y}\Phi\paren*{-h(x, y')}
\end{align}
with the constraint that $\sum_{y\in \sY}h(x,y)=0$. We present a
series of positive results by proving multi-class $\sH$-consistency
bounds when using as an auxiliary function the hinge-loss, the squared
hinge-loss, the exponential loss, and the $\rho$-margin loss, as
summarized in Table~\ref{table:bound-constrained}. As with the binary
case \citep{awasthi2022Hconsistency}, the bound admits a linear
dependency for $\Phi_{\mathrm{hinge}}^{\mathrm{cstnd}}$ and
$\Phi_{\rho}^{\mathrm{cstnd}}$, in contrast with a square-root
dependency for $\Phi_{\mathrm{sq-hinge}}^{\mathrm{cstnd}}$ and
$\Phi_{\mathrm{exp}}^{\mathrm{cstnd}}$, as illustrated in
Figure~\ref{fig:auxiliary}. The proofs with corresponding summarized
Theorems~\ref{Thm:bound_lee_hinge}, \ref{Thm:bound_lee_sq-hinge},
\ref{Thm:bound_lee_exp} and \ref{Thm:bound_lee_rho} are included in
Appendix~\ref{app:deferred_proofs_lee} for completeness. For
$\Phi_{\rho}^{\mathrm{cstnd}}$, the symmetric and complete assumption
can be relaxed to be symmetric and satisfy that for any $x\in \sX$,
there exists a hypothesis $h \in \sH$ such that $h(x,y) \leq -\rho$
for any $y \neq y_{\max}$, as shown in
Theorem~\ref{Thm:bound_lee_rho}.

The main idea of the proofs in this section is to leverage the
constraint condition of \citet{lee2004multicategory} that the scores
sum to zero, and appropriately choose a hypothesis $\ov h$ that
differs from $h$ only by its scores for $\hh(x)$ and $y_{\max}$. We
can then upper-bound the minimal conditional risk by the conditional
risk of $\ov h$, without having to derive the closed form expression
of the minimal conditional risk.

As shown by \citet[Theorem~3.2]{steinwart2007compare}, for the family
of all measurable functions, the minimizability gaps vanish:
$\sM_{\ell_{0-1}}\paren*{\sH_{\mathrm{all}}}
= \sM_{\Phi^{\mathrm{sum}}}\paren*{\sH_{\mathrm{all}}}
= \sM_{\Phi^{\mathrm{cstnd}}}\paren*{\sH_{\mathrm{all}}}
= 0$, for $\Phi = \Phi_{\mathrm{hinge}}$, $\Phi_{\mathrm{sq-hinge}}$,
$\Phi_{\mathrm{exp}}$ and $\Phi_{\rho}$. Therefore, when
$\sH = \sH_{\mathrm{all}}$, our quantitative bounds in
Table~\ref{table:bound-sum} and Table~\ref{table:bound-constrained}
imply the asymptotic consistency results of those multi-class losses
in \citep{tewari2007consistency}, which shows that our results are
stronger and more significant. We also provide bounds for multi-class
losses using a non-convex auxiliary function, which are not studied in
the previous work.

\begin{table}[t]
  \caption{$\sH$-consistency bounds for constrained losses with symmetric and complete hypothesis sets.}
  \label{table:bound-constrained}
  \centering
  \resizebox{\textwidth}{!}{
  \begin{tabular}{@{\hspace{0cm}}l@{\hspace{.2cm}}l@{\hspace{0cm}}}
    \toprule
    Constrained loss        & $\sH$-consistency bound (Theorems~\ref{Thm:bound_lee_hinge}, \ref{Thm:bound_lee_sq-hinge}, \ref{Thm:bound_lee_exp} and \ref{Thm:bound_lee_rho}) \\
    \midrule
    $\Phi_{\mathrm{hinge}}^{\mathrm{cstnd}}$  & $\sR_{\ell_{0-1}}(h)-\sR_{\ell_{0-1}}^*(\sH)\leq\sR_{\Phi_{\mathrm{hinge}}^{\mathrm{cstnd}}}(h)-\sR_{\Phi_{\mathrm{hinge}}^{\mathrm{cstnd}}}^*(\sH)+\sM_{\Phi_{\mathrm{hinge}}^{\mathrm{cstnd}}}(\sH)-\sM_{\ell_{0-1}}(\sH)$ \\
    $\Phi_{\mathrm{sq-hinge}}^{\mathrm{cstnd}}$  & $\sR_{\ell_{0-1}}(h)-\sR_{\ell_{0-1}}^*(\sH)\leq\paren*{\sR_{\Phi_{\mathrm{sq-hinge}}^{\mathrm{cstnd}}}(h)-\sR_{\Phi_{\mathrm{sq-hinge}}^{\mathrm{cstnd}}}^*(\sH)+\sM_{\Phi_{\mathrm{sq-hinge}}^{\mathrm{cstnd}}}(\sH)}^{\frac12}-\sM_{\ell_{0-1}}(\sH)$ \\
    $\Phi_{\mathrm{exp}}^{\mathrm{cstnd}}$   & $\sR_{\ell_{0-1}}(h)-\sR_{\ell_{0-1}}^*(\sH)\leq \sqrt{2}\paren*{\sR_{\Phi_{\mathrm{exp}}^{\mathrm{cstnd}}}(h)-\sR_{\Phi_{\mathrm{exp}}^{\mathrm{cstnd}}}^*(\sH)+\sM_{\Phi_{\mathrm{exp}}^{\mathrm{cstnd}}}(\sH)}^{\frac12}-\sM_{\ell_{0-1}}(\sH)$ \\
    $\Phi_{\rho}^{\mathrm{cstnd}}$   & $\sR_{\ell_{0-1}}(h)- \sR_{\ell_{0-1}}^*(\sH) \leq   \sR_{\Phi_{\rho}^{\mathrm{cstnd}}}(h)-\sR_{\Phi_{\rho}^{\mathrm{cstnd}}}^*(\sH)+\sM_{\Phi_{\rho}^{\mathrm{cstnd}}}(\sH)-\sM_{\ell_{0-1}}(\sH)$ \\
    \bottomrule
  \end{tabular}
  }
\end{table}

\section{Adversarial \texorpdfstring{$\sH$}{H}-consistency bounds}
\label{sec:adv-mhcb}

In this section, we analyze multi-class $\sH$-consistency bounds in the
adversarial scenario ($\ell_2 = \ell_\gamma$).
\ignore{
thus is for surrogate losses $\ell_1$ of the adversarial multi-class $0/1$ loss $\ell_{\gamma}$. 
}

For any $x \in \sX$, we denote
by $\sH_\gamma(x)$
the set of hypotheses $h$ with
a positive margin on the ball of radius $\gamma$ around $x$,
$\sH_\gamma(x) = \curl*{h\in\sH:\inf_{x'\colon \norm*{x - x'}_p \leq \gamma}\rho_h(x', \hh(x)) > 0}$, and by $\mathsf H_\gamma (x)$ the set of labels generated by these hypotheses,
$\mathsf H_\gamma (x) = \curl*{\hh(x)
\colon h\in \sH_{\gamma}(x)}$.
\ignore{
For convenience, we define the sets $\sH_\gamma(x) = \curl*{h\in\sH:\inf_{x'\colon \norm*{x-x'}_p\leq
    \gamma}\rho_h(x', \hh(x)) > 0}$ and $\mathsf H_\gamma (x) = \curl*{y
  \in \sY\colon \exists h\in \sH_{\gamma} (x) \text{ such that }
  \hh(x) = y}$.}
\ignore{
By definition, we have $\sH_\gamma (x) = \emptyset$
iff $\mathsf H_\gamma (x) = \emptyset$. }
When $\sH$ is
symmetric, we have $\mathsf H_{\gamma}(x) = \sY$ iff $\sH_{\gamma}(x)\neq\emptyset$. The following lemma characterizes the
conditional $\epsilon$-regret for adversarial $0/1$ loss, which will
be helpful for applying Theorem~\ref{Thm:excess_bounds_Psi-mhcb} and Theorem~\ref{Thm:excess_bounds_Gamma-mhcb} to the adversarial scenario.
\begin{restatable}{lemma}{ExplicitAssumptionAdvMhcb}
\label{lemma:explicit_assumption_adv-mhcb}
For any $x \in \sX$, the minimal conditional
$\ell_{\gamma}$-risk and the
conditional $\epsilon$-regret for $\ell_{\gamma}$ can be expressed as follows:
\begin{align*}
\sC^*_{\ell_{\gamma}}(\sH, x) & = 1 - \max_{y\in \mathsf H_{\gamma}(x)} \sfp(y \!\mid\! x) \mathds{1}_{\sH_{\gamma}(x)\neq\emptyset}\\
\bracket*{\Delta\sC_{\ell_{\gamma},\sH}(h,x)}_{\e} & =
\begin{cases}
\bracket*{\max_{y\in\mathsf H_{\gamma}(x)} \sfp(y \!\mid\! x) - \sfp(\hh(x) \!\mid\! x)\mathds{1}_{h\in \sH_{\gamma}(x)}}_{\e} & \text{if }\sH_{\gamma}(x)\neq \emptyset \\
0 & \text{otherwise.}
\end{cases}
\end{align*}
\ignore{
In particular, for a symmetric $\sH$, the minimal conditional $\ell_{\gamma}$-risk can be expressed as
\begin{align*}
\sC^*_{\ell_{\gamma}}(\sH, x)=1-\max_{y\in \sY} \sfp(y \!\mid\! x) \mathds{1}_{\sH_{\gamma}(x)\neq\emptyset}.
\end{align*}
The conditional $\epsilon$-regret for $\ell_{\gamma}$ can be characterized as
\begin{align*}
\bracket*{\Delta\sC_{\ell_{\gamma},\sH}(h,x)}_{\e}=
\begin{cases}
\bracket*{\max_{y\in \sY} \sfp(y \!\mid\! x)-\sfp(\hh(x) \!\mid\! x)\mathds{1}_{h\in \sH_{\gamma}(x)}}_{\e} & \sH_{\gamma}(x)\neq \emptyset \\
0 & \text{otherwise.}
\end{cases}
\end{align*}}
\end{restatable}
The proof of Lemma~\ref{lemma:explicit_assumption_adv-mhcb} is presented in Appendix~\ref{app:deferred_proofs_cond}. By Lemma~\ref{lemma:explicit_assumption_adv-mhcb}, Theorems~\ref{Thm:excess_bounds_Psi-mhcb} and \ref{Thm:excess_bounds_Gamma-mhcb} can be instantiated as
Theorems~\ref{Thm:excess_bounds_Psi_adv_general} and \ref{Thm:excess_bounds_Gamma_adv_general} in the adversarial scenario as follows, where $\sH$-consistency bounds are provided between the adversarial multi-class $0/1$ loss and a surrogate loss $\ell$.

\begin{theorem}[\textbf{Adversarial distribution-dependent $\Psi$-bound}]
\label{Thm:excess_bounds_Psi_adv_general}
Assume that there exists a convex function
$\Psi\colon \mathbb{R_{+}} \to \Rset$ with $\Psi(0)=0$ and
$\epsilon\geq0$ such that the following holds for all $h\in \sH$, $x\in \curl*{x\in\sX:\sH_{\gamma}(x)\neq \emptyset}$ and $\sD\in \sP$:
\begin{align}
\label{eq:condition_Psi_general_adv-mhcb}
    \Psi\paren*{\bracket*{\max_{y\in \mathsf H_{\gamma}(x)} \sfp(y \!\mid\! x)-\sfp(\hh(x) \!\mid\! x)\mathds{1}_{h\in \sH_{\gamma}(x)}}_{\e}}
    \leq \Delta\sC_{\ell,\sH}(h,x).
\end{align}
Then, for any hypothesis $h \in \sH$ and any distribution $\sD\in \sP$, we have
\begin{align}
\label{eq:bound_Psi_adv}
     \Psi\paren*{\sR_{\ell_{\gamma}}(h)- \sR_{\ell_{\gamma}}^*(\sH)+\sM_{\ell_{\gamma}}(\sH)}
     \leq  \sR_{\ell}(h)-\sR_{\ell}^*(\sH)+\sM_{\ell}(\sH)+\max\curl*{0,\Psi(\e)}.
\end{align}
\end{theorem}

\begin{theorem}[\textbf{Adversarial distribution-dependent $\Gamma$-bound}]
\label{Thm:excess_bounds_Gamma_adv_general}
Assume that there exists a non-negative concave function $\Gamma\colon \mathbb{R_{+}}\to \Rset$
and $\epsilon\geq0$ such that the following holds for all $h\in \sH$, $x\in \curl*{x\in\sX:\sH_{\gamma}(x)\neq \emptyset}$ and $\sD\in \sP$:
\begin{align}
\label{eq:condition_Gamma_general_adv-mhcb}
    \bracket*{\max_{y\in \mathsf H_{\gamma}(x)} \sfp(y \!\mid\! x)-\sfp(\hh(x) \!\mid\! x)\mathds{1}_{h\in \sH_{\gamma}(x)}}_{\e}
    \leq \Gamma\paren*{ \Delta\sC_{\ell,\sH}(h,x)}.
\end{align}
Then, for any hypothesis $h\in\sH$ and any distribution $\sD\in \sP$, we have
\begin{align}
\label{eq:bound_Gamma_adv}
     \sR_{\ell_{\gamma}}(h)- \sR_{\ell_{\gamma}}^*(\sH)
     \leq  \Gamma\paren*{\sR_{\ell}(h)-\sR_{\ell}^*(\sH)+\sM_{\ell}(\sH)}
     -\sM_{\ell_{\gamma}}(\sH)+\epsilon.
\end{align}
\end{theorem}
Next, we will apply
Theorem~\ref{Thm:excess_bounds_Psi_adv_general} and Theorem~\ref{Thm:excess_bounds_Gamma_adv_general} to study various
hypothesis sets and adversarial surrogate loss functions in
Sections~\ref{sec:neg-adv} for negative results and
Section~\ref{sec:max-adv}, \ref{sec:sum-adv}, and \ref{sec:con-adv} for positive results. A
careful analysis is presented in each case (see
Appendix~\ref{app:deferred_proofs_adv_negative}, \ref{app:deferred_proofs_adv_max}, \ref{app:deferred_proofs_adv_sum} and \ref{app:deferred_proofs_adv_lee}).

\subsection{Negative results for adversarial robustness}
\label{sec:neg-adv}
The following result rules out the $\sH$-consistency guarantee of multi-class losses with a convex auxiliary function, which are commonly used in practice. The proof is given in Appendix~\ref{app:deferred_proofs_adv_negative}.
\begin{restatable}[\textbf{Negative results for convex functions}]{theorem}{NegativeAdv}
\label{Thm:negative_adv}
Fix $c = 2$. Suppose that $\Phi$ is convex and non-increasing, and $\sH$ contains $0$ and satisfies the condition that there exists $x\in \sX$ such that $\sH_{\gamma}(x)\neq \emptyset$. If for a non-decreasing function $f\colon\Rset_{+} \to \Rset_{+}$, the following $\sH$-consistency bound holds for any hypothesis $h\in\sH$ and any distribution $\sD$:
\begin{align}
\label{eq:bound_convex_adv}
   \sR_{\ell_{\gamma}}(h) - \sR_{\ell_{\gamma}}^*(\sH)
    \leq f\paren*{\sR_{\wt \ell}(h)-\sR_{\wt \ell}^*(\sH)},
\end{align}
then, $f$ is lower bounded by $\frac12$, for $\wt \ell=\wt \Phi^{\mathrm{max}}$, $\wt \Phi^{\mathrm{sum}}$ and $\wt \Phi^{\mathrm{cstnd}}$.
\end{restatable}
Instead, we show in Sections~\ref{sec:max-adv}, \ref{sec:sum-adv}, and \ref{sec:con-adv} that the max, sum and constrained losses using as auxiliary function the non-convex $\rho$-margin loss admit favorable $\sH$-consistency bounds in the multi-class setting, thereby significantly generalizing the binary counterpart in \citep{awasthi2022Hconsistency}.

\subsection{Adversarial max losses}
\label{sec:max-adv}
We first consider the adversarial max loss $\wt \Phi^{\mathrm{max}}$ defined as the supremum based counterpart of \eqref{eq:max_loss}:
\begin{align}
\label{eq:max_loss_adv}
\wt\Phi^{\mathrm{max}}(h,x,y)=\sup_{x':\norm*{x-x'}_p\leq \gamma}\Phi\paren*{\rho_h(x', y)}.
\end{align}
For the adversarial max loss with $\Phi=\Phi_{\rho}$, we can obtain $\sH$-consistency  bounds as follows.
\begin{restatable}[\textbf{$\sH$-consistency bound of $\wt{\Phi}_{\rho}^{\mathrm{max}}$}]
  {theorem}{BoundMaxRhoAdv}
\label{Thm:bound_max_rho_adv}
Suppose that $\sH$ is symmetric. Then, for any hypothesis $h\in\sH$
and any distribution $\sD$, we have
\begin{align}
\label{eq:bound_max_rho_adv}
     \sR_{\ell_{\gamma}}(h)- \sR_{\ell_{\gamma}}^*(\sH) \leq \frac{  \sR_{\wt{\Phi}_{\rho}^{\mathrm{max}}}(h)-\sR_{\wt{\Phi}_{\rho}^{\mathrm{max}}}^*(\sH)+\sM_{\wt{\Phi}_{\rho}^{\mathrm{max}}}(\sH)}{\min\curl*{1,\frac{\inf_{x\in \curl*{x\in\sX:\sH_{\gamma}(x)\neq \emptyset}}\sup_{h\in \sH_{\gamma}(x)}\inf_{x':\norm*{x-x'}_p\leq\gamma}\rho_h(x',\hh(x))}{\rho}}}-\sM_{\ell_{\gamma}}(\sH).
\end{align}
\end{restatable}

\subsection{Adversarial sum losses}
\label{sec:sum-adv}
Next, we consider the adversarial sum loss $\wt \Phi^{\mathrm{sum}}$ defined as the supremum based counterpart of \eqref{eq:sum_loss}:
\begin{align}
\label{eq:sum_loss_adv}
\wt\Phi^{\mathrm{sum}}(h,x,y)=\sup_{x':\norm*{x-x'}_p\leq \gamma}\sum_{y'\neq y}\Phi\paren*{h(x',y)-h(x',y')}.
\end{align}
Using the auxiliary Lemma~\ref{lemma:sum_auxiliary} in Appendix~\ref{app:deferred_proofs_sum_auxiliary}, we can obtain the $\sH$-consistency bound of $\wt{\Phi}_{\rho}^{\mathrm{sum}}$.
\begin{restatable}[\textbf{$\sH$-consistency bound of $\wt{\Phi}_{\rho}^{\mathrm{sum}}$}]
  {theorem}{BoundSumRhoAdv}
\label{Thm:bound_sum_rho_adv}
Assume that $\sH$ is symmetric and that for any $x\in \sX$,
there exists a hypothesis $h \in \sH$
inducing the same ordering of the labels for any $x'\in \curl*{x'\colon \norm*{x - x'}_p\leq \gamma}$
and such that
$\inf_{x'\colon \norm*{x - x'}_p\leq \gamma}\abs*{h(x', i) - h(x', j)}\geq \rho$ for
any $i\neq j \in \sY$. Then, for any
hypothesis $h\in\sH$ and any distribution $\sD$, the following inequality holds:
\begin{align}
\label{eq:bound_sum_rho_adv}
     \sR_{\ell_{\gamma}}(h)- \sR_{\ell_{\gamma}}^*(\sH) \leq   \sR_{\wt{\Phi}_{\rho}^{\mathrm{sum}}}(h)-\sR_{\wt{\Phi}_{\rho}^{\mathrm{sum}}}^*(\sH)+\sM_{\wt{\Phi}_{\rho}^{\mathrm{sum}}}(\sH)-\sM_{\ell_{\gamma}}(\sH).
\end{align}
\end{restatable}

\subsection{Adversarial constrained loss}
\label{sec:con-adv}
Similarly, we define the adversarial constrained loss $\wt \Phi^{\mathrm{cstnd}}$ as supremum based counterpart of \eqref{eq:lee_loss}:
\begin{align}
\label{eq:lee_loss_adv}
\wt\Phi^{\mathrm{cstnd}}(h,x,y)=\sup_{x':\norm*{x-x'}_p\leq \gamma}\sum_{y'\neq y}\Phi\paren*{-h(x',y')}
\end{align}
with the constraint that $\sum_{y\in \sY}h(x,y)=0$. For the adversarial constrained loss with $\Phi=\Phi_{\rho}$, we can obtain the $\sH$-consistency bound of $\wt{\Phi}_{\rho}^{\mathrm{cstnd}}$ as follows.
\begin{restatable}[\textbf{$\sH$-consistency bound of $\wt{\Phi}_{\rho}^{\mathrm{cstnd}}$}]
  {theorem}{BoundLeeRhoAdv}
\label{Thm:bound_lee_rho_adv}
Suppose that $\sH$ is symmetric and satisfies that for any $x\in \sX$, there exists a hypothesis $h \in \sH$ with the constraint $\sum_{y\in \sY}h(x,y)=0$ such that $\sup_{x':\norm*{x-x'}_p\leq \gamma}h(x',y) \leq -\rho$ for any $y \neq y_{\max}$. Then, for any hypothesis $h\in\sH$ and any distribution,
\begin{align}
\label{eq:bound_lee_rho_adv}
     \sR_{\ell_{\gamma}}(h)- \sR_{\ell_{\gamma}}^*(\sH) \leq   \sR_{\wt{\Phi}_{\rho}^{\mathrm{cstnd}}}(h)-\sR_{\wt{\Phi}_{\rho}^{\mathrm{cstnd}}}^*(\sH)+\sM_{\wt{\Phi}_{\rho}^{\mathrm{cstnd}}}(\sH)-\sM_{\ell_{\gamma}}(\sH).
\end{align}
\end{restatable}
The proofs of Theorems~\ref{Thm:bound_max_rho_adv}, \ref{Thm:bound_sum_rho_adv} and \ref{Thm:bound_lee_rho_adv} are included in Appendix~\ref{app:deferred_proofs_adv_max}, \ref{app:deferred_proofs_adv_sum} and \ref{app:deferred_proofs_adv_lee} respectively. These results
are significant since they apply to general hypothesis sets.
In particular, symmetric hypothesis sets $\sH_{\mathrm{all}}$, $\sH_{\mathrm{lin}}$ and $\sH_{\mathrm{NN}}$ with $B=+ \infty$ all verify the conditions of those theorems. When $B<+ \infty$, the conditions in Theorems~\ref{Thm:bound_sum_rho_adv} and \ref{Thm:bound_lee_rho_adv} can still be verified with a suitable
choice of $\rho$, where we can consider the hypotheses such that $w_y=0$ in $\sH_{\mathrm{lin}}$ and $\sH_{\mathrm{NN}}$, while Theorem~\ref{Thm:bound_max_rho_adv} holds for any $\rho>0$.

\section{Conclusion}

We presented a comprehensive study of $\sH$-consistency bounds for
multi-class classification, including the analysis of the three most
commonly used families of multi-class surrogate losses (max losses,
sum losses and constrained losses) and including the study of
surrogate losses for the adversarial robustness. Our theoretical
analysis helps determine which surrogate losses admit a favorable
guarantee for a given hypothesis set $\sH$. Our bounds can help guide
the design of multi-class classification algorithms for both the
adversarial and non-adversarial settings. They also help compare
different surrogate losses for the same setting and the same hypothesis
set. Of course, in addition to the functional form of the
$\sH$-consistency bound, the approximation property of a surrogate
loss function combined with the hypothesis set plays an important
role.

\chapter{Multi-Class Classification: Comp-Sum Losses} \label{ch4}
In this chapter, we present the first $\sH$-consistency bounds for the
logistic loss, which can be used to derive directly guarantees for
current algorithms used in the machine learning community.  More
generally, we will consider a broader family of loss functions that we
refer to as \emph{comp-sum losses}, that is loss functions obtained by
composition of a concave function, such as logarithm in the case of
the logistic loss, with a sum of functions of differences of score,
such as the negative exponential. We prove $\sH$-consistency bounds
for a wide family of comp-sum losses, which includes as special cases
the logistic loss
\citep{Verhulst1838,Verhulst1845,Berkson1944,Berkson1951}, the
\emph{generalized cross-entropy loss} \citep{zhang2018generalized},
and the \emph{mean absolute error loss} \citep{ghosh2017robust}.
We further show that our bounds are \emph{tight} and thus cannot be
improved.

$\sH$-consistency bounds are expressed in terms of a quantity called
\emph{minimizability gap}, which only depends on the loss function and
the hypothesis set $\sH$ used. It is the difference of the best-in
class expected loss and the expected pointwise infimum of the loss.
For the loss functions we consider, the minimizability gap vanishes
when $\sH$ is the full family of measurable functions. However, in
general, the gap is non-zero and plays an important role, depending on
the property of the loss function and the hypothesis set. Thus, to
better understand $\sH$-consistency bounds for comp-sum losses, we
specifically analyze their minimizability gaps, which we
use to compare their guarantees.

A recent challenge in the application of neural networks is their
robustness to imperceptible perturbations
\citep{szegedy2013intriguing}. While neural networks trained on large
datasets often achieve a remarkable performance
\citep{SutskeverVinyalsLe2014,KrizhevskySutskeverHinton2012}, their
accuracy remains substantially lower in the presence of such
perturbations. One key issue in this scenario is the definition of a
useful surrogate loss for the adversarial loss. To tackle this
problem, we introduce a family of loss functions designed for
adversarial robustness that we call \emph{smooth adversarial comp-sum
loss functions}. These are loss functions derived from their comp-sum
counterparts by augmenting them with a natural smooth term.  We show
that these loss functions are beneficial in the adversarial setting by
proving that they admit $\sH$-consistency bounds. This leads to a
family of algorithms for adversarial robustness that consist of
minimizing a regularized smooth adversarial comp-sum loss.

While our main purpose is a theoretical analysis, we also present an
extensive empirical analysis. We compare the empirical performance of
comp-sum losses for different tasks and relate that to their
theoretical properties. We further report the results of experiments
with the CIFAR-10, CIFAR-100 and SVHN datasets comparing the
performance of our algorithms based on smooth adversarial comp-sum
losses with that of the state-of-the-art algorithm for this task
\textsc{trades} \citep{zhang2019theoretically}. The results show that
our adversarial algorithms outperform \textsc{trades} and also achieve
a substantially better non-adversarial (clean) accuracy.

The rest of this paper is organized as follows. In
Section~\ref{sec:preliminaries-comp}, we introduce some basic concepts and
definitions related to comp-sum loss functions.  In
Section~\ref{sec:H-consistency-bounds}, we present our
$\sH$-consistency bounds for comp-sum losses.  We further carefully
compare their minimizability gaps in Section~\ref{sec:comparison}. In
Section~\ref{sec:adversarial}, we define and motivate our smooth
adversarial comp-sum losses, for which we prove $\sH$-consistency
bounds, and briefly discuss corresponding adversarial algorithms. In
Section~\ref{sec:experiments}, we report the results of our
experiments both to compare comp-sum losses in several tasks, and to
compare the performance of our algorithms based on smooth adversarial
comp-sum losses. In Section~\ref{sec:future-work}, we discuss avenues
for future research.

The presentation in this chapter is based on \citep{mao2023cross}.

\section{Preliminaries}
\label{sec:preliminaries-comp}

We consider the familiar multi-class classification
setting and denote by $\sX$ the input space,
by $\sY = [n] = \set{1, \ldots, n}$ the set of 
classes or categories ($n \geq 2$) and by $\sD$ a distribution
over $\sX \times \sY$.

We study general loss functions $\ell\colon \sH_{\rm{all}} \times \sX
\times \sY \to \Rset$ where $\sH_{\rm{all}}$ is the family of all
measurable functions $h\colon \sX \times \sY \to \Rset$. In
particular, the zero-one classification loss is defined, for all $h
\in \sH_{\rm{all}}$, $x \in \sX$ and $y \in \sY$, by $\ell_{0-1}(h, x,
y) = 1_{\hh(x) \neq y}$, where $\hh(x) = \argmax_{y \in \sY}h(x, y)$
with an arbitrary but fixed deterministic strategy used for breaking
the ties. For simplicity, we fix that strategy to be
the one selecting the label with the highest index under the natural
ordering of labels.

We denote by $\sR_\ell(h)$ the generalization error or expected loss
of a hypothesis $h \colon \sX \times \sY \to \Rset$: $\sR_\ell(h) =
\E_{(x, y) \sim \sD}[\ell(h, x, y)]$. For a hypothesis set $\sH
\subseteq \sH_{\rm{all}}$ of functions mapping from $\sX \times \sY$
to $\Rset$, $\sR^*_\ell(\sH)$ denotes the best-in class expected loss:
$\sR^*_\ell(\sH) = \inf_{h \in \sH} \sR_\ell(h)$.

We will prove $\sH$-consistency bounds, which are inequalities
relating the zero-one classification estimation loss $\ell_{0-1}$ of
any hypothesis $h \in \sH$ to that of its surrogate loss $\ell$
\citep*{awasthi2022Hconsistency,awasthi2022multi}. They
take the following form: $\forall h \in \sH, \sR_{\ell_{0-1}}(h) -
\sR^*_{\ell_{0-1}}(\sH) \leq f(\sR_\ell(h) - \sR^*_\ell(\sH))$, where
$f$ is a non-decreasing real-valued function. Thus, they show that the
estimation zero-one loss of $h$, $\sR_{\ell_{0-1}}(h) -
\sR^*_{\ell_{0-1}}(\sH)$, is bounded by $f(\e)$ when its surrogate
estimation loss, $\sR_\ell(h) - \sR^*_\ell(\sH)$, is bounded by
$\e$. These guarantees are thus non-asymptotic and depend on the
hypothesis set $\sH$ considered.

$\sH$-consistency bounds are expressed in terms of a quantity
depending on the hypothesis set $\sH$ and the loss function $\ell$
called \emph{minimizability gap}, and defined by $\sM_\ell(\sH) =
\sR^*_{\ell}(\sH) - \E_x\bracket[\big]{\inf_{h \in \sH} \E_y
  \bracket*{\ell(h, X, y) \mid X = x}}$. By the super-additivity of
the infimum, since \[\sR^*_{\ell}(\sH) = \inf_{h \in \sH}
\E_x\bracket[\big]{\E_y \bracket*{\ell(h, X, y) \mid X = x}},\] the
minimizability gap is always non-negative.  It measures the difference
between the best-in-class expected loss and the expected infimum of
the pointwise expected loss.  When the loss function $\ell$ only
depends on $h(x,\cdot)$ for all $h$, $x$, and $y$, that is $\ell(h, x,
y) = \Psi(h(x,1), \ldots, h(x,n), y)$, for some function $\Psi$, then
it is not hard to show that the minimizability gap vanishes for the
family of all measurable functions: $\sM(\sH_{\rm{all}}) = 0$
\citep[lemma~2.5]{steinwart2007compare}.  In general, however, the
minimizabiliy gap is non-zero for a restricted hypothesis set $\sH$
and is therefore important to analyze.  Note that the minimizabiliy
gap can be upper-bounded by the approximation error $\sA(\sH) =
\sR^*_{\ell}(\sH) - \E_x\bracket[\big]{\inf_{h \in \sH_{\rm{all}}}
  \E_y \bracket*{\ell(h, X, y) \mid X = x}}$.  It is however a finer
quantity than the approximation error and can thus lead to more
favorable guarantees.

\ignore{
  This holds for many loss functions used in practice but typically
  not for loss functions used for adversarial robustness.
}

\ignore{
— definition of comp-sum losses.
— notions of calibration function C and minimizability gaps.
— definition of H-cons bounds.
— desired H-cons bounds for comp-sum losses.
}

\begin{figure}[t]
\begin{center}
\includegraphics[scale=0.6]{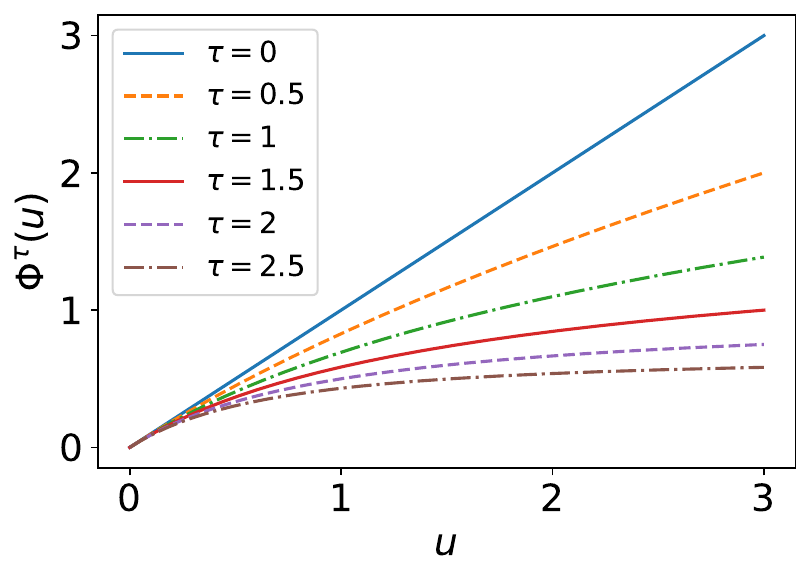}
\caption{Function $\Phi^{\tau}$ with different values of $\tau$.}
\label{fig:phi}
\end{center}
\end{figure}

\textbf{Comp-sum losses}.
In this paper, we derive guarantees for \emph{comp-sum losses}, a
family of functions including the logistic loss
that is defined via a composition of two functions $\Phi_1$ and
$\Phi_2$\ignore{, as in \citep{KuznetsovMohriSyed2014}}:
\begin{align}
\label{eq:comp-sum_loss}
\ell_{\Phi_1[\Phi_2]}^{\mathrm{comp}}(h, x, y)
= \Phi_1\paren[\bigg]{\sum_{y'\neq y}\Phi_2\paren*{h(x, y) - h(x, y')}},
\end{align}
where $\Phi_2$ is a non-increasing function upper-bounding
$\mathds{1}_{u\leq 0}$ over $u\in\Rset$ and $\Phi_1$ a non-decreasing
auxiliary function.  We will specifically consider $\Phi_2(u) =
\exp(-u)$ as with the loss function related to AdaBoost
\citep{freund1997decision} and $\Phi_1$ chosen out of the following
family of functions $\Phi^\tau$, $\tau \geq 0$, defined for all $u \geq
0$ by
\begin{equation}
\label{eq:Phi1}
\Phi^{\tau}(u) =
\begin{cases}
\frac{1}{1 - \tau} \paren*{(1 + u)^{1 - \tau} - 1} & \tau \geq 0, \tau \neq 1 \\
\log(1 + u) & \tau = 1.
\end{cases}
\end{equation}
Figure~\ref{fig:phi} shows the plot of function $\Phi^{\tau}$ for
different values of $\tau$.  Functions $\Phi^\tau$ verify the
following identities:
\begin{align}
\label{eq:Phi1-derivative}
\frac{\partial \Phi^{\tau}}{\partial u}(u)
= \frac{1}{(1 + u)^{\tau}}, \quad \Phi^{\tau}(0) = 0.
\end{align}
In view of that, by l’H\^opital's rule, $\Phi^\tau$ is continuous as a
function of $\tau$ at $\tau = 1$. To simplify the notation, we will
use $\ell_{\tau}^{\rm{comp}}$ as a short-hand for
$\ell_{\Phi_1[\Phi_2]}^{\mathrm{comp}}$ when $\Phi_1 = \Phi^{\tau}$
and $\Phi_2(u) = \exp(-u)$. $\ell_{\tau}^{\rm{comp}}(h, x, y)$ can be
expressed as follows for any $h$, $x$, $y$ and $\tau \geq 0$:
\begin{equation}
\label{eq:comp-loss}
\ell_{\tau}^{\rm{comp}}(h, x, y)
= \Phi^{\tau}\paren*{\sum_{y'\in \sY} e^{h(x, y') - h(x, y)}-1}
=
\begin{cases}
  \frac{1}{1 - \tau}
  \paren*{\bracket*{\sum_{y'\in\sY} e^{{h(x, y') - h(x, y)}}}^{1 - \tau} - 1}
  & \tau \neq 1 \\
\log\paren*{\sum_{y'\in \sY} e^{h(x, y') - h(x, y)}} & \tau = 1.
\end{cases}
\end{equation}
When $\tau = 0$, $\ell_{\tau}^{\rm{comp}}$ coincides with the
sum-exponential loss
\citep{weston1998multi,awasthi2022multi}:
\begin{align*}
  \ell_{\tau = 0}^{\mathrm{comp}}(h, x, y)
  = \sum_{y'\neq y} e^{h(x, y') - h(x, y)}.
\end{align*}
When $\tau = 1$, it coincides with the (multinomial) logistic loss
\citep{Verhulst1838,Verhulst1845,Berkson1944,Berkson1951}:
\begin{align*}
  \ell_{\tau = 1}^{\mathrm{comp}}(h, x, y)
  =- \log \bracket*{\frac{e^{h(x,y)}}{\sum_{y' \in \sY} e^{h(x,y')}}}.
\end{align*}
For $1 < \tau < 2$, it matches the
\emph{generalized cross entropy loss} \citep{zhang2018generalized}:
\begin{align*}
  \ell_{1 < \tau < 2}^{\mathrm{comp}}(h, x, y)
  = \frac{1}{\tau - 1}\bracket*{1 - \bracket*{\frac{e^{h(x,y)}}
    {\sum_{y'\in \sY} e^{h(x,y')}}}^{\tau - 1}},
\end{align*}
for $\tau = 2$, the \emph{mean absolute error loss}
\citep{ghosh2017robust}:
\begin{align*}
  \ell_{\tau = 2}^{\mathrm{comp}}(h, x, y) =
  1 - \frac{e^{h(x,y)}}{\sum_{y'\in \sY} e^{h(x, y')}}.
\end{align*}
Since for any $\tau\geq0$, $\frac{\partial \Phi^{\tau}}{\partial u}$
is non-increasing and satisfies $\frac{\partial \Phi^{\tau}}{\partial
  u}(0)=1$, $\Phi^{\tau}(0) = 0$, for any $\tau \geq 0$,
$\Phi^{\tau}$ is concave, non-decreasing, differentiable,
$1$-Lipschitz, and satisfies that
\begin{equation}
\label{eq:Phi1-property}
\forall u \geq 0,~\Phi^{\tau} (u) \leq u.
\end{equation}

\section{\texorpdfstring{$\sH$}{H}-Consistency Bounds for Comp-Sum Losses}
\label{sec:H-consistency-bounds}

In this section, we present and discuss $\sH$-consistency bounds for
comp-sum losses in the standard multi-class classification
scenario. We say that a hypothesis set is \emph{symmetric} when it
does not depend on a specific ordering of the classes, that is,
there exists a family $\sF$ of functions $f$ mapping from $\sX$ to
$\Rset$ such that $\curl*{\bracket*{h(x, 1),\ldots,h(x, c)}\colon h\in
  \sH} = \curl*{\bracket*{f_1(x),\ldots, f_c(x)}\colon f_1, \ldots,
  f_c\in \sF}$, for any $x \in \sX$. We say that a hypothesis set
$\sH$ is \emph{complete} if the set of scores it generates spans
$\Rset$, that is, $\curl*{h(x, y)\colon h\in \sH} = \Rset$, for any
$(x, y)\in \sX \times \sY$. The hypothesis sets widely used in
practice are all symmetric and complete.

\subsection{\texorpdfstring{$\sH$}{H}-Consistency Guarantees}

The following holds for all comp-sum loss functions and
all symmetric and complete hypothesis sets, which includes
those typically considered in applications.

\begin{figure}[t]
\begin{center}
\includegraphics[scale=0.25]{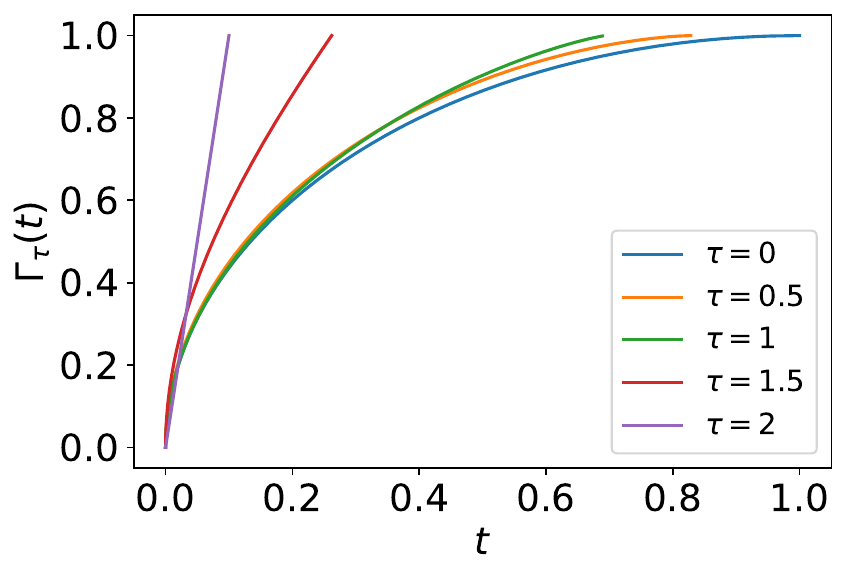}
\caption{Function $\Gamma_{\tau}$ with different values of $\tau$ for $n = 10$.}
\label{fig:Gamma}
\end{center}
\end{figure}

\begin{restatable}[\textbf{$\sH$-consistency bounds for comp-sum losses}]
  {theorem}{BoundCompSum}
\label{Thm:bound_comp_sum}
Assume that $\sH$ is symmetric and complete. Then, for any $\tau\in
[0,\infty)$ and any $h \in \sH$, the following inequality holds:
\begin{align*}
\sR_{\ell_{0-1}}(h)-\sR_{\ell_{0-1}}^*(\sH)
\leq \Gamma_{\tau}
  \paren*{\sR_{\ell_{\tau}^{\rm{comp}}}(h) - \sR_{\ell_{\tau}^{\rm{comp}}}^*(\sH)
    + \sM_{\ell_{\tau}^{\rm{comp}}}(\sH)}
- \sM_{\ell_{0-1}}(\sH),
\end{align*}
where $\Gamma_{\tau}(t)= \sT_{\tau}^{-1}(t)$ is the inverse of
$\sH$-consistency comp-sum transformation, defined for all $\beta \in
    [0,1]$ by
$\sT_{\tau}(\beta)
= \begin{cases}
\frac{2^{1-\tau}}{1-\tau}\bracket*{1 -\bracket*{\frac{\paren*{1 + \beta}^{\frac1{2 - \tau }} +  \paren*{1 - \beta}^{\frac1{2 - \tau }}}{2}}^{2 - \tau }} & \tau \in [0,1)\\
\frac{1+\beta}{2}\log\bracket*{1+\beta} + \frac{1-\beta}{2}\log\bracket*{1-\beta} & \tau =1 \\
\frac{1}{(\tau-1)n^{\tau-1}}\bracket*{\bracket*{\frac{\paren*{1 + \beta}^{\frac1{2 - \tau }} +  \paren*{1 - \beta}^{\frac1{2 - \tau }}}{2}}^{2 - \tau } -1} & \tau \in (1,2)\\
\frac{1}{(\tau-1)n^{\tau-1}}\,
\beta & \tau \in [2,+ \infty).
\end{cases}$ 
\end{restatable}
By l’H\^opital's rule, $\sT_{\tau}$ is continuous as a function of
$\tau$ at $\tau = 1$. Since $\lim_{x\to
  0^{+}}\paren*{a^{\frac1x}+b^{\frac1x}}^x= \max\curl*{a,b}$,
$\sT_{\tau}$ is continuous as a function of $\tau$ at $\tau =
2$. Furthermore, for any $\tau\in [0,+\infty)$, $\sT_{\tau}$ is a
  convex and increasing function, and satisfies that
  $\sT_{\tau}(0)=0$. Note that for the sum-exponential loss ($\tau=0$)
  and logistic loss ($\tau=1$), the expression $\sT_{\tau}$ matches
  that of their binary $\sH$-consistency estimation error
  transformation $1-\sqrt{1-t^2}$ and
  $\frac{1+t}{2}\log(1+t)+\frac{1-t}{2}\log(1-t)$ in the binary
  classification setting \cite{awasthi2022Hconsistency}, which were
  proven to be tight. We will show that, for these loss functions and
  in this multi-class classification setting, $\sT_{\tau}$s admit a
  tight functional forms as well. We illustrate the function
  $\Gamma_{\tau}$ with different values of $\tau$ in
  Figure~\ref{fig:Gamma}.

\begin{figure*}[t]
\begin{center}
\includegraphics[scale=0.25]{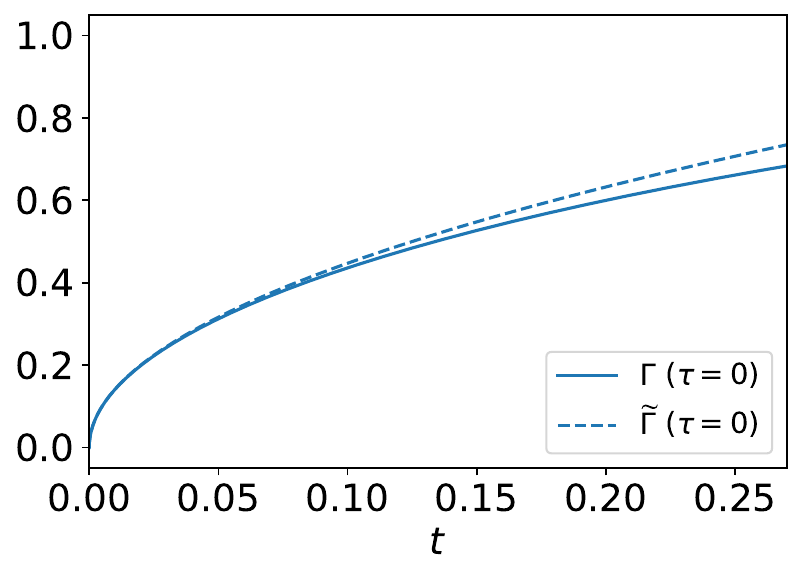}
\includegraphics[scale=0.25]{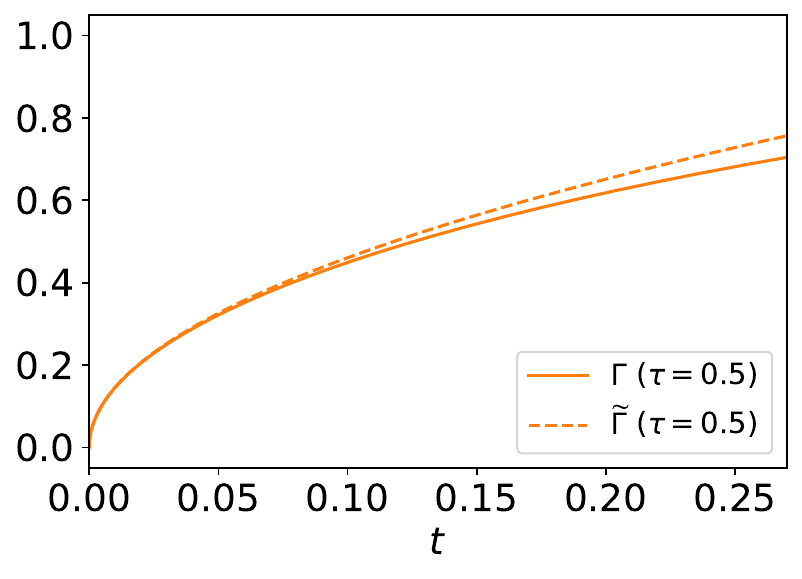}
\includegraphics[scale=0.25]{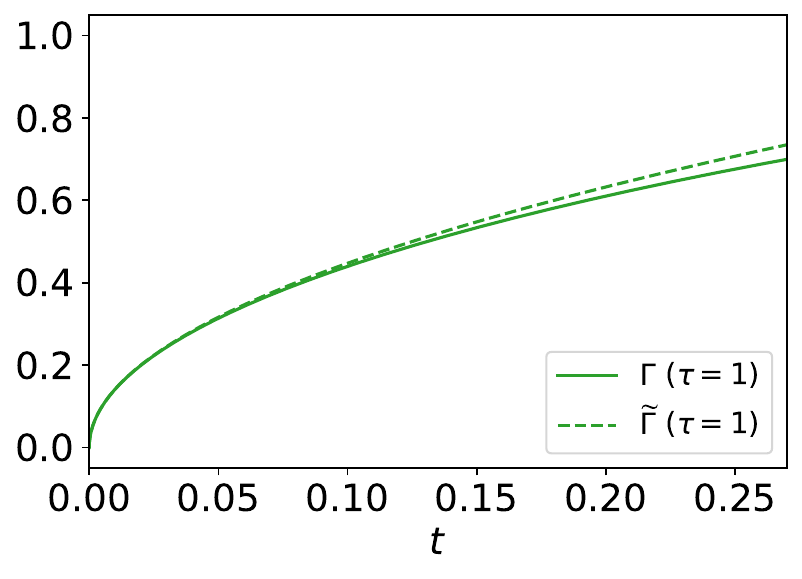}
\includegraphics[scale=0.25]{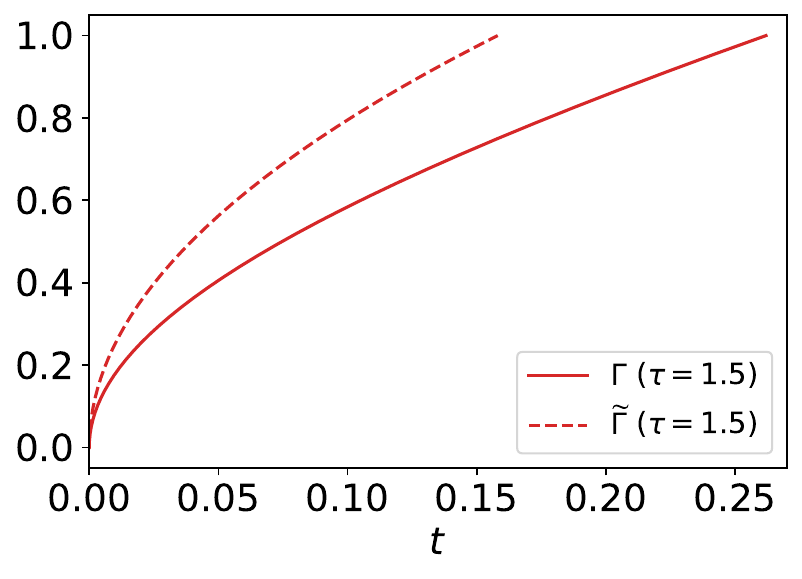}
\caption{Function $\Gamma_{\tau}$ and its upper bound $\wt\Gamma_{\tau}$ with different values of $\tau$ and $n = 10$.}
\label{fig:Gamma-wt}
\end{center}
\end{figure*}

By using Taylor expansion, $\sT_{\tau}(\beta)$ can be lower bounded by
its polynomial approximation with the tightest order as
\begin{equation}
\label{eq:wt-sT}
\sT_{\tau}(\beta)\geq \wt\sT_{\tau}(\beta)=
\begin{cases}
\frac{\beta^2}{2^{\tau}(2-\tau)} & \tau \in [0,1)\\
\frac{\beta^2}{2n^{\tau-1}} & \tau \in [1,2)\\
\frac{\beta}{(\tau-1)n^{\tau-1}}& \tau \in [2,+ \infty).
\end{cases}
\end{equation}
Accordingly, $\Gamma_{\tau}(t)$ can be upper-bounded by the inverse of
$\wt\sT_{\tau}$, which is denoted by $\wt \Gamma_{\tau}(t)=
\wt\sT_{\tau}^{-1}(t)$, as shown below
\begin{equation}
\label{eq:wt-Gamma}
\Gamma_{\tau}(t)\leq  \wt \Gamma_{\tau}(t)= \begin{cases}
\sqrt{2^{\tau}(2-\tau) t} & \tau\in [0,1)\\
\sqrt{2n^{\tau-1} t } & \tau\in [1,2) \\
(\tau - 1) n^{\tau - 1} t & \tau \in [2,+ \infty).
\end{cases}
\end{equation}
A detailed derivation is given in
Appendix~\ref{app:Gamma-upper-bound}. The plots of function
$\Gamma_{\tau}$ and their corresponding upper bound $\wt\Gamma_{\tau}$
($n=10$) are shown in Figure~\ref{fig:Gamma-wt}, for different values
of $\tau$; they illustrate the quality of the approximations via
$\wt\Gamma_{\tau}$.

Recall that the minimizability gaps vanish when $\sH$ is the family of
all measurable functions or when $\sH$ contains the Bayes predictor.
In their absence, the theorem
shows that if the estimation loss
$(\sR_{\ell_{\tau}^{\rm{comp}}}(h) -
\sR_{\ell_{\tau}^{\rm{comp}}}^*(\sH))$ is reduced to $\e$, then, for
$\tau \in [0, 2)$, in particular for the logistic loss ($\tau = 1$)
  and the generalized cross-entropy loss ($\tau \in (1, 2)$), modulo a
  multiplicative constant, the zero-one estimation loss
  $(\sR_{\ell_{0-1}}(h) - \sR_{\ell_{0-1}}^*(\sH))$ is bounded by
  $\sqrt{\e}$. For the logistic loss, the following 
  guarantee holds for all $h \in \sH$:
\[
\sR_{\ell_{0-1}}(h) - \sR_{\ell_{0-1}}^*(\sH)
\leq 
  \sqrt{2 \paren[\big]{\sR_{\ell_{1}^{\rm{comp}}}(h) - \sR_{\ell_{1}^{\rm{comp}}}^*(\sH)}}.
\]
The bound is even more favorable for the mean absolute error loss
($\tau = 2$) or for comp-sum losses $\ell_{\tau}^{\rm{comp}}$ with
$\tau \in (2, +\infty)$ since in that case, modulo a multiplicative
constant, the zero-one estimation loss $(\sR_{\ell_{0-1}}(h) -
\sR_{\ell_{0-1}}^*(\sH))$ is bounded by $\e$. In general, the
minimizability gaps are not null however and, in addition to the
functional form of $\Gamma_\tau$, two other key features help compare
comp-sum losses: (i) the magnitude of the minimizability gap
$\sM_{\ell_{\tau}^{\rm{comp}}}(\sH)$; and (ii) the dependency of the
multiplicative constant on the number of classes, which makes it less
favorable for $\tau \in (1, +\infty)$. Thus, we will specifically
further analyze the minimizability gaps in the next section
(Section~\ref{sec:comparison}).

The proof of the theorem is given in
Appendix~\ref{app:bound_comp-sum}. It consists of using the general
$\sH$-consistency bound tools given by \citet{awasthi2022Hconsistency,
  awasthi2022multi} and of analyzing the calibration gap
of the loss function $\ell_{\tau}^{\rm{comp}}$ for different values of
$\tau$ in order to lower bound it in terms of the zero-one loss
calibration gap. As pointed out by
\citet{awasthi2022multi}, deriving such bounds is
non-trivial in the multi-class classification setting. In the proof,
we specifically choose auxiliary functions $\ov h_{\mu}$ target to the
comp-sum losses, which satisfies the property $\sum_{y\in \sY}e^{h(x,
  y)} = \sum_{y\in \sY}e^{\ov h_{\mu}(x, y)}$. Using this property, we
then establish several general lemmas that are applicable to any
$\tau\in [0,\infty)$ and are helpful to lower bound the calibration
  gap of $\ell_{\tau}^{\rm{comp}}$. This is significantly different
  from the proofs of \citet{awasthi2022multi}
  whose analysis depends on concrete loss functions case by
  case. Furthermore, our proof technique actually leads to the
  tightest bounds as shown below. Our proofs are novel and cover the
  full comp-sum loss family, which includes the logistic loss. Next,
  we further prove that the functional form of our bounds
  $\sH$-consistency bounds cannot be improved.

\begin{restatable}[\textbf{Tightness}]{theorem}{TightnessComp}
\label{Thm:tightness-comp}
Assume that $\sH$ is symmetric and complete. Then, for any $\tau\in [0,1]$ and $\beta \in[0,1]$, there exist a distribution $\sD$ and a
hypothesis $h\in\sH$ such that $\sR_{\ell_{0-1}}(h)-
\sR_{\ell_{0-1},\sH}^*+\sM_{\ell_{0-1},\sH}= \beta$ and
$\sR_{\ell_{\tau}^{\rm{comp}}}(h) - \sR_{\ell_{\tau}^{\rm{comp}}}^*(\sH) + \sM_{\ell_{\tau}^{\rm{comp}}}(\sH)=
\sT_{\tau}(\beta)$.
\end{restatable}
The proof is given in Appendix~\ref{app:bound_comp-sum}.  The theorem
shows that the bounds given by the $\sH$-consistency comp-sum
transformation $\sT_{\tau}$, or, equivalently, by its inverse
$\Gamma_{\tau}$ in Theorem~\ref{Thm:bound_comp_sum} is tight for any
$\tau\in [0, 1]$, which includes as special cases the logistic loss
($\tau = 1$).

\subsection{Learning Bounds}

Our $\sH$-consistency bounds can be used to derive zero-one learning
bounds for a hypothesis set $\sH$. For a sample size $m$, let
$\Rad_m^\tau(\sH)$ denote the Rademacher complexity of the family of
functions $\curl*{(x, y) \mapsto \ell_{\tau}^{\rm{comp}}(h, x, y)
  \colon h \in \sH}$ and $B_\tau$
an upper bound on the loss $\ell_{\tau}^{\rm{comp}}$.
\begin{restatable}{theorem}{GenBound}
\label{th:genbound}
With probability at least $1 - \delta$ over the draw of a sample $S$
from $\sD^m$, the following zero-one loss estimation bound holds
for an empirical minimizer $\h h_S \in \sH$ of the comp-sum loss
$\ell_{\tau}^{\rm{comp}}$ over $S$:
\begin{align*}
\sR_{\ell_{0-1}}(\h h_S) - \sR_{\ell_{0-1}}^*(\sH)
\leq \Gamma_{\tau}
  \paren[\bigg]{\sM_{\ell_{\tau}^{\rm{comp}}}(\sH) + 4 \Rad_m^\tau(\sH) +
2 B_\tau \sqrt{\tfrac{\log \frac{2}{\delta}}{2m}}}
- \sM_{\ell_{0-1}}(\sH).
\end{align*}
\end{restatable}
The proof is given in Appendix~\ref{app:genbound}. To our knowledge,
these are the first zero-one estimation loss guarantees for empirical
minimizers of a comp-sum loss such as the logistic loss. Our previous
comments about the properties of $\Gamma_\tau$, in particular its
functional form or its dependency on the number of classes $n$,
similarly apply here. These are precise bounds that take into account
the minimizability gaps.

\section{Comparison of Minimizability Gaps}
\label{sec:comparison}

We now further analyze these quantities and make our guarantees even
more explicit.
Consider a composed loss function defined by $(\Phi_1 \circ \ell_2)(h,
x, y)$, for all $h \in \sH$ and $(x, y) \in \sX \times \sY$, with
$\Phi_1$ concave and non-decreasing.
Then, by Jensen's inequality, we can write:
\begin{equation}
\begin{aligned}
\label{eq:concave-Phi1}
  \sR^*_{\Phi_1 \circ \ell_2}(\sH)
  & = \inf_{h \in \sH} \curl*{\E_{(x, y) \sim \sD}[(\Phi_1 \circ \ell_2)(h, x, y)] }\\
  & \leq \inf_{h \in \sH} \curl*{\Phi_1\paren*{\E_{(x, y) \sim \sD}[\ell_2(h, x, y)]}}\\
  & = \Phi_1\paren*{\inf_{h \in \sH} \curl*{\E_{(x, y) \sim \sD}[\ell_2(h, x, y)]}}\\
  & = \Phi_1\paren*{\sR^*_{\ell_2}(\sH)}.
\end{aligned}
\end{equation}
Recall that the comp-sum losses $\ell_{\tau}^{\rm{comp}}$ can be
written as $\ell_{\tau}^{\rm{comp}}= \Phi^{\tau}\circ
\ell_{\tau=0}^{\rm{comp}}$, where $\ell_{\tau=0}^{\rm{comp}}(h, x, y)
= \sum_{y'\neq y}\exp\paren*{h(x, y') - h(x, y)}$. $\Phi^\tau$ is
concave since we have $\frac{\partial^2 \Phi^\tau}{\partial^2 u}(u) =
\frac{-\tau}{(1 + u)^{\tau + 1}} \leq 0$ for all $\tau \geq 0$ and $u
\geq 0$. Using these observations, the following results can be shown.
\begin{restatable}[\textbf{Characterization of minimizability gaps - stochastic case}]
  {theorem}{GapUpperBound}
\label{Thm:gap-upper-bound}
Assume that $\sH$ is symmetric and complete. Then, for the comp-sum losses $\ell_{\tau}^{\rm{comp}}$, the minimizability gaps can be upper-bounded as follows:
\begin{align}
\label{eq:gap-upper-bound}
\sM_{\ell_{\tau}^{\rm{comp}}}(\sH)
\leq \Phi^{\tau}\paren*{\sR^*_{\ell_{\tau=0}^{\rm{comp}}}(\sH)} - \E_x[\sC^*_{\ell_{\tau}^{\rm{comp}}}(\sH, x)],
\end{align}
where $\sC^*_{\ell_{\tau}^{\rm{comp}}}(\sH, x)$ is given by 
\begin{equation}
\label{eq:comp-sum-Cstar}
\begin{cases}
\frac{1}{1 - \tau} \paren*{\bracket*{\sum_{y\in \sY}p(x,y)^{\frac{1}{2-\tau}}}^{2 - \tau} - 1} & \tau\geq 0, \tau\neq1, \tau \neq 2\\
-\sum_{y\in \sY} p(x,y) \log\bracket*{p(x,y) } & \tau=1\\
1 - \max_{y\in \sY}p(x,y) & \tau =2.
\end{cases}
\end{equation}
\ignore{is a decreasing function of $\tau \geq 0$ for any distribution.}
\end{restatable}
Note that the expressions for $\sC^*_{\ell_{\tau}^{\rm{comp}}}(\sH, x)$ in
\eqref{eq:comp-sum-Cstar} can be formulated in terms of the $(2 -
\tau)$-R\'enyi entropy.

\begin{restatable}[\textbf{Characterization of minimizability gaps - deterministic case}]
  {theorem}{GapUpperBoundDetermi}
\label{Thm:gap-upper-bound-determi}
Assume that for any $x \in \sX$, we have $\curl*{\paren*{h(x, 1),
    \ldots, h(x, n)}\colon h \in \sH}$ = $[-\Lambda,
  +\Lambda]^n$. Then, for comp-sum losses $\ell_{\tau}^{\rm{comp}}$
and any deterministic distribution, the minimizability gaps can be
upper-bounded as follows:
\begin{align}
\label{eq:gap-upper-bound-determi}
\sM_{\ell_{\tau}^{\rm{comp}}}(\sH)
\leq \Phi^{\tau}\paren*{\sR^*_{\ell_{\tau=0}^{\rm{comp}}}(\sH)} - \sC^*_{\ell_{\tau}^{\rm{comp}}}(\sH, x),
\end{align}
where $\sC^*_{\ell_{\tau}^{\rm{comp}}}(\sH, x)$ is given by 
\begin{equation}
\label{eq:comp-sum-Cstar-determi}
\begin{cases}
\frac{1}{1 - \tau} \paren*{\bracket*{1 + e^{-2 \Lambda}(n - 1)}^{1 - \tau} - 1} & \tau\geq 0, \tau\neq1\\
\log\bracket*{1 + e^{-2 \Lambda}(n - 1) } & \tau=1.
\end{cases}
\end{equation}
\end{restatable}
The proofs of these theorems are given in
Appendix~\ref{app:gap-upper-bound}. Note that, when $\tau = 0$,
$\Phi^{\tau}(u)=u$ gives the sum exponential loss $\Phi^{\tau}\circ
\ell_{\tau=0}^{\rm{comp}} = \ell_{\tau=0}^{\rm{comp}}$. For
deterministic distributions, by \eqref{eq:comp-sum-Cstar-determi}, we
obtain $\sC^*_{\ell_{\tau=0}^{\rm{comp}}}(\sH, x) = e^{-2 \Lambda}(n -
1)$.  Therefore, \eqref{eq:comp-sum-Cstar-determi} can be rewritten as
$\sC^*_{\ell_{\tau}^{\rm{comp}}}(\sH, x) =
\Phi^{\tau}\paren*{\sC^*_{\ell_{\tau=0}^{\rm{comp}}}(\sH, x)}$.  Thus,
inequality~\eqref{eq:gap-upper-bound-determi} can be rewritten as
follows:
\begin{align}
\label{eq:gap-upper-bound-determi-final}
\sM_{\ell_{\tau}^{\rm{comp}}}(\sH)
\leq \Phi^{\tau}\paren*{\sR^*_{\ell_{\tau=0}^{\rm{comp}}}(\sH)} - \Phi^{\tau}\paren*{\sC^*_{\ell_{\tau=0}^{\rm{comp}}}(\sH, x)}.
\end{align}
We will denote the right-hand side by $\wt \sM_{\ell_{\tau}^{\rm{comp}}}(\sH)$, 
$\wt \sM_{\ell_{\tau}^{\rm{comp}}}(\sH) = \Phi^{\tau}\paren*{\sR^*_{\ell_{\tau=0}^{\rm{comp}}}(\sH)} - \Phi^{\tau}\paren*{\sC^*_{\ell_{\tau=0}^{\rm{comp}}}(\sH, x)}$.
Note that we always have $\sR^*_{\ell_{\tau=0}^{\rm{comp}}}(\sH)\geq
\E_x\bracket*{\sC^*_{\ell_{\tau=0}^{\rm{comp}}}(\sH, x)}$. Here,
$\E_x\bracket[big]{\sC^*_{\ell_{\tau=0}^{\rm{comp}}}(\sH,
  x)}= \sC^*_{\ell_{\tau=0}^{\rm{comp}}}(\sH, x)$ since
$\sC^*_{\ell_{\tau=0}^{\rm{comp}}}(\sH, x)$ is independent of $x$ as shown in \eqref{eq:comp-sum-Cstar-determi}. Then,
\eqref{eq:gap-upper-bound-determi-final} can be used to compare the
minimizability gaps for different $\tau$. 

\begin{restatable}{lemma}{LemmaCompare}
\label{lemma:lemma-compare}
For any $u_1 \geq u_2\geq 0$, $\Phi^{\tau}(u_1)-\Phi^{\tau}(u_2)$ is non-increasing with respect to $\tau$.
\end{restatable}
The proof is given in
Appendix~\ref{app:lemma-compare}.
Lemma~\ref{lemma:lemma-compare} implies that $\wt \sM_{\ell_{\tau}^{\rm{comp}}}(\sH)$ is a non-increasing function of $\tau$. Thus, given a hypothesis set $\sH$, we have:
\begin{equation}
\label{eq:order-M}
\wt \sM_{\ell_{\tau=0}}(\sH) \geq \wt \sM_{\ell_{\tau=1}}(\sH) \geq  \wt\sM_{\ell_{1<\tau<2}}(\sH)\geq \wt \sM_{\ell_{\tau=2}}(\sH).
\end{equation}
By Section~\ref{sec:preliminaries-comp}, these minimizability gaps
specifically correspond to that of sum-exponential loss ($\tau=0$),
logistic loss ($\tau=1$), generalized cross-entropy loss ($1<\tau<2$)
and mean absolute error loss ($\tau=2$) respectively. Note that for
those loss functions, by Theorem~\ref{Thm:bound_comp_sum}, when the
estimation error $\sR_{\ell_{\tau}^{\rm{comp}}}(h) -
\sR_{\ell_{\tau}^{\rm{comp}}}^*(\sH)$ is minimized to zero, the
estimation error of zero-one classification loss is upper-bounded by
$\wt \Gamma_{\tau}\paren*{\sM_{\ell_{\tau}}}$. Therefore,
\eqref{eq:order-M} combined with the form of $\wt\Gamma_{\tau}$ helps
compare the sum-exponential loss ($\tau=0$), logistic loss ($\tau=0$),
generalized cross-entropy loss ($1<\tau<2$) and mean absolute error
loss ($\tau=2$) in practice. See Section~\ref{sec:experiments-non-adv}
for a discussion of the empirical results in light of these
theoretical findings.

\section{Smooth Adversarial Comp-Sum Losses}
\label{sec:adversarial}

A recent challenge in the application of neural networks is their
robustness to small perturbations
\citep{szegedy2013intriguing}. While neural networks trained on large
datasets have achieved breakthroughs in speech and visual recognition
tasks in recent years
\citep{SutskeverVinyalsLe2014,KrizhevskySutskeverHinton2012}, their
accuracy remains substantially lower in the presence of such
perturbations even for state-of-the-art robust algorithms. One key
factor in the design of robust algorithms is the choice of the
surrogate loss function used for training since directly optimizing
the target adversarial zero-one loss with most hypothesis sets is
NP-hard. To tackle this problem, we introduce a family of loss
functions designed for adversarial robustness that we call
\emph{smooth adversarial comp-sum loss functions}. These are loss
functions obtained by augmenting comp-sum losses with a natural
corresponding smooth term.  We show that these loss functions are
beneficial in the adversarial setting by proving that they admit
$\sH$-consistency bounds. This leads to a family of algorithms for
adversarial robustness that consist of minimizing a regularized smooth
adversarial comp-sum loss.

\subsection{Definition}

In adversarial robustness, the target adversarial zero-one
classification loss is defined as the worst loss incurred over an
$\ell_p$ perturbation ball of $x$ with perturbation size $\gamma$, $p
\in [1, +\infty]$, $\sfB_p(x, \gamma) = \curl*{x'\colon \norm*{x -
    x'}_p \leq \gamma}$:
\[
\ell_{\gamma}(h, x, y) = \sup_{x' \in \sfB_p(x, \gamma)} \ell_{0-1}(h, x', y).
\]
We first introduce the adversarial comp-sum
$\rho$-margin losses, which is defined as the supremum based counterpart of comp-sum losses \eqref{eq:comp-sum_loss} with $\Phi_1 = \Phi^{\tau}$ and \[\Phi_2(u) =
\Phi_{\mathrm{\rho}}(u) = \min\curl*{\max\curl*{0, 1 -
    \frac{u}{\rho}}, 1},\] the $\rho$-margin loss function (see for
example \citep{MohriRostamizadehTalwalkar2018}):
\begin{align*}
\wt
\ell^{\mathrm{comp}}_{\tau,\rho}(h, x, y)
=
\sup_{x':\norm*{x - x'}_p\leq \gamma}
\Phi^{\tau}
\paren*{\sum_{y' \neq y}\Phi_{\rho}\paren*{h(x', y') - h(x', y)}}.
\end{align*}
In the next section, we will show that $\wt
\ell^{\mathrm{comp}}_{\tau,\rho}$ admits an $\sH$-consistency bound
with respect to the adversarial zero-one loss $\ell_{\gamma}$. Since
$\Phi_{\rho}$ is not-convex, we will further derive the \emph{smooth
adversarial comp-sum loss} based on $\wt
\ell^{\mathrm{comp}}_{\tau,\rho}$, that has similar $\sH$-consistency
guarantees and is better to optimize.  By the expression of the
derivative of $\Phi^\tau$ in \eqref{eq:Phi1-derivative}, for all $\tau
\geq 0$ and $u \geq 0$, we have $\abs*{\frac{\partial
    \Phi^{\tau}}{\partial u}(u)} = \frac{1}{(1 + u)^{\tau}} \leq 1$,
thus $\Phi^{\tau}$ is $1$-Lipschitz over $\Rset_+$. Define
$\Delta_h(x, y, y') = h(x, y) - h(x, y')$ and let $\ov \Delta_h(x, y)$
denote the $(n - 1)$-dimensional vector $\paren[big]{\Delta_h(x, y,
  1), \ldots, \Delta_h(x, y, y - 1), \Delta_h(x, y, y + 1), \ldots,
  \Delta_h(x, y, n)}$.  For any $\tau \geq 0$, since $\Phi^{\tau}$ is
$1$-Lipschitz and non-decreasing, we have:
\begin{align*}
\wt \ell^{\mathrm{comp}}_{\tau,\rho}(h, x, y) 
- \ell^{\mathrm{comp}}_{\tau,\rho}(h, x, y)
\leq \sup_{x' \in \sfB(x, \gamma)}\sum_{y'\neq y}\Phi_{\rho}\paren*{-\Delta_h(x',y,y')}
- \ignore{\sum_{y'\neq y}} \Phi_{\rho}\paren*{-\Delta_h(x, y, y')}.
\end{align*}
Since $\Phi_{\rho}(u)$ is $\frac{1}{\rho}$-Lipschitz, 
by the Cauchy-Schwarz inequality, for any
$\nu \geq \frac{\sqrt{n - 1}}{\rho} \geq \frac{1}{\rho}$, we have
\begin{align*}
\wt \ell^{\mathrm{comp}}_{\tau,\rho}(h, x, y)
  & \leq  \ell^{\mathrm{comp}}_{\tau,\rho}(h, x, y) + \nu \sup_{x' \in \sfB(x, \gamma)} \norm*{\ov \Delta_h(x',y)-\ov \Delta_h(x,y)}_2\\
  & \leq  \ell^{\mathrm{comp}}_{\tau}\paren*{\frac{h}{\rho},x,y} + \nu \sup_{x' \in \sfB(x, \gamma)} \norm*{\ov \Delta_h(x',y)-\ov \Delta_h(x,y)}_2,
\end{align*}
where we used the inequality $\exp\paren*{-u/\rho}\geq
\Phi_{\rho}(u)$.  We will refer to a loss function defined by the last
expression as a \emph{smooth adversarial comp-sum loss} and denote it
by $\ell^{\mathrm{comp}}_{\mathrm{smooth}}$. In the next section, we
will provide strong $\sH$-consistency guarantees for
$\ell^{\mathrm{comp}}_{\mathrm{smooth}}$.

\subsection{Adversarial \texorpdfstring{$\sH$}{H}-Consistency Guarantees}

To derive guarantees for our smooth adversarial comp-sum loss, we
first prove an adversarial $\sH$-consistency bound for adversarial
comp-sum $\rho$-margin losses $\wt \ell^{\mathrm{comp}}_{\tau,\rho}$
for any symmetric and \emph{locally $\rho$-consistent} hypothesis set.
\begin{definition}
We say that a hypothesis set $\sH$ is \emph{locally $\rho$-consistent}
if for any $x\in \sX$, there exists a hypothesis $h \in \sH$ such that
$\inf_{x'\colon \norm*{x - x'}\leq \gamma}\abs*{h(x', i) - h(x',
  j)}\geq \rho>0$ for any $i\neq j \in \sY$ and for any $x'\in
\curl*{x'\colon \norm*{x - x'}\leq \gamma}$, $\curl*{h(x',y):y\in
  \sY}$ has the same ordering.
\end{definition}
Common hypothesis sets used in practice, such as the family of linear
models, that of neural networks and of course that of all measurable
functions are all locally $\rho$-consistent for some $\rho > 0$. The
guarantees given in the following result are thus general and widely
applicable.
\begin{restatable}[\textbf{$\sH$-consistency bound of $\wt \ell^{\mathrm{comp}}_{\tau,\rho}$}]
{theorem}{BoundCompRhoAdv}
\label{Thm:bound_comp_rho_adv}
Assume that $\sH$ is symmetric and locally $\rho$-consistent. Then,
for any choice of the hyperparameters $\tau, \rho > 0$, any hypothesis
$h \in \sH$, the following inequality holds:
\begin{equation*}
      \sR_{\ell_{\gamma}}(h)- \sR^*_{\ell_{\gamma}}(\sH)
      \leq \Phi^{\tau}(1)
      \paren*{\sR_{\wt \ell^{\mathrm{comp}}_{\tau,\rho}}(h)-\sR^*_{\wt \ell^{\mathrm{comp}}_{\tau,\rho}}(\sH) + \sM_{\wt \ell^{\mathrm{comp}}_{\tau,\rho}}(\sH)} - \sM_{\ell_{\gamma}}(\sH).
\end{equation*}
\end{restatable}
The proof is given in
Appendix~\ref{app:deferred_proofs_adv_comp}. Using the inequality
$\ell^{\mathrm{comp}}_{\mathrm{smooth}}\geq \wt
\ell^{\mathrm{comp}}_{\tau,\rho}$ yields the following similar
guarantees for smooth adversarial comp-sum loss under the same
condition of hypothesis sets.
\begin{restatable}[\textbf{Guarantees for smooth adversarial comp-sum losses}]
  {corollary}{SmoothCompRhoAdv}
\label{cor:smooth_comp_rho_adv}
Assume that $\sH$ is symmetric and locally $\rho$-consistent. Then,
for any choice of the hyperparameters $\tau, \rho >0$, any hypothesis
$h \in \sH$, the following inequality
holds:
\begin{align}
\sR_{\ell_{\gamma}}(h)- \sR^*_{\ell_{\gamma}}(\sH) \leq \Phi^{\tau}(1)
\bracket*{\sR_{\ell^{\mathrm{comp}}_{\mathrm{smooth}}}(h)-\sR^*_{\wt \ell^{\mathrm{comp}}_{\tau,\rho}}(\sH) + \sM_{\wt \ell^{\mathrm{comp}}_{\tau,\rho}}(\sH)\! } \!-\!\! \sM_{\ell_{\gamma}}(\sH).\nonumber
\end{align}
\end{restatable}
This is the first $\sH$-consistency bound for the comp-sum loss in the adversarial robustness.
As with the non-adversarial scenario in
Section~\ref{sec:H-consistency-bounds}, the minimizability gaps
appearing in those bounds in Theorem~\ref{Thm:bound_comp_rho_adv} and
Corollary~\ref{cor:smooth_comp_rho_adv} actually equal to zero in most
common cases. More precisely, Theorem \ref{Thm:bound_comp_rho_adv}
guarantees $\sH$-consistency for distributions such that the
minimizability gaps vanish:
\begin{align*}
 \sR_{\ell_{\gamma}}(h)- \sR^*_{\ell_{\gamma}}(\sH) \leq \Phi^{\tau}(1) \bracket*{\sR_{\wt \ell^{\mathrm{comp}}_{\tau,\rho}}(h) -
  \sR^*_{\wt \ell^{\mathrm{comp}}_{\tau,\rho}}(\sH)}.  
\end{align*}
For $\tau \in [0, \infty)$ and $\rho>0$, if the estimation loss
  $(\sR_{\wt \ell^{\mathrm{comp}}_{\tau,\rho}}(h) - \sR_{\wt
    \ell^{\mathrm{comp}}_{\tau,\rho}}^*(\sH))$ is reduced to $\e$,
  then, the adversarial zero-one estimation loss
  $(\sR_{\ell_{\gamma}}(h) - \sR_{\ell_{\gamma}}^*(\sH))$ is bounded
  by $\e$ modulo a multiplicative constant. A similar guarantee
  applies to smooth adversarial comp-sum loss as well. These
  guarantees suggest an adversarial robustness algorithm that consists
  of minimizing a regularized empirical smooth adversarial comp-sum
  loss, $\ell^{\mathrm{comp}}_{\mathrm{smooth}}$. We call this
  algorithm \textsc{adv-comp-sum}. In the next section, we report
  empirical results for \textsc{adv-comp-sum}, demonstrating that it significantly outperforms the current state-of-the-art loss/algorithm
  \textsc{trades}.
\ignore{Note that \textsc{trades} corresponds to a specific loss
  function \citep{zhang2019theoretically} and it is not clear if
  \textsc{trades} benefits from adversarial $\sH$-consistency bounds.
}

\section{Experiments}
\label{sec:experiments}

We first report empirical results comparing the
performance of comp-sum losses for different values of $\tau$.
Next, we report a series of empirical results comparing
our adversarial robust algorithm \textsc{adv-comp-sum} with
several baselines.

\subsection{Standard Multi-Class Classification}
\label{sec:experiments-non-adv}

We compared comp-sum losses with different values of $\tau$ on
CIFAR-10 and CIFAR-100 datasets
\citep{Krizhevsky09learningmultiple}\ignore{and SVHN
  \citep{Netzer2011}}. All models were trained via Stochastic Gradient
Descent (SGD) with Nesterov momentum \citep{nesterov1983method}, batch
size $1\mathord,024$ and weight decay $1\times 10^{-4}$.  \ignore{We
  use ResNet-$34$ for CIFAR-10 and CIFAR-100, and ResNet-$16$ for
  SVHN.} We used ResNet-$34$ and trained for $200$ epochs using the
cosine decay learning rate schedule \citep{loshchilov2016sgdr} without
restarts. The initial learning rate was selected from $\curl{0.01,
  0.1, 1.0}$; the best model is reported for each surrogate loss. We
report the zero-one classification accuracy of the models and the
standard deviation for three trials.

\begin{table}[ht]
\caption{Zero-one classification accuracy for comp-sum surrogates;
  mean $\pm$ standard deviation over three runs for different $\tau$.}
    \label{tab:comparison-standrad}
\begin{center}
    \begin{tabular}{@{\hspace{0pt}}llllll@{\hspace{0pt}}}
      $\tau$ & $0$ & $0.5$ & $1.0$ & $1.5$ & $2.0$ \\
      \midrule
     CIFAR-10  & 87.37  & 90.28 &  92.59 &  92.03  & 90.35 \\
     $\pm$ &  0.57 & 0.10 &   0.10 & 0.08 &  0.24\\
      \midrule
      CIFAR-100 & 57.87  & 65.52& 70.93 & 69.87  & 8.99  \\
      $\pm$ &
      0.60 & 0.34 &  0.34 &  0.39 &  0.98
    \end{tabular}
\end{center}
\end{table}

Table~\ref{tab:comparison-standrad} shows that on CIFAR-10 and
CIFAR-100, the logistic loss ($\tau = 1$) outperforms the comp-sum
loss ($\tau = 0.5$) and, by an even larger margin, the sum-exponential
loss ($\tau = 0$). This is consistent with our theoretical analysis
based on $\sH$-consistency bounds in Theorem~\ref{Thm:bound_comp_sum}
since all three losses have the same square-root functional form and
since, by Lemma~\ref{lemma:lemma-compare} and \eqref{eq:wt-Gamma}, the
magnitude of the minimizability gap decreases with $\tau$.

Table~\ref{tab:comparison-standrad} also shows that on CIFAR-10 and
CIFAR-100, the logistic loss ($\tau = 1$) and the generalized
cross-entropy loss ($\tau = 1.5$) achieve relatively close results
that are clearly superior to that of mean absolute error loss
($\tau=2$). This empirical observation agrees with our theoretically
analysis based on their $\sH$-consistency bounds
(Theorem~\ref{Thm:bound_comp_sum}): by
Lemma~\ref{lemma:lemma-compare}, the minimizability gap of $\tau=1.5$
and $\tau=2$ is smaller than that of $\tau = 1$; however, by
\eqref{eq:wt-Gamma}, the dependency of the multiplicative constant on
the number of classes appears for $\tau=1.5$ in the form of
$\sqrt{n}$, which makes the generalized cross-entropy loss less
favorable, and for $\tau = 2$ in the form of $n$, which makes the mean
absolute error loss least favorable. Another reason for the inferior
performance of the mean absolute error loss ($\tau=2$) is that, as
observed in our experiments, it is difficult to optimize in practice,
using deep neural networks on complex datasets. This has also been
previously reported by \citet{zhang2018generalized}. In fact, the mean
absolute error loss can be formulated as an $\ell_1$-distance and is
therefore not smooth; but it has the advantage of robustness, as shown
in \citep{ghosh2017robust}.

\begin{table}[t]
\caption{Clean accuracy and robust accuracy under
  PGD$^{40}_{\mathrm{margin}}$ and AutoAttack; mean $\pm$ standard
  deviation over three runs for both \textsc{adv-comp-sum} and the
  state-of-the-art \textsc{trades} in
  \citep{gowal2020uncovering}. Accuracies of some well-known
  adversarial defense models are included for
  completeness. \textsc{adv-comp-sum} significantly outperforms
  \textsc{trades} for both robust and clean accuracy in all the
  settings.}
    \label{tab:comparison}
\begin{center}
    \resizebox{\textwidth}{!}{
    \begin{tabular}{@{\hspace{0pt}}lllll@{\hspace{0pt}}}
      Method & Dataset & Clean & PGD$^{40}_{\mathrm{margin}}$ & AutoAttack \\
    \midrule
    \citet{gowal2020uncovering} (WRN-70-16) & \multirow{10}{*}{CIFAR-10} & 85.34 $\pm$ 0.04 & 57.90 $\pm$ 0.13 & 57.05 $\pm$ 0.17\\
    \textbf{\textsc{adv-comp-sum} (WRN-70-16)} &  & \textbf{86.16 $\pm$ 0.16} & \textbf{59.35 $\pm$ 0.07} &  \textbf{57.77 $\pm$ 0.08} \\
    \citet{gowal2020uncovering} (WRN-34-20) & &  85.21 $\pm$ 0.16 & 57.54 $\pm$ 0.18 & 56.70 $\pm$ 0.14 \\
    \textbf{\textsc{adv-comp-sum} (WRN-34-20)} &  &  \textbf{85.59 $\pm$ 0.17} & \textbf{58.92 $\pm$ 0.06} &  \textbf{57.41 $\pm$ 0.06}\\
    \citet{gowal2020uncovering} (WRN-28-10) & &  84.33 $\pm$ 0.18 & 55.92 $\pm$ 0.20 & 55.19 $\pm$ 0.23 \\
    \textbf{\textsc{adv-comp-sum} (WRN-28-10)} & & \textbf{84.50 $\pm$ 0.33} & \textbf{57.28 $\pm$ 0.05} &  \textbf{55.79 $\pm$ 0.06}\\
    \cmidrule{1-1} \cmidrule{3-5}
    \citet{pang2020bag} (WRN-34-20)&  &  86.43& \NA&54.39\\
    \citet{DBLP:conf/icml/RiceWK20} (WRN-34-20)&  &  85.34& \NA&53.42 \\
    \citet{wu2020adversarial} (WRN-34-10)&  &  85.36& \NA&56.17 \\
    \citet{qin2019adversarial} (WRN-40-8)&  &  86.28& \NA&52.84 \\
    \midrule
    \citet{gowal2020uncovering} (WRN-70-16) & \multirow{2}{*}{CIFAR-100} & 60.56 $\pm$ 0.31 & 31.39 $\pm$ 0.19 & 29.93 $\pm$ 0.14\\
    \textbf{\textsc{adv-comp-sum} (WRN-70-16)}  & &  \textbf{63.10 $\pm$ 0.24} & \textbf{33.76 $\pm$ 0.18} & \textbf{31.05 $\pm$ 0.15}\\
    \midrule
    \citet{gowal2020uncovering} (WRN-34-20) & \multirow{2}{*}{SVHN} & 93.03 $\pm$ 0.13 & 61.01 $\pm$ 0.16 & 57.84 $\pm$ 0.19\\
    \textbf{\textsc{adv-comp-sum} (WRN-34-20)}  & &  \textbf{93.98 $\pm$ 0.12} & \textbf{62.97 $\pm$ 0.05} &  \textbf{58.13 $\pm$ 0.12}\\
    \end{tabular}
    }
\end{center}
\end{table}

\subsection{Adversarial Multi-Class Classification}

Here, we report empirical results for our adversarial robustness
algorithm \textsc{adv-comp-sum} on CIFAR-10, CIFAR-100
\citep{Krizhevsky09learningmultiple} and SVHN \citep{Netzer2011}
datasets\ignore{, showing that it outperforms the current
  state-of-the-art algorithm, \textsc{trades},}. No generated data or
extra data was used.

\textbf{Experimental settings.}  We followed exactly the experimental
settings of \citet{gowal2020uncovering} and adopted precisely the same
training procedure and neural network architectures, which are
WideResNet (WRN) \citep{zagoruyko2016wide} with SiLU activations
\citep{hendrycks2016gaussian}. Here, WRN-$n$-$k$ denotes a residual
network with $n$ convolutional layers and a widening factor $k$. For
CIFAR-10 and CIFAR-100, the simple data augmentations, 4-pixel padding
with $32 \times 32$ random crops and random horizontal flips, were
applied. We used 10-step Projected Gradient-Descent (PGD) with random
starts to generate training attacks. All models were trained via
Stochastic Gradient Descent (SGD) with Nesterov momentum
\citep{nesterov1983method}, batch size $1\mathord,024$ and weight
decay $5\times 10^{-4}$. We trained for $400$ epochs using the cosine
decay learning rate schedule \citep{loshchilov2016sgdr} without
restarts. The initial learning rate is set to $0.4$. We used model
weight averaging \citep{DBLP:conf/uai/IzmailovPGVW18} with decay rate
$0.9975$. For \textsc{trades}, we adopted exactly the same setup as
\citet{gowal2020uncovering}. For our smooth adversarial comp-sum
losses, we set both $\rho$ and $\nu$ to $1$ by default. In practice,
they can be selected by cross-validation and that could potentially
lead to better performance. The per-epoch computational cost of our
method is similar to that of \textsc{trades}.

\textbf{Evaluation.} We used early stopping on a held-out validation
set of $1\mathord,024$ samples by evaluating its robust accuracy
throughout training with 40-step PGD on the margin loss, denoted by
PGD$^{40}_{\mathrm{margin}}$, and selecting the best check-point
\citep{DBLP:conf/icml/RiceWK20}.
We report the \emph{clean accuracy}, that is the standard
classification accuracy on the test set, and the robust accuracy with
$\ell_{\infty}$-norm perturbations bounded by $\gamma = 8/255$ under
PGD attack, measured by PGD$^{40}_{\mathrm{margin}}$ on the full test
set, as well as under AutoAttack \citep{croce2020reliable} ({\small
  \url{https://github.com/fra31/auto-attack}}), the state-of-the-art
attack for measuring empirically adversarial robustness. We averaged
accuracies over three runs and report the standard deviation for both
\textsc{adv-comp-sum} and \textsc{trades}, reproducing the results
reported for \textsc{trades} in \citep{gowal2020uncovering}.

\textbf{Results}.  Table~\ref{tab:comparison} shows that
\textsc{adv-comp-sum} outperforms \textsc{trades} on CIFAR-10 for all
the neural network architectures adopted (WRN-70-16, WRN-34-20 and
WRN-28-10). Here, \textsc{adv-comp-sum} was implemented with
$\tau=0.4$. Other common choices of $\tau$ yield similar results,
including $\tau=1$ (logistic loss). In all the settings, robust
accuracy under AutoAttack is higher by at least 0.6\% for
\textsc{adv-comp-sum}, by at least 1.36\% under the
PGD$^{40}_{\mathrm{margin}}$ attack.

It is worth pointing out that the improvement in robustness accuracy
for our models does not come at the expense of a worse clean accuracy
than \textsc{trades}. In fact, \textsc{adv-comp-sum} consistently
outperforms \textsc{trades} for the clean accuracy as well. For the
largest model WRN-70-16, the improvement is over 0.8\%. For
completeness, we also include in Table~\ref{tab:comparison} the
results for some other well-known adversarial defense
models. \textsc{adv-comp-sum} with the smallest model WRN-28-10
surpasses \citep{pang2020bag, DBLP:conf/icml/RiceWK20,
  qin2019adversarial}. \citep{wu2020adversarial} is significantly
outperformed by \textsc{adv-comp-sum} with a slightly larger model
WRN-34-20, by more than 1.2\% in the robust accuracy and also in the
clean accuracy.

To show the generality of our approach, we carried out experiments
with other datasets, including CIFAR-100 and SVHN. For WRN-70-16 on
CIFAR-100, \textsc{adv-comp-sum} outperforms \textsc{trades} by 1.12\%
in the robust accuracy and 2.54\% in the clean accuracy. For WRN-34-20
on SVHN, \textsc{adv-comp-sum} also outperforms \textsc{trades} by
0.29\% in the robust accuracy and 0.95\% in the clean accuracy.

Let us underscore that outperforming the state-of-the-art results of
\citet{gowal2020uncovering} in the same scenario and without resorting
to additional unlabeled data has turned out to be very challenging:
despite the large research emphasis on this topic in the last several
years and the many publications, none was reported to surpass that
performance, using an alternative surrogate loss.\ignore{ Thus, the
  improvements we report both in clean accuracy and robust accuracy
  with respect to the results of \citep{gowal2020uncovering} are
  significant.}

\section{Discussion}
\label{sec:future-work}

\textbf{Applications of $\sH$-consistency bounds}.  Given a hypothesis
set $\sH$, our quantitative $\sH$-consistency bounds can help select
the most favorable surrogate loss, which depends on (i) the functional
form of the $\sH$-consistency bound: for instance, the bound for the
mean absolute error loss exhibits a linear dependency, while that of
the logistic loss and generalized cross-entropy losses exhibit a
square-root dependency, resulting in a less favorable convergence
rate; (ii) the smoothness of the loss and, more generally, its
optimization properties; for example, the mean absolute error loss is
less smooth than the logistic loss, and surrogate losses with more
favorable bounds may lead to more challenging optimizations; in fact,
the zero-one loss serves as its own surrogate with the tightest bound
for any hypothesis set, but is known to result in NP-complete
optimization problems for many common choices of $\sH$; (iii)
approximation properties of the surrogate loss function: for instance,
given a choice of $\sH$, the minimizability gap for a surrogate loss
may be more or less favorable; (iv) the dependency of the
multiplicative constant on the number of classes: for example, the
linear dependency of $n$ in the bound for the mean absolute error loss
makes it less favorable than the logistic loss.

Another application is the derivation of generalization bounds for
surrogate loss minimizers (see Theorem ~\ref{th:genbound}), expressed
in terms of the quantities discussed above.

\textbf{Concurrent work}. The concurrent and independent study of
\citet{zheng2023revisiting} also provides an $\sH$-consistency bound
for the logistic loss. Their bound holds for the special case of $\sH$
being a constrained linear hypothesis set, subject to an additional
assumption on the distribution. In contrast, our bounds do not require
any distributional assumption. However, it should be noted that our
results are only applicable to complete hypothesis sets.  In upcoming
work, we present $\sH$-consistency bounds for non-complete hypothesis
sets and arbitrary distributions.

\textbf{Future work}. In addition to the extension to non-complete
hypothesis sets just mentioned, it would be valuable to investigate
the application or generalization of $\sH$-consistency bounds in
scenarios involving noisy labels
\citep{ghosh2017robust,zhang2018generalized}.
For comp-sum losses, this paper focuses on the case where $\Phi_2$ is
the exponential loss and $\Phi_1$ is based on \eqref{eq:Phi1}. This
includes the cross-entropy loss (or logistic loss), generalized
cross-entropy, the mean absolute error and other cross-entropy-like
functions, which are the most widely used ones in the family of
comp-sum losses. The study of other such loss functions and the
comparison with other families of multi-class loss functions
\citep{awasthi2022multi} is left to the future work.
Although our algorithm demonstrates improvements over the
current state-of-the-art technique, adversarial robustness remains a
challenging problem. A key issue seems to be that of generalization
for complex families of neural networks (see for example
\citep*{awasthi2020adversarial}). A more detailed study of that
problem might help enhance the performance of our algorithm.
Finally, in addition to their immediate implications, our results and
techniques have broader applications in analyzing surrogate losses and
algorithms across different learning scenarios. For instance, they can
be used in the context of ranking, as demonstrated in recent work by
\citet*{MaoMohriZhong2023ranking}. Furthermore, they can be extended to
address the challenges of learning with abstention
\citep*{cortes2016learning,cortes2016boosting}. Additionally, our
findings can be valuable in non-i.i.d.\ learning settings, such as
drifting \citep{MohriMunozMedina2012} or time series prediction
\citep{KuznetsovMohri2018,KuznetsovMohri2020}.

\section{Conclusion}
\label{sec:conclusion}

We presented a detailed analysis of the theoretical properties of a
family of surrogate losses that includes the logistic loss (or
cross-entropy with the softmax). These are more precise and more
informative guarantees than Bayes consistency since they are
non-asymptotic and specific to the hypothesis set used. Our bounds are
tight and can be made more explicit, when combined with our analysis
of minimizability gaps. These inequalities can help compare different
surrogate losses and evaluate their advantages in different scenarios.
We showcased one application of this analysis by extending comp-sum
losses to the adversarial robustness setting, which yields principled
surrogate losses and algorithms for that scenario. We believe that our
analysis can be helpful to the design of algorithms in many other
scenarios.

\chapter{Characterization and Extensions} \label{ch5}
In this chapter, we provide both a general characterization and an extension
of $\sH$-consistency bounds for multi-class classification. Previous
approaches to deriving these bounds required the development of new
proofs for each specific case. In contrast, we introduce the general
concept of an \emph{error transformation function} that serves as a
very general tool for deriving such guarantees with tightness
guarantees. We show that deriving an $\sH$-consistency bound for
comp-sum losses and constrained losses for both complete and bounded
hypothesis sets can be reduced to the calculation of their
corresponding error transformation function. Our general tools and
tight bounds show several remarkable advantages: first, they improve
existing bounds for complete hypothesis sets previously proven in
\citep{awasthi2022multi}; second, they encompass all previously
comp-sum and constrained losses studied thus far as well as many new
ones \citep{awasthi2022Hconsistency, mao2023cross}; third, they extend
beyond the completeness assumption adopted in previous work; fourth,
they provide novel guarantees for bounded hypothesis sets; and, finally,
they help prove a much stronger and more significant guarantee for
logistic loss with linear hypothesis set than
\citep{zheng2023revisiting}.

We present new and tight
$\sH$-consistency bounds for both the family of comp-sum losses
(Section~\ref{sec:bounds-comp}) and that of constrained losses
(Section~\ref{sec:bounds-cstnd}), which cover the familiar
cross-entropy, or logistic loss applied to the outputs of a neural
network. We further extend our analysis beyond the completeness
assumptions adopted in previous studies and cover more realistic
bounded hypothesis sets (Section~\ref{sec:extension-comp}
and~\ref{sec:extension_cstnd}).  Our characterizations are based on
error transformations, which are explicitly defined for each
formulation. We illustrate the application of our general results
through several special examples. A by-product of our analysis is the
observation that a recently derived multi-class $\sH$-consistency
bound for cross-entropy reduces to an excess bound independent of the
hypothesis set. Instead, we prove a much stronger and more significant
guarantee (Section~\ref{sec:extension-comp}). We start with some basic definitions
and notation in Section~\ref{sec:pre}. 

The presentation in this chapter is based on \citep{MaoMohriZhong2023characterization}.

\section{Preliminaries}
\label{sec:pre}

We denote by $\sX$ the input space, by $\sY$ the output space, and by
$\sD$ a distribution over $\sX\times \sY$.  We consider the standard
scenario of multi-class classification, where $\sY = \curl*{1, \ldots,
  n}$. Given a hypothesis set $\sH$ of functions mapping $\sX \times
\sY$ to $\Rset$, the multi-class classification problem consists of
finding a hypothesis $h\in \sH$ with small generalization error
$\sR_{\ell_{0-1}}(h)$, defined by $\sR_{\ell_{0-1}}(h) = \E_{(x, y)
  \sim \sD}[\ell_{0-1}(h, x, y)]$, where $\ell_{0-1}(h, x, y) =
\1_{\hh(x)\neq y}$ is the multi-class zero-one loss with $\hh(x) =
\argmax_{y\in \sY}h(x, y)$ the prediction of $h$
for the input point $x$. We also denote by $\mathsf H(x)$
the set of all predictions associated to input $x$ generated by
functions in $\sH$, that is, $\mathsf H(x) = \curl*{\hh(x) \colon h
  \in \sH}$.

We will analyze the guarantees of surrogate multi-class losses in
terms of the zero-one loss. We denote by $\ell$ a surrogate loss and
by $\sR_{\ell}(h)$ its generalization error, $\sR_{\ell}(h) = \E_{(x,
  y) \sim \sD}[\ell(h, x, y)]$. For a loss function $\ell$, we define
the best-in-class generalization error within a hypothesis set $\sH$
as $\sR^*_{\ell}(\sH) = \inf_{h \in \sH}\sR_{\ell}(h)$, and refer to
$\sR_{\ell}(h) - \sR^*_{\ell}(\sH)$ as the \emph{estimation
error}. We will study the key notion of \emph{$\sH$-consistency
bounds} \citep{awasthi2022Hconsistency,awasthi2022multi}, which are
upper bounds on the zero-one estimation error of any predictor in a
hypothesis set, expressed in terms of its surrogate loss estimation
error, for some real-valued function $f$ that is non-decreasing:
\begin{equation*}
  \forall h \in \sH,\; \sR_{\ell_{0-1}}(h) - \sR^*_{\ell_{0-1}}(\sH)
  \leq f \paren*{\sR_{\ell}(h) - \sR^*_{\ell}(\sH)}.
\end{equation*}
These bounds imply that the zero-one estimation error is at most
$f(\e)$ whenever the surrogate loss estimation error is bounded by
$\e$. Thus, the learning guarantees provided by $\sH$-consistency
bounds are both non-asymptotic and hypothesis set-specific. The
function $f$ appearing in these bounds is expressed in terms of a
\emph{minimizability gap}, which is a quantity measuring the
difference of best-in-class error $\sR^*_{\ell}(\sH)$ and the expected
\emph{best-in-class conditional error} $\E_{x} \bracket*
     {\sC_{\ell}^*(\sH,x)}$: \[\sM_{\ell}(\sH) = \sR^*_{\ell}(\sH) -
     \E_{X}\bracket* {\sC_{\ell}^*(\sH, x)},\] where
     $\sC_{\ell}(h,x)=\E_{y|x}\bracket*{\ell(h, x, y)}$ and
     $\sC_{\ell}^*(\sH,x)=\inf_{h\in \sH}\sC_{\ell}(h,x)$ are the
     \emph{conditional error} and \emph{best-in-class conditional
     error} respectively. We further write $\Delta\sC_{\ell,\sH} =
     \sC_{\ell}(h, x) - \sC_{\ell}^*(\sH, x)$ to denote the
     \emph{conditional regret}. Note that that the minimizability gap
     is an inherent quantity depending on a hypothesis set $\sH$ and
     the loss function $\ell$.
  
By Lemma~\ref{lemma:explicit_assumption_01-chcb}, the minimizability gap
for the zero-one loss, $\sM_{\ell_{0-1}}(\sH)$, coincides with its
approximation error $\sA_{\ell_{0-1}}(\sH) = \sR^*_{\ell_{0-1}}(\sH) -
\sR^*_{\ell_{0-1}}(\sH_{\rm{all}})$ when the set of all possible
predictions generated by $\sH$ covers the label space $\sY$. This
holds for typical hypothesis sets used in practice. However, for a
surrogate loss $\ell$, the minimizability gap $\sM_{\ell}(\sH)$ is
always upper-bounded by and in general finer than its approximation
error $\sA_{\ell}(\sH) =
\sR^*_{\ell}(\sH)-\sR^*_{\ell}(\sH_{\rm{all}})$ since $\sM_{\ell}(\sH)
= \sA_{\ell}(\sH) - I_{\ell}(\sH)$, where $\sH_{\rm{all}}$ is the
family of all measurable functions and $I_{\ell}(\sH) =
\mathbb{E}_{x}\left[\sC_{\ell}^*(\sH,x) -
  \sC_{\ell}^*(\sH_{\rm{all}},x)\right]$ (see
Appendix~\ref{app:minimizability-gap} for a more detailed
discussion). Thus, an $\sH$-consistency bound, expressed as follows
for some increasing function $\Gamma$:
  \begin{equation}
  \label{eq:H-consistency-bound}
  \sR_{\ell_{0-1}}(h)-\sR^*_{\ell_{0-1}}(\sH) +
  \sM_{\ell_{0-1}}(\sH)\leq \Gamma\paren*{\sR_{\ell}(h)
    - \sR^*_{\ell}(\sH)+\sM_{\ell}(\sH)},
  \end{equation}
is more favorable than an excess error bound expressed in terms of
approximation errors \[\sR_{\ell_{0-1}}(h) - \sR^*_{\ell_{0-1}}(\sH) +
\sA_{\ell_{0-1}}(\sH) \leq \Gamma\paren*{\sR_{\ell}(h) -
  \sR^*_{\ell}(\sH) + \sA_{\ell}(\sH)}.\] Here, $\Gamma$ is typically
linear or the square-root function modulo constants. When $\sH =
\sH_{\rm{all}}$, the family of all measurable functions, an
$\sH$-consistency bound coincides with the excess error bound and
implies Bayes-consistency by taking the limit. It is therefore
a stronger guarantee than an excess error bound
and Bayes-consistency.

The minimizability gap is always non-negative, since the infimum of
the expectation is greater than or equal to the expectation of
infimum. Furthermore, as shown in
Appendix~\ref{app:minimizability-gap}, when $\sH$ is the family of all
measurable functions or when the Bayes-error coincides with the
best-in-class error, that is,
$\sR^*_{\ell}(\sH) = \sR^*_{\ell}(\sH_{\rm{all}})$, the minimizability
gap vanishes. In such cases, \eqref{eq:H-consistency-bound} implies
the $\sH$-consistency of a surrogate loss $\ell$ with respect to the
zero-one loss $\ell_{0-1}$:
\begin{equation*}
  \sR_{\ell}(h_n)
  - \sR^*_{\ell}(\sH) \xrightarrow{n \rightarrow +\infty} 0
  \implies
  \sR_{\ell_{0-1}}(h_n) -
  \sR^*_{\ell_{0-1}}(\sH) \xrightarrow{n \rightarrow +\infty} 0.
\end{equation*}
In the next sections, we will provide both a general characterization
and an extension of $\sH$-consistency bounds for multi-class
classification. Before proceeding, we first introduce a useful lemma
from \citep{awasthi2022multi} which characterizes the conditional
regret of zero-one loss explicitly. We denote by $p(x) = \paren*{\sfp(1 \!\mid\! x), \ldots, \sfp(n \!\mid\! x)}$ as the conditional distribution of $y$ given
$x$.
 
\begin{restatable}{lemma}{ExplicitAssumptionChcb}
\label{lemma:explicit_assumption_01-chcb}
For zero-one loss $\ell_{0-1}$, the best-in-class conditional error
and the conditional regret for $\ell_{0-1}$ can be expressed as
follows: for any $x \in \sX$, we have
\begin{align*}
  \sC^*_{\ell_{0-1}}(\sH,x)
  = 1 - \max_{y \in \mathsf H(x)} \sfp(y \!\mid\! x) \quad \text{and} \quad
  \Delta\sC_{\ell_{0-1},\sH}(h,x)
  = \max_{y \in \mathsf H(x)} \sfp(y \!\mid\! x) - \sfp(\hh(x) \!\mid\! x).
\end{align*}
\end{restatable}

\section{Comparison with previous work}
\label{sec:previous}

Here, we briefly discuss previous studies of $\sH$-consistency bounds
\citep{awasthi2022Hconsistency, awasthi2022multi, zheng2023revisiting,
  mao2023cross} in standard binary or multi-class classification and
compare their results with those we present.

\citet{awasthi2022Hconsistency} studied $\sH$-consistency bounds in
binary classification. They provided a series of \emph{tight}
$\sH$-consistency bounds for the \emph{bounded} hypothesis set of
linear models $\sH_{\mathrm{lin}}^{\rm{bi}}$ and one-hidden-layer
neural networks $\sH_{\mathrm{NN}}^{\rm{bi}}$, defined as follows:
\begin{align*}
  & \sH_{\mathrm{lin}}^{\rm{bi}}
  = \big\{x\mapsto w \cdot x + b
  \mid \norm*{w}\leq W,\abs*{b}\leq B\big\}\\
  & \sH_{\mathrm{NN}}^{\rm{bi}} = \big\{x\mapsto \sum_{j =
    1}^n u_j(w_j \cdot x+b)_{+} \mid \|u \|_{1}\leq
  \Lambda,\|w_j\|\leq W, \abs*{b}\leq B\big\},
\end{align*}
where $B$, $W$, and $\Lambda$ are positive constants and where
$(\cdot)_+ = \max(\cdot,0)$. We will show that our bounds recover
these binary classification $\sH$-consistency bounds.

The scenario of multi-class classification is more challenging and
more crucial in applications.  Recent work by \citet{awasthi2022multi}
showed that \emph{max losses} \citep{crammer2001algorithmic}, defined
as $\ell^{\mathrm{max}}(h,x,y) = \max_{y'\neq
  y}\Phi\paren*{h(x,y)-h(x,y')}$ for some convex and non-increasing
function $\Phi$, cannot admit meaningful $\sH$-consistency bounds,
unless the distribution is deterministic. They also presented a series
of $\sH$-consistency bounds for \emph{sum losses}
\citep{weston1998multi} and \emph{constrained losses}
\citep{lee2004multicategory} for \emph{symmetric} and \emph{complete}
hypothesis sets, that is such that:
\begin{align*}
  & \sH
  = \curl*{h:\sX\times\sY \to \Rset \colon h(\cdot,y)\in \sF,\forall y\in \sY}  \tag{symmetry} \\
  & 
  \forall x\in \sX, \curl*{f(x)\colon
    f\in \sF} = \Rset, \tag{completeness}
\end{align*}
for some family $\sF$ of functions mapping from $\sX$ to $\Rset$.  The
completeness assumption rules out the bounded hypothesis sets
typically used in practice such as $\sH_{\rm{lin}}$. Moreover, the
final bounds derived from \citep{awasthi2022multi} are based on ad hoc
proofs and may not be tight. In contrast, we will study both the
complete and bounded hypothesis sets, and provide a very general tool
to derive $\sH$-consistency bounds. Our bounds are tighter than those
of \citet{awasthi2022multi} given for complete hypothesis sets and
extend beyond the completeness assumption.

\citep{mao2023cross} complemented the work of \citep{awasthi2022multi}
by studying a wide family of \emph{comp-sum losses} in multi-class
classification, which generalized the \emph{sum-losses} and included
as special cases the logistic loss
\citep{Verhulst1838,Verhulst1845,Berkson1944,Berkson1951}, the
\emph{generalized cross-entropy loss} \citep{zhang2018generalized},
and the \emph{mean absolute error loss} \citep{ghosh2017robust}. Here
too, the completeness assumption was adopted, thus 
their $\sH$-consistency bounds do not apply to common bounded
hypothesis sets used in practice. We illustrate the application of our
general results through a broader set of surrogate losses than
\citep{mao2023cross} and significantly generalize the bounds of
\citep{mao2023cross} to bounded hypothesis sets.

Recently, \citet{zheng2023revisiting} proved $\sH$-consistency bounds
for logistic loss with linear hypothesis sets in the multi-class
classification:
  \[\sH_{\mathrm{lin}}
  = \curl*{x \mapsto w_y \cdot x + b_y
\mid \norm*{w_y} \leq W, \abs*{b_y} \leq B, y \in \sY}.\]
However, their bounds require a crucial distributional assumption
under which, the minimizability gaps
$\sM_{\ell_{0-1}}\paren*{\sH_{\rm{lin}}}$ and
$\sM_{\ell_{\rm{log}}}\paren*{\sH_{\rm{lin}}}$ coincide with the
approximation errors
$\sR_{\ell_{0-1}}\paren*{\sH_{\rm{lin}}}-\sR_{\ell_{0-1}}^*\paren*{\sH_{\rm{all}}}$
and
$\sR_{\ell_{\rm{log}}}\paren*{\sH_{\rm{lin}}}-\sR_{\ell_{\rm{log}}}^*\paren*{\sH_{\rm{all}}}$
respectively (see the note before
\citep[Appendix~F]{zheng2023revisiting}). Thus, their bounds can be
recovered as excess error bounds
\[\sR_{\ell_{0-1}}\paren*{h} - \sR_{\ell_{0-1}}^*\paren*{\sH_{\rm{all}}}
\leq \sqrt{2}
\paren*{\sR_{\ell_{\rm{log}}}\paren*{h}-\sR_{\ell_{\rm{log}}}^*
  \paren*{\sH_{\rm{all}}}}^{\frac12},\]
which are less significant. In contrast, our
$\sH_{\rm{lin}}$-consistency bound are much finer and take into
account the role of the parameter $B$ and that of the number of labels
$n$. Thus, we provide stronger and more significant guarantees for
logistic loss with linear hypothesis set than
\citep{zheng2023revisiting}.

In summary, our general tools offer the remarkable advantages of
deriving tight bounds, which improve upon the existing bounds of
\citet{awasthi2022multi} given for complete hypothesis sets, cover the
comp-sum and constrained losses considered in
\citep{awasthi2022Hconsistency,mao2023cross} as well as new ones,
extend beyond the completeness assumption with novel guarantees valid
for bounded hypothesis sets, and are much stronger and more
significant guarantees for logistic loss with linear hypothesis sets
than those of \citet{zheng2023revisiting}.

\section{Comp-sum losses}
\label{sec:comp}

In this section, we present a general characterization of
$\sH$-consistency bounds for \emph{comp-sum losses}, a family of loss
functions including the \emph{logistic loss}
\citep{Verhulst1838,Verhulst1845,Berkson1944,Berkson1951}, the
\emph{sum exponential loss} \citep{weston1998multi,awasthi2022multi},
the \emph{generalized cross entropy loss}
\citep{zhang2018generalized}, the \emph{mean absolute error loss}
\citep{ghosh2017robust}, and many other loss functions used in
applications.

This is a family of loss functions defined via the composition of a
non-negative and non-decreasing function $\Psi$ with the
sum exponential losses (see \citep{mao2023cross}):
\begin{align}
\label{eq:comp-sum_loss-chcb}
\forall h \in \sH, \forall (x, y) \times \sX \times \sY,  \quad
\ell^{\rm{comp}}(h, x, y)
= \Psi\paren*{\sum_{y'\neq \sY}e^{h(x, y') - h(x, y)}}.
\end{align}
This expression can be equivalently written as $\ell^{\rm{comp}}(h, x,
y) = \Phi\paren*{\frac{e^{h(x, y)}}{\sum_{y'\in \sY}e^{h(x,y')}}}$,
where $\Phi\colon u \mapsto \Psi (\frac{1 - u}{u})$ is a
non-increasing auxiliary function from $[0, 1]$ to $\Rset_{+} \cup
\curl{+\infty}$. As an example, the logistic loss corresponds to the
choice $\Phi\colon u \mapsto -\log(u)$ and the sum exponential loss to
$\Phi \colon u \mapsto \frac{1-u}{u}$.

\subsection{\texorpdfstring{$\sH$}{H}-consistency bounds}
\label{sec:bounds-comp}

In previous work, deriving $\sH$-consistency bounds has required
giving new proofs for each instance. The following result provides a
very general tool for deriving such bounds with tightness guarantees. We
introduce an \emph{error transformation function} and show that
deriving an $\sH$-consistency bound for comp-sum losses can be reduced
to the calculation of this function.

\begin{restatable}[\textbf{$\sH$-consistency bound for comp-sum losses}]
  {theorem}{BoundComp}
\label{Thm:bound_comp}
Assume that $\sH$ is symmetric and complete and that $\sT^{\rm{comp}}$
is convex.  Then, the following inequality holds for any hypothesis $h
\in \sH$ and any distribution
\begin{align}
\label{eq:bound_comp}
     \sT^{\rm{comp}}\paren*{\sR_{\ell_{0-1}}(h)-\sR^*_{\ell_{0-1}}(\sH)+\sM_{\ell_{0-1}}(\sH)}\leq \sR_{\ell^{\mathrm{comp}}}(h)-\sR^*_{\ell^{\mathrm{comp}}}(\sH)+\sM_{\ell^{\mathrm{comp}}}(\sH),
\end{align}
with $\sT^{\rm{comp}}$ an \emph{$\sH$-estimation error transformation for
  comp-sum losses} defined for all $t\in
  \left[0, 1\right]$ by
\begin{align*}
& \sT^{\rm{comp}}(t) = \\
& \begin{cases}
\inf\limits_{\tau\in \bracket*{0,\frac12}}\sup\limits_{\mu\in [-\tau,1 - \tau]} \curl*{\frac{1 + t}{2}\bracket*{\Phi\paren*{\tau} - \Phi\paren*{1 - \tau - \mu}} + \frac{1 - t}{2}\bracket*{ \Phi\paren*{1 - \tau} - \Phi\paren*{\tau + \mu}}} & n =2 \\
\inf\limits_{\psum\in \bracket*{\frac{1}{n-1}\vee t,1}}\inf\limits_{\substack{ \tau_1\geq \max(\tau_2,1/n)\\ \tau_1+\tau_2\leq 1,\tau_2\geq 0}}\sup\limits_{\mu\in [-\tau_2, \tau_1]} \curl*{\frac{\psum + t}{2}\bracket*{\Phi\paren*{\tau_2} - \Phi\paren*{\tau_1-\mu}} + \frac{\psum-t}{2}\bracket*{ \Phi\paren*{\tau_1} - \Phi\paren*{\tau_2+\mu}}} & n > 2. 
\end{cases}
\end{align*}
Furthermore, for any $t \in[0,1]$, there exist a distribution $\sD$
and a hypothesis $h \in \sH$ such that $\sR_{\ell_{0-1}}(h)-
\sR^*_{\ell_{0-1}}(\sH)+\sM_{\ell_{0-1}}(\sH)=t$ and
$\sR_{\ell^{\mathrm{comp}}}(h) - \sR_{\ell^{\mathrm{comp}}}^*(\sH) +
\sM_{\ell^{\mathrm{comp}}}(\sH)= \sT^{\rm{comp}}(t)$.
\end{restatable}
Thus, Theorem~\ref{Thm:bound_comp} shows that, when $\sT^{\rm{comp}}$
is convex, to make these guarantees explicit, all that is needed is to
calculate $\sT^{\rm{comp}}$. Moreover, the last statement shows the
\emph{tightness} of the guarantees derived using this function. The
constraints in $\sT^{\rm{comp}}$ are due to the forms that the
conditional probability vector and scoring functions take. These forms
become more flexible for $n>2$, leading to intricate constraints. Note
that our $\sH$-consistency bounds are distribution-independent and we
cannot claim tightness across all distributions.

The general expression of $\sT^{\rm{comp}}$ in
Theorem~\ref{Thm:bound_comp} is complex, but it can be considerably
simplified under some broad assumptions, as shown by the following
result.

\begin{restatable}[\textbf{characterization of $\sT^{\rm{comp}}$}]
  {theorem}{CharComp}
\label{Thm:char_comp}
Assume that $\Phi$ is convex, differentiable at $\frac12$ and $\Phi'\paren*{\frac12}<0$. Then, $\sT^{\rm{comp}}$ can be expressed as follows:
\begin{align*}
\sT^{\rm{comp}}(t)
& = \begin{cases}
 \Phi\paren*{\frac12} - \inf_{\mu\in\bracket*{-\frac12,\frac12}}\curl*{\frac{1 - t}{2}\Phi\paren*{\frac12+\mu}+\frac{1 + t}{2}\Phi\paren*{\frac12-\mu}} & n=2\\
 \inf_{\tau\in\bracket*{\frac1n,\frac12}}\curl*{\Phi(\tau) -\inf_{\mu\in [-\tau,\tau]}\curl*{\frac{1 + t}{2}\Phi(\tau-\mu)+\frac{1 - t}{2}\Phi\paren*{\tau+\mu}}} & n>2.
\end{cases}
\end{align*}
\end{restatable}
The proof of this result as well as that of other theorems in this
section are given in Appendix~\ref{app:comp-sum}.

\textbf{Examples.}
We now illustrate the application of our theory through several
examples. To do so, we compute the $\sH$-estimation error
transformation $\sT^{\rm{comp}}$ for comp-sum losses and present the
results in Table~\ref{tab:transformation_comp}. Remarkably, by
applying Theorem~\ref{Thm:bound_comp}, we are able to obtain the same
$\sH$-consistency bounds for comp-sum losses with $\Phi(t)=-\log(t)$,
$\frac{1}{t}-1$, $\frac{1}{q}\paren*{1 - t^{q}},q\in (0,1)$ and $1 -
t$ as those derived using ad hoc methods in \citep{mao2023cross}, and a novel tight $\sH$-consistency bound for the new comp-sum loss $\ell_{\rm{sq}} = \bracket*{1 - \frac{e^{h(x, y)}}{\sum_{y'\in \sY}e^{h(x, y')}}}^2$ with $\Phi(t)=(1-t)^2$ in Theorem~\ref{thm:sq}.

The calculation of $\sT^{\rm{comp}}$ for all entries of Table~\ref{tab:transformation_comp} is detailed in Appendix~\ref{app:examples-comp}. To illustrate the effectiveness of our general tools, here, we show how the error transformation function can be  straightforwardly calculated in the case of the new surrogate loss $\ell_{\rm{sq}}$.

\begin{theorem}[\textbf{$\sH$-consistency bound for a new comp-sum loss}]
\label{thm:sq}
Assume that $\sH$ is symmetric and complete. Then, for all $h\in \sH$ and any distribution, the following tight bound holds.
\begin{equation*}
\sR_{\ell_{0-1}}(h)-\sR^*_{\ell_{0-1}}(\sH)\leq 2\paren*{\sR_{\ell_{\rm{sq}}}(h)-\sR^*_{\ell_{\rm{sq}}}(\sH)+\sM_{\ell_{\rm{sq}}}(\sH)}^{\frac12}-\sM_{\ell_{0-1}}(\sH).
\end{equation*}
\end{theorem}
\begin{proof}
For $n = 2$, plugging in $\Phi(t)=(1 - t)^2$ in Theorem~\ref{Thm:char_comp}, gives
\begin{align*}
\sT^{\rm{comp}}
& = \frac{1}{4}-\inf_{\mu\in\bracket*{-\frac12,\frac12}}\curl*{\frac{1 - t}{2}\paren*{\frac12-\mu}^2+\frac{1 + t}{2}\paren*{\frac12+\mu}^2}=\frac{1}{4}-\frac{1-t^2}{4}=\frac{t^2}{4}.
\end{align*}
Similarly, for $n > 2$, plugging in $\Phi(t)=(1-t)^2$ in Theorem~\ref{Thm:char_comp} yields
\begin{align*}
\sT^{\rm{comp}}
&=\inf_{\tau\in\bracket*{\frac1n,\frac12}}\curl*{(1-\tau)^2 -\inf_{\mu\in [-\tau,\tau]}\curl*{\frac{1 + t}{2}(1-\tau+\mu)^2+\frac{1 - t}{2}\paren*{1-\tau-\mu}^2}}\\
&=\inf_{\tau\in\bracket*{\frac1n,\frac12}}\curl*{(1-\tau)^2 -(1-\tau)^2(1-t^2)}
\tag{minimum achieved at $\mu = t(\tau - 1)$}\\
&=\frac{t^2}{4}.\tag{minimum achieved at $\tau = \frac12$}
\end{align*}
By Theorem~\ref{Thm:bound_comp}, the bound obtained is tight, which completes the proof.
\end{proof}

\begin{table}[t]
\caption{$\sH$-estimation error
  transformation for common comp-sum losses.}
  \label{tab:transformation_comp}
\begin{center}
\resizebox{\columnwidth}{!}{
\begin{tabular}{l|l|l|l|l|l}
\toprule
      Auxiliary function $\Phi$ & $-\log(t)$ & $\frac{1}{t}-1$ & $\frac{1}{q}\paren*{1 - t^{q}},q\in (0,1)$ & $1 - t$ & $(1-t)^2$ \\
\midrule
      Transformation $\sT^{\rm{comp}}$ & $\frac{1 + t}{2}\log(1 + t)+\frac{1 - t}{2}\log(1 - t)$ & $1-\sqrt{1 - t^2} $ & $\frac{1}{q n^{q}}\paren*{\frac{\paren*{1 + t}^{\frac1{1-q }} +  \paren*{1 - t}^{\frac1{1-q }}}{2}}^{1-q }-\frac{1}{q n^{q}}$ & $\frac{t}{n}$ & $\frac{t^2}{4}$\\
\bottomrule
\end{tabular}
}
\end{center}
\end{table}

\subsection{Extension to non-complete/bounded hypothesis sets:
  comp-sum losses}
\label{sec:extension-comp}

As pointed out earlier, the hypothesis sets typically used in practice
are bounded.  Let $\sF$ be a family of real-valued functions $f$ with
$|f(x)| \leq \Lambda(x)$ for all $x \in \sX$ and such that all values
in $[-\Lambda(x), +\Lambda(x)]$ can be reached, where $\Lambda(x)> 0$
is a fixed function on $\sX$.  We will study hypothesis sets
$\ov\sH$ in which each scoring function is bounded:
\begin{equation}
\label{eq:bounded-H}
\ov\sH = \curl*{h:\sX\times\sY \to \Rset \mid h(\cdot,y)\in
  \sF, \forall y\in \sY}.
\end{equation}
This holds for most hypothesis sets used in practice.  The symmetric
and complete hypothesis sets studied in previous work correspond to
the special case of $\ov\sH$ where $\Lambda(x) = +\infty$ for all
$x\in \sX$. The hypothesis set of linear models $\sH_{\rm{lin}}$,
defined by
\begin{align*}
  \sH_{\mathrm{lin}}
  = \big\{(x,y)\mapsto w_y \cdot x + b_y \mid
\norm*{w_y}\leq W,\abs*{b_y}\leq B,y\in\sY\big\},
\end{align*}
is also a special instance of $\ov\sH$ where $\Lambda(x) = W\norm*{x}
+ B$.  Let us emphasize that previous studies did not establish any
$\sH$-consistency bound for these general hypothesis sets, $\ov \sH$.

\begin{restatable}[\textbf{$\ov\sH$-consistency bound for comp-sum losses}]
  {theorem}{BoundCompBD}
\label{Thm:bound_comp_BD}
Assume that $\ov\sT^{\rm{comp}}$ is convex.  Then, the following
inequality holds for any hypothesis $h\in\ov\sH$ and any distribution:
\begin{align*}
     \ov\sT^{\rm{comp}}\paren*{\sR_{\ell_{0-1}}(h)-\sR^*_{\ell_{0-1}}(\ov\sH)+\sM_{\ell_{0-1}}(\ov\sH)}\leq \sR_{\ell^{\mathrm{comp}}}(h)-\sR^*_{\ell^{\mathrm{comp}}}(\ov\sH)+\sM_{\ell^{\mathrm{comp}}}(\ov\sH)
\end{align*}
with $\ov\sT^{\rm{comp}}$ the $\ov\sH$-estimation error transformation
for comp-sum losses defined for all $t\in \left[0,1\right]$ by
$\ov\sT^{\rm{comp}}(t)=$
\begin{align*}
\begin{cases}
\inf\limits_{\tau\in \bracket*{0,\frac12}}\sup\limits_{\mu\in \bracket*{s_{\min}-\tau,1-\tau-s_{\min}}} \curl*{\frac{1 + t}{2}\bracket*{\Phi\paren*{\tau}-\Phi\paren*{1-\tau-\mu}} + \frac{1 - t}{2}\bracket*{ \Phi\paren*{1-\tau}-\Phi\paren*{\tau+\mu}}} & n =2 \\
\inf\limits_{\psum\in \bracket*{\frac{1}{n-1}\vee t,1}}\inf\limits_{\substack{S_{\min}\leq \tau_2\leq \tau_1\leq S_{\max}\\ \tau_1+\tau_2\leq 1}}\sup\limits_{\mu\in \C} \curl*{\frac{\psum + t}{2}\bracket*{\Phi\paren*{\tau_2}-\Phi\paren*{\tau_1-\mu}} + \frac{\psum-t}{2}\bracket*{ \Phi\paren*{\tau_1}-\Phi\paren*{\tau_2+\mu}}} & n  > 2,
\end{cases}
\end{align*}
where $\C=\bracket*{\max\curl*{s_{\min}-\tau_2,\tau_1- s_{\max}},\min\curl*{s_{\max}-\tau_2,\tau_1- s_{\min}}}$, $s_{\max}=\frac{1}{1+(n-1)e^{-2\inf_{x}\Lambda(x)}}$ and  $s_{\min}=\frac{1}{1+(n-1)e^{2\inf_{x}\Lambda(x)}}$.
Furthermore, for any $t \in[0,1]$, there exist a distribution $\sD$ and 
$h\in\sH$ such that $\sR_{\ell_{0-1}}(h)-
\sR^*_{\ell_{0-1}}(\sH)+\sM_{\ell_{0-1}}(\sH)=t$ and
$\sR_{\ell^{\mathrm{comp}}}(h) - \sR_{\ell^{\mathrm{comp}}}^*(\sH) + \sM_{\ell^{\mathrm{comp}}}(\sH)=
\sT^{\rm{comp}}(t)$.
\end{restatable}
This theorem significantly broadens the applicability of our framework
as it encompasses bounded hypothesis sets.  The last statement of the
theorem further shows the tightness of the $\sH$-consistency bounds
derived using this error transformation function.  We now illustrate
the application of our theory through several examples.

\textbf{A. Example: logistic loss.}  We first consider the multinomial
logistic loss, that is $\ell^{\rm{comp}}$ with $\Phi(u) = -\log(u)$, for
which we give the following guarantee.

\begin{restatable}[\textbf{$\ov \sH$-consistency bounds for logistic loss}]{theorem}{Logistic}
\label{thm:bound-logistic}
For any $h\in \ov \sH$ and any distribution, we have
\begin{align*}
\sR_{\ell_{0-1}}\paren*{h}-\sR_{\ell_{0-1}}^*\paren*{\ov \sH}+\sM_{\ell_{0-1}}\paren*{\ov \sH}
& \leq \Psi^{-1}
\paren*{\sR_{\ell_{\rm{log}}}\paren*{h}
  - \sR_{\ell_{\rm{log}}}^*\paren*{\ov \sH}+\sM_{\ell_{\rm{log}}}\paren*{\ov \sH}},
\end{align*}
where $\ell_{\rm{log}}
= -\log\paren*{\frac{e^{h(x,y)}}{\sum_{y'\in \sY}e^{h(x,y')}}}$ and  $\Psi(t)=\begin{cases}
 \frac{1 + t}{2}\log(1 + t)+\frac{1 - t}{2}\log(1 - t)& t\leq \frac{s_{\max}-s_{\min}}{s_{\min}+s_{\max}}\\
 \frac{t}{2}\log\paren*{\frac{s_{\max}}{s_{\min}}}+\log\paren*{\frac{2\sqrt{s_{\max}s_{\min}}}{s_{\max}+s_{\min}}}& \mathrm{otherwise}.
\end{cases}$
\end{restatable}
The proof of Theorem~\ref{thm:bound-logistic} is given in
Appendix~\ref{app:bound-logistic}. With the help of some simple
calculations, we can derive a simpler upper bound:
\begin{equation*}
\Psi^{-1}(t)\leq \Gamma(t)=\begin{cases}
 \sqrt{2t} & t\leq \frac{(s_{\max}-s_{\min})^2}{2(s_{\min}+s_{\max})^2}\\
 \frac{2(s_{\min}+s_{\max})}{s_{\max}-s_{\min}} t& \mathrm{otherwise}.
\end{cases}    
\end{equation*}
When the relative difference between $s_{\min}$ and $s_{\max}$ is
small, the coefficient of the linear term in $\Gamma$ explodes. On the
other hand, making that difference large essentially turns $\Gamma$
into a square-root function for all values. In general, $\Lambda$ is
not infinite since a regularization is used, which controls both the
complexity of the hypothesis set and the magnitude of the scores.

\textbf{Comparison with \citep{mao2023cross}.} For the symmetric and
complete hypothesis sets $\sH$ considered in \citep{mao2023cross}, $\Lambda(x) = + \infty$, $s_{\max} = 1$, $s_{\min} = 0$,
$\Psi(t)=\frac{1 + t}{2}\log(1 + t) + \frac{1 - t}{2} \log(1 - t)$ and
$\Gamma(t)= \sqrt{2t}$. By Theorem~\ref{thm:bound-logistic}, this
yields an $\sH$-consistency bound for the logistic loss.

\begin{corollary}
[\textbf{$\sH$-consistency bounds for logistic loss}]
\label{cor:bound-logistic-complete}
Assume that $\sH$ is symmetric and complete. Then, for any $h\in \sH$ and any distribution, we have
\begin{align*}
\sR_{\ell_{0-1}}\paren*{h} - \sR_{\ell_{0-1}}^*\paren*{\sH}
& \leq \Psi^{-1}
\paren*{\sR_{\ell_{\rm{log}}}\paren*{h} - \sR_{\ell_{\rm{log}}}^*\paren*{\sH}+\sM_{\ell_{\rm{log}}}\paren*{\sH}} - \sM_{\ell_{0-1}}\paren*{\sH}
\end{align*}
where $\Psi(t) = \frac{1 + t}{2}\log(1 + t)+\frac{1 - t}{2}\log(1 -
t)$ and $\Psi^{-1}(t)\leq \sqrt{2t}$.
\end{corollary}
Corollary~\ref{cor:bound-logistic-complete} recovers the
$\sH$-consistency bounds of \citet{mao2023cross}.

\textbf{Comparison with \citep{awasthi2022Hconsistency} and
  \citep{zheng2023revisiting}.} For the linear models
$\sH_{\mathrm{lin}}=\big\{(x,y)\mapsto w_y \cdot x + b_y \mid
\norm*{w_y}\leq W,\abs*{b_y} \leq B\big\}$, we have $\Lambda(x) =
W\norm*{x}+B$.  By Theorem~\ref{thm:bound-logistic}, we obtain
$\sH_{\rm{lin}}$-consistency bounds for logistic loss.
\begin{corollary}
[\textbf{$\sH_{\rm{lin}}$-consistency bounds for logistic loss}]
\label{cor:bound-logistic-lin}
For any $h\in \sH_{\rm{lin}}$ and any distribution,
\begin{align*}
\sR_{\ell_{0-1}}\paren*{h}-\sR_{\ell_{0-1}}^*\paren*{\sH_{\rm{lin}}}
&\leq
\Psi^{-1}
\paren*{\sR_{\ell_{\rm{log}}}\paren*{h}-\sR_{\ell_{\rm{log}}}^*\paren*{\sH_{\rm{lin}}}+\sM_{\ell_{\rm{log}}}\paren*{\sH_{\rm{lin}}}}-\sM_{\ell_{0-1}}\paren*{\sH_{\rm{lin}}}
\end{align*}
where
$\Psi(t)
= \begin{cases}
 \frac{1 + t}{2}\log(1 + t)+\frac{1 - t}{2}\log(1 - t) & t\leq  \frac{(n-1)(e^{2B}-e^{-2B})}{2+(n-1)(e^{2B}+e^{-2B})}\\
 \frac{t}{2}\log\paren*{\frac{1+(n-1)e^{2B}}{1+(n-1)e^{-2B}}}+\log\paren*{\frac{2\sqrt{(1+(n-1)e^{2B})(1+(n-1)e^{-2B})}}{2+(n-1)(e^{2B}+e^{-2B})}}& \mathrm{otherwise}.
\end{cases}$
\end{corollary}
For $n = 2$, we have
$\Psi(t) =
\begin{cases}
\frac{t+1}{2}\log(t+1)+\frac{1 - t}{2}\log(1 - t) &  t\leq \frac{e^{2B}-1}{e^{2B}+1}\\
\frac{t}{2}\log\paren*{\frac{1+e^{2B}}{1+e^{-2B}}}+\log\paren*{\frac{2\sqrt{(1+e^{2B})(1+e^{-2B})}}{2+e^{2B}+e^{-2B}}} & \mathrm{otherwise},
\end{cases}$
which coincides with the $\sH_{\rm{lin}}$-estimation error
transformation in \citep{awasthi2022Hconsistency}.  Thus,
Corollary~\ref{cor:bound-logistic-lin} includes as a special case the
$\sH_{\rm{lin}}$-consistency bounds given by
\citet{awasthi2022Hconsistency} for binary classification.

Our bounds of Corollary~\ref{cor:bound-logistic-lin} improves upon the
multi-class $\sH_{\rm{lin}}$-consistency bounds of recent work
\citep[Theorem~3.3]{zheng2023revisiting} in the following ways: i)
their bound holds only for restricted distributions while our bound
holds for any distribution; ii) their bound holds only for restricted
values of the estimation error $\sR_{\ell_{\rm{log}}}\paren*{h} -
\sR_{\ell_{\rm{log}}}^* \paren*{\sH_{\rm{lin}}}$ while ours holds for
any value in $\mathbb{R}$ and more precisely admits a piecewise
functional form; iii) under their distributional assumption, the
minimizability gaps $\sM_{\ell_{0-1}}\paren*{\sH_{\rm{lin}}}$ and
$\sM_{\ell_{\rm{log}}}\paren*{\sH_{\rm{lin}}}$ coincide with the
approximation errors $\sR_{\ell_{0-1}}\paren*{\sH_{\rm{lin}}} -
\sR_{\ell_{0-1}}^*\paren*{\sH_{\rm{all}}}$ and
$\sR_{\ell_{\rm{log}}}\paren*{\sH_{\rm{lin}}} -
\sR_{\ell_{\rm{log}}}^*\paren*{\sH_{\rm{all}}}$ respectively (see the
note before \citep[Appendix~F]{zheng2023revisiting}). Thus, their
bounds can be recovered as an excess error bound
$\sR_{\ell_{0-1}}\paren*{h} -
\sR_{\ell_{0-1}}^*\paren*{\sH_{\rm{all}}} \leq \sqrt{2}
\bracket*{\sR_{\ell_{\rm{log}}}\paren*{h} -
  \sR_{\ell_{\rm{log}}}^*\paren*{\sH_{\rm{all}}}}^{\frac12}$,
which is not specific to the hypothesis set $\sH$ and thus not as
significant.  In contrast, our $\sH_{\rm{lin}}$-consistency bound is
finer and takes into account the role of the parameter $B$ as
well as the number of labels $n$; iv)
\citep[Theorem~3.3]{zheng2023revisiting} only offers approximate
bounds that are not tight; in contrast, by
Theorem~\ref{Thm:bound_comp_BD}, our bound is tight.

Note that our $\sH$-consistency bound in
Theorem~\ref{thm:bound-logistic} are not limited to specific
hypothesis set forms. They are directly applicable to various types of
hypothesis sets including neural networks.  For example, the same
derivation can be extended to one-hidden-layer neural networks studied
in \citep{awasthi2022Hconsistency} and their multi-class
generalization by calculating and substituting the corresponding
$\Lambda(x)$. As a result, we can obtain novel and tight
$\sH$-consistency bounds for bounded neural network hypothesis sets in
multi-class classification, which highlights the versatility of our
general tools.

\textbf{B. Example: sum exponential loss}. We then consider the sum
exponential loss, that is $\ell^{\rm{comp}}$ with
$\Phi(u)=\frac{1-u}{u}$. By computing the error transformation in
Theorem~\ref{Thm:bound_comp_BD}, we obtain the following result.

\begin{restatable}[\textbf{$\ov\sH$-consistency bounds for
      sum exponential loss}]{theorem}{Exponential}
\label{thm:bound-exponential}
For any $h\in \sH$ and any distribution,
\begin{align*}
\sR_{\ell_{0-1}}\paren*{h}-\sR_{\ell_{0-1}}^*\paren*{\ov \sH}+\sM_{\ell_{0-1}}\paren*{\ov \sH}
& \leq \Psi^{-1}
\paren*{\sR_{\ell_{\rm{exp}}}\paren*{h}-\sR_{\ell_{\rm{exp}}}^*\paren*{\ov \sH}+\sM_{\ell_{\rm{exp}}}\paren*{\ov \sH}}
\end{align*}
where $\ell_{\rm{exp}}= \sum_{y'\neq y}e^{h(x,y')-h(x,y)}$ and $\Psi(t)=\begin{cases}
 1-\sqrt{1 - t^2} & t\leq \frac{s_{\max}^2-s_{\min}^2}{s_{\min}^2+s_{\max}^2}\\
 \frac{s_{\max}-s_{\min}}{2s_{\max}s_{\min}}t-\frac{\paren*{s_{\max}-s_{\min}}^2}{2s_{\max}s_{\min}\paren*{s_{\max}+s_{\min}}}& \mathrm{otherwise}.
\end{cases}$.
\end{restatable}
The proof of Theorem~\ref{thm:bound-exponential} is given in
Appendix~\ref{app:bound-exponential}. Observe that $1-\sqrt{1-t^2}\geq
t^2/2$. By Theorem~\ref{thm:bound-exponential}, making $s_{\min}$
close to zero, that is making $\Lambda$ close to infinite for any
$x\in \sX$, essentially turns $\Psi$ into a square function for all
values. In general, $\Lambda$ is not infinite since a regularization
is used in practice, which controls both the complexity of the
hypothesis set and the magnitude of the scores.

\textbf{C. Example: generalized cross-entropy loss and mean absolute
  error loss}. Due to space limitations, we present the results for
these loss functions in Appendix~\ref{app:extension-comp}.

\section{Constrained losses}
\label{sec:cstnd}

In this section, we present a general characterization of
$\sH$-consistency bounds for \emph{constrained loss}, that is loss
functions defined via a constraint, as in
\citep{lee2004multicategory}:
\begin{align}
\label{eq:cstnd_loss}
\ell^{\mathrm{cstnd}}(h, x, y) = \sum_{y'\neq y}\Phi\paren*{-h(x, y')}
\end{align}
with the constraint that $\sum_{y\in \sY}h(x,y) = 0$ for a
non-negative and non-increasing auxiliary function $\Phi$.

\subsection{\texorpdfstring{$\sH$}{H}-consistency bounds}
\label{sec:bounds-cstnd}

As in the previous section, we prove a result that supplies a very
general tool, an \emph{error transformation function} for deriving
$\sH$-consistency bounds for constrained losses. When
$\sT^{\rm{cstnd}}$ is convex, to make these guarantees explicit, we
only need to calculate $\sT^{\rm{cstnd}}$.

\begin{restatable}[\textbf{$\sH$-consistency bound for constrained losses}]
  {theorem}{BoundCstnd}
\label{Thm:bound_cstnd}
Assume that $\sH$ is symmetric and complete. Assume that
$\sT^{\rm{cstnd}}$ is convex.  Then, for any hypothesis $h \in \sH$
and any distribution,
\begin{align*}
     \sT^{\rm{cstnd}}\paren*{\sR_{\ell_{0-1}}(h)-\sR_{\ell_{0-1}}^*(\sH)+\sM_{\ell_{0-1}}(\sH)}\leq \sR_{\ell^{\mathrm{cstnd}}}(h)-\sR_{\ell^{\mathrm{cstnd}}}^*(\sH)+\sM_{\ell^{\mathrm{cstnd}}}(\sH)
\end{align*}
with $\sH$-estimation error transformation for constrained losses defined on $t\in \left[0,1\right]$ by $\sT^{\rm{cstnd}}(t)=$
\begin{align*}
\begin{cases}
\inf\limits_{\tau\geq 0}\sup\limits_{\mu\in \Rset} \curl*{\frac{1 - t}{2}\bracket*{ \Phi\paren*{\tau}-\Phi\paren*{-\tau+\mu}} +  \frac{1 + t}{2}\bracket*{ \Phi\paren*{-\tau}-\Phi\paren*{\tau-\mu}}} & n =2 \\
\inf\limits_{\psum\in \bracket*{\frac{1}{n-1},1}}\inf\limits_{\tau_1\geq \max\curl*{\tau_2,0}}\sup\limits_{\mu\in \Rset} \curl*{\frac{2-\psum-t}{2}\bracket*{ \Phi\paren*{-\tau_2}-\Phi\paren*{-\tau_1+\mu}} +  \frac{2-\psum + t}{2}\bracket*{ \Phi\paren*{-\tau_1}-\Phi\paren*{-\tau_2-\mu}}} & n >2. 
\end{cases}
\end{align*}
Furthermore, for any $t \in[0,1]$, there exist a distribution $\sD$ and a
hypothesis $h \in \sH$ such that $\sR_{\ell_{0-1}}(h)-
\sR^*_{\ell_{0-1}}(\sH)+\sM_{\ell_{0-1}}(\sH)=t$ and
$\sR_{\ell^{\mathrm{cstnd}}}(h) - \sR_{\ell^{\mathrm{cstnd}}}^*(\sH) + \sM_{\ell^{\mathrm{cstnd}}}(\sH)=
\sT^{\rm{cstnd}}(t)$.
\end{restatable}

Here too, the theorem guarantees the tightness of the bound. This
general expression of $\sT^{\rm{cstnd}}$ can be considerably
simplified under some broad assumptions, as shown by the following
result.

\begin{restatable}[\textbf{characterization of $\sT^{\rm{cstnd}}$}]
  {theorem}{CharCstnd}
\label{Thm:char_cstnd}
Assume that $\Phi$ is convex, differentiable at zero and $\Phi'(0)<0$. Then, $\sT^{\rm{cstnd}}$ can be expressed as follows:
\begin{align*}
\sT^{\rm{cstnd}}(t)
&=\begin{cases}
 \Phi(0) -\inf_{\mu\in \mathbb{R}}\curl*{\frac{1 - t}{2}\Phi(\mu)+\frac{1 + t}{2}\Phi(-\mu)} & n=2\\
 \inf_{\tau\geq 0}\curl*{\paren*{2-\frac{1}{n-1}}\Phi(-\tau) -\inf_{\mu\in \mathbb{R}}\curl*{\frac{2-t-\frac{1}{n-1}}{2}\Phi(-\tau+\mu)+\frac{2+t-\frac{1}{n-1}}{2}\Phi(-\tau-\mu)}} & n>2
\end{cases}\\
&\geq 
\begin{cases}
 \Phi(0) -\inf_{\mu\in \mathbb{R}}\curl*{\frac{1 - t}{2}\Phi(\mu)+\frac{1 + t}{2}\Phi(-\mu)} & n=2\\
 \inf_{\tau\geq 0}\curl*{2\Phi(-\tau) -\inf_{\mu\in \mathbb{R}}\curl*{\frac{2-t}{2}\Phi(-\tau+\mu)+\frac{2+t}{2}\Phi(-\tau-\mu)}} & n>2.
\end{cases}
\end{align*}
\end{restatable}
The proof of all the results in this section are given in
Appendix~\ref{app:constrained_losses}.

\textbf{Examples.}
We now compute the $\sH$-estimation error transformation for
constrained losses and present the results in
Table~\ref{tab:transformation_cstnd}. Here, we present the simplified
$\sT^{\rm{cstnd}}$ by using the lower bound in
Theorem~\ref{Thm:char_cstnd}.
Remarkably, by applying Theorem~\ref{Thm:bound_cstnd}, we are able to
obtain tighter $\sH$-consistency bounds for constrained losses with
$\Phi=\Phi_{\rm{hinge}},\Phi_{\rm{sq-hinge}}, \Phi_{\rm{exp}}$ than
those derived using ad hoc methods in \citep{awasthi2022multi}, and a
novel $\sH$-consistency bound for the new constrained loss
$\ell^{\mathrm{cstnd}}(h, x, y) = \sum_{y'\neq y}(1+h(x, y'))^2$ with
$\Phi(t)=(1-t)^2$.

\begin{table}[t]
\caption{$\sH$-estimation error
  transformation for common constrained losses.}
  \label{tab:transformation_cstnd}
\begin{center}
\resizebox{\columnwidth}{!}{
\begin{tabular}{l|l|l|l|l}
\toprule
      Auxiliary function $\Phi$ & $\Phi_{\rm{exp}}(t)=e^{-t}$ & $\Phi_{\rm{hinge}}(t)=\max\curl*{0,1 - t}$ & $\Phi_{\rm{sq-hinge}}(t)=(1 - t)^2 \mathds{1}_{t\leq 1}$ & $\Phi_{\rm{sq}}=(1-t)^2$\\
\midrule
      Transformation $\sT^{\rm{cstnd}}$ & $\sT^{\rm{cstnd}}(t)=2-\sqrt{4-t^2}$ & $\sT^{\rm{cstnd}}(t)=t$ & $\sT^{\rm{cstnd}}(t)=\frac{t^2}{2}$ & $\sT^{\rm{cstnd}}(t)=\frac{t^2}{2}$ \\
\bottomrule
\end{tabular}
}
\end{center}
\end{table}

\subsection{Extension to non-complete or bounded hypothesis sets}
\label{sec:extension_cstnd}

As in the case of comp-sum losses, we extend our results beyond the
completeness assumption adopted in previous work and establish
$\ov\sH$-consistency bounds for bounded hypothesis sets. This
significantly broadens the applicability of our framework.

\begin{restatable}[\textbf{$\ov\sH$-consistency bound for
      constrained losses}]
  {theorem}{BoundCstndBD}
\label{Thm:bound_cstnd_BD}
Assume that $\ov\sT^{\rm{cstnd}}$ is convex.  Then, the following
inequality holds for any hypothesis $h\in\ov\sH$ and any distribution:
\begin{align}
\label{eq:bound_cstnd_nc}
     \ov\sT^{\rm{cstnd}}\paren*{\sR_{\ell_{0-1}}(h) -
       \sR^*_{\ell_{0-1}}(\ov\sH)+\sM_{\ell_{0-1}}(\ov\sH)}\leq
     \sR_{\ell^{\mathrm{cstnd}}}(h) -
     \sR^*_{\ell^{\mathrm{cstnd}}}(\ov\sH)+\sM_{\ell^{\mathrm{cstnd}}}(\ov\sH).
\end{align}
with $\ov\sT^{\rm{cstnd}}$ the $\ov\sH$-estimation error
transformation for constrained losses defined for all $t\in \left[0,
  1\right]$ by $\ov\sT^{\rm{cstnd}}(t)=$
\begin{align*}
\begin{cases}
\inf\limits_{\tau\geq 0}\sup\limits_{\mu\in \bracket*{\tau-\Lambda_{\rm{min}},\tau+\Lambda_{\rm{min}}}} \curl*{\frac{1 - t}{2}\bracket*{ \Phi\paren*{\tau}-\Phi\paren*{-\tau+\mu}} +  \frac{1 + t}{2}\bracket*{ \Phi\paren*{-\tau}-\Phi\paren*{\tau-\mu}}} & n =2 \\
\inf\limits_{\psum\in \bracket*{\frac{1}{n-1},1}}\inf\limits_{\tau_1\geq \max\curl*{\tau_2,0}}\sup\limits_{\mu\in \C} \curl*{\frac{2-\psum-t}{2}\bracket*{ \Phi\paren*{-\tau_2}-\Phi\paren*{-\tau_1+\mu}} +  \frac{2-\psum + t}{2}\bracket*{ \Phi\paren*{-\tau_1}-\Phi\paren*{-\tau_2-\mu}}} & n > 2,
\end{cases}
\end{align*}
where $\C=\bracket*{\max\curl*{\tau_1,-\tau_2}-\Lambda_{\rm{min}},\min\curl*{\tau_1,-\tau_2}+\Lambda_{\rm{min}}}$ and $\Lambda_{\rm{min}}=\inf_{x\in \sX}\Lambda(x)$.
Furthermore, for any $t \in[0,1]$, there exist a distribution $\sD$ and a
hypothesis $h\in\sH$ such that $\sR_{\ell_{0-1}}(h)-
\sR^*_{\ell_{0-1}}(\sH)+\sM_{\ell_{0-1}}(\sH)=t$ and
$\sR_{\ell^{\mathrm{cstnd}}}(h) - \sR_{\ell^{\mathrm{cstnd}}}^*(\sH) + \sM_{\ell^{\mathrm{cstnd}}}(\sH)=
\sT^{\rm{cstnd}}(t)$.
\end{restatable}
The proof is presented in Appendix~\ref{app:bound_cstnd_BD}. Next, we
illustrate the application of our theory through an example of
constrained exponential losses, that is $\ell^{\rm{cstnd}}$ with
$\Phi(t)=e^{-t}$. By using the error transformation in
Theorem~\ref{Thm:bound_cstnd_BD}, we obtain new $\ov \sH$-consistency
bounds in Theorem~\ref{thm:bound-exponential-cstnd} (see
Appendix~\ref{app:bound-exponential-cstnd} for the proof) for bounded hypothesis sets
$\ov \sH$.

\begin{restatable}[\textbf{$\ov\sH$-consistency bounds for
      constrained exponential loss}]{theorem}{ExponentialCstnd}
\label{thm:bound-exponential-cstnd}
Let $\Phi(t)=e^{-t}$. For any $h\in \ov \sH$ and any distribution,
\begin{align*}
\sR_{\ell_{0-1}}\paren*{h}-\sR_{\ell_{0-1}}^*\paren*{\ov \sH}+\sM_{\ell_{0-1}}\paren*{\ov \sH}
& \leq \Psi^{-1}
\paren*{\sR_{\ell^{\rm{cstnd}}}\paren*{h}-\sR_{\ell^{\rm{cstnd}}}^*\paren*{\ov \sH}+\sM_{\ell^{\rm{cstnd}}}\paren*{\ov \sH}}
\end{align*}
where $\Psi(t)=\begin{cases}
 1-\sqrt{1-t^2} & t\leq \frac{e^{2\Lambda_{\rm{min}}}-1}{e^{2\Lambda_{\rm{min}}}+1}\\
\frac{t}{2}\paren*{e^{\Lambda_{\rm{min}}}-e^{-\Lambda_{\rm{min}}}}+\frac{2-e^{\Lambda_{\rm{min}}}-e^{-\Lambda_{\rm{min}}}}{2}& \mathrm{otherwise}.
\end{cases}$.
\end{restatable}
\citet{awasthi2022multi} proves $\sH$-consistency
bounds for constrained exponential losses when $\sH$ is symmetric and
complete. Theorem~\ref{thm:bound-exponential-cstnd} significantly generalizes those results to the non-complete hypothesis sets. Different from the complete case, the functional form of our new bounds has two pieces which corresponds to the linear and the square root convergence respectively, modulo the constants. Furthermore, the coefficient of the linear piece depends on the the magnitude of $\Lambda_{\rm{min}}$. When $\Lambda_{\rm{min}}$ is
small, the coefficient of the linear term in $\Psi^{-1}$ explodes. On the
other hand, making $\Lambda_{\rm{min}}$ large essentially turns $\Psi^{-1}$
into a square-root function.

\section{Discussion}

Here, we further elaborate on the practical value of our tools and
$\sH$-consistency bounds. Our contributions include a more general and
convenient mathematical tool for proving $\sH$-consistency bounds,
along with tighter bounds that enable a better comparison of surrogate
loss functions and extensions beyond previous completeness
assumptions. As mentioned by \citep{awasthi2022multi}, given a
hypothesis set $\sH$, $\sH$-consistency bounds can be used to compare
different surrogate loss functions and select the most favorable one,
which depends on the functional form of the $\sH$-consistency bound;
the smoothness of the surrogate loss and its optimization properties;
approximation properties of the surrogate loss function controlled by
minimizability gaps; and the dependency on the number of classes in
the multiplicative constant. Consequently, a tighter $\sH$-consistency
bound provides a more accurate comparison, as a loose bound might not
adequately capture the full advantage of using one surrogate loss. In
contrast, Bayes-consistency does not take into account the hypothesis
set and is an asymptotic property, thereby failing to guide the
comparison of different surrogate losses.

Another application of our $\sH$-consistency bounds involves deriving
generalization bounds for surrogate loss minimizers
\citep{mao2023cross}, expressed in terms of the same quantities
previously discussed. Therefore, when dealing with finite samples, a
tighter $\sH$-consistency bound could also result in a corresponding
tighter generalization bound. Moreover, our novel results extend
beyond previous completeness assumptions, offering guarantees
applicable to bounded hypothesis sets commonly used with
regularization. This enhancement provides meaningful learning
guarantees. Technically, our error transformation function serves as a
very general tool for deriving $\sH$-consistency bounds with tightness
guarantees. These functions are defined within each class of loss
functions including comp-sum losses and constrained losses, and their
formulation depends on the structure of the individual loss function
class, the range of the hypothesis set and the number of classes. To
derive explicit bounds, all that is needed is to calculate these error
transformation functions. Under some broad assumptions on the
auxiliary function within a loss function, these error transformation
functions can be further distilled into more simplified forms, making
them straightforward to compute.

\section{Conclusion}
\label{sec:conclusion-chcb}

We presented a general characterization and extension of
$\sH$-consistency bounds for multi-class classification. We introduced
new tools for deriving such bounds with tightness guarantees and
illustrated their benefits through several applications and
examples. Our proposed method is a significant advance in the theory
of $\sH$-consistency bounds for multi-class classification. It can
provide a general and powerful tool for deriving tight bounds for a
wide variety of other loss functions and hypothesis sets. We believe
that our work will open up new avenues of research in the field of
multi-class classification consistency.

\chapter{A Universal Growth Rate} \label{ch6}
In this chapter, we present a comprehensive analysis of
the growth rate of $\sH$-consistency bounds for all margin-based
surrogate losses in binary classification, as well as for
\emph{comp-sum losses} and \emph{constrained losses} in multi-class
classification.
We establish a square-root growth rate near zero for margin-based
surrogate losses $\ell$ defined by $\ell(h, x, y) = \Phi(-yh(x))$,
assuming only that $\Phi$ is convex and twice continuously
differentiable with $\Phi'(0) \!>\! 0$ and $\Phi''(0) \!>\! 0$
(Section~\ref{sec:binary}).  This includes both upper and lower bounds
(Theorem~\ref{thm:binary-lower}). These results directly apply to
excess error bounds as well. Importantly, our lower bound requires
weaker conditions than \citep[Theorem~4]{frongillo2021surrogate}, and
our upper bound is entirely novel.  This work demonstrates that the
$\sH$-consistency bound growth rate for these loss functions is
precisely square-root, refining the ``at least square-root'' finding of
these authors (for excess error bounds). It is known that polyhedral
losses admit a linear grow rate \citep{frongillo2021surrogate}. Thus,
a striking dichotomy emerges that reflects previous observations by
these authors: $\sH$-consistency bounds for polyhedral losses exhibit
a linear growth rate in binary classification, while they follow a
square-root rate for smooth loss functions.

Moreover, we significantly extend our findings to key multi-class
surrogate loss families, including \emph{comp-sum losses}
\citep{mao2023cross} (e.g., logistic loss or cross-entropy with
softmax \citep{Berkson1944}, sum-losses \citep{weston1998multi},
generalized cross entropy loss \citep{zhang2018generalized}), and
\emph{constrained losses}
\citep{lee2004multicategory,awasthi2022multi}
(Section~\ref{sec:multi}).
In Section~\ref{sec:comp-srd}, we prove that the growth rate of
$\sH$-consistency bounds for comp-sum losses is exactly
square-root. This applies when the auxiliary function $\Phi$ they are
based upon is convex and twice continuously differentiable with
$\Phi'(u) \!<\! 0$ and $\Phi''(u) \!>\! 0$ for all $u$ in $(0,
  \frac{1}{2}]$. These conditions hold for all common loss functions
used in practice.
Further, in Section~\ref{sec:cstnd-srd}, we demonstrate that the
square-root growth rate also extends to $\sH$-consistency bounds for
constrained losses.  This requires the auxiliary function $\Phi$ to be
convex and twice continuously differentiable with $\Phi'(u) \!>\! 0$
and $\Phi''(u) \!>\! 0$ for any $u \geq 0$, alongside an additional
technical condition.  These are satisfied by all constrained losses
typically encountered in practice.

These results reveal a universal square-root growth rate for smooth
surrogate losses, the predominant choice in neural network training
(over polyhedral losses) for both binary and multi-class
classification in applications.
\ignore{
Crucially, under conditions on
\emph{minimizability gaps} detailed later, this implies a direct
relationship between the surrogate estimation loss and the target
zero-one estimation error: when the surrogate estimation loss is
reduced to a sufficiently small $\e > 0$, the zero-one estimation
error scales precisely as $\sqrt{\e}$. Our results apply to the growth
rate of the first surrogate estimation loss term
$\Gamma\paren*{\sE_{\ell}(h) -\sE^*_{\ell}(\sH)}$ (square-root), with
the second term $\Gamma\paren*{\sM_{\ell}(\sH)}$ being a constant.
}
Given this universal growth rate, how do we choose between different
surrogate losses? Section~\ref{sec:M-gaps} addresses this question
in detail. To start, we examine how $\sH$-consistency bounds vary
across surrogates based on the number of classes. Then, focusing on
behavior near zero (ignoring constants), we isolate minimizability
gaps as the key differentiating factor in these bounds.  These gaps
depend solely on the chosen surrogate loss and hypothesis set.
We provide a detailed analysis of minimizability gaps, covering:
comparisons across different comp-sum losses, conditions where gaps
become zero, and general conditions leading to small gaps.  These
findings help guide surrogate loss selection. Additionally, we
demonstrate the key role of minimizability gaps in comparing excess
error bounds and $\sH$-consistency bounds
(Appendix~\ref{app:excess-bounds}). Importantly, combining
$\sH$-consistency bounds with surrogate loss Rademacher complexity
bounds allows us to derive zero-one loss (estimation) learning bounds
for surrogate loss minimizers
(Appendix~\ref{app:generalization-bound}). We start with the introduction of
necessary concepts and definitions.

The presentation in this chapter is based on \citep{mao2024universal}.

\section{Preliminaries}
\label{sec:pre-srd}

\textbf{Notation and definitions.} We denote the input space by $\sX$
and the label space by $\sY$, a finite set of cardinality $n$ with
elements $\curl*{1, \ldots, n}$.  $\sD$ denotes a distribution over
$\sX \times \sY$.

We write $\sH_{\rm{all}}$ to denote the family of all real-valued
measurable functions defined over $\sX \times \sY$ and denote by $\sH$
a subset, $\sH \subseteq \sH_{\rm{all}}$.  The label assigned by $h
\in \sH$ to an input $x \in \sX$ is denoted by $\hh(x)$ and defined by
$\hh(x) = \argmax_{y \in \sY} h(x, y)$, with an arbitrary but fixed
deterministic strategy used for breaking the ties. For simplicity, we
fix that strategy to be the one selecting the label with the highest
index under the natural ordering of labels.

We will consider general loss functions $\ell \colon \sH \times \sX
\times \sY \to \Rset_{+}$. For many loss functions used in practice, the
loss value at $(x, y)$, $\ell(h, x, y)$, only depends on the value $h$
takes at $x$ and not on its values on other
points. That is, there exists a measurable function $\hat \ell \colon \Rset^n \times \sY \to \Rset_{+}$ such that $\ell(h, x, y) = \hat \ell(h(x), y)$, where $h(x) = \bracket*{h(x, 1), \ldots, h(x, n)}$ is
the score vector of the predictor $h$. We will then say that $\ell$ is a \emph{pointwise loss
function}.
We denote by $\sE_\ell(h)$ the generalization error or expected loss
of a hypothesis $h \in \sH$ and by $\sE^*_\ell(\sH)$ the \emph{best-in
class error}:
$
\sE_\ell(h) = \E_{(x, y) \sim \sD}[\ell(h, x, y)],
\sE^*_\ell(\sH) = \inf_{h \in \sH} \sE_\ell(h).
$
$\sE^*_\ell\paren*{\sH_{\rm{all}}}$ is also
known as the \emph{Bayes error}.
We write $\sfp(y \!\mid\! x) = \sD(Y = y \!\mid\! X = x)$ to denote the conditional
probability of $Y = y$ given $X = x$ and $p(x) = (\sfp(1 \!\mid\! x),
\ldots, \sfp(n \!\mid\! x))$ for the conditional probability vector for any $x \in
\sX$. We denote by $\sC_{\ell}(h, x)$ the \emph{conditional error} of
$h \in \sH$ at a point $x \in \sX$ and by $\sC_{\ell}^*(\sH, x)$ the
\emph{best-in-class conditional error}:
$
\sC_{\ell}(h, x) = \E_y \bracket*{\ell(h, x, y) \mid x} =
\sum_{y \in \sY} \sfp(y \!\mid\! x) \, \ell(h, x, y),
\sC_{\ell}^*(\sH, x) = \inf_{h\in \sH}\sC_{\ell}(h, x),
$
and use the shorthand
$\Delta\sC_{\ell,\sH}(h,x) = \sC_{\ell}(h,x) - \sC_{\ell}^*(\sH, x)$
for the \emph{calibration gap} or \emph{conditional regret} for
$\ell$.
The generalization error of $h$ can be written as $\sE_{\ell}(h) =
\E_{x}\bracket*{\sC_{\ell}(h, x)}$. For convenience, we also
define, for any vector $p = (p_1, \ldots, p_n) \in \Delta^n$, where
$\Delta^n$ is the probability simplex of $\Rset^n$, $\sC_{\ell}(h, x,
p) = \sum_{y\in \sY} p_y \, \ell(h,x,y)$, $\sC_{\ell,\sH}^*(x, p) =
\inf_{h \in \sH} \sC_{\ell}(h, x, p)$ and $\Delta\sC_{\ell, \sH}(h, x,
p) = \sC_{\ell}(h, x, p) -\sC_{\ell,\sH}^*(x, p)$. Thus, we have
$\Delta\sC_{\ell,\sH}(h, x, p(x)) = \Delta\sC_{\ell, \sH}(h, x)$.

We will study the properties of a surrogate loss function $\ell_1$ for
a target loss function $\ell_2$.  In multi-class classification,
$\ell_2$ is typically the zero-one multi-class classification loss
function $\ell_{0-1}$ defined by $\ell_{0-1}(h, x, y) = 1_{\hh(x) \neq
  y}$. Some surrogate loss functions $\ell_1$ include the max losses \citep{crammer2001algorithmic}, comp-sum losses \citep{mao2023cross} and constrained losses \citep{lee2004multicategory}.

\textbf{Binary classification.} The definitions just presented were
given for the general multi-class classification setting.  In the
special case of binary classification (two classes), the standard
formulation and definitions are slightly different. For convenience,
the label space is typically defined as $\sY = \curl*{-1, +1}$.
Instead of two scoring functions, one for each label, a single
real-valued function is used whose sign determines the predicted
class. Thus, here, a hypothesis set $\sH$ is a family of measurable
real-valued functions defined over $\sX$ and $\sH_{\rm all}$ is the
family of all such functions. $\ell$ is \emph{pointwise} if there
exists a measurable function $\hat \ell \colon \Rset \times \sY \to
\Rset_{+}$ such that $\ell(h, x, y) = \hat \ell(h(x), y)$. The target
loss function is typically the binary loss $\ell_{0-1}$, defined by
$\ell_{0-1}(h, x, y) = 1_{\sign(h(x)) \neq y}$, where $\sign(h(x)) =
1_{h(x) \geq 0} - 1_{h(x) < 0}$. Some widely used surrogate losses
$\ell_1$ for $\ell_{0-1}$ are margin-based losses, which are defined
by $\ell_1(h, x, y) = \Phi\paren*{-yh(x)}$, for some non-decreasing
convex function $\Phi\colon \Rset \to \Rset_{+}$. Instead of two
conditional probabilities, one for each label, a single conditional
probability corresponding to the positive class $ +1 $ is used. That
is, let $\eta(x) = \sD(Y = + 1 \mid X = x)$ denote the conditional
probability of $Y = + 1$ given $X = x$. The conditional error can then
be expressed as:
\[
\sC_{\ell}(h,x) = \E_y \bracket*{\ell(h, x, y) \mid x} =
\eta(x) \ell(h, x, +1) + (1 - \eta(x)) \ell(h, x, -1).
\]
For convenience, we also define, for
any $p \in [0, 1]$, $\sC_{\ell}(h, x, p) = p \ell(h, x, +1) + (1 - p)\ell(h, x, -1)$, $\sC^*_{\ell, \sH}(x, p) = \inf_{h \in \sH} \sC_{\ell}(h, x, p)$  and $\Delta\sC_{\ell, \sH}(h, x, p) = \sC_{\ell}(h, x, p) - \inf_{h \in \sH}\sC_{\ell}(h, x, p)$. Thus, we have
$\Delta\sC_{\ell,\sH}(h, x, \eta(x)) = \Delta\sC_{\ell, \sH}(h, x)$.

To simplify matters, we will use the same notation for binary and
multi-class classification, such as $\sY$ for the label space or $\sH$
for a hypothesis set. We rely on the reader to adapt to the
appropriate definitions based on the context.

\textbf{Estimation, approximation, and excess errors.}  For a
hypothesis $h$, the difference
$\sE_{\ell}(h)-\sE_{\ell}^*\paren*{\sH_{\mathrm{all}}}$ is known as 
the \emph{excess error}. It can be decomposed into the sum of two
terms, the \emph{estimation error}, $\paren*{\sE_{\ell}(h) -
  \sE_{\ell}^*(\sH)}$ and the \emph{approximation error}
$\sA_{\ell}(\sH) = \paren*{\sE_{\ell}^*(\sH) -
  \sE_{\ell}^*\paren*{\sH_{\mathrm{all}}}}$:
\begin{equation}
\label{eq:excess-error-decomp} 
\begin{aligned}
    \sE_{\ell}(h) - \sE_{\ell}^*\paren*{\sH_{\mathrm{all}}}
    & = \paren*{\sE_{\ell}(h) - \sE_{\ell}^*(\sH)}
    + \paren*{\sE_{\ell}^*(\sH) - \sE_{\ell}^*\paren*{\sH_{\mathrm{all}}}}.
\end{aligned}
\end{equation}
\ignore{ The approximation error is thus the difference of best-in
  class error and Bayes error. It an be viewed as a measure of
  complexity: the larger the hypothesis set $\sH$, the smaller is
  $\sA_{\ell}(\sH)$.  }
A fundamental result for a pointwise loss function $\ell$ is that the
Bayes error and the approximation error admit the following simpler
expressions.  We give a concise proof of this lemma in
Appendix~\ref{app:lemma}, where we establish the measurability of the
function $x \mapsto \sC^*_{\ell}\paren*{\sH_{\rm{all}}, x}$.
\ignore{
For a pointwise loss function $\ell$, the Bayes error can be expressed in terms of the
expectation of the Bayes conditional error and thus the approximation
error can be rewritten as $\sA_{\ell}(\sH) = \sE^*_\ell(\sH) -
\E_{x}\bracket*{\sC_{\ell}^*(\sH_{\rm{all}},x)}$. The proof is included in Appendix~\ref{app:lemma}.
}
\begin{restatable}{lemma}{ApproximationError}
\label{lemma:approximation-error}
  Let $\ell$ be a pointwise loss function. Then, the Bayes error and
  the approximation error can be expressed as follows:
  $\sE^*_\ell\paren*{\sH_{\rm{all}}} =
  \E_{x}\bracket*{\sC_{\ell}^*(\sH_{\rm{all}},x)}$ and
  $\sA_{\ell}(\sH) = \sE^*_\ell(\sH) -
  \E_{x}\bracket*{\sC_{\ell}^*(\sH_{\rm{all}},x)}$.
\end{restatable}
For restricted hypothesis sets ($\sH\neq \sH_{\rm{all}}$), the
infimum's super-additivity implies that $\sE^*_\ell\paren*{\sH} \geq
\E_x \bracket*{\sC_\ell^*(\sH, x)}$.  This inequality is generally
strict, and the difference, $\sE^*_\ell\paren*{\sH} - \E_x
\bracket*{\sC_\ell^*(\sH, x)}$, plays a crucial role in our analysis.

\ignore{
Note that for a restricted hypothesis set $\sH\neq \sH_{\rm{all}}$, by
the super-additivity of the infimum, $\sE^*_\ell\paren*{\sH}$ is
always lower bounded by $\E_x \bracket*{\sC_\ell^*(\sH,
  x)}$. Moreover, the two terms are in general not equal. The
difference $\sE^*_\ell\paren*{\sH} - \E_x \bracket*{\sC_\ell^*(\sH,
  x)}$ will actually be a key quantity in our analysis.
}

\section{\texorpdfstring{$\sH$}{H}-consistency bounds}
\label{sec:min}

A widely used notion of consistency is that of
\emph{Bayes-consistency} given below
\citep{\ignore{Zhang2003,bartlett2006convexity,}steinwart2007compare}.

\begin{definition}[\textbf{Bayes-consistency}]
A loss function $\ell_1$ is \emph{Bayes-consistent} with respect to a
loss function $\ell_2$, if for any distribution $\sD$ and any sequence
$\{h_n\}_{n\in \Nset} \subset \sH_{\rm{all}}$, $\lim_{n \to +\infty}
\sE_{\ell_1}(h_n) - \sE_{\ell_1}^*\paren*{\sH_{\mathrm{all}}} = 0$
implies $\lim_{n \to +\infty} \sE_{\ell_2}(h_n) -
\sE_{\ell_2}^*\paren*{\sH_{\mathrm{all}}} = 0$.
\end{definition}
Thus, when this property holds, asymptotically, a nearly optimal
minimizer of $\ell_1$ over the family of all measurable functions is
also a nearly optimal optimizer of $\ell_2$.  But, Bayes-consistency
does not supply any information about a hypothesis set $\sH$ not
containing the full family $\sH_{\rm all}$, that is a typical
hypothesis set used for learning.  Furthermore, it is only an
asymptotic property and provides no convergence guarantee. In
particular, it does not give any guarantee for approximate
minimizers. Instead, we will consider upper bounds on the target
estimation error expressed in terms of the surrogate estimation error,
\emph{$\sH$-consistency bounds}
\citep{awasthi2022Hconsistency,awasthi2022multi, mao2023cross},
which account for the hypothesis set $\sH$ adopted.

\begin{definition}[\textbf{$\sH$-consistency bounds}]
Given a hypothesis set $\sH$, an \emph{$\sH$-consistency bound}
relating the loss function $\ell_1$ to the loss function $\ell_2$ for
a hypothesis set $\sH$ is an inequality of the form
\begin{equation}
\label{eq:est-bound-srd}
    \forall h \in \sH, \quad
\sE_{\ell_2}(h) - \sE^*_{\ell_2}(\sH)
+ \sM_{\ell_2}(\sH)
\leq \Gamma\paren*{\sE_{\ell_1}(h)
-\sE^*_{\ell_1}(\sH) + \sM_{\ell_1}(\sH)},
\end{equation}
that holds for any distribution $\sD$, where $\Gamma \colon \Rset_{+}
\to \Rset_{+}$ is a non-decreasing concave function with $\Gamma \geq
0$ \citep{awasthi2022Hconsistency,awasthi2022multi}. Here,
$\sM_{\ell_1}(\sH)$ and $\sM_{\ell_2}(\sH)$ are \emph{minimizability
gaps} for the respective loss functions. The minimizability gap for a
hypothesis set $\sH$ and loss function $\ell$ is denoted by
$\sM_\ell(\sH)$ and defined as: $ \sM_\ell(\sH) =
\sE^*_\ell\paren*{\sH} - \E_x \bracket*{\sC_\ell^*(\sH, x)}$. It
quantifies the discrepancy between the best possible expected loss
within a hypothesis class and the expected infimum of pointwise
expected losses. This gap is always non-negative: $\sM_\ell(\sH) = \inf_{h
  \in \sH} \E_x[\sC_\ell(h, x)] - \E_x [\inf_{h \in \sH} \sC_\ell(\sH,
  x)] \geq 0$, by the infimum's super-additivity, and is bounded above
by the approximation error $\sA_{\ell}(\sH) = \inf_{h \in \sH}
\E_x[\sC_\ell(h, x)] - \E_x [\inf_{h \in \sH_{\rm{all}}} \sC_\ell(\sH,
  x)]$.  We further study the key role of minimizability gaps in
$\sH$-consistency bounds and their properties in
Section~\ref{sec:M-gaps} and Appendix~\ref{app:properties}.  As shown
in Appendix~\ref{app:explicit-form}, under general assumptions,
minimizability gaps are essential quantities required in any bound
that relates the estimation errors of two loss functions with an
arbitrary hypothesis set $\sH$.
\end{definition}
Thus, an $\sH$-consistency bound provides the guarantee that when the
surrogate estimation loss $\sE_{\ell}(h) - \sE^*_{\ell}(\sH)$ is
minimized to $\e$, the following upper bound holds for the zero-one
estimation error:
\begin{align*}
  \sE_{\ell}(h) - \sE^*_{\ell}(\sH)
  \leq \Gamma(\e + \sM_{\ell}(\sH)) - \sM_{\ell_{0-1}}(\sH)
  \leq \Gamma(\e) + \Gamma(\sM_{\ell}(\sH)) - \sM_{\ell_{0-1}}(\sH),
\end{align*}
where the second inequality follows from the sub-additivity of a
concave function $\Gamma$ over $\Rset_+$.  We will demonstrate that,
for smooth surrogate losses, $\Gamma(\e)$ scales as $\sqrt{\e}$.
Note, however, that, while $\Gamma(\e)$ tends to zero when $\e \to 0$
for functions $\Gamma$ derived in $\sH$-consistency bounds, the
remaining terms in the bound are constant. This is not surprising as,
in general, minimizing the surrogate estimation error to zero
\emph{cannot} guarantee that the zero-one estimation error will also
converge to zero.  This is well-known, for example, in the case of
linear models \citep{ben2012minimizing}. Instead, an $\sH$-consistency
bound provides the tightest possible upper bound on the estimation
error for the zero-one loss when the surrogate estimation error is
minimized.

The upper bound simplifies to $\Gamma(\e)$ when the minimizability
gaps are zero, which occurs when either $\sH = \sH_{\rm all}$ (the set
of all measurable functions) or in realizable cases, which are
particularly relevant to the practical use of complex neural networks
in applications.  In Appendix~\ref{app:small-M-gaps}, we examine more
general cases of small minimizability gaps, taking into account the
complexity of $\sH$ and the distribution.

Our results cover in particular the special case of excess bounds
($\sH = \sH_{\rm all}$). Let us emphasize that, for $\sH \neq \sH_{\rm
  all}$, $\sH$-consistency bounds offer tighter and more favorable
guarantees on the estimation error compared to those derived from
excess bounds analysis alone (see Appendix~\ref{app:excess-bounds}).

When $\ell_2 = \ell_{0-1}$, the zero-one loss, we say that
$\sT$ is the \emph{$\sH$-estimation error transformation function
of a surrogate loss $\ell$} if the following holds:
\[
\forall h \in \sH, \quad
\sT \paren*{\sE_{\ell_{0-1}}(h) - \sE^*_{\ell_{0-1}}(\sH)
+ \sM_{\ell_{0-1}}(\sH)}
\leq \sE_{\ell}(h)-\sE^*_{\ell}(\sH) + \sM_{\ell}(\sH),
\]
and the bound is \emph{tight}. That is, for any $t \in [0, 1]$, there
exists a hypothesis $h \in \sH$ and a distribution such that
$\sE_{\ell_{0-1}}(h) - \sE^*_{\ell_{0-1}}(\sH) + \sM_{\ell_{0-1}}(\sH)
= t$ and $\sE_{\ell}(h) - \sE^*_{\ell}(\sH) + \sM_{\ell}(\sH) =
\sT(t)$.
An explicit form of $\sT$ has been characterized for binary
margin-based losses \citep{awasthi2022Hconsistency}, as well as comp-sum losses
and constrained losses in multi-class classification
\citep{MaoMohriZhong2023characterization}. In the following sections,
we will prove the property $\sT(t) = \Theta(t^2)$ (under mild
assumptions), demonstrating a square-root growth rate for
$\sH$-consistency bounds.\ignore{ Crucially, under conditions on
\emph{minimizability gaps} detailed later, this implies a direct
relationship between the surrogate estimation loss and the target
zero-one estimation error: when the surrogate loss is reduced to a
sufficiently small $\e > 0$, the zero-one error scales precisely as
$\sqrt{\e}$.} Appendix~\ref{app:bounds-example} provides
examples of $\sH$-consistency bounds for both binary and multi-class
classification.
Our analysis also suggests choosing appropriately $\sH$ and the
function $\Gamma$ to ensure a small minimizability gap and to take
into account the number of classes and other properties, as discussed
in Section~\ref{sec:M-gaps}.

\section{Binary classification}
\label{sec:binary}

We consider the broad family of margin-based loss functions $\ell$
defined for any $h \in \sH$, and $(x, y) \in \sX \times \sY$ by
$\ell(h, x, y) = \Phi(-y h(x))$, where $\Phi$ is a non-decreasing
convex function upper-bounding the zero-one loss. Margin-based loss
functions include most loss functions used in binary
classification. As an example, $\Phi(u) = \log(1 + e^{u})$ for the
logistic loss or $\Phi(u) = \exp(u)$ for the exponential loss. We say
that a hypothesis set $\sH$ is \emph{complete}, if for all $x \in
\sX$, we have $\curl*{h(x) \colon h \in \sH} = \Rset$. As shown by
\citet{awasthi2022Hconsistency}, the transformation $\sT$ has the following form
for complete hypothesis sets:
\begin{equation*}
\sT(t) \colon = \inf_{u \leq 0} f_t(u) - \inf_{u \in \Rset} f_t(u).
\end{equation*}
Here, for any $t \in [0, 1]$, $f_t$ is defined by:
$\forall u \in \Rset,\, f_t(u) = \frac{1 - t}{2} \Phi(u) + \frac{1 + t}{2} \Phi(-u)$. The following result is useful for proving the growth rate in binary classification.

\begin{theorem}
\label{thm:binary-char}
Let $\sH$ be a complete hypothesis set.
Assume that $\Phi$ is convex and differentiable at zero and satisfies the inequality $\Phi'(0) > 0$. Then, the transformation $\sT$ can be expressed as follows:
\begin{equation*}
\forall t \in [0, 1], \quad \sT(t) = f_t(0) -\inf_{u \in \Rset} f_t(u).
\end{equation*}
\end{theorem}
\begin{proof}
By the convexity of $\Phi$, for any $t \in [0, 1]$ and $u \in \Rset_-$, we have
\[
f_t(u) = \frac{1 - t}{2} \Phi(u) + \frac{1 + t}{2} \Phi(-u)
\geq \Phi(0) - t u \Phi'(0) \geq \Phi(0).
\]
Thus,  we can write
$\sT(t)
= \inf_{u \leq 0} f_t(u) - \inf_{u \in \Rset} f_t(u)
\geq \Phi(0) - \inf_{u \in \Rset} f_t(u) = f_t(0) - \inf_{u \in \Rset} f_t(u)$,
\ignore{
\begin{align*}
  \sT(t)
  = \inf_{u \leq 0} f_t(u) - \inf_{u \in \Rset} f_t(u)
  \geq \Phi(0) - \inf_{u \in \Rset} f_t(u) = f_t(0) - \inf_{u \in \Rset} f_t(u),
  & \geq \inf_{u \leq 0} \paren*{\Phi(0) - t u \Phi'(0)}
  - \inf_{u \in \Rset} f_t(u)\\
  & = \Phi(0) - \inf_{u \in \Rset} f_t(u) = f_t(0) - \inf_{u \in \Rset} f_t(u) \tag{$u < 0$, $\Phi'(0) > 0$},
\end{align*}}
where equality is achieved when $u = 0$.
\end{proof}
\ignore{
As an example,
for the logistics loss, $a^*_t$ is given by $a^*_t = \log \paren*{\frac{1 + t}{1 - t}}$, and for the exponential loss by $a^*_t = \frac{1}{2} \log \paren*{\frac{1 + t}{1 - t}}$, for any $t \in [0, 1]$
}

\begin{restatable}[Upper and lower bound for binary margin-based losses]{theorem}{BinaryLower}
\label{thm:binary-lower}
Let $\sH$ be a complete hypothesis set. Assume that $\Phi$ is convex, twice continuously differentiable, and satisfies the inequalities $\Phi'(0) > 0$ and $\Phi''(0) > 0$. Then, the following property holds: $\sT(t) = \Theta (t^2)$; that is, there exist positive constants $C > 0$, $c > 0$, and $T > 0$ such that $C t^2 \geq \sT(t) \geq c t^2 $, for all $0 < t \leq T$.
\end{restatable}
\noindent \textbf{Proof sketch} First, we demonstrate that, by applying the implicit function theorem, $\inf_{u \in \Rset} f_t(u)$ is attained uniquely by $a^*_t$, and that $a^*_t$ is continuously differentiable over $[0, \e]$ for some $\e > 0$. The minimizer $a^*_t$ satisfies the following condition:
$
f'_t(a^*_t) = \frac{1 - t}{2} \Phi'(a^*_t) - \frac{1 + t}{2} \Phi'(-a^*_t) = 0.
$
Specifically, at $t = 0$, we have $\Phi'(a^*_0) =
\Phi'(-a^*_0)$. Then, by the convexity of $\Phi$ and monotonicity of
the derivative $\Phi'$, we must have $a^*_0 = 0$ and since $\Phi'$ is
non-decreasing and $\Phi''(0) > 0$, we have $a^*_t > 0$ for all $t \in
(0, \e]$. Furthermore, since $a^*_t$ is a function of class $C^1$, we
  can differentiate this condition with respect to $t$ and take the
  limit $t \to 0$, which gives the following equality: $\frac{d
    a_t^*}{d t}(0) = \frac{\Phi'(0)}{\Phi''(0)} > 0$. Since $\lim_{t
    \to 0} \frac{a_t^*}{t} = \frac{d a_t^*}{d t}(0) =
  \frac{\Phi'(0)}{\Phi''(0)} > 0$, we have $a^*_t = \Theta(t)$.  By
  Theorem~\ref{thm:binary-char} and Taylor's theorem with an integral
  remainder, $\sT$ can be expressed as follows: for any $t \in [0,
    \e]$, $\sT(t) = f_t(0) -\inf_{u \in \Rset} f_t(u) = \int_0^{a^*_t}
  u f''_t(u) \, du = \int_0^{a^*_t} u \bracket*{\frac{1 - t}{2}
    \Phi''(u) + \frac{1 + t}{2} \Phi''(-u)} \, du$. Since $\Phi''(0) >
  0$ and $\Phi''$ is continuous, there is a non-empty interval
  $[- \alpha, + \alpha]$ over which $\Phi''$ is
  positive. Since $a^*_0 = 0$ and $a^*_t$ is continuous, there exists
  a sub-interval $[0, \epsilon'] \subseteq [0, \epsilon]$ over which
  $a^*_t \leq \alpha$. Since $\Phi''$ is continuous, it admits a
  minimum and a maximum over any compact set and we can define $c =
  \min_{u \in [-\alpha, \alpha]} \Phi''(u)$ and $C = \max_{u \in
    [-\alpha, \alpha]} \Phi''(u)$. $c$ and $C$ are both positive since
  we have $\Phi''(0) > 0$. Thus, for $t$ in $[0, \epsilon']$, the
  following inequality holds:
$
C \frac{(a^*_t)^2}{2} = \int_0^{a^*_t}  u C \, du \geq \sT(t) = \int_0^{a^*_t} u \bracket*{\frac{1 - t}{2} \Phi''(u) + \frac{1 + t}{2} \Phi''(-u)} \, du
\geq \int_0^{a^*_t}  u c \, du
= c \frac{(a^*_t)^2}{2}.
$
This implies that $\sT(t) = \Theta(t^2)$. The full proof is included in Appendix~\ref{app:binary-lower}.

Theorem~\ref{thm:binary-lower} directly applies to excess error bounds
as well, when $\sH = \sH_{\rm{all}}$. Importantly, our lower bound
requires weaker conditions than
\citep[Theorem~4]{frongillo2021surrogate}, and our upper bound is
entirely novel.  This result demonstrates that the growth rate for
these loss functions is precisely square-root, refining the ``at least
square-root'' finding of these authors. It is known that polyhedral
losses admit a linear grow rate \citep{frongillo2021surrogate}. Thus,
a striking dichotomy emerges: $\sH$-consistency bounds for polyhedral
losses exhibit a linear growth rate, while they follow a square-root
rate for smooth loss functions (see Appendix~\ref{app:poly-smooth} for
a detailed comparison).

\section{Multi-class classification}
\label{sec:multi}

\ignore{Let $\sX$ be the input space and $\sY = \curl*{1, \ldots, n}$
  the label space. We will specifically consider the multi-class case
  where $n > 2$. We denote by $\sH$ a hypothesis set. A hypothesis $h
  \in \sH$ is defined based on a scoring function $h \colon \sX \times
  \sY \to \Rset$.  Let $\ell \colon \sH \times \sX \times \sY \to
  \Rset$ be a loss function. We will specifically consider the target
  multi-class zero-one loss $\ell_{0-1}\colon (h, x, y) \mapsto
  1_{\hh(x) \neq y}$, where $\hh(x) = \argmax_{y \in \sY} h(x, y)$,
  with ties broken by selecting the label with the highest index under
  the natural ordering of labels.}
In this section, we will study two families of surrogate losses in
multi-class classification: comp-sum losses and constrained losses,
defined in Section~\ref{sec:comp-srd} and Section~\ref{sec:cstnd-srd}
respectively. Comp-sum losses and constrained losses are general and
cover all loss functions commonly used in practice. We will consider
any hypothesis set $\sH$ that is \emph{symmetric} and
\emph{complete}. We say that a hypothesis set is \emph{symmetric} when
it does not depend on a specific ordering of the classes, that is,
when there exists a family $\sF$ of functions $f$ mapping from $\sX$
to $\Rset$ such that $\curl*{\bracket*{h(x, 1), \ldots, h(x, n)}
  \colon h \in \sH} = \curl*{\bracket*{f_1(x), \ldots, f_n(x)} \colon
  f_1, \ldots, f_n \in \sF}$, for any $x \in \sX$. We say that a
hypothesis set $\sH$ is \emph{complete} if the set of scores it
generates spans $\Rset$, that is, $\curl*{h(x, y) \colon h \in \sH} =
\Rset$, for any $(x, y) \in \sX \times \sY$.

\ignore{Let $y_{\max}=\argmax_{y\in \sY} p_y$, where we choose the
  label with the highest index under the natural ordering of labels as
  the tie-breaking strategy, as with $\hh(x) = \argmax_{y\in \sY}h(x,
  y)$.}

\subsection{Comp-sum losses}
\label{sec:comp-srd}

Here, we consider comp-sum losses \citep*{mao2023cross}, defined as
\[
\forall h \in \sH, \forall (x, y) \times \sX \times \sY,  \quad
\ell^{\rm{comp}}(h, x, y)
= \Phi \paren*{\frac{e^{ h(x, y)}}{\sum_{y'\in \sY} e^{h(x, y')}}},
\]
where $\Phi \colon \Rset \to \Rset_{+}$ is a non-increasing
function. For example, $\Phi$ can be chosen as the negative log
function $u \mapsto -\log(u)$ for the comp-sum losses, which leads to
the multinomial logistic loss.  As shown by
\citet*{MaoMohriZhong2023characterization}, for symmetric and complete
hypothesis sets, the transformation $\sT$ for the family of comp-sum
losses can be characterized as follows.
\begin{theorem}[{\citet[Theorem~3]{MaoMohriZhong2023characterization}}]
\label{Thm:char_comp-srd}
Let $\sH$ be a symmetric and complete hypothesis set.  Assume that
$\Phi$ is convex, differentiable at $\frac12$ and satisfies the
inequality $\Phi'(\frac12) < 0$. Then, the transformation $\sT$ can be
expressed as
\begin{align*}
  \sT(t) = \inf_{\tau \in \bracket*{\frac1n, \frac12}}
  \sup_{|u| \leq \tau} \curl*{\Phi(\tau) -  \frac{1 - t}{2}\Phi(\tau + u)
    - \frac{1 + t}{2} \Phi \paren*{\tau - u}}.
\end{align*}
\end{theorem}
Next, we will show that as with the binary case, for the comp-sum
losses, the properties $\sT(t) = \Omega(t^2)$ and $\sT(t) = O(t^2)$
hold.\ignore{We first introduce a corollary of
  Theorem~\ref{thm:a_implicit}, which characterizes the optimal
  solution of a class of constrained convex optimization problems.}
We first introduce a generalization of the classical implicit function
theorem where the function takes the value zero over a set of points
parameterized by a compact set. We treat the special case of a
function $F$ defined over $\Rset^3$ and denote by $(t, a, \tau)
\in \Rset^3$ its arguments. The theorem holds more generally for the
arguments being in $\Rset^{n_1} \times \Rset^{n_2} \times \Rset^{n_3}$
and with the condition on the partial derivative being non-zero
replaced with a partial Jacobian being non-singular.

\begin{theorem}[Implicit function theorem with a compact set]
\label{thm:a_implicit}
Let $F\colon \Rset \times \Rset \times \Rset \to \Rset$ be a
continuously differentiable function in a neighborhood of $(0, 0,
\tau)$, for any $\tau$ in a non-empty compact set $\sC$, with $F(0, 0,
\tau) = 0$. Then, if $\frac{\partial F}{\partial a} (0, 0, \tau)$ is
non-zero for all $\tau$ in $\sC$, then, there exist a neighborhood
$\sO$ of $0$ and a unique function $\bar a$ defined over $\sO \times
\sC$ that is continuously differentiable and satisfies
\[
\forall (t, \tau) \in \sO \times \sC,
\quad F(t, \bar a(t, \tau), \tau) = 0.
\]
\end{theorem}
\begin{proof}
  By the implicit function theorem (see for example
  \citep{DontchevRockafellar2009}), for any $\tau \in \sC$, there
  exists an open set $\sU_\tau = (-t_\tau, +t_\tau) \times (\tau -
  \e_\tau, \tau + \e_\tau)$, ($t_\tau > 0$ and $\e_\tau > 0$), and a
  unique function $\bar a_\tau \colon \sU_\tau \to \Rset$ that is in
  $C^1$ and such that for all $(t, \tau) \in \sU_\tau$, $F(t,
  \bar a_\tau(t), \tau) = 0$.

  By the uniqueness of $\bar a_\tau$, for any $\tau \neq \tau'$ and
  $(t_1, \tau_1) \in \sU_\tau \cap \sU_{\tau'}$, we have $\bar
  a_\tau(t_1, \tau_1) = \bar a_{\tau'}(t_1, \tau_1)$. Thus, we can
  define a function $\bar a$ over $\sU = \bigcup_{\tau \in \sC}
  \sU_\tau$ that is of class $C^1$ and such that for any $(t, \tau)
  \in \sU$, $F(t, \bar a(t, \tau), \tau) = 0$.

  Now, $\bigcup_{\tau \in \sC} (\tau - \e_\tau, \tau + \e_\tau)$ is a
  cover of the compact set $\sC$ via open sets. Thus, we can extract
  from it a finite cover $\bigcup_{\tau \in I} (\tau - \e_\tau, \tau +
  \e_\tau)$, for some finite cardinality set $I$. Define $(-t_0, +t_0)
  = \bigcap_{\tau \in I} (-t_\tau, +t_\tau)$, which is a non-empty
  open interval as an intersection of (embedded) open intervals
  containing zero. Then, $\bar a$ is continously differentiable over
  $(-t_0, +t_0) \times \sC$ and for any $(t, \tau) \in (-t_0, +t_0)
  \times \sC$, we have $F(t, \bar a(t, \tau), \tau) = 0$.
\end{proof}

\ignore{
The proof will make use of the following lemma.
\begin{lemma}
  \label{lemma:cont}
  Let $\Phi$ be a non-negative and continuous function defined over
  $\Rset$. Then, for any $t \in [0, 1]$, the function $g$ defined over
  $\Rset_+$ by
  \[
g(\tau) =  \sup_{|u| \leq \tau} \curl*{
  \frac{1 - t}{2} \Phi(\tau + u) + \frac{1 + t}{2} \Phi\paren*{\tau - u}}
\]
is continuous.
  
\end{lemma}
\begin{proof}
  Fix $t \in [0, 1]$. For any $\tau \geq 0$, $g(\tau)$ can be equivalently
  expressed as follows:
\begin{align*}
  g(\tau)
  & =  \sup_{|u| \leq \tau} \curl*{
    \frac{1 - t}{2} \Phi(\tau + u) + \frac{1 + t}{2} \Phi\paren*{\tau - u}}\\
  & =  \sup_{|u| \leq 1} \curl*{
    \frac{1 - t}{2} \Phi\paren*{(1 + u) \tau} + \frac{1 + t}{2} \Phi\paren*{(1 - u)\tau}}.
\end{align*}
Since $\Phi$ is continuous, for any $u \in [-1, +1]$,
$\tau \mapsto \frac{1 - t}{2} \Phi((1 + u) \tau) + \frac{1 + t}{2} \Phi((1 - u)\tau)$ is continuous. Since the supremum over a fixed compact
set of a family of continuous functions is continuous, this shows that
that $g$ is a continuous function.
\end{proof}
}

\begin{restatable}[Upper and lower bound for comp-sum losses]{theorem}{CompLower}
\label{thm:comp-lower}
Assume that $\Phi$ is convex, twice continuously differentiable, and satisfies the properties $\Phi'(u) < 0$ and $\Phi''(u) > 0$ for any $u \in (0, \frac12]$.
Then, the following property holds: 
$\sT(t) = \Theta(t^2)$.
\end{restatable}
\noindent \textbf{Proof sketch}
For any $\tau \in \bracket*{\frac1n, \frac12}$, define the function $\sT_\tau$ by
$ \sT_\tau(t) = f_{t, \tau}(0) - \inf_{|u| \leq \tau} f_{t, \tau}(u),
$
where
$
f_{t, \tau}(u)
= \frac{1 - t}{2} \Phi_{\tau}(u) + \frac{1 + t}{2} \Phi_{\tau}(-u)$, $t \in [0, 1]$ and $
\Phi_{\tau}(u) = \Phi(\tau + u).
$

We aim to establish a lower and upper bound for $\inf_{\tau \in
  \bracket*{\frac1n, \frac12}} \sT_\tau(t)$.  For any fixed $\tau \in
\bracket*{\frac1n, \frac12}$, this situation is parallel to that of
binary classification (Theorem~\ref{thm:binary-char} and
Theorem~\ref{thm:binary-lower}), since we have $\Phi'_{\tau}(0) =
\Phi'(\tau) < 0$ and $\Phi''_{\tau}(0) = \Phi''(\tau) > 0$.  By
Theorem~\ref{thm:a_implicit} and the proof of
Theorem~\ref{thm:binary-lower}, adopting a similar notation, while
incorporating the $\tau$ subscript to distinguish different functions
$\Phi_\tau$ and $f_{t, \tau}$, we can write $ \forall t \in [0,
  t_0],\, \sT_\tau(t) = \int_0^{-a^*_{t, \tau}} u \bracket*{\frac{1 -
    t}{2} \Phi''_{\tau}(-u) + \frac{1 + t}{2} \Phi''_{\tau}(u)} \, du
$, where $a^*_{t, \tau}$ verifies $ a_{0, \tau}^* = 0$ and $
\frac{\partial a_{t, \tau}^*}{\partial t}(0) =
\frac{\Phi'_{\tau}(0)}{\Phi''_{\tau}(0)} = c_\tau < 0.  $ Then, by
further analyzing this equality, we can show the lower bound
$\inf_{\tau \in \bracket*{\frac1n, \frac12}} -a_{t, \tau}^* =
\Omega(t)$ and the upper bound $\sup_{\tau \in \bracket*{\frac1n,
    \frac12}} -a_{t, \tau}^* = O(t)$ for some $t \in [0, t_1]$, $t_1 >
0$. Finally, using the fact that $\Phi''$ reaches its maximum and
minimum over a compact set, we obtain that $ \sT(t) = \inf_{\tau \in
  \bracket*{\frac1n, \frac12}} \sT_\tau(t) = \Theta(t^2)$. The full
proof is included in Appendix~\ref{app:comp-lower}.

Theorem~\ref{thm:comp-lower} significantly extends
Theorem~\ref{thm:binary-lower} to multi-class comp-sum losses, which
include the logistic loss or cross-entropy used with a softmax
activation function. It shows that the growth rate of
$\sH$-consistency bounds for comp-sum losses is exactly square-root,
provided that the auxiliary function $\Phi$ they are based upon is
convex, twice continuously differentiable, and satisfies $\Phi'(u) <
0$ and $\Phi''(u) > 0$ for any $u \in (0, \frac{1}{2}]$, which holds
for most loss functions used in practice.

\subsection{Constrained losses}
\label{sec:cstnd-srd}

Here, we consider constrained losses (see
\citep{lee2004multicategory}), defined as
\[
\forall h \in \sH, \forall (x, y) \times \sX \times \sY,  \quad
\ell^{\mathrm{cstnd}}(h, x, y)
= \sum_{y'\neq y}\Phi\paren*{h(x, y')} \text{ subject to }
\sum_{y\in \sY} h(x, y) = 0,
\]
where $\Phi \colon \Rset \to \Rset_{+}$ is a non-decreasing
function. On possible choice for $\Phi$ is the exponential function.
As shown by \citet{MaoMohriZhong2023characterization}, for symmetric and complete
hypothesis sets, the transformation $\sT$ for the family of
constrained losses can be characterized as follows.

\begin{theorem}[{\citet[Theorem~11]{MaoMohriZhong2023characterization}}]
\label{Thm:char_cstnd-srd}
Let $\sH$ be a symmetric and complete hypothesis set.  Assume that $\Phi$ is convex,
differentiable at zero and satisfies the inequality $\Phi'(0) >
0$. Then, the transformation $\sT$ can be expressed as
\begin{align*}
\sT(t) =
\inf_{\tau \geq 0}\sup_{u \in \Rset}
\curl*{\paren[\Big]{2 - \frac{1}{n-1}} \Phi(\tau)
  - \frac{2 - \frac{1}{n-1} - t}{2} \Phi(\tau + u)
  - \frac{2 - \frac{1}{n-1} + t}{2}\Phi(\tau - u)}.
\end{align*}
\end{theorem}
Next, we will show that for the constrained losses, the properties
$\sT(t) = \Omega(t)$ and $\sT(t) = O(t)$ hold as well. Note that by
Theorem~\ref{Thm:char_cstnd-srd}, we have
\begin{equation*}
  \sT\paren*{\paren[\Big]{2 - \frac{1}{n-1}}t}
  = \paren*{2 - \frac{1}{n-1}} \inf_{\tau \geq 0} \sup_{u \in \Rset}
  \curl*{\Phi(\tau) - \frac{1 - t}{2} \Phi(\tau + u) - \frac{1 + t}{2}\Phi(\tau - u)}.
\end{equation*}
Therefore, to prove $\sT(t) = \Theta(t^2)$, we only need to show 
\begin{align*}
\inf_{\tau \geq 0} \sup_{u \in \Rset} \curl*{\Phi(\tau) - \frac{1 - t}{2} \Phi(\tau + u) - \frac{1 + t}{2}\Phi(\tau - u)} &= \Theta(t^2).
\end{align*}
For simplicity, we assume that the infimum over $\tau \geq 0$ can be
reached within some finite interval $[0, A]$, $A > 0$. This assumption
holds for common choices of $\Phi$, as discussed in
\citep{MaoMohriZhong2023characterization}. Furthermore, as
demonstrated in Appendix~\ref{app:analysis}, under certain conditions
on $\Phi''$, the infimum over $\tau \in [0, A]$ is reached at zero for
sufficiently small values of $t$. For specific examples, see
\citep[Appendix D.3]{MaoMohriZhong2023characterization}, where
$\Phi(t) = e^{t}$ is considered.

\begin{restatable}[Upper and lower bound for constrained losses]{theorem}{CstndLower}
\label{thm:cstnd-lower}
Assume that $\Phi$ is convex, twice continuously differentiable, and
satisfies the properties $\Phi'(u) > 0$ and $\Phi''(u) > 0$ for any $u
\geq 0$.  Then, for any $A > 0$, the following property holds:
\[
\inf_{\tau \in [0, A]} \sup_{u \in \Rset}
\curl*{\Phi(\tau) - \frac{1 - t}{2} \Phi(\tau + u) - \frac{1 + t}{2}\Phi(\tau - u) }
= \Theta(t^2).
\]
\end{restatable}
\noindent \textbf{Proof sketch} For any $\tau \in [0, A]$, define the
function $\sT_\tau$ by $ \sT_\tau(t) = f_{t, \tau}(0) - \inf_{u \in
  \Rset} f_{t, \tau}(u)$, where $f_{t, \tau}(u) = \frac{1 - t}{2}
\Phi_{\tau}(u) + \frac{1 + t}{2} \Phi_{\tau}(-u)$, $t \in [0, 1]$ and
$\Phi_{\tau}(u) = \Phi(\tau + u)$.  We aim to establish a lower and
upper bound for $\inf_{\tau \in [0, A]} \sT_\tau(t)$.  For any fixed
$\tau \in [0, A]$, this situation is parallel to that of binary
classification (Theorem~\ref{thm:binary-char} and
Theorem~\ref{thm:binary-lower}), since we also have $\Phi'_{\tau}(0) =
\Phi'(\tau) > 0$ and $\Phi''_{\tau}(0) = \Phi''(\tau) > 0$.  By
applying Theorem~\ref{thm:a_implicit} and leveraging the proof of
Theorem~\ref{thm:binary-lower}, adopting a similar notation, while
incorporating the $\tau$ subscript to distinguish different functions
$\Phi_\tau$ and $f_{t, \tau}$, we can write $ \forall t \in [0, t_0],
\, \sT_\tau(t) = \int_0^{a^*_{t, \tau}} u \bracket*{\frac{1 - t}{2}
  \Phi''_{\tau}(u) + \frac{1 + t}{2} \Phi''_{\tau}(-u)} \, du, $ where
$a^*_{t, \tau}$ verifies $ a_{0, \tau}^* = 0$ and $ \frac{\partial
  a_{t, \tau}^*}{\partial t}(0) =
\frac{\Phi'_{\tau}(0)}{\Phi''_{\tau}(0)} = c_\tau > 0.  $ Then, by
further analyzing this equality, we can show the lower bound
$\inf_{\tau \in [0, A]} a_{t, \tau}^* = \Omega(t)$ and the upper bound
$\sup_{\tau \in [0, A]} a_{t, \tau}^* = O(t)$ for some $t \in [0,
  t_1]$, $t_1 > 0$. Finally, using the fact that $\Phi''$ reaches its
maximum and minimum over some compact set, we obtain that $ \sT(t) =
\inf_{\tau \in [0, A]} \sT_\tau(t) = \Theta(t^2)$. The full proof is
included in Appendix~\ref{app:cstnd-lower}.

Theorem~\ref{thm:cstnd-lower} significantly expands our findings to
multi-class constrained losses.  It demonstrates that, under some
assumptions, which are commonly satisfied by smooth constrained losses
used in practice, constrained loss $\sH$-consistency bounds also
exhibit a square-root growth rate.

\section{Minimizability gaps}
\label{sec:M-gaps}

As shown in Sections~\ref{sec:binary} and \ref{sec:multi},
$\sH$-consistency bounds for smooth loss functions in both binary and
multi-class classification all admit a square-root growth rate near
zero.  In this section, we start by examining how the number of
classes impacts these bounds. We then turn our attention to the
minimizability gaps, which are the only distinguishing factors between
the bounds.

\subsection{Dependency on number of classes}
  
Even with identical growth rates, surrogate losses can vary in their
$\sH$-consistency bounds due to the number of classes. This factor
becomes crucial to consider when the class count is large.
Consider the family of comp-sum loss functions
$\ell_{\tau}^{\rm{comp}}$ with $\tau
\in [0, 2)$, defined as 
\begin{equation*}
\ell_{\tau}^{\rm{comp}}(h, x, y)
= \Phi^{\tau} \paren*{ \frac{ e^{ h(x, y)}}{\sum_{y'\in \sY} e^{h(x, y') } } } 
=
\begin{cases}
  \frac{1}{1 - \tau}
  \paren*{\bracket*{\sum_{y'\in\sY} e^{{h(x, y') - h(x, y)}}}^{1 - \tau} - 1}
  & \tau \neq 1,  \tau \in [0, 2) \\
\log\paren*{\sum_{y'\in \sY} e^{h(x, y') - h(x, y)}} & \tau = 1,
\end{cases}
\end{equation*}
where $\Phi^{\tau}(u) = -\log (u) 1_{\tau = 1} + \frac{1}{1 - \tau}
\paren*{u^{\tau - 1} - 1} 1_{\tau \neq 1}$, for any $\tau \in [0,
  2)$. \citet[Eq.~(7) \& Theorem~3.1]{mao2023cross}, established
  the following bound for any $h \in \sH$ and $\tau \in [1, 2)$,
\begin{align*}
\sR_{\ell_{0-1}}(h) - \sR_{\ell_{0-1}}^*(\sH)
\leq \wt \Gamma_{\tau}
  \paren*{\sR_{\ell_{\tau}^{\rm{comp}}}(h) - \sR_{\ell_{\tau}^{\rm{comp}}}^*(\sH)
    + \sM_{\ell_{\tau}^{\rm{comp}}}(\sH)}
- \sM_{\ell_{0-1}}(\sH),
\end{align*}
where $\wt \Gamma_{\tau}(t) = \sqrt{2n^{\tau-1} t}$.  Thus, while all
these loss functions show square-root growth, the number of classes
acts as a critical scaling factor.

\subsection{Comparison across comp-sum losses}

In Appendix~\ref{app:M-gaps-comp}, we compare minimizability gaps
cross comp-sum losses.  We will see that minimizability gaps decrease
as $\tau$ increases. This might suggest favoring $\tau$ close to $2$.
But when accounting for $n$, $\ell_{\tau}^{\rm{comp}}$ with $\tau = 1$
(logistic loss) is optimal since $n$ then vanishes. Thus, both class
count and minimizability gaps are essential in loss selection. In
Appendix~\ref{app:small-M-gaps-multi}, we will show that the
minimizability gaps can become zero or relatively small under certain
conditions in multi-class classification.  In such scenarios,
the logistic loss is favored,
which can partly explain its widespread practical application.

\subsection{Small surrogate minimizability gaps}
\label{sec:small-M-gaps}

While minimizability gaps vanish in special scenarios (e.g.,
unrestricted hypothesis sets, best-in-class error matching Bayes
error), we now seek broader conditions for zero or small surrogate
minimizability gaps to make our bounds more meaningful.

Due to space constraints, we focus on binary classification here, with
multi-class results given in Appendix~\ref{app:small-M-gaps-multi}. We
address pointwise surrogate losses which take the form $\ell(h(x), y)$
for a labeled point $(x, y)$.
We write $A = \curl*{h(x) \colon h \in \sH}$ to denote the set of
predictor values at $x$, which we assume to be independent of
$x$. All proofs for this section are presented in
Appendix~\ref{app:small-M-gaps}.

\ignore{
We have seen that the minimizability gaps vanish in some special cases
where the hypothesis set is the family of all measurable functions or
the best-in-class error coincides with the Bayes error. To make the
bounds more significant, we now further analyze these quantities and
examine general conditions under which the surrogate minimizability
gap is zero or small.

Due to space limitations, we only presents results for binary
classification in this section. Corresponding results for multi-class
classification are provided in
Appendix~\ref{app:small-M-gaps-multi}. We consider pointwise surrogate
losses, for which the loss of a predictor $h$ at a labeled point $(x,
y)$ can be expressed by $\ell(h(x), y)$.
We denote by $A$ the set of values taken by predictors in $\sH$ at
$x$, which we assume to be independent of $x$: $A = \curl*{h(x) \colon
  h \in \sH}$, for all $x \in \sX$.  The proof of results in this
section are included in Appendix~\ref{app:small-M-gaps}.
}

\textbf{Deterministic scenario}. We first consider the deterministic
scenario, where the conditional probability $\sfp(y \!\mid\! x)$ is either zero or
one. For a deterministic distribution, we denote by $\sX_+$ the subset
of $\sX$ over which the label is $+1$ and by $\sX_-$ the subset of
$\sX$ over which the label is $-1$. For convenience, let $\ell_+ =
\inf_{\alpha \in A} \ell(\alpha, +1)$ and $\ell_- = \inf_{\alpha \in
  A} \ell(\alpha, -1)$.

\begin{restatable}{theorem}{ZeroMinGap}
\label{th:ZeroMinGap}
Assume that $\sD$ is deterministic and that the best-in-class error is
achieved by some $h^* \in \sH$. Then, the minimizability gap is null,
$\sM(\sH) = 0$, iff
\begin{align*}
  \ell(h^*(x), +1)  = \ell_+ \text{ a.s.\ over $\sX_+$}, \quad
  \ell(h^*(x), -1)  = \ell_- \text{ a.s.\ over $\sX_-$}.
\end{align*}
If further $\alpha \mapsto \ell(\alpha, +1)$ and $\alpha \mapsto
\ell(\alpha, -1)$ are injective and $\ell_+ = \ell(\alpha_+, +1)$,
$\ell_- = \ell(\alpha_-, -1)$, then, the condition is equivalent to
$h^*(x) = \alpha_+ 1_{x \in \sX_+} + \alpha_- 1_{x \in \sX_-}$
Furthermore, the minimizability gap is bounded by $\e$ iff $p
\paren*{\E \bracket*{\ell(h^*(x), +1) \mid y = +1} - \ell_+ } + (1 -
p) \paren*{\E\bracket*{\ell(h^*(x), -1) \mid y = -1} - \ell_-} \leq \e
$. In particular, the condition implies:
\begin{align*}
  & \E \bracket*{\ell(h^*(x), +1) \mid y = +1} - \ell_{+} \leq \frac{\e}{p}
  \quad \text{and} \quad
  \E \bracket*{\ell(h^*(x), -1) \mid y = -1} - \ell_- \leq \frac{\e}{1 - p}.
\end{align*}
\end{restatable}

\setlength{\intextsep}{0pt}
\setlength{\columnsep}{10pt}
\begin{wrapfigure}{r}{0.24\textwidth}
  \includegraphics[width=0.24\textwidth]{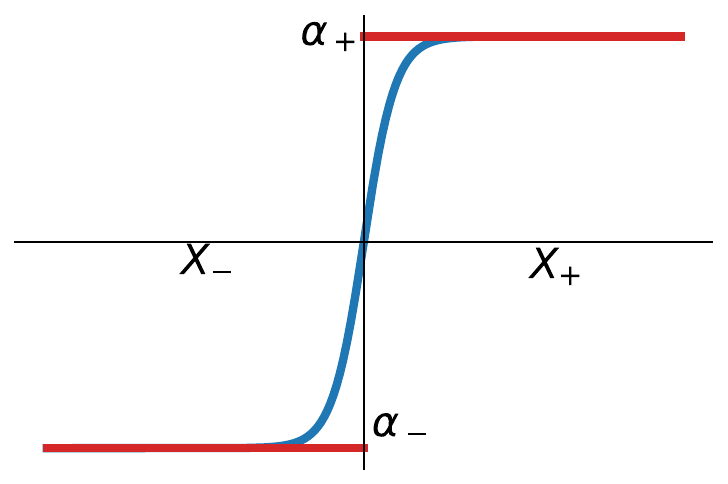}
\captionsetup{format=plain}
\caption{Approximation provided by sigmoid activation function.}
\label{fig:illustration}
\end{wrapfigure}
The theorem suggests that, under those assumptions, for the surrogate
minimizability gap to be zero, the best-in-class hypothesis must be
piecewise constant with specific values on $\sX_+$ and $\sX_-$. The
existence of such a hypothesis in $\sH$ depends both on the complexity
of the decision surface separating $\sX_+$ and $\sX_-$ and on that of
the hypothesis set $\sH$.  More generally, when the best-in-class
classifier $\e$-approximates $\alpha_+$ over $\sX_+$ and $\alpha_-$
over $\sX_-$, then the minimizability gap is bounded by
$\e$.\ignore{The existence of such a hypothesis in $\sH$ depends on
  the complexity of the decision surface.}  As an example, when the
decision surface is a hyperplane, a hypothesis set of linear functions
combined with a sigmoid activation function can provide such a good
approximation (see Figure~\ref{fig:illustration} for an illustration
in a simple case).

\textbf{Stochastic scenario}. Here, we present a general result that
is a direct extension of that of the deterministic scenario.  We show
that the minimizability gap is zero when there exists $h^*\in \sH$
that matches $\alpha^*(x)$ for all $x$, where $\alpha^*(x)$ is the
minimizer of the conditional error.  We also show that the
minimizability gap is bounded by $\e$ when there exists $h^*\in \sH$
whose conditional error $\e$-approximates best-in-class conditional
error for all $x$.

\begin{restatable}{theorem}{ZeroMinGapStochastic}
\label{th:ZeroMinGapStochastic}
  The best-in-class error is achieved by some $h^*\in \sH$ and the
  minimizability gap is null, $\sM(\sH) = 0$, iff there exists $h^*\in
  \sH$ such that for all $x$,
  \begin{align}
  \label{eq:cond-zero-stochastic}
  \E_{y}[\ell(h^*(x), y) \mid x]
  = \inf_{\alpha \in A} \E_{y}[\ell(\alpha, y) \mid x] \text{ a.s.\ over $\sX$}.
  \end{align}
  If further $\alpha \mapsto \E_{y}[\ell(\alpha, y) \mid x]$ is injective
  and $\inf_{\alpha \in A} \E_{y}[\ell(\alpha, y) \mid x] = \E_{y}[\ell(\alpha^*(x), y) \mid x]$, then, the condition is equivalent to
  $h^*(x) = \alpha^*(x) \text{ a.s.\ for $x \in \sX$}$. Furthermore,
  the minimizability gap is bounded by $\e$, $\sM(\sH) \leq \e$, iff
  there exists $h^*\in \sH$ such that
\begin{align}
\label{eq:cond-epsilon-stochastic}
\E_{x}\bracket*{\E_{y}[\ell(h^*(x), y) \mid x]
  - \inf_{\alpha \in A} \E_{y}[\ell(\alpha, y) \mid x]} \leq \e.
\end{align}
\end{restatable}
In deterministic settings,
condition~\eqref{eq:cond-epsilon-stochastic} coincides with that of
Theorem~\ref{th:ZeroMinGap}. However, in stochastic scenarios, the
existence of such a hypothesis depends on both decision surface
complexity and the conditional distribution's properties.  For
illustration, see Appendix~\ref{app:examples} where we analyze
the exponential, logistic (binary), and multi-class logistic losses.

We thoroughly analyzed minimizability gaps, comparing them across
comp-sum losses, and identifying conditions for zero or small gaps,
which help inform surrogate loss selection.
In Appendix~\ref{app:excess-bounds}, we show the crucial role of
minimizability gaps in comparing excess bounds with $\sH$-consistency
bounds.  Importantly, combining $\sH$-consistency bounds with
surrogate loss Rademacher complexity bounds yields zero-one loss
(estimation) learning bounds for surrogate loss minimizers (see
Appendix~\ref{app:generalization-bound}).

\section{Conclusion}

We established a universal square-root growth rate for the widely-used
class of smooth surrogate losses in both binary and multi-class
classification. This underscores the minimizability gap as a
crucial discriminator among surrogate losses. Our detailed analysis of
these gaps can provide guidance for loss selection.

\chapter*{Conclusion} \label{chp-conclusion}
\addcontentsline{toc}{chapter}{Conclusion}

In this thesis, we considered upper bounds on the target estimation error
expressed in terms of the surrogate estimation error, which we referred
to as $\sH$-consistency bounds, since they
account for the hypothesis set $\sH$ adopted. These guarantees are
significantly stronger than Bayes-consistency, $\sH$-calibration or $\sH$-consistency. They are also more informative than
\emph{excess error bounds} derived in the literature, which correspond
to the special case where $\sH$ is the family of all measurable
functions.

We presented an exhaustive study of $\sH$-consistency
bounds for binary classification, including a series of new guarantees for both the
non-adversarial zero-one loss function and the adversarial zero-one
loss function. Our hypothesis-dependent guarantees are significantly
stronger than the consistency or calibration ones.

We also presented a comprehensive study of $\sH$-consistency bounds for
multi-class classification, including the analysis of the three most
commonly used families of multi-class surrogate losses (max losses,
sum losses and constrained losses) and including the study of
surrogate losses for the adversarial robustness. Our theoretical
analysis helps determine which surrogate losses admit a favorable
guarantee for a given hypothesis set $\sH$. Our bounds can help guide
the design of multi-class classification algorithms for both the
adversarial and non-adversarial settings. They also help compare
different surrogate losses for the same setting and the same hypothesis
set. Of course, in addition to the functional form of the
$\sH$-consistency bound, the approximation property of a surrogate
loss function combined with the hypothesis set plays an important
role.

We further presented a detailed analysis of the theoretical properties of a
family of surrogate losses that includes the logistic loss (or
cross-entropy with the softmax). These are more precise and more
informative guarantees than Bayes consistency since they are
non-asymptotic and specific to the hypothesis set used. Our bounds are
tight and can be made more explicit, when combined with our analysis
of minimizability gaps. These inequalities can help compare different
surrogate losses and evaluate their advantages in different scenarios.
We showcased one application of this analysis by extending comp-sum
losses to the adversarial robustness setting, which yields principled
surrogate losses and algorithms for that scenario. We believe that our
analysis can be helpful to the design of algorithms in many other
scenarios.

Next, we presented a general characterization and extension of
$\sH$-consistency bounds for multi-class classification. We introduced
new tools for deriving such bounds with tightness guarantees and
illustrated their benefits through several applications and
examples. Our proposed method is a significant advance in the theory
of $\sH$-consistency bounds for multi-class classification. It can
provide a general and powerful tool for deriving tight bounds for a
wide variety of other loss functions and hypothesis sets. We believe
that our work will open up new avenues of research in the field of
multi-class classification consistency.

Finally, we established a universal square-root growth rate for the widely-used
class of smooth surrogate losses in both binary and multi-class
classification. This underscores the minimizability gap as a
crucial discriminator among surrogate losses. Our detailed analysis of
these gaps can provide guidance for loss selection.

This thesis
includes a series of theoretical and conceptual tools helpful for the
analysis of other loss functions and other hypothesis sets in other learning scenarios, including 
ranking \citep{MaoMohriZhong2023ranking, MaoMohriZhong2023rankingabs},
regression \citep{mao2024h}, learning with abstentions 
\citep{MaoMohriZhong2023score,MaoMohriZhong2023predictor}, structured prediction \citep{MaoMohriZhong2023structured}, learning with deferral
\citep{MaoMohriZhong2023deferral,MaoMohriMohriZhong2023twostage,mao2024regression,mao2024realizable,mao2025theory,MaoMohriZhong2025mastering,desalvo2025budgeted}, top-$k$ classification
\citep{cortes2024cardinality}, multi-label learning
\citep{mao2024multi}, imbalanced learning \citep{cortes2025balancing,cortes2025improved}, optimization of generalized metrics \citep{MaoMohriZhong2025principled}, application in large language models (LLMs) \citep{MohriAndorChoiCollinsMaoZhong2023learning}, etc. They can be extended to include distribution dependence, providing more favorable distribution- and predictor-dependent bounds \citep{mao2025enhanced,mohri2025beyond}, and can be further extended to the analysis of non-i.i.d. settings, such as that of drifting distributions \citep{HelmboldLong1994,Long1999,BarveLong1997,BartlettBenDavidKulkarni2000,
MohriMunozMedina2012,Gama2014} or, more generally, time series prediction \citep{Engle1982,Bollerslev1986,BrockwellDavis1986,BoxJenkins1990,
Hamilton1994,Meir2000,KuznetsovMohri2015,KuznetsovMohri2017,KuznetsovMohri2020}. Our results can also be extended to many other loss functions, using our general proof techniques or a similar analysis.

Theoretically, the general tools and results presented in this thesis can be used or extended to address the questions: \emph{Which surrogate losses should be used, and which benefit from theoretical guarantees specific to the hypothesis set used?} This includes providing $\sH$-consistency bounds or demonstrating that no non-trivial $\sH$-consistency bound can be established in some cases for existing surrogate loss functions across a variety of learning tasks. Our theoretical analysis helps determine which surrogate losses admit favorable guarantees for a given hypothesis set $\sH$. It also helps compare
different surrogate losses for the same setting and the same hypothesis
set $\sH$. In this thesis, we explored these questions within the scenarios of binary and multi-class classification. Other scenarios, such as pairwise ranking, bounded regression, ordinal regression, learning with noisy labels, learning from label proportions, and, more broadly, weakly supervised learning, remain intriguing avenues for further investigation.

Algorithmically, the general tools and results presented in this thesis can be used or extended to address the questions: \emph{Can we design surrogate losses that benefit from $\sH$-consistency bounds? Can such loss functions be used to design effective algorithms?} Our $\sH$-consistency bounds can guide the design of algorithms in various scenarios.  The resulting algorithms can establish themselves as the first to achieve $\sH$-consistency in their respective scenarios and have the potential to surpass the current state-of-the-art in comprehensive empirical analyses. In this thesis, we explored these questions within the scenario of adversarial robustness. Other scenarios, such as learning with abstentions, structured prediction, learning with deferral, top-$k$ classification, multi-label learning, time series prediction, learning from imbalanced data, general cost-sensitive learning, and others, remain intriguing avenues for further investigation.


\appendix

\chapter{Appendix to Chapter~\ref{ch2}}

\disableatoc
\section{Deferred Tables}
\label{app:table}

\begin{table}[ht]
\caption{Non-adversarial $\sH_{\mathrm{lin}}$-estimation error
  transformation ($\epsilon=0)$ and $\sH_{\mathrm{lin}}$-consistency
  bounds. All the bounds are hypothesis set-dependent (parameter $B$ in $\sH_{\mathrm{lin}}$) and provide
  novel guarantees as discussed in Section~\ref{sec:non-adv-lin}. The
  minimizability gaps appearing in the bounds for the surrogates are
  concluded in Table~\ref{tab:loss}. The detailed derivation is
  included in Appendix~\ref{app:derivation-lin}.}
    \label{tab:compare_inverse}
\begin{center}
    \resizebox{\columnwidth}{!}{
    \begin{tabular}{l|lll}
    \toprule
      Surrogates & $\sT_{\Phi}(t),\, t\in [0,1]$   & $\sT_{\Phi}^{-1}(t),\, t
      \in\Rset_{+}$ & Bound \\
    \midrule
      Hinge & $\min \curl*{B, 1} \, t $  & $\frac{t}{\min\curl*{B, 1}}$ & \eqref{eq:hinge-lin-est} \\
      Logistic & $\begin{cases}
\frac{t+1}{2}\log_2(t+1)+\frac{1-t}{2}\log_2(1-t),\quad &  t\leq \frac{e^B-1}{e^{B}+1},\\
1-\frac{t+1}{2}\log_2(1+e^{-B})-\frac{1-t}{2}\log_2(1+e^B),\quad & t> \frac{e^B-1}{e^B+1}.
\end{cases}$ & upper-bounded by $\begin{cases}
\sqrt{2t}, & t\leq \frac{1}{2}\paren*{\frac{e^B-1}{e^B+1}}^2,\\
2\paren*{\frac{e^B+1}{e^B-1}}\, t, & t> \frac{1}{2}\paren*{\frac{e^B-1}{e^B+1}}^2.
\end{cases}$ & \eqref{eq:log-lin-est}\\
      Exponential & $\begin{cases}
1-\sqrt{1-t^2}, & t\leq \frac{e^{2B}-1}{e^{2B}+1},\\
1-\frac{t+1}{2}e^{-B}-\frac{1-t}{2}e^B, & t> \frac{e^{2B}-1}{e^{2B}+1}.
\end{cases}$ & upper-bounded by $\begin{cases}
\sqrt{2t}, & t\leq \frac{1}{2}\paren*{\frac{e^{2B}-1}{e^{2B}+1}}^2,\\
2\paren*{\frac{e^{2B}+1}{e^{2B}-1}}\, t, & t> \frac{1}{2}\paren*{\frac{e^{2B}-1}{e^{2B}+1}}^2.
\end{cases}$ & \eqref{eq:exp-lin-est}\\
      Quadratic & $\begin{cases}
t^2, & t\leq B,\\
2B \,t-B^2, & t> B.
\end{cases}$ & $\begin{cases}
\sqrt{t}, & t \leq B^2, \\
\frac{t}{2B}+\frac{B}{2}, & t > B^2.
\end{cases}$ & \eqref{eq:quad-lin-est}\\
      Sigmoid & $\tanh(kB) \, t$ & $\frac{t}{\tanh(kB)}$ & \eqref{eq:sig-lin-est}\\
      $\rho$-Margin & $\frac{\min\curl*{B,\rho}}{\rho} \, t$ & $\frac{\rho }{\min\curl*{B,\rho}} \, t$ & \eqref{eq:rho-lin-est}\\
    \bottomrule
    \end{tabular}}
    \end{center}
\end{table}

\begin{table}[ht]
    \caption{Non-adversarial $\sH_{\mathrm{NN}}$-estimation error
      transformation ($\epsilon=0)$ and
      $\sH_{\mathrm{NN}}$-consistency bounds. All the
      bounds are hypothesis set-dependent (parameter $\Lambda$
      and $B$ in $\sH_{\mathrm{NN}}$) and provide novel guarantees as
      discussed in Section~\ref{sec:non-adv-NN}. The minimizability
      gaps appearing in the bounds for the surrogates are concluded in
      Table~\ref{tab:loss}. The detailed derivation is included in
      Appendix~\ref{app:derivation-NN}.}
    \label{tab:compare_inverse-NN}
\begin{center}
    \resizebox{\columnwidth}{!}{
    \begin{tabular}{l|lll}
    \toprule
      Surrogates & $\sT_{\Phi}(t),\, t\in [0,1]$   & $\sT_{\Phi}^{-1}(t),\, t
      \in\Rset_{+}$ & Bound \\
    \midrule
      Hinge & $\min \curl*{\Lambda B, 1} \, t $  & $\frac{t}{\min\curl*{\Lambda B, 1}}$ & \eqref{eq:hinge-NN-est} \\
      Logistic & $\begin{cases}
\frac{t+1}{2}\log_2(t+1)+\frac{1-t}{2}\log_2(1-t),\quad & t\leq \frac{e^{\Lambda B}-1}{e^{\Lambda B}+1},\\
1-\frac{t+1}{2}\log_2(1+e^{-\Lambda B})-\frac{1-t}{2}\log_2(1+e^{\Lambda B}),\quad & t> \frac{e^{\Lambda B}-1}{e^{\Lambda B}+1}.
\end{cases}$ & upper-bounded by $\begin{cases}
\sqrt{2t}, & t\leq \frac{1}{2}\paren*{\frac{e^{\Lambda B}-1}{e^{\Lambda B}+1}}^2,\\
2\paren*{\frac{e^{\Lambda B}+1}{e^{\Lambda B}-1}}\, t, & t> \frac{1}{2}\paren*{\frac{e^{\Lambda B}-1}{e^{\Lambda B}+1}}^2.
\end{cases}$ & \eqref{eq:log-NN-est}\\
      Exponential & $\begin{cases}
1-\sqrt{1-t^2}, & t\leq \frac{e^{2\Lambda B}-1}{e^{2\Lambda B}+1},\\
1-\frac{t+1}{2}e^{-\Lambda B}-\frac{1-t}{2}e^{\Lambda B}, & t> \frac{e^{2\Lambda B}-1}{e^{2\Lambda B}+1}.
\end{cases}$ & upper-bounded by $\begin{cases}
\sqrt{2t}, & t\leq \frac{1}{2}\paren*{\frac{e^{2\Lambda B}-1}{e^{2B}+1}}^2,\\
2\paren*{\frac{e^{2\Lambda B}+1}{e^{2\Lambda B}-1}}\, t, & t> \frac{1}{2}\paren*{\frac{e^{2\Lambda B}-1}{e^{2\Lambda B}+1}}^2.
\end{cases}$ & \eqref{eq:exp-NN-est}\\
      Quadratic & $\begin{cases}
t^2,~t\leq \Lambda B,\\
2\Lambda B t-(\Lambda B)^2,~t> \Lambda B.
\end{cases}$ & $\begin{cases}
\sqrt{t}, & t \leq (\Lambda B)^2 \\
\frac{t}{2\Lambda B}+\frac{\Lambda B}{2}, & t > (\Lambda B)^2
\end{cases}$ & \eqref{eq:quad-NN-est}\\
      Sigmoid & $\tanh(k\Lambda B) \, t$ & $\frac{t}{\tanh(k\Lambda B)}$ & \eqref{eq:sig-NN-est}\\
      $\rho$-Margin & $\frac{\min\curl*{\Lambda B,\rho}}{\rho} \, t$ & $\frac{\rho }{\min\curl*{\Lambda B,\rho}} \, t$ & \eqref{eq:rho-NN-est}\\
    \bottomrule
    \end{tabular}
    }
    \end{center}
\end{table}


\newpage
\section{Deferred Theorems}
\label{app:theorems}

\begin{theorem}[\textbf{Non-adversarial distribution-dependent $\Psi$-bound}]
\label{Thm:excess_bounds_Psi_01_general}
Suppose that $\sH$ satisfies the condition of
Lemma~\ref{lemma:explicit_assumption_01} and that $\Phi$ is a
margin-based loss function. Assume there exist a convex function
$\Psi\colon \mathbb{R_{+}} \to \Rset$ with $\Psi(0)=0$ and
$\epsilon\geq0$ such that the following holds for any $x\in \sX$:
\begin{align}
\label{eq:condition_Psi_general}
    \Psi\paren*{\tri*{2 \abs*{\Delta \eta(x)}}_{\e}}
    \leq \inf_{h \in \ov \sH(x)}\Delta\sC_{\Phi,\sH}(h,x).
\end{align}
Then, for any hypothesis $h \in \sH$,
\begin{equation}
     \Psi\paren*{\sR_{\ell_{0-1}}(h)- \sR_{\ell_{0-1}}^*(\sH)+\sM_{\ell_{0-1}}(\sH)}
     \leq  \sR_{\Phi}(h)-\sR_{\Phi}^*(\sH)+\sM_{\Phi}(\sH)+\max\curl*{0,\Psi(\e)}.
\end{equation}
\end{theorem}
\begin{theorem}[\textbf{Non-adversarial distribution-dependent $\Gamma$-bound}]
\label{Thm:excess_bounds_Gamma_01_general}
Suppose that $\sH$ satisfies the condition of
Lemma~\ref{lemma:explicit_assumption_01} and that $\Phi$ is a
margin-based loss function. Assume there exist a non-negative and
non-decreasing concave function $\Gamma\colon \mathbb{R_{+}}\to \Rset$
and $\epsilon\geq0$ such that the following holds for any $x\in \sX$:
\begin{align}
\label{eq:condition_Gamma_general}
    \tri*{2 \abs*{\Delta \eta(x)}}_{\e} \leq \Gamma\paren*{\inf_{h\in \ov \sH(x)}\Delta\sC_{\Phi,\sH}(h,x)}.
\end{align}
Then, for any hypothesis $h\in\sH$,
\begin{equation}
     \sR_{\ell_{0-1}}(h)- \sR_{\ell_{0-1}}^*(\sH)
     \leq  \Gamma\paren*{\sR_{\Phi}(h)-\sR_{\Phi}^*(\sH)+\sM_{\Phi}(\sH)}
     -\sM_{\ell_{0-1}}(\sH)+\epsilon.
\end{equation}
\end{theorem}
\begin{theorem}[\textbf{Adversarial distribution-dependent $\Psi$-bound}]
\label{Thm:excess_bounds_Psi_01_general_adv}
Suppose that $\sH$ is symmetric and that $\wt{\Phi}$ is a
supremum-based margin loss function. Assume there exist a convex
function $\Psi\colon \mathbb{R_{+}}\to \Rset$ with $\Psi(0)=0$ and
$\epsilon\geq0$ such that the following holds for any $x\in \sX$:
\begin{equation}
\label{eq:condition_Psi_general_adv}
\begin{aligned}
&\Psi\paren*{\tri*{\abs*{\Delta \eta(x)}+1/2}_{\e}}\leq
\inf_{h\in \ov \sH_\gamma(x)}
\Delta\sC_{\Phi,\sH}(h,x),\\
&\Psi\paren*{\tri*{2\Delta \eta(x)}_{\e}}\leq
\inf_{h\in\sH: \ov h_\gamma(x)<0}
\Delta\sC_{\Phi,\sH}(h,x),\\
&\Psi\paren*{\tri*{-2\Delta \eta(x)}_{\e}}\leq
\inf_{h\in\sH: \uv h_\gamma(x)>0}
\Delta\sC_{\Phi,\sH}(h,x).
\end{aligned}
\end{equation}
Then, for any hypothesis $h\in\sH$,
\begin{equation}
     \Psi\paren*{\sR_{\ell_{\gamma}}(h)- \sR_{\ell_{\gamma}}^*(\sH)+\sM_{\ell_{\gamma}}(\sH)}
     \leq  \sR_{\wt{\Phi}}(h)-\sR_{\wt{\Phi}}^*(\sH)+\sM_{\wt{\Phi}}(\sH)+\max\curl*{0,\Psi(\e)}.
\end{equation}
\end{theorem}
\begin{theorem}[\textbf{Adversarial distribution-dependent $\Gamma$-bound}]
\label{Thm:excess_bounds_Gamma_01_general_adv}
Suppose that $\sH$ is symmetric and that $\wt{\Phi}$ is a
supremum-based margin loss function. Assume there exist a non-negative
and non-decreasing concave function $\Gamma\colon \mathbb{R_{+}}\to
\Rset$ and $\epsilon\geq0$ such that the following holds for any $x\in
\sX$:
\begin{equation}
\label{eq:condition_Gamma_general_adv}
\begin{aligned}
&\tri*{\abs*{\Delta \eta(x)}+1/2}_{\e}\leq \Gamma\paren*{
\inf_{h\in \ov \sH_\gamma(x)}
\Delta\sC_{\Phi,\sH}(h,x)},\\
&\tri*{2\Delta \eta(x)}_{\e}\leq \Gamma\paren*{
\inf_{h\in\sH: \ov h_\gamma(x)<0}
\Delta\sC_{\Phi,\sH}(h,x)},\\
&\tri*{-2\Delta \eta(x)}_{\e}\leq \Gamma\paren*{
\inf_{h\in\sH: \uv h_\gamma(x)>0}
\Delta\sC_{\Phi,\sH}(h,x)}.
\end{aligned}
\end{equation}
Then, for any hypothesis $h\in\sH$,
\begin{equation}
     \sR_{\ell_{\gamma}}(h)- \sR_{\ell_{\gamma}}^*(\sH)
     \leq  \Gamma\paren*{\sR_{\wt{\Phi}}(h)-\sR_{\wt{\Phi}}^*(\sH)+\sM_{\wt{\Phi}}(\sH)}-\sM_{\ell_{\gamma}}(\sH)+\epsilon.
\end{equation}
\end{theorem}

\begin{restatable}[\textbf{Distribution-independent $\Gamma$-bound}]{theorem}{ExcessBoundsGammaUniform}
\label{Thm:excess_bounds_Gamma_uniform}
Suppose that $\sH$ satisfies the condition of
Lemma~\ref{lemma:explicit_assumption_01} and that $\Phi$ is a
margin-based loss function. Assume there exist a non-negative and
non-decreasing concave function $\Gamma\colon \mathbb{R_{+}}\to \Rset$
and $\epsilon\geq0$ such that the following holds for any for any
$t\in\left[1/2,1\right]\colon$
\begin{align*}
\tri*{2t-1}_{\e}\leq \Gamma\paren*{\inf_{x\in \sX,h\in\sH:h(x)< 0}\Delta\sC_{\Phi,\sH}(h,x,t)}. 
\end{align*}
Then, for any hypothesis $h\in\sH$ and any distribution,
\begin{equation}
\label{eq:bound_Gamma_01}
     \sR_{\ell_{0-1}}(h)- \sR_{\ell_{0-1}}^*(\sH)
     \leq  \Gamma\paren*{\sR_{\Phi}(h)-\sR_{\Phi}^*(\sH)+\sM_{\Phi}(\sH)}
     -\sM_{\ell_{0-1}}(\sH)+\epsilon.
\end{equation}
\end{restatable}

\begin{restatable}[\textbf{Adversarial distribution-independent $\Gamma$-bound}]{theorem}{ExcessBoundsGammaUniformAdv}
\label{Thm:excess_bounds_Gamma_uniform-adv}
Suppose that $\sH$ is symmetric and that $\wt{\Phi}$ is a supremum-based margin loss function. Assume there exist a non-negative and non-decreasing concave function $\Gamma\colon \mathbb{R_{+}}\to \Rset$ and $\epsilon\geq0$ such that the following holds for any for any $t\in\left[1/2,1\right]\colon$
\begin{align*}
&\tri*{t}_{\e}  \leq \Gamma\paren*{\inf_{x\in \sX,h\in \ov \sH_\gamma(x)\subsetneqq \sH}\Delta\sC_{\wt{\Phi},\sH}(h,x,t)},\\
&\tri*{2t-1}_{\e} \leq \Gamma\paren*{\inf_{x\in \sX,h\in\sH\colon \ov h_\gamma(x)< 0}\Delta\sC_{\wt{\Phi},\sH}(h,x,t)}.
\end{align*}
Then, for any hypothesis $h\in\sH$ and any distribution,
\begin{equation}
\label{eq:bound_Gamma_01_adv}
     \sR_{\ell_{\gamma}}(h)- \sR_{\ell_{\gamma}}^*(\sH)
     \leq  \Gamma\paren*{\sR_{\wt{\Phi}}(h)-\sR_{\wt{\Phi}}^*(\sH)+\sM_{\wt{\Phi}}(\sH)}-\sM_{\ell_{\gamma}}(\sH)+\epsilon.
\end{equation}
\end{restatable}

\section{Proof of Theorem~\ref{Thm:excess_bounds_Psi} and Theorem~\ref{Thm:excess_bounds_Gamma}}
\label{app:excess_bounds}

\ExcessBoundsPsi*
\begin{proof}
For any $h\in \sH$, since $\Psi\paren*{\Delta\sC_{\ell_2,\sH}(h,x)\mathds{1}_{\Delta\sC_{\ell_2,\sH}(h,x)>\epsilon}}\leq \Delta\sC_{\ell_1,\sH}(h,x)$ for all $x\in \sX$, we have
    \begin{align*}
       &\Psi\paren*{\sR_{\ell_2}(h)-\sR_{\ell_2}^*(\sH)+\sM_{\ell_2}(\sH)}\\
       &=\Psi\paren*{\mathbb{E}_{X}  \bracket*{\sC_{\ell_2}(h,x)-\sC^*_{\ell_2}(\sH, x)}}\\
       &=\Psi\paren*{\mathbb{E}_{X}  \bracket*{\Delta\sC_{\ell_2,\sH}(h,x)}}\\
       &\leq\mathbb{E}_{X} \bracket*{\Psi\paren*{\Delta\sC_{\ell_2,\sH}(h,x)}} & (\text{Jensen's ineq.}) \\
        &=\mathbb{E}_{X}  \bracket*{\Psi\paren*{\Delta\sC_{\ell_2,\sH}(h,x)\mathds{1}_{\Delta\sC_{\ell_2,\sH}(h,x)>\epsilon}+\Delta\sC_{\ell_2,\sH}(h,x)\mathds{1}_{\Delta\sC_{\ell_2,\sH}(h,x)\leq\epsilon}}}\\ 
       &\leq\mathbb{E}_{X} \bracket*{\Psi\paren*{\Delta\sC_{\ell_2,\sH}(h,x)\mathds{1}_{\Delta\sC_{\ell_2,\sH}(h,x)>\epsilon}}+\Psi\paren*{\Delta\sC_{\ell_2,\sH}(h,x)\mathds{1}_{\Delta\sC_{\ell_2,\sH}(h,x)\leq\epsilon}}}  &\paren*{\Psi(0)\geq 0}\\ 
       &\leq\mathbb{E}_{X} \bracket*{\Delta\sC_{\ell_1,\sH}(h,x)}+\sup_{t\in[0,\epsilon]}\Psi(t) &\paren*{\text{assumption}}\\
       &=\sR_{\ell_1}(h)-\sR_{\ell_1}^*(\sH)+\sM_{\ell_1}(\sH)+\max\curl*{\Psi(0),\Psi(\e)}, &\paren*{\text{convexity of $\Psi$}}
    \end{align*}
    which proves the theorem.
\end{proof}

\ExcessBoundsGamma*
\begin{proof}
For any $h\in \sH$, since $\Delta\sC_{\ell_2,\sH}(h,x)\mathds{1}_{\Delta\sC_{\ell_2,\sH}(h,x)>\epsilon}\leq \Gamma\paren*{\Delta\sC_{\ell_1,\sH}(h,x)}$ for all $ x\in \sX$, we have
    \begin{align*}
       &\sR_{\ell_2}(h)-\sR_{\ell_2}^*(\sH)+\sM_{\ell_2}(\sH)\\
       &=\mathbb{E}_{X}  \bracket*{\sC_{\ell_2}(h,x)-\sC^*_{\ell_2}(\sH, x)}\\
       &=\mathbb{E}_{X}  \bracket*{\Delta\sC_{\ell_2,\sH}(h,x)}\\
        &=\mathbb{E}_{X}  \bracket*{\Delta\sC_{\ell_2,\sH}(h,x)\mathds{1}_{\Delta\sC_{\ell_2,\sH}(h,x)>\epsilon}+\Delta\sC_{\ell_2,\sH}(h,x)\mathds{1}_{\Delta\sC_{\ell_2,\sH}(h,x)\leq\epsilon}}\\ 
        &\leq\mathbb{E}_{X}  \bracket*{\Gamma\paren*{\Delta\sC_{\ell_1,\sH}(h,x)}}+\epsilon &\paren*{\text{assumption}}\\ 
        &\leq\Gamma\paren*{\mathbb{E}_{X}  \bracket*{\Delta\sC_{\ell_1,\sH}(h,x)}}+\epsilon &\paren*{\text{concavity of $\Gamma$}}\\ 
       &=\Gamma\paren*{\sR_{\ell_1}(h)-\sR_{\ell_1}^*(\sH)+\sM_{\ell_1}(\sH)}+\epsilon,
    \end{align*}
    which proves the theorem.
\end{proof}

\section{Proof of Lemma~\ref{lemma:explicit_assumption_01} and Lemma~\ref{lemma:explicit_assumption_01_adv}}
\label{app:explicit_assumption}
\ExplicitAssumption*
\begin{proof}
By the definition, the conditional $\ell_{0-1}$-risk is 
\begin{align*}
\sC_{\ell_{0-1}}(h,x) 
& = \eta(x)\mathds{1}_{h(x)< 0}+(1-\eta(x))\mathds{1}_{h(x)\geq 0} \\
& = 
\begin{cases}
\eta(x) & \text{if} ~ h(x)<0,\\
1-\eta(x) & \text{if} ~ h(x)\geq 0.
\end{cases}
\end{align*}
By the assumption, for any $x\in \sX$, there exists $h^*\in \sH$ such that $\sign(h^*(x))=\sign(\Delta \eta(x))$, where $\Delta \eta(x)$ is the Bayes classifier such that \[\sC_{\ell_{0-1}}\paren*{\Delta \eta(x),x}=\sC^*_{\ell_{0-1}}\paren*{\sH_{\mathrm{all}}, x}=\min\curl*{\eta(x),1-\eta(x)}.\] Therefore,  the optimal conditional $\ell_{0-1}$-risk is
\begin{align*}
\sC^*_{\ell_{0-1}}(\sH, x)= \sC_{\ell_{0-1}}\paren*{h^*,x}=\sC_{\ell_{0-1}}\paren*{\Delta \eta(x),x}=\min\curl*{\eta(x),1-\eta(x)}
\end{align*}
which proves the first part of lemma. By the definition,
\begin{align*}
\Delta\sC_{\ell_{0-1},\sH}(h,x)
&=\sC_{\ell_{0-1}}(h,x) - \sC^*_{\ell_{0-1}}(\sH, x)\\
& = \eta(x)\mathds{1}_{h(x)< 0}+(1-\eta(x))\mathds{1}_{h(x)\geq 0} - \min\curl*{\eta(x),1-\eta(x)}\\
& = 
\begin{cases}
2 \abs*{\Delta \eta(x)},  & h \in \ov \sH(x), \\
0, & \text{otherwise} .
\end{cases}
\end{align*}
This leads to
\begin{align*}
\tri*{\Delta\sC_{\ell_{0-1},\sH}(h,x)}_{\e}=\tri*{2 \abs*{\Delta \eta(x)}}_{\e}\mathds{1}_{h \in \ov \sH(x)}\,.
\end{align*}
\end{proof}

\ExplicitAssumptionAdv*
\begin{proof}
By the definition, the conditional $\ell_{\gamma}$-risk is
\begin{align*}
  \sC_{\ell_{\gamma}}(h,x)
  &=\eta(x) \mathds{1}_{\left\{\uv h_\gamma(x)\leq 0\right\}}+(1-\eta(x)) \mathds{1}_{\left\{\ov h_\gamma(x)\geq 0\right\}}\\
  &=\begin{cases}
   1 & \text{if} ~ h\in \ov \sH_\gamma(x),\\
   \eta(x) & \text{if} ~ \ov h_\gamma(x)<0,\\
   1-\eta(x) & \text{if} ~ \uv h_\gamma(x)> 0.\\
  \end{cases}
\end{align*}
Since $\sH$ is symmetric, for any $x \in \sX$, either there exists $h\in\sH$ such that $\uv h_\gamma(x)>0$, or $\ov \sH_\gamma(x)=\sH$. When $\ov \sH_\gamma(x)=\sH$, $\{h\in \sH:\ov h_\gamma(x)<0\}$ and $\{h\in \sH:\uv h_\gamma(x)>0\}$ are both empty sets. Thus $\sC^*_{\ell_{\gamma}}(\sH, x)=1$.  When $\ov \sH_\gamma(x)\neq\sH$, there exists $h\in \sH$ such that $\sC_{\ell_{\gamma}}(h,x)=\min\curl*{\eta(x),1-\eta(x)}=\sC^*_{\ell_{\gamma}}(\sH, x)$. Therefore, the minimal conditional $\ell_{\gamma}$-risk is
\begin{align*}
     \sC^*_{\ell_{\gamma}}(\sH, x)=\begin{cases}
        1, & \ov \sH_\gamma(x)=\sH\,,\\
     	\min\curl*{\eta(x),1-\eta(x)}, & \ov \sH_\gamma(x)\neq\sH\,.
     \end{cases}
\end{align*}
When $\ov \sH_\gamma(x)=\sH$, $\sC_{\ell_{\gamma}}(h,x)\equiv1$, which implies that $\Delta\sC_{\ell_{\gamma},\sH}(h,x)\equiv0$. For $h\in \ov \sH_\gamma(x)\subsetneqq \sH$, $\Delta\sC_{\ell_{\gamma},\sH}(h,x)=1-\min\curl*{\eta(x),1-\eta(x)}=\abs*{\Delta\eta(x)}+1/2$; for $h\in\sH$ such that $\ov h_\gamma(x)<0$, we have $\Delta\sC_{\ell_{\gamma},\sH}(h,x)=\eta(x)-\min\curl*{\eta(x),1-\eta(x)}=\max\curl*{0,2\Delta \eta(x)}$; for $h\in \sH$ such that $\uv h_\gamma(x)>0$, $\Delta\sC_{\ell_{\gamma},\sH}(h,x)=1-\eta(x)-\min\curl*{\eta(x),1-\eta(x)}=\max\curl*{0,-2\Delta \eta(x)}$. Therefore,
\begin{align*}
  \Delta\sC_{\ell_{\gamma},
    \sH}(h,x)=
    \begin{cases}
    \abs*{\Delta\eta(x)}+1/2 &  h\in \ov \sH_\gamma(x)\subsetneqq \sH,\\
    \max\curl*{0,2\Delta \eta(x)} & \ov h_\gamma(x)<0,\\
    \max\curl*{0,-2\Delta \eta(x)} & \uv h_\gamma(x)>0,\\
    0 & \text{otherwise}.
    \end{cases}
\end{align*}
This leads to
\begin{align*}
\tri*{\Delta\sC_{\ell_{\gamma},\sH}(h,x)}_{\e}
=
\begin{cases}
\tri*{\abs*{\Delta \eta(x)}+\frac12}_{\e} 
&h \in \ov \sH_\gamma(x)\subsetneqq \sH\\
\tri*{2\Delta \eta(x)}_{\e}
& \ov h_\gamma(x)<0\\
\tri*{-2\Delta \eta(x)}_{\e}
&\uv h_\gamma(x)>0 \\
0 
&\text{otherwise}
\end{cases}
\end{align*}
\end{proof}

\section{Comparison with Previous Results when \texorpdfstring{$\sH=\sH_{\mathrm{all}}$}{all}}
\label{app:compare-all-measurable}
\subsection{Comparison with \texorpdfstring{\citep[Theorem 4.7]{MohriRostamizadehTalwalkar2018}}{ref}}
Assume $\Phi$ is convex and non-increasing. For any $x\in \sX$, by the convexity, we have
\begin{align}
\label{eq:convex}
\sC_{\Phi}(h,x)=\eta(x)\Phi(h(x))+(1-\eta(x))\Phi(-h(x))\geq \Phi(2\Delta \eta(x)h(x)).
\end{align}
Then,
\begin{align*}
\inf_{h\in \ov{\sH_{\mathrm{all}}}(x)}\Delta\sC_{\Phi,\sH_{\mathrm{all}}}(h,x)
&\geq
\inf_{h\in\sH_{\mathrm{all}}:2\Delta \eta(x)h(x)\leq 0}\Delta\sC_{\Phi,\sH_{\mathrm{all}}}(h,x) \tag{$h\in \ov{\sH_{\mathrm{all}}}(x)  \implies h(x)\Delta \eta(x)\leq 0$}\\
& \geq \inf_{h\in\sH_{\mathrm{all}}:2\Delta \eta(x)h(x)\leq 0}\Phi(2\Delta \eta(x)h(x))-\sC^*_{\Phi}\paren*{\sH_{\mathrm{all}}, x}  \tag{\eqref{eq:convex}}\\
&= \sC_{\Phi}(0,x)- \sC^*_{\Phi}\paren*{\sH_{\mathrm{all}}, x} & \tag{$\Phi$ is non-increasing}
\end{align*}
Thus the condition of Theorem~4.7 in \citep{MohriRostamizadehTalwalkar2018} implies the condition in Corollary~\ref{cor:excess_bounds_Psi_01_M}:
\begin{align*}
& \abs*{\Delta \eta(x)}\leq c~\bracket*{\sC_{\Phi}(0,x)- \sC^*_{\Phi}\paren*{\sH_{\mathrm{all}}, x}}^{\frac1s},\; \forall x\in \sX\\
& \qquad \implies
\abs*{\Delta \eta(x)}\leq c~\inf_{h\in \ov{\sH_{\mathrm{all}}}(x)}\bracket*{\Delta\sC_{\Phi,\sH_{\mathrm{all}}}(h,x)}^{\frac1s},\; \forall x\in \sX.
\end{align*}
Therefore, Theorem~4.7 in \citep{MohriRostamizadehTalwalkar2018} is a special case of Corollary~\ref{cor:excess_bounds_Psi_01_M}.

\subsection{Comparison with \texorpdfstring{\citep[Theorem 1.1]{bartlett2006convexity}}{ref}}
We show that the $\psi$-transform in \citep{bartlett2006convexity} verifies the condition in Corollary~\ref{cor:excess_bounds_Psi_01_B} for all distributions.
First, by Definition 2 in \citep{bartlett2006convexity}, we know that $\psi$ is convex, $\psi(0)=0$ and $\psi\leq \wt{\psi}$. 
Then,
\begin{align*}
\psi\paren*{2 \abs*{\Delta \eta(x)}}
&\leq \wt{\psi} \paren*{2 \abs*{\Delta \eta(x)}} & \tag{$\psi\leq \wt{\psi}$} \\
&= \inf_{\alpha\leq0}\paren*{\max\curl*{\eta(x),1-\eta(x)}\Phi(\alpha)+\min\curl*{\eta(x),1-\eta(x)}\Phi(-\alpha)}\\
&-\inf_{\alpha\in \Rset}\paren*{\max\curl*{\eta(x),1-\eta(x)}\Phi(\alpha)+\min\curl*{\eta(x),1-\eta(x)}\Phi(-\alpha)} &\tag{$\text{def. of } \wt{\psi}$}\\
&= \inf_{\alpha\Delta \eta(x)\leq0}\paren*{\eta(x)\Phi(\alpha)+\paren*{1-\eta(x)}\Phi(-\alpha)}-\inf_{\alpha\in \Rset}\paren*{\eta(x)\Phi(\alpha)+\paren*{1-\eta(x)}\Phi(-\alpha)} & \tag{$\text{symmetry}$}\\
&= \inf_{h\in\sH_{\mathrm{all}}:h(x)\Delta \eta(x)\leq 0}\Delta\sC_{\Phi,\sH_{\mathrm{all}}}(h,x)\\
&\leq\inf_{h\in \ov{\sH_{\mathrm{all}}}(x)}\Delta\sC_{\Phi,\sH_{\mathrm{all}}}(h,x) & \tag{$h\in \ov{\sH_{\mathrm{all}}}(x)  \implies h(x)\Delta \eta(x)\leq 0$}
\end{align*}
Therefore, Theorem 1.1 in \citep{bartlett2006convexity} is a special case of Corollary~\ref{cor:excess_bounds_Psi_01_B}.

\section{Proof of Theorem~\ref{Thm:excess_bounds_Psi_uniform} and Theorem~\ref{Thm:excess_bounds_Gamma_uniform}}
\label{app:uniform}
\ExcessBoundsPsiUniform*
\begin{proof}
Note the condition~\eqref{eq:condition_Psi_general} in Theorem~\ref{Thm:excess_bounds_Psi_01_general} is symmetric about $\Delta \eta(x)=0$. Thus, condition~\eqref{eq:condition_Psi_general} uniformly holds for all distributions is equivalent to the following holds for any $t\in\left[1/2,1\right]\colon$
\begin{align*}
\Psi \paren*{\tri*{2t-1}_{\e}}\leq \inf_{x\in \sX,h\in\sH:h(x)<0}\Delta\sC_{\Phi,\sH}(h,x,t),
\end{align*}
which proves the theorem.
\end{proof}

\ExcessBoundsGammaUniform*
\begin{proof}
Note the condition~\eqref{eq:condition_Gamma_general} in Theorem~\ref{Thm:excess_bounds_Gamma_01_general} is symmetric about $\Delta \eta(x)=0$. Thus, condition~\eqref{eq:condition_Gamma_general} uniformly holds for all distributions is equivalent to the following holds for any $t\in\left[1/2,1\right]\colon$
\begin{align*}
\Psi \paren*{\tri*{2t-1}_{\e}}\leq \inf_{x\in \sX,h\in\sH:h(x)<0}\Delta\sC_{\Phi,\sH}(h,x,t),
\end{align*}
which proves the theorem.
\end{proof}

\section{Proof of Theorem~\ref{Thm:excess_bounds_Psi_uniform-adv} and Theorem~\ref{Thm:excess_bounds_Gamma_uniform-adv}}
\label{app:uniform-adv}
\ExcessBoundsPsiUniformAdv*
\begin{proof}
Note the condition~\eqref{eq:condition_Psi_general_adv} in Theorem~\ref{Thm:excess_bounds_Psi_01_general_adv} is symmetric about $\Delta \eta(x)=0$. Thus, condition~\eqref{eq:condition_Psi_general_adv} uniformly holds for all distributions is equivalent to the following holds for any $t\in\left[1/2,1\right]\colon$
\begin{align*}
&\Psi\paren*{\tri*{t}_{\e}}  \leq \inf_{x\in\sX,h\in \ov \sH_\gamma(x)\subsetneqq \sH}\Delta\sC_{\wt{\Phi},\sH}(h,x,t),\\
&\Psi\paren*{\tri*{2t-1}_{\e}} \leq \inf_{x\in \sX,h\in\sH\colon  \ov h_\gamma(x)< 0}\Delta\sC_{\wt{\Phi},\sH}(h,x,t),
\end{align*}
which proves the theorem.
\end{proof}

\ExcessBoundsGammaUniformAdv*
\begin{proof}
Note the condition~\eqref{eq:condition_Gamma_general_adv} in Theorem~\ref{Thm:excess_bounds_Gamma_01_general_adv} is symmetric about $\Delta \eta(x)=0$. Thus, condition~\eqref{eq:condition_Gamma_general_adv} uniformly holds for all distributions is equivalent to the following holds for any $t\in\left[1/2,1\right]\colon$
\begin{align*}
&\tri*{t}_{\e}  \leq \Gamma\paren*{\inf_{x\in \sX,h\in \ov \sH_\gamma(x)\subsetneqq \sH}\Delta\sC_{\wt{\Phi},\sH}(h,x,t)},\\
&\tri*{2t-1}_{\e} \leq \Gamma\paren*{\inf_{x\in \sX,h\in\sH\colon \ov h_\gamma(x)< 0}\Delta\sC_{\wt{\Phi},\sH}(h,x,t)},
\end{align*}
which proves the theorem.
\end{proof}

\section{Proof of Theorem~\ref{Thm:tightness} and Theorem~\ref{Thm:tightness-adv} }
\label{app:tightness}
\Tightness*
\begin{proof}
By Theorem~\ref{Thm:excess_bounds_Psi_uniform}, if $\sT_{\Phi}$ is convex with $\sT_{\Phi}(0)=0$, the first inequality holds. For any $t\in [0,1]$, consider the distribution that supports on a singleton $\curl*{x_0}$ and satisfies that $\eta(x_0)=\frac12+\frac{t}{2}$. Thus
\begin{align*}
\inf_{x\in\sX,h\in\sH:h(x)<0}\Delta\sC_{\Phi,\sH}\paren*{h,x,\eta(x_0)}=\inf_{h\in\sH:h(x_0)<0}\Delta\sC_{\Phi,\sH}\paren*{h,x_0,\eta(x_0)}= \inf_{h\in\sH:h(x_0)<0}\Delta\sC_{\Phi,\sH}(h,x_0).
\end{align*}
For any $\delta>0$, take $h_0\in \sH$ such that $h_0(x_0)< 0$ and
\begin{align*}
\Delta\sC_{\Phi,\sH}(h_0,x_0)\leq\inf_{h\in\sH:h(x_0)<0}\Delta\sC_{\Phi,\sH}(h,x_0)+\delta= \inf_{x\in\sX,h\in\sH:h(x)<0}\Delta\sC_{\Phi,\sH}\paren*{h,x,\eta(x_0)}+\delta.    
\end{align*}
Then, we have
\begin{align*}
\sR_{\ell_{0-1}}(h_0)- \sR_{\ell_{0-1}}^*(\sH)+\sM_{\ell_{0-1}}(\sH) & = \sR_{\ell_{0-1}}(h_0) - \mathbb{E}_{X} \bracket* {\sC^*_{\ell_{0-1}}(\sH, x)}\\
& =\Delta\sC_{\ell_{0-1},\sH}(h_0,x_0)\\ 
& =2\eta(x_0)-1\\
& = t,\\
\sR_{\Phi}(h_0)-\sR_{\Phi}^*(\sH)+\sM_{\Phi}(\sH) 
& = \sR_{\Phi}(h_0) - \mathbb{E}_{X} \bracket* {\sC^*_{\Phi}(\sH, x)}\\
& = \Delta\sC_{\Phi,\sH}(h_0,x_0)\\
& \leq \inf_{x\in\sX,h\in\sH:h(x)<0}\Delta\sC_{\Phi,\sH}\paren*{h,x,\eta(x_0)} + \delta \\
& = \sT_{\Phi}\paren*{2\eta(x_0)-1}+\delta\\
& = \sT_{\Phi}(t) + \delta,
\end{align*}
which completes the proof.
\quad
\end{proof}
\TightnessAdv*
\begin{proof}
By Theorem~\ref{Thm:excess_bounds_Psi_uniform-adv}, if $\sT_{\wt{\Phi}}$ is convex with $\sT_{\wt{\Phi}}(0)=0$, the first inequality holds. For any $t\in [0,1]$, consider the distribution that supports on a singleton $\curl*{x_0}$, which satisfies that $\eta(x_0)=\frac12+\frac{t}{2}$ and $\ov \sH_{\gamma}(x_0)\neq \sH$.
Thus
\begin{align*}
\inf_{x\in\sX,h\in\sH:\ov h_\gamma(x)<0}\Delta\sC_{\wt{\Phi},\sH}\paren*{h,x,\eta(x_0)}=\inf_{h\in\sH:\ov h_\gamma(x_0)<0}\Delta\sC_{\wt{\Phi},\sH}\paren*{h,x_0,\eta(x_0)}= \inf_{h\in\sH:\ov h_\gamma(x_0)<0}\Delta\sC_{\wt{\Phi},\sH}(h,x_0).
\end{align*}
For any $\delta>0$, take $h\in \sH$ such that $\ov h_\gamma(x_0)< 0$ and
\begin{align*}
\Delta\sC_{\wt{\Phi},\sH}(h,x_0)\leq \inf_{h\in\sH:\ov h_\gamma(x_0)<0}\Delta\sC_{\wt{\Phi},\sH}(h,x_0)+\delta=\inf_{x\in\sX,h\in\sH:\ov h_\gamma(x)<0}\Delta\sC_{\wt{\Phi},\sH}\paren*{h,x,\eta(x_0)}+\delta.    
\end{align*}
Then, we have
\begin{align*}
\sR_{\ell_{\gamma}}(h)- \sR_{\ell_{\gamma}}^*(\sH)+\sM_{\ell_{\gamma}}(\sH) & = \sR_{\ell_{\gamma}}(h) - \mathbb{E}_{X} \bracket* {\sC^*_{\ell_{\gamma}}(\sH, x)}\\
& =\Delta\sC_{\ell_{\gamma},\sH}(h,x_0)\\
& =2\eta(x_0)-1\\
& = t,\\
\sR_{\wt{\Phi}}(h)-\sR_{\wt{\Phi}}^*(\sH)+\sM_{\wt{\Phi}}(\sH) 
& = \sR_{\wt{\Phi}}(h) - \mathbb{E}_{X} \bracket* {\sC^*_{\wt{\Phi}}(\sH, x)}\\
& = \Delta\sC_{\wt{\Phi},\sH}(h,x_0)\\
& \leq \inf_{x\in\sX,h\in\sH:\ov h_\gamma(x)< 0}\Delta\sC_{\wt{\Phi},\sH}\paren*{h,x,\eta(x_0)} + \delta \\
& = \sT_{2}\paren*{2\eta(x_0)-1}+\delta\\
& = \sT_{\wt{\Phi}}(2\eta(x_0)-1) + \delta & (\sT_2\leq \sT_1)\\
& =\sT_{\wt{\Phi}}(t) + \delta
\end{align*}
which completes the proof.
\quad
\end{proof}


\section{Proof of Theorem~\ref{Thm:negative_convex_adv}}
\label{app:negative_adv}
\NegativeConvexAdv*
\begin{proof}
Assume $x_0\in \sX$ is distinguishing. Consider the distribution that supports on $\curl*{x_0}$. Let $\eta(x_0)=1/2$ and $h_0=0\in \sH$. Then, for any $h\in \sH$,
\begin{align*}
\sR_{\ell_{\gamma}}(h)=\sC_{\ell_{\gamma}}(h,x_0)
=1/2\mathds{1}_{\uv h_\gamma(x_0)\leq 0} +1/2 \mathds{1}_{\ov h_\gamma(x_0)\geq 0} \geq 1/2,
\end{align*}
where the equality can be achieved for some $h\in \sH$ since $x_0$ is distinguishing. Therefore, 
\begin{align*}
\sR_{\ell_{\gamma}}^*(\sH)=\sC^*_{\ell_{\gamma}}(\sH, x_0)=\inf_{h\in \sH}\sC_{\ell_{\gamma}}(h,x_0)=1/2.
\end{align*}
Note $\sR_{\ell_{\gamma}}(h_0)=1/2 +1/2 =1$. 
For the supremum-based convex loss $\wt{\Phi}$, for any $h\in \sH$,
\begin{align*}
\sR_{\wt{\Phi}}(h)=\sC_{\wt{\Phi}}(h,x_0)
&=1/2\Phi\paren*{\uv h_\gamma(x_0)} +1/2\Phi\paren*{-\ov h_\gamma(x_0)}\\  
& \geq \Phi\paren*{1/2 \uv h_\gamma(x_0) -1/2 \ov h_\gamma(x_0)} &\quad \paren*{\text{convexity of }\Phi}\\
& \geq \Phi(0),  &\quad \paren*{\Phi \text{ is non-increasing}}
\end{align*}
where both equality can be achieved by $h_0=0$.
Therefore,
\begin{align*}
\sR_{\wt{\Phi}}^*(\sH)=\sC^*_{\wt{\Phi}}(\sH, x_0)=\sR_{\wt{\Phi}}(h_0)=\Phi(0).
\end{align*}
If \eqref{eq:est-bound} holds for some non-decreasing function $f$, then, we obtain for any $h\in \sH$,
\begin{align*}
\sR_{\ell_{\gamma}}(h)-1/2\leq  f\paren*{\sR_{\wt{\Phi}}(h) - \Phi(0)}.
\end{align*}
Let $h=h_0$, then $f(0)\geq 1/2$. Since $f$ is non-decreasing, for any $t\in [0,1]$, $f(t)\geq 1/2$.

\ignore{
For the supremum-based sigmoid loss $\wt{\Phi}_{\mathrm{sig}}$, for any $h\in \sH$,
\begin{align*}
\sR_{\wt{\Phi}_{\mathrm{sig}}}(h)=\sC_{\wt{\Phi}_{\mathrm{sig}}}(h,x_0)
&=1/2\Phi_{\mathrm{sig}}\paren*{\uv h_\gamma(x_0)} +1/2\Phi_{\mathrm{sig}}\paren*{-\ov h_\gamma(x_0)}\\
& = 1 +1/2 \bracket*{\tanh(k \ov h_\gamma(x_0))-\tanh(k \uv h_\gamma(x_0))} \\
& \geq 1
\end{align*}
where the equality can be achieved by $h_0=0$.
Therefore,
\begin{align*}
\sR_{\wt{\Phi}_{\mathrm{sig}}}^*(\sH)=\sC^*_{\wt{\Phi}_{\mathrm{sig}}}(\sH, x_0)=\sR_{\wt{\Phi}_{\mathrm{sig}}}(h_0)=1
\end{align*}
If \eqref{eq:est-bound} holds for some non-decreasing function $f$, then, we obtain for any $h\in \sH$,
\begin{align*}
\sR_{\ell_{\gamma}}(h)-1/2\leq  f\paren*{\sR_{\wt{\Phi}_{\mathrm{sig}}}(h) - 1}.
\end{align*}
Let $h=h_0$, then $f(0)\geq 1/2$. Since $f$ is non-decreasing, for any $t\in [0,1]$, $f(t)\geq 1/2$.}

For the supremum-based symmetric loss $\wt{\Phi}_{\mathrm{sym}}$, there exists a constant $C\geq 0$ such that, for any $h\in \sH$, 
\begin{align*}
\sR_{\wt{\Phi}_{\mathrm{sym}}}(h)=\sC_{\wt{\Phi}_{\mathrm{sym}}}(h,x_0)
&=1/2\Phi_{\mathrm{sym}}\paren*{\uv h_\gamma(x_0)} +1/2\Phi_{\mathrm{sym}}\paren*{-\ov h_\gamma(x_0)}\\
& \geq 1/2\Phi_{\mathrm{sym}}\paren*{\ov h_\gamma(x_0)} +1/2\Phi_{\mathrm{sym}}\paren*{-\ov h_\gamma(x_0)}\\
& \geq \frac{C}{2}
\end{align*}
where the equality can be achieved by $h_0=0$.
Therefore,
\begin{align*}
\sR_{\wt{\Phi}_{\mathrm{sym}}}^*(\sH)=\sC^*_{\wt{\Phi}_{\mathrm{sym}}}(\sH, x_0)=\sR_{\wt{\Phi}_{\mathrm{sym}}}(h_0)=\frac{C}{2}.
\end{align*}
If \eqref{eq:est-bound} holds for some non-decreasing function $f$, then, we obtain for any $h\in \sH$,
\begin{align*}
\sR_{\ell_{\gamma}}(h)-1/2\leq  f\paren*{\sR_{\wt{\Phi}_{\mathrm{sym}}}(h) - \frac{C}{2}}.
\end{align*}
Let $h=h_0$, then $f(0)\geq 1/2$. Since $f$ is non-decreasing, for any $t\in [0,1]$, $f(t)\geq 1/2$.
\end{proof}

\section{Derivation of Non-Adversarial \texorpdfstring{$\sH$}{H}-Consistency Bounds}
\label{app:derivation-non-adv}
\subsection{Linear Hypotheses}
\label{app:derivation-lin}
Since $\sH_{\mathrm{lin}}$ satisfies the condition of Lemma~\ref{lemma:explicit_assumption_01}, by Lemma~\ref{lemma:explicit_assumption_01} the $\paren*{\ell_{0-1},\sH_{\mathrm{lin}}}$-minimizability gap can be expressed as follows:
\begin{equation}
\begin{aligned}
\label{eq:M-01-lin}
\sM_{\ell_{0-1}}\paren*{\sH_{\mathrm{lin}}}
& = \sR_{\ell_{0-1}}^*\paren*{\sH_{\mathrm{lin}}}-\mathbb{E}_{X}\bracket*{\min\curl*{\eta(x),1-\eta(x)}} \\
&= \sR_{\ell_{0-1}}^*\paren*{\sH_{\mathrm{lin}}} - \sR_{\ell_{0-1}}^*\paren*{\sH_{\mathrm{all}}}.
\end{aligned}
\end{equation}
Therefore, the $\paren*{\ell_{0-1},\sH_{\mathrm{lin}}}$-minimizability gap coincides with
the $\paren*{\ell_{0-1},\sH_{\mathrm{lin}}}$-approximation error.
By the definition of $\sH_{\mathrm{lin}}$, for any $x \in \sX$, $\curl[\big]{h(x) \mid h \in \sH_{\mathrm{lin}}} = \bracket*{-W \norm*{x}_p-B, W\norm*{x}_p + B}$.
\subsubsection{Hinge Loss}
For the hinge loss $\Phi_{\mathrm{hinge}}(\alpha)\colon=\max\curl*{0,1 - \alpha}$, for all $h\in \sH_{\mathrm{lin}}$ and $x\in \sX$:
\begin{equation*}
\begin{aligned}
\sC_{\Phi_{\mathrm{hinge}}}(h,x,t)
& =t \Phi_{\mathrm{hinge}}(h(x))+(1-t)\Phi_{\mathrm{hinge}}(-h(x))\\
& =t\max\curl*{0,1-h(x)}+(1-t)\max\curl*{0,1+h(x)}.\\
\inf_{h\in\sH_{\mathrm{lin}}}\sC_{\Phi_{\mathrm{hinge}}}(h,x,t)
& = 1-\abs*{2t-1}\min\curl*{W\norm*{x}_p+B,1}.
\end{aligned}
\end{equation*}
Therefore, the $\paren*{\Phi_{\mathrm{hinge}},\sH_{\mathrm{lin}}}$-minimizability gap can be expressed as follows:
\begin{equation}
\begin{aligned}
\label{eq:M-hinge-lin}
\sM_{\Phi_{\mathrm{hinge}}}\paren*{\sH_{\mathrm{lin}}}
& = \sR_{\Phi_{\mathrm{hinge}}}^*\paren*{\sH_{\mathrm{lin}}}-\mathbb{E}_{X}\bracket*{1-\inf_{h\in\sH_{\mathrm{lin}}}\sC_{\Phi_{\mathrm{hinge}}}(h,x,\eta(x))}.\\
& = \sR_{\Phi_{\mathrm{hinge}}}^*\paren*{\sH_{\mathrm{lin}}}-\mathbb{E}_{X}\bracket*{1-\abs*{2\eta(x)-1}\min\curl*{W\norm*{x}_p+B,1}}.
\end{aligned}
\end{equation}
Note the $\paren*{\Phi_{\mathrm{hinge}},\sH_{\mathrm{lin}}}$-minimizability gap coincides with
the $\paren*{\Phi_{\mathrm{hinge}},\sH_{\mathrm{lin}}}$-approximation error: $\sR_{\Phi_{\mathrm{hinge}}}^*\paren*{\sH_{\mathrm{lin}}}-\mathbb{E}_{X}\bracket*{1-\abs*{2\eta(x)-1}}$ for $B \geq 1$.

For $\frac{1}2< t\leq1$, we have
\begin{align*}
\inf_{h\in\sH_{\mathrm{lin}}:h(x)<0}\sC_{\Phi_{\mathrm{hinge}}}(h,x,t)
& = t\max\curl*{0,1-0}+(1-t)\max\curl*{0,1+0}\\
& =1.\\
\inf_{x\in \sX} \inf_{h\in\sH_{\mathrm{lin}}:h(x)<0} \Delta\sC_{\Phi_{\mathrm{hinge}},\sH_{\mathrm{lin}}}(h,x,t)
& = \inf_{x\in \sX} \curl*{\inf_{h\in\sH_{\mathrm{lin}}:h(x)<0}\sC_{\Phi_{\mathrm{hinge}}}(h,x,t)-\inf_{h\in\sH_{\mathrm{lin}}}\sC_{\Phi_{\mathrm{hinge}}}(h,x,t)}\\
&=\inf_{x\in \sX}\paren*{2t-1}\min\curl*{W\norm*{x}_p+B,1}\\
&=(2t-1)\min\curl*{B,1}\\
&=\sT(2t - 1),
\end{align*}
where $\sT$ is the increasing and convex function on $[0,1]$ defined by
\begin{align*}
\forall t \in [0,1], \quad \sT(t) = \min \curl*{B, 1} \, t .
\end{align*}
By Definition~\ref{def:trans}, for any $\epsilon\geq 0$, the $\sH_{\mathrm{lin}}$-estimation error transformation of the hinge loss is as follows:
\begin{align*}
\sT_{\Phi_{\mathrm{hinge}}}= \min \curl*{B, 1} \, t, \quad t \in [0,1],
\end{align*}
Therefore, $\sT_{\Phi_{\mathrm{hinge}}}$ is convex, non-decreasing, invertible and satisfies that $\sT_{\Phi_{\mathrm{hinge}}}(0)=0$. By Theorem~\ref{Thm:tightness}, we can choose $\Psi(t)=\min\curl*{B,1} \, t$ in Theorem~\ref{Thm:excess_bounds_Psi_uniform}, or, equivalently, $\Gamma(t) = \frac{t}{\min\curl*{B, 1}}$ in Theorem~\ref{Thm:excess_bounds_Gamma_uniform}, which are optimal when $\e=0$. Thus, by Theorem~\ref{Thm:excess_bounds_Psi_uniform} or Theorem~\ref{Thm:excess_bounds_Gamma_uniform}, setting $\e = 0$ yields the $\sH_{\mathrm{lin}}$-consistency bound for the hinge loss, valid for all $h \in \sH_{\mathrm{lin}}$:
\begin{align}
\label{eq:hinge-lin-est}
     \sR_{\ell_{0-1}}(h)- \sR_{\ell_{0-1}}^*\paren*{\sH_{\mathrm{lin}}}
     \leq \frac{\sR_{\Phi_{\mathrm{hinge}}}(h)- \sR_{\Phi_{\mathrm{hinge}}}^*\paren*{\sH_{\mathrm{lin}}}+\sM_{\Phi_{\mathrm{hinge}}}\paren*{\sH_{\mathrm{lin}}}}{\min\curl*{B,1}}-\sM_{\ell_{0-1}, \sH_{\mathrm{lin}}}.
\end{align}
Since the $\paren*{\ell_{0-1},\sH_{\mathrm{lin}}}$-minimizability gap coincides with
the $\paren*{\ell_{0-1},\sH_{\mathrm{lin}}}$-approximation error and 
$\paren*{\Phi_{\mathrm{hinge}},\sH_{\mathrm{lin}}}$-minimizability gap coincides with
the $\paren*{\Phi_{\mathrm{hinge}},\sH_{\mathrm{lin}}}$-approximation error for $B \geq 1$,
the inequality can be rewritten as follows:
\begin{align*}
     \sR_{\ell_{0-1}}(h)- \sR_{\ell_{0-1}}^*\paren*{\sH_{\mathrm{all}}}
     \leq 
     \begin{cases}
     \sR_{\Phi_{\mathrm{hinge}}}(h) - \sR_{\Phi_{\mathrm{hinge}}}^*\paren*{\sH_{\mathrm{all}}} & \text{if } B \geq 1\\
     \frac{1}{B} \bracket[\Big]{\sR_{\Phi_{\mathrm{hinge}}}(h)
     - \mathbb{E}_{X}\bracket*{1-\abs*{2\eta(x)-1}\min\curl*{W\norm*{x}_p+B,1}} } & \text{otherwise}.
     \end{cases}
\end{align*}
The inequality for $B \geq 1$ coincides with the consistency excess error bound
known for the hinge loss \citep{Zhang2003,bartlett2006convexity,MohriRostamizadehTalwalkar2018} but the one for $B < 1$ is distinct and novel. For $B<1$, we have
\begin{align*}
& \mathbb{E}_{X}\bracket*{1-\abs*{2\eta(x)-1}\min\curl*{W\norm*{x}_p+B,1}}\\
&\qquad > \mathbb{E}_{X}\bracket*{1-\abs*{2\eta(x)-1}}= 2\mathbb{E}_X\bracket*{\min\curl*{\eta(x), 1 - \eta(x)}}= \sR_{\Phi_{\mathrm{hinge}}}^*\paren*{\sH_{\mathrm{all}}}.
\end{align*}
Therefore for $B<1$,
\begin{align*}
\sR_{\Phi_{\mathrm{hinge}}}(h) - \mathbb{E}_{X}\bracket*{1-\abs*{2\eta(x)-1}\min\curl*{W\norm*{x}_p+B,1}} < \sR_{\Phi_{\mathrm{hinge}}}(h) - \sR_{\Phi_{\mathrm{hinge}}}^*\paren*{\sH_{\mathrm{all}}}.
\end{align*}
Note that: $\sR_{\Phi_{\mathrm{hinge}}}^*\paren*{\sH_{\mathrm{all}}} = 2 \sR_{\ell_{0-1}}^*\paren*{\sH_{\mathrm{all}}} =2\mathbb{E}_X\bracket*{\min\curl*{\eta(x), 1 - \eta(x)}}$. Thus, the first
inequality (case $B \geq 1$) can be equivalently written as follows:
\begin{align*}
    \forall h \in \sH_{\mathrm{lin}},\; \sR_{\ell_{0-1}}(h) 
     \leq \sR_{\Phi_{\mathrm{hinge}}}(h) - \mathbb{E}_X\bracket*{\min\curl*{\eta(x), 1 - \eta(x)}},
\end{align*}
which is a more informative upper bound than the standard
inequality $\sR_{\ell_{0-1}}(h) 
     \leq \sR_{\Phi_{\mathrm{hinge}}}(h)$.
     
\subsubsection{Logistic Loss}
For the logistic loss $\Phi_{\mathrm{log}}(\alpha)\colon=\log_2(1+e^{-\alpha})$, for all $h\in \sH_{\mathrm{lin}}$ and $x\in \sX$:
\begin{equation*}
\begin{aligned}
& \sC_{\Phi_{\mathrm{log}}}(h,x,t)
 = t \Phi_{\mathrm{log}}(h(x))+(1-t)\Phi_{\mathrm{log}}(-h(x))\\
& = t\log_2\paren*{1+e^{-h(x)}}+(1-t)\log_2\paren*{1+e^{h(x)}}.\\
& \inf_{h\in\sH_{\mathrm{lin}}}\sC_{\Phi_{\mathrm{log}}}(h,x,t)\\
&=\begin{cases}
-t\log_2(t)-(1-t)\log_2(1-t) \\
\qquad \text{if }\log\abs*{\frac{t}{1-t}}\leq W\norm*{x}_p+B,\\
\max\curl*{t,1-t}\log_2\paren*{1+e^{-(W\norm*{x}_p+B)}}+\min\curl*{t,1-t}\log_2\paren*{1+e^{W\norm*{x}_p+B}}\\
\qquad \text{if }\log\abs*{\frac{t}{1-t}}> W\norm*{x}_p+B.
\end{cases}
\end{aligned}
\end{equation*}
Therefore, the $\paren*{\Phi_{\mathrm{log}},\sH_{\mathrm{lin}}}$-minimizability gap can be expressed as follows:
\begin{equation}
\begin{aligned}
\label{eq:M-log-lin}
& \sM_{\Phi_{\mathrm{log}}}\paren*{\sH_{\mathrm{lin}}}\\
& = \sR_{\Phi_{\mathrm{log}}}^*\paren*{\sH_{\mathrm{lin}}}-
\mathbb{E}_{X}\bracket*{\inf_{h\in\sH_{\mathrm{lin}}}\sC_{\Phi_{\mathrm{log}}}(h,x,\eta(x))}\\
& = \sR_{\Phi_{\mathrm{log}}}^*\paren*{\sH_{\mathrm{lin}}} -
\mathbb{E}_{X}\bracket*{-\eta(x)\log_2(\eta(x))-(1-\eta(x))\log_2(1-\eta(x))\mathds{1}_{\log\abs*{\frac{\eta(x)}{1-\eta(x)}}\leq W\norm*{x}_p+B}}\\
& - \mathbb{E}_{X}\bracket*{\max\curl*{\eta(x),1-\eta(x)}\log_2\paren*{1+e^{-(W\norm*{x}_p+B)}}\mathds{1}_{\log\abs*{\frac{\eta(x)}{1-\eta(x)}}> W\norm*{x}_p+B}}\\
&-\mathbb{E}_{X}\bracket*{\min\curl*{\eta(x),1-\eta(x)}\log_2\paren*{1+e^{W\norm*{x}_p+B}}\mathds{1}_{\log\abs*{\frac{\eta(x)}{1-\eta(x)}}> W\norm*{x}_p+B}}
\end{aligned}
\end{equation}
Note $\paren*{\Phi_{\mathrm{log}},\sH_{\mathrm{lin}}}$-minimizability gap coincides with
the $\paren*{\Phi_{\mathrm{log}},\sH_{\mathrm{lin}}}$-approximation error:\\
$\sR_{\Phi_{\mathrm{log}}}^*\paren*{\sH_{\mathrm{lin}}}-
\mathbb{E}_{X}\bracket*{-\eta(x)\log_2(\eta(x))-(1-\eta(x))\log_2(1-\eta(x))}$ for $B =+\infty$.

For $\frac{1}2< t\leq1$, we have
\begin{align*}
& \inf_{h\in\sH_{\mathrm{lin}}:h(x)<0}\sC_{\Phi_{\mathrm{log}}}(h,x,t)\\
& = t\log_2\paren*{1+e^{-0}}+(1-t)\log_2\paren*{1+e^{0}} \\
& = 1, \\
& \inf_{x\in \sX}\inf_{h\in\sH_{\mathrm{lin}}:h(x)<0}\Delta\sC_{\Phi_{\mathrm{log}},\sH_{\mathrm{lin}}}(h,x,t)\\
&=\inf_{x\in \sX}\paren*{\inf_{h\in\sH_{\mathrm{lin}}:h(x)<0}\sC_{\Phi_{\mathrm{log}}}(h,x,t)-\inf_{h\in\sH_{\mathrm{lin}}}\sC_{\Phi_{\mathrm{log}}}(h,x,t)}\\
&=\inf_{x\in \sX}\begin{cases}
1+t\log_2(t)+(1-t)\log_2(1-t)\\
\text{if }\log\frac{t}{1-t}\leq W\norm*{x}_p+B,\\
1-t\log_2\paren*{1+e^{-(W\norm*{x}_p+B)}}-(1-t)\log_2\paren*{1+e^{W\norm*{x}_p+B}}\\
\text{if }\log\frac{t}{1-t}> W\norm*{x}_p+B.
\end{cases}\\
&=\begin{cases}
1+t\log_2(t)+(1-t)\log_2(1-t) & \text{if }\log\frac{t}{1-t}\leq B,\\
1-t\log_2\paren*{1+e^{-B}}-(1-t)\log_2\paren*{1+e^{B}} & \text{if }\log\frac{t}{1-t}> B.
\end{cases}\\
&=\sT(2t-1),
\end{align*}
where $\sT$ is the increasing and convex function on $[0,1]$ defined by
\begin{align*}
\forall t\in[0,1], \quad
\sT(t)=\begin{cases}
\frac{t+1}{2}\log_2(t+1)+\frac{1-t}{2}\log_2(1-t),\quad &  t\leq \frac{e^B-1}{e^B+1},\\
1-\frac{t+1}{2}\log_2(1+e^{-B})-\frac{1-t}{2}\log_2(1+e^B),\quad & t> \frac{e^B-1}{e^B+1}.
\end{cases}
\end{align*}
By Definition~\ref{def:trans}, for any $\epsilon\geq 0$, the $\sH_{\mathrm{lin}}$-estimation error transformation of the logistic loss is as follows:
\begin{align*}
\sT_{\Phi_{\mathrm{log}}}= 
\begin{cases}
\sT(t), & t\in \left[\epsilon,1\right], \\
\frac{\sT(\epsilon)}{\epsilon}\, t, &  t\in \left[0,\epsilon\right).
\end{cases}
\end{align*}
Therefore, when $\epsilon=0$, $\sT_{\Phi_{\mathrm{log}}}$ is convex, non-decreasing, invertible and satisfies that $\sT_{\Phi_{\mathrm{log}}}(0)=0$. By Theorem~\ref{Thm:tightness}, we can choose $\Psi(t)=\sT_{\Phi_{\mathrm{log}}}(t)$ in Theorem~\ref{Thm:excess_bounds_Psi_uniform}, or equivalently $\Gamma(t)=\sT_{\Phi_{\mathrm{log}}}^{-1}(t)$ in Theorem~\ref{Thm:excess_bounds_Gamma_uniform}, which are optimal. To simplify the expression, using the fact that
\begin{align*}
\frac{t+1}{2}\log_2(t+1)+\frac{1-t}{2}\log_2(1-t)
&=
1-\paren*{-\frac{t+1}{2}\log_2\paren*{\frac{t+1}{2}}-\frac{1-t}{2}\log_2\paren*{\frac{1-t}{2}}}\\
&\geq 1 - \sqrt{4\frac{1-t}{2} \frac{t+1}{2}}\\
& = 1- \sqrt{1-t^2}\\
& \geq \frac{t^2}{2},\\
1-\frac{t+1}{2}\log_2(1+e^{-B})-\frac{1-t}{2}\log_2(1+e^B)
&=\frac{1}{2}\log_2\paren*{\frac{4}{2+e^{-B}+e^B}}+1/2\log_2\paren*{\frac{1+e^B}{1+e^{-B}}}\, t,
\end{align*}
$\sT_{\Phi_{\mathrm{log}}}$ can be lower bounded by
\begin{align*}
\wt{\sT}_{\Phi_{\mathrm{log}}}(t)= \begin{cases}
\frac{t^2}{2},& t\leq \frac{e^B-1}{e^B+1},\\
\frac{1}{2}\paren*{\frac{e^B-1}{e^B+1}}\, t, & t> \frac{e^B-1}{e^B+1}.
\end{cases}   
\end{align*}
Thus, we adopt an upper bound of $\sT_{\Phi_{\mathrm{log}}}^{-1}$ as follows:
\begin{align*}
\wt{\sT}_{\Phi_{\mathrm{log}}}^{-1}(t)=\begin{cases}
\sqrt{2t}, & t\leq \frac{1}{2}\paren*{\frac{e^B-1}{e^B+1}}^2,\\
2\paren*{\frac{e^B+1}{e^B-1}}\, t, & t> \frac{1}{2}\paren*{\frac{e^B-1}{e^B+1}}^2.
\end{cases}
\end{align*}
Therefore, by Theorem~\ref{Thm:excess_bounds_Psi_uniform} or Theorem~\ref{Thm:excess_bounds_Gamma_uniform}, setting $\e = 0$ yields the $\sH_{\mathrm{lin}}$-consistency bound for the logistic loss, valid for all $h \in \sH_{\mathrm{lin}}$:
\begin{multline}
\label{eq:log-lin-est}
     \sR_{\ell_{0-1}}(h)-\sR_{\ell_{0-1}}^*\paren*{\sH_{\mathrm{lin}}}+\sM_{\ell_{0-1}, \sH_{\mathrm{lin}}}\\
     \leq 
     \begin{cases}
     \sqrt{2}\,\paren*{\sR_{\Phi_{\mathrm{log}}}(h)- \sR_{\Phi_{\mathrm{log}}}^*\paren*{\sH_{\mathrm{lin}}}+\sM_{\Phi_{\mathrm{log}}}\paren*{\sH_{\mathrm{lin}}}}^{\frac12}, \\ \qquad \text{if } \sR_{\Phi_{\mathrm{log}}}(h)- \sR_{\Phi_{\mathrm{log}}}^*\paren*{\sH_{\mathrm{lin}}}\leq \frac{1}{2}\paren*{\frac{e^B-1}{e^B+1}}^2-\sM_{\Phi_{\mathrm{log}}}\paren*{\sH_{\mathrm{lin}}}\\
     2\paren*{\frac{e^B+1}{e^B-1}}\paren*{\sR_{\Phi_{\mathrm{log}}}(h)- \sR_{\Phi_{\mathrm{log}}}^*\paren*{\sH_{\mathrm{lin}}}+\sM_{\Phi_{\mathrm{log}}}\paren*{\sH_{\mathrm{lin}}}}, \\ \qquad \text{otherwise}
     \end{cases}
\end{multline}
Since the $\paren*{\ell_{0-1},\sH_{\mathrm{lin}}}$-minimizability gap coincides with
the $\paren*{\ell_{0-1},\sH_{\mathrm{lin}}}$-approximation error and 
$\paren*{\Phi_{\mathrm{log}},\sH_{\mathrm{lin}}}$-minimizability gap coincides with
the $\paren*{\Phi_{\mathrm{log}},\sH_{\mathrm{lin}}}$-approximation error for $B =+\infty$,
the inequality can be rewritten as follows:
\begin{align*}
     &\sR_{\ell_{0-1}}(h)- \sR_{\ell_{0-1}}^*\paren*{\sH_{\mathrm{all}}}\\
     &\quad \leq 
     \begin{cases}
      \sqrt{2}\,\bracket*{\sR_{\Phi_{\mathrm{log}}}(h) - \sR_{\Phi_{\mathrm{log}}}^*\paren*{\sH_{\mathrm{all}}}}^{\frac12} & \text{if } B = +\infty \\
     \begin{cases}
    \sqrt{2}\,\bracket*{\sR_{\Phi_{\mathrm{log}}}(h)- \sR_{\Phi_{\mathrm{log}}}^*\paren*{\sH_{\mathrm{lin}}}+\sM_{\Phi_{\mathrm{log}}}\paren*{\sH_{\mathrm{lin}}}}^{\frac12}  \\ \qquad \text{if } \sR_{\Phi_{\mathrm{log}}}(h)- \sR_{\Phi_{\mathrm{log}}}^*\paren*{\sH_{\mathrm{lin}}}\leq \frac{1}{2}\paren*{\frac{e^B-1}{e^B+1}}^2-\sM_{\Phi_{\mathrm{log}}}\paren*{\sH_{\mathrm{lin}}} \\
    2\paren*{\frac{e^B+1}{e^B-1}}\paren*{\sR_{\Phi_{\mathrm{log}}}(h)- \sR_{\Phi_{\mathrm{log}}}^*\paren*{\sH_{\mathrm{lin}}}+\sM_{\Phi_{\mathrm{log}}}\paren*{\sH_{\mathrm{lin}}}} \\ \qquad \text{otherwise}
    \end{cases} & \text{otherwise}
     \end{cases}
\end{align*}
where the $\paren*{\Phi_{\mathrm{log}},\sH_{\mathrm{lin}}}$-minimizability gap $\sM_{\Phi_{\mathrm{log}}}\paren*{\sH_{\mathrm{lin}}}$ is characterized as below, which is less than
the $\paren*{\Phi_{\mathrm{log}},\sH_{\mathrm{lin}}}$-approximation error when $B<+ \infty$:
\begin{align*}
& \sM_{\Phi_{\mathrm{log}}}\paren*{\sH_{\mathrm{lin}}}\\
& = \sR_{\Phi_{\mathrm{log}}}^*\paren*{\sH_{\mathrm{lin}}}-
\mathbb{E}_{X}\bracket*{-\eta(x)\log_2(\eta(x))-(1-\eta(x))\log_2(1-\eta(x))\mathds{1}_{\log\abs*{\frac{\eta(x)}{1-\eta(x)}}\leq W\norm*{x}_p+B}}\\
& - \mathbb{E}_{X}\bracket*{\max\curl*{\eta(x),1-\eta(x)}\log_2\paren*{1+e^{-(W\norm*{x}_p+B)}}\mathds{1}_{\log\abs*{\frac{\eta(x)}{1-\eta(x)}}> W\norm*{x}_p+B}}\\
&-\mathbb{E}_{X}\bracket*{\min\curl*{\eta(x),1-\eta(x)}\log_2\paren*{1+e^{W\norm*{x}_p+B}}\mathds{1}_{\log\abs*{\frac{\eta(x)}{1-\eta(x)}}> W\norm*{x}_p+B}}\\
& < \sR_{\Phi_{\mathrm{log}}}^*\paren*{\sH_{\mathrm{lin}}} - \mathbb{E}_{X}\bracket*{-\eta(x)\log_2(\eta(x))-(1-\eta(x))\log_2(1-\eta(x))}\\
& = \sR_{\Phi_{\mathrm{log}}}^*\paren*{\sH_{\mathrm{lin}}} - \sR_{\Phi_{\mathrm{log}}}^*\paren*{\sH_{\mathrm{all}}} .
\end{align*}
Therefore, the inequality for $B = + \infty$ coincides with the consistency excess error bound
known for the logistic loss \citep{Zhang2003,MohriRostamizadehTalwalkar2018} but the one for $B< + \infty$ is distinct and novel.

\subsubsection{Exponential Loss}
For the exponential loss $\Phi_{\mathrm{exp}}(\alpha)\colon=e^{-\alpha}$, for all $h\in \sH_{\mathrm{lin}}$ and $x\in \sX$:
\begin{equation*}
\begin{aligned}
\sC_{\Phi_{\mathrm{exp}}}(h,x,t)
&=t \Phi_{\mathrm{exp}}(h(x))+(1-t)\Phi_{\mathrm{exp}}(-h(x))\\
&=te^{-h(x)}+(1-t)e^{h(x)}.\\
\inf_{h\in\sH_{\mathrm{lin}}}\sC_{\Phi_{\mathrm{exp}}}(h,x,t)
&=\begin{cases}
2\sqrt{t(1-t)} \\ \qquad \text{if }1/2\log\abs*{\frac{t}{1-t}}\leq W\norm*{x}_p+B\\
\max\curl*{t,1-t}e^{-(W\norm*{x}_p+B)}+\min\curl*{t,1-t}e^{W\norm*{x}_p+B} \\ \qquad \text{if }1/2\log\abs*{\frac{t}{1-t}}> W\norm*{x}_p+B.
\end{cases}
\end{aligned}
\end{equation*}
Therefore, the $\paren*{\Phi_{\mathrm{exp}},\sH_{\mathrm{lin}}}$-minimizability gap can be expressed as follows:
\begin{equation}
\begin{aligned}
\label{eq:M-exp-lin}
\sM_{\Phi_{\mathrm{exp}}}\paren*{\sH_{\mathrm{lin}}}
& = \sR_{\Phi_{\mathrm{exp}}}^*\paren*{\sH_{\mathrm{lin}}}-
\mathbb{E}_{X}\bracket*{\inf_{h\in\sH_{\mathrm{lin}}}\sC_{\Phi_{\mathrm{exp}}}(h,x,\eta(x))}\\
& = \sR_{\Phi_{\mathrm{exp}}}^*\paren*{\sH_{\mathrm{lin}}}-
\mathbb{E}_{X}\bracket*{2\sqrt{\eta(x)(1-\eta(x))}\mathds{1}_{1/2\log\abs*{\frac{\eta(x)}{1-\eta(x)}}\leq W\norm*{x}_p+B}}\\
& - \mathbb{E}_{X}\bracket*{\max\curl*{\eta(x),1-\eta(x)}e^{-(W\norm*{x}_p+B)}\mathds{1}_{1/2\log\abs*{\frac{\eta(x)}{1-\eta(x)}}> W\norm*{x}_p+B}}\\
&-\mathbb{E}_{X}\bracket*{\min\curl*{\eta(x),1-\eta(x)}e^{W\norm*{x}_p+B}\mathds{1}_{1/2\log\abs*{\frac{\eta(x)}{1-\eta(x)}}> W\norm*{x}_p+B}}.
\end{aligned}
\end{equation}
Note $\paren*{\Phi_{\mathrm{exp}},\sH_{\mathrm{lin}}}$-minimizability gap coincides with
the $\paren*{\Phi_{\mathrm{exp}},\sH_{\mathrm{lin}}}$-approximation error:\\ $\sR_{\Phi_{\mathrm{exp}}}^*\paren*{\sH_{\mathrm{lin}}}-
\mathbb{E}_{X}\bracket*{2\sqrt{\eta(x)(1-\eta(x))}}$ for $B =+\infty$.

For $\frac{1}2< t\leq1$, we have
\begin{align*}
&\inf_{h\in\sH_{\mathrm{lin}}:h(x)<0}\sC_{\Phi_{\mathrm{exp}}}(h,x,t)\\
&=te^{-0}+(1-t)e^{0}\\
&=1.\\
&\inf_{x\in \sX}\inf_{h\in\sH_{\mathrm{lin}}:h(x)<0}\Delta\sC_{\Phi_{\mathrm{exp}},\sH_{\mathrm{lin}}}(h,x,t)\\
&=\inf_{x\in \sX}\paren*{\inf_{h\in\sH_{\mathrm{lin}}:h(x)<0}\sC_{\Phi_{\mathrm{exp}}}(h,x,t)-\inf_{h\in\sH_{\mathrm{lin}}}\sC_{\Phi_{\mathrm{exp}}}(h,x,t)}\\
&=\inf_{x\in \sX}\begin{cases}
1-2\sqrt{t(1-t)} & \text{if }1/2\log\frac{t}{1-t}\leq W\norm*{x}_p+B,\\
1-te^{-(W\norm*{x}_p+B)}-(1-t)e^{W\norm*{x}_p+B} & \text{if }1/2\log\frac{t}{1-t}> W\norm*{x}_p+B.
\end{cases}\\
&=\begin{cases}
1-2\sqrt{t(1-t)}, & 1/2\log\frac{t}{1-t}\leq B\\
1-te^{-B}-(1-t)e^{B}, & 1/2\log\frac{t}{1-t}> B
\end{cases}\\
&=\sT(2t-1),
\end{align*}
where $\sT$ is the increasing and convex function on $[0,1]$ defined by
\begin{align*}
\forall t\in[0,1], \quad 
\sT(t)=\begin{cases}
1-\sqrt{1-t^2}, & t\leq \frac{e^{2B}-1}{e^{2B}+1},\\
1-\frac{t+1}{2}e^{-B}-\frac{1-t}{2}e^B, & t> \frac{e^{2B}-1}{e^{2B}+1}.
\end{cases}
\end{align*}
By Definition~\ref{def:trans}, for any $\epsilon\geq 0$, the $\sH_{\mathrm{lin}}$-estimation error transformation of the exponential loss is as follows:
\begin{align*}
\sT_{\Phi_{\mathrm{exp}}}= 
\begin{cases}
\sT(t), & t\in \left[\epsilon,1\right], \\
\frac{\sT(\epsilon)}{\epsilon}\, t, &  t\in \left[0,\epsilon\right).
\end{cases}
\end{align*}
Therefore, when $\epsilon=0$, $\sT_{\Phi_{\mathrm{exp}}}$ is convex, non-decreasing, invertible and satisfies that $\sT_{\Phi_{\mathrm{exp}}}(0)=0$. By Theorem~\ref{Thm:tightness}, we can choose $\Psi(t)=\sT_{\Phi_{\mathrm{exp}}}(t)$ in Theorem~\ref{Thm:excess_bounds_Psi_uniform}, or equivalently $\Gamma(t)=\sT_{\Phi_{\mathrm{exp}}}^{-1}(t)$ in Theorem~\ref{Thm:excess_bounds_Gamma_uniform}, which are optimal. To simplify the expression, using the fact that
\begin{align*}
1- \sqrt{1-t^2} & \geq \frac{t^2}{2}, \\
1-\frac{t+1}{2}e^{-B}-\frac{1-t}{2}e^B & = 1-1/2 e^B-1/2 e^{-B}+\frac{e^B-e^{-B}}2\, t,
\end{align*}
$\sT_{\Phi_{\mathrm{exp}}}$ can be lower bounded by
\begin{align*}
\wt{\sT}_{\Phi_{\mathrm{exp}}}(t)= \begin{cases}
\frac{t^2}{2},& t\leq \frac{e^{2B}-1}{e^{2B}+1},\\
\frac{1}{2}\paren*{\frac{e^{2B}-1}{e^{2B}+1}}\, t, & t> \frac{e^{2B}-1}{e^{2B}+1}.
\end{cases}   
\end{align*}
Thus, we adopt an upper bound of $\sT_{\Phi_{\mathrm{exp}}}^{-1}$ as follows:
\begin{align*}
\wt{\sT}_{\Phi_{\mathrm{exp}}}^{-1}(t)=
\begin{cases}
\sqrt{2t}, & t\leq \frac{1}{2}\paren*{\frac{e^{2B}-1}{e^{2B}+1}}^2,\\
2\paren*{\frac{e^{2B}+1}{e^{2B}-1}}\, t, & t> \frac{1}{2}\paren*{\frac{e^{2B}-1}{e^{2B}+1}}^2.
\end{cases}
\end{align*}
Therefore, by Theorem~\ref{Thm:excess_bounds_Psi_uniform} or Theorem~\ref{Thm:excess_bounds_Gamma_uniform}, setting $\e = 0$ yields the $\sH_{\mathrm{lin}}$-consistency bound for the exponential loss, valid for all $h \in \sH_{\mathrm{lin}}$:
\begin{multline}
\label{eq:exp-lin-est}
     \sR_{\ell_{0-1}}(h)-\sR_{\ell_{0-1}}^*\paren*{\sH_{\mathrm{lin}}}+\sM_{\ell_{0-1}, \sH_{\mathrm{lin}}}\\
     \leq 
     \begin{cases}
     \sqrt{2}\,\paren*{\sR_{\Phi_{\mathrm{exp}}}(h)- \sR_{\Phi_{\mathrm{exp}}}^*\paren*{\sH_{\mathrm{lin}}}+\sM_{\Phi_{\mathrm{exp}}}\paren*{\sH_{\mathrm{lin}}}}^{\frac12}, \\ \qquad \text{if } \sR_{\Phi_{\mathrm{exp}}}(h)- \sR_{\Phi_{\mathrm{exp}}}^*\paren*{\sH_{\mathrm{lin}}}\leq \frac{1}{2}\paren*{\frac{e^{2B}-1}{e^{2B}+1}}^2-\sM_{\Phi_{\mathrm{exp}}}\paren*{\sH_{\mathrm{lin}}},\\
     2\paren*{\frac{e^{2B}+1}{e^{2B}-1}}\paren*{\sR_{\Phi_{\mathrm{exp}}}(h)- \sR_{\Phi_{\mathrm{exp}}}^*\paren*{\sH_{\mathrm{lin}}}+\sM_{\Phi_{\mathrm{exp}}}\paren*{\sH_{\mathrm{lin}}}}, \\ \qquad \text{otherwise}.
     \end{cases}
\end{multline}
Since the $\paren*{\ell_{0-1},\sH_{\mathrm{lin}}}$-minimizability gap coincides with
the $\paren*{\ell_{0-1},\sH_{\mathrm{lin}}}$-approximation error and
$\paren*{\Phi_{\mathrm{exp}},\sH_{\mathrm{lin}}}$-minimizability gap coincides with
the $\paren*{\Phi_{\mathrm{exp}},\sH_{\mathrm{lin}}}$-approximation error for $B =+\infty$,
the inequality can be rewritten as follows:
\begin{multline*}
     \sR_{\ell_{0-1}}(h)- \sR_{\ell_{0-1}}^*\paren*{\sH_{\mathrm{all}}}\\
     \leq 
     \begin{cases}
      \sqrt{2}\,\bracket*{\sR_{\Phi_{\mathrm{exp}}}(h) - \sR_{\Phi_{\mathrm{exp}}}^*\paren*{\sH_{\mathrm{all}}}}^{\frac12} & \text{if } B = +\infty, \\
     \begin{cases}
    \sqrt{2}\,\bracket*{\sR_{\Phi_{\mathrm{exp}}}(h)- \sR_{\Phi_{\mathrm{exp}}}^*\paren*{\sH_{\mathrm{lin}}}+\sM_{\Phi_{\mathrm{exp}}}\paren*{\sH_{\mathrm{lin}}}}^{\frac12} \\ \qquad \text{if } \sR_{\Phi_{\mathrm{exp}}}(h)- \sR_{\Phi_{\mathrm{exp}}}^*\paren*{\sH_{\mathrm{lin}}}\leq \frac{1}{2}\paren*{\frac{e^{2B}-1}{e^{2B}+1}}^2-\sM_{\Phi_{\mathrm{exp}}}\paren*{\sH_{\mathrm{lin}}}, \\
    2\paren*{\frac{e^{2B}+1}{e^{2B}-1}}\paren*{\sR_{\Phi_{\mathrm{exp}}}(h)- \sR_{\Phi_{\mathrm{exp}}}^*\paren*{\sH_{\mathrm{lin}}}+\sM_{\Phi_{\mathrm{exp}}}\paren*{\sH_{\mathrm{lin}}}} \\ \qquad \text{otherwise}.
    \end{cases} & \text{otherwise}.
     \end{cases}
\end{multline*}
where the $\paren*{\Phi_{\mathrm{exp}},\sH_{\mathrm{lin}}}$-minimizability gap $\sM_{\Phi_{\mathrm{exp}}}\paren*{\sH_{\mathrm{lin}}}$ is characterized as below, which is less than
the $\paren*{\Phi_{\mathrm{exp}},\sH_{\mathrm{lin}}}$-approximation error when $B<+ \infty$:
\begin{align*}
\sM_{\Phi_{\mathrm{exp}}}\paren*{\sH_{\mathrm{lin}}}
& = \sR_{\Phi_{\mathrm{exp}}}^*\paren*{\sH_{\mathrm{lin}}}-
\mathbb{E}_{X}\bracket*{2\sqrt{\eta(x)(1-\eta(x))}\mathds{1}_{1/2\log\abs*{\frac{\eta(x)}{1-\eta(x)}}\leq W\norm*{x}_p+B}}\\
& - \mathbb{E}_{X}\bracket*{\max\curl*{\eta(x),1-\eta(x)}e^{-(W\norm*{x}_p+B)}\mathds{1}_{1/2\log\abs*{\frac{\eta(x)}{1-\eta(x)}}> W\norm*{x}_p+B}}\\
&-\mathbb{E}_{X}\bracket*{\min\curl*{\eta(x),1-\eta(x)}e^{W\norm*{x}_p+B}\mathds{1}_{1/2\log\abs*{\frac{\eta(x)}{1-\eta(x)}}> W\norm*{x}_p+B}}\\
& < \sR_{\Phi_{\mathrm{exp}}}^*\paren*{\sH_{\mathrm{lin}}} - \mathbb{E}_{X}\bracket*{2\sqrt{\eta(x)(1-\eta(x))}}\\
& = \sR_{\Phi_{\mathrm{exp}}}^*\paren*{\sH_{\mathrm{lin}}} - \sR_{\Phi_{\mathrm{exp}}}^*\paren*{\sH_{\mathrm{all}}} .
\end{align*}
Therefore, the inequality for $B = + \infty$ coincides with the consistency excess error bound
known for the exponential loss \citep{Zhang2003,MohriRostamizadehTalwalkar2018} but the one for $B< + \infty$ is distinct and novel.

\subsubsection{Quadratic Loss}
For the quadratic loss $\Phi_{\mathrm{quad}}(\alpha)\colon=(1-\alpha)^2\mathds{1}_{\alpha\leq 1}$, for all $h\in \sH_{\mathrm{lin}}$ and $x\in \sX$:
\begin{equation*}
\begin{aligned}
\sC_{\Phi_{\mathrm{quad}}}(h,x,t)
&=t \Phi_{\mathrm{quad}}(h(x))+(1-t)\Phi_{\mathrm{quad}}(-h(x))\\
&=t\paren*{1-h(x)}^2\mathds{1}_{h(x)\leq 1}+(1-t)\paren*{1+h(x)}^2\mathds{1}_{h(x)\geq -1}.\\
\inf_{h\in\sH_{\mathrm{lin}}}\sC_{\Phi_{\mathrm{quad}}}(h,x,t)
&=\begin{cases}
4t(1-t), \\ \qquad \abs*{2t-1}\leq W\norm*{x}_p+B,\\
\max\curl*{t,1-t}\paren*{1-\paren*{W\norm*{x}_p+B}}^2
+\min\curl*{t,1-t}\paren*{1+W\norm*{x}_p+B}^2, \\ \qquad \abs*{2t-1}> W\norm*{x}_p+B.
\end{cases}
\end{aligned}
\end{equation*}
Therefore, the $\paren*{\Phi_{\mathrm{quad}},\sH_{\mathrm{lin}}}$-minimizability gap can be expressed as follows:
\begin{equation}
\begin{aligned}
\label{eq:M-quad-lin}
\sM_{\Phi_{\mathrm{quad}}}\paren*{\sH_{\mathrm{lin}}}
& = \sR_{\Phi_{\mathrm{quad}}}^*\paren*{\sH_{\mathrm{lin}}}-\mathbb{E}_{X}\bracket*{4\eta(x)(1-\eta(x))\mathds{1}_{\abs*{2\eta(x)-1}\leq W\norm*{x}_p+B}}\\
& - \mathbb{E}_{X}\bracket*{\max\curl*{\eta(x),1-\eta(x)}\paren*{1-\paren*{W\norm*{x}_p+B}}^2\mathds{1}_{\abs*{2\eta(x)-1}> W\norm*{x}_p+B}}\\
& - \mathbb{E}_{X}\bracket*{\min\curl*{\eta(x),1-\eta(x)}\paren*{1+\paren*{W\norm*{x}_p+B}}^2\mathds{1}_{\abs*{2\eta(x)-1}> W\norm*{x}_p+B}}
\end{aligned}
\end{equation}
Note $\paren*{\Phi_{\mathrm{quad}},\sH_{\mathrm{lin}}}$-minimizability gap coincides with
the $\paren*{\Phi_{\mathrm{quad}},\sH_{\mathrm{lin}}}$-approximation error:\\ $\sR_{\Phi_{\mathrm{quad}}}^*\paren*{\sH_{\mathrm{lin}}}-
\mathbb{E}_{X}\bracket*{4\eta(x)(1-\eta(x))}$ for $B \geq 1$.

For $\frac{1}2< t\leq1$, we have
\begin{align*}
&\inf_{h\in\sH_{\mathrm{lin}}:h(x)<0}\sC_{\Phi_{\mathrm{quad}}}(h,x,t)\\
&=t+(1-t)\\
&=1\\
&\inf_{x\in \sX}\inf_{h\in\sH_{\mathrm{lin}}:h(x)<0}\Delta\sC_{\Phi_{\mathrm{quad}},\sH_{\mathrm{lin}}}(h,x,t)\\
& =\inf_{x\in \sX}\paren*{\inf_{h\in\sH_{\mathrm{lin}}:h(x)<0}\sC_{\Phi_{\mathrm{quad}}}(h,x,t)-\inf_{h\in\sH_{\mathrm{lin}}}\sC_{\Phi_{\mathrm{quad}}}(h,x,t)}\\
&=\inf_{x\in \sX}\begin{cases}
1-4t(1-t),& 2t-1\leq W\norm*{x}_p+B,\\
1-t\paren*{1-\paren*{W\norm*{x}_p+B}}^2-(1-t)\paren*{1+W\norm*{x}_p+B}^2, & 2t-1> W\norm*{x}_p+B.
\end{cases}\\
&=\begin{cases}
1-4t(1-t),& 2t-1\leq B,\\
1-t\paren*{1-B}^2-(1-t)\paren*{1+B}^2,& 2t-1> B.
\end{cases}\\
&=\sT(2t-1)
\end{align*}
where $\sT$ is the increasing and convex function on $[0,1]$ defined by
\begin{align*}
\forall t\in[0,1], \quad
\sT(t)=\begin{cases}
t^2, & t\leq B,\\
2B \,t-B^2, & t> B.
\end{cases}
\end{align*}
By Definition~\ref{def:trans}, for any $\epsilon\geq 0$, the $\sH_{\mathrm{lin}}$-estimation error transformation of the quadratic loss is as follows:
\begin{align*}
\sT_{\Phi_{\mathrm{quad}}}= 
\begin{cases}
\sT(t), & t\in \left[\epsilon,1\right], \\
\frac{\sT(\epsilon)}{\epsilon}\, t, &  t\in \left[0,\epsilon\right).
\end{cases}
\end{align*}
Therefore, when $\epsilon=0$, $\sT_{\Phi_{\mathrm{quad}}}$ is convex, non-decreasing, invertible and satisfies that $\sT_{\Phi_{\mathrm{quad}}}(0)=0$. By Theorem~\ref{Thm:tightness}, we can choose $\Psi(t)=\sT_{\Phi_{\mathrm{quad}}}(t)$ in Theorem~\ref{Thm:excess_bounds_Psi_uniform}, or equivalently $\Gamma(t) = \sT_{\Phi_{\mathrm{quad}}}^{-1}(t)=
\begin{cases}
\sqrt{t}, & t \leq B^2 \\
\frac{t}{2B}+\frac{B}{2}, & t > B^2
\end{cases}$, in Theorem~\ref{Thm:excess_bounds_Gamma_uniform}, which are optimal. Thus, by Theorem~\ref{Thm:excess_bounds_Psi_uniform} or Theorem~\ref{Thm:excess_bounds_Gamma_uniform}, setting $\e = 0$ yields the $\sH_{\mathrm{lin}}$-consistency bound for the quadratic loss, valid for all $h \in \sH_{\mathrm{lin}}$:
\begin{multline}
\label{eq:quad-lin-est}
    \sR_{\ell_{0-1}}(h)- \sR_{\ell_{0-1}}^*\paren*{\sH_{\mathrm{lin}}}+\sM_{\ell_{0-1}}\paren*{\sH_{\mathrm{lin}}}\\
    \leq
    \begin{cases}
    \bracket*{\sR_{\Phi_{\mathrm{quad}}}(h)- \sR_{\Phi_{\mathrm{quad}}}^*\paren*{\sH_{\mathrm{lin}}}+\sM_{\Phi_{\mathrm{quad}}}\paren*{\sH_{\mathrm{lin}}}}^{\frac12}  \\ \qquad \text{if } \sR_{\Phi_{\mathrm{quad}}}(h)- \sR_{\Phi_{\mathrm{quad}}}^*\paren*{\sH_{\mathrm{lin}}}\leq B^2-\sM_{\Phi_{\mathrm{quad}}}\paren*{\sH_{\mathrm{lin}}} \\
    \frac{\sR_{\Phi_{\mathrm{quad}}}(h)- \sR_{\Phi_{\mathrm{quad}}}^*\paren*{\sH_{\mathrm{lin}}}+\sM_{\Phi_{\mathrm{quad}}}\paren*{\sH_{\mathrm{lin}}}}{2B}+\frac{B}{2} \\ \qquad \text{otherwise}
    \end{cases}
\end{multline}
Since the $\paren*{\ell_{0-1},\sH_{\mathrm{lin}}}$-minimizability gap coincides with
the $\paren*{\ell_{0-1},\sH_{\mathrm{lin}}}$-approximation error and
$\paren*{\Phi_{\mathrm{quad}},\sH_{\mathrm{lin}}}$-minimizability gap coincides with
the $\paren*{\Phi_{\mathrm{quad}},\sH_{\mathrm{lin}}}$-approximation error for $B \geq 1$,
the inequality can be rewritten as follows:
\begin{multline*}
     \sR_{\ell_{0-1}}(h)- \sR_{\ell_{0-1}}^*\paren*{\sH_{\mathrm{all}}}\\
     \leq 
     \begin{cases}
     \bracket*{\sR_{\Phi_{\mathrm{quad}}}(h) - \sR_{\Phi_{\mathrm{quad}}}^*\paren*{\sH_{\mathrm{all}}}}^{\frac12} & \text{if } B \geq 1\\
     \begin{cases}
    \bracket*{\sR_{\Phi_{\mathrm{quad}}}(h)- \sR_{\Phi_{\mathrm{quad}}}^*\paren*{\sH_{\mathrm{lin}}}+\sM_{\Phi_{\mathrm{quad}}}\paren*{\sH_{\mathrm{lin}}}}^{\frac12}  \\ \qquad \text{if } \sR_{\Phi_{\mathrm{quad}}}(h)- \sR_{\Phi_{\mathrm{quad}}}^*\paren*{\sH_{\mathrm{lin}}}\leq B^2-\sM_{\Phi_{\mathrm{quad}}}\paren*{\sH_{\mathrm{lin}}} \\
    \frac{\sR_{\Phi_{\mathrm{quad}}}(h)- \sR_{\Phi_{\mathrm{quad}}}^*\paren*{\sH_{\mathrm{lin}}}+\sM_{\Phi_{\mathrm{quad}}}\paren*{\sH_{\mathrm{lin}}}}{2B}+\frac{B}{2} \\ \qquad \text{otherwise}
    \end{cases} & \text{otherwise}
     \end{cases}
\end{multline*}
where the $\paren*{\Phi_{\mathrm{quad}},\sH_{\mathrm{lin}}}$-minimizability gap $\sM_{\Phi_{\mathrm{quad}},\sH_{\mathrm{lin}}}$ is characterized as below, which is less than
the $\paren*{\Phi_{\mathrm{quad}}}\paren*{\sH_{\mathrm{lin}}}$-approximation error when $B<1$:
\begin{align*}
\sM_{\Phi_{\mathrm{quad}}}\paren*{\sH_{\mathrm{lin}}}
& = \sR_{\Phi_{\mathrm{quad}}}^*\paren*{\sH_{\mathrm{lin}}}-\mathbb{E}_{X}\bracket*{4\eta(x)(1-\eta(x))\mathds{1}_{\abs*{2\eta(x)-1}\leq W\norm*{x}_p+B}}\\
& - \mathbb{E}_{X}\bracket*{\max\curl*{\eta(x),1-\eta(x)}\paren*{1-\paren*{W\norm*{x}_p+B}}^2\mathds{1}_{\abs*{2\eta(x)-1}> W\norm*{x}_p+B}}\\
& - \mathbb{E}_{X}\bracket*{\min\curl*{\eta(x),1-\eta(x)}\paren*{1+\paren*{W\norm*{x}_p+B}}^2\mathds{1}_{\abs*{2\eta(x)-1}> W\norm*{x}_p+B}}\\
& \leq \sR_{\Phi_{\mathrm{quad}}}^*\paren*{\sH_{\mathrm{lin}}} - \mathbb{E}_{X}\bracket*{4\eta(x)(1-\eta(x))}\\
& = \sR_{\Phi_{\mathrm{quad}}}^*\paren*{\sH_{\mathrm{lin}}} - \sR_{\Phi_{\mathrm{quad}}}^*\paren*{\sH_{\mathrm{all}}} .
\end{align*}
Therefore, the inequality for $B \geq 1$ coincides with the consistency excess error bound
known for the quadratic loss \citep{Zhang2003,bartlett2006convexity} but the one for $B< 1$ is distinct and novel.

\subsubsection{Sigmoid Loss}
\label{app:sig-lin}
For the sigmoid loss $\Phi_{\mathrm{sig}}(\alpha)\colon=1-\tanh(k\alpha),~k>0$,
for all $h\in \sH_{\mathrm{lin}}$ and $x\in \sX$:
\begin{equation*}
\begin{aligned}
\sC_{\Phi_{\mathrm{sig}}}(h,x,t)
&=t \Phi_{\mathrm{sig}}(h(x))+(1-t)\Phi_{\mathrm{sig}}(-h(x)),\\
&=t\paren*{1-\tanh(kh(x))}+(1-t)\paren*{1+\tanh(kh(x))}.\\
\inf_{h\in\sH_{\mathrm{lin}}}\sC_{\Phi_{\mathrm{sig}}}(h,x,t)
&=1-\abs*{1-2t}\tanh\paren*{k\paren*{W\norm*{x}_p+B}}
\end{aligned}
\end{equation*}
Therefore, the $\paren*{\Phi_{\mathrm{sig}},\sH_{\mathrm{lin}}}$-minimizability gap can be expressed as follows:
\begin{equation}
\begin{aligned}
\label{eq:M-sig-lin}
\sM_{\Phi_{\mathrm{sig}}}\paren*{\sH_{\mathrm{lin}}}
&= \sR_{\Phi_{\mathrm{sig}}}^*\paren*{\sH_{\mathrm{lin}}}-\mathbb{E}_{X}\bracket*{\inf_{h\in\sH_{\mathrm{lin}}}\sC_{\Phi_{\mathrm{sig}}}(h,x,\eta(x))}\\
&= \sR_{\Phi_{\mathrm{sig}}}^*\paren*{\sH_{\mathrm{lin}}}-\mathbb{E}_{X}\bracket*{1-\abs*{1-2\eta(x)}\tanh\paren*{k\paren*{W\norm*{x}_p+B}}}.
\end{aligned}
\end{equation}
Note $\paren*{\Phi_{\mathrm{sig}},\sH_{\mathrm{lin}}}$-minimizability gap coincides with
the $\paren*{\Phi_{\mathrm{sig}},\sH_{\mathrm{lin}}}$-approximation error:\\ $\sR_{\Phi_{\mathrm{sig}}}^*\paren*{\sH_{\mathrm{lin}}}-\mathbb{E}_{X}\bracket*{1-\abs*{1-2\eta(x)}}$ for $B = + \infty$.

For $\frac{1}2< t\leq1$, we have
\begin{align*}
\inf_{h\in\sH_{\mathrm{lin}}:h(x)<0}\sC_{\Phi_{\mathrm{sig}}}(h,x,t)
&=1-\abs*{1-2t}\tanh(0)\\
&=1.\\
\inf_{x\in \sX}\inf_{h\in\sH_{\mathrm{lin}}:h(x)<0}\Delta\sC_{\Phi_{\mathrm{sig}},\sH_{\mathrm{lin}}}(h,x,t)
&=\inf_{x\in \sX}\paren*{\inf_{h\in\sH_{\mathrm{lin}}:h(x)<0}\sC_{\Phi_{\mathrm{sig}}}(h,x,t)-\inf_{h\in\sH_{\mathrm{lin}}}\sC_{\Phi_{\mathrm{sig}}}(h,x,t)}\\
&=\inf_{x\in \sX}(2t-1)\tanh\paren*{k\paren*{W\norm*{x}_p+B}}\\
&=(2t-1)\tanh(kB)\\
&=\sT(2t-1)
\end{align*}
where $\sT$ is the increasing and convex function on $[0,1]$ defined by
\begin{align*}
\forall t\in[0,1],\; \sT(t)=\tanh(kB) \, t .
\end{align*}
By Definition~\ref{def:trans}, for any $\epsilon\geq 0$, the $\sH_{\mathrm{lin}}$-estimation error transformation of the sigmoid loss is as follows:
\begin{align*}
\sT_{\Phi_{\mathrm{sig}}}= \tanh(kB) \, t, \quad t \in [0,1],
\end{align*}
Therefore, $\sT_{\Phi_{\mathrm{sig}}}$ is convex, non-decreasing, invertible and satisfies that $\sT_{\Phi_{\mathrm{sig}}}(0)=0$. By Theorem~\ref{Thm:tightness}, we can choose $\Psi(t)=\tanh(kB)\,t$ in Theorem~\ref{Thm:excess_bounds_Psi_uniform}, or equivalently $\Gamma(t)=\frac{t}{\tanh(kB)}$ in Theorem~\ref{Thm:excess_bounds_Gamma_uniform}, which are optimal when $\e=0$.
Thus, by Theorem~\ref{Thm:excess_bounds_Psi_uniform} or Theorem~\ref{Thm:excess_bounds_Gamma_uniform}, setting $\e = 0$ yields the $\sH_{\mathrm{lin}}$-consistency bound for the sigmoid loss, valid for all $h \in \sH_{\mathrm{lin}}$:
\begin{align}
\label{eq:sig-lin-est}
     \sR_{\ell_{0-1}}(h)- \sR_{\ell_{0-1}}^*\paren*{\sH_{\mathrm{lin}}}\leq \frac{\sR_{\Phi_{\mathrm{sig}}}(h)- \sR_{\Phi_{\mathrm{sig}}}^*\paren*{\sH_{\mathrm{lin}}}+\sM_{\Phi_{\mathrm{sig}}}\paren*{\sH_{\mathrm{lin}}}}{\tanh(kB)}-\sM_{\ell_{0-1}}\paren*{\sH_{\mathrm{lin}}}.
\end{align}
Since the $\paren*{\ell_{0-1},\sH_{\mathrm{lin}}}$-minimizability gap coincides with
the $\paren*{\ell_{0-1},\sH_{\mathrm{lin}}}$-approximation error and
$\paren*{\Phi_{\mathrm{sig}},\sH_{\mathrm{lin}}}$-minimizability gap coincides with
the $\paren*{\Phi_{\mathrm{sig}},\sH_{\mathrm{lin}}}$-approximation error for $B = + \infty$,
the inequality can be rewritten as follows:
\begin{multline}
\label{eq:sig-lin-est-2}
     \sR_{\ell_{0-1}}(h)- \sR_{\ell_{0-1}}^*\paren*{\sH_{\mathrm{all}}}\\
     \leq 
     \begin{cases}
     \sR_{\Phi_{\mathrm{sig}}}(h) - \sR_{\Phi_{\mathrm{sig}}}^*\paren*{\sH_{\mathrm{all}}} & \text{if } B = + \infty\\
     \frac{1}{\tanh(kB)} \bracket[\Big]{\sR_{\Phi_{\mathrm{sig}}}(h)
     - \mathbb{E}_{X}\bracket*{1-\abs*{1-2\eta(x)}\tanh\paren*{k\paren*{W\norm*{x}_p+B}}} } & \text{otherwise}.
     \end{cases}
\end{multline}
The inequality for $B = + \infty$ coincides with the consistency excess error bound
known for the sigmoid loss \citep{Zhang2003,bartlett2006convexity,MohriRostamizadehTalwalkar2018} but the one for $B < + \infty$ is distinct and novel.  For $B<+ \infty$, we have
\begin{align*}
& \mathbb{E}_{X}\bracket*{1-\abs*{1-2\eta(x)}\tanh\paren*{k\paren*{W\norm*{x}_p+B}}}\\
&\qquad > \mathbb{E}_{X}\bracket*{1-\abs*{2\eta(x)-1}}= 2\mathbb{E}_X\bracket*{\min\curl*{\eta(x), 1 - \eta(x)}}= \sR_{\Phi_{\mathrm{hinge}}}^*\paren*{\sH_{\mathrm{all}}}.
\end{align*}
Therefore for $B<+ \infty$,
\begin{align*}
\sR_{\Phi_{\mathrm{sig}}}(h) - \mathbb{E}_{X}\bracket*{1-\abs*{1-2\eta(x)}\tanh\paren*{k\paren*{W\norm*{x}_p+B}}} < \sR_{\Phi_{\mathrm{sig}}}(h) - \sR_{\Phi_{\mathrm{sig}}}^*\paren*{\sH_{\mathrm{all}}}.
\end{align*}
Note that: $\sR_{\Phi_{\mathrm{sig}}}^*\paren*{\sH_{\mathrm{all}}} = 2 \sR_{\ell_{0-1}}^*\paren*{\sH_{\mathrm{all}}} =2\mathbb{E}_X\bracket*{\min\curl*{\eta(x), 1 - \eta(x)}}$. Thus, the first
inequality (case $B = + \infty$) can be equivalently written as follows:
\begin{align*}
    \forall h \in \sH_{\mathrm{lin}}, \quad \sR_{\ell_{0-1}}(h) 
     \leq \sR_{\Phi_{\mathrm{sig}}}(h) - \mathbb{E}_X\bracket*{\min\curl*{\eta(x), 1 - \eta(x)}},
\end{align*}
which is a more informative upper bound than the standard
inequality $\sR_{\ell_{0-1}}(h) 
     \leq \sR_{\Phi_{\mathrm{sig}}}(h)$.

\subsubsection{\texorpdfstring{$\rho$}{rho}-Margin Loss}
\label{app:rho-lin}
For the $\rho$-margin loss $\Phi_{\rho}(\alpha)\colon=\min\curl*{1,\max\curl*{0,1-\frac{\alpha}{\rho}}},~\rho>0$,
for all $h\in \sH_{\mathrm{lin}}$ and $x\in \sX$:
\begin{equation*}
\begin{aligned}
\sC_{\Phi_{\rho}}(h,x,t)
&=t \Phi_{\rho}(h(x))+(1-t)\Phi_{\rho}(-h(x)),\\
&=t\min\curl*{1,\max\curl*{0,1-\frac{h(x)}{\rho}}}+(1-t)\min\curl*{1,\max\curl*{0,1+\frac{h(x)}{\rho}}}.\\
\inf_{h\in\sH_{\mathrm{lin}}}\sC_{\Phi_{\rho}}(h,x,t)
&=\min\curl*{t,1-t}+\max\curl*{t,1-t}\paren*{1-\frac{\min\curl*{W\norm*{x}_p+B,\rho}}{\rho}}.
\end{aligned}
\end{equation*}
Therefore, the $\paren*{\Phi_{\rho},\sH_{\mathrm{lin}}}$-minimizability gap can be expressed as follows:
\begin{equation}
\label{eq:M-rho-lin}
\begin{aligned}
& \sM_{\Phi_{\rho}}\paren*{\sH_{\mathrm{lin}}}\\
& = \sR_{\Phi_{\rho}}^*\paren*{\sH_{\mathrm{lin}}}-\mathbb{E}_{X}\bracket*{\inf_{h\in\sH_{\mathrm{lin}}}\sC_{\Phi_{\rho}}(h,x,\eta(x))}\\
& = \sR_{\Phi_{\rho}}^*\paren*{\sH_{\mathrm{lin}}}-\mathbb{E}_{X}\bracket*{\min\curl*{\eta(x),1-\eta(x)}+\max\curl*{\eta(x),1-\eta(x)}\paren*{1-\frac{\min\curl*{W\norm*{x}_p+B,\rho}}{\rho}}}.
 \end{aligned}
\end{equation}
Note the $\paren*{\Phi_{\rho},\sH_{\mathrm{lin}}}$-minimizability gap coincides with
the $\paren*{\Phi_{\rho},\sH_{\mathrm{lin}}}$-approximation error:\\ $\sR_{\Phi_{\rho}}^*\paren*{\sH_{\mathrm{lin}}}-\mathbb{E}_{X}\bracket*{\min\curl*{\eta(x),1-\eta(x)}}$ for $B \geq \rho$.

For $\frac{1}2< t\leq1$, we have
\begin{align*}
\inf_{h\in\sH_{\mathrm{lin}}:h(x)<0}\sC_{\Phi_{\rho}}(h,x,t)
&=t+(1-t)\paren*{1-\frac{\min\curl*{W\norm*{x}_p+B,\rho}}{
\rho}}.\\
\inf_{x\in \sX}\inf_{h\in\sH_{\mathrm{lin}}:h(x)<0}\Delta\sC_{\Phi_{\rho},\sH_{\mathrm{lin}}}(h,x)
&=\inf_{x\in \sX}\paren*{\inf_{h\in\sH_{\mathrm{lin}}:h(x)<0}\sC_{\Phi_{\rho}}(h,x,t)-\inf_{h\in\sH_{\mathrm{lin}}}\sC_{\Phi_{\rho}}(h,x,t)}\\
&=\inf_{x\in \sX}(2t-1)\frac{\min\curl*{W\norm*{x}_p+B,\rho}}{\rho}\\
&=(2t-1)\frac{\min\curl*{B,\rho}}{\rho}\\
&=\sT(2t-1)
\end{align*}
where $\sT$ is the increasing and convex function on $[0,1]$ defined by
\begin{align*}
\forall t\in [0,1],\; \sT(t)=\frac{\min\curl*{B,\rho}}{\rho} \, t.    
\end{align*} 
By Definition~\ref{def:trans}, for any $\epsilon\geq 0$, the $\sH_{\mathrm{lin}}$-estimation error transformation of the $\rho$-margin loss is as follows:
\begin{align*}
\sT_{\Phi_{\rho}}= \frac{\min\curl*{B,\rho}}{\rho} \, t, \quad t \in [0,1],
\end{align*}
Therefore, $\sT_{\Phi_{\rho}}$ is convex, non-decreasing, invertible and satisfies that $\sT_{\Phi_{\rho}}(0)=0$. By Theorem~\ref{Thm:tightness}, we can choose $\Psi(t)=\frac{\min\curl*{B,\rho}}{\rho} \, t$ in Theorem~\ref{Thm:excess_bounds_Psi_uniform}, or equivalently $\Gamma(t)=\frac{\rho }{\min\curl*{B,\rho}} \, t$ in Theorem~\ref{Thm:excess_bounds_Gamma_uniform}, which are optimal when $\e=0$.
Thus, by Theorem~\ref{Thm:excess_bounds_Psi_uniform} or Theorem~\ref{Thm:excess_bounds_Gamma_uniform}, setting $\e = 0$ yields the $\sH_{\mathrm{lin}}$-consistency bound for the $\rho$-margin loss, valid for all $h \in \sH_{\mathrm{lin}}$:
\begin{align}
\label{eq:rho-lin-est}
     \sR_{\ell_{0-1}}(h)- \sR_{\ell_{0-1}}^*\paren*{\sH_{\mathrm{lin}}}\leq \frac{\rho\paren*{\sR_{\Phi_{\rho}}(h)- \sR_{\Phi_{\rho}}^*\paren*{\sH_{\mathrm{lin}}}+\sM_{\Phi_{\rho}}\paren*{\sH_{\mathrm{lin}}}}}{\min\curl*{B,\rho}}-\sM_{\ell_{0-1}}\paren*{\sH_{\mathrm{lin}}}.
\end{align}
Since the $\paren*{\ell_{0-1},\sH_{\mathrm{lin}}}$-minimizability gap coincides with
the $\paren*{\ell_{0-1},\sH_{\mathrm{lin}}}$-approximation error and
$\paren*{\Phi_{\rho},\sH_{\mathrm{lin}}}$-minimizability gap coincides with
the $\paren*{\Phi_{\rho},\sH_{\mathrm{lin}}}$-approximation error for $B \geq \rho$,
the inequality can be rewritten as follows:
\begin{multline*}
     \sR_{\ell_{0-1}}(h)- \sR_{\ell_{0-1}}^*\paren*{\sH_{\mathrm{all}}}\\
     \leq 
     \begin{cases}
     \sR_{\Phi_{\rho}}(h) - \sR_{\Phi_{\rho}}^*\paren*{\sH_{\mathrm{all}}} & \text{if } B \geq \rho\\
     \frac{\rho \paren*{\sR_{\Phi_{\rho}}(h)
     - \mathbb{E}_{X}\bracket*{\min\curl*{\eta(x),1-\eta(x)}+\max\curl*{\eta(x),1-\eta(x)}\paren*{1-\frac{\min\curl*{W\norm*{x}_p+B,\rho}}{\rho}} } }}{B}  & \text{otherwise}. 
     \end{cases}
\end{multline*}
Note that: $\sR_{\Phi_{\rho}}^*\paren*{\sH_{\mathrm{all}}} =  \sR_{\ell_{0-1}}^*\paren*{\sH_{\mathrm{all}}} =\mathbb{E}_X\bracket*{\min\curl*{\eta(x), 1 - \eta(x)}}$. Thus, the first
inequality (case $B \geq \rho$) can be equivalently written as follows:
\begin{align}
\label{eq:rho-lin-est-2}
    \forall h \in \sH_{\mathrm{lin}}, \quad \sR_{\ell_{0-1}}(h) 
     \leq \sR_{\Phi_{\rho}}(h).
\end{align}
The case $B \geq \rho$ is one of the ``trivial cases'' mentioned in Section~\ref{sec:general}, where the trivial inequality $\sR_{\ell_{0-1}}(h) \leq \sR_{\Phi_{\rho}}(h)$ can be obtained directly using the fact that $\ell_{0-1}$ is upper-bounded by $\Phi_{\rho}$. This, however, does not imply that non-adversarial $\sH_{\mathrm{lin}}$-consistency bound for the $\rho$-margin loss is trivial when $B>\rho$ since it is optimal.

\subsection{One-Hidden-Layer ReLU Neural Network}
\label{app:derivation-NN}
As with the linear case, $\sH_{\mathrm{NN}}$ also satisfies the condition of Lemma~\ref{lemma:explicit_assumption_01} and thus $\paren*{\ell_{0-1},\sH_{\mathrm{NN}}}$-minimizability gap coincides with
the $\paren*{\ell_{0-1},\sH_{\mathrm{NN}}}$-approximation error:
\begin{equation}
\begin{aligned}
\label{eq:M-01-NN}
\sM_{\ell_{0-1}}\paren*{\sH_{\mathrm{NN}}}
& = \sR_{\ell_{0-1}}^*\paren*{\sH_{\mathrm{NN}}}-\mathbb{E}_{X}\bracket*{\min\curl*{\eta(x),1-\eta(x)}} \\
&= \sR_{\ell_{0-1}}^*\paren*{\sH_{\mathrm{NN}}} - \sR_{\ell_{0-1}}^*\paren*{\sH_{\mathrm{all}}}.
\end{aligned}
\end{equation}
By the definition of $\sH_{\mathrm{NN}}$, for any $x \in \sX$, 
\begin{align*}
\curl*{h(x)\mid h\in\sH_{\mathrm{NN}}}= \bracket*{-\Lambda\paren*{W\norm*{x}_p+B}, \Lambda\paren*{W\norm*{x}_p+B}}.
\end{align*}

\subsubsection{Hinge Loss}
For the hinge loss $\Phi_{\mathrm{hinge}}(\alpha)\colon=\max\curl*{0,1 - \alpha}$, for all $h\in \sH_{\mathrm{NN}}$ and $x\in \sX$:
\begin{equation*}
\begin{aligned}
\sC_{\Phi_{\mathrm{hinge}}}(h,x,t)
& =t \Phi_{\mathrm{hinge}}(h(x))+(1-t)\Phi_{\mathrm{hinge}}(-h(x))\\
& =t\max\curl*{0,1-h(x)}+(1-t)\max\curl*{0,1+h(x)}.\\
\inf_{h\in\sH_{\mathrm{NN}}}\sC_{\Phi_{\mathrm{hinge}}}(h,x,t)
& = 1-\abs*{2t-1}\min\curl*{\Lambda W\norm*{x}_p+\Lambda B,1}.
\end{aligned}
\end{equation*}
Therefore, the $\paren*{\Phi_{\mathrm{hinge}},\sH_{\mathrm{NN}}}$-minimizability gap can be expressed as follows:
\begin{equation}
\begin{aligned}
\label{eq:M-hinge-NN}
\sM_{\Phi_{\mathrm{hinge}}}\paren*{\sH_{\mathrm{NN}}}
& = \sR_{\Phi_{\mathrm{hinge}}}^*\paren*{\sH_{\mathrm{NN}}}-\mathbb{E}_{X}\bracket*{1-\inf_{h\in\sH_{\mathrm{NN}}}\sC_{\Phi_{\mathrm{hinge}}}(h,x,\eta(x))}.\\
& = \sR_{\Phi_{\mathrm{hinge}}}^*\paren*{\sH_{\mathrm{NN}}}-\mathbb{E}_{X}\bracket*{1-\abs*{2\eta(x)-1}\min\curl*{\Lambda W\norm*{x}_p+\Lambda B,1}}.
\end{aligned}
\end{equation}
Note the $\paren*{\Phi_{\mathrm{hinge}},\sH_{\mathrm{NN}}}$-minimizability gap coincides with
the $\paren*{\Phi_{\mathrm{hinge}},\sH_{\mathrm{NN}}}$-approximation error $\sR_{\Phi_{\mathrm{hinge}}}^*\paren*{\sH_{\mathrm{NN}}}-\mathbb{E}_{X}\bracket*{1-\abs*{2\eta(x)-1}}$ for $\Lambda B \geq 1$.

For $\frac{1}2< t\leq1$, we have
\begin{align*}
\inf_{h\in\sH_{\mathrm{NN}}:h(x)<0}\sC_{\Phi_{\mathrm{hinge}}}(h,x,t)
& = t\max\curl*{0,1-0}+(1-t)\max\curl*{0,1+0}\\
& =1.\\
\inf_{x\in \sX} \inf_{h\in\sH_{\mathrm{NN}}:h(x)<0} \Delta\sC_{\Phi_{\mathrm{hinge}},\sH_{\mathrm{NN}}}(h,x,t)
& = \inf_{x\in \sX} \curl*{\inf_{h\in\sH_{\mathrm{NN}}:h(x)<0}\sC_{\Phi_{\mathrm{hinge}}}(h,x,t)-\inf_{h\in\sH_{\mathrm{NN}}}\sC_{\Phi_{\mathrm{hinge}}}(h,x,t)}\\
&=\inf_{x\in \sX}\paren*{2t-1}\min\curl*{\Lambda W\norm*{x}_p+\Lambda B,1}\\
&=(2t-1)\min\curl*{\Lambda B,1}\\
&=\sT(2t - 1),
\end{align*}
where $\sT$ is the increasing and convex function on $[0,1]$ defined by
\begin{align*}
\forall t \in [0,1],\; \sT(t) = \min \curl*{\Lambda B, 1} \, t .
\end{align*}
By Definition~\ref{def:trans}, for any $\epsilon\geq 0$, the $\sH_{\mathrm{NN}}$-estimation error transformation of the hinge loss is as follows:
\begin{align*}
\sT_{\Phi_{\mathrm{hinge}}}= \min \curl*{\Lambda B, 1} \, t, \quad t \in [0,1],
\end{align*}
Therefore, $\sT_{\Phi_{\mathrm{hinge}}}$ is convex, non-decreasing, invertible and satisfies that $\sT_{\Phi_{\mathrm{hinge}}}(0)=0$. By Theorem~\ref{Thm:tightness}, we can choose $\Psi(t)=\min\curl*{\Lambda B,1} \, t$ in Theorem~\ref{Thm:excess_bounds_Psi_uniform}, or, equivalently, $\Gamma(t) = \frac{t}{\min\curl*{\Lambda B, 1}}$ in Theorem~\ref{Thm:excess_bounds_Gamma_uniform}, which are optimal when $\e=0$. Thus, by Theorem~\ref{Thm:excess_bounds_Psi_uniform} or Theorem~\ref{Thm:excess_bounds_Gamma_uniform}, setting $\e = 0$ yields the $\sH_{\mathrm{NN}}$-consistency bound for the hinge loss, valid for all $h \in \sH_{\mathrm{NN}}$:
\begin{align}
\label{eq:hinge-NN-est}
     \sR_{\ell_{0-1}}(h)- \sR_{\ell_{0-1}}^*\paren*{\sH_{\mathrm{NN}}}
     \leq \frac{\sR_{\Phi_{\mathrm{hinge}}}(h)- \sR_{\Phi_{\mathrm{hinge}}}^*\paren*{\sH_{\mathrm{NN}}}+\sM_{\Phi_{\mathrm{hinge}}}\paren*{\sH_{\mathrm{NN}}}}{\min\curl*{\Lambda B,1}}-\sM_{\ell_{0-1}, \sH_{\mathrm{NN}}}.
\end{align}
Since the $\paren*{\ell_{0-1},\sH_{\mathrm{NN}}}$-minimizability gap coincides with
the $\paren*{\ell_{0-1},\sH_{\mathrm{NN}}}$-approximation error and
$\paren*{\Phi_{\mathrm{hinge}},\sH_{\mathrm{NN}}}$-minimizability gap coincides with
the $\paren*{\Phi_{\mathrm{hinge}},\sH_{\mathrm{NN}}}$-approximation error for $\Lambda B \geq 1$,
the inequality can be rewritten as follows:
\begin{multline*}
     \sR_{\ell_{0-1}}(h)- \sR_{\ell_{0-1}}^*\paren*{\sH_{\mathrm{all}}}\\
      \leq 
     \begin{cases}
     \sR_{\Phi_{\mathrm{hinge}}}(h) - \sR_{\Phi_{\mathrm{hinge}}}^*\paren*{\sH_{\mathrm{all}}} & \text{if } \Lambda B \geq 1\\
     \frac{1}{\Lambda B} \bracket[\Big]{\sR_{\Phi_{\mathrm{hinge}}}(h)
     - \mathbb{E}_{X}\bracket*{1-\abs*{2\eta(x)-1}\min\curl*{\Lambda W\norm*{x}_p+\Lambda B,1}\
     } } & \text{otherwise}.
     \end{cases}
\end{multline*}*
The inequality for $\Lambda B \geq 1$ coincides with the consistency excess error bound
known for the hinge loss \citep{Zhang2003,bartlett2006convexity,MohriRostamizadehTalwalkar2018} but the one for $\Lambda B < 1$ is distinct and novel.  For $\Lambda B < 1$, we have
\begin{align*}
& \mathbb{E}_{X}\bracket*{1-\abs*{2\eta(x)-1}\min\curl*{\Lambda W\norm*{x}_p+\Lambda B,1}}\\
&\qquad > \mathbb{E}_{X}\bracket*{1-\abs*{2\eta(x)-1}}= 2\mathbb{E}_X\bracket*{\min\curl*{\eta(x), 1 - \eta(x)}}= \sR_{\Phi_{\mathrm{hinge}}}^*\paren*{\sH_{\mathrm{all}}}.
\end{align*}
Therefore for $\Lambda B < 1$, 
\begin{align*}
\sR_{\Phi_{\mathrm{hinge}}}(h) - \mathbb{E}_{X}\bracket*{1-\abs*{2\eta(x)-1}\min\curl*{\Lambda W\norm*{x}_p+\Lambda B,1}} < \sR_{\Phi_{\mathrm{hinge}}}(h) - \sR_{\Phi_{\mathrm{hinge}}}^*\paren*{\sH_{\mathrm{all}}}.
\end{align*}
Note that: $\sR_{\Phi_{\mathrm{hinge}}}^*\paren*{\sH_{\mathrm{all}}} = 2 \sR_{\ell_{0-1}}^*\paren*{\sH_{\mathrm{all}}} =2\mathbb{E}_X\bracket*{\min\curl*{\eta(x), 1 - \eta(x)}}$. Thus, the first
inequality (case $\Lambda B \geq 1$) can be equivalently written as follows:
\begin{align*}
    \forall h \in \sH_{\mathrm{NN}}, \quad \sR_{\ell_{0-1}}(h) 
     \leq \sR_{\Phi_{\mathrm{hinge}}}(h) - \mathbb{E}_X\bracket*{\min\curl*{\eta(x), 1 - \eta(x)}},
\end{align*}
which is a more informative upper bound than the standard
inequality $\sR_{\ell_{0-1}}(h) 
     \leq \sR_{\Phi_{\mathrm{hinge}}}(h)$.

\subsubsection{Logistic Loss}

For the logistic loss $\Phi_{\mathrm{log}}(\alpha)\colon=\log_2(1+e^{-\alpha})$, for all $h\in \sH_{\mathrm{NN}}$ and $x\in \sX$:
\begin{equation*}
\begin{aligned}
&\sC_{\Phi_{\mathrm{log}}}(h,x,t)\\
& = t \Phi_{\mathrm{log}}(h(x))+(1-t)\Phi_{\mathrm{log}}(-h(x)),\\
& = t\log_2\paren*{1+e^{-h(x)}}+(1-t)\log_2\paren*{1+e^{h(x)}}.\\
& \inf_{h\in\sH_{\mathrm{NN}}}\sC_{\Phi_{\mathrm{log}}}(h,x,t)\\
&=\begin{cases}
-t\log_2(t)-(1-t)\log_2(1-t) \\ \qquad \text{if }\log\abs*{\frac{t}{1-t}}\leq \Lambda W\norm*{x}_p+ \Lambda B,\\
\max\curl*{t,1-t}\log_2\paren*{1+e^{-(\Lambda W\norm*{x}_p+\Lambda B)}}+\min\curl*{t,1-t}\log_2\paren*{1+e^{\Lambda W\norm*{x}_p+\Lambda B}} \\ \qquad \text{if }\log\abs*{\frac{t}{1-t}}> \Lambda W\norm*{x}_p+\Lambda B.
\end{cases}
\end{aligned}
\end{equation*}
Therefore, the $\paren*{\Phi_{\mathrm{log}},\sH_{\mathrm{NN}}}$-minimizability gap can be expressed as follows:
\begin{equation}
\begin{aligned}
\label{eq:M-log-NN}
& \sM_{\Phi_{\mathrm{log}}}\paren*{\sH_{\mathrm{NN}}}\\
& = \sR_{\Phi_{\mathrm{log}}}^*\paren*{\sH_{\mathrm{NN}}}-
\mathbb{E}_{X}\bracket*{\inf_{h\in\sH_{\mathrm{NN}}}\sC_{\Phi_{\mathrm{log}}}(h,x,\eta(x))}\\
& = \sR_{\Phi_{\mathrm{log}}}^*\paren*{\sH_{\mathrm{NN}}}-
\mathbb{E}_{X}\bracket*{-\eta(x)\log_2(\eta(x))-(1-\eta(x))\log_2(1-\eta(x))\mathds{1}_{\log\abs*{\frac{\eta(x)}{1-\eta(x)}}\leq \Lambda W\norm*{x}_p+\Lambda B}}\\
& - \mathbb{E}_{X}\bracket*{\max\curl*{\eta(x),1-\eta(x)}\log_2\paren*{1+e^{-(\Lambda W\norm*{x}_p+\Lambda B)}}\mathds{1}_{\log\abs*{\frac{\eta(x)}{1-\eta(x)}}> \Lambda W\norm*{x}_p+\Lambda B}}\\
&-\mathbb{E}_{X}\bracket*{\min\curl*{\eta(x),1-\eta(x)}\log_2\paren*{1+e^{\Lambda W\norm*{x}_p+\Lambda B}}\mathds{1}_{\log\abs*{\frac{\eta(x)}{1-\eta(x)}}> \Lambda W\norm*{x}_p+\Lambda B}}
\end{aligned}
\end{equation}
Note $\paren*{\Phi_{\mathrm{log}},\sH_{\mathrm{NN}}}$-minimizability gap coincides with
the $\paren*{\Phi_{\mathrm{log}},\sH_{\mathrm{NN}}}$-approximation error:\\
$\sR_{\Phi_{\mathrm{log}}}^*\paren*{\sH_{\mathrm{NN}}}-
\mathbb{E}_{X}\bracket*{-\eta(x)\log_2(\eta(x))-(1-\eta(x))\log_2(1-\eta(x))}$ for $\Lambda B =+\infty$.

For $\frac{1}2< t\leq1$, we have
\begin{align*}
&\inf_{h\in\sH_{\mathrm{NN}}:h(x)<0}\sC_{\Phi_{\mathrm{log}}}(h,x,t)\\
& = t\log_2\paren*{1+e^{-0}}+(1-t)\log_2\paren*{1+e^{0}} \\
& = 1, \\
&\inf_{x\in \sX}\inf_{h\in\sH_{\mathrm{NN}}:h(x)<0}\Delta\sC_{\Phi_{\mathrm{log}},\sH_{\mathrm{NN}}}(h,x,t)\\
&=\inf_{x\in \sX}\paren*{\inf_{h\in\sH_{\mathrm{NN}}:h(x)<0}\sC_{\Phi_{\mathrm{log}}}(h,x,t)-\inf_{h\in\sH_{\mathrm{NN}}}\sC_{\Phi_{\mathrm{log}}}(h,x,t)}\\
&=\inf_{x\in \sX}\begin{cases}
1+t\log_2(t)+(1-t)\log_2(1-t)\\
\text{if }\log\frac{t}{1-t}\leq \Lambda W\norm*{x}_p+\Lambda B,\\
1-t\log_2\paren*{1+e^{-(\Lambda W\norm*{x}_p+\Lambda B)}}-(1-t)\log_2\paren*{1+e^{\Lambda W\norm*{x}_p+\Lambda B}}\\
\text{if }\log\frac{t}{1-t}> \Lambda W\norm*{x}_p+\Lambda B.
\end{cases}\\
&=\begin{cases}
1+t\log_2(t)+(1-t)\log_2(1-t) & \text{if }\log\frac{t}{1-t}\leq \Lambda B,\\
1-t\log_2\paren*{1+e^{-\Lambda B}}-(1-t)\log_2\paren*{1+e^{\Lambda B}} & \text{if }\log\frac{t}{1-t}> \Lambda B.
\end{cases}\\
&=\sT(2t-1),
\end{align*}
where $\sT$ is the increasing and convex function on $[0,1]$ defined by
\begin{align*}
\forall t\in[0,1], \quad
\sT(t)=\begin{cases}
\frac{t+1}{2}\log_2(t+1)+\frac{1-t}{2}\log_2(1-t),\quad &  t\leq \frac{e^{\Lambda B}-1}{e^{\Lambda B}+1},\\
1-\frac{t+1}{2}\log_2(1+e^{-\Lambda B})-\frac{1-t}{2}\log_2(1+e^{\Lambda B}),\quad & t> \frac{e^{\Lambda B}-1}{e^{\Lambda B}+1}.
\end{cases}
\end{align*}
By Definition~\ref{def:trans}, for any $\epsilon\geq 0$, the $\sH_{\mathrm{NN}}$-estimation error transformation of the logistic loss is as follows:
\begin{align*}
\sT_{\Phi_{\mathrm{log}}}= 
\begin{cases}
\sT(t), & t\in \left[\epsilon,1\right], \\
\frac{\sT(\epsilon)}{\epsilon}\, t, &  t\in \left[0,\epsilon\right).
\end{cases}
\end{align*}
Therefore, when $\epsilon=0$, $\sT_{\Phi_{\mathrm{log}}}$ is convex, non-decreasing, invertible and satisfies that $\sT_{\Phi_{\mathrm{log}}}(0)=0$. By Theorem~\ref{Thm:tightness}, we can choose $\Psi(t)=\sT_{\Phi_{\mathrm{log}}}(t)$ in Theorem~\ref{Thm:excess_bounds_Psi_uniform}, or equivalently $\Gamma(t)=\sT_{\Phi_{\mathrm{log}}}^{-1}(t)$ in Theorem~\ref{Thm:excess_bounds_Gamma_uniform}, which are optimal. To simplify the expression, using the fact that
\begin{align*}
\frac{t+1}{2}\log_2(t+1)+\frac{1-t}{2}\log_2(1-t)
&=
1-\paren*{-\frac{t+1}{2}\log_2\paren*{\frac{t+1}{2}}-\frac{1-t}{2}\log_2\paren*{\frac{1-t}{2}}}\\
&\geq 1 - \sqrt{4\frac{1-t}{2} \frac{t+1}{2}}\\
& = 1- \sqrt{1-t^2}\\
& \geq \frac{t^2}{2},\\
1-\frac{t+1}{2}\log_2(1+e^{-\Lambda B})-\frac{1-t}{2}\log_2(1+e^{\Lambda B})
&=\frac{1}{2}\log_2\paren*{\frac{4}{2+e^{-\Lambda B}+e^{\Lambda B}}}+1/2\log_2\paren*{\frac{1+e^{\Lambda B}}{1+e^{-\Lambda B}}}\, t,
\end{align*}
$\sT_{\Phi_{\mathrm{log}}}$ can be lower bounded by
\begin{align*}
\wt{\sT}_{\Phi_{\mathrm{log}}}(t)= \begin{cases}
\frac{t^2}{2},& t\leq \frac{e^{\Lambda B}-1}{e^{\Lambda B}+1},\\
\frac{1}{2}\paren*{\frac{e^{\Lambda B}-1}{e^{\Lambda B}+1}}\, t, & t> \frac{e^{\Lambda B}-1}{e^{\Lambda B}+1}.
\end{cases}   
\end{align*}
Thus, we adopt an upper bound of $\sT_{\Phi_{\mathrm{log}}}^{-1}$ as follows:
\begin{align*}
\wt{\sT}_{\Phi_{\mathrm{log}}}^{-1}(t)=\begin{cases}
\sqrt{2t}, & t\leq \frac{1}{2}\paren*{\frac{e^{\Lambda B}-1}{e^{\Lambda B}+1}}^2,\\
2\paren*{\frac{e^{\Lambda B}+1}{e^{\Lambda B}-1}}\, t, & t> \frac{1}{2}\paren*{\frac{e^{\Lambda B}-1}{e^{\Lambda B}+1}}^2.
\end{cases}
\end{align*}
Therefore, by Theorem~\ref{Thm:excess_bounds_Psi_uniform} or Theorem~\ref{Thm:excess_bounds_Gamma_uniform}, setting $\e = 0$ yields the $\sH_{\mathrm{NN}}$-consistency bound for the logistic loss, valid for all $h \in \sH_{\mathrm{NN}}$:
\begin{multline}
\label{eq:log-NN-est}
     \sR_{\ell_{0-1}}(h)-\sR_{\ell_{0-1}}^*\paren*{\sH_{\mathrm{NN}}}+\sM_{\ell_{0-1}, \sH_{\mathrm{NN}}}\\
     \leq 
     \begin{cases}
     \sqrt{2}\,\paren*{\sR_{\Phi_{\mathrm{log}}}(h)- \sR_{\Phi_{\mathrm{log}}}^*\paren*{\sH_{\mathrm{NN}}}+\sM_{\Phi_{\mathrm{log}}}\paren*{\sH_{\mathrm{NN}}}}^{\frac12}, \\ \qquad \text{if } \sR_{\Phi_{\mathrm{log}}}(h)- \sR_{\Phi_{\mathrm{log}}}^*\paren*{\sH_{\mathrm{NN}}}\leq \frac{1}{2}\paren*{\frac{e^{\Lambda B}-1}{e^{\Lambda B}+1}}^2-\sM_{\Phi_{\mathrm{log}}}\paren*{\sH_{\mathrm{NN}}}\\
     2\paren*{\frac{e^{\Lambda B}+1}{e^{\Lambda B}-1}}\paren*{\sR_{\Phi_{\mathrm{log}}}(h)- \sR_{\Phi_{\mathrm{log}}}^*\paren*{\sH_{\mathrm{NN}}}+\sM_{\Phi_{\mathrm{log}}}\paren*{\sH_{\mathrm{NN}}}}, \\ \qquad \text{otherwise}
     \end{cases}
\end{multline}
Since the $\paren*{\ell_{0-1},\sH_{\mathrm{NN}}}$-minimizability gap coincides with
the $\paren*{\ell_{0-1},\sH_{\mathrm{NN}}}$-approximation error and
$\paren*{\Phi_{\mathrm{log}},\sH_{\mathrm{NN}}}$-minimizability gap coincides with
the $\paren*{\Phi_{\mathrm{log}},\sH_{\mathrm{NN}}}$-approximation error for $\Lambda B =+\infty$,
the inequality can be rewritten as follows:
\begin{align*}
     &\sR_{\ell_{0-1}}(h)- \sR_{\ell_{0-1}}^*\paren*{\sH_{\mathrm{all}}}\\
     &\quad \leq 
     \begin{cases}
      \sqrt{2}\,\bracket*{\sR_{\Phi_{\mathrm{log}}}(h) - \sR_{\Phi_{\mathrm{log}}}^*\paren*{\sH_{\mathrm{all}}}}^{\frac12} & \text{if } \Lambda B = +\infty \\
     \begin{cases}
    \sqrt{2}\,\bracket*{\sR_{\Phi_{\mathrm{log}}}(h)- \sR_{\Phi_{\mathrm{log}}}^*\paren*{\sH_{\mathrm{NN}}}+\sM_{\Phi_{\mathrm{log}}}\paren*{\sH_{\mathrm{NN}}}}^{\frac12}  \\ \qquad \text{if } \sR_{\Phi_{\mathrm{log}}}(h)- \sR_{\Phi_{\mathrm{log}}}^*\paren*{\sH_{\mathrm{NN}}}\leq \frac{1}{2}\paren*{\frac{e^{\Lambda B}-1}{e^{\Lambda B}+1}}^2-\sM_{\Phi_{\mathrm{log}}}\paren*{\sH_{\mathrm{NN}}} \\
    2\paren*{\frac{e^{\Lambda B}+1}{e^{\Lambda B}-1}}\paren*{\sR_{\Phi_{\mathrm{log}}}(h)- \sR_{\Phi_{\mathrm{log}}}^*\paren*{\sH_{\mathrm{NN}}}+\sM_{\Phi_{\mathrm{log}}}\paren*{\sH_{\mathrm{NN}}}} \\ \qquad \text{otherwise}
    \end{cases} & \text{otherwise}
     \end{cases}
\end{align*}
where the $\paren*{\Phi_{\mathrm{log}},\sH_{\mathrm{NN}}}$-minimizability gap $\sM_{\Phi_{\mathrm{log}}}\paren*{\sH_{\mathrm{NN}}}$ is characterized as below,which is less than
the $\paren*{\Phi_{\mathrm{log}},\sH_{\mathrm{NN}}}$-approximation error when $\Lambda B<+ \infty$:
\begin{align*}
& \sM_{\Phi_{\mathrm{log}}}\paren*{\sH_{\mathrm{NN}}}\\
& = \sR_{\Phi_{\mathrm{log}}}^*\paren*{\sH_{\mathrm{NN}}}-
\mathbb{E}_{X}\bracket*{-\eta(x)\log_2(\eta(x))-(1-\eta(x))\log_2(1-\eta(x))\mathds{1}_{\log_2\abs*{\frac{\eta(x)}{1-\eta(x)}}\leq \Lambda W\norm*{x}_p+\Lambda B}}\\
& - \mathbb{E}_{X}\bracket*{\max\curl*{\eta(x),1-\eta(x)}\log_2\paren*{1+e^{-(\Lambda W\norm*{x}_p+\Lambda B)}}\mathds{1}_{\log_2\abs*{\frac{\eta(x)}{1-\eta(x)}}> \Lambda W\norm*{x}_p+\Lambda B}}\\
&-\bracket*{\min\curl*{\eta(x),1-\eta(x)}\log_2\paren*{1+e^{\Lambda W\norm*{x}_p+\Lambda B}}\mathds{1}_{\log_2\abs*{\frac{\eta(x)}{1-\eta(x)}}> \Lambda W\norm*{x}_p+\Lambda B}}\\
& < \sR_{\Phi_{\mathrm{log}}}^*\paren*{\sH_{\mathrm{NN}}} - \mathbb{E}_{X}\bracket*{-\eta(x)\log_2(\eta(x))-(1-\eta(x))\log_2(1-\eta(x))}\\
& = \sR_{\Phi_{\mathrm{log}}}^*\paren*{\sH_{\mathrm{NN}}} - \sR_{\Phi_{\mathrm{log}}}^*\paren*{\sH_{\mathrm{all}}} .
\end{align*}
Therefore, the inequality for $\Lambda B = + \infty$ coincides with the consistency excess error bound
known for the logistic loss \citep{Zhang2003,MohriRostamizadehTalwalkar2018} but the one for $\Lambda B< + \infty$ is distinct and novel.

\subsubsection{Exponential Loss}
For the exponential loss $\Phi_{\mathrm{exp}}(\alpha)\colon=e^{-\alpha}$, for all $h\in \sH_{\mathrm{NN}}$ and $x\in \sX$:
\begin{equation*}
\begin{aligned}
\sC_{\Phi_{\mathrm{exp}}}(h,x,t)
&=t \Phi_{\mathrm{exp}}(h(x))+(1-t)\Phi_{\mathrm{exp}}(-h(x))\\
&=te^{-h(x)}+(1-t)e^{h(x)}.\\
\inf_{h\in\sH_{\mathrm{NN}}}\sC_{\Phi_{\mathrm{exp}}}(h,x,t)
&=\begin{cases}
2\sqrt{t(1-t)} \\ \qquad \text{if }1/2\log\abs*{\frac{t}{1-t}}\leq \Lambda W\norm*{x}_p+\Lambda B\\
\max\curl*{t,1-t}e^{-(\Lambda W\norm*{x}_p+ \Lambda B)}+\min\curl*{t,1-t}e^{\Lambda W\norm*{x}_p+\Lambda B} \\ \qquad \text{if }1/2\log\abs*{\frac{t}{1-t}}> \Lambda W\norm*{x}_p+ \Lambda B.
\end{cases}
\end{aligned}
\end{equation*}
Therefore, the $\paren*{\Phi_{\mathrm{exp}},\sH_{\mathrm{NN}}}$-minimizability gap can be expressed as follows:
\begin{equation}
\begin{aligned}
\label{eq:M-exp-NN}
\sM_{\Phi_{\mathrm{exp}}}\paren*{\sH_{\mathrm{NN}}}
& = \sR_{\Phi_{\mathrm{exp}}}^*\paren*{\sH_{\mathrm{NN}}}-
\mathbb{E}_{X}\bracket*{\inf_{h\in\sH_{\mathrm{NN}}}\sC_{\Phi_{\mathrm{exp}}}(h,x,\eta(x))}\\
& = \sR_{\Phi_{\mathrm{exp}}}^*\paren*{\sH_{\mathrm{NN}}}-
\mathbb{E}_{X}\bracket*{2\sqrt{\eta(x)(1-\eta(x))}\mathds{1}_{1/2\log\abs*{\frac{\eta(x)}{1-\eta(x)}}\leq \Lambda W\norm*{x}_p+\Lambda B}}\\
& - \mathbb{E}_{X}\bracket*{\max\curl*{\eta(x),1-\eta(x)}e^{-(\Lambda W\norm*{x}_p+\Lambda B)}\mathds{1}_{1/2\log\abs*{\frac{\eta(x)}{1-\eta(x)}}> \Lambda W\norm*{x}_p+\Lambda B}}\\
&-\mathbb{E}_{X}\bracket*{\min\curl*{\eta(x),1-\eta(x)}e^{\Lambda W\norm*{x}_p+\Lambda B}\mathds{1}_{1/2\log\abs*{\frac{\eta(x)}{1-\eta(x)}}>\Lambda W\norm*{x}_p+\Lambda B}}.
\end{aligned}
\end{equation}
Note $\paren*{\Phi_{\mathrm{exp}},\sH_{\mathrm{NN}}}$-minimizability gap coincides with
the $\paren*{\Phi_{\mathrm{exp}},\sH_{\mathrm{NN}}}$-approximation error:\\
$\sR_{\Phi_{\mathrm{exp}}}^*\paren*{\sH_{\mathrm{NN}}}-
\mathbb{E}_{X}\bracket*{2\sqrt{\eta(x)(1-\eta(x))}}$ for $\Lambda B =+\infty$.

For $\frac{1}2< t\leq1$, we have
\begin{align*}
&\inf_{h\in\sH_{\mathrm{NN}}:h(x)<0}\sC_{\Phi_{\mathrm{exp}}}(h,x,t)\\
&=te^{-0}+(1-t)e^{0}\\
&=1.\\
&\inf_{x\in \sX}\inf_{h\in\sH_{\mathrm{NN}}:h(x)<0}\Delta\sC_{\Phi_{\mathrm{exp}},\sH_{\mathrm{NN}}}(h,x,t)\\
&=\inf_{x\in \sX}\paren*{\inf_{h\in\sH_{\mathrm{NN}}:h(x)<0}\sC_{\Phi_{\mathrm{exp}}}(h,x,t)-\inf_{h\in\sH_{\mathrm{NN}}}\sC_{\Phi_{\mathrm{exp}}}(h,x,t)}\\
&=\inf_{x\in \sX}\begin{cases}
1-2\sqrt{t(1-t)} & \text{if }1/2\log\frac{t}{1-t}\leq \Lambda W\norm*{x}_p+\Lambda B,\\
1-te^{-(\Lambda W\norm*{x}_p+\Lambda B)}-(1-t)e^{\Lambda W\norm*{x}_p+\Lambda B} & \text{if }1/2\log\frac{t}{1-t}>\Lambda W\norm*{x}_p+\Lambda B.
\end{cases}\\
&=\begin{cases}
1-2\sqrt{t(1-t)}, & 1/2\log\frac{t}{1-t}\leq \Lambda B\\
1-te^{-\Lambda B}-(1-t)e^{\Lambda B}, & 1/2\log\frac{t}{1-t}> \Lambda B
\end{cases}\\
&=\sT(2t-1),
\end{align*}
where $\sT$ is the increasing and convex function on $[0,1]$ defined by
\begin{align*}
\forall t\in[0,1], \quad 
\sT(t)=\begin{cases}
1-\sqrt{1-t^2}, & t\leq \frac{e^{2\Lambda B}-1}{e^{2\Lambda B}+1},\\
1-\frac{t+1}{2}e^{-\Lambda B}-\frac{1-t}{2}e^{\Lambda B}, & t> \frac{e^{2\Lambda B}-1}{e^{2\Lambda B}+1}.
\end{cases}
\end{align*}
By Definition~\ref{def:trans}, for any $\epsilon\geq 0$, the $\sH_{\mathrm{lin}}$-estimation error transformation of the exponential loss is as follows:
\begin{align*}
\sT_{\Phi_{\mathrm{exp}}}= 
\begin{cases}
\sT(t), & t\in \left[\epsilon,1\right], \\
\frac{\sT(\epsilon)}{\epsilon}\, t, &  t\in \left[0,\epsilon\right).
\end{cases}
\end{align*}
Therefore, when $\epsilon=0$, $\sT_{\Phi_{\mathrm{exp}}}$ is convex, non-decreasing, invertible and satisfies that $\sT_{\Phi_{\mathrm{exp}}}(0)=0$. By Theorem~\ref{Thm:tightness}, we can choose $\Psi(t)=\sT_{\Phi_{\mathrm{exp}}}(t)$ in Theorem~\ref{Thm:excess_bounds_Psi_uniform}, or equivalently $\Gamma(t)=\sT_{\Phi_{\mathrm{exp}}}^{-1}(t)$ in Theorem~\ref{Thm:excess_bounds_Gamma_uniform}, which are optimal.  To simplify the expression, using the fact that
\begin{align*}
1- \sqrt{1-t^2} & \geq \frac{t^2}{2}, \\
1-\frac{t+1}{2}e^{-\Lambda B}-\frac{1-t}{2}e^{\Lambda B} & = 1-1/2 e^{\Lambda B}-1/2 e^{-\Lambda B}+\frac{e^{\Lambda B}-e^{-\Lambda B}}2\, t,
\end{align*}
$\sT_{\Phi_{\mathrm{exp}}}$ can be lower bounded by
\begin{align*}
\wt{\sT}_{\Phi_{\mathrm{exp}}}(t)= \begin{cases}
\frac{t^2}{2},& t\leq \frac{e^{2\Lambda B}-1}{e^{2\Lambda B}+1},\\
\frac{1}{2}\paren*{\frac{e^{2\Lambda B}-1}{e^{2\Lambda B}+1}}\, t, & t> \frac{e^{2\Lambda B}-1}{e^{2\Lambda B}+1}.
\end{cases}   
\end{align*}
Thus, we adopt an upper bound of $\sT_{\Phi_{\mathrm{exp}}}^{-1}$ as follows:
\begin{align*}
\wt{\sT}_{\Phi_{\mathrm{exp}}}^{-1}(t)=
\begin{cases}
\sqrt{2t}, & t\leq \frac{1}{2}\paren*{\frac{e^{2\Lambda B}-1}{e^{2B}+1}}^2,\\
2\paren*{\frac{e^{2\Lambda B}+1}{e^{2\Lambda B}-1}}\, t, & t> \frac{1}{2}\paren*{\frac{e^{2\Lambda B}-1}{e^{2\Lambda B}+1}}^2.
\end{cases}
\end{align*}
Therefore, by Theorem~\ref{Thm:excess_bounds_Psi_uniform} or Theorem~\ref{Thm:excess_bounds_Gamma_uniform}, setting $\e = 0$ yields the $\sH_{\mathrm{NN}}$-consistency bound for the exponential loss, valid for all $h \in \sH_{\mathrm{NN}}$:
\begin{multline}
\label{eq:exp-NN-est}
     \sR_{\ell_{0-1}}(h)-\sR_{\ell_{0-1}}^*\paren*{\sH_{\mathrm{NN}}}+\sM_{\ell_{0-1}, \sH_{\mathrm{NN}}}\\
     \leq 
     \begin{cases}
     \sqrt{2}\,\paren*{\sR_{\Phi_{\mathrm{exp}}}(h)- \sR_{\Phi_{\mathrm{exp}}}^*\paren*{\sH_{\mathrm{NN}}}+\sM_{\Phi_{\mathrm{exp}}}\paren*{\sH_{\mathrm{NN}}}}^{\frac12}, \\ \qquad \text{if } \sR_{\Phi_{\mathrm{exp}}}(h)- \sR_{\Phi_{\mathrm{exp}}}^*\paren*{\sH_{\mathrm{NN}}}\leq \frac{1}{2}\paren*{\frac{e^{2\Lambda B}-1}{e^{2\Lambda B}+1}}^2-\sM_{\Phi_{\mathrm{exp}}}\paren*{\sH_{\mathrm{NN}}},\\
     2\paren*{\frac{e^{2\Lambda B}+1}{e^{2\Lambda B}-1}}\paren*{\sR_{\Phi_{\mathrm{exp}}}(h)- \sR_{\Phi_{\mathrm{exp}}}^*\paren*{\sH_{\mathrm{NN}}}+\sM_{\Phi_{\mathrm{exp}}}\paren*{\sH_{\mathrm{NN}}}}, \\ \qquad \text{otherwise}.
     \end{cases}
\end{multline}
Since the $\paren*{\ell_{0-1},\sH_{\mathrm{NN}}}$-minimizability gap coincides with
the $\paren*{\ell_{0-1},\sH_{\mathrm{NN}}}$-approximation error and
$\paren*{\Phi_{\mathrm{log}},\sH_{\mathrm{NN}}}$-minimizability gap coincides with
the $\paren*{\Phi_{\mathrm{log}},\sH_{\mathrm{NN}}}$-approximation error for $\Lambda B =+\infty$,
the inequality can be rewritten as follows:
\begin{multline*}
     \sR_{\ell_{0-1}}(h)- \sR_{\ell_{0-1}}^*\paren*{\sH_{\mathrm{all}}}\leq \\
     \begin{cases}
      \sqrt{2}\,\bracket*{\sR_{\Phi_{\mathrm{exp}}}(h) - \sR_{\Phi_{\mathrm{exp}}}^*\paren*{\sH_{\mathrm{all}}}}^{\frac12} & \text{if } B = +\infty \\
     \begin{cases}
    \sqrt{2}\,\bracket*{\sR_{\Phi_{\mathrm{exp}}}(h)- \sR_{\Phi_{\mathrm{exp}}}^*\paren*{\sH_{\mathrm{NN}}}+\sM_{\Phi_{\mathrm{exp}}}\paren*{\sH_{\mathrm{NN}}}}^{\frac12} \\ \qquad \text{if } \sR_{\Phi_{\mathrm{exp}}}(h)- \sR_{\Phi_{\mathrm{exp}}}^*\paren*{\sH_{\mathrm{NN}}}\leq \frac{1}{2}\paren*{\frac{e^{2\Lambda B}-1}{e^{2\Lambda B}+1}}^2-\sM_{\Phi_{\mathrm{exp}}}\paren*{\sH_{\mathrm{NN}}} \\
    2\paren*{\frac{e^{2\Lambda B}+1}{e^{2\Lambda B}-1}}\paren*{\sR_{\Phi_{\mathrm{exp}}}(h)- \sR_{\Phi_{\mathrm{exp}}}^*\paren*{\sH_{\mathrm{NN}}}+\sM_{\Phi_{\mathrm{exp}}}\paren*{\sH_{\mathrm{NN}}}} \\ \qquad  \text{otherwise}
    \end{cases} & \text{otherwise}
     \end{cases}
\end{multline*}
where the $\paren*{\Phi_{\mathrm{exp}},\sH_{\mathrm{NN}}}$-minimizability gap $\sM_{\Phi_{\mathrm{exp}}}\paren*{\sH_{\mathrm{NN}}}$ is characterized as below, which is less than
the $\paren*{\Phi_{\mathrm{exp}},\sH_{\mathrm{NN}}}$-approximation error when $\Lambda B<+ \infty$:
\begin{align*}
\sM_{\Phi_{\mathrm{exp}}}\paren*{\sH_{\mathrm{NN}}}
& = \sR_{\Phi_{\mathrm{exp}}}^*\paren*{\sH_{\mathrm{NN}}}-
\mathbb{E}_{X}\bracket*{2\sqrt{\eta(x)(1-\eta(x))}\mathds{1}_{1/2\log_2\abs*{\frac{\eta(x)}{1-\eta(x)}}\leq \Lambda W\norm*{x}_p+\Lambda B}}\\
& - \mathbb{E}_{X}\bracket*{\max\curl*{\eta(x),1-\eta(x)}e^{-(\Lambda W\norm*{x}_p+\Lambda B)}\mathds{1}_{1/2\log_2\abs*{\frac{\eta(x)}{1-\eta(x)}}> \Lambda W\norm*{x}_p+\Lambda B}}\\
&-\bracket*{\min\curl*{\eta(x),1-\eta(x)}e^{\Lambda W\norm*{x}_p+\Lambda B}\mathds{1}_{1/2\log_2\abs*{\frac{\eta(x)}{1-\eta(x)}}> \Lambda W\norm*{x}_p+\Lambda B}}\\
& < \sR_{\Phi_{\mathrm{exp}}}^*\paren*{\sH_{\mathrm{NN}}} - \mathbb{E}_{X}\bracket*{2\sqrt{\eta(x)(1-\eta(x))}}\\
& = \sR_{\Phi_{\mathrm{exp}}}^*\paren*{\sH_{\mathrm{NN}}} - \sR_{\Phi_{\mathrm{exp}}}^*\paren*{\sH_{\mathrm{all}}} .
\end{align*}
Therefore, the inequality for $\Lambda B = + \infty$ coincides with the consistency excess error bound
known for the exponential loss \citep{Zhang2003,MohriRostamizadehTalwalkar2018} but the one for $\Lambda B< + \infty$ is distinct and novel.

\subsubsection{Quadratic Loss}
For the quadratic loss $\Phi_{\mathrm{quad}}(\alpha)\colon=(1-\alpha)^2\mathds{1}_{\alpha\leq 1}$, for all $h\in \sH_{\mathrm{NN}}$ and $x\in \sX$:
\begin{equation*}
\begin{aligned}
&\sC_{\Phi_{\mathrm{quad}}}(h,x,t)\\
&=t \Phi_{\mathrm{quad}}(h(x))+(1-t)\Phi_{\mathrm{quad}}(-h(x))\\
&=t\paren*{1-h(x)}^2\mathds{1}_{h(x)\leq 1}+(1-t)\paren*{1+h(x)}^2\mathds{1}_{h(x)\geq -1}.\\
&\inf_{h\in\sH_{\mathrm{NN}}}\sC_{\Phi_{\mathrm{quad}}}(h,x,t)\\
&=\begin{cases}
4t(1-t), \\ \qquad  \abs*{2t-1}\leq\Lambda W\norm*{x}_p+\Lambda B,\\
\max\curl*{t,1-t}\paren*{1-\paren*{\Lambda W\norm*{x}_p+\Lambda B}}^2
+\min\curl*{t,1-t}\paren*{1+\Lambda W\norm*{x}_p+\Lambda B}^2, \\ \qquad  \abs*{2t-1}> \Lambda W\norm*{x}_p+\Lambda B.
\end{cases}
\end{aligned}
\end{equation*}
Therefore, the $\paren*{\Phi_{\mathrm{quad}},\sH_{\mathrm{NN}}}$-minimizability gap can be expressed as follows:
\begin{equation}
\begin{aligned}
\label{eq:M-quad-NN}
\sM_{\Phi_{\mathrm{quad}}}\paren*{\sH_{\mathrm{NN}}}
& = \sR_{\Phi_{\mathrm{quad}}}^*\paren*{\sH_{\mathrm{NN}}}-\mathbb{E}_{X}\bracket*{4\eta(x)(1-\eta(x))\mathds{1}_{\abs*{2\eta(x)-1}\leq \Lambda W\norm*{x}_p+\Lambda B}}\\
& - \mathbb{E}_{X}\bracket*{\max\curl*{\eta(x),1-\eta(x)}\paren*{1-\paren*{\Lambda W\norm*{x}_p+\Lambda B}}^2\mathds{1}_{\abs*{2\eta(x)-1}> \Lambda W\norm*{x}_p+\Lambda B}}\\
& - \mathbb{E}_{X}\bracket*{\min\curl*{\eta(x),1-\eta(x)}\paren*{1+\paren*{\Lambda W\norm*{x}_p+\Lambda B}}^2\mathds{1}_{\abs*{2\eta(x)-1}> \Lambda W\norm*{x}_p+\Lambda B}}
\end{aligned}
\end{equation}
Note $\paren*{\Phi_{\mathrm{quad}},\sH_{\mathrm{NN}}}$-minimizability gap coincides with
the $\paren*{\Phi_{\mathrm{quad}},\sH_{\mathrm{NN}}}$-approximation error:\\
$\sR_{\Phi_{\mathrm{quad}}}^*\paren*{\sH_{\mathrm{NN}}}-
\mathbb{E}_{X}\bracket*{4\eta(x)(1-\eta(x))}$ for $\Lambda B \geq 1$.

For $\frac{1}2< t\leq1$, we have
\begin{align*}
&\inf_{h\in\sH_{\mathrm{NN}}:h(x)<0}\sC_{\Phi_{\mathrm{quad}}}(h,x,t)\\
&=t+(1-t)\\
&=1\\
&\inf_{x\in \sX}\inf_{h\in\sH_{\mathrm{NN}}:h(x)<0}\Delta\sC_{\Phi_{\mathrm{quad}},\sH_{\mathrm{NN}}}(h,x,t)\\
& =\inf_{x\in \sX}\paren*{\inf_{h\in\sH_{\mathrm{NN}}:h(x)<0}\sC_{\Phi_{\mathrm{quad}}}(h,x,t)-\inf_{h\in\sH_{\mathrm{NN}}}\sC_{\Phi_{\mathrm{quad}}}(h,x,t)}\\
&=\inf_{x\in \sX}\begin{cases}
1-4t(1-t),& 2t-1\leq \Lambda W\norm*{x}_p+\Lambda B,\\
1-t\paren*{1-\paren*{\Lambda W\norm*{x}_p+\Lambda B}}^2-(1-t)\paren*{1+\Lambda W\norm*{x}_p+\Lambda B}^2, & 2t-1> \Lambda W\norm*{x}_p+\Lambda B.
\end{cases}\\
&=\begin{cases}
1-4t(1-t),& 2t-1\leq \Lambda B,\\
1-t\paren*{1-\Lambda B}^2-(1-t)\paren*{1+\Lambda B}^2,& 2t-1> \Lambda B.
\end{cases}\\
&=\sT(2t-1)
\end{align*}
where $\sT$ is the increasing and convex function on $[0,1]$ defined by
\begin{align*}
\forall t\in[0,1], \quad
\sT(t)=\begin{cases}
t^2, & t\leq \Lambda B,\\
2\Lambda B \,t-(\Lambda B)^2, & t> \Lambda B.
\end{cases}
\end{align*}
By Definition~\ref{def:trans}, for any $\epsilon\geq 0$, the $\sH_{\mathrm{NN}}$-estimation error transformation of the quadratic loss is as follows:
\begin{align*}
\sT_{\Phi_{\mathrm{quad}}}= 
\begin{cases}
\sT(t), & t\in \left[\epsilon,1\right], \\
\frac{\sT(\epsilon)}{\epsilon}\, t, &  t\in \left[0,\epsilon\right).
\end{cases}
\end{align*}
Therefore, when $\epsilon=0$, $\sT_{\Phi_{\mathrm{quad}}}$ is convex, non-decreasing, invertible and satisfies that $\sT_{\Phi_{\mathrm{quad}}}(0)=0$. By Theorem~\ref{Thm:tightness}, we can choose $\Psi(t)=\sT_{\Phi_{\mathrm{quad}}}(t)$ in Theorem~\ref{Thm:excess_bounds_Psi_uniform}, or equivalently $\Gamma(t) = \sT_{\Phi_{\mathrm{quad}}}^{-1}(t)=
\begin{cases}
\sqrt{t}, & t \leq (\Lambda B)^2 \\
\frac{t}{2\Lambda B}+\frac{\Lambda B}{2}, & t > (\Lambda B)^2
\end{cases}$, in Theorem~\ref{Thm:excess_bounds_Gamma_uniform}, which are optimal. Thus, by Theorem~\ref{Thm:excess_bounds_Psi_uniform} or Theorem~\ref{Thm:excess_bounds_Gamma_uniform}, setting $\e = 0$ yields the $\sH_{\mathrm{NN}}$-consistency bound for the quadratic loss, valid for all $h \in \sH_{\mathrm{NN}}$:
\begin{multline}
\label{eq:quad-NN-est}
    \sR_{\ell_{0-1}}(h)- \sR_{\ell_{0-1}}^*\paren*{\sH_{\mathrm{NN}}}+\sM_{\ell_{0-1}}\paren*{\sH_{\mathrm{NN}}}\\
    \leq
    \begin{cases}
    \bracket*{\sR_{\Phi_{\mathrm{quad}}}(h)- \sR_{\Phi_{\mathrm{quad}}}^*\paren*{\sH_{\mathrm{NN}}}+\sM_{\Phi_{\mathrm{quad}}}\paren*{\sH_{\mathrm{NN}}}}^{\frac12}  \\ \qquad  \text{if } \sR_{\Phi_{\mathrm{quad}}}(h)- \sR_{\Phi_{\mathrm{quad}}}^*\paren*{\sH_{\mathrm{NN}}}\leq (\Lambda B)^2-\sM_{\Phi_{\mathrm{quad}}}\paren*{\sH_{\mathrm{NN}}} \\
    \frac{\sR_{\Phi_{\mathrm{quad}}}(h)- \sR_{\Phi_{\mathrm{quad}}}^*\paren*{\sH_{\mathrm{NN}}}+\sM_{\Phi_{\mathrm{quad}}}\paren*{\sH_{\mathrm{NN}}}}{2\Lambda B}+\frac{\Lambda B}{2} \\ \qquad  \text{otherwise}
    \end{cases}
\end{multline}
Since the $\paren*{\ell_{0-1},\sH_{\mathrm{NN}}}$-minimizability gap coincides with
the $\paren*{\ell_{0-1},\sH_{\mathrm{NN}}}$-approximation error and
$\paren*{\Phi_{\mathrm{quad}},\sH_{\mathrm{NN}}}$-minimizability gap coincides with
the $\paren*{\Phi_{\mathrm{quad}},\sH_{\mathrm{NN}}}$-approximation error for $\Lambda B \geq 1$, the inequality can be rewritten as follows:
\begin{multline*}
     \sR_{\ell_{0-1}}(h)- \sR_{\ell_{0-1}}^*\paren*{\sH_{\mathrm{all}}}\\
     \leq 
     \begin{cases}
     \bracket*{\sR_{\Phi_{\mathrm{quad}}}(h) - \sR_{\Phi_{\mathrm{quad}}}^*\paren*{\sH_{\mathrm{all}}}}^{\frac12} & \text{if } \Lambda B \geq 1\\
     \begin{cases}
    \bracket*{\sR_{\Phi_{\mathrm{quad}}}(h)- \sR_{\Phi_{\mathrm{quad}}}^*\paren*{\sH_{\mathrm{NN}}}+\sM_{\Phi_{\mathrm{quad}}}\paren*{\sH_{\mathrm{NN}}}}^{\frac12}  \\ \qquad  \text{if } \sR_{\Phi_{\mathrm{quad}}}(h)- \sR_{\Phi_{\mathrm{quad}}}^*\paren*{\sH_{\mathrm{NN}}}\leq (\Lambda B)^2-\sM_{\Phi_{\mathrm{quad}}}\paren*{\sH_{\mathrm{NN}}} \\
    \frac{\sR_{\Phi_{\mathrm{quad}}}(h)- \sR_{\Phi_{\mathrm{quad}}}^*\paren*{\sH_{\mathrm{NN}}}+\sM_{\Phi_{\mathrm{quad}}}\paren*{\sH_{\mathrm{NN}}}}{2\Lambda B}+\frac{\Lambda B}{2} \\ \qquad  \text{otherwise}
    \end{cases} & \text{otherwise}
     \end{cases}
\end{multline*}
where the $\paren*{\Phi_{\mathrm{quad}},\sH_{\mathrm{NN}}}$-minimizability gap $\sM_{\Phi_{\mathrm{quad}},\sH_{\mathrm{NN}}}$ is characterized as below,  which is less than
the $\paren*{\Phi_{\mathrm{quad}},\sH_{\mathrm{NN}}}$-approximation error when $\Lambda B<1$:
\begin{align*}
\sM_{\Phi_{\mathrm{quad}}}\paren*{\sH_{\mathrm{NN}}}
& = \sR_{\Phi_{\mathrm{quad}}}^*\paren*{\sH_{\mathrm{NN}}}-\mathbb{E}_{X}\bracket*{4\eta(x)(1-\eta(x))\mathds{1}_{\abs*{2\eta(x)-1}\leq \Lambda W\norm*{x}_p+\Lambda B}}\\
& - \mathbb{E}_{X}\bracket*{\max\curl*{\eta(x),1-\eta(x)}\paren*{1-\paren*{\Lambda W\norm*{x}_p+\Lambda B}}^2\mathds{1}_{\abs*{2\eta(x)-1}> \Lambda W\norm*{x}_p+\Lambda B}}\\
& - \mathbb{E}_{X}\bracket*{\min\curl*{\eta(x),1-\eta(x)}\paren*{1+\paren*{\Lambda W\norm*{x}_p+\Lambda B}}^2\mathds{1}_{\abs*{2\eta(x)-1}> \Lambda W\norm*{x}_p+\Lambda B}}\\
& < \sR_{\Phi_{\mathrm{quad}}}^*\paren*{\sH_{\mathrm{NN}}} - \mathbb{E}_{X}\bracket*{4\eta(x)(1-\eta(x))}\\
& = \sR_{\Phi_{\mathrm{quad}}}^*\paren*{\sH_{\mathrm{NN}}} - \sR_{\Phi_{\mathrm{quad}}}^*\paren*{\sH_{\mathrm{all}}} .
\end{align*}
Therefore, the inequality for $\Lambda B \geq 1$ coincides with the consistency excess error bound
known for the quadratic loss \citep{Zhang2003,bartlett2006convexity} but the one for $\Lambda B< 1$ is distinct and novel.

\subsubsection{Sigmoid Loss}
For the sigmoid loss $\Phi_{\mathrm{sig}}(\alpha)\colon=1-\tanh(k\alpha),~k>0$,
for all $h\in \sH_{\mathrm{NN}}$ and $x\in \sX$:
\begin{equation*}
\begin{aligned}
\sC_{\Phi_{\mathrm{sig}}}(h,x,t)
&=t \Phi_{\mathrm{sig}}(h(x))+(1-t)\Phi_{\mathrm{sig}}(-h(x)),\\
&=t\paren*{1-\tanh(kh(x))}+(1-t)\paren*{1+\tanh(kh(x))}.\\
\inf_{h\in\sH_{\mathrm{NN}}}\sC_{\Phi_{\mathrm{sig}}}(h,x,t)
&=1-\abs*{1-2t}\tanh\paren*{k\paren*{\Lambda W\norm*{x}_p+\Lambda B}}
\end{aligned}
\end{equation*}
Therefore, the $\paren*{\Phi_{\mathrm{sig}},\sH_{\mathrm{NN}}}$-minimizability gap can be expressed as follows:
\begin{equation}
\begin{aligned}
\label{eq:M-sig-NN}
\sM_{\Phi_{\mathrm{sig}}}\paren*{\sH_{\mathrm{NN}}}
&= \sR_{\Phi_{\mathrm{sig}}}^*\paren*{\sH_{\mathrm{NN}}}-\mathbb{E}_{X}\bracket*{\inf_{h\in\sH_{\mathrm{NN}}}\sC_{\Phi_{\mathrm{sig}}}(h,x,\eta(x))}\\
&= \sR_{\Phi_{\mathrm{sig}}}^*\paren*{\sH_{\mathrm{NN}}}-\mathbb{E}_{X}\bracket*{1-\abs*{1-2\eta(x)}\tanh\paren*{k\paren*{\Lambda W\norm*{x}_p+\Lambda B}}}.
\end{aligned}
\end{equation}
Note $\paren*{\Phi_{\mathrm{sig}},\sH_{\mathrm{NN}}}$-minimizability gap coincides with
the $\paren*{\Phi_{\mathrm{sig}},\sH_{\mathrm{NN}}}$-approximation error:\\
$\sR_{\Phi_{\mathrm{sig}}}^*\paren*{\sH_{\mathrm{NN}}}-\mathbb{E}_{X}\bracket*{1-\abs*{1-2\eta(x)}}$ for $\Lambda B = + \infty$.

For $\frac{1}2< t\leq1$, we have
\begin{align*}
\inf_{h\in\sH_{\mathrm{NN}}:h(x)<0}\sC_{\Phi_{\mathrm{sig}}}(h,x,t)
&=1-\abs*{1-2t}\tanh(0)\\
&=1.\\
\inf_{x\in \sX}\inf_{h\in\sH_{\mathrm{NN}}:h(x)<0}\Delta\sC_{\Phi_{\mathrm{sig}},\sH_{\mathrm{NN}}}(h,x,t)
&=\inf_{x\in \sX}\paren*{\inf_{h\in\sH_{\mathrm{NN}}:h(x)<0}\sC_{\Phi_{\mathrm{sig}}}(h,x,t)-\inf_{h\in\sH_{\mathrm{NN}}}\sC_{\Phi_{\mathrm{sig}}}(h,x,t)}\\
&=\inf_{x\in \sX}(2t-1)\tanh\paren*{k\paren*{\Lambda W\norm*{x}_p+\Lambda B}}\\
&=(2t-1)\tanh(k\Lambda B)\\
&=\sT(2t-1)
\end{align*}
where $\sT$ is the increasing and convex function on $[0,1]$ defined by
\begin{align*}
\forall t\in[0,1],\; \sT(t)=\tanh(k\Lambda B) \, t .
\end{align*}
By Definition~\ref{def:trans}, for any $\epsilon\geq 0$, the $\sH_{\mathrm{NN}}$-estimation error transformation of the sigmoid loss is as follows:
\begin{align*}
\sT_{\Phi_{\mathrm{sig}}}= \tanh(k\Lambda B) \, t, \quad t \in [0,1],
\end{align*}
Therefore, $\sT_{\Phi_{\mathrm{sig}}}$ is convex, non-decreasing, invertible and satisfies that $\sT_{\Phi_{\mathrm{sig}}}(0)=0$. By Theorem~\ref{Thm:tightness}, we can choose $\Psi(t)=\tanh(k\Lambda B)\,t$ in Theorem~\ref{Thm:excess_bounds_Psi_uniform}, or equivalently $\Gamma(t)=\frac{t}{\tanh(k\Lambda B)}$ in Theorem~\ref{Thm:excess_bounds_Gamma_uniform}, which are optimal when $\e=0$.
Thus, by Theorem~\ref{Thm:excess_bounds_Psi_uniform} or Theorem~\ref{Thm:excess_bounds_Gamma_uniform},  setting $\e = 0$ yields the $\sH_{\mathrm{NN}}$-consistency bound for the sigmoid loss, valid for all $h \in \sH_{\mathrm{NN}}$:
\begin{align}
\label{eq:sig-NN-est}
     \sR_{\ell_{0-1}}(h)- \sR_{\ell_{0-1}}^*\paren*{\sH_{\mathrm{NN}}}\leq \frac{\sR_{\Phi_{\mathrm{sig}}}(h)- \sR_{\Phi_{\mathrm{sig}}}^*\paren*{\sH_{\mathrm{NN}}}+\sM_{\Phi_{\mathrm{sig}}}\paren*{\sH_{\mathrm{NN}}}}{\tanh(k\Lambda B)}-\sM_{\ell_{0-1}}\paren*{\sH_{\mathrm{NN}}}.
\end{align}
Since the $\paren*{\ell_{0-1},\sH_{\mathrm{NN}}}$-minimizability gap coincides with
the $\paren*{\ell_{0-1},\sH_{\mathrm{NN}}}$-approximation error, and since
$\paren*{\Phi_{\mathrm{sig}},\sH_{\mathrm{NN}}}$-minimizability gap coincides with
the $\paren*{\Phi_{\mathrm{sig}},\sH_{\mathrm{NN}}}$-approximation error for $\Lambda B = + \infty$,
the inequality can be rewritten as follows:
\begin{multline*}
     \sR_{\ell_{0-1}}(h)- \sR_{\ell_{0-1}}^*\paren*{\sH_{\mathrm{all}}}\\
      \leq 
     \begin{cases}
     \sR_{\Phi_{\mathrm{sig}}}(h) - \sR_{\Phi_{\mathrm{sig}}}^*\paren*{\sH_{\mathrm{all}}} & \text{if } \Lambda B = + \infty\\
     \frac{1}{\tanh(k\Lambda B)} \bracket[\Big]{\sR_{\Phi_{\mathrm{sig}}}(h)
     - \mathbb{E}_{X}\bracket*{1-\abs*{1-2\eta(x)}\tanh\paren*{k\paren*{\Lambda W\norm*{x}_p+\Lambda B}}\ } } & \text{otherwise}.
     \end{cases}
\end{multline*}
The inequality for $\Lambda B = + \infty$ coincides with the consistency excess error bound
known for the sigmoid loss \citep{Zhang2003,bartlett2006convexity,MohriRostamizadehTalwalkar2018} but the one for $\Lambda B < + \infty$ is distinct and novel. 
For $\Lambda B<+ \infty$, we have
\begin{align*}
& \mathbb{E}_{X}\bracket*{1-\abs*{1-2\eta(x)}\tanh\paren*{k\paren*{\Lambda W\norm*{x}_p+\Lambda B}}}\\
& \qquad > \mathbb{E}_{X}\bracket*{1-\abs*{2\eta(x)-1}}= 2\mathbb{E}_X\bracket*{\min\curl*{\eta(x), 1 - \eta(x)}}= \sR_{\Phi_{\mathrm{hinge}}}^*\paren*{\sH_{\mathrm{all}}}.
\end{align*}
Therefore for $\Lambda B<+ \infty$,
\begin{align*}
\sR_{\Phi_{\mathrm{sig}}}(h) - \mathbb{E}_{X}\bracket*{1-\abs*{1-2\eta(x)}\tanh\paren*{k\paren*{\Lambda W\norm*{x}_p+\Lambda B}}} < \sR_{\Phi_{\mathrm{sig}}}(h) - \sR_{\Phi_{\mathrm{sig}}}^*\paren*{\sH_{\mathrm{all}}}.
\end{align*}
Note that: $\sR_{\Phi_{\mathrm{sig}}}^*\paren*{\sH_{\mathrm{all}}} = 2 \sR_{\ell_{0-1}}^*\paren*{\sH_{\mathrm{all}}} =2\mathbb{E}_X\bracket*{\min\curl*{\eta(x), 1 - \eta(x)}}$. Thus, the first
inequality (case $\Lambda B = + \infty$) can be equivalently written as follows:
\begin{align*}
    \forall h \in \sH_{\mathrm{NN}},\; \sR_{\ell_{0-1}}(h) 
     \leq \sR_{\Phi_{\mathrm{sig}}}(h) - \mathbb{E}_X\bracket*{\min\curl*{\eta(x), 1 - \eta(x)}},
\end{align*}
which is a more informative upper bound than the standard
inequality $\sR_{\ell_{0-1}}(h) 
     \leq \sR_{\Phi_{\mathrm{sig}}}(h)$.

\subsubsection{\texorpdfstring{$\rho$}{rho}-Margin Loss}
For the $\rho$-margin loss $\Phi_{\rho}(\alpha)\colon=\min\curl*{1,\max\curl*{0,1-\frac{\alpha}{\rho}}},~\rho>0$,
for all $h\in \sH_{\mathrm{NN}}$ and $x\in \sX$:
\begin{equation*}
\begin{aligned}
\sC_{\Phi_{\rho}}(h,x,t)
&=t \Phi_{\rho}(h(x))+(1-t)\Phi_{\rho}(-h(x)),\\
&=t\min\curl*{1,\max\curl*{0,1-\frac{h(x)}{\rho}}}+(1-t)\min\curl*{1,\max\curl*{0,1+\frac{h(x)}{\rho}}}.\\
\inf_{h\in\sH_{\mathrm{NN}}}\sC_{\Phi_{\rho}}(h,x,t)
&=\min\curl*{t,1-t}+\max\curl*{t,1-t}\paren*{1-\frac{\min\curl*{\Lambda W\norm*{x}_p+\Lambda B,\rho}}{\rho}}.
\end{aligned}
\end{equation*}
Therefore, the $\paren*{\Phi_{\rho},\sH_{\mathrm{NN}}}$-minimizability gap can be expressed as follows:
\begin{equation}
\label{eq:M-rho-NN}
\begin{aligned}
& \sM_{\Phi_{\rho}}\paren*{\sH_{\mathrm{NN}}}\\
& = \sR_{\Phi_{\rho}}^*\paren*{\sH_{\mathrm{NN}}}-\mathbb{E}_{X}\bracket*{\inf_{h\in\sH_{\mathrm{NN}}}\sC_{\Phi_{\rho}}(h,x,\eta(x))}\\
& = \sR_{\Phi_{\rho}}^*\paren*{\sH_{\mathrm{NN}}}-\mathbb{E}_{X}\bracket[\Bigg]{\min\curl*{\eta(x),1-\eta(x)}\\
& \qquad +\max\curl*{\eta(x),1-\eta(x)}\paren*{1-\frac{\min\curl*{\Lambda W\norm*{x}_p+\Lambda B,\rho}}{\rho}}}.
 \end{aligned}
\end{equation}
Note the $\paren*{\Phi_{\rho},\sH_{\mathrm{NN}}}$-minimizability gap coincides with
the $\paren*{\Phi_{\rho},\sH_{\mathrm{NN}}}$-approximation error:\\
$\sR_{\Phi_{\rho}}^*\paren*{\sH_{\mathrm{NN}}}-\mathbb{E}_{X}\bracket*{\min\curl*{\eta(x),1-\eta(x)}}$ for $\Lambda B \geq \rho$.

For $\frac{1}2< t\leq1$, we have
\begin{align*}
\inf_{h\in\sH_{\mathrm{NN}}:h(x)<0}\sC_{\Phi_{\rho}}(h,x,t)
&=t+(1-t)\paren*{1-\frac{\min\curl*{\Lambda W\norm*{x}_p+\Lambda B,\rho}}{
\rho}}.\\
\inf_{x\in \sX}\inf_{h\in\sH_{\mathrm{NN}}:h(x)<0}\Delta\sC_{\Phi_{\rho},\sH_{\mathrm{NN}}}(h,x)
&=\inf_{x\in \sX}\paren*{\inf_{h\in\sH_{\mathrm{NN}}:h(x)<0}\sC_{\Phi_{\rho}}(h,x,t)-\inf_{h\in\sH_{\mathrm{NN}}}\sC_{\Phi_{\rho}}(h,x,t)}\\
&=\inf_{x\in \sX}(2t-1)\frac{\min\curl*{\Lambda W\norm*{x}_p+\Lambda B,\rho}}{\rho}\\
&=(2t-1)\frac{\min\curl*{\Lambda B,\rho}}{\rho}\\
&=\sT(2t-1)
\end{align*}
where $\sT$ is the increasing and convex function on $[0,1]$ defined by
\begin{align*}
\forall t\in [0,1],\; \sT(t)=\frac{\min\curl*{\Lambda B,\rho}}{\rho} \, t.    
\end{align*} 
By Definition~\ref{def:trans}, for any $\epsilon\geq 0$, the $\sH_{\mathrm{NN}}$-estimation error transformation of the $\rho$-margin loss is as follows:
\begin{align*}
\sT_{\Phi_{\rho}}= \frac{\min\curl*{\Lambda B,\rho}}{\rho} \, t, \quad t \in [0,1],
\end{align*}
Therefore, $\sT_{\Phi_{\rho}}$ is convex, non-decreasing, invertible and satisfies that $\sT_{\Phi_{\rho}}(0)=0$. By Theorem~\ref{Thm:tightness}, we can choose $\Psi(t)=\frac{\min\curl*{\Lambda B,\rho}}{\rho} \, t$ in Theorem~\ref{Thm:excess_bounds_Psi_uniform}, or equivalently $\Gamma(t)=\frac{\rho }{\min\curl*{\Lambda B,\rho}} \, t$ in Theorem~\ref{Thm:excess_bounds_Gamma_uniform}, which are optimal when $\e=0$.
Thus, by Theorem~\ref{Thm:excess_bounds_Psi_uniform} or Theorem~\ref{Thm:excess_bounds_Gamma_uniform}, setting $\e = 0$ yields the $\sH_{\mathrm{NN}}$-consistency bound for the $\rho$-margin loss, valid for all $h \in \sH_{\mathrm{NN}}$:
\begin{align}
\label{eq:rho-NN-est}
     \sR_{\ell_{0-1}}(h)- \sR_{\ell_{0-1}}^*\paren*{\sH_{\mathrm{NN}}}\leq \frac{\rho\paren*{\sR_{\Phi_{\rho}}(h)- \sR_{\Phi_{\rho}}^*\paren*{\sH_{\mathrm{NN}}}+\sM_{\Phi_{\rho}}\paren*{\sH_{\mathrm{NN}}}}}{\min\curl*{\Lambda B,\rho}}-\sM_{\ell_{0-1}}\paren*{\sH_{\mathrm{NN}}}.
\end{align}
Since the $\paren*{\ell_{0-1},\sH_{\mathrm{NN}}}$-minimizability gap coincides with
the $\paren*{\ell_{0-1},\sH_{\mathrm{NN}}}$-approximation error and
$\paren*{\Phi_{\rho},\sH_{\mathrm{NN}}}$-minimizability gap coincides with
the $\paren*{\Phi_{\rho},\sH_{\mathrm{NN}}}$-approximation error for $\Lambda B \geq \rho$,
the inequality can be rewritten as follows:
\begin{multline*}
     \sR_{\ell_{0-1}}(h)- \sR_{\ell_{0-1}}^*\paren*{\sH_{\mathrm{all}}}\\
      \leq 
     \begin{cases}
     \sR_{\Phi_{\rho}}(h) - \sR_{\Phi_{\rho}}^*\paren*{\sH_{\mathrm{all}}} & \text{if } \Lambda B \geq \rho\\
     \frac{\rho \paren*{\sR_{\Phi_{\rho}}(h)
     - \mathbb{E}_{X}\bracket*{\min\curl*{\eta(x),1-\eta(x)}+\max\curl*{\eta(x),1-\eta(x)}\paren*{1-\frac{\min\curl*{\Lambda W\norm*{x}_p+\Lambda B,\rho}}{\rho}} } }}{\Lambda B}  & \text{otherwise}. 
     \end{cases}
\end{multline*}
Note that: $\sR_{\Phi_{\rho}}^*\paren*{\sH_{\mathrm{all}}} =  \sR_{\ell_{0-1}}^*\paren*{\sH_{\mathrm{all}}} =\mathbb{E}_X\bracket*{\min\curl*{\eta(x), 1 - \eta(x)}}$. Thus, the first
inequality (case $\Lambda B \geq \rho$) can be equivalently written as follows:
\begin{align*}
    \forall h \in \sH_{\mathrm{NN}}, \quad \sR_{\ell_{0-1}}(h) 
     \leq \sR_{\Phi_{\rho}}(h).
\end{align*}
The case $\Lambda B \geq \rho$ is one of the ``trivial cases'' mentioned in Section~\ref{sec:general}, where the trivial inequality $\sR_{\ell_{0-1}}(h) \leq \sR_{\Phi_{\rho}}(h)$ can be obtained directly using the fact that $\ell_{0-1}$ is upper-bounded by $\Phi_{\rho}$. This, however, does not imply that non-adversarial $\sH_{\mathrm{NN}}$-consistency bound for the $\rho$-margin loss is trivial when $\Lambda B>\rho$ since it is optimal.

\section{Derivation of Adversarial \texorpdfstring{$\sH$}{H}-Consistency Bounds}
\label{app:derivation-adv}
\subsection{Linear Hypotheses}
\label{app:derivation-lin-adv}
By the definition of $\sH_{\mathrm{lin}}$, for any $x \in \sX$, 
\begin{align*}
&\uv h_\gamma(x)
  =w \cdot x-\gamma \|w\|_q+b \\
& \in 
  \begin{cases}
\bracket*{-W \norm*{x}_p-\gamma W-B, W\norm*{x}_p - \gamma W+ B}
&  \norm*{x}_p \geq \gamma\\
\bracket*{-W \norm*{x}_p-\gamma W-B, B}
&  \norm*{x}_p < \gamma
\end{cases},\\
&\ov h_\gamma(x)=w \cdot x+\gamma \|w\|_q+b \\
& \in 
\begin{cases}
\bracket*{-W \norm*{x}_p+\gamma W-B, W\norm*{x}_p + \gamma W+ B}
&  \norm*{x}_p \geq \gamma\\
\bracket*{-B, W\norm*{x}_p + \gamma W+ B}
&  \norm*{x}_p < \gamma
\end{cases}.
\end{align*}
Note $\sH_{\mathrm{lin}}$ is symmetric. For any $x\in \sX$, there exist $w=0$ and any $0<b\leq B$ such that $w \cdot x-\gamma \|w\|_q+b>0$. Thus by Lemma~\ref{lemma:explicit_assumption_01_adv}, for any $x\in \sX$, $\sC^*_{\ell_{\gamma}}\paren*{\sH_{\mathrm{lin}}, x} =\min\curl*{\eta(x), 1 - \eta(x)}$. The $\paren*{\ell_{\gamma},\sH_{\mathrm{lin}}}$-minimizability gap can be expressed as follows:
\begin{align}
\label{eq:M-01-lin-adv}
\sM_{\ell_{\gamma}}\paren*{\sH_{\mathrm{lin}}}
& = \sR_{\ell_{\gamma}}^*\paren*{\sH_{\mathrm{lin}}}-\mathbb{E}_{X}\bracket*{\min\curl*{\eta(x),1-\eta(x)}}.
\end{align}

\subsubsection{Supremum-Based \texorpdfstring{$\rho$}{rho}-Margin Loss}
\label{app:rho-lin-adv}
For the supremum-based $\rho$-margin loss 
\begin{align*}
\wt{\Phi}_{\rho}\colon=\sup_{x'\colon \|x-x'\|_p\leq \gamma}\Phi_{\rho}(y h(x')),  \quad \text{where } \Phi_{\rho}(\alpha)=\min\curl*{1,\max\curl*{0,1-\frac{\alpha}{\rho}}},~\rho>0,   
\end{align*}
for all $h\in \sH_{\mathrm{lin}}$ and $x\in \sX$:
\begin{equation*}
\begin{aligned}
\sC_{\wt{\Phi}_{\rho}}(h,x,t) 
&=t \wt{\Phi}_{\rho}(h(x))+(1-t)\wt{\Phi}_{\rho}(-h(x))\\
& = t\Phi_{\rho}\paren*{\uv h_\gamma(x)}+(1-t)\Phi_{\rho}\paren*{-\ov h_\gamma(x)}\\
& =t\min\curl*{1,\max\curl*{0,1-\frac{\uv h_\gamma(x)}{\rho}}}+(1-t)\min\curl*{1,\max\curl*{0,1+\frac{\ov h_\gamma(x)}{\rho}}}.\\
\inf_{h\in\sH_{\mathrm{lin}}}\sC_{\wt{\Phi}_{\rho}}(h,x,t)
& =\max\curl*{t,1-t}\paren*{1-\frac{\min\curl*{W\max \curl*{\norm*{x}_p,\gamma}-\gamma W+B,\rho}}{\rho}} + \min\curl*{t,1-t}.
\end{aligned}
\end{equation*}
Therefore, the $\paren*{\wt{\Phi}_{\rho},\sH_{\mathrm{lin}}}$-minimizability gap can be expressed as follows:
\begin{equation}
\begin{aligned}
\label{eq:M-rho-lin-adv}
\sM_{\wt{\Phi}_{\rho}}\paren*{\sH_{\mathrm{lin}}} 
& = \sR_{\wt{\Phi}_{\rho}}^*\paren*{\sH_{\mathrm{lin}}}- \mathbb{E}_{X}\bracket*{\inf_{h\in\sH_{\mathrm{lin}}}\sC_{\wt{\Phi}_{\rho}}(h,x,\eta(x))}\\
& = \sR_{\wt{\Phi}_{\rho}}^*\paren*{\sH_{\mathrm{lin}}}- \mathbb{E}_{X}\bracket*{\max\curl*{\eta(x),1-\eta(x)}\paren*{1-\frac{\min\curl*{W\max\curl*{\norm*{x}_p,\gamma}-\gamma W+B,\rho}}{\rho}}}\\
& -\mathbb{E}_{X}\bracket*{\min\curl*{\eta(x),1-\eta(x)}}.    
\end{aligned}
\end{equation}
For $\frac{1}2< t\leq1$, we have
\begin{align*}
\inf_{h\in\sH_{\mathrm{lin}}:\uv h_\gamma(x)\leq 0 \leq \ov h_\gamma(x)}\sC_{\wt{\Phi}_{\rho}}(h,x,t)
& = t+(1-t)\\
& =1\\
\inf_{x\in \sX} \inf_{h\in\sH_{\mathrm{lin}}\colon \uv h_\gamma(x)\leq 0 \leq \ov h_\gamma(x)}\Delta\sC_{\wt{\Phi}_{\rho},\sH_{\mathrm{lin}}}(h,x,t)
& = \inf_{x\in \sX} \curl*{\inf_{h\in\sH_{\mathrm{lin}}:\uv h_\gamma(x)\leq 0 \leq \ov h_\gamma(x)}\sC_{\wt{\Phi}_{\rho}}(h,x,t)-\inf_{h\in\sH_{\mathrm{lin}}}\sC_{\wt{\Phi}_{\rho}}(h,x,t)}\\
&=\inf_{x\in \sX}\frac{\min\curl*{W\max \curl*{\norm*{x}_p,\gamma}-\gamma W+B,\rho}}{\rho}\,t\\
&=\frac{\min\curl*{B,\rho}}{\rho}\,t\\
&=\sT_1(t),
\end{align*}
where $\sT_1$ is the increasing and convex function on $\bracket*{0,1}$ defined by
\begin{align*}
\forall t \in \bracket*{0,1}, \quad \sT_1(t) = \frac{\min\curl*{B,\rho}}{\rho} \, t \,;
\end{align*}
\begin{align*}
\inf_{h\in\sH_{\mathrm{lin}}:\ov h_\gamma(x)<0}\sC_{\wt{\Phi}_{\rho}}(h,x,t)
& = t+(1-t)\paren*{1-\frac{\min\curl*{W\max \curl*{\norm*{x}_p,\gamma}-\gamma W+B,\rho}}{\rho}}\\
\inf_{x\in \sX} \inf_{h\in\sH_{\mathrm{lin}}\colon \ov h_\gamma(x)< 0}\Delta\sC_{\wt{\Phi}_{\rho},\sH_{\mathrm{lin}}}(h,x,t)
& = \inf_{x\in \sX} \curl*{\inf_{h\in\sH_{\mathrm{lin}}:\ov h_\gamma(x)< 0}\sC_{\wt{\Phi}_{\rho}}(h,x,t)-\inf_{h\in\sH_{\mathrm{lin}}}\sC_{\wt{\Phi}_{\rho}}(h,x,t)}\\
&=\inf_{x\in \sX}(2t-1)\frac{\min\curl*{W\max \curl*{\norm*{x}_p,\gamma}-\gamma W+B,\rho}}{\rho}\\
&=(2t-1)\frac{\min\curl*{B,\rho}}{\rho}\\
&=\sT_2(2t - 1),
\end{align*}
where $\sT_2$ is the increasing and convex function on $\bracket*{0,1}$ defined by
\begin{align*}
\forall t \in [0,1], \quad \sT_2(t) = \frac{\min\curl*{B,\rho}}{\rho} \, t \,;
\end{align*}
By Definition~\ref{def:trans-adv}, for any $\epsilon\geq 0$, the adversarial $\sH_{\mathrm{lin}}$-estimation error transformation of the supremum-based $\rho$-margin loss is as follows:
\begin{align*}
\sT_{\wt{\Phi}_{\rho}}= \frac{\min\curl*{B,\rho}}{\rho} \, t, \quad t \in [0,1],
\end{align*}
Therefore, $\sT_1=\sT_2$ and $\sT_{\wt{\Phi}_{\rho}}$ is convex, non-decreasing, invertible and satisfies that $\sT_{\wt{\Phi}_{\rho}}(0)=0$. By Theorem~\ref{Thm:tightness-adv}, we can choose $\Psi(t)=\frac{\min\curl*{B,\rho}}{\rho}\,t$ in Theorem~\ref{Thm:excess_bounds_Psi_uniform-adv}, or equivalently $\Gamma(t) = \frac{\rho}{\min\curl*{B,\rho}}\,t$ in Theorem~\ref{Thm:excess_bounds_Gamma_uniform-adv}, which are optimal when $\e=0$.
Thus, by Theorem~\ref{Thm:excess_bounds_Psi_uniform-adv} or Theorem~\ref{Thm:excess_bounds_Gamma_uniform-adv}, setting $\e = 0$ yields the adversarial $\sH_{\mathrm{lin}}$-consistency bound for the supremum-based $\rho$-margin loss, valid for all $h \in \sH_{\mathrm{lin}}$:
\begin{align}
\label{eq:rho-lin-est-adv}
     \sR_{\ell_{\gamma}}(h)- \sR_{\ell_{\gamma}}^*\paren*{\sH_{\mathrm{lin}}}
     \leq \frac{\rho\paren*{\sR_{\wt{\Phi}_{\rho}}(h)- \sR_{\wt{\Phi}_{\rho}}^*\paren*{\sH_{\mathrm{lin}}}+\sM_{\wt{\Phi}_{\rho}}\paren*{\sH_{\mathrm{lin}}}}}{\min\curl*{B,\rho}}-\sM_{\ell_{\gamma}, \sH_{\mathrm{lin}}}.
\end{align}
Since
\begin{align*}
\sM_{\ell_{\gamma}}\paren*{\sH_{\mathrm{lin}}}
& = \sR_{\ell_{\gamma}}^*\paren*{\sH_{\mathrm{lin}}}-\mathbb{E}_{X}\bracket*{\min\curl*{\eta(x),1-\eta(x)}},\\
\sM_{\wt{\Phi}_{\rho}}\paren*{\sH_{\mathrm{lin}}} 
& = \sR_{\wt{\Phi}_{\rho}}^*\paren*{\sH_{\mathrm{lin}}} - \mathbb{E}_{X}\bracket*{\max\curl*{\eta(x),1-\eta(x)}\paren*{1-\frac{\min\curl*{W\max \curl*{\norm*{x}_p,\gamma}-\gamma W+B,\rho}}{\rho}}}\\
& -\mathbb{E}_{X}\bracket*{\min\curl*{\eta(x),1-\eta(x)}},
\end{align*}
inequality \eqref{eq:rho-lin-est-adv} can be rewritten as follows:
\begin{align}
\label{eq:rho-lin-est-adv-2}
     \sR_{\ell_{\gamma}}(h) \leq
     \begin{cases}
     \sR_{\wt{\Phi}_{\rho}}(h)- \mathbb{E}_{X}\bracket*{\max\curl*{\eta(x),1-\eta(x)}\paren*{1-\frac{\min\curl*{W\max \curl*{\norm*{x}_p,\gamma}-\gamma W+B,\rho}}{\rho}}} & \text{if } B \geq \rho\\
    \frac{\rho \paren*{\sR_{\wt{\Phi}_{\rho}}(h)- \mathbb{E}_{X}\bracket*{\max\curl*{\eta(x),1-\eta(x)}\paren*{1-\frac{\min\curl*{W\max \curl*{\norm*{x}_p,\gamma}-\gamma W+B,\rho}}{\rho}}}}}{\min\curl*{B,\rho}}\\
    +\paren*{1-\frac{\rho}{\min\curl*{B,\rho}}}\mathbb{E}_{X}\bracket*{\min\curl*{\eta(x),1-\eta(x)}} & \text{otherwise}.
     \end{cases}
\end{align}
Note that: $\min\curl*{W\max \curl*{\norm*{x}_p,\gamma}-\gamma W+B,\rho} = \rho$ if $B \geq \rho$. Thus, the first
inequality (case $B \geq \rho$) can be equivalently written as follows:
\begin{align}
\label{eq:rho-lin-est-adv-3}
    \forall h \in \sH_{\mathrm{lin}}, \quad \sR_{\ell_{\gamma}}(h) 
     \leq \sR_{\wt{\Phi}_{\rho}}(h).
\end{align}
The case $B \geq \rho$ is one of the ``trivial cases'' mentioned in Section~\ref{sec:general}, where the trivial inequality $\sR_{\ell_{\gamma}}(h) \leq \sR_{\wt{\Phi}_{\rho}}(h)$ can be obtained directly using the fact that $\ell_{\gamma}$ is upper-bounded by $\wt{\Phi}_{\rho}$. This, however, does not imply that adversarial $\sH_{\mathrm{lin}}$-consistency bound for the supremum-based $\rho$-margin loss is trivial when $B>\rho$ since it is optimal.

\subsection{One-Hidden-Layer ReLU Neural Networks}
\label{app:derivation-NN-adv}
By the definition of $\sH_{\mathrm{NN}}$, for any $x \in \sX$, 
\begin{align*}
&\uv h_\gamma(x)=\inf_{x'\colon \|x-x'\|_p\leq \gamma}\sum_{j = 1}^n u_j(w_j \cdot x'+b)_{+}\\
&\ov h_\gamma(x)=\sup_{x'\colon \|x-x'\|_p\leq \gamma}\sum_{j = 1}^n u_j(w_j \cdot x'+b)_{+} 
\end{align*}
Note $\sH_{\mathrm{NN}}$ is symmetric. For any $x\in \sX$, there exist $u=\paren*{\frac{1}{\Lambda},\ldots,\frac{1}{\Lambda}}$, $w=0$ and any $0<b\leq B$ satisfy that $\uv h_\gamma(x)>0$. Thus by Lemma~\ref{lemma:explicit_assumption_01_adv}, for any $x\in \sX$, $\sC^*_{\ell_{\gamma}}\paren*{\sH_{\mathrm{NN}}, x} =\min\curl*{\eta(x), 1 - \eta(x)}$. The $\paren*{\ell_{\gamma},\sH_{\mathrm{NN}}}$-minimizability gap can be expressed as follows:
\begin{align}
\label{eq:M-01-NN-adv}
\sM_{\ell_{\gamma}}\paren*{\sH_{\mathrm{NN}}}
& = \sR_{\ell_{\gamma}}^*\paren*{\sH_{\mathrm{NN}}}-\mathbb{E}_{X}\bracket*{\min\curl*{\eta(x),1-\eta(x)}}.
\end{align}

\subsubsection{Supremum-Based \texorpdfstring{$\rho$}{rho}-Margin Loss}
\label{app:derivation-NN-adv-rho}
For the supremum-based $\rho$-margin loss 
\begin{align*}
\wt{\Phi}_{\rho}=\sup_{x'\colon \|x-x'\|_p\leq \gamma}\Phi_{\rho}(y h(x')),  \quad \text{where } \Phi_{\rho}(\alpha)=\min\curl*{1,\max\curl*{0,1-\frac{\alpha}{\rho}}},~\rho>0,   
\end{align*}
for all $h\in \sH_{\mathrm{NN}}$ and $x\in \sX$:
\begin{equation*}
\begin{aligned}
\sC_{\wt{\Phi}_{\rho}}(h,x,t) 
&=t \wt{\Phi}_{\rho}(h(x))+(1-t)\wt{\Phi}_{\rho}(-h(x))\\
& = t\Phi_{\rho}\paren*{\uv h_\gamma(x)}+(1-t)\Phi_{\rho}\paren*{-\ov h_\gamma(x)}\\
& =t\min\curl*{1,\max\curl*{0,1-\frac{\uv h_\gamma(x)}{\rho}}}+(1-t)\min\curl*{1,\max\curl*{0,1+\frac{\ov h_\gamma(x)}{\rho}}}.\\
\inf_{h\in\sH_{\mathrm{NN}}}\sC_{\wt{\Phi}_{\rho}}(h,x,t)
& =\max\curl*{t,1-t}\paren*{1-\frac{\min\curl*{\sup_{h\in\sH_{\mathrm{NN}}}\uv h_\gamma(x),\rho}}{\rho}} + \min\curl*{t,1-t}.
\end{aligned}
\end{equation*}
Therefore, the $\paren*{\wt{\Phi}_{\rho},\sH_{\mathrm{NN}}}$-minimizability gap can be expressed as follows:
\begin{equation}
\begin{aligned}
\label{eq:M-rho-NN-adv}
\sM_{\wt{\Phi}_{\rho}}\paren*{\sH_{\mathrm{NN}}}
& = \sR_{\wt{\Phi}_{\rho}}^*\paren*{\sH_{\mathrm{NN}}} - \mathbb{E}_{X}\bracket*{\inf_{h\in\sH_{\mathrm{NN}}}\sC_{\wt{\Phi}_{\rho}}(h,x,\eta(x))}\\
& = \sR_{\wt{\Phi}_{\rho}}^*\paren*{\sH_{\mathrm{NN}}} - \mathbb{E}_{X}\bracket*{\max\curl*{\eta(x),1-\eta(x)}\paren*{1-\frac{\min\curl*{\sup_{h\in\sH_{\mathrm{NN}}}\uv h_\gamma(x),\rho}}{\rho}}}\\
& -\mathbb{E}_{X}\bracket*{\min\curl*{\eta(x),1-\eta(x)}}.
\end{aligned}
\end{equation}
For $\frac{1}2<t\leq1$, we have
\begin{align*}
& \inf_{h\in\sH_{\mathrm{NN}}:\uv h_\gamma(x)\leq 0 \leq \ov h_\gamma(x)}\sC_{\wt{\Phi}_{\rho}}(h,x,t) \\
& = t+(1-t)\\
& = 1\\
& \inf_{x\in \sX} \inf_{h\in\sH_{\mathrm{NN}}\colon \uv h_\gamma(x)\leq 0 \leq \ov h_\gamma(x)}\Delta\sC_{\wt{\Phi}_{\rho},\sH_{\mathrm{NN}}}(h,x,t)\\
& = \inf_{x\in \sX} \curl*{\inf_{h\in\sH_{\mathrm{NN}}:\uv h_\gamma(x)\leq 0 \leq \ov h_\gamma(x)}\sC_{\wt{\Phi}_{\rho}}(h,x,t)-\inf_{h\in\sH_{\mathrm{NN}}}\sC_{\wt{\Phi}_{\rho}}(h,x,t)}\\
&=\inf_{x\in \sX}\frac{\min\curl*{\sup_{h\in\sH_{\mathrm{NN}}}\uv h_\gamma(x),\rho}}{\rho}\,t\\
&=\frac{\min\curl*{\inf_{x\in\sX}\sup_{h\in\sH_{\mathrm{NN}}}\uv h_\gamma(x),\rho}}{\rho}\,t\\
&=\sT_1(\eta(x)),
\end{align*}
where $\sT_1$ is the increasing and convex function on $\bracket*{0,1}$ defined by
\begin{align*}
\forall t \in \bracket*{0,1}, \quad \sT_1(t) = \frac{\min\curl*{\inf_{x\in\sX}\sup_{h\in\sH_{\mathrm{NN}}}\uv h_\gamma(x),\rho}}{\rho} \, t \,;
\end{align*}
\begin{align*}
\inf_{h\in\sH_{\mathrm{NN}}:\ov h_\gamma(x)<0}\sC_{\wt{\Phi}_{\rho}}(h,x,t)
& = t+(1-t)\paren*{1-\frac{\min\curl*{\sup_{h\in\sH_{\mathrm{NN}}}\uv h_\gamma(x),\rho}}{\rho}}\\
\inf_{x\in \sX} \inf_{h\in\sH_{\mathrm{NN}}\colon \ov h_\gamma(x)< 0}\Delta\sC_{\wt{\Phi}_{\rho},\sH_{\mathrm{NN}}}(h,x,t)
& = \inf_{x\in \sX} \curl*{\inf_{h\in\sH_{\mathrm{NN}}:\ov h_\gamma(x)< 0}\sC_{\wt{\Phi}_{\rho}}(h,x,t)-\inf_{h\in\sH_{\mathrm{NN}}}\sC_{\wt{\Phi}_{\rho}}(h,x,t)}\\
&=\inf_{x\in \sX}(2t-1)\frac{\min\curl*{\sup_{h\in\sH_{\mathrm{NN}}}\uv h_\gamma(x),\rho}}{\rho}\\
&=(2t-1)\frac{\min\curl*{\inf_{x\in\sX}\sup_{h\in\sH_{\mathrm{NN}}}\uv h_\gamma(x),\rho}}{\rho}\\
&=\sT_2(2t - 1),
\end{align*}
where $\sT_2$ is the increasing and convex function on $\bracket*{0,1}$ defined by
\begin{align*}
\forall t \in [0,1], \quad \sT_2(t) = \frac{\min\curl*{\inf_{x\in\sX}\sup_{h\in\sH_{\mathrm{NN}}}\uv h_\gamma(x),\rho}}{\rho} \, t \,;
\end{align*}
By Definition~\ref{def:trans-adv}, for any $\epsilon\geq 0$, the adversarial $\sH_{\mathrm{NN}}$-estimation error transformation of the supremum-based $\rho$-margin loss is as follows:
\begin{align*}
\sT_{\wt{\Phi}_{\rho}}= \frac{\min\curl*{\inf_{x\in\sX}\sup_{h\in\sH_{\mathrm{NN}}}\uv h_\gamma(x),\rho}}{\rho}\,t, \quad t \in [0,1],
\end{align*}
Therefore, $\sT_1=\sT_2$ and $\sT_{\wt{\Phi}_{\rho}}$ is convex, non-decreasing, invertible and satisfies that $\sT_{\wt{\Phi}_{\rho}}(0)=0$. By Theorem~\ref{Thm:tightness-adv}, we can choose $\Psi(t)=\frac{\min\curl*{\inf_{x\in\sX}\sup_{h\in\sH_{\mathrm{NN}}}\uv h_\gamma(x),\rho}}{\rho}\,t$ in Theorem~\ref{Thm:excess_bounds_Psi_uniform-adv}, or equivalently $\Gamma(t) = \frac{\rho}{\min\curl*{\inf_{x\in\sX}\sup_{h\in\sH_{\mathrm{NN}}}\uv h_\gamma(x),\rho}} \, t$ in Theorem~\ref{Thm:excess_bounds_Gamma_uniform-adv}, which are optimal when $\e=0$.
Thus, by Theorem~\ref{Thm:excess_bounds_Psi_uniform-adv} or Theorem~\ref{Thm:excess_bounds_Gamma_uniform-adv}, setting $\e = 0$ yields the adversarial $\sH_{\mathrm{NN}}$-consistency bound for the supremum-based $\rho$-margin loss, valid for all $h \in \sH_{\mathrm{NN}}$:
\begin{align}
\label{eq:rho-NN-est-adv}
     \sR_{\ell_{\gamma}}(h)- \sR_{\ell_{\gamma}}^*\paren*{\sH_{\mathrm{NN}}}
     \leq \frac{\rho\paren*{\sR_{\wt{\Phi}_{\rho}}(h)- \sR_{\wt{\Phi}_{\rho}}^*\paren*{\sH_{\mathrm{NN}}}+\sM_{\wt{\Phi}_{\rho}}\paren*{\sH_{\mathrm{NN}}}}}{\min\curl*{\inf_{x\in\sX}\sup_{h\in\sH_{\mathrm{NN}}}\uv h_\gamma(x),\rho}}-\sM_{\ell_{\gamma}, \sH_{\mathrm{NN}}}.
\end{align}
Observe that
\begin{align*}
\inf_{x\in\sX}\sup_{h\in\sH_{\mathrm{NN}}}\uv h_\gamma(x)
&\geq
\sup_{h\in\sH_{\mathrm{NN}}}\inf_{x\in\sX}\uv h_\gamma(x)\\
& = \sup_{\|u \|_{1}\leq \Lambda,~\|w_j\|_q\leq W,~\abs*{b}\leq B}\inf_{x\in\sX}\inf_{\|s\|_p\leq \gamma}\sum_{j = 1}^n u_j(w_j \cdot x + w_j \cdot s +b)_{+}\\
& \geq \sup_{\|u \|_{1}\leq \Lambda,~\abs*{b}\leq B} \inf_{x\in\sX}\inf_{\|s\|_p\leq \gamma} \sum_{j = 1}^n u_j(0 \cdot x + 0 \cdot s +b)_{+}\\
& = \sup_{\|u \|_{1}\leq \Lambda,~\abs*{b}\leq B} \sum_{j = 1}^n u_j(b)_{+}\\
& = \Lambda B.
\end{align*}
Thus, the inequality can be relaxed as follows:
\begin{align}
\label{eq:rho-NN-est-adv-2}
     \sR_{\ell_{\gamma}}(h)- \sR_{\ell_{\gamma}}^*\paren*{\sH_{\mathrm{NN}}}
     \leq \frac{\rho\paren*{\sR_{\wt{\Phi}_{\rho}}(h)- \sR_{\wt{\Phi}_{\rho}}^*\paren*{\sH_{\mathrm{NN}}}+\sM_{\wt{\Phi}_{\rho}}\paren*{\sH_{\mathrm{NN}}}}}{\min\curl*{\Lambda B,\rho}}-\sM_{\ell_{\gamma}, \sH_{\mathrm{NN}}}.
\end{align}
Since
\begin{align*}
\sM_{\ell_{\gamma}}\paren*{\sH_{\mathrm{NN}}}
& = \sR_{\ell_{\gamma}}^*\paren*{\sH_{\mathrm{NN}}}-\mathbb{E}_{X}\bracket*{\min\curl*{\eta(x),1-\eta(x)}},\\
\sM_{\wt{\Phi}_{\rho}}\paren*{\sH_{\mathrm{NN}}}
& = \sR_{\wt{\Phi}_{\rho}}^*\paren*{\sH_{\mathrm{NN}}} - \mathbb{E}_{X}\bracket*{\max\curl*{\eta(x),1-\eta(x)}\paren*{1-\frac{\min\curl*{\sup_{h\in\sH_{\mathrm{NN}}}\uv h_\gamma(x),\rho}}{\rho}}}\\
& -\mathbb{E}_{X}\bracket*{\min\curl*{\eta(x),1-\eta(x)}},
\end{align*}
inequality \eqref{eq:rho-NN-est-adv} can be rewritten as follows:
\begin{align*}
\sR_{\ell_{\gamma}}(h)
     \leq
     \begin{cases}
     \sR_{\wt{\Phi}_{\rho}}(h)- \mathbb{E}_{X}\bracket*{\max\curl*{\eta(x),1-\eta(x)}\paren*{1-\frac{\min\curl*{\sup_{h\in\sH_{\mathrm{NN}}}\uv h_\gamma(x),\rho}}{\rho}}} & \text{if } \Lambda B \geq \rho\\
    \frac{\rho \paren*{\sR_{\wt{\Phi}_{\rho}}(h)- \mathbb{E}_{X}\bracket*{\max\curl*{\eta(x),1-\eta(x)}\paren*{1-\frac{\min\curl*{\sup_{h\in\sH_{\mathrm{NN}}}\uv h_\gamma(x),\rho}}{\rho}}}}}{\min\curl*{\Lambda B,\rho}}\\
    +\paren*{1-\frac{\rho}{\min\curl*{\Lambda B,\rho}}}\mathbb{E}_{X}\bracket*{\min\curl*{\eta(x),1-\eta(x)}} & \text{otherwise}.
     \end{cases}
\end{align*}
Observe that
\begin{align*}
\sup_{h\in\sH_{\mathrm{NN}}}\uv h_\gamma(x)
&=
\sup_{ \|u \|_{1}\leq \Lambda,~\|w_j\|_q\leq W,~\abs*{b}\leq B}\inf_{x'\colon \|x-x'\|_p\leq \gamma}\sum_{j = 1}^n u_j(w_j \cdot x'+b)_{+}\\
& \leq \inf_{x'\colon \|x-x'\|_p\leq \gamma} \sup_{ \|u \|_{1}\leq \Lambda,~\|w_j\|_q\leq W,~\abs*{b}\leq B} \sum_{j = 1}^n u_j(w_j \cdot x'+b)_{+} \\
& = \inf_{x'\colon \|x-x'\|_p\leq \gamma} \Lambda\paren*{W\norm*{x'}_p+B}\\
& =
\begin{cases}
\Lambda\paren*{W\norm*{x}_p-\gamma W + B} & \text{if } \norm*{x}_p \geq \gamma\\
\Lambda B & \text{if } \norm*{x}_p < \gamma
\end{cases}\\
& = \Lambda\paren*{W\max \curl*{\norm*{x}_p,\gamma}-\gamma W + B}.
\end{align*}
Thus, the inequality can be further relaxed as follows:
\begin{align}
\label{eq:rho-NN-est-adv-3}
\sR_{\ell_{\gamma}}(h)
     \leq
     \begin{cases}
     \sR_{\wt{\Phi}_{\rho}}(h)- \mathbb{E}_{X}\bracket*{\max\curl*{\eta(x),1-\eta(x)}\paren*{1-\frac{\min\curl*{\Lambda\paren*{W\max \curl*{\norm*{x}_p,\gamma}-\gamma W + B},\rho}}{\rho}}} & \text{if } \Lambda B \geq \rho\\
    \frac{\rho \paren*{\sR_{\wt{\Phi}_{\rho}}(h)- \mathbb{E}_{X}\bracket*{\max\curl*{\eta(x),1-\eta(x)}\paren*{1-\frac{\min\curl*{\Lambda\paren*{W\max \curl*{\norm*{x}_p,\gamma}-\gamma W + B},\rho}}{\rho}}}}}{\min\curl*{\Lambda B,\rho}}\\
    +\paren*{1-\frac{\rho}{\min\curl*{\Lambda B,\rho}}}\mathbb{E}_{X}\bracket*{\min\curl*{\eta(x),1-\eta(x)}} & \text{otherwise}.
     \end{cases}
\end{align}
Note the relaxed adversarial $\sH_{\mathrm{NN}}$-consistency bounds \eqref{eq:rho-NN-est-adv} and \eqref{eq:rho-NN-est-adv-3} for the supremum-based $\rho$-margin loss are identical to the bounds \eqref{eq:rho-lin-est-adv} and \eqref{eq:rho-lin-est-adv-2} in the linear case respectively  modulo the replacement of $B$ by $\Lambda B$.


\section{Derivation of Non-Adversarial
\texorpdfstring{$\sH_{\mathrm{all}}$}{all}-Consistency Bounds under Massart's Noise Condition}
\label{app:derivation-all_noise}
With Massart's noise condition, we introduce a modified $\sH$-estimation error transformation. 
We assume that $\epsilon=0$ throughout this section.
\begin{restatable}{proposition}{TightnessNoise}
\label{prop:prop-noise}
Under Massart's noise condition with $\beta$, the modified $\sH$-estimation error transformation of $\Phi$ for $\epsilon=0$ is defined on $t\in \left[0,1\right]$ by,
\begin{align*}
\sT^M_{\Phi}\paren*{t}= \sT(t)\mathds{1}_{t\in \left[2\beta,1\right]}+(\sT(2\beta)/2\beta)\,t\mathds{1}_{t\in \left[0,2\beta\right)},
\end{align*}
with $\sT(t)$ defined in Definition~\ref{def:trans}. Suppose that $\sH$ satisfies the condition of
Lemma~\ref{lemma:explicit_assumption_01} and $\wt{\sT}^M_{\Phi}$ is any lower bound of $\sT^M_{\Phi}$ such that $\wt{\sT}^M_{\Phi}\leq \sT^M_{\Phi}$. If
$\wt{\sT}^M_{\Phi}$ is convex with $\wt{\sT}^M_{\Phi}(0)=0$, then, for any hypothesis $h\in\sH$ and any distribution under Massart's noise condition with $\beta$,
\begin{equation*}
     \wt{\sT}^M_{\Phi}\paren*{\sR_{\ell_{0-1}}(h)- \sR_{\ell_{0-1}}^*(\sH)+\sM_{\ell_{0-1}}(\sH)}
     \leq  \sR_{\Phi}(h)-\sR_{\Phi}^*(\sH)+\sM_{\Phi}(\sH).
\end{equation*}
\end{restatable}
\begin{proof}
Note the condition~\eqref{eq:condition_Psi_general} in Theorem~\ref{Thm:excess_bounds_Psi_01_general} is symmetric about $\Delta \eta(x)=0$. Thus, condition~\eqref{eq:condition_Psi_general} uniformly holds for all distributions is equivalent to the following holds for any $t\in\left[1/2+\beta,1\right]\colon$
\begin{align}
\label{eq:condition_Psi_general-massarts}
\Psi \paren*{\tri*{2t-1}_{\e}}\leq \inf_{x\in \sX,h\in\sH:h(x)<0}\Delta\sC_{\Phi,\sH}(h,x,t),
\end{align}
It is clear that any lower bound $\wt{\sT}^M_{\Phi}$ of the modified $\sH$-estimation error transformation verified condition~\eqref{eq:condition_Psi_general-massarts}. Then by Theorem~\ref{Thm:excess_bounds_Psi_01_general}, the proof is completed.
\end{proof}
\subsection{Quadratic Loss}
For the quadratic loss $\Phi_{\mathrm{quad}}(\alpha)\colon=(1-\alpha)^2\mathds{1}_{\alpha\leq 1}$, 
for all $h\in \sH_{\mathrm{all}}$ and $x\in \sX$:
\begin{align*}
\sC_{\Phi_{\mathrm{quad}}}(h,x,t)
&=t \Phi_{\mathrm{quad}}(h(x))+(1-t)\Phi_{\mathrm{quad}}(-h(x))\\
&=t\paren*{1-h(x)}^2\mathds{1}_{h(x)\leq 1}+(1-t)\paren*{1+h(x)}^2\mathds{1}_{h(x)\geq -1}.\\
\inf_{h\in\sH_{\mathrm{all}}}\sC_{\Phi_{\mathrm{quad}}}(h,x,t)&=4t(1-t)\\
\sM_{\Phi_{\mathrm{quad}}}\paren*{\sH_{\mathrm{all}}}
& = \sR_{\Phi_{\mathrm{quad}}}^*\paren*{\sH_{\mathrm{all}}}-\mathbb{E}_{X}\bracket*{\inf_{h\in\sH_{\mathrm{all}}}\sC_{\Phi_{\mathrm{quad}}}(h,x,\eta(x))}\\
& = \sR_{\Phi_{\mathrm{quad}}}^*\paren*{\sH_{\mathrm{all}}} - \mathbb{E}_{X}\bracket*{4\eta(x)(1-\eta(x))}\\
&=0
\end{align*}
Thus, for $\frac{1}2< t\leq1$, we have
\begin{align*}
\inf_{h\in\sH_{\mathrm{all}}:h(x)<0}\sC_{\Phi_{\mathrm{quad}}}(h,x,t)
&=t+(1-t)\\
&=1\\
\inf_{x\in \sX}\inf_{h\in\sH_{\mathrm{all}}:h(x)<0}\Delta\sC_{\Phi_{\mathrm{quad}},\sH_{\mathrm{all}}}(h,x,t)
& =\inf_{x\in \sX}\paren*{\inf_{h\in\sH_{\mathrm{all}}:h(x)<0}\sC_{\Phi_{\mathrm{quad}}}(h,x,t)-\inf_{h\in\sH_{\mathrm{all}}}\sC_{\Phi_{\mathrm{quad}}}(h,x,t)}\\
&=\inf_{x\in \sX}\paren*{1-4t(1-t)}\\
&=1-4t(1-t)\\
&=\sT(2t-1)
\end{align*}
where $\sT$ is the increasing and convex function on $[0,1]$ defined by
\begin{align*}
\forall t\in[0,1], \quad
\sT(t)=t^2.
\end{align*}
By
Proposition~\ref{prop:prop-noise}, for $\epsilon= 0$, the modified $\sH_{\mathrm{all}}$-estimation error transformation of the quadratic loss under Massart's noise condition with $\beta$ is as follows:
\begin{align*}
\sT^M_{\Phi_{\mathrm{quad}}}(t)= 
\begin{cases}
2\beta \, t, & t\in \left[0, 2\beta\right], \\
t^2, & t\in \left[2\beta,1\right].
\end{cases}
\end{align*}
Therefore, $\sT^M_{\Phi_{\mathrm{quad}}}$ is convex, non-decreasing, invertible and satisfies that $\sT^M_{\Phi_{\mathrm{quad}}}(0)=0$. By
Proposition~\ref{prop:prop-noise}, we obtain the $\sH_{\mathrm{all}}$-consistency bound for the quadratic loss, valid for all $h \in \sH_{\mathrm{all}}$ such that $\sR_{\Phi_{\mathrm{quad}}}(h)- \sR_{\Phi_{\mathrm{quad}}}^*\paren*{\sH_{\mathrm{all}}}\leq \sT(2\beta)=4\beta^2$ and distributions $\sD$ satisfies Massart's noise condition with $\beta$:
\begin{align}
    \sR_{\ell_{0-1}}(h)- \sR_{\ell_{0-1}}^*\paren*{\sH_{\mathrm{all}}} \leq
    \frac{\sR_{\Phi_{\mathrm{quad}}}(h)- \sR_{\Phi_{\mathrm{quad}}}^*\paren*{\sH_{\mathrm{all}}}}{2\beta}
\end{align}

\subsection{Logistic Loss}
For the logistic loss $\Phi_{\mathrm{log}}(\alpha)\colon=\log_2(1+e^{-\alpha})$, for all $h\in \sH_{\mathrm{all}}$ and $x\in \sX$:
\begin{align*}
\sC_{\Phi_{\mathrm{log}}}(h,x,t)
& = t \Phi_{\mathrm{log}}(h(x))+(1-t)\Phi_{\mathrm{log}}(-h(x)),\\
& = t\log_2\paren*{1+e^{-h(x)}}+(1-t)\log_2\paren*{1+e^{h(x)}}.\\
\inf_{h\in\sH_{\mathrm{all}}}\sC_{\Phi_{\mathrm{log}}}(h,x,t)
&=-t\log_2(t)-(1-t)\log_2(1-t)\\
\sM_{\Phi_{\mathrm{log}}}\paren*{\sH_{\mathrm{all}}}
& = \sR_{\Phi_{\mathrm{log}}}^*\paren*{\sH_{\mathrm{all}}}-
\mathbb{E}_{X}\bracket*{\inf_{h\in\sH_{\mathrm{all}}}\sC_{\Phi_{\mathrm{log}}}(h,x,\eta(x))}\\
& = \sR_{\Phi_{\mathrm{log}}}^*\paren*{\sH_{\mathrm{all}}}- \mathbb{E}_{X}\bracket*{-\eta(x)\log_2(\eta(x))-(1-\eta(x))\log_2(1-\eta(x))}\\
& = 0
\end{align*}
Thus, for $\frac{1}2< t\leq1$, we have
\begin{align*}
\inf_{h\in\sH_{\mathrm{all}}:h(x)<0}\sC_{\Phi_{\mathrm{log}}}(h,x,t)
& = t\log_2\paren*{1+e^{-0}}+(1-t)\log_2\paren*{1+e^{0}} \\
& = 1, \\
\inf_{x\in \sX}\inf_{h\in\sH_{\mathrm{all}}:h(x)<0}\Delta\sC_{\Phi_{\mathrm{log}},\sH_{\mathrm{all}}}(h,x,t)
&=\inf_{x\in \sX}\paren*{\inf_{h\in\sH_{\mathrm{all}}:h(x)<0}\sC_{\Phi_{\mathrm{log}}}(h,x,t)-\inf_{h\in\sH_{\mathrm{all}}}\sC_{\Phi_{\mathrm{log}}}(h,x,t)}\\
&=\inf_{x\in \sX}\paren*{1+t\log_2(t)+(1-t)\log_2(1-t}\\
&=1+t\log_2(t)+(1-t)\log_2(1-t)\\
&=\sT(2t-1),
\end{align*}
where $\sT$ is the increasing and convex function on $[0,1]$ defined by
\begin{align*}
\forall t\in[0,1], \quad
\sT(t)=\frac{t+1}{2}\log_2(t+1)+\frac{1-t}{2}\log_2(1-t)
\end{align*}
By Proposition~\ref{prop:prop-noise}, for $\epsilon= 0$, the modified $\sH_{\mathrm{all}}$-estimation error transformation of the logistic loss under Massart's noise condition with $\beta$ is as follows:
\begin{align*}
\sT^M_{\Phi_{\mathrm{log}}}= 
\begin{cases}
\sT(t), & t\in \left[2\beta,1\right], \\
\frac{\sT(2\beta)}{2\beta}\, t, &  t\in \left[0,2\beta\right).
\end{cases}
\end{align*}
Therefore, $\sT^M_{\Phi_{\mathrm{log}}}$ is convex, non-decreasing, invertible and satisfies that $\sT^M_{\Phi_{\mathrm{log}}}(0)=0$. By  Proposition~\ref{prop:prop-noise}, we obtain the $\sH_{\mathrm{all}}$-consistency bound for the logistic loss, valid for all $h \in \sH_{\mathrm{all}}$ such that $\sR_{\Phi_{\mathrm{log}}}(h)- \sR_{\Phi_{\mathrm{log}}}^*\paren*{\sH_{\mathrm{all}}}\leq \sT(2\beta)=\frac{2\beta+1}{2}\log_2(2\beta+1)+\frac{1-2\beta}{2}\log_2(1-2\beta)$ and distributions $\sD$ satisfies Massart's noise condition with $\beta$:
\begin{align}
     \sR_{\ell_{0-1}}(h)-\sR_{\ell_{0-1}}^*\paren*{\sH_{\mathrm{all}}} \leq 
     \frac{2\beta\paren*{\sR_{\Phi_{\mathrm{log}}}(h)- \sR_{\Phi_{\mathrm{log}}}^*\paren*{\sH_{\mathrm{all}}}}}{\frac{2\beta+1}{2}\log_2(2\beta+1)+\frac{1-2\beta}{2}\log_2(1-2\beta)}
\end{align}

\subsection{Exponential Loss}
For the exponential loss $\Phi_{\mathrm{exp}}(\alpha)\colon=e^{-\alpha}$, for all $h\in \sH_{\mathrm{all}}$ and $x\in \sX$:
\begin{align*}
\sC_{\Phi_{\mathrm{exp}}}(h,x,t)
&=t \Phi_{\mathrm{exp}}(h(x))+(1-t)\Phi_{\mathrm{exp}}(-h(x))\\
&=t e^{-h(x)}+(1-t)e^{h(x)}.\\
\inf_{h\in\sH_{\mathrm{all}}}\sC_{\Phi_{\mathrm{exp}}}(h,x,t)
&=2\sqrt{t(1-t)}\\
\sM_{\Phi_{\mathrm{exp}}}\paren*{\sH_{\mathrm{all}}}
& = \sR_{\Phi_{\mathrm{exp}}}^*\paren*{\sH_{\mathrm{all}}}-
\mathbb{E}_{X}\bracket*{\inf_{h\in\sH_{\mathrm{all}}}\sC_{\Phi_{\mathrm{exp}}}(h,x,\eta(x))}\\
& = \sR_{\Phi_{\mathrm{exp}}}^*\paren*{\sH_{\mathrm{all}}}-
\mathbb{E}_{X}\bracket*{2\sqrt{\eta(x)(1-\eta(x))}}\\
& = 0.
\end{align*}
Thus, for $\frac{1}2< t\leq1$, we have
\begin{align*}
\inf_{h\in\sH_{\mathrm{all}}:h(x)<0}\sC_{\Phi_{\mathrm{exp}}}(h,x,t)
&=te^{-0}+(1-t)e^{0}\\
&=1.\\
\inf_{x\in \sX}\inf_{h\in\sH_{\mathrm{all}}:h(x)<0}\Delta\sC_{\Phi_{\mathrm{exp}},\sH_{\mathrm{all}}}(h,x)
&=\inf_{x\in \sX}\paren*{\inf_{h\in\sH_{\mathrm{all}}:h(x)<0}\sC_{\Phi_{\mathrm{exp}}}(h,x)-\inf_{h\in\sH_{\mathrm{all}}}\sC_{\Phi_{\mathrm{exp}}}(h,x)}\\
&=\inf_{x\in \sX}\paren*{1-2\sqrt{t(1-t)}}\\
&=1-2\sqrt{t(1-t)}\\
&=\sT(2t-1),
\end{align*}
where $\sT$ is the increasing and convex function on $[0,1]$ defined by
\begin{align*}
\forall t\in[0,1], \quad 
\sT(t)=1-\sqrt{1-t^2}.
\end{align*}
By Proposition~\ref{prop:prop-noise}, for $\epsilon= 0$, the modified $\sH_{\mathrm{all}}$-estimation error transformation of the exponential loss under Massart's noise condition with $\beta$ is as follows:
\begin{align*}
\sT^M_{\Phi_{\mathrm{exp}}}= 
\begin{cases}
\sT(t), & t\in \left[2\beta,1\right], \\
\frac{\sT(2\beta)}{2\beta}\, t, &  t\in \left[0,2\beta\right).
\end{cases}
\end{align*}
Therefore, $\sT^M_{\Phi_{\mathrm{exp}}}$ is convex, non-decreasing, invertible and satisfies that $\sT^M_{\Phi_{\mathrm{exp}}}(0)=0$. By Proposition~\ref{prop:prop-noise}, we obtain the $\sH_{\mathrm{all}}$-consistency bound for the exponential loss, valid for all $h \in \sH_{\mathrm{all}}$ such that $\sR_{\Phi_{\mathrm{exp}}}(h)- \sR_{\Phi_{\mathrm{exp}}}^*\paren*{\sH_{\mathrm{all}}}\leq \sT(2\beta)=1-\sqrt{1-4\beta^2}$ and distributions $\sD$ satisfies Massart's noise condition with $\beta$:
\begin{align}
     &\sR_{\ell_{0-1}}(h)-\sR_{\ell_{0-1}}^*\paren*{\sH_{\mathrm{all}}}\leq 
     \frac{2\beta\paren*{\sR_{\Phi_{\mathrm{exp}}}(h)- \sR_{\Phi_{\mathrm{exp}}}^*\paren*{\sH_{\mathrm{all}}}}}{1-\sqrt{1-4\beta^2}}
\end{align}

\section{Derivation of Adversarial
\texorpdfstring{$\sH$}{H}-Consistency Bounds under Massart's Noise Condition}
\label{app:derivation-adv_noise}
As with the non-adversarial scenario in Section~\ref{sec:noise-non-adv},  we
introduce a modified adversarial $\sH$-estimation error transformation. We assume that $\epsilon=0$ throughout this section.
\begin{restatable}{proposition}{TransNoiseAdv}
\label{prop-adv-noise}
Under Massart's noise condition with $\beta$, the modified adversarial $\sH$-estimation error transformation of $\wt{\Phi}$ for $\epsilon=0$ is defined on $t\in \left[0,1\right]$ by 
\begin{align*}
\sT^M_{\wt{\Phi}}\paren*{t}=
\min\curl*{\sT^M_1(t), \sT^M_2(t)},
\end{align*}
where 
\begin{align*}
    &\sT^M_1(t):=\h{\sT}_1(t)\mathds{1}_{t\in \left[\frac{1}{2}+\beta,1\right]}+ 2/(1+2\beta)\,\h{\sT}_1\paren*{\frac{1}{2}+\beta}\, t\mathds{1}_{t\in \left[0,\frac{1}{2}+\beta\right)},\\
    &\sT^M_2(t):=\h{\sT}_2(t)\mathds{1}_{t\in \left[2\beta,1\right]}+ \frac{\h{\sT}_2(2\beta)}{2\beta}\,t\mathds{1}_{t\in \left[0,2\beta\right)},
\end{align*}
with $\h{\sT}_1(t)$ and $\h{\sT}_2(t)$ defined in Definition~\ref{def:trans-adv}. Suppose that $\sH$ is symmetric and $\wt{\sT}^M_{\wt{\Phi}}$ is any lower bound of $\sT^M_{\wt{\Phi}}$ such that $\wt{\sT}^M_{\wt{\Phi}}\leq \sT^M_{\wt{\Phi}}$. If
$\wt{\sT}^M_{\wt{\Phi}}$ is convex with
$\wt{\sT}^M_{\wt{\Phi}}(0)=0$, then, for any hypothesis $h\in\sH$ and any distribution under Massart's noise condition with $\beta$,
\begin{equation*}
     \wt{\sT}^M_{\wt{\Phi}}\paren*{\sR_{\ell_{\gamma}}(h) - \sR_{\ell_{\gamma}}^*(\sH) + \sM_{\ell_{\gamma}}(\sH)}
     \leq  \sR_{\wt{\Phi}}(h)
     - \sR_{\wt{\Phi}}^*(\sH) + \sM_{\wt{\Phi}}(\sH).
\end{equation*}
\end{restatable}
\begin{proof}
Note the condition~\eqref{eq:condition_Psi_general_adv} in Theorem~\ref{Thm:excess_bounds_Psi_01_general_adv} is symmetric about $\Delta \eta(x)=0$. Thus, condition~\eqref{eq:condition_Psi_general_adv} uniformly holds for all distributions under Massart's noise condition with $\beta$ is equivalent to the following holds for any $t\in\left[1/2+\beta,1\right]\colon$
\begin{equation}
\label{eq:condition_Psi_general_adv-massarts}
\begin{aligned}
&\Psi\paren*{\tri*{t}_{\e}}  \leq \inf_{x\in\sX,h\in \ov \sH_\gamma(x)\subsetneqq \sH}\Delta\sC_{\wt{\Phi},\sH}(h,x,t),\\
&\Psi\paren*{\tri*{2t-1}_{\e}} \leq \inf_{x\in \sX,h\in\sH\colon  \ov h_\gamma(x)< 0}\Delta\sC_{\wt{\Phi},\sH}(h,x,t),
\end{aligned}
\end{equation}
It is clear that any lower bound $\wt{\sT}^M_{\wt{\Phi}}$ of the modified adversarial $\sH$-estimation error transformation verified condition~\eqref{eq:condition_Psi_general_adv-massarts}. Then by Theorem~\ref{Thm:excess_bounds_Psi_01_general_adv}, the proof is completed.
\end{proof}
\subsection{Linear Hypotheses}
\label{app:derivation-lin-adv_noise}
By the definition of $\sH_{\mathrm{lin}}$, for any $x \in \sX$, 
\begin{align*}
&\uv h_\gamma(x)
  =w \cdot x-\gamma \|w\|_q+b \\
& \in 
  \begin{cases}
\bracket*{-W \norm*{x}_p-\gamma W-B, W\norm*{x}_p - \gamma W+ B}
&  \norm*{x}_p \geq \gamma\\
\bracket*{-W \norm*{x}_p-\gamma W-B, B}
&  \norm*{x}_p < \gamma
\end{cases},\\
&\ov h_\gamma(x)=w \cdot x+\gamma \|w\|_q+b \\
& \in 
\begin{cases}
\bracket*{-W \norm*{x}_p+\gamma W-B, W\norm*{x}_p + \gamma W+ B}
&  \norm*{x}_p \geq \gamma\\
\bracket*{-B, W\norm*{x}_p + \gamma W+ B}
&  \norm*{x}_p < \gamma
\end{cases}.
\end{align*}
Note $\sH_{\mathrm{lin}}$ is symmetric. For any $x\in \sX$, there exist $w=0$ and any $0<b\leq B$ such that $w \cdot x-\gamma \|w\|_q+b>0$. Thus by Lemma~\ref{lemma:explicit_assumption_01_adv}, for any $x\in \sX$, $\sC^*_{\ell_{\gamma}}\paren*{\sH_{\mathrm{lin}}, x} =\min\curl*{\eta(x), 1 - \eta(x)}$. The $\paren*{\ell_{\gamma},\sH_{\mathrm{lin}}}$-minimizability gap can be expressed as follows:
\begin{align*}
\sM_{\ell_{\gamma}}\paren*{\sH_{\mathrm{lin}}}
& = \sR_{\ell_{\gamma}}^*\paren*{\sH_{\mathrm{lin}}}-\mathbb{E}_{X}\bracket*{\min\curl*{\eta(x),1-\eta(x)}}.
\end{align*}

\subsubsection{Supremum-Based Hinge Loss}

For the supremum-based hinge loss 
$
\wt{\Phi}_{\mathrm{hinge}}\colon=\sup_{x'\colon \|x-x'\|_p\leq \gamma}\Phi_{\mathrm{hinge}}(y h(x'))$, with $ \Phi_{\mathrm{hinge}}(\alpha)=\max\curl*{0,1 - \alpha},  
$
for all $h\in \sH_{\mathrm{lin}}$ and $x\in \sX$:
\begin{align*}
&\sC_{\wt{\Phi}_{\mathrm{hinge}}}(h,x,t)\\ 
&=t \wt{\Phi}_{\mathrm{hinge}}(h(x))+(1-t)\wt{\Phi}_{\mathrm{hinge}}(-h(x))\\
& = t\Phi_{\mathrm{hinge}}\paren*{\uv h_\gamma(x)}+(1-t)\Phi_{\mathrm{hinge}}\paren*{-\ov h_\gamma(x)}\\
& =t\max\curl*{0,1-\uv h_\gamma(x)} +(1-t)\max\curl*{0,1+\ov h_\gamma(x)}\\
& \geq \bracket*{ t\max\curl*{0,1-\ov h_\gamma(x)} +(1-t)\max\curl*{0,1+\ov h_\gamma(x)} } \\ & \qquad \wedge \bracket*{ t\max\curl*{0,1-\uv h_\gamma(x)} +(1-t)\max\curl*{0,1+\uv h_\gamma(x)}} \\
&\inf_{h\in\sH_{\mathrm{lin}}}\sC_{\wt{\Phi}_{\mathrm{hinge}}}(h,x,t)\\ &\geq \inf_{h\in\sH_{\mathrm{lin}}}\bracket*{ t\max\curl*{0,1-\ov h_\gamma(x)} +(1-t)\max\curl*{0,1+\ov h_\gamma(x)} } \\ & \qquad \wedge \inf_{h\in\sH_{\mathrm{lin}}}\bracket*{ t\max\curl*{0,1-\uv h_\gamma(x)} +(1-t)\max\curl*{0,1+\uv h_\gamma(x)}} \\
& = 1-\abs*{2t-1}\min\curl*{W\max\curl*{\norm*{x}_p,\gamma}-\gamma W+B,1}\\
&\inf_{h\in\sH_{\mathrm{lin}}}\sC_{\wt{\Phi}_{\mathrm{hinge}}}(h,x,t)\\
&=\inf_{h\in\sH_{\mathrm{lin}}}\bracket*{t\max\curl*{0,1-\uv h_\gamma(x)} +(1-t)\max\curl*{0,1+\ov h_\gamma(x)}}\\
& = \inf_{h\in\sH_{\mathrm{lin}}}
\bracket*{t\max\curl*{0,1-w \cdot x+\gamma \|w\|_q-b}+(1-t)\max\curl*{0,1+w \cdot x+\gamma \|w\|_q+b}}\\
& \leq \inf_{b\in [-B,B]}\bracket*{t\max\curl*{0,1-b}+(1-t)\max\curl*{0,1+b}} = 1-\abs*{2t-1}\min\curl*{B,1}\\
& \sM_{\wt{\Phi}_{\mathrm{hinge}}}\paren*{\sH_{\mathrm{lin}}} \\
& = \sR_{\wt{\Phi}_{\mathrm{hinge}}}^*\paren*{\sH_{\mathrm{lin}}} - \mathbb{E}\bracket*{\inf_{h\in\sH_{\mathrm{lin}}}\sC_{\wt{\Phi}_{\mathrm{hinge}}}(h,x,\eta(x))}\\
& \leq \sR_{\wt{\Phi}_{\mathrm{hinge}}}^*\paren*{\sH_{\mathrm{lin}}} - \mathbb{E}\bracket*{1-\abs*{2\eta(x)-1}\min\curl*{W\max\curl*{\norm*{x}_p,\gamma}-\gamma W+B,1}}
\end{align*}
Thus, for $\frac{1}2< t\leq1$, we have
\begin{align*}
\inf_{h\in\sH_{\mathrm{lin}}:\uv h_\gamma(x)\leq 0 \leq \ov h_\gamma(x)}\sC_{\wt{\Phi}_{\mathrm{hinge}}}(h,x,t)
& = t+(1-t)\\
& =1\\
\inf_{x\in \sX} \inf_{h\in\sH_{\mathrm{lin}}\colon \uv h_\gamma(x)\leq 0 \leq \ov h_\gamma(x)}\Delta\sC_{\wt{\Phi}_{\mathrm{hinge}},\sH_{\mathrm{lin}}}(h,x,t)
& = \inf_{x\in \sX} \curl*{1-\inf_{h\in\sH_{\mathrm{lin}}}\sC_{\wt{\Phi}_{\mathrm{hinge}}}(h,x,t)}\\
&\geq\inf_{x\in \sX}\paren*{2t-1}\min\curl*{B,1}\\
&=\paren*{2t-1}\min\curl*{B,1}\\
&=\sT_1(t),
\end{align*}
where $\sT_1$ is the increasing and convex function on $\bracket*{0,1}$ defined by\\
$
\sT_1(t) =
\begin{cases}
\min\curl*{B,1} \, (2t-1), & t \in \bracket*{1/2+\beta,1},\\ \min\curl*{B,1}\frac{4\beta}{1+2\beta}\, t, & t\in \left[0,1/2+\beta\right).
\end{cases}
$
\begin{align*}
& \inf_{h\in\sH_{\mathrm{lin}}:\ov h_\gamma(x)<0}\sC_{\wt{\Phi}_{\mathrm{hinge}}}(h,x,t)\\
& \geq \inf_{h\in\sH_{\mathrm{lin}}:\ov h_\gamma(x)<0}\, \bracket*{t\max\curl*{0,1-\ov h_\gamma(x)} +(1-t)\max\curl*{0,1+\ov h_\gamma(x)}}\\
& = t\max\curl*{0,1-0}+(1-t)\max\curl*{0,1+0} = 1\\
& \inf_{x\in \sX} \inf_{h\in\sH_{\mathrm{lin}}\colon \ov h_\gamma(x)< 0}\Delta\sC_{\wt{\Phi}_{\mathrm{hinge}},\sH_{\mathrm{lin}}}(h,x,t)\\
& = \inf_{x\in \sX} \curl*{\inf_{h\in\sH_{\mathrm{lin}}:\ov h_\gamma(x)< 0}\sC_{\wt{\Phi}_{\mathrm{hinge}}}(h,x,t)-\inf_{h\in\sH_{\mathrm{lin}}}\sC_{\wt{\Phi}_{\mathrm{hinge}}}(h,x,t)}\\
&\geq\inf_{x\in \sX}\paren*{2t-1}\min\curl*{B,1}\\
&=\paren*{2t-1}\min\curl*{B,1}\\
&=\sT_2(2t - 1),
\end{align*}
where $\sT_2$ is the increasing and convex function on $\bracket*{0,1}$ defined by
\begin{align*}
\forall t \in [0,1], \quad \sT_2(t) = \min\curl*{B,1} \, t.
\end{align*}
By Proposition~\ref{prop-adv-noise}, for $\epsilon= 0$, the modified adversarial $\sH_{\mathrm{lin}}$-estimation error transformation of the supremum-based hinge loss under Massart's noise condition with $\beta$ is lower bounded as follows:
\begin{align*}
\sT^M_{\wt{\Phi}_{\mathrm{hinge}}}\geq\wt{\sT}^M_{\wt{\Phi}_{\mathrm{hinge}}}:=\min\curl*{\sT_1,\sT_2}=\begin{cases}
\min\curl*{B,1} \, (2t-1), & t \in \bracket*{1/2+\beta,1},\\ \min\curl*{B,1}\frac{4\beta}{1+2\beta}\, t, & t\in \left[0,1/2+\beta\right).
\end{cases}
\end{align*}
Note $\wt{\sT}^M_{\wt{\Phi}_{\mathrm{hinge}}}$ is convex, non-decreasing, invertible and satisfies that $\wt{\sT}^M_{\wt{\Phi}_{\mathrm{hinge}}}(0)=0$. By Proposition~\ref{prop-adv-noise}, using the fact that
$\wt{\sT}^M_{\wt{\Phi}_{\mathrm{hinge}}}\geq \min\curl*{B,1}\frac{4\beta}{1+2\beta}\, t$ yields the adversarial $\sH_{\mathrm{lin}}$-consistency bound for the supremum-based hinge loss, valid for all $h \in \sH_{\mathrm{lin}}$ and distributions $\sD$ satisfies Massart's noise condition with $\beta$:
\begin{align}
\label{eq:hinge-lin-est-adv}
     & \sR_{\ell_{\gamma}}(h)- \sR_{\ell_{\gamma}}^*\paren*{\sH_{\mathrm{lin}}} \leq 
     \frac{1+2\beta}{4\beta}\frac{\sR_{\wt{\Phi}_{\mathrm{hinge}}}(h)-\sR_{\wt{\Phi}_{\mathrm{hinge}}}^*\paren*{\sH_{\mathrm{lin}}}+\sM_{\wt{\Phi}_{\mathrm{hinge}}}\paren*{\sH_{\mathrm{lin}}}}{\min\curl*{B,1}}-\sM_{\ell_{\gamma}, \sH_{\mathrm{lin}}}
\end{align}
Since
\begin{align*}
\sM_{\ell_{\gamma}}\paren*{\sH_{\mathrm{lin}}}
& = \sR_{\ell_{\gamma}}^*\paren*{\sH_{\mathrm{lin}}}-\mathbb{E}_{X}\bracket*{\min\curl*{\eta(x),1-\eta(x)}},\\
\sM_{\wt{\Phi}_{\mathrm{hinge}}}\paren*{\sH_{\mathrm{lin}}} 
& \leq \sR_{\wt{\Phi}_{\mathrm{hinge}}}^*\paren*{\sH_{\mathrm{lin}}} - \mathbb{E}\bracket*{1-\abs*{2\eta(x)-1}\min\curl*{W\max\curl*{\norm*{x}_p,\gamma}-\gamma W+B,1}},
\end{align*}
the inequality can be relaxed as follows:
\begin{align*}
     \sR_{\ell_{\gamma}}(h) & \leq \frac{1+2\beta}{4\beta}\frac{\sR_{\wt{\Phi}_{\mathrm{hinge}}}(h)}{\min\curl*{B,1}}+\mathbb{E}_{X}\bracket*{\min\curl*{\eta(x),1-\eta(x)}}\\
     &\qquad -\frac{1+2\beta}{4\beta}\frac{\mathbb{E}\bracket*{1-\abs*{2\eta(x)-1}\min\curl*{W\max\curl*{\norm*{x}_p,\gamma}-\gamma W+B,1}}}{\min\curl*{B,1}}
\end{align*}
Note that: $\min\curl*{W\max\curl*{\norm*{x}_p,\gamma}-\gamma W+B,1}\leq 1$ and $1-\abs*{1-2\eta(x)}=2\min\curl*{\eta(x),1-\eta(x)}$. Thus the inequality can be further relaxed as follows:
\begin{align*}
     \sR_{\ell_{\gamma}}(h)
     \leq \frac{1+2\beta}{4\beta}\frac{\sR_{\wt{\Phi}_{\mathrm{hinge}}}(h)}{ \min\curl*{B,1}}-\paren*{\frac{1+2\beta}{2\beta \min\curl*{B,1}}-1}\,\mathbb{E}_{X}\bracket*{\min\curl*{\eta(x),1-\eta(x)}}.
\end{align*}
When $B\geq 1$, it can be equivalently written as follows:
\begin{align}
\label{eq:hinge-lin-est-adv-2}
     \sR_{\ell_{\gamma}}(h)
     \leq \frac{1+2\beta}{4\beta}\,\sR_{\wt{\Phi}_{\mathrm{hinge}}}(h)-\frac{1}{2\beta }\,\mathbb{E}_{X}\bracket*{\min\curl*{\eta(x),1-\eta(x)}}.
\end{align}

\subsubsection{Supremum-Based Sigmoid Loss}
For the supremum-based sigmoid loss 
\begin{align*}
\wt{\Phi}_{\mathrm{sig}}\colon=\sup_{x'\colon \|x-x'\|_p\leq \gamma}\Phi_{\mathrm{sig}}(y h(x')),  \quad \text{where } \Phi_{\mathrm{sig}}(\alpha)=1-\tanh(k\alpha),~k>0,   
\end{align*}
for all $h\in \sH_{\mathrm{lin}}$ and $x\in \sX$:
\begin{align*}
\sC_{\wt{\Phi}_{\mathrm{sig}}}(h,x,t) 
&=t \wt{\Phi}_{\mathrm{sig}}(h(x))+(1-t)\wt{\Phi}_{\mathrm{sig}}(-h(x))\\
& = t\Phi_{\mathrm{sig}}\paren*{\uv h_\gamma(x)}+(1-t)\Phi_{\mathrm{sig}}\paren*{-\ov h_\gamma(x)}\\
& =t\paren*{1-\tanh(k \uv h_\gamma(x))} +(1-t)\paren*{1+\tanh(k \ov h_\gamma(x))}\\
& \geq \max\curl*{1+\paren{1-2t}\tanh(k\ov h_\gamma(x)), 1+\paren{1-2t}\tanh(k\uv h_\gamma(x))}\\
\inf_{h\in\sH_{\mathrm{lin}}}\sC_{\wt{\Phi}_{\mathrm{sig}}}(h,x,t)
& \geq \max\curl*{\inf_{h\in\sH_{\mathrm{lin}}}\bracket*{1+\paren{1-2t}\tanh(k\ov h_\gamma(x))}, \inf_{h\in\sH_{\mathrm{lin}}}\bracket*{1+\paren{1-2t}\tanh(k\uv h_\gamma(x))}}\\
& = 1 - \abs*{1-2t}\tanh\paren*{k\paren*{W\max\curl*{\norm*{x}_p,\gamma}-\gamma W+B}}\\
\inf_{h\in\sH_{\mathrm{lin}}}\sC_{\wt{\Phi}_{\mathrm{sig}}}(h,x,t)
&= \inf_{h\in\sH_{\mathrm{lin}}}\bracket[\big]{t\paren*{1-\tanh(k \paren*{w \cdot x-\gamma \|w\|_q+b})}\\ &\qquad +(1-t)\paren*{1+\tanh(k\paren*{ w \cdot x+\gamma \|w\|_q+b})}}\\
&\leq \inf_{b\in [-B,B]}\bracket*{t\paren*{1-\tanh(kb)} +(1-t)\paren*{1+\tanh(kb)}}\\
& = \max\curl*{t,1-t}\paren*{1-\tanh\paren*{k B }}+\min\curl*{t,1-t}\paren*{1+\tanh\paren*{k B}}\\
& = 1 - \abs*{1-2t}\tanh\paren*{k B}\\
\sM_{\wt{\Phi}_{\mathrm{sig}}}\paren*{\sH_{\mathrm{lin}}} 
& = \sR_{\wt{\Phi}_{\mathrm{sig}}}^*\paren*{\sH_{\mathrm{lin}}} - \mathbb{E}\bracket*{\inf_{h\in\sH_{\mathrm{lin}}}\sC_{\wt{\Phi}_{\mathrm{sig}}}(h,x,\eta(x))}\\
& \leq \sR_{\wt{\Phi}_{\mathrm{sig}}}^*\paren*{\sH_{\mathrm{lin}}} - \mathbb{E}\bracket*{1 - \abs*{1-2\eta(x)}\tanh\paren*{k\paren*{W\max\curl*{\norm*{x}_p,\gamma}-\gamma W+B}}}
\end{align*}
Thus, for $\frac{1}2< t\leq1$, we have
\begin{align*}
\inf_{h\in\sH_{\mathrm{lin}}:\uv h_\gamma(x)\leq 0 \leq \ov h_\gamma(x)}\sC_{\wt{\Phi}_{\mathrm{sig}}}(h,x,t) 
& = t+(1-t)\\
& =1\\
\inf_{x\in \sX} \inf_{h\in\sH_{\mathrm{lin}}\colon \uv h_\gamma(x)\leq 0 \leq \ov h_\gamma(x)}\Delta\sC_{\wt{\Phi}_{\mathrm{sig}},\sH_{\mathrm{lin}}}(h,x,t)
& = \inf_{x\in \sX} \curl*{1-\inf_{h\in\sH_{\mathrm{lin}}}\sC_{\wt{\Phi}_{\mathrm{sig}}}(h,x,t)}\\
&\geq \inf_{x\in \sX} (2t-1)\tanh\paren*{k B}\\
&=(2t-1)\tanh\paren*{k B}\\
&=\sT_1(t),
\end{align*}
where $\sT_1$ is the increasing and convex function on $\bracket*{0,1}$ defined by
\begin{align*}
 \sT_1(t) = \begin{cases}
\tanh\paren*{k B} \frac{4\beta}{1+2\beta} \, t, & t\in\bracket*{0,1/2+\beta},\\
\tanh\paren*{k B} \, (2t-1),  & t \in \bracket*{1/2+\beta,1}.
\end{cases}    
\end{align*}
\begin{align*}
\inf_{h\in\sH_{\mathrm{lin}}:\ov h_\gamma(x)<0}\sC_{\wt{\Phi}_{\mathrm{sig}}}(h,x,t)
& \geq \inf_{h\in\sH_{\mathrm{lin}}:\ov h_\gamma(x)<0} 
\bracket*{1+\paren{1-2t}\tanh(k\ov h_\gamma(x))}\\
& = 1\\
\inf_{x\in \sX} \inf_{h\in\sH_{\mathrm{lin}}\colon \ov h_\gamma(x)< 0}\Delta\sC_{\wt{\Phi}_{\mathrm{sig}},\sH_{\mathrm{lin}}}(h,x,t)
& = \inf_{x\in \sX} \curl*{\inf_{h\in\sH_{\mathrm{lin}}:\ov h_\gamma(x)< 0}\sC_{\wt{\Phi}_{\mathrm{sig}}}(h,x,t)-\inf_{h\in\sH_{\mathrm{lin}}}\sC_{\wt{\Phi}_{\mathrm{sig}}}(h,x,t)}\\
&\geq \inf_{x\in \sX}(2t-1)\tanh\paren*{k B}\\
&=(2t-1)\tanh\paren*{k B}\\
&=\sT_2(2t-1),
\end{align*}
where $\sT_2$ is the increasing and convex function on $\bracket*{2\beta,1}$ defined by
\begin{align*}
\forall t \in [0,1], \quad \sT_2(t) = \tanh\paren*{k B} \, t \,;
\end{align*}
By Proposition~\ref{prop-adv-noise}, for $\epsilon= 0$, the modified adversarial $\sH_{\mathrm{lin}}$-estimation error transformation of the supremum-based sigmoid loss under Massart's noise condition with $\beta$ is lower bounded as follows:
\begin{align*}
\sT^M_{\wt{\Phi}_{\mathrm{sig}}}\geq \wt{\sT}^M_{\wt{\Phi}_{\mathrm{sig}}}=\min\curl*{\sT_1,\sT_2}=\begin{cases}
\tanh\paren*{k B} \frac{4\beta}{1+2\beta} \, t, & t\in\bracket*{0,1/2+\beta},\\
\tanh\paren*{k B} \, (2t-1),  & t \in \bracket*{1/2+\beta,1}.
\end{cases}    
\end{align*}
Note $\wt{\sT}^M_{\wt{\Phi}_{\mathrm{sig}}}$ is convex, non-decreasing, invertible and satisfies that $\wt{\sT}^M_{\wt{\Phi}_{\mathrm{sig}}}(0)=0$. By Proposition~\ref{prop-adv-noise}, using the fact that
$\wt{\sT}^M_{\wt{\Phi}_{\mathrm{sig}}}\geq \tanh\paren*{k B} \frac{4\beta}{1+2\beta} \, t$ yields the adversarial $\sH_{\mathrm{lin}}$-consistency bound for the supremum-based sigmoid loss, valid for all $h \in \sH_{\mathrm{lin}}$ and distributions $\sD$ satisfies Massart's noise condition with $\beta$:
\begin{align}
\label{eq:sig-lin-est-adv}
     \sR_{\ell_{\gamma}}(h)- \sR_{\ell_{\gamma}}^*\paren*{\sH_{\mathrm{lin}}} 
     \leq 
     \frac{1+2\beta}{4\beta}\frac{\sR_{\wt{\Phi}_{\mathrm{sig}}}(h)-\sR_{\wt{\Phi}_{\mathrm{sig}}}^*\paren*{\sH_{\mathrm{lin}}}+\sM_{\wt{\Phi}_{\mathrm{sig}}}\paren*{\sH_{\mathrm{lin}}}}{\tanh\paren*{k B}}-\sM_{\ell_{\gamma}, \sH_{\mathrm{lin}}}.
\end{align}
Since
\begin{align*}
\sM_{\ell_{\gamma}}\paren*{\sH_{\mathrm{lin}}}
& = \sR_{\ell_{\gamma}}^*\paren*{\sH_{\mathrm{lin}}}-\mathbb{E}_{X}\bracket*{\min\curl*{\eta(x),1-\eta(x)}},\\
\sM_{\wt{\Phi}_{\mathrm{sig}}}\paren*{\sH_{\mathrm{lin}}} 
& \leq \sR_{\wt{\Phi}_{\mathrm{sig}}}^*\paren*{\sH_{\mathrm{lin}}} - \mathbb{E}\bracket*{1 - \abs*{1-2\eta(x)}\tanh\paren*{k\paren*{W\max\curl*{\norm*{x}_p,\gamma}-\gamma W+B}}},
\end{align*}
the inequality can be relaxed as follows:
\begin{align*}
     \sR_{\ell_{\gamma}}(h) 
     & \leq 
     \frac{1+2\beta}{4\beta}\frac{\sR_{\wt{\Phi}_{\mathrm{sig}}}(h)}{\tanh\paren*{k B}}+\mathbb{E}_{X}\bracket*{\min\curl*{\eta(x),1-\eta(x)}}\\
     & \qquad -\frac{1+2\beta}{4\beta}\frac{\mathbb{E}\bracket*{1 - \abs*{1-2\eta(x)}\tanh\paren*{k\paren*{W\max\curl*{\norm*{x}_p,\gamma}-\gamma W+B}}}}{\tanh\paren*{k B}}.
\end{align*}
Note that: $\tanh\paren*{k\paren*{W\max\curl*{\norm*{x}_p,\gamma}-\gamma W+B}}\leq 1$ and $1-\abs*{1-2\eta(x)}=2\min\curl*{\eta(x),1-\eta(x)}$. Thus the inequality can be further relaxed as follows:
\begin{align*}
    \sR_{\ell_{\gamma}}(h)
     \leq \frac{1+2\beta}{4\beta}\frac{\sR_{\wt{\Phi}_{\mathrm{sig}}}(h)}{\tanh\paren*{k B}}-\paren*{\frac{1+2\beta}{2\beta \tanh(kB)}-1}\,\mathbb{E}_{X}\bracket*{\min\curl*{\eta(x),1-\eta(x)}}.
\end{align*}
When $B=+ \infty$, it can be equivalently written as follows:
\begin{align}
\label{eq:sig-lin-est-adv-2}
    \sR_{\ell_{\gamma}}(h)
     \leq \frac{1+2\beta}{4\beta}\,\sR_{\wt{\Phi}_{\mathrm{sig}}}(h)-\frac{1}{2\beta}\,\mathbb{E}_{X}\bracket*{\min\curl*{\eta(x),1-\eta(x)}}.
\end{align}

\subsection{One-Hidden-Layer ReLU Neural Networks}
\label{app:derivation-NN-adv_noise}

By the definition of $\sH_{\mathrm{NN}}$, for any $x \in \sX$, 
\begin{align*}
&\uv h_\gamma(x)=\inf_{x'\colon \|x-x'\|_p\leq \gamma}\sum_{j = 1}^n u_j(w_j \cdot x'+b)_{+}\\
&\ov h_\gamma(x)=\sup_{x'\colon \|x-x'\|_p\leq \gamma}\sum_{j = 1}^n u_j(w_j \cdot x'+b)_{+} 
\end{align*}
Note $\sH_{\mathrm{NN}}$ is symmetric. For any $x\in \sX$, there exist $u=\paren*{\frac{1}{\Lambda},\ldots,\frac{1}{\Lambda}}$, $w=0$ and any $0<b\leq B$ satisfy that $\uv h_\gamma(x)>0$. Thus by Lemma~\ref{lemma:explicit_assumption_01_adv}, for any $x\in \sX$, $\sC^*_{\ell_{\gamma}}\paren*{\sH_{\mathrm{NN}}, x} =\min\curl*{\eta(x), 1 - \eta(x)}$. The $\paren*{\ell_{\gamma},\sH_{\mathrm{NN}}}$-minimizability gap can be expressed as follows:
\begin{align*}
\sM_{\ell_{\gamma}}\paren*{\sH_{\mathrm{NN}}}
& = \sR_{\ell_{\gamma}}^*\paren*{\sH_{\mathrm{NN}}}-\mathbb{E}_{X}\bracket*{\min\curl*{\eta(x),1-\eta(x)}}.
\end{align*}

\subsubsection{Supremum-Based Hinge Loss}
For the supremum-based hinge loss 
$
\wt{\Phi}_{\mathrm{hinge}}\colon=\sup_{x'\colon \|x-x'\|_p\leq \gamma}\Phi_{\mathrm{hinge}}(y h(x'))$ with $\Phi_{\mathrm{hinge}}(\alpha)=\max\curl*{0,1 - \alpha},  
$
for all $h\in \sH_{\mathrm{NN}}$ and $x\in \sX$:
\begin{align*}
& \sC_{\wt{\Phi}_{\mathrm{hinge}}}(h,x,t) \\
&=t \wt{\Phi}_{\mathrm{hinge}}(h(x))+(1-t)\wt{\Phi}_{\mathrm{hinge}}(-h(x))\\
& = t\Phi_{\mathrm{hinge}}\paren*{\uv h_\gamma(x)}+(1-t)\Phi_{\mathrm{hinge}}\paren*{-\ov h_\gamma(x)}\\
& =t\max\curl*{0,1-\uv h_\gamma(x)} +(1-t)\max\curl*{0,1+\ov h_\gamma(x)}\\
& \geq \bracket*{ t\max\curl*{0,1-\ov h_\gamma(x)} +(1-t)\max\curl*{0,1+\ov h_\gamma(x)} } \\ & \qquad \wedge \bracket*{ t\max\curl*{0,1-\uv h_\gamma(x)} +(1-t)\max\curl*{0,1+\uv h_\gamma(x)}} \\
& \inf_{h\in\sH_{\mathrm{NN}}}\sC_{\wt{\Phi}_{\mathrm{hinge}}}(h,x,t)\\
& \geq \inf_{h\in\sH_{\mathrm{NN}}}\bracket*{ t\max\curl*{0,1-\ov h_\gamma(x)} +(1-t)\max\curl*{0,1+\ov h_\gamma(x)} } \\ & \qquad \wedge \inf_{h\in\sH_{\mathrm{NN}}}\bracket*{ t\max\curl*{0,1-\uv h_\gamma(x)} +(1-t)\max\curl*{0,1+\uv h_\gamma(x)}} \\
& = 1-\abs*{2t-1}\min\curl*{\sup_{h\in\sH_{\mathrm{NN}}}\uv h_\gamma(x),1}\\
& \inf_{h\in\sH_{\mathrm{NN}}}\sC_{\wt{\Phi}_{\mathrm{hinge}}}(h,x,t)\\
& \leq \inf_{h\in\sH_{\mathrm{NN}:w=0}}\sC_{\wt{\Phi}_{\mathrm{hinge}}}(h,x,t)\\
& = 1-\abs*{2t-1}\min\curl*{\Lambda B,1}\\
& \sM_{\wt{\Phi}_{\mathrm{hinge}}}\paren*{\sH_{\mathrm{NN}}} \\
& = \sR_{\wt{\Phi}_{\mathrm{hinge}}}^*\paren*{\sH_{\mathrm{NN}}} - \mathbb{E}\bracket*{\inf_{h\in\sH_{\mathrm{NN}}}\sC_{\wt{\Phi}_{\mathrm{hinge}}}(h,x,\eta(x))}\\
& \leq \sR_{\wt{\Phi}_{\mathrm{hinge}}}^*\paren*{\sH_{\mathrm{NN}}} - \mathbb{E}\bracket*{1-\abs*{2\eta(x)-1}\min\curl*{\sup_{h\in\sH_{\mathrm{NN}}}\uv h_\gamma(x),1}}
\end{align*}
Thus, for $\frac{1}2< t\leq1$, we have
\begin{align*}
\inf_{h\in\sH_{\mathrm{NN}}:\uv h_\gamma(x)\leq 0 \leq \ov h_\gamma(x)}\sC_{\wt{\Phi}_{\mathrm{hinge}}}(h,x,t)
& = t+(1-t)\\
& =1\\
\inf_{x\in \sX} \inf_{h\in\sH_{\mathrm{NN}}\colon \uv h_\gamma(x)\leq 0 \leq \ov h_\gamma(x)}\Delta\sC_{\wt{\Phi}_{\mathrm{hinge}},\sH_{\mathrm{NN}}}(h,x,t)
& = \inf_{x\in \sX} \curl*{1-\inf_{h\in\sH_{\mathrm{NN}}}\sC_{\wt{\Phi}_{\mathrm{hinge}}}(h,x,t)}\\
&\geq\inf_{x\in \sX}\paren*{2t-1}\min\curl*{\Lambda B,1}\\
&=\paren*{2t-1}\min\curl*{\Lambda B,1}\\
&=\sT_1(t),
\end{align*}
where $\sT_1$ is the increasing and convex function on $\bracket*{0,1}$ defined by\\
$
\sT_1(t) = \begin{cases}
\min\curl*{\Lambda B,1} \frac{4\beta}{1+2\beta} \, t, & t\in\bracket*{0,1/2+\beta},\\
\min\curl*{\Lambda B,1} \, (2t-1),  & t \in \bracket*{1/2+\beta,1}.
\end{cases}   
$
\begin{align*}
& \inf_{h\in\sH_{\mathrm{NN}}:\ov h_\gamma(x)<0}\sC_{\wt{\Phi}_{\mathrm{hinge}}}(h,x,t)\\
& \geq \inf_{h\in\sH_{\mathrm{NN}}:\ov h_\gamma(x)<0}\, [t\max\curl*{0,1-\ov h_\gamma(x)}t +(1-t)\max\curl*{0,1+\ov h_\gamma(x)}]\\
& = t\max\curl*{0,1-0}+(1-t)\max\curl*{0,1+0} = 1\\
& \inf_{x\in \sX} \inf_{h\in\sH_{\mathrm{NN}}\colon \ov h_\gamma(x)< 0}\Delta\sC_{\wt{\Phi}_{\mathrm{hinge}},\sH_{\mathrm{NN}}}(h,x,t)\\
& = \inf_{x\in \sX} \curl*{\inf_{h\in\sH_{\mathrm{NN}}:\ov h_\gamma(x)< 0}\sC_{\wt{\Phi}_{\mathrm{hinge}}}(h,x,t)-\inf_{h\in\sH_{\mathrm{NN}}}\sC_{\wt{\Phi}_{\mathrm{hinge}}}(h,x,t)}\\
&\geq\inf_{x\in \sX}\paren*{2t-1}\min\curl*{\Lambda B,1}\\
&=\paren*{2t-1}\min\curl*{\Lambda B,1}\\
&=\sT_2(2t - 1),
\end{align*}
where $\sT_2$ is the increasing and convex function on $\bracket*{0,1}$ defined by
\begin{align*}
\forall t \in [0,1], \quad \sT_2(t) = \min\curl*{\Lambda B,1} \, t \,;
\end{align*}
By Proposition~\ref{prop-adv-noise}, for $\epsilon= 0$, the modified adversarial $\sH_{\mathrm{NN}}$-estimation error transformation of the supremum-based hinge loss under Massart's noise condition with $\beta$ is lower bounded as follows:
\begin{align*}
\sT^M_{\wt{\Phi}_{\mathrm{hinge}}}\geq\wt{\sT}^M_{\wt{\Phi}_{\mathrm{hinge}}}:=\min\curl*{\sT_1,\sT_2}=\begin{cases}
\min\curl*{\Lambda B,1} \, (2t-1), & t \in \bracket*{1/2+\beta,1},\\ \min\curl*{\Lambda B,1}\frac{4\beta}{1+2\beta}\, t, & t\in \left[0,1/2+\beta\right).
\end{cases}
\end{align*}
Note $\wt{\sT}^M_{\wt{\Phi}_{\mathrm{hinge}}}$ is convex, non-decreasing, invertible and satisfies that $\wt{\sT}^M_{\wt{\Phi}_{\mathrm{hinge}}}(0)=0$. By Proposition~\ref{prop-adv-noise}, using the fact that
$\wt{\sT}^M_{\wt{\Phi}_{\mathrm{hinge}}}\geq \min\curl*{\Lambda B,1}\frac{4\beta}{1+2\beta}\, t$ yields the adversarial $\sH_{\mathrm{NN}}$-consistency bound for the supremum-based hinge loss, valid for all $h \in \sH_{\mathrm{NN}}$ and distributions $\sD$ satisfies Massart's noise condition with $\beta$
\begin{align}
\label{eq:hinge-NN-est-adv}
     \sR_{\ell_{\gamma}}(h)- \sR_{\ell_{\gamma}}^*\paren*{\sH_{\mathrm{NN}}} 
     \leq 
      \frac{1+2\beta}{4\beta}\frac{\sR_{\wt{\Phi}_{\mathrm{hinge}}}(h)-\sR_{\wt{\Phi}_{\mathrm{hinge}}}^*\paren*{\sH_{\mathrm{NN}}}+\sM_{\wt{\Phi}_{\mathrm{hinge}}}\paren*{\sH_{\mathrm{NN}}}}{\min\curl*{\Lambda B,1}}-\sM_{\ell_{\gamma}, \sH_{\mathrm{NN}}}
\end{align}
Since
\begin{align*}
\sM_{\ell_{\gamma}}\paren*{\sH_{\mathrm{NN}}}
& = \sR_{\ell_{\gamma}}^*\paren*{\sH_{\mathrm{NN}}}-\mathbb{E}_{X}\bracket*{\min\curl*{\eta(x),1-\eta(x)}},\\
\sM_{\wt{\Phi}_{\mathrm{hinge}}}\paren*{\sH_{\mathrm{NN}}} 
& \leq \sR_{\wt{\Phi}_{\mathrm{hinge}}}^*\paren*{\sH_{\mathrm{NN}}} - \mathbb{E}\bracket*{1-\abs*{2\eta(x)-1}\min\curl*{\sup_{h\in\sH_{\mathrm{NN}}}\uv h_\gamma(x),1}},
\end{align*}
the inequality can be relaxed as follows:
\begin{align*}
     \sR_{\ell_{\gamma}}(h) 
     & \leq 
      \frac{1+2\beta}{4\beta}\frac{\sR_{\wt{\Phi}_{\mathrm{hinge}}}(h)}{\min\curl*{\Lambda B,1}}+\mathbb{E}_{X}\bracket*{\min\curl*{\eta(x),1-\eta(x)}}\\
      & \qquad -\frac{1+2\beta}{4\beta}\frac{\mathbb{E}\bracket*{1-\abs*{2\eta(x)-1}\min\curl*{\sup_{h\in\sH_{\mathrm{NN}}}\uv h_\gamma(x),1}}}{\min\curl*{\Lambda B,1}}.
\end{align*}
Observe that
\begin{align*}
\sup_{h\in\sH_{\mathrm{NN}}}\uv h_\gamma(x)
&=
\sup_{ \|u \|_{1}\leq \Lambda,~\|w_j\|_q\leq W,~\abs*{b}\leq B}\inf_{x'\colon \|x-x'\|_p\leq \gamma}\sum_{j = 1}^n u_j(w_j \cdot x'+b)_{+}\\
& \leq \inf_{x'\colon \|x-x'\|_p\leq \gamma} \sup_{ \|u \|_{1}\leq \Lambda,~\|w_j\|_q\leq W,~\abs*{b}\leq B} \sum_{j = 1}^n u_j(w_j \cdot x'+b)_{+} \\
& = \inf_{x'\colon \|x-x'\|_p\leq \gamma} \Lambda\paren*{W\norm*{x'}_p+B}\\
& =
\begin{cases}
\Lambda\paren*{W\norm*{x}_p-\gamma W + B} & \text{if } \norm*{x}_p \geq \gamma\\
\Lambda B & \text{if } \norm*{x}_p < \gamma
\end{cases}\\
& = \Lambda\paren*{W\max \curl*{\norm*{x}_p,\gamma}-\gamma W + B}.
\end{align*}
Thus, the inequality can be further relaxed as follows:
\begin{align*}
     \sR_{\ell_{\gamma}}(h)
     & \leq 
     \frac{1+2\beta}{4\beta}\frac{\sR_{\wt{\Phi}_{\mathrm{hinge}}}(h)}{\min\curl*{\Lambda B,1}}+\mathbb{E}_{X}\bracket*{\min\curl*{\eta(x),1-\eta(x)}}\\
     &\qquad -\frac{1+2\beta}{4\beta}\frac{\mathbb{E}\bracket*{1-\abs*{2\eta(x)-1}\min\curl*{ \Lambda\paren*{W\max \curl*{\norm*{x}_p,\gamma}-\gamma W + B},1}}}{\min\curl*{\Lambda B,1}}.
\end{align*}
Note that: $\min\curl*{ \Lambda\paren*{W\max \curl*{\norm*{x}_p,\gamma}-\gamma W + B},1}\leq 1$. Thus the  inequality can be further relaxed as follows:
\begin{align}
\label{eq:hinge-NN-est-adv-2}
     \sR_{\ell_{\gamma}}(h)
     \leq \frac{1+2\beta}{4\beta}\frac{\sR_{\wt{\Phi}_{\mathrm{hinge}}}(h)}{ \min\curl*{\Lambda B,1}}-\paren*{\frac{1+2\beta}{2\beta \min\curl*{\Lambda B,1}}-1}\,\mathbb{E}_{X}\bracket*{\min\curl*{\eta(x),1-\eta(x)}}.
\end{align}
When $\Lambda B\geq 1$, it can be equivalently written as follows:
\begin{align*}
     \sR_{\ell_{\gamma}}(h)
     \leq \frac{1+2\beta}{4\beta}\,\sR_{\wt{\Phi}_{\mathrm{hinge}}}(h)-\frac{1}{2\beta }\,\mathbb{E}_{X}\bracket*{\min\curl*{\eta(x),1-\eta(x)}}.
\end{align*}

\subsubsection{Supremum-Based Sigmoid Loss}
For the supremum-based sigmoid loss 
\begin{align*}
\wt{\Phi}_{\mathrm{sig}}\colon=\sup_{x'\colon \|x-x'\|_p\leq \gamma}\Phi_{\mathrm{sig}}(y h(x')),  \quad \text{where } \Phi_{\mathrm{sig}}(\alpha)=1-\tanh(k\alpha),~k>0,   
\end{align*}
for all $h\in \sH_{\mathrm{NN}}$ and $x\in \sX$:
\begin{align*}
\sC_{\wt{\Phi}_{\mathrm{sig}}}(h,x,t) 
&=t \wt{\Phi}_{\mathrm{sig}}(h(x))+(1-t)\wt{\Phi}_{\mathrm{sig}}(-h(x))\\
& = t\Phi_{\mathrm{sig}}\paren*{\uv h_\gamma(x)}+(1-t)\Phi_{\mathrm{sig}}\paren*{-\ov h_\gamma(x)}\\
& =t\paren*{1-\tanh(k \uv h_\gamma(x))} +(1-t)\paren*{1+\tanh(k \ov h_\gamma(x))}\\
& \geq \max\curl*{1+\paren{1-2t}\tanh(k\ov h_\gamma(x)), 1+\paren{1-2t}\tanh(k\uv h_\gamma(x))}\\
\inf_{h\in\sH_{\mathrm{NN}}}\sC_{\wt{\Phi}_{\mathrm{sig}}}(h,x,t)
& \geq \max\curl*{\inf_{h\in\sH_{\mathrm{NN}}}\bracket*{1+\paren{1-2t}\tanh(k\ov h_\gamma(x))}, \inf_{h\in\sH_{\mathrm{NN}}}\bracket*{1+\paren{1-2t}\tanh(k\uv h_\gamma(x))}}\\
& = 1 - \abs*{1-2t}\tanh\paren*{k\sup_{h\in\sH_{\mathrm{NN}}}\uv h_\gamma(x)}\\
\inf_{h\in\sH_{\mathrm{NN}}}\sC_{\wt{\Phi}_{\mathrm{sig}}}(h,x,t)
& \leq \max\curl*{t,1-t}\paren*{1-\tanh\paren*{k \Lambda B }}+\min\curl*{t,1-t}\paren*{1+\tanh\paren*{k \Lambda B}}\\
& = 1 - \abs*{1-2t}\tanh\paren*{k \Lambda B}\\
\sM_{\wt{\Phi}_{\mathrm{sig}}}\paren*{\sH_{\mathrm{NN}}} 
& = \sR_{\wt{\Phi}_{\mathrm{sig}}}^*\paren*{\sH_{\mathrm{NN}}} - \mathbb{E}\bracket*{\inf_{h\in\sH_{\mathrm{NN}}}\sC_{\wt{\Phi}_{\mathrm{sig}}}(h,x,\eta(x))}\\
& \leq \sR_{\wt{\Phi}_{\mathrm{sig}}}^*\paren*{\sH_{\mathrm{NN}}} - \mathbb{E}\bracket*{1 - \abs*{1-2\eta(x)}\tanh\paren*{k\sup_{h\in\sH_{\mathrm{NN}}}\uv h_\gamma(x)}}
\end{align*}
For $\frac{1}2< t\leq1$, we have
\begin{align*}
\inf_{h\in\sH_{\mathrm{NN}}:\uv h_\gamma(x)\leq 0 \leq \ov h_\gamma(x)}\sC_{\wt{\Phi}_{\mathrm{sig}}}(h,x,t) 
& = t+(1-t)\\
& =1\\
\inf_{x\in \sX} \inf_{h\in\sH_{\mathrm{NN}}\colon \uv h_\gamma(x)\leq 0 \leq \ov h_\gamma(x)}\Delta\sC_{\wt{\Phi}_{\mathrm{sig}},\sH_{\mathrm{NN}}}(h,x,t)
& = \inf_{x\in \sX} \curl*{1-\inf_{h\in\sH_{\mathrm{NN}}}\sC_{\wt{\Phi}_{\mathrm{sig}}}(h,x,t)}\\
&\geq \inf_{x\in \sX} (2t-1)\tanh\paren*{k \Lambda B}\\
&=(2t-1)\tanh\paren*{k \Lambda B}\\
&=\sT_1(t),
\end{align*}
where $\sT_1$ is the increasing and convex function on $\bracket*{0,1}$ defined by
\begin{align*}
\sT_1(t) = \begin{cases}
\tanh\paren*{k \Lambda B} \frac{4\beta}{1+2\beta} \, t, & t\in\bracket*{0,1/2+\beta},\\
\tanh\paren*{k \Lambda B} \, (2t-1),  & t \in \bracket*{1/2+\beta,1}.
\end{cases}    
\end{align*}
\begin{align*}
\inf_{h\in\sH_{\mathrm{NN}}:\ov h_\gamma(x)<0}\sC_{\wt{\Phi}_{\mathrm{sig}}}(h,x,t)
& \geq \inf_{h\in\sH_{\mathrm{NN}}:\ov h_\gamma(x)<0} 1+\paren{1-2t}\tanh(k\ov h_\gamma(x))\\
& = 1\\
\inf_{x\in \sX} \inf_{h\in\sH_{\mathrm{NN}}\colon \ov h_\gamma(x)< 0}\Delta\sC_{\wt{\Phi}_{\mathrm{sig}},\sH_{\mathrm{NN}}}(h,x)
& = \inf_{x\in \sX} \curl*{\inf_{h\in\sH_{\mathrm{NN}}:\ov h_\gamma(x)< 0}\sC_{\wt{\Phi}_{\mathrm{sig}}}(h,x,t)-\inf_{h\in\sH_{\mathrm{NN}}}\sC_{\wt{\Phi}_{\mathrm{sig}}}(h,x,t)}\\
&\geq \inf_{x\in \sX}(2t-1)\tanh\paren*{k \Lambda B}\\
&=(2t-1)\tanh\paren*{k \Lambda B}\\
&=\sT_2(2t-1),
\end{align*}
where $\sT_2$ is the increasing and convex function on $\bracket*{0,1}$ defined by
\begin{align*}
\forall t \in [0,1], \quad \sT_2(t) = \tanh\paren*{k \Lambda B} \, t \,;
\end{align*}
By Proposition~\ref{prop-adv-noise}, for $\epsilon= 0$, the modified adversarial $\sH_{\mathrm{NN}}$-estimation error transformation of the supremum-based sigmoid loss under Massart's noise condition with $\beta$ is lower bounded as follows:
\begin{align*}
\sT^M_{\wt{\Phi}_{\mathrm{sig}}}\geq \wt{\sT}^M_{\wt{\Phi}_{\mathrm{sig}}}=\min\curl*{\sT_1,\sT_2}=\begin{cases}
\tanh\paren*{k \Lambda B} \frac{4\beta}{1+2\beta} \, t, & t\in\bracket*{0,1/2+\beta},\\
\tanh\paren*{k \Lambda B} \, (2t-1),  & t \in \bracket*{1/2+\beta,1}.
\end{cases}    
\end{align*}
Note $\wt{\sT}^M_{\wt{\Phi}_{\mathrm{sig}}}$ is convex, non-decreasing, invertible and satisfies that $\wt{\sT}^M_{\wt{\Phi}_{\mathrm{sig}}}(0)=0$. By Proposition~\ref{prop-adv-noise}, using the fact that
$\wt{\sT}^M_{\wt{\Phi}_{\mathrm{sig}}}\geq \tanh\paren*{k \Lambda B} \frac{4\beta}{1+2\beta} \, t$ yields the adversarial $\sH_{\mathrm{NN}}$-consistency bound for the supremum-based sigmoid loss, valid for all $h \in \sH_{\mathrm{NN}}$ and distributions $\sD$ satisfies Massart's noise condition with $\beta$:
\begin{align}
\label{eq:sig-NN-est-adv}
     \sR_{\ell_{\gamma}}(h)- \sR_{\ell_{\gamma}}^*\paren*{\sH_{\mathrm{NN}}} 
     \leq 
     \frac{1+2\beta}{4\beta}\frac{\sR_{\wt{\Phi}_{\mathrm{sig}}}(h)-\sR_{\wt{\Phi}_{\mathrm{sig}}}^*\paren*{\sH_{\mathrm{NN}}}+\sM_{\wt{\Phi}_{\mathrm{sig}}}\paren*{\sH_{\mathrm{NN}}}}{\tanh\paren*{k \Lambda B}}-\sM_{\ell_{\gamma}, \sH_{\mathrm{NN}}}
\end{align}
Since
\begin{align*}
\sM_{\ell_{\gamma}}\paren*{\sH_{\mathrm{NN}}}
& = \sR_{\ell_{\gamma}}^*\paren*{\sH_{\mathrm{NN}}}-\mathbb{E}_{X}\bracket*{\min\curl*{\eta(x),1-\eta(x)}},\\
\sM_{\wt{\Phi}_{\mathrm{sig}}}\paren*{\sH_{\mathrm{NN}}} 
& \leq \sR_{\wt{\Phi}_{\mathrm{sig}}}^*\paren*{\sH_{\mathrm{NN}}} - \mathbb{E}\bracket*{1 - \abs*{1-2\eta(x)}\tanh\paren*{k\sup_{h\in\sH_{\mathrm{NN}}}\uv h_\gamma(x)}},
\end{align*}
the inequality can be relaxed as follows:
\begin{align*}
     \sR_{\ell_{\gamma}}(h) 
     & \leq \frac{1+2\beta}{4\beta}\frac{\sR_{\wt{\Phi}_{\mathrm{sig}}}(h)}{\tanh\paren*{k \Lambda B}}+\mathbb{E}_{X}\bracket*{\min\curl*{\eta(x),1-\eta(x)}}\\
     & \qquad -\frac{1+2\beta}{4\beta}\frac{\mathbb{E}\bracket*{1 - \abs*{1-2\eta(x)}\tanh\paren*{k\sup_{h\in\sH_{\mathrm{NN}}}\uv h_\gamma(x)}}}{\tanh\paren*{k \Lambda B}}.
\end{align*}
Observe that
\begin{align*}
\sup_{h\in\sH_{\mathrm{NN}}}\uv h_\gamma(x)
&=
\sup_{ \|u \|_{1}\leq \Lambda,~\|w_j\|_q\leq W,~\abs*{b}\leq B}\inf_{x'\colon \|x-x'\|_p\leq \gamma}\sum_{j = 1}^n u_j(w_j \cdot x'+b)_{+}\\
& \leq \inf_{x'\colon \|x-x'\|_p\leq \gamma} \sup_{ \|u \|_{1}\leq \Lambda,~\|w_j\|_q\leq W,~\abs*{b}\leq B} \sum_{j = 1}^n u_j(w_j \cdot x'+b)_{+} \\
& = \inf_{x'\colon \|x-x'\|_p\leq \gamma} \Lambda\paren*{W\norm*{x'}_p+B}\\
& =
\begin{cases}
\Lambda\paren*{W\norm*{x}_p-\gamma W + B} & \text{if } \norm*{x}_p \geq \gamma\\
\Lambda B & \text{if } \norm*{x}_p < \gamma
\end{cases}\\
& = \Lambda\paren*{W\max \curl*{\norm*{x}_p,\gamma}-\gamma W + B}.
\end{align*}
Thus, the inequality can be further relaxed as follows:
\begin{align*}
     \sR_{\ell_{\gamma}}(h) & \leq \frac{1+2\beta}{4\beta}\frac{\sR_{\wt{\Phi}_{\mathrm{sig}}}(h)}{\tanh\paren*{k \Lambda B}}+\mathbb{E}_{X}\bracket*{\min\curl*{\eta(x),1-\eta(x)}}\\
     & \qquad -\frac{1+2\beta}{4\beta}\frac{\mathbb{E}\bracket*{1 - \abs*{1-2\eta(x)}\tanh\paren*{k\Lambda\paren*{W\max \curl*{\norm*{x}_p,\gamma}-\gamma W + B}}}}{\tanh\paren*{k \Lambda B}}.
\end{align*}
Note that: $\tanh\paren*{k\Lambda\paren*{W\max \curl*{\norm*{x}_p,\gamma}-\gamma W + B}}\leq 1$ and $1-\abs*{1-2\eta(x)}=2\min\curl*{\eta(x),1-\eta(x)}$. Thus the inequality can be further relaxed as follows:
\begin{align}
\label{eq:sig-NN-est-adv-2}
     \sR_{\ell_{\gamma}}(h)
     \leq \frac{1+2\beta}{4\beta}\frac{\sR_{\wt{\Phi}_{\mathrm{sig}}}(h)}{\tanh\paren*{k\Lambda B}}-\paren*{\frac{1+2\beta}{2\beta \tanh(k\Lambda B)}-1}\,\mathbb{E}_{X}\bracket*{\min\curl*{\eta(x),1-\eta(x)}}.
\end{align}
When $\Lambda B=+ \infty$, it can be equivalently written as follows:
\begin{align*}
     \sR_{\ell_{\gamma}}(h)
     \leq \frac{1+2\beta}{4\beta}\,\sR_{\wt{\Phi}_{\mathrm{sig}}}(h)-\frac{1}{2\beta}\,\mathbb{E}_{X}\bracket*{\min\curl*{\eta(x),1-\eta(x)}}.
\end{align*}
\restoreatoc

\chapter{Appendix to Chapter~\ref{ch3}}

\disableatoc
\section{Discussion on multi-class \texorpdfstring{$0/1$}{0/1} loss}
\label{app:dicussion-01}
The multi-class $0/1$ loss can be defined in multiple ways, e.g. $\mathds{1}_{\rho_{h}(x,y)\leq 0}$, $\mathds{1}_{\rho_{h}(x,y)< 0}$ and $\mathds{1}_{\hh(x)\neq y}$ where $\hh(x) = \argmax_{y \in \sY} h(x, y)$ with an arbitrary but fixed
deterministic strategy used for breaking ties. The counterparts of these three formulas in binary classification are $\mathds{1}_{yh(x)\leq 0}$, $\mathds{1}_{yh(x)< 0}$ and $\mathds{1}_{\sgn(h(x))\neq y}$ where $\sgn(0)$ is defined as $+1$ or $-1$.
To be consistent with the literature on Bayes-consistency \citep{bartlett2006convexity,tewari2007consistency}, in this paper we adopt the last formula $\mathds{1}_{\hh(x)\neq y}$ of multi-class $0/1$ loss. Moreover, to be consistent with the binary case \citep{awasthi2022Hconsistency},
 we assume that
in case of a tie, $\hh(x)$ is 
defined as the label with the
highest index under the natural ordering of labels. This assumption corresponds to the binary case where we always predict $+1$ in case of a tie, that is, the case where the binary $0/1$ loss is defined by $\mathds{1}_{\sgn(h(x))\neq y}$ with $\sgn(0) = +1$, as in 
\citep{awasthi2022Hconsistency}.
Nevertheless, other deterministic strategies would lead to similar results.

\section{Discussion on finite sample bounds}
\label{app:finite-sample-bounds}
Here, we discuss several ways to derive the finite
sample bounds on the estimation error for the target $0/1$ loss. One can directly derive estimation error bounds for the $0/1$ loss, typically for Empirical Risk Minimization (ERM), e.g. $\sR_{\ell_{0-1}}\paren*{h_{S}^{\mathrm{ERM}}}-\sR^*_{\ell_{0-1}}(\sH)$ with $h_{S}^{\mathrm{ERM}}=\argmin_{h\in \sH} \h{\sR}_{S}(h)$ can be upper-bounded using the standard generalization bounds, as shown in \citep{MohriRostamizadehTalwalkar2018}.
But, those bounds would not say anything about the use of a surrogate loss.

An alternative is to use the excess error bound for the target $0/1$ loss and split the excess error of the surrogate loss into an estimation term and an approximation term, i.e. for some function $f\colon \Rset_{+}\to \Rset_{+}$, the following inequality holds:
\begin{align*}
\sR_{\ell_{0-1}}\paren*{h}-\sR^*_{\ell_{0-1}}\paren*{\sH_{\mathrm{all}}}\leq f \paren*{\sR_{\ell_{\mathrm{sur}}}\paren*{h}-\sR^*_{\ell_{\mathrm{sur}}}(\sH)+\sR^*_{\ell_{\mathrm{sur}}}(\sH)-\sR^*_{\ell_{\mathrm{sur}}}\paren*{\sH_{\mathrm{all}}}}.
\end{align*}
Then, an estimation error bound for the surrogate loss can be used to upper-bound $\sR_{\ell_{\mathrm{sur}}}\paren*{h}-\sR^*_{\ell_{\mathrm{sur}}}(\sH)$, as shown in \citep{bartlett2006convexity}. But, those bounds would not be an estimation error guarantee for the target loss $\ell_{0-1}$. 

\ignore{ Another alternative is to upper-bound the empirical error,
  $\h{\sR}_{S}(h)=\frac{1}{m}\sum_{i=1}^m\ell_{0-1}(h,x_i,y_i)$ where
  $m$ is the sample size, by the surrogate counterpart
  $\frac{1}{m}\sum_{i=1}^m\ell_{\mathrm{sur}}(h,x_i,y_i)$ in the
  standard generalization bounds of $0/1$ loss and seek to similarly
  derive an ``estimation bound'' of the surrogate loss. But, that
  cannot be done because we only have one-sided inequalities when
  using a surrogate loss.  }

Finally, using the $\sH$-consistency bound proposed by \citet{awasthi2022Hconsistency}, that is, for some non-decreasing function $f\colon \Rset_{+}\to \Rset_{+}$,
\begin{align*}
\sR_{\ell_{0-1}}\paren*{h}-\sR^*_{\ell_{0-1}}(\sH)\leq f \paren*{\sR_{\ell_{\mathrm{sur}}}\paren*{h}-\sR^*_{\ell_{\mathrm{sur}}}(\sH)},   
\end{align*}
we can directly derive the estimation error bound for the target $0/1$ loss by upper-bounding $\sR_{\ell_{\mathrm{sur}}}\paren*{h}-\sR^*_{\ell_{\mathrm{sur}}}(\sH)$ with the estimation error bound for the surrogate loss. In conclusion, the $\sH$-consistency bound is a useful tool to derive non-trivial finite sample bounds on the estimation error for the target $0/1$ loss.

\section{General \texorpdfstring{$\sH$}{H}-consistency bounds}
\label{app:deferred_proofs_general}

\ExcessBoundsPsiMhcb*
\begin{proof}
For any $h\in \sH$ and $\sD\in \sP$, since $\Psi\paren*{\Delta\sC_{\ell_2,\sH}(h,x)\mathds{1}_{\Delta\sC_{\ell_2,\sH}(h,x)>\epsilon}}\leq \Delta\sC_{\ell_1,\sH}(h,x),\forall x\in \sX$, we can write
    \begin{align*}
       &\Psi\paren*{\sR_{\ell_2}(h)-\sR_{\ell_2}^*(\sH)+\sM_{\ell_2}(\sH)}\\
       &=\Psi\paren*{\mathbb{E}_{X}  \bracket*{\sC_{\ell_2}(h,x)-\sC^*_{\ell_2}(\sH, x)}}\\
       &=\Psi\paren*{\mathbb{E}_{X}  \bracket*{\Delta\sC_{\ell_2,\sH}(h,x)}}\\
       &\leq\mathbb{E}_{X} \bracket*{\Psi\paren*{\Delta\sC_{\ell_2,\sH}(h,x)}} & (\text{Jensen's ineq.}) \\
        &=\mathbb{E}_{X}  \bracket*{\Psi\paren*{\Delta\sC_{\ell_2,\sH}(h,x)\mathds{1}_{\Delta\sC_{\ell_2,\sH}(h,x)>\epsilon}+\Delta\sC_{\ell_2,\sH}(h,x)\mathds{1}_{\Delta\sC_{\ell_2,\sH}(h,x)\leq\epsilon}}}\\ 
       &\leq\mathbb{E}_{X} \bracket*{\Psi\paren*{\Delta\sC_{\ell_2,\sH}(h,x)\mathds{1}_{\Delta\sC_{\ell_2,\sH}(h,x)>\epsilon}}+\Psi\paren*{\Delta\sC_{\ell_2,\sH}(h,x)\mathds{1}_{\Delta\sC_{\ell_2,\sH}(h,x)\leq\epsilon}}}  &\paren*{\Psi(0)\geq 0}\\ 
       &\leq\mathbb{E}_{X} \bracket*{\Delta\sC_{\ell_1,\sH}(h,x)}+\sup_{t\in[0,\epsilon]}\Psi(t) &\paren*{\text{assumption}}\\
       &=\sR_{\ell_1}(h)-\sR_{\ell_1}^*(\sH)+\sM_{\ell_1}(\sH)+\max\curl*{\Psi(0),\Psi(\e)}, &\paren*{\text{convexity of $\Psi$}}
    \end{align*}
    which proves the theorem.
\end{proof}

\ExcessBoundsGamma*
\begin{proof}
For any $h\in \sH$ and $\sD\in \sP$, since $\Delta\sC_{\ell_2,\sH}(h,x)\mathds{1}_{\Delta\sC_{\ell_2,\sH}(h,x)>\epsilon}\leq \Gamma\paren*{\Delta\sC_{\ell_1,\sH}(h,x)},\forall x\in \sX$, we can write
    \begin{align*}
       &\sR_{\ell_2}(h)-\sR_{\ell_2}^*(\sH)+\sM_{\ell_2}(\sH)\\
       &=\mathbb{E}_{X}  \bracket*{\sC_{\ell_2}(h,x)-\sC^*_{\ell_2}(\sH, x)}\\
       &=\mathbb{E}_{X}  \bracket*{\Delta\sC_{\ell_2,\sH}(h,x)}\\
        &=\mathbb{E}_{X}  \bracket*{\Delta\sC_{\ell_2,\sH}(h,x)\mathds{1}_{\Delta\sC_{\ell_2,\sH}(h,x)>\epsilon}+\Delta\sC_{\ell_2,\sH}(h,x)\mathds{1}_{\Delta\sC_{\ell_2,\sH}(h,x)\leq\epsilon}}\\ 
        &\leq\mathbb{E}_{X}  \bracket*{\Gamma\paren*{\Delta\sC_{\ell_1,\sH}(h,x)}}+\epsilon &\paren*{\text{assumption}}\\ 
        &\leq\Gamma\paren*{\mathbb{E}_{X}  \bracket*{\Delta\sC_{\ell_1,\sH}(h,x)}}+\epsilon &\paren*{\text{concavity of $\Gamma$}}\\ 
       &=\Gamma\paren*{\sR_{\ell_1}(h)-\sR_{\ell_1}^*(\sH)+\sM_{\ell_1}(\sH)}+\epsilon,
    \end{align*}
    which proves the theorem.
\end{proof}

\section{Non-adversarial and adversarial conditional regrets}
\label{app:deferred_proofs_cond}

\ExplicitAssumption*
\begin{proof}
By the definition, the conditional $\ell_{0-1}$-risk can be expressed as follows: 
\begin{align}
\label{eq:cond}
\sC_{\ell_{0-1}}(h,x)  =  \sum_{y\in \sY} \sfp(y \!\mid\! x) \mathds{1}_{\hh(x)\neq y}=
1-\sfp(\hh(x) \!\mid\! x).
\end{align}
Since $\curl*{\hh(x):h\in \sH}=\mathsf H(x)$, the minimal conditional $\ell_{0-1}$-risk can be expressed as follows:
\begin{align*}
\sC^*_{\ell_{0-1}}(\sH, x)=1-\max_{y\in \mathsf H(x)} \sfp(y \!\mid\! x),
\end{align*}
which proves the first part of the lemma. By the definition,
\begin{align*}
\Delta\sC_{\ell_{0-1},\sH}(h,x)
= \sC_{\ell_{0-1}}(h,x)- \sC^*_{\ell_{0-1}}(\sH, x) =
\max_{y\in \mathsf H(x)} \sfp(y \!\mid\! x) - \sfp(\hh(x) \!\mid\! x).
\end{align*}
This leads to
\begin{align*}
\bracket*{\Delta\sC_{\ell_{0-1},\sH}(h,x)}_{\e}=
\bracket*{\max_{y\in \mathsf H(x)} \sfp(y \!\mid\! x) - \sfp(\hh(x) \!\mid\! x)}_{\e}.
\end{align*}
\end{proof}

\ExplicitAssumptionAdv*
\begin{proof}
By the definition, the conditional $\ell_{\gamma}$-risk can be expressed as follows: 
\begin{align}
\label{eq:cond_adv}
\sC_{\ell_{\gamma}}(h,x)  =  \sum_{y\in \sY} \sfp(y \!\mid\! x) \sup_{x':\norm*{x-x'}_p\leq \gamma}\mathds{1}_{\rho_h(x', y) \leq 0}=
\begin{cases}
1-\sfp(\hh(x) \!\mid\! x) & h\in \sH_{\gamma}(x) \\
1 & \text{otherwise.}
\end{cases}
\end{align}
When $\sH_\gamma(x)=\emptyset$, \eqref{eq:cond_adv} implies that $\sC^*_{\ell_{\gamma}}(\sH, x)=1$. When $\sH_\gamma(x)\neq \emptyset$, $\mathsf H_\gamma(x)$ is also non-empty.
By \eqref{eq:cond_adv}, $y\in \sY_\gamma(x)$ if and only if there exists $h\in \sH_\gamma$ such that $\sC_{\ell_{\gamma}}(h,x)=1-\sfp(y \!\mid\! x)$.
Therefore, the minimal conditional $\ell_{\gamma}$-risk can be expressed as follows:
\begin{align*}
\sC^*_{\ell_{\gamma}}(\sH, x)=1-\max_{y\in \mathsf H_\gamma(x)} \sfp(y \!\mid\! x) \mathds{1}_{\sH_\gamma(x)\neq\emptyset},
\end{align*}
which proves the first part of lemma.
When $\sH_{\gamma}(x)=\emptyset$, $\sC_{\ell_{\gamma}}(h,x)\equiv 1$, which implies that $\Delta\sC_{\ell_{\gamma},\sH}(h,x) \equiv 0$. When $\sH_{\gamma}(x)\neq\emptyset$, $\mathsf H_\gamma(x)$ is also non-empty, for $h\in \sH_{\gamma}(x)$, $\Delta\sC_{\ell_{\gamma},\sH}(h,x)=1-\sfp(\hh(x) \!\mid\! x)-\paren*{1-\max_{y\in \mathsf H_\gamma(x)} \sfp(y \!\mid\! x)}=\max_{y\in \mathsf H_\gamma(x)} \sfp(y \!\mid\! x)-\sfp(\hh(x) \!\mid\! x)$; for $h\notin \sH_{\gamma}(x)$, $\Delta\sC_{\ell_{\gamma},\sH}(h,x)=1-\paren*{1-\max_{y\in \mathsf H_\gamma(x)} \sfp(y \!\mid\! x)}=\max_{y\in \mathsf H_\gamma(x)} \sfp(y \!\mid\! x)$. Therefore,
\begin{align*}
\Delta\sC_{\ell_{\gamma},\sH}(h,x)= 
\begin{cases}
\max_{y\in \mathsf H_\gamma(x)} \sfp(y \!\mid\! x)-\sfp(\hh(x) \!\mid\! x)\mathds{1}_{h\in \sH_{\gamma}(x)} & \sH_{\gamma}(x)\neq \emptyset \\
0 & \text{otherwise.}
\end{cases}
\end{align*}
This leads to
\begin{align*}
\bracket*{\Delta\sC_{\ell_{\gamma},\sH}(h,x)}_{\e}=
\begin{cases}
\bracket*{\max_{y\in \mathsf H_\gamma(x)} \sfp(y \!\mid\! x)-\sfp(\hh(x) \!\mid\! x)\mathds{1}_{h\in \sH_{\gamma}(x)}}_{\e} & \sH_{\gamma}(x)\neq \emptyset \\
0 & \text{otherwise.}
\end{cases}
\end{align*}
\ignore{
The second part of the lemma follows immediately by the fact that $\mathsf H_\gamma(x) =\sY$ when $\sH_\gamma(x)\neq\emptyset$ under the assumption that $\sH$ is symmetric.}
\end{proof}

\section{Proof of negative results and \texorpdfstring{$\sH$}{H}-consistency bounds for max losses \texorpdfstring{$ \Phi^{\mathrm{max}}$}{max}}
\label{app:deferred_proofs_max}

\NegativeMax*
\begin{proof}
Consider the distribution that supports on a singleton domain $\curl*{x}$ with $x$ satisfying that $\abs*{\mathsf H(x)}\geq 2$. Take $y_1 \in \mathsf H(x)$ such that $y_1 \neq c$ and $y_2\in \sY$ such that $y_2\neq y_1,\, y_2\neq c$. We define $p(x)$ as $\sfp(y_1 \!\mid\! x)=\sfp(y_2 \!\mid\! x)=\frac12$ and $\sfp(y \!\mid\! x)=0$ for other $y\in \sY$. Let $h_0\in \sH$ such that $h_0(x,1)=h_0(x,2)=\ldots=h_0(x,c)$. By Lemma~\ref{lemma:explicit_assumption_01-mhcb} and the fact that $y_1\in \mathsf H(x)$, the minimal conditional $\ell_{0-1}$-risk is
\begin{align*}
\sR^*_{\ell_{0-1}}(\sH)=
\sC^*_{\ell_{0-1}}(\sH, x)=1-\max_{y\in \mathsf H(x)}\sfp(y \!\mid\! x)=1-\sfp(y_1 \!\mid\! x)=\frac12.
\end{align*}
For $h=h_0$, we have
\begin{align*}
\sR_{\ell_{0-1}}(h_0)=\sC_{\ell_{0-1}}(h_0,x)=\sum_{y\in \sY} \sfp(y \!\mid\! x) \mathds{1}_{\hh_0(x)\neq y}=
1-\sfp(\hh_0(x) \!\mid\! x)=1-\sfp(c \!\mid\! x)=1.
\end{align*} For the max loss, the conditional $\Phi^{\mathrm{max}}$-risk can be expressed as follows:
\begin{align*}
\sC_{\Phi^{\mathrm{max}}}(h,x)  &=  \sum_{y\in \sY} \sfp(y \!\mid\! x) \Phi\paren*{\rho_h(x, y)}=\frac12 \Phi\paren*{\rho_h(x, y_1)} + \frac12 \Phi\paren*{\rho_h(x, y_2)}.
\end{align*}
If $\Phi$ is convex and non-increasing, we obtain for any $h\in \sH$,
\begin{align*}
&\sR_{\Phi^{\mathrm{max}}}(h)\\
&=\sC_{\Phi^{\mathrm{max}}}(h,x)\\
&=\frac12 \Phi\paren*{\rho_h(x, y_1)} + \frac12 \Phi\paren*{\rho_h(x, y_2)}\\
& \geq \Phi\paren*{\frac12 \rho_h(x, y_1)+ \frac12 \rho_h(x, y_2)} & \paren*{\Phi \text{ is convex}}\\
& = \Phi\paren*{\frac12 \paren*{h(x,y_1)+h(x,y_2)-\max_{y\neq y_1}h(x,y)-\max_{y\neq y_2}h(x,y)}}\\
& \geq \Phi(0), & \paren*{\Phi \text{ is non-increasing}}
\end{align*}
where both equality can be achieved by $h_0$.
Therefore,
\begin{align*}
\sR_{\Phi^{\mathrm{max}}}^*(\sH)=\sC^*_{\Phi^{\mathrm{max}}}(\sH, x)=\sR_{\Phi^{\mathrm{max}}}(h_0)=\Phi(0).
\end{align*}
If \eqref{eq:bound_max_convex} holds for some non-decreasing function $f$, then, we obtain for any $h\in \sH$,
\begin{align*}
\sR_{\ell_{0-1}}(h)-\frac12\leq  f\paren*{\sR_{\Phi^{\mathrm{max}}}(h) - \Phi(0)}.
\end{align*}
Let $h=h_0$, then $f(0)\geq 1/2$. Since $f$ is non-decreasing, for any $t\geq 0$, $f(t)\geq 1/2$.
\end{proof}

\BoundMaxRho*
\begin{proof}
By the definition, the conditional $\Phi_{\rho}^{\mathrm{max}}$-risk can be expressed as follows:
\begin{equation}
\label{eq:cond_rho}
\begin{aligned}
\sC_{\Phi_{\rho}^{\mathrm{max}}}(h,x)  &=  \sum_{y\in \sY} \sfp(y \!\mid\! x) \Phi_{\rho}\paren*{\rho_h(x, y)}\\
& =
1-\sfp(\hh(x) \!\mid\! x)+ \max\curl*{0,1-\frac{\rho_h(x, \hh(x))}{\rho}}\,\sfp(\hh(x) \!\mid\! x) \\
& =
1- \min\curl*{1,\frac{\rho_h(x, \hh(x))}{\rho}}\,\sfp(\hh(x) \!\mid\! x)
\end{aligned}
\end{equation}
Since $\sH$ is symmetric, for any $x\in \sX$ and $y\in \sY$, 
\begin{align*}
\sup_{h\in \curl*{h\in\sH: \hh(x)=y}}\rho_h(x,\hh(x))=\sup_{h\in \sH}\rho_h(x,\hh(x))   
\end{align*}
Therefore, the minimal conditional $\Phi_{\rho}^{\mathrm{max}}$-risk can be expressed as follows:
\begin{align*}
\sC^*_{\Phi_{\rho}^{\mathrm{max}}}(\sH, x)=1-\min\curl*{1,\frac{\sup_{h\in \sH}\rho_h(x,\hh(x))}{\rho}}\max_{y\in \sY}\sfp(y \!\mid\! x).
\end{align*}
By the definition and using the fact that $\mathsf H(x)=\sY$ when $\sH$ is symmetric, we obtain
\begin{align*}
& \Delta\sC_{\Phi_{\rho}^{\mathrm{max}},\sH}(h,x)\\
& = \sC_{\Phi_{\rho}^{\mathrm{max}}}(h,x) - \sC^*_{\Phi_{\rho}^{\mathrm{max}}}(\sH, x) \\
& = \min\curl*{1,\frac{\sup_{h\in \sH}\rho_h(x,\hh(x))}{\rho}}\max_{y\in \sY}\sfp(y \!\mid\! x) - \min\curl*{1,\frac{\rho_h(x, \hh(x))}{\rho}}\,\sfp(\hh(x) \!\mid\! x)\\
& \geq \min\curl*{1,\frac{\sup_{h\in \sH}\rho_h(x,\hh(x))}{\rho}}\paren*{\max_{y\in \sY}\sfp(y \!\mid\! x) - \sfp(\hh(x) \!\mid\! x)}\\
& \geq \min\curl*{1,\frac{\sup_{h\in \sH}\rho_h(x,\hh(x))}{\rho}} \Delta\sC_{\ell_{0-1},\sH}(h,x) & \paren*{\mathsf H(x)=\sY}\\
& \geq \min\curl*{1,\frac{\sup_{h\in \sH}\rho_h(x,\hh(x))}{\rho}}\bracket*{\Delta\sC_{\ell_{0-1},\sH}(h,x)}_{\e} & \paren*{\bracket*{x}_{\e}\leq x}\\
& \geq \min\curl*{1,\frac{\inf_{x\in \sX}\sup_{h\in \sH}\rho_h(x,\hh(x))}{\rho}}\bracket*{\Delta\sC_{\ell_{0-1},\sH}(h,x)}_{\e}\\
\end{align*}
for any $\e\geq 0$. Therefore, taking $\sP$ be the set of all distributions, $\sH$ be the symmetric hypothesis set, $\e=0$ and
\begin{align*}
\Psi(t)=\min\curl*{1,\frac{\inf_{x\in \sX}\sup_{h\in \sH}\rho_h(x,\hh(x))}{\rho}}\,t
\end{align*}
in Theorem~\ref{Thm:excess_bounds_Psi_01_general-mhcb}, or, equivalently, $\Gamma(t) = \Psi^{-1}(t)$ in Theorem~\ref{Thm:excess_bounds_Gamma_01_general-mhcb}, we obtain for any hypothesis $h\in\sH$ and any distribution,
\begin{align*}
\sR_{\ell_{0-1}}(h)-\sR_{\ell_{0-1}}^*(\sH)\leq\frac{\sR_{\Phi_{\rho}^{\mathrm{max}}}(h)-\sR_{\Phi_{\rho}^{\mathrm{max}}}^*(\sH)+\sM_{\Phi_{\rho}^{\mathrm{max}}}(\sH)}{\min\curl*{1,\frac{\inf_{x\in \sX}\sup_{h\in\sH}\rho_h(x,\hh(x))}{\rho}}}-\sM_{\ell_{0-1}}(\sH).
\end{align*}
\end{proof}

\BoundMaxRe*
\begin{proof}
Under the $\sH$-realizability assumption of distribution, for any $x\in \sX$, there exists $y\in \sY$ such that $\sfp(y \!\mid\! x)=1$. Then, the conditional $\Phi^{\mathrm{max}}$-risk can be expressed as follows:
\begin{equation}
\label{eq:cond_max_re}
\begin{aligned}
\sC_{\Phi^{\mathrm{max}}}(h,x)  &=  \sum_{y\in \sY} \sfp(y \!\mid\! x) \Phi\paren*{\rho_h(x, y)}\\
& = \Phi\paren*{\rho_h(x, y_{\max})}.
\end{aligned}
\end{equation}
Since $\sH$ is symmetric and complete, there exists $h\in \sH$ such that $\hh(x)=y_{\max}$ and we have 
\begin{align*}
\sup_{h\in \curl*{h\in\sH: \hh(x)=y_{\max}}}\rho_h(x,\hh(x))&=\sup_{h\in \sH}\rho_h(x,\hh(x))\\
&=\sup_{h\in\sH}\paren*{\max_{y\in \sY}h(x,y)-\max_{y\neq \hh(x)} h(x,y)}\\
&= + \infty.
\end{align*}
Thus, using the fact that $\lim_{t\to+\infty}\Phi(t)=0$, the minimal conditional $\Phi^{\mathrm{max}}$-risk can be expressed as follows:
\begin{align*}
\sC^*_{\Phi^{\mathrm{max}}}(\sH, x)
&=\inf_{h\in \sH}\sC_{\Phi^{\mathrm{max}}}(h,x)\\
& = \inf_{h\in\sH} \Phi\paren*{\rho_h(x, \hh(x))}\\
& =\Phi\paren*{\sup_{h\in\sH}\rho_h(x, \hh(x))}\tag{$\Phi$ is non-increasing}\\
& = 0 \tag{$\lim_{t\to+\infty}\Phi(t)=0$}
\end{align*}
By the definition and using the fact that $\mathsf H(x)=\sY$ when $\sH$ is symmetric, we obtain
\begin{align*}
\Delta \sC_{\Phi^{\mathrm{max}},\sH}(h,x)
&=\sC_{\Phi^{\mathrm{max}}}(h,x) - \sC^*_{\Phi^{\mathrm{max}}}(\sH, x)\\
& = \Phi\paren*{\rho_h(x, y_{\max})}\\
&\geq \Phi(0)\mathds{1}_{y_{\max}\neq \hh(x)} \tag{$\Phi$ is non-increasing}\\
& \geq \max_{y\in \sY}\sfp(y \!\mid\! x) - \sfp(\hh(x) \!\mid\! x)\\
& =  \Delta\sC_{\ell_{0-1},\sH}(h,x) \tag{by Lemma~\ref{lemma:explicit_assumption_01-mhcb} and $\mathsf H(x)=\sY$}\\
& \geq \bracket*{\Delta\sC_{\ell_{0-1},\sH}(h,x)}_{\e} \tag{$\bracket*{t}_{\e}\leq t$}
\end{align*}
for any $\e\geq 0$. Note that $\sM_{\ell_{0-1}}(\sH)=0$ under the realizability assumption. Therefore, taking $\sP$ be the set of $\sH$-realizable distributions, $\sH$ be the symmetric and complete hypothesis set, $\e=0$ and $\Psi(t)=t$ in Theorem~\ref{Thm:excess_bounds_Psi_01_general-mhcb}, or, equivalently, $\Gamma(t) = t$ in Theorem~\ref{Thm:excess_bounds_Gamma_01_general-mhcb}, we obtain for any hypothesis $h\in\sH$ and any $\sH$-realizable distribution,
\begin{align*}
 \sR_{\ell_{0-1}}(h)-\sR_{\ell_{0-1}}^*(\sH)\leq \sR_{\Phi^{\mathrm{max}}}(h)-\sR_{\Phi^{\mathrm{max}}}^*(\sH)+\sM_{\Phi^{\mathrm{max}}}(\sH)
\end{align*}
\end{proof}
\section{Proof of \texorpdfstring{$\sH_{\mathrm{all}},\sH_{\mathrm{lin}},\sH_{\mathrm{NN}}$}{H}-consistency bounds for max \texorpdfstring{$\rho$}{rho}-margin loss \texorpdfstring{$ \Phi_{\rho}^{\mathrm{max}}$}{maxrho}}
\label{app:bound_max_rho}
\begin{restatable}[\textbf{$\sH_{\mathrm{all}}$-consistency bound of $\Phi_{\rho}^{\mathrm{max}}$}]
  {corollary}{BoundMaxRhoAll}
\label{cor:bound_max_rho_all}
For any hypothesis $h\in\sH_{\mathrm{all}}$ and any distribution,
\begin{align}
\label{eq:bound_max_rho_all}
     \sR_{\ell_{0-1}}(h)-\sR_{\ell_{0-1}}^*\paren*{\sH_{\mathrm{all}}}\leq\sR_{\Phi_{\rho}^{\mathrm{max}}}(h)-\sR_{\Phi_{\rho}^{\mathrm{max}}}^*\paren*{\sH_{\mathrm{all}}}.
\end{align}
\end{restatable}
\begin{proof}
For $\sH=\sH_{\mathrm{all}}$, we have for all $x\in \sX$, $\sup_{h\in\sH_{\mathrm{all}}}\rho_h(x,\hh(x))>\rho$. Furthermore,  as shown by \citet[Theorem~3.2]{steinwart2007compare}, the minimizability gaps $\sM_{\ell_{0-1}}\paren*{\sH_{\mathrm{all}}}=\sM_{\Phi_{\rho}^{\mathrm{max}}}\paren*{\sH_{\mathrm{all}}}=0$. Therefore, by Theorem~\ref{Thm:bound_max_rho}, the $\sH_{\mathrm{all}}$-consistency bound of $\Phi_{\rho}^{\mathrm{max}}$ can be expressed as follows:
\begin{align*}
\sR_{\ell_{0-1}}(h)-\sR_{\ell_{0-1}}^*\paren*{\sH_{\mathrm{all}}}\leq\sR_{\Phi_{\rho}^{\mathrm{max}}}(h)-\sR_{\Phi_{\rho}^{\mathrm{max}}}^*\paren*{\sH_{\mathrm{all}}}.   
\end{align*}
\end{proof}

\begin{restatable}[\textbf{$\sH_{\mathrm{lin}}$-consistency bound of $\Phi_{\rho}^{\mathrm{max}}$}]
  {corollary}{BoundMaxRhoLin}
\label{cor:bound_max_rho_lin}
For any hypothesis $h\in\sH_{\mathrm{lin}}$ and any distribution,
\begin{align}
\label{eq:bound_max_rho_lin}
     \sR_{\ell_{0-1}}(h)-\sR_{\ell_{0-1}}^*\paren*{\sH_{\mathrm{lin}}}\leq\frac{\sR_{\Phi_{\rho}^{\mathrm{max}}}(h)-\sR_{\Phi_{\rho}^{\mathrm{max}}}^*\paren*{\sH_{\mathrm{lin}}}+\sM_{\Phi_{\rho}^{\mathrm{max}}}\paren*{\sH_{\mathrm{lin}}}}{\min\curl*{1,\frac{2B}{\rho}}}-\sM_{\ell_{0-1}}\paren*{\sH_{\mathrm{lin}}},
\end{align}
where $\sM_{\ell_{0-1}}\paren*{\sH_{\mathrm{lin}}}=\sR^*_{\ell_{0-1}}\paren*{\sH_{\mathrm{lin}}}-\mathbb{E}_{X}\bracket*{1-\max_{y\in \sY}\sfp(y \!\mid\! x)}$ and $\sM_{\Phi_{\rho}^{\mathrm{max}}}\paren*{\sH_{\mathrm{lin}}}=\sR^*_{\Phi_{\rho}^{\mathrm{max}}}\paren*{\sH_{\mathrm{lin}}}-\mathbb{E}_{X}\bracket*{1-\min\curl*{1,\frac{2 \paren*{W\norm*{x}_p + B}}{\rho}}\max_{y\in \sY}\sfp(y \!\mid\! x)}$.
\end{restatable}
\begin{proof}
For $\sH=\sH_{\mathrm{lin}}$, we have for all $x\in \sX$,
\begin{equation}
\label{eq:margin_max_rho_lin}
\begin{aligned}
\sup_{h\in\sH_{\mathrm{lin}}}\rho_h(x,\hh(x)) & = \sup_{h\in\sH_{\mathrm{lin}}}\paren*{\max_{y\in \sY}h(x,y)-\max_{y\neq \hh(x)} h(x,y)}\\
& = \max_{\norm*{w}_q\leq W,\abs*{b}\leq B} \paren*{w\cdot x +b} - \min_{\norm*{w}_q\leq W,\abs*{b}\leq B} \paren*{w\cdot x +b}\\
& = 2 \paren*{W\norm*{x}_p + B}
\end{aligned}
\end{equation}
Thus, $\inf_{x\in \sX}\sup_{h\in\sH_{\mathrm{lin}}}\rho_h(x,\hh(x))=\inf_{x\in \sX} 2 \paren*{W\norm*{x}_p + B} = 2B$. Since $\sH=\sH_{\mathrm{lin}}$ is symmetric, by lemma~\ref{lemma:explicit_assumption_01-mhcb}, we have
\begin{align}
\label{eq:gap_01_lin}
\sM_{\ell_{0-1}}\paren*{\sH_{\mathrm{lin}}}=\sR^*_{\ell_{0-1}}\paren*{\sH_{\mathrm{lin}}}-\mathbb{E}_{X}\bracket*{1-\max_{y\in \sY}\sfp(y \!\mid\! x)}.
\end{align}
By the definition, the conditional $\Phi_{\rho}^{\mathrm{max}}$-risk can be expressed as follows:
\begin{equation*}
\begin{aligned}
\sC_{\Phi_{\rho}^{\mathrm{max}}}(h,x)  &=  \sum_{y\in \sY} \sfp(y \!\mid\! x) \Phi_{\rho}\paren*{\rho_h(x, y)}\\
& =
1-\sfp(\hh(x) \!\mid\! x)+ \max\curl*{0,1-\frac{\rho_h(x, \hh(x))}{\rho}}\,\sfp(\hh(x) \!\mid\! x) \\
& =
1- \min\curl*{1,\frac{\rho_h(x, \hh(x))}{\rho}}\,\sfp(\hh(x) \!\mid\! x)
\end{aligned}
\end{equation*}
Since $\sH_{\mathrm{lin}}$ is symmetric, for any $x\in \sX$ and $y\in \sY$, 
\begin{align*}
\sup_{h\in \curl*{h\in\sH_{\mathrm{lin}}: \hh(x)=y}}\rho_h(x,\hh(x))=\sup_{h\in \sH_{\mathrm{lin}}}\rho_h(x,\hh(x)).   
\end{align*}
Thus, using \eqref{eq:margin_max_rho_lin}, the minimal conditional $\Phi_{\rho}^{\mathrm{max}}$-risk can be expressed as follows:
\begin{align*}
\sC^*_{\Phi_{\rho}^{\mathrm{max}}}\paren*{\sH_{\mathrm{lin}}, x} & =1-\min\curl*{1,\frac{\sup_{h\in \sH_{\mathrm{lin}}}\rho_h(x,\hh(x))}{\rho}}\max_{y\in \sY}\sfp(y \!\mid\! x)\\
& = 1-\min\curl*{1,\frac{2 \paren*{W\norm*{x}_p + B}}{\rho}}\max_{y\in \sY}\sfp(y \!\mid\! x) & \paren*{\text{by } \eqref{eq:margin_max_rho_lin}}
\end{align*}
Therefore, the $\paren*{\Phi_{\rho}^{\mathrm{max}}, \sH_{\mathrm{lin}}}$-minimizability gap is
\begin{align}
\label{eq:gap_max_rho_lin}
\sM_{\Phi_{\rho}^{\mathrm{max}}}\paren*{\sH_{\mathrm{lin}}}=\sR^*_{\Phi_{\rho}^{\mathrm{max}}}\paren*{\sH_{\mathrm{lin}}}-\mathbb{E}_{X}\bracket*{1-\min\curl*{1,\frac{2 \paren*{W\norm*{x}_p + B}}{\rho}}\max_{y\in \sY}\sfp(y \!\mid\! x)}.
\end{align}
By Theorem~\ref{Thm:bound_max_rho}, the $\sH_{\mathrm{lin}}$-consistency  bound of $\Phi_{\rho}^{\mathrm{max}}$ can be expressed as follows:
\begin{align*}
     \sR_{\ell_{0-1}}(h)-\sR_{\ell_{0-1}}^*\paren*{\sH_{\mathrm{lin}}}\leq\frac{\sR_{\Phi_{\rho}^{\mathrm{max}}}(h)-\sR_{\Phi_{\rho}^{\mathrm{max}}}^*\paren*{\sH_{\mathrm{lin}}}+\sM_{\Phi_{\rho}^{\mathrm{max}}}\paren*{\sH_{\mathrm{lin}}}}{\min\curl*{1,\frac{2B}{\rho}}}-\sM_{\ell_{0-1}}\paren*{\sH_{\mathrm{lin}}}.
\end{align*}
where $\sM_{\ell_{0-1}}\paren*{\sH_{\mathrm{lin}}}$ and $\sM_{\Phi_{\rho}^{\mathrm{max}}}\paren*{\sH_{\mathrm{lin}}}$ are given by \eqref{eq:gap_01_lin} and \eqref{eq:gap_max_rho_lin} respectively.
\end{proof}

\begin{restatable}[\textbf{$\sH_{\mathrm{NN}}$-consistency bound of $\Phi_{\rho}^{\mathrm{max}}$}]
  {corollary}{BoundMaxRhoNN}
\label{cor:bound_max_rho_NN}
For any hypothesis $h\in\sH_{\mathrm{NN}}$ and any distribution,
\begin{align}
\label{eq:bound_max_rho_NN}
     \sR_{\ell_{0-1}}(h)-\sR_{\ell_{0-1}}^*\paren*{\sH_{\mathrm{NN}}}\leq\frac{\sR_{\Phi_{\rho}^{\mathrm{max}}}(h)-\sR_{\Phi_{\rho}^{\mathrm{max}}}^*\paren*{\sH_{\mathrm{NN}}}+\sM_{\Phi_{\rho}^{\mathrm{max}}}\paren*{\sH_{\mathrm{NN}}}}{\min\curl*{1,\frac{2\Lambda B}{\rho}}}-\sM_{\ell_{0-1}}\paren*{\sH_{\mathrm{NN}}},
\end{align}
where $\sM_{\ell_{0-1}}\paren*{\sH_{\mathrm{NN}}}=\sR^*_{\ell_{0-1}}\paren*{\sH_{\mathrm{NN}}}-\mathbb{E}_{X}\bracket*{1-\max_{y\in \sY}\sfp(y \!\mid\! x)}$ and $\sM_{\Phi_{\rho}^{\mathrm{max}}}\paren*{\sH_{\mathrm{NN}}}=\sR^*_{\Phi_{\rho}^{\mathrm{max}}}\paren*{\sH_{\mathrm{NN}}}-\mathbb{E}_{X}\bracket*{1-\min\curl*{1,\frac{2 \Lambda\paren*{W\norm*{x}_p+B}}{\rho}}\max_{y\in \sY}\sfp(y \!\mid\! x)}$.
\end{restatable}
\begin{proof}
For $\sH=\sH_{\mathrm{NN}}$, we have for all $x\in \sX$,
\begin{equation}
\label{eq:margin_max_rho_NN}
\begin{aligned}
& \sup_{h\in\sH_{\mathrm{NN}}}\rho_h(x,\hh(x)) = \sup_{h\in\sH_{\mathrm{NN}}}\paren*{\max_{y\in \sY}h(x,y)-\max_{y\neq \hh(x)} h(x,y)}\\
& = \max_{\|u\|_{1}\leq
\Lambda,\|w_{j}\|_q\leq W, \abs*{b_j}\leq B} \paren*{\sum_{j =1}^n u_{j}(w_{j} \cdot x+b_j)_{+}} - \min_{\|u\|_{1}\leq
\Lambda,\|w_{j}\|_q\leq W, \abs*{b_j}\leq B} \paren*{\sum_{j =1}^n u_{j}(w_{j} \cdot x+b_j)_{+}}\\
& = 2 \Lambda\paren*{W\norm*{x}_p+B}
\end{aligned}
\end{equation}
Thus, $\inf_{x\in \sX}\sup_{h\in\sH_{\mathrm{NN}}}\rho_h(x,\hh(x))=\inf_{x\in \sX} 2 \Lambda\paren*{W\norm*{x}_p+B} = 2\Lambda B$. Since $\sH=\sH_{\mathrm{NN}}$ is symmetric, by lemma~\ref{lemma:explicit_assumption_01-mhcb}, we have
\begin{align}
\label{eq:gap_01_NN}
\sM_{\ell_{0-1}}\paren*{\sH_{\mathrm{NN}}}=\sR^*_{\ell_{0-1}}\paren*{\sH_{\mathrm{NN}}}-\mathbb{E}_{X}\bracket*{1-\max_{y\in \sY}\sfp(y \!\mid\! x)}.
\end{align}
By the definition, the conditional $\Phi_{\rho}^{\mathrm{max}}$-risk can be expressed as follows:
\begin{equation*}
\begin{aligned}
\sC_{\Phi_{\rho}^{\mathrm{max}}}(h,x)  &=  \sum_{y\in \sY} \sfp(y \!\mid\! x) \Phi_{\rho}\paren*{\rho_h(x, y)}\\
& =
1-\sfp(\hh(x) \!\mid\! x)+ \max\curl*{0,1-\frac{\rho_h(x, \hh(x))}{\rho}}\,\sfp(\hh(x) \!\mid\! x) \\
& =
1- \min\curl*{1,\frac{\rho_h(x, \hh(x))}{\rho}}\,\sfp(\hh(x) \!\mid\! x)
\end{aligned}
\end{equation*}
Since $\sH_{\mathrm{NN}}$ is symmetric, for any $x\in \sX$ and $y\in \sY$, 
\begin{align*}
\sup_{h\in \curl*{h\in\sH_{\mathrm{NN}}: \hh(x)=y}}\rho_h(x,\hh(x))=\sup_{h\in \sH_{\mathrm{NN}}}\rho_h(x,\hh(x)).   
\end{align*}
Thus, using \eqref{eq:margin_max_rho_NN}, the minimal conditional $\Phi_{\rho}^{\mathrm{max}}$-risk can be expressed as follows:
\begin{align*}
\sC^*_{\Phi_{\rho}^{\mathrm{max}}}\paren*{\sH_{\mathrm{NN}}, x} & =1-\min\curl*{1,\frac{\sup_{h\in \sH_{\mathrm{NN}}}\rho_h(x,\hh(x))}{\rho}}\max_{y\in \sY}\sfp(y \!\mid\! x)\\
& = 1-\min\curl*{1,\frac{2 \Lambda\paren*{W\norm*{x}_p+B}}{\rho}}\max_{y\in \sY}\sfp(y \!\mid\! x) & \paren*{\text{by } \eqref{eq:margin_max_rho_NN}}
\end{align*}
Therefore, the $\paren*{\Phi_{\rho}^{\mathrm{max}}, \sH_{\mathrm{NN}}}$-minimizability gap is
\begin{align}
\label{eq:gap_max_rho_NN}
\sM_{\Phi_{\rho}^{\mathrm{max}}}\paren*{\sH_{\mathrm{NN}}}=\sR^*_{\Phi_{\rho}^{\mathrm{max}}}\paren*{\sH_{\mathrm{NN}}}-\mathbb{E}_{X}\bracket*{1-\min\curl*{1,\frac{2 \Lambda\paren*{W\norm*{x}_p+B}}{\rho}}\max_{y\in \sY}\sfp(y \!\mid\! x)}.
\end{align}
By Theorem~\ref{Thm:bound_max_rho}, the $\sH_{\mathrm{NN}}$-consistency  bound of $\Phi_{\rho}^{\mathrm{max}}$ can be expressed as follows:
\begin{align*}
     \sR_{\ell_{0-1}}(h)-\sR_{\ell_{0-1}}^*\paren*{\sH_{\mathrm{NN}}}\leq\frac{\sR_{\Phi_{\rho}^{\mathrm{max}}}(h)-\sR_{\Phi_{\rho}^{\mathrm{max}}}^*\paren*{\sH_{\mathrm{NN}}}+\sM_{\Phi_{\rho}^{\mathrm{max}}}\paren*{\sH_{\mathrm{NN}}}}{\min\curl*{1,\frac{2\Lambda B}{\rho}}}-\sM_{\ell_{0-1}}\paren*{\sH_{\mathrm{NN}}}.
\end{align*}
where $\sM_{\ell_{0-1}}\paren*{\sH_{\mathrm{NN}}}$ and $\sM_{\Phi_{\rho}^{\mathrm{max}}}\paren*{\sH_{\mathrm{NN}}}$
are given by \eqref{eq:gap_01_NN} and \eqref{eq:gap_max_rho_NN} respectively.
\end{proof}

\section{Auxiliary Lemma for sum losses}
\label{app:deferred_proofs_sum_auxiliary}
\begin{restatable}{lemma}{SumAuxiliary}
\label{lemma:sum_auxiliary}
Fix a vector $\tau=(\tau_1,\ldots,\tau_c)$ in the probability simplex of $\Rset^c$ and any real values $a_1\leq a_2 \leq \cdots \leq a_c$ in increasing order. Then, for any permutation $\sigma$ of the set $\curl*{1, \ldots, c}$,
\begin{align*}
\begin{bmatrix}
a_1\\
a_2\\
\vdots\\
a_c
\end{bmatrix}
\cdot
\begin{bmatrix}
\tau_{\sigma(1)}\\
\tau_{\sigma(2)}\\
\vdots\\
\tau_{\sigma(c)}
\end{bmatrix}
\leq 
\begin{bmatrix}
a_1\\
a_2\\
\vdots\\
a_c
\end{bmatrix}
\cdot
\begin{bmatrix}
\tau_{\bracket*{1}}\\
\tau_{\bracket*{2}}\\
\vdots\\
\tau_{\bracket*{c}}
\end{bmatrix},
\end{align*}
where we define $\tau_{\bracket*{1}}, \tau_{\bracket*{2}}, \ldots, \tau_{\bracket*{c}}$ by sorting the probabilities  $\curl*{\tau_y:y\in \curl*{1, \ldots, c}}$ in increasing order.
\end{restatable}
\begin{proof}
For any permutation $\sigma$ of the set $\curl*{1, \ldots, c}$, we prove by induction. At the first step, if $\sigma(c)=\bracket*{c}$, then let $\sigma_1=\sigma$. Otherwise, denote $k_1\in \curl*{1,\ldots,c-1}$ such that $\sigma(k_1)=\bracket*{c}$ and choose $\sigma_1$ to be the permutation that differs from $\sigma$ only by permuting $c$ and $k_1$. Thus,
\begin{align*}
\begin{bmatrix}
a_1\\
a_2\\
\vdots\\
a_c
\end{bmatrix}
\cdot
\begin{bmatrix}
\tau_{\sigma(1)}\\
\tau_{\sigma(2)}\\
\vdots\\
\tau_{\sigma(c)}
\end{bmatrix}
-
\begin{bmatrix}
a_1\\
a_2\\
\vdots\\
a_c
\end{bmatrix}
\cdot
\begin{bmatrix}
\tau_{\sigma_1(1)}\\
\tau_{\sigma_1(2)}\\
\vdots\\
\tau_{\sigma_1(c)}
\end{bmatrix}
&=a_{k_1}\tau_{\bracket*{c}}+a_{c}\tau_{\sigma(c)}-\paren*{a_{k_1}\tau_{\sigma(c)}+a_{c}\tau_{\bracket*{c}}}\\
&=\paren*{a_{k_1}-a_{c}}\paren*{\tau_{\bracket*{c}}-\tau_{\sigma(c)}}\leq 0.
\end{align*}
At the second step, if $\sigma_1(c-1)=\bracket*{c-1}$, then let $\sigma_2=\sigma_1$. Otherwise, denote $k_2\in \curl*{1,\ldots,c-2}$ such that $\sigma_1(k_2)=\bracket*{c-1}$ and choose $\sigma_2$ to be the permutation that differs from $\sigma_1$ only by permuting $c-1$ and $k_2$. Thus,
\begin{align*}
\begin{bmatrix}
a_1\\
a_2\\
\vdots\\
a_c
\end{bmatrix}
\cdot
\begin{bmatrix}
\tau_{\sigma_1(1)}\\
\tau_{\sigma_1(2)}\\
\vdots\\
\tau_{\sigma_1(c)}
\end{bmatrix}
-
\begin{bmatrix}
a_1\\
a_2\\
\vdots\\
a_c
\end{bmatrix}
\cdot
\begin{bmatrix}
\tau_{\sigma_2(1)}\\
\tau_{\sigma_2(2)}\\
\vdots\\
\tau_{\sigma_2(c)}
\end{bmatrix}
=\paren*{a_{k_2}-a_{c-1}}\paren*{\tau_{\bracket*{c-1}}-\tau_{\sigma_1(c-1)}}\leq 0.
\end{align*}
And so on, at the $n$th step, if $\sigma_{n-1}(c-n+1)=\bracket*{c-n+1}$, then let $\sigma_{n}=\sigma_{n-1}$. Otherwise, denote $k_n\in \curl*{1,\ldots,c-n}$ such that $\sigma_{n-1}(k_n)=\bracket*{c-n+1}$ and choose $\sigma_{n}$ to be the permutation that differs from $\sigma_{n-1}$ only by permuting $c-n+1$ and $k_n$. We have 
\begin{align*}
 \begin{bmatrix}
a_1\\
a_2\\
\vdots\\
a_c
\end{bmatrix}
\cdot
\begin{bmatrix}
\tau_{\sigma_{n-1}(1)}\\
\tau_{\sigma_{n-1}(2)}\\
\vdots\\
\tau_{\sigma_{n-1}(c)}
\end{bmatrix}
\leq
\begin{bmatrix}
a_1\\
a_2\\
\vdots\\
a_c
\end{bmatrix}
\cdot
\begin{bmatrix}
\tau_{\sigma_{n}(1)}\\
\tau_{\sigma_{n}(2)}\\
\vdots\\
\tau_{\sigma_{n}(c)}
\end{bmatrix}.
\end{align*}
Finally, after $c$ steps, we will obtain $\sigma_{c}$ which satisfies
$\sigma_{c}(y)=\bracket*{y}$ for any $y\in \curl*{1,\ldots,c}$. Therefore, we obtain
\begin{align*}
 \begin{bmatrix}
a_1\\
a_2\\
\vdots\\
a_c
\end{bmatrix}
\cdot
\begin{bmatrix}
\tau_{\sigma(1)}\\
\tau_{\sigma(2)}\\
\vdots\\
\tau_{\sigma(c)}
\end{bmatrix}
\leq
\begin{bmatrix}
a_1\\
a_2\\
\vdots\\
a_c
\end{bmatrix}
\begin{bmatrix}
\tau_{\sigma_1(1)}\\
\tau_{\sigma_1(2)}\\
\vdots\\
\tau_{\sigma_1(c)}
\end{bmatrix}
\leq 
\ldots 
\leq 
\begin{bmatrix}
a_1\\
a_2\\
\vdots\\
a_c
\end{bmatrix}
\begin{bmatrix}
\tau_{\sigma_n(1)}\\
\tau_{\sigma_n(2)}\\
\vdots\\
\tau_{\sigma_n(c)}
\end{bmatrix}
\leq
\ldots
\leq 
\begin{bmatrix}
a_1\\
a_2\\
\vdots\\
a_c
\end{bmatrix}
\cdot
\begin{bmatrix}
\tau_{\bracket*{1}}\\
\tau_{\bracket*{2}}\\
\vdots\\
\tau_{\bracket*{c}}
\end{bmatrix}
\end{align*}
which proves the lemma.
\end{proof}

\section{Proof of negative and \texorpdfstring{$\sH$}{H}-consistency bounds for sum losses \texorpdfstring{$ \Phi^{\mathrm{sum}}$}{sum}}
\label{app:deferred_proofs_sum}
By the definition, the conditional $\Phi^{\mathrm{sum}}$-risk can be expressed as follows:
\begin{equation}
\label{eq:cond_sum}
\begin{aligned}
\sC_{\Phi^{\mathrm{sum}}}(h,x) 
& =  \sum_{y\in \sY} \sfp(y \!\mid\! x) \sum_{y'\neq y} \Phi\paren*{h(x,y)-h(x,y')}\\
& = \sum_{y\in \sY} \sfp(y \!\mid\! x) \sum_{y'\in \sY} \Phi\paren*{h(x,y)-h(x,y')} - \Phi(0)
\end{aligned}
\end{equation}

\NegativeSum*
\begin{proof}
Consider the distribution that supports on a singleton domain $\curl*{x}$. We define $p(x)$ as $\sfp(1 \!\mid\! x)=\frac12-\e$, $\sfp(2 \!\mid\! x)=\frac13$, $\sfp(3 \!\mid\! x)=\frac{1}{6}+\e$ and $\sfp(y \!\mid\! x)=0$ for other $y\in \sY$, where $0< \e < \frac{1}{6}$. Note $\sfp(1 \!\mid\! x)>\sfp(2 \!\mid\! x)>\sfp(3 \!\mid\! x)>\sfp(y \!\mid\! x)=0$, $y\not \in \curl*{1, 2, 3}$. Let $h_0\in \sH$ such that $h_0(x,1)=1$, $h_0(x,2)=1$, $h_0(x,3)=0$ and $h_0(x,y)=-1$ for other $y\in \sY$. By the completeness of $\sH$, the hypothesis $h$ is in $\sH$ . By Lemma~\ref{lemma:explicit_assumption_01-mhcb} and the fact that $\mathsf H(x)=\sY$ when $\sH$ is symmetric, the minimal conditional $\ell_{0-1}$-risk is
\begin{align*}
\sR^*_{\ell_{0-1}}(\sH)=
\sC^*_{\ell_{0-1}}(\sH, x)=1-\max_{y\in \sY}\sfp(y \!\mid\! x)=1-\sfp(1 \!\mid\! x)=\frac12+\e.
\end{align*}
For $h=h_0$, we have
\begin{align*}
\sR_{\ell_{0-1}}(h_0)=\sC_{\ell_{0-1}}(h_0,x)=\sum_{y\in \sY} \sfp(y \!\mid\! x) \mathds{1}_{\hh_0(x)\neq y}=
1-\sfp(\hh_0(x) \!\mid\! x)=1-\sfp(2 \!\mid\! x)=\frac23.
\end{align*} 
For the sum hinge loss, by \eqref{eq:cond_sum}, the conditional $\Phi_{\mathrm{hinge}}^{\mathrm{sum}}$-risk can be expressed as follows:
\begin{align*}
\sC_{\Phi_{\mathrm{hinge}}^{\mathrm{sum}}}(h,x)
& = \sum_{y\in \sY} \sfp(y \!\mid\! x) \sum_{y'\neq y}\max\curl*{0,1+h(x,y')-h(x,y)}\\
& = \sum_{y\in \curl*{1,2,3}} \sfp(y \!\mid\! x) \sum_{y'\neq y}\max\curl*{0,1+h(x,y')-h(x,y)}\\
& \geq  \sum_{y\in \curl*{1,2,3}} \sfp(y \!\mid\! x) \sum_{y'\neq y, y'\in \curl*{1,2,3}}\max\curl*{0,1+h(x,y')-h(x,y)}\\
& = \paren*{\frac12-\e}\bracket*{\max\curl*{0,1+h(x,2)-h(x,1)}+\max\curl*{0,1+h(x,3)-h(x,1)}}\\
& +\frac13\bracket*{\max\curl*{0,1+h(x,1)-h(x,2)}+\max\curl*{0,1+h(x,3)-h(x,2)}}\\
& + \paren*{\frac16+\e}\bracket*{\max\curl*{0,1+h(x,1)-h(x,3)}+\max\curl*{0,1+h(x,2)-h(x,3)}}\\
& = g(h).
\end{align*}
Note $\sC_{\Phi_{\mathrm{hinge}}^{\mathrm{sum}}}(h_0,x)=3\e+\frac32$. Since $\frac12-\e>\frac13>\frac16+\e$, by Lemma~\ref{lemma:sum_auxiliary}, we have
\begin{align*}
\inf_{h\in \sH} g(h) = \inf_{h\in \sH:\, h(x,1)\geq h(x,2)\geq h(x,3)} g(h).
\end{align*}
When $h(x,1)\geq h(x,2)\geq h(x,3)$, $g(h)$ can be written as
\begin{align*}
g(h) & = \paren*{\frac12-\e}\bracket*{\max\curl*{0,1+h(x,2)-h(x,1)}+\max\curl*{0,1+h(x,3)-h(x,1)}}\\
& +\frac13\bracket*{\paren*{1+h(x,1)-h(x,2)}+\max\curl*{0,1+h(x,3)-h(x,2)}}\\
& + \paren*{\frac16+\e}\bracket*{\paren*{1+h(x,1)-h(x,3)}+\paren*{1+h(x,2)-h(x,3)}}\\
\end{align*}
If $h(x,1)-h(x,2)> 1$, define the hypothesis $\ov h\in \sH$ by
\begin{align*}
\ov h(x,y)=
\begin{cases}
h(x,1)-\frac{h(x,1)-h(x,2)-1}{2}, & \text{if } y=1\\
h(x,y) & \text{otherwise}.
\end{cases}
\end{align*}
By the completeness of $\sH$ and some computation, the new hypothesis $\ov h$ is in $\sH$ and satisfies that $g(\ov h)< g(h)$.
Similarly, if $h(x,2)-h(x,3)> 1$, define the hypothesis $\ov h\in \sH$ by
\begin{align*}
\ov h(x,y)=
\begin{cases}
h(x,2)-\frac{h(x,2)-h(x,3)-1}{2}, & \text{if } y=2\\
h(x,y) & \text{otherwise}.
\end{cases}
\end{align*}
By the completeness of $\sH$ and some computation, the new hypothesis $\ov h$ is in $\sH$ and satisfies that $g(\ov h)< g(h)$. Therefore,
\begin{align*}
\inf_{h\in \sH} g(h)  = \inf_{h\in \sH:\, h(x,1)\geq h(x,2)\geq h(x,3)} g(h)=\inf_{h\in \sH:\, h(x,1)\geq h(x,2)\geq h(x,3),\, h(x,1)-h(x,2)\leq 1,\, h(x_2)-h(x,3)\leq 1} g(h)
\end{align*}
When $h(x,1)\geq h(x,2)\geq h(x,3)$, $h(x,1)-h(x,2)\leq 1$ and $ h(x_2)-h(x,3)\leq 1$, $g(h)$ can be written as
\begin{align*}
g(h) & = \paren*{\frac12-\e}\bracket*{\paren*{1+h(x,2)-h(x,1)}+\max\curl*{0,1+h(x,3)-h(x,1)}}\\
& +\frac13\bracket*{\paren*{1+h(x,1)-h(x,2)}+\paren*{1+h(x,3)-h(x,2)}}\\
& + \paren*{\frac16+\e}\bracket*{\paren*{1+h(x,1)-h(x,3)}+\paren*{1+h(x,2)-h(x,3)}}\\
\end{align*}
If $h(x,1)-h(x,3)>1$, define the hypothesis $\ov h\in \sH$ by
\begin{align*}
\ov h(x,y)=
\begin{cases}
h(x,1)-\frac{h(x,1)-h(x,3)-1}{2}, & \text{if } y=1\\
h(x,y) & \text{otherwise}.
\end{cases}
\end{align*}
By the completeness of $\sH$ and some computation using the fact that $0<\e< \frac16$, the new hypothesis $\ov h$ is in $\sH$ and satisfies that $g(\ov h)< g(h)$. Therefore,
\begin{align*}
\inf_{h\in \sH} g(h) & = \inf_{h\in \sH:\, h(x,1)\geq h(x,2)\geq h(x,3),\, h(x,1)-h(x,2)\leq 1,\, h(x_2)-h(x,3)\leq 1,\,h(x,1)-h(x,3)\leq 1} g(h)\\
& = \inf_{h\in \sH:\, h(x,1)\geq h(x,2)\geq h(x,3),\, h(x,1)-h(x,2)\leq 1,\, h(x_2)-h(x,3)\leq 1,\,h(x,1)-h(x,3)\leq 1} \bracket[\bigg]{ \\
&\qquad  \paren*{3\e -\frac12}\paren*{h(x,1)-h(x,3)} + 2 }\\
& = 3\e +\frac 32.
\end{align*}
Thus, we obtain for any $h\in\sH$,
\begin{align*}
\sR_{\Phi_{\mathrm{hinge}}^{\mathrm{sum}}}(h)=\sC_{\Phi_{\mathrm{hinge}}^{\mathrm{sum}}}(h,x)
\geq g(h)\geq 3\e+\frac32 = \sC_{\Phi_{\mathrm{hinge}}^{\mathrm{sum}}}(h_0,x)
\end{align*}
Therefore,
\begin{align*}
\sR_{\Phi_{\mathrm{hinge}}^{\mathrm{sum}}}^*(\sH)=\sC^*_{\Phi_{\mathrm{hinge}}^{\mathrm{sum}}}(\sH, x)=\sR_{\Phi_{\mathrm{hinge}}^{\mathrm{sum}}}(h_0)=3\e+\frac32.
\end{align*}
If \eqref{eq:bound_sum_hinge} holds for some non-decreasing function $f$, then, we obtain for any $h\in \sH$,
\begin{align*}
\sR_{\ell_{0-1}}(h)-\frac12-\e \leq  f\paren*{\sR_{\Phi_{\mathrm{hinge}}^{\mathrm{sum}}}(h) - \sR_{\Phi_{\mathrm{hinge}}^{\mathrm{sum}}}(h_0)}.
\end{align*}
Let $h=h_0$, then $f(0)\geq 1/6-\e$. Since $f$ is non-decreasing, for any $t\geq 0$ and $0<\e<\frac16$, $f(t)\geq 1/6-\e$. Let $\e \to 0$, we obtain that $f$ is lower bounded by $\frac16$.
\end{proof}

\begin{restatable}[\textbf{$\sH$-consistency bound of $\Phi_{\mathrm{sq-hinge}}^{\mathrm{sum}}$}]
  {theorem}{BoundSumSqHinge}
\label{Thm:bound_sum_sq-hinge}
Suppose that $\sH$ is symmetric and complete. Then, for any hypothesis $h\in\sH$ and any distribution,
\begin{align}
\label{eq:bound_sum_sq-hinge}
     \sR_{\ell_{0-1}}(h)-\sR_{\ell_{0-1}}^*(\sH)\leq\paren*{\sR_{\Phi_{\mathrm{sq-hinge}}^{\mathrm{sum}}}(h)-\sR_{\Phi_{\mathrm{sq-hinge}}^{\mathrm{sum}}}^*(\sH)+\sM_{\Phi_{\mathrm{sq-hinge}}^{\mathrm{sum}}}(\sH)}^{\frac12}-\sM_{\ell_{0-1}}(\sH).
\end{align}
\end{restatable}
\begin{proof}
For the sum squared hinge loss $\Phi_{\mathrm{sq-hinge}}^{\mathrm{sum}}$, by \eqref{eq:cond_sum}, we have
\begin{equation*}
\label{eq:cond_sum_sq-hinge}
\begin{aligned}
\sC_{\Phi_{\mathrm{sq-hinge}}^{\mathrm{sum}}}(h,x)
& = \sum_{y\in \sY} \sfp(y \!\mid\! x) \sum_{y'\neq y}\max\curl*{0,1+h(x,y')-h(x,y)}^2 \\
& = \sfp(y_{\max} \!\mid\! x) \sum_{y'\neq y_{\max}} \max\curl*{0,1+h(x,y')-h(x,y)}^2\\
&\qquad +\sum_{y\neq y_{\max}}\sfp(y \!\mid\! x) \sum_{y'\neq y} \max\curl*{0,1+h(x,y')-h(x,y)}^2\\
& = \sfp(y_{\max} \!\mid\! x) \sum_{y'\neq y_{\max}} \max\curl*{0,1+h(x,y')-h(x,y_{\max})}^2\\
& \qquad +\sum_{y\neq y_{\max}}\sfp(y \!\mid\! x) \max\curl*{0,1+h(x,y_{\max})-h(x,y)}^2\\
& \qquad \quad + \sum_{y\neq y_{\max}}\sfp(y \!\mid\! x)\sum_{y'\not \in \curl*{y_{\max}, y}} \max\curl*{0,1+h(x,y')-h(x,y)}^2
\end{aligned}
\end{equation*}
For any $h\in \sH$, define the hypothesis $\ov h_{\lambda} \in \sH$ by
\begin{align*}
\ov h_{\lambda}(x,y) = 
\begin{cases}
  h(x, y) & \text{if $y \neq y_{\max}$}\\
  \lambda & \text{if $y = y_{\max}$}
\end{cases} 
\end{align*}
for any $\lambda \in \Rset$. By the completeness of $\sH$, the new hypothesis $\ov h_{\lambda}$ is in $\sH$.
Therefore, the minimal conditional $\Phi_{\mathrm{sq-hinge}}^{\mathrm{sum}}$-risk satisfies that for any $\lambda\in \Rset$, $\sC^*_{\Phi_{\mathrm{sq-hinge}}^{\mathrm{sum}}}(\sH, x)
 \leq  \sC_{\Phi_{\mathrm{sq-hinge}}^{\mathrm{sum}}}(\ov h_{\lambda}, x)$.
Let $h \in \sH$ be a
hypothesis such that $\hh(x) \neq y_{\max}$. By the definition and using the fact that $\mathsf H(x)=\sY$ when $\sH$ is symmetric, we obtain
\begin{align*}
&\Delta\sC_{\Phi_{\mathrm{sq-hinge}}^{\mathrm{sum}},\sH}(h,x)
= \sC_{\Phi_{\mathrm{sq-hinge}}^{\mathrm{sum}}}(h,x) - \sC^*_{\Phi_{\mathrm{sq-hinge}}^{\mathrm{sum}}}(\sH, x) \\
& \geq \sC_{\Phi_{\mathrm{sq-hinge}}^{\mathrm{sum}}}(h,x) - \sC_{\Phi_{\mathrm{sq-hinge}}^{\mathrm{sum}}}(\ov h_{\lambda},x)\\
& \geq \sfp(y_{\max} \!\mid\! x) \max\curl*{0,1+h(x,\hh(x))-h(x,y_{\max})}^2\\
&\qquad +\sfp(\hh(x) \!\mid\! x) \max\curl*{0,1+h(x,y_{\max})-h(x,\hh(x))}^2 - \frac{4\sfp(y_{\max} \!\mid\! x)\sfp(\hh(x) \!\mid\! x)}{\sfp(y_{\max} \!\mid\! x)+\sfp(\hh(x) \!\mid\! x)} \tag{taking supremum with respect to $\lambda$}\\
& \geq \sfp(y_{\max} \!\mid\! x) + \sfp(\hh(x) \!\mid\! x)  - \frac{4\sfp(y_{\max} \!\mid\! x)\sfp(\hh(x) \!\mid\! x)}{\sfp(y_{\max} \!\mid\! x)+\sfp(\hh(x) \!\mid\! x)} \tag{$h(x,\hh(x))- h(x,y_{\max})\geq 0$} \\
& = \frac{\paren*{\sfp(y_{\max} \!\mid\! x)-\sfp(\hh(x) \!\mid\! x)}^2}{\sfp(y_{\max} \!\mid\! x)+\sfp(\hh(x) \!\mid\! x)}\\
& \geq \paren*{\max_{y\in \sY}\sfp(y \!\mid\! x) - \sfp(\hh(x) \!\mid\! x)}^2 \tag{$0\leq \sfp(y_{\max} \!\mid\! x)+\sfp(\hh(x) \!\mid\! x)\leq 1$}\\
& = \paren*{ \Delta\sC_{\ell_{0-1},\sH}(h,x)}^2 \tag{by Lemma~\ref{lemma:explicit_assumption_01-mhcb} and $\mathsf H(x)=\sY$}\\
& \geq \paren*{\bracket*{\Delta\sC_{\ell_{0-1},\sH}(h,x)}_{\e}}^2 \tag{$\bracket*{t}_{\e}\leq t$}
\end{align*}
for any $\e\geq 0$. Therefore, taking $\sP$ be the set of all distributions, $\sH$ be the symmetric and complete hypothesis set, $\e=0$ and $\Psi(t)=t^2$ in Theorem~\ref{Thm:excess_bounds_Psi_01_general-mhcb}, or, equivalently, $\Gamma(t) = \sqrt{t}$ in Theorem~\ref{Thm:excess_bounds_Gamma_01_general-mhcb}, we obtain for any hypothesis $h\in\sH$ and any distribution,
\begin{align*}
\sR_{\ell_{0-1}}(h)-\sR_{\ell_{0-1}}^*(\sH)\leq \paren*{\sR_{\Phi_{\mathrm{sq-hinge}}^{\mathrm{sum}}}(h)-\sR_{\Phi_{\mathrm{sq-hinge}}^{\mathrm{sum}}}^*(\sH)+\sM_{\Phi_{\mathrm{sq-hinge}}^{\mathrm{sum}}}(\sH)}^{\frac12}-\sM_{\ell_{0-1}}(\sH).
\end{align*}
\end{proof}

\begin{restatable}[\textbf{$\sH$-consistency bound of $\Phi_{\mathrm{exp}}^{\mathrm{sum}}$}]
  {theorem}{BoundSumExp}
\label{Thm:bound_sum_exp}
Suppose that $\sH$ is symmetric and complete. Then, for any hypothesis $h\in\sH$ and any distribution,
\begin{align}
\label{eq:bound_sum_exp}
     \sR_{\ell_{0-1}}(h)-\sR_{\ell_{0-1}}^*(\sH)\leq \sqrt{2}\paren*{\sR_{\Phi_{\mathrm{exp}}^{\mathrm{sum}}}(h)-\sR_{\Phi_{\mathrm{exp}}^{\mathrm{sum}}}^*(\sH)+\sM_{\Phi_{\mathrm{exp}}^{\mathrm{sum}}}(\sH)}^{\frac12}-\sM_{\ell_{0-1}}(\sH).
\end{align}
\end{restatable}
\begin{proof}
For the sum exponential loss $\Phi_{\mathrm{exp}}^{\mathrm{sum}}$, by \eqref{eq:cond_sum}, the conditional $\Phi_{\mathrm{exp}}^{\mathrm{sum}}$-risk can be expressed as follows:
\begin{equation*}
\label{eq:cond_sum_exp}
\begin{aligned}
& \sC_{\Phi_{\mathrm{exp}}^{\mathrm{sum}}}(h,x)
= \sum_{y\in \sY} \sfp(y \!\mid\! x) \sum_{y'\neq y} \exp\paren*{h(x,y')-h(x,y)}\\
& = \sfp(y_{\max} \!\mid\! x) \sum_{y'\neq y_{\max}} \exp\paren*{h(x,y')-h(x,y_{\max})}+\sum_{y\neq y_{\max}}\sfp(y \!\mid\! x) \sum_{y'\neq y} \exp\paren*{h(x,y')-h(x,y)}\\
& = \sfp(y_{\max} \!\mid\! x) \sum_{y'\neq y_{\max}} \exp\paren*{h(x,y')-h(x,y_{\max})} + \sum_{y\neq y_{\max}}\sfp(y \!\mid\! x)\exp\paren*{h(x,y_{\max})-h(x,y)}\\
& + \sum_{y\neq y_{\max}}\sfp(y \!\mid\! x) \sum_{y'\not \in \curl*{y_{\max},y}} \exp\paren*{h(x,y')-h(x,y)}\\
\end{aligned}
\end{equation*}
For any $h \in \sH$, define the hypothesis $\ov h_{\lambda} \in \sH$ by
\begin{align*}
\ov h_{\lambda}(x,y) = 
\begin{cases}
  h(x, y) & \text{if $y \not \in \curl*{y_{\max}, \hh(x)}$}\\
  \log\paren*{\exp\bracket*{h(x, y_{\max})} + \lambda} & \text{if $y = \hh(x)$}\\
  \log\paren*{\exp\bracket*{h(x,\hh(x))} -\lambda} & \text{if $y = y_{\max}$}
\end{cases} 
\end{align*}
for any $\lambda \in \Rset$. By the completeness of $\sH$, the new hypothesis $\ov h_{\lambda}$ is in $\sH$.
Therefore, the minimal conditional $\Phi_{\mathrm{exp}}^{\mathrm{sum}}$-risk satisfies that for any $\lambda\in \Rset$, $\sC^*_{\Phi_{\mathrm{exp}}^{\mathrm{sum}}}(\sH, x)
 \leq  \sC_{\Phi_{\mathrm{exp}}^{\mathrm{sum}}}(\ov h_{\lambda}, x)$.
Let $h \in \sH$ be a
hypothesis such that $\hh(x) \neq y_{\max}$. By the definition and using the fact that $\mathsf H(x)=\sY$ when $\sH$ is symmetric, we obtain
\begin{align*}
&\Delta\sC_{\Phi_{\mathrm{exp}}^{\mathrm{sum}},\sH}(h,x)
= \sC_{\Phi_{\mathrm{exp}}^{\mathrm{sum}}}(h,x) - \sC^*_{\Phi_{\mathrm{exp}}^{\mathrm{sum}}}(\sH, x) \\
& \geq \sC_{\Phi_{\mathrm{exp}}^{\mathrm{sum}}}(h,x) - \sC_{\Phi_{\mathrm{exp}}^{\mathrm{sum}}}(\ov h_{\lambda},x) \\
& \geq \sum_{y'\in \sY} e^{h(x,y')} \bracket*{\sfp(y_{\max} \!\mid\! x)e^{-h(x,y_{\max})}+\sfp(\hh(x) \!\mid\! x) e^{-h(x,\hh(x))}-\frac{\paren*{\sqrt{\sfp(y_{\max} \!\mid\! x)}+\sqrt{\sfp(\hh(x) \!\mid\! x)}}^2}{e^{h(x,\hh(x))}+e^{h(x,y_{\max})}}}\tag{taking supremum with respect to $\lambda$}\\
& \geq \paren*{\sqrt{\sfp(y_{\max} \!\mid\! x)}-\sqrt{\sfp(\hh(x) \!\mid\! x)}}^2 
\tag{$h(x,\hh(x))\geq h(x,y_{\max})$ and $\sfp(\hh(x) \!\mid\! x)\leq \sfp(y_{\max} \!\mid\! x)$}\\
& = \paren*{\frac{\sfp(y_{\max} \!\mid\! x)-\sfp(\hh(x) \!\mid\! x)}{\sqrt{\sfp(\hh(x) \!\mid\! x)}+\sqrt{\sfp(y_{\max} \!\mid\! x)}}}^2\\
& \geq \frac12 \paren*{\max_{y\in \sY}\sfp(y \!\mid\! x) - \sfp(\hh(x) \!\mid\! x)}^2 \tag{$0\leq \sfp(y_{\max} \!\mid\! x)+\sfp(\hh(x) \!\mid\! x)\leq 1$}\\
& = \frac12 \paren*{ \Delta\sC_{\ell_{0-1},\sH}(h,x)}^2 \tag{by Lemma~\ref{lemma:explicit_assumption_01-mhcb} and $\mathsf H(x)=\sY$}\\
& \geq \frac12 \paren*{\bracket*{\Delta\sC_{\ell_{0-1},\sH}(h,x)}_{\e}}^2 \tag{$\bracket*{t}_{\e}\leq t$}
\end{align*}
for any $\e\geq 0$. Therefore, taking $\sP$ be the set of all distributions, $\sH$ be the symmetric and complete hypothesis set, $\e=0$ and $\Psi(t)=\frac{t^2}{2}$ in Theorem~\ref{Thm:excess_bounds_Psi_01_general-mhcb}, or, equivalently, $\Gamma(t) = \sqrt{2t}$ in Theorem~\ref{Thm:excess_bounds_Gamma_01_general-mhcb}, we obtain for any hypothesis $h\in\sH$ and any distribution,
\begin{align*}
\sR_{\ell_{0-1}}(h)-\sR_{\ell_{0-1}}^*(\sH)\leq \sqrt{2}\paren*{\sR_{\Phi_{\mathrm{exp}}^{\mathrm{sum}}}(h)-\sR_{\Phi_{\mathrm{exp}}^{\mathrm{sum}}}^*(\sH)+\sM_{\Phi_{\mathrm{exp}}^{\mathrm{sum}}}(\sH)}^{\frac12}-\sM_{\ell_{0-1}}(\sH).
\end{align*}
\end{proof}

\begin{restatable}[\textbf{$\sH$-consistency bound of $\Phi_{\rho}^{\mathrm{sum}}$}]{theorem}{BoundSumRho}
\label{Thm:bound_sum_rho}
Suppose that $\sH$ is symmetric and satisfies that for any $x\in \sX$, there exists a hypothesis $h \in \sH$ such that $\abs*{h(x,i)-h(x,j)}\geq \rho$ for any $i\neq j \in \sY$. Then, for any hypothesis $h\in\sH$ and any distribution,
\begin{align}
\label{eq:bound_sum_rho}
     \sR_{\ell_{0-1}}(h)- \sR_{\ell_{0-1}}^*(\sH) \leq   \sR_{\Phi_{\rho}^{\mathrm{sum}}}(h)-\sR_{\Phi_{\rho}^{\mathrm{sum}}}^*(\sH)+\sM_{\Phi_{\rho}^{\mathrm{sum}}}(\sH)-\sM_{\ell_{0-1}}(\sH).
\end{align}
\end{restatable}
\begin{proof}
For any $x\in \sX$, we define \[p_{\bracket*{1}}(x), p_{\bracket*{2}}(x), \ldots, p_{\bracket*{c}}(x)\] by sorting the probabilities  $\curl*{\sfp(y \!\mid\! x):y\in \sY}$ in increasing order. Similarly,
for any $x\in \sX$ and $h\in \sH$, we define $h\paren*{x, \curl*{1}_x}, h\paren*{x, \curl*{2}_x},\ldots, h\paren*{x, \curl*{c}_x}$ by sorting the scores $\curl*{h(x,y):y\in \sY}$ in increasing order. In particular, we have
\begin{align*}
h\paren*{x, \curl*{1}_x}=\min_{y\in \sY}h(x,y), \quad  h\paren*{x, \curl*{c}_x}=\max_{y\in \sY}h(x,y), \quad h\paren*{x, \curl*{i}_x}\leq h\paren*{x, \curl*{j}_j},\,\forall i\leq j.
\end{align*}
If there is a tie for the maximum, we pick the label with the highest index under the natural ordering of labels, i.e. $\curl*{c}_x=\hh(x)$.
By the definition, the conditional $\Phi_{\rho}^{\mathrm{sum}}$-risk can be expressed as follows:
\begin{align*}
& \sC_{\Phi_{\rho}^{\mathrm{sum}}}(h,x) =  \sum_{y\in \sY} \sfp(y \!\mid\! x) \sum_{y'\neq y} \Phi_{\rho}\paren*{h(x,y)-h(x,y')}\\
& =
\sum_{i=1}^c \sfp(\curl*{i}_{x} \!\mid\! x) \bracket*{\sum_{j=1}^{i-1}\Phi_{\rho}\paren*{h(x,\curl*{i}_{x})-h(x,\curl*{j}_{x})}+ \sum_{j=i+1}^{c}\Phi_{\rho}\paren*{h(x,\curl*{i}_{x})-h(x,\curl*{j}_{x})}}\\
& = \sum_{i=1}^c  \sfp(\curl*{i}_{x} \!\mid\! x) \bracket*{\sum_{j=1}^{i-1}\Phi_{\rho}\paren*{h(x,\curl*{i}_{x})-h(x,\curl*{j}_{x})} + c-i} \tag{$\Phi_{\rho}(t)=1$ for $t\leq 0$}
\end{align*}   
By the assumption, there exists a hypotheses $h \in \sH$ such that $\abs*{h(x,i)-h(x,j)}\geq \rho$ for any $i\neq j \in \sY$. Since $\sH$ is symmetric, we can always choose $h^*$ among these hypotheses such that $h^*$ and $p(x)$ induce the same ordering of the labels, i.e. $\sfp(\curl*{k}_{x} \!\mid\! x)=p_{\bracket*{k}}(x)$ for any $k\in \sY$. Then, we have
\begin{align*}
\sC^*_{\Phi_{\rho}^{\mathrm{sum}}}(\sH, x)
& \leq \sC_{\Phi_{\rho}^{\mathrm{sum}}}(h^*, x)\\
& =\sum_{i=1}^c \sfp(\curl*{i}_{x} \!\mid\! x) \bracket*{\sum_{j=1}^{i-1}\Phi_{\rho}\paren*{h^*(x,\curl*{i}_{x})-h^*(x,\curl*{j}_{x})} + c-i} \\
& = \sum_{i=1}^c  \sfp(\curl*{i}_{x} \!\mid\! x) (c-i) \tag{$\abs*{h^*(x,i)-h^*(x,j)}\geq \rho$ for any $i\neq j$ and $\Phi_{\rho}(t)=0,\,\forall t\geq \rho$}\\
& = \sum_{i=1}^c p_{\bracket*{i}}(x) (c-i) \tag{$h^*$ and $p(x)$ induce the same ordering of the labels}\\
& = c - \sum_{i=1}^c i\, p_{\bracket*{i}}(x) \tag{$\sum_{i=1}^c p_{\bracket*{i}}(x)=1$}
\end{align*}
By the definition and using the fact that $\mathsf H(x)=\sY$ when $\sH$ is symmetric, we obtain
\begin{align*}
& \Delta\sC_{\Phi_{\rho}^{\mathrm{sum}},\sH}(h,x)\\
& =  \sC_{\Phi_{\rho}^{\mathrm{sum}}}(h,x)-\sC^*_{\Phi_{\rho}^{\mathrm{sum}}}(\sH, x)\\
& =  \sum_{i=1}^c  \sfp(\curl*{i}_{x} \!\mid\! x) \bracket*{\sum_{j=1}^{i-1}\Phi_{\rho}\paren*{h(x,\curl*{i}_{x})-h(x,\curl*{j}_{x})} + c-i} - \paren*{c - \sum_{i=1}^c i\, p_{\bracket*{i}}(x)}\\
&\geq \sum_{i=1}^c \sfp(\curl*{i}_{x} \!\mid\! x) \paren*{c-i} - \paren*{c - \sum_{i=1}^c i\, p_{\bracket*{i}}(x)} \tag{$\Phi_{\rho}\geq 0$}\\
& = \sum_{i=1}^c i\, p_{\bracket*{i}}(x) - \sum_{i=1}^c i\, \sfp(\curl*{i}_{x} \!\mid\! x) \tag{$\sum_{i=1}^c \sfp(\curl*{i}\!\mid\! x)=1$}\\
& = \max_{y\in \sY}\sfp(y \!\mid\! x)-\sfp(\hh(x) \!\mid\! x) + \begin{bmatrix}
c-1\\
c-1\\
c-2\\
\vdots\\
1
\end{bmatrix}
\cdot
\begin{bmatrix}
p_{\bracket*{c}}(x)\\
p_{\bracket*{c-1}}(x)\\
p_{\bracket*{c-2}}(x)\\
\vdots\\
p_{\bracket*{1}}(x)\\
\end{bmatrix}
-
\begin{bmatrix}
c-1\\
c-1\\
c-2\\
\vdots\\
1
\end{bmatrix}
\cdot
\begin{bmatrix}
\sfp(\curl*{c}_{x} \!\mid\! x)\\
\sfp(\curl*{c-1}_{x} \!\mid\! x)\\
\sfp(\curl*{c-2}_{x} \!\mid\! x)\\
\vdots\\
\sfp(\curl*{1}_{x} \!\mid\! x)\\
\end{bmatrix}
\tag{$p_{[c]}(x)=\max_{y\in \sY}\sfp(y \!\mid\! x)$ and $\curl*{c}_x = \hh(x)$}\\
& \geq \max_{y\in \sY}\sfp(y \!\mid\! x)-\sfp(\hh(x) \!\mid\! x) \tag{by Lemma~\ref{lemma:sum_auxiliary}}\\
& = \Delta\sC_{\ell_{0-1},\sH}(h,x)\tag{by Lemma~\ref{lemma:explicit_assumption_01-mhcb}}\\
& \geq \bracket*{\Delta\sC_{\ell_{0-1},\sH}(h,x)}_{\e} \tag{$\bracket*{t}_{\e}\leq t$}
\end{align*}
for any $\e\geq 0$.
Therefore, taking $\sP$ be the set of all distributions, $\sH$ be the symmetric hypothesis set, $\e=0$ and
$\Psi(t)=t$ in Theorem~\ref{Thm:excess_bounds_Psi_adv_general}, or, equivalently, $\Gamma(t) = t$ in Theorem~\ref{Thm:excess_bounds_Gamma_01_general-mhcb}, we obtain for any hypothesis $h\in\sH$ and any distribution,
\begin{align*}
\sR_{\ell_{0-1}}(h)- \sR_{\ell_{0-1}}^*(\sH) \leq   \sR_{\Phi_{\rho}^{\mathrm{sum}}}(h)-\sR_{\Phi_{\rho}^{\mathrm{sum}}}^*(\sH)+\sM_{\Phi_{\rho}^{\mathrm{sum}}}(\sH)-\sM_{\ell_{0-1}}(\sH).
\end{align*}
\end{proof}

\section{Proof of \texorpdfstring{$\sH$}{H}-consistency bounds for constrained losses \texorpdfstring{$ \Phi^{\mathrm{cstnd}}$}{cstnd}}
\label{app:deferred_proofs_lee}
Recall that $\hh(x)$ and $y_{\max}$ are defined as follows: \[\hh(x)= \argmax_{y \in \sY} h(x, y) \quad \text{and} \quad y_{\max} = \argmax_{y \in \sY} \sfp(y \!\mid\! x).\] If there is a tie, we pick the label with the highest index under the natural ordering of labels.
The main idea of the proofs in this section is to leverage the constraint condition of \citet{lee2004multicategory} that the scores sum to zero, and appropriately choose a hypothesis $\ov h$ that differs from $h$ only for its scores for $\hh(x)$ and $y_{\max}$. Then, we can upper-bound the minimal conditional risk by the conditional risk of $\ov h$ without requiring complicated computation of the minimal conditional risk. By the definition, the conditional $\Phi^{\mathrm{cstnd}}$-risk can be expressed as follows:
\begin{equation}
\label{eq:cond_LLW}
\begin{aligned}
\sC_{\Phi^{\mathrm{cstnd}}}(h,x)  &=  \sum_{y\in \sY} \sfp(y \!\mid\! x) \sum_{y'\neq y} \Phi\paren*{-h(x,y')}\\
& = \sum_{y\in \sY}  \Phi\paren*{-h(x,y)} \sum_{y'\neq y} \sfp(y' \!\mid\! x) \\
& = \sum_{y\in \sY} \paren*{1-\sfp(y \!\mid\! x)}\Phi\paren*{-h(x,y)}
\end{aligned}
\end{equation}

\begin{restatable}[\textbf{$\sH$-consistency bound of $\Phi_{\mathrm{hinge}}^{\mathrm{cstnd}}$}]{theorem}{BoundLeeHinge}
\label{Thm:bound_lee_hinge}
Suppose that $\sH$ is symmetric and complete. Then, for any hypothesis $h\in\sH$ and any distribution,
\begin{align}
\label{eq:bound_lee_hinge}
     \sR_{\ell_{0-1}}(h)-\sR_{\ell_{0-1}}^*(\sH)\leq\sR_{\Phi_{\mathrm{hinge}}^{\mathrm{cstnd}}}(h)-\sR_{\Phi_{\mathrm{hinge}}^{\mathrm{cstnd}}}^*(\sH)+\sM_{\Phi_{\mathrm{hinge}}^{\mathrm{cstnd}}}(\sH)-\sM_{\ell_{0-1}}(\sH).
\end{align}
\end{restatable}
\begin{proof}
For the constrained hinge loss $\Phi_{\mathrm{hinge}}^{\mathrm{cstnd}}$, by \eqref{eq:cond_LLW}, the conditional $\Phi_{\mathrm{hinge}}^{\mathrm{cstnd}}$-risk can be expressed as follows:
\begin{equation*}
\label{eq:cond_LLW_hinge}
\begin{aligned}
& \sC_{\Phi_{\mathrm{hinge}}^{\mathrm{cstnd}}}(h,x)
= \sum_{y\in \sY} \paren*{1-\sfp(y \!\mid\! x)}\max\curl*{0,1+h(x,y)}\\
& = \sum_{y\in \curl*{y_{\max}, \hh(x)}}\paren*{1-\sfp(y \!\mid\! x)}\max\curl*{0,1+h(x,y)}+\sum_{y\not \in \curl*{y_{\max}, \hh(x)}}\paren*{1-\sfp(y \!\mid\! x)}\max\curl*{0,1+h(x,y)}
\end{aligned}
\end{equation*}
Let $h \in \sH$ be a
hypothesis such that $\hh(x) \neq y_{\max}$. For any $x\in \sX$, if $h(x,y_{\max})\leq -1$, define the hypothesis $\ov h \in \sH$ by
\begin{align*}
\ov h(x,y) = 
\begin{cases}
  h(x, y) & \text{if $y \not \in \curl*{y_{\max}, \hh(x)}$}\\
  h(x, y_{\max}) & \text{if $y = \hh(x)$}\\
  h(x,\hh(x)) & \text{if $y = y_{\max}$}.
\end{cases} 
\end{align*}
Otherwise, define the hypothesis $\ov h\in \sH$ by
\begin{align*}
\ov h(x,y) = 
\begin{cases}
  h(x, y) & \text{if $y \not \in \curl*{y_{\max}, \hh(x)}$}\\
  -1 & \text{if $y = \hh(x)$}\\
  h(x, y_{\max})+h(x,\hh(x))+1 & \text{if $y = y_{\max}$}.
\end{cases}   
\end{align*}
By the completeness of $\sH$, the new hypothesis $\ov h$ is in $\sH$ and satisfies that $\sum_{y\in \sY}\ov h(x,y)=0$. Since
$\sum_{y \in \sY} h(x, y) = 0$, there must be non-negative scores. By definition
of $\hh(x)$ as a maximizer, we must thus have $h(x,\hh(x)) \geq 0$.
Therefore, the minimal conditional $\Phi_{\mathrm{hinge}}^{\mathrm{cstnd}}$-risk satisfies:
\begin{align*}
& \sC^*_{\Phi_{\mathrm{hinge}}^{\mathrm{cstnd}}}(\sH, x)
 \leq \sC_{\Phi_{\mathrm{hinge}}^{\mathrm{cstnd}}}(\ov h, x)\\
& =\begin{cases}
\paren*{1-\sfp(y_{\max} \!\mid\! x)}(1+ h(x, \hh(x))) +\sum_{y\not \in \curl*{y_{\max},\hh(x)}}\paren*{1-\sfp(y \!\mid\! x)}(1+h(x,y))
\\ \qquad \text{if $h(x,y_{\max})\leq -1$}\\
\paren*{1-\sfp(y_{\max} \!\mid\! x)}(h(x, y_{\max})+h(x,\hh(x))+2)+\sum_{y\not \in \curl*{y_{\max},\hh(x)}}\paren*{1-\sfp(y \!\mid\! x)}(1+h(x,y)) \\ \qquad \text{otherwise.}
\end{cases}
\end{align*}
By the definition and using the fact that $\mathsf H(x)=\sY$ when $\sH$ is symmetric, we obtain
\begin{align*}
\Delta\sC_{\Phi_{\mathrm{hinge}}^{\mathrm{cstnd}},\sH}(h,x)
& = \sC_{\Phi_{\mathrm{hinge}}^{\mathrm{cstnd}}}(h,x) - \sC^*_{\Phi_{\mathrm{hinge}}^{\mathrm{cstnd}}}(\sH, x) \\
& \geq \sC_{\Phi_{\mathrm{hinge}}^{\mathrm{cstnd}}}(h,x) - \sC_{\Phi_{\mathrm{hinge}}^{\mathrm{cstnd}}}(\ov h,x)\\
& = \paren*{1+h(x,\hh(x))}\paren*{\sfp(y_{\max} \!\mid\! x)-\sfp(\hh(x) \!\mid\! x)} \\
& \geq \max_{y\in \sY}\sfp(y \!\mid\! x) - \sfp(\hh(x) \!\mid\! x) \tag{$h(x,\hh(x))\geq 0$}\\
& = \Delta\sC_{\ell_{0-1},\sH}(h,x) \tag{by Lemma~\ref{lemma:explicit_assumption_01-mhcb} and $\mathsf H(x)=\sY$}\\
& \geq \bracket*{\Delta\sC_{\ell_{0-1},\sH}(h,x)}_{\e} \tag{$\bracket*{t}_{\e}\leq t$}
\end{align*}
for any $\e\geq 0$. Therefore, taking $\sP$ be the set of all distributions, $\sH$ be the symmetric and complete hypothesis set, $\e=0$ and $\Psi(t)=t$ in Theorem~\ref{Thm:excess_bounds_Psi_01_general-mhcb}, or, equivalently, $\Gamma(t) = t$ in Theorem~\ref{Thm:excess_bounds_Gamma_01_general-mhcb}, we obtain for any hypothesis $h\in\sH$ and any distribution,
\begin{align*}
\sR_{\ell_{0-1}}(h)-\sR_{\ell_{0-1}}^*(\sH)\leq\sR_{\Phi_{\mathrm{hinge}}^{\mathrm{cstnd}}}(h)-\sR_{\Phi_{\mathrm{hinge}}^{\mathrm{cstnd}}}^*(\sH)+\sM_{\Phi_{\mathrm{hinge}}^{\mathrm{cstnd}}}(\sH)-\sM_{\ell_{0-1}}(\sH).
\end{align*}
\end{proof}

\begin{restatable}[\textbf{$\sH$-consistency bound of $\Phi_{\mathrm{sq-hinge}}^{\mathrm{cstnd}}$}]
  {theorem}{BoundLeeSqHinge}
\label{Thm:bound_lee_sq-hinge}
Suppose that $\sH$ is symmetric and complete. Then, for any hypothesis $h\in\sH$ and any distribution,
\begin{align}
\label{eq:bound_lee_sq-hinge}
     \sR_{\ell_{0-1}}(h)-\sR_{\ell_{0-1}}^*(\sH)\leq\paren*{\sR_{\Phi_{\mathrm{sq-hinge}}^{\mathrm{cstnd}}}(h)-\sR_{\Phi_{\mathrm{sq-hinge}}^{\mathrm{cstnd}}}^*(\sH)+\sM_{\Phi_{\mathrm{sq-hinge}}^{\mathrm{cstnd}}}(\sH)}^{\frac12}-\sM_{\ell_{0-1}}(\sH).
\end{align}
\end{restatable}
\begin{proof}
For the constrained squared hinge loss $\Phi_{\mathrm{sq-hinge}}^{\mathrm{cstnd}}$, by \eqref{eq:cond_LLW}, the conditional $\Phi_{\mathrm{sq-hinge}}^{\mathrm{cstnd}}$-risk can be expressed as follows:
\begin{equation*}
\label{eq:cond_LLW_sq-hinge}
\begin{aligned}
& \sC_{\Phi_{\mathrm{sq-hinge}}^{\mathrm{cstnd}}}(h,x)
= \sum_{y\in \sY} \paren*{1-\sfp(y \!\mid\! x)}\max\curl*{0,1+h(x,y)}^2\\
& = \sum_{y\in \curl*{y_{\max}, \hh(x)}}\paren*{1-\sfp(y \!\mid\! x)}\max\curl*{0,1+h(x,y)}^2+\sum_{y\not \in \curl*{y_{\max}, \hh(x)}}\paren*{1-\sfp(y \!\mid\! x)}\max\curl*{0,1+h(x,y)}^2
\end{aligned}
\end{equation*}
Let $h \in \sH$ be a
hypothesis such that $\hh(x) \neq y_{\max}$. For any $x\in \sX$, if $h(x,y_{\max})\leq -1$, define the hypothesis $\ov h \in \sH$ by
\begin{align*}
\ov h(x,y) = 
\begin{cases}
  h(x, y) & \text{if $y \not \in \curl*{y_{\max}, \hh(x)}$}\\
  h(x, y_{\max}) & \text{if $y = \hh(x)$}\\
  h(x,\hh(x)) & \text{if $y = y_{\max}$}.
\end{cases} 
\end{align*}
Otherwise, define the hypothesis $\ov h\in \sH$ by
\begin{align*}
\ov h(x,y) = 
\begin{cases}
  h(x, y) & \text{if $y \not \in \curl*{y_{\max}, \hh(x)}$}\\
  \frac{1-\sfp(y_{\max} \!\mid\! x)}{2-\sfp(y_{\max} \!\mid\! x)-\sfp(\hh(x) \!\mid\! x)}(2+h(x,y_{\max})+h(x,\hh(x)))-1 & \text{if $y = \hh(x)$}\\
  \frac{1-\sfp(\hh(x) \!\mid\! x)}{2-\sfp(y_{\max} \!\mid\! x)-\sfp(\hh(x) \!\mid\! x)}(2+h(x,y_{\max})+h(x,\hh(x)))-1 & \text{if $y = y_{\max}$}.
\end{cases}   
\end{align*}
By the completeness of $\sH$, the new hypothesis $\ov h$ is in $\sH$ and satisfies that $\sum_{y\in \sY}\ov h(x,y)=0$. Since
$\sum_{y \in \sY} h(x, y) = 0$, there must be non-negative scores. By definition
of $\hh(x)$ as a maximizer, we must thus have $h(x,\hh(x)) \geq 0$.
Therefore, the minimal conditional $\Phi_{\mathrm{sq-hinge}}^{\mathrm{cstnd}}$-risk satisfies:
\begin{align*}
& \sC^*_{\Phi_{\mathrm{sq-hinge}}^{\mathrm{cstnd}}}(\sH, x)
 \leq \sC_{\Phi_{\mathrm{sq-hinge}}^{\mathrm{cstnd}}}(\ov h, x)\\
& =\begin{cases}
\paren*{1-\sfp(y_{\max} \!\mid\! x)}(1+ h(x, \hh(x)))^2 +\sum_{y\not \in \curl*{y_{\max},\hh(x)}}\paren*{1-\sfp(y \!\mid\! x)}(1+h(x,y))
\\ \qquad \text{if $h(x,y_{\max})\leq -1$}\\
\frac{\paren*{1-\sfp(y_{\max} \!\mid\! x)}\paren*{1-\sfp(\hh(x) \!\mid\! x)}(2+h(x,y_{\max})+h(x,\hh(x)))^2}{2-\sfp(y_{\max} \!\mid\! x)-\sfp(y \!\mid\! x)}+\sum_{y\not \in \curl*{y_{\max},\hh(x)}}\paren*{1-\sfp(y \!\mid\! x)}(1+h(x,y)) \\ \qquad \text{otherwise.}
\end{cases}
\end{align*}
By the definition and using the fact that $\mathsf H(x)=\sY$ when $\sH$ is symmetric, we obtain
\begin{align*}
&\Delta\sC_{\Phi_{\mathrm{sq-hinge}}^{\mathrm{cstnd}},\sH}(h,x)
= \sC_{\Phi_{\mathrm{sq-hinge}}^{\mathrm{cstnd}}}(h,x) - \sC^*_{\Phi_{\mathrm{sq-hinge}}^{\mathrm{cstnd}}}(\sH, x) \\
& \geq \sC_{\Phi_{\mathrm{sq-hinge}}^{\mathrm{cstnd}}}(h,x) - \sC_{\Phi_{\mathrm{sq-hinge}}^{\mathrm{cstnd}}}(\ov h,x)\\
& = 
\begin{cases}
\paren*{1+h(x,\hh(x))}^2\paren*{\sfp(y_{\max} \!\mid\! x)-\sfp(\hh(x) \!\mid\! x)} & \text{if $h(x,y_{\max})\leq -1$}\\
g\paren*{1-\sfp(y_{\max} \!\mid\! x),1-\sfp(\hh(x) \!\mid\! x),1+h(x,y_{\max}), 1+h(x,\hh(x))} & \text{otherwise}
\end{cases}\\
& \geq \paren*{1+h(x,\hh(x))}^2 \paren*{\max_{y\in \sY}\sfp(y \!\mid\! x) - \sfp(\hh(x) \!\mid\! x)}^2 \tag{property of $g$ and $ \sfp(y_{\max} \!\mid\! x)\leq 1$}\\
& \geq \paren*{\max_{y\in \sY}\sfp(y \!\mid\! x) - \sfp(\hh(x) \!\mid\! x)}^2 \tag{$h(x,\hh(x))\geq 0$}\\
& = \paren*{ \Delta\sC_{\ell_{0-1},\sH}(h,x)}^2 \tag{by Lemma~\ref{lemma:explicit_assumption_01-mhcb} and $\mathsf H(x)=\sY$}\\
& \geq \paren*{\bracket*{\Delta\sC_{\ell_{0-1},\sH}(h,x)}_{\e}}^2 \tag{$\bracket*{t}_{\e}\leq t$}
\end{align*}
for any $\e\geq 0$, where $g(x,y,\alpha,\beta)=\frac{x^2\alpha^2+y^2\beta^2-2xy\alpha\beta}{x+y}\geq \beta^2(x-y)^2$ when $0\leq x\leq y\leq 1$, $x+y\geq 1$ and $1 \leq  \alpha\leq \beta$. Therefore, taking $\sP$ be the set of all distributions, $\sH$ be the symmetric and complete hypothesis set, $\e=0$ and $\Psi(t)=t^2$ in Theorem~\ref{Thm:excess_bounds_Psi_01_general-mhcb}, or, equivalently, $\Gamma(t) = \sqrt{t}$ in Theorem~\ref{Thm:excess_bounds_Gamma_01_general-mhcb}, we obtain for any hypothesis $h\in\sH$ and any distribution,
\begin{align*}
\sR_{\ell_{0-1}}(h)-\sR_{\ell_{0-1}}^*(\sH)\leq\paren*{\sR_{\Phi_{\mathrm{sq-hinge}}^{\mathrm{cstnd}}}(h)-\sR_{\Phi_{\mathrm{sq-hinge}}^{\mathrm{cstnd}}}^*(\sH)+\sM_{\Phi_{\mathrm{sq-hinge}}^{\mathrm{cstnd}}}(\sH)}^{\frac12}-\sM_{\ell_{0-1}}(\sH).
\end{align*}
\end{proof}

\begin{restatable}[\textbf{$\sH$-consistency bound of $\Phi_{\mathrm{exp}}^{\mathrm{cstnd}}$}]
  {theorem}{BoundLeeExp}
\label{Thm:bound_lee_exp}
Suppose that $\sH$ is symmetric and complete. Then, for any hypothesis $h\in\sH$ and any distribution,
\begin{align}
\label{eq:bound_lee_exp}
     \sR_{\ell_{0-1}}(h)-\sR_{\ell_{0-1}}^*(\sH)\leq \sqrt{2}\paren*{\sR_{\Phi_{\mathrm{exp}}^{\mathrm{cstnd}}}(h)-\sR_{\Phi_{\mathrm{exp}}^{\mathrm{cstnd}}}^*(\sH)+\sM_{\Phi_{\mathrm{exp}}^{\mathrm{cstnd}}}(\sH)}^{\frac12}-\sM_{\ell_{0-1}}(\sH).
\end{align}
\end{restatable}

\begin{proof}
For the constrained exponential loss $\Phi_{\mathrm{exp}}^{\mathrm{cstnd}}$, by \eqref{eq:cond_LLW}, the conditional $\Phi_{\mathrm{exp}}^{\mathrm{cstnd}}$-risk can be expressed as follows:
\begin{equation*}
\label{eq:cond_LLW_exp}
\begin{aligned}
& \sC_{\Phi_{\mathrm{exp}}^{\mathrm{cstnd}}}(h,x)
= \sum_{y\in \sY} \paren*{1-\sfp(y \!\mid\! x)}\exp\paren*{h(x,y)}\\
& = \sum_{y\in \curl*{y_{\max}, \hh(x)}}\paren*{1-\sfp(y \!\mid\! x)}\exp\paren*{h(x,y)}+\sum_{y\not \in \curl*{y_{\max}, \hh(x)}}\exp\paren*{h(x,y)}
\end{aligned}
\end{equation*}
Let $h \in \sH$ be a
hypothesis such that $\hh(x) \neq y_{\max}$. For any $x\in \sX$, define the hypothesis $\ov h_{\mu} \in \sH$ by
\begin{align*}
\ov h_{\mu}(x,y) = 
\begin{cases}
  h(x, y) & \text{if $y \not \in \curl*{y_{\max}, \hh(x)}$}\\
  h(x, y_{\max})+\mu & \text{if $y = \hh(x)$}\\
  h(x,\hh(x))-\mu & \text{if $y = y_{\max}$}
\end{cases} 
\end{align*}
for any $\mu \in \Rset$. By the completeness of $\sH$, the new hypothesis $\ov h_{\mu}$ is in $\sH$ and satisfies that $\sum_{y\in \sY}\ov h_{\mu}(x,y)=0$. Since
$\sum_{y \in \sY} h(x, y) = 0$, there must be non-negative scores. By definition
of $\hh(x)$ as a maximizer, we must thus have $h(x,\hh(x)) \geq 0$.
Therefore, the minimal conditional $\Phi_{\mathrm{exp}}^{\mathrm{cstnd}}$-risk satisfies that for any $\mu\in \Rset$,
\begin{align*}
\sC^*_{\Phi_{\mathrm{exp}}^{\mathrm{cstnd}}}(\sH, x)
& \leq  \sC_{\Phi_{\mathrm{exp}}^{\mathrm{cstnd}}}(\ov h_{\mu}, x)\\
& =(1-\sfp(y_{\max} \!\mid\! x))e^{h(x, \hh(x))-\mu}+(1-\sfp(\hh(x) \!\mid\! x))e^{h(x, y_{\max})+\mu}\\
&\qquad +\sum_{y\not \in \curl*{y_{\max}, \hh(x)}}\paren*{1-\sfp(y \!\mid\! x)}\exp\paren*{h(x,y)}.
\end{align*}
By the definition and using the fact that $\mathsf H(x)=\sY$ when $\sH$ is symmetric, we obtain
\begin{align*}
&\Delta\sC_{\Phi_{\mathrm{exp}}^{\mathrm{cstnd}},\sH}(h,x)
= \sC_{\Phi_{\mathrm{exp}}^{\mathrm{cstnd}}}(h,x) - \sC^*_{\Phi_{\mathrm{exp}}^{\mathrm{cstnd}}}(\sH, x) \\
& \geq \sC_{\Phi_{\mathrm{exp}}^{\mathrm{cstnd}}}(h,x) - \sC_{\Phi_{\mathrm{exp}}^{\mathrm{cstnd}}}(\ov h_{\mu},x)\\
& \geq \paren*{\sqrt{(1-\sfp(\hh(x) \!\mid\! x))e^{h(x, \hh(x))}}-\sqrt{(1-\sfp(y_{\max} \!\mid\! x))e^{h(x, y_{\max})}}}^2 \tag{taking supremum with respect to $\mu$}\\
& \geq e^{h(x, \hh(x))} \paren*{\sqrt{(1-\sfp(\hh(x) \!\mid\! x))}-\sqrt{(1-\sfp(y_{\max} \!\mid\! x))}}^2  \tag{$e^{h(x, \hh(x))}\geq e^{h(x, y_{\max})}$ and $\sfp(\hh(x) \!\mid\! x)\leq \sfp(y_{\max} \!\mid\! x)$}\\
& \geq \paren*{\sqrt{(1-\sfp(\hh(x) \!\mid\! x))}-\sqrt{(1-\sfp(y_{\max} \!\mid\! x))}}^2 \tag{$h(x,\hh(x))\geq 0$}\\
& = \paren*{\frac{\sfp(y_{\max} \!\mid\! x)-\sfp(\hh(x) \!\mid\! x)}{\sqrt{(1-\sfp(\hh(x) \!\mid\! x))}+\sqrt{(1-\sfp(y_{\max} \!\mid\! x))}}}^2\\
& \geq \frac12 \paren*{\max_{y\in \sY}\sfp(y \!\mid\! x) - \sfp(\hh(x) \!\mid\! x)}^2 \tag{$0\leq \sfp(y_{\max} \!\mid\! x)+\sfp(\hh(x) \!\mid\! x)\leq 1$}\\
& = \frac12 \paren*{ \Delta\sC_{\ell_{0-1},\sH}(h,x)}^2 \tag{by Lemma~\ref{lemma:explicit_assumption_01-mhcb} and $\mathsf H(x)=\sY$}\\
& \geq \frac12 \paren*{\bracket*{\Delta\sC_{\ell_{0-1},\sH}(h,x)}_{\e}}^2 \tag{$\bracket*{t}_{\e}\leq t$}
\end{align*}
for any $\e\geq 0$. Therefore, taking $\sP$ be the set of all distributions, $\sH$ be the symmetric and complete hypothesis set, $\e=0$ and $\Psi(t)=\frac{t^2}{2}$ in Theorem~\ref{Thm:excess_bounds_Psi_01_general-mhcb}, or, equivalently, $\Gamma(t) = \sqrt{2t}$ in Theorem~\ref{Thm:excess_bounds_Gamma_01_general-mhcb}, we obtain for any hypothesis $h\in\sH$ and any distribution,
\begin{align*}
\sR_{\ell_{0-1}}(h)-\sR_{\ell_{0-1}}^*(\sH)\leq \sqrt{2}\paren*{\sR_{\Phi_{\mathrm{exp}}^{\mathrm{cstnd}}}(h)-\sR_{\Phi_{\mathrm{exp}}^{\mathrm{cstnd}}}^*(\sH)+\sM_{\Phi_{\mathrm{exp}}^{\mathrm{cstnd}}}(\sH)}^{\frac12}-\sM_{\ell_{0-1}}(\sH).
\end{align*}
\end{proof}

\begin{restatable}[\textbf{$\sH$-consistency bound of $\Phi_{\rho}^{\mathrm{cstnd}}$}]
  {theorem}{BoundLeeRho}
\label{Thm:bound_lee_rho}
Suppose that $\sH$ is symmetric and satisfies that for any $x\in \sX$, there exists a hypothesis $h \in \sH$ such that $h(x,y) \leq -\rho$ for any $y \neq y_{\max}$. Then, for any hypothesis $h\in\sH$ and any distribution,
\begin{align}
\label{eq:bound_lee_rho}
     \sR_{\ell_{0-1}}(h)- \sR_{\ell_{0-1}}^*(\sH) \leq   \sR_{\Phi_{\rho}^{\mathrm{cstnd}}}(h)-\sR_{\Phi_{\rho}^{\mathrm{cstnd}}}^*(\sH)+\sM_{\Phi_{\rho}^{\mathrm{cstnd}}}(\sH)-\sM_{\ell_{0-1}}(\sH).
\end{align}
\end{restatable}
\begin{proof}
Since $\sum_{y \in \sY} h(x, y) = 0$, by the definition
of $\hh(x)$ as a maximizer, we must thus have $h(x,\hh(x)) \geq 0$. For the constrained $\rho$-margin loss $\Phi_{\rho}^{\mathrm{cstnd}}$, by \eqref{eq:cond_LLW}, the conditional $\Phi_{\rho}^{\mathrm{cstnd}}$-risk can be expressed as follows:
\begin{equation*}
\begin{aligned}
\sC_{\Phi_{\rho}^{\mathrm{cstnd}}}(h,x)
& = \sum_{y\in \sY} \paren*{1-\sfp(y \!\mid\! x)}\min\curl*{\max\curl*{0,1+\frac{h(x,y)}{\rho}},1}\\
& = \sum_{y\in \sY:h(x,y)\geq 0}\paren*{1-\sfp(y \!\mid\! x)} + \sum_{y\in \sY:h(x,y)<0}\paren*{1-\sfp(y \!\mid\! x)}\max\curl*{0,1+\frac{h(x,y)}{\rho}} \\
& \geq 1-\sfp(\hh(x) \!\mid\! x)\\
& \geq 1-\max_{y\in \sY} \sfp(y \!\mid\! x).
\end{aligned}
\end{equation*}
By the assumption, the equality can be achieved by some $h_{\rho}^* \in \sH$ with constraint $\sum_{y\in \sY}h(x,y)=0$ such that
 $h_{\rho}^*(x,y) \leq -\rho$ for any $y \neq y_{\max}$ and $h_{\rho}^*(x,y_{\max})=-\sum_{y'\neq y_{\max}}h_{\rho}^*(x,y')\geq 0$. Therefore, the minimal conditional $\Phi_{\rho}^{\mathrm{cstnd}}$-risk can be expressed as follows:
\begin{align*}
\sC^*_{\Phi_{\rho}^{\mathrm{cstnd}}}(\sH, x)=1-\max_{y\in \sY} \sfp(y \!\mid\! x).
\end{align*}
By the definition and using the fact that $\mathsf H(x)=\sY$ when $\sH$ is symmetric, we obtain
\begin{align*}
& \Delta\sC_{\Phi_{\rho}^{\mathrm{cstnd}},\sH}(h,x)  = \sC_{\Phi_{\rho}^{\mathrm{cstnd}}}(h,x) - \sC^*_{ \Phi_{\rho}^{\mathrm{cstnd}}}(\sH, x) \\
& = \sum_{y\in \sY:h(x,y)\geq 0}\paren*{1-\sfp(y \!\mid\! x)} + \sum_{y\in \sY:h(x,y)< 0}\paren*{1-\sfp(y \!\mid\! x)}\max\curl*{0,1+\frac{h(x,y)}{\rho}} - \paren*{1-\max_{y\in \sY} \sfp(y \!\mid\! x)}\\
& \geq 1-\sfp(\hh(x) \!\mid\! x) - \paren*{1-\max_{y\in \sY} \sfp(y \!\mid\! x)} \\
& = \max_{y\in \sY}\sfp(y \!\mid\! x) - \sfp(\hh(x) \!\mid\! x)\\
& = \Delta\sC_{\ell_{0-1},\sH}(h,x) \tag{by Lemma~\ref{lemma:explicit_assumption_01-mhcb} and $\mathsf H(x)=\sY$}\\
& \geq \bracket*{\Delta\sC_{\ell_{0-1},\sH}(h,x)}_{\e} \tag{$\bracket*{t}_{\e}\leq t$}
\end{align*}
for any $\e\geq 0$. Therefore, taking $\sP$ be the set of all distributions, $\sH$ be the symmetric hypothesis set,
$\e=0$ and $\Psi(t)=t$ in Theorem~\ref{Thm:excess_bounds_Psi_01_general-mhcb}, or, equivalently, $\Gamma(t) = t$ in Theorem~\ref{Thm:excess_bounds_Gamma_01_general-mhcb},  we obtain for any hypothesis $h\in\sH$ and any distribution,
\begin{align*}
\sR_{\ell_{0-1}}(h)- \sR_{\ell_{0-1}}^*(\sH) \leq   \sR_{\Phi_{\rho}^{\mathrm{cstnd}}}(h)-\sR_{\Phi_{\rho}^{\mathrm{cstnd}}}^*(\sH)+\sM_{\Phi_{\rho}^{\mathrm{cstnd}}}(\sH)-\sM_{\ell_{0-1}}(\sH).
\end{align*}
\end{proof}

\section{Proof of negative results for adversarial robustness}
\label{app:deferred_proofs_adv_negative}
\NegativeAdv*
\begin{proof}
Consider the distribution that supports on a singleton domain $\curl*{x}$ with $x$ satisfying that $\sH_{\gamma}(x) \neq \emptyset$. When $\sH_{\gamma}(x) \neq \emptyset$, $\mathsf H_{\gamma}(x)$ is also non-empty. Take $y_1\in \sH_{\gamma}(x)$ and let $y_2\neq y_1$.  We define $p(x)$ as $\sfp(y_1 \!\mid\! x)=\sfp(y_2 \!\mid\! x)=\frac12$. Let $h_0=0\in \sH$. By Lemma~\ref{lemma:explicit_assumption_adv-mhcb} and the fact that $\sH_{\gamma}(x) \neq \emptyset$ and $y_1\in \mathsf H_{\gamma}(x)$, the minimal conditional $\ell_{\gamma}$-risk is
\begin{align*}
\sR^*_{\ell_{\gamma}}(\sH)=
\sC^*_{\ell_{\gamma}}(\sH, x)=1-\max_{y\in \mathsf H_{\gamma}(x)}\sfp(y \!\mid\! x)=1-\sfp(y_1 \!\mid\! x)=\frac12.
\end{align*}
For $h=h_0$, we have
\begin{align*}
\sR_{\ell_{\gamma}}(h_0)=\sC_{\ell_{\gamma}}(h_0,x)=\sum_{y\in \sY} \sfp(y \!\mid\! x) \sup_{x':\norm*{x-x'}_p\leq \gamma}\mathds{1}_{\rho_h(x', y) \leq 0}=1.
\end{align*} 
For the adversarial max loss with non-increasing $\Phi$, the conditional $\wt \Phi^{\mathrm{max}}$-risk can be expressed as follows:
\begin{align*}
\sC_{\wt \Phi^{\mathrm{max}}}(h,x)  
&=  \sum_{y\in \sY} \sfp(y \!\mid\! x) \sup_{x':\norm*{x-x'}_p\leq\gamma}\Phi\paren*{\rho_h(x', y)}\\
& = \sum_{y\in \sY} \sfp(y \!\mid\! x) \Phi\paren*{\inf_{x':\norm*{x-x'}_p\leq\gamma}\rho_h(x', y)}\\
& =\frac12 \Phi\paren*{\inf_{x':\norm*{x-x'}_p\leq\gamma}\paren*{h(x', y_1)-h(x',y_2)}} + \frac12 \Phi\paren*{\inf_{x':\norm*{x-x'}_p\leq\gamma}\paren*{h(x',y_2)-h(x', y_1)}}\\
& =\frac12 \Phi\paren*{\inf_{x':\norm*{x-x'}_p\leq\gamma}\paren*{h(x', y_1)-h(x',y_2)}} + \frac12 \Phi\paren*{-\sup_{x':\norm*{x-x'}_p\leq\gamma}\paren*{h(x',y_1)-h(x', y_2)}}
\end{align*}
If $\Phi$ is convex and non-increasing, we obtain for any $h\in \sH$,
\begin{align*}
\sR_{\wt \Phi^{\mathrm{max}}}(h)
&=\sC_{\wt \Phi^{\mathrm{max}}}(h,x)\\
&=\frac12 \Phi\paren*{\inf_{x':\norm*{x-x'}_p\leq\gamma}\paren*{h(x', y_1)-h(x',y_2)}} + \frac12 \Phi\paren*{\sup_{x':\norm*{x-x'}_p\leq\gamma}\paren*{h(x',y_1)-h(x', y_2)}}\\
& \geq \Phi\paren*{\frac12 \inf_{x':\norm*{x-x'}_p\leq\gamma}\paren*{h(x', y_1)-h(x',y_2)}- \frac12 \sup_{x':\norm*{x-x'}_p\leq\gamma}\paren*{h(x',y_1)-h(x', y_2)}} \tag{$\Phi$ is convex}\\
& \geq \Phi(0), \tag{$\Phi$ is non-increasing}
\end{align*}
where the equality can be achieved by $h_0$.
Therefore,
\begin{align*}
\sR_{\wt \Phi^{\mathrm{max}}}^*(\sH)=\sC^*_{\wt \Phi^{\mathrm{max}}}(\sH, x)=\sR_{\wt \Phi^{\mathrm{max}}}(h_0)=\Phi(0).
\end{align*}
If \eqref{eq:bound_convex_adv} holds for some non-decreasing function $f$ and $\wt \ell = \wt \Phi^{\mathrm{max}}$, then, we obtain for any $h\in \sH$,
\begin{align*}
\sR_{\ell_{\gamma}}(h)-\frac12\leq  f\paren*{\sR_{\wt \Phi^{\mathrm{max}}}(h) - \Phi(0)}.
\end{align*}
Let $h=h_0$, then $f(0)\geq 1/2$. Since $f$ is non-decreasing, for any $t\geq 0$, $f(t)\geq 1/2$.

For the adversarial sum loss with non-increasing $\Phi$, the conditional $\wt \Phi^{\mathrm{sum}}$-risk can be expressed as follows:
\begin{align*}
\sC_{\wt \Phi^{\mathrm{sum}}}(h,x) &=  \sum_{y\in \sY} \sfp(y \!\mid\! x) \sup_{x':\norm*{x-x'}_p\leq\gamma}\sum_{y'\neq y}\Phi\paren*{h(x',y)-h(x',y')}\\
& =\frac12 \Phi\paren*{\inf_{x':\norm*{x-x'}_p\leq\gamma}\paren*{h(x', y_1)-h(x',y_2)}} + \frac12 \Phi\paren*{\inf_{x':\norm*{x-x'}_p\leq\gamma}\paren*{h(x',y_2)-h(x', y_1)}}\\
& =\frac12 \Phi\paren*{\inf_{x':\norm*{x-x'}_p\leq\gamma}\paren*{h(x', y_1)-h(x',y_2)}} + \frac12 \Phi\paren*{-\sup_{x':\norm*{x-x'}_p\leq\gamma}\paren*{h(x',y_1)-h(x', y_2)}}
\end{align*}
If $\Phi$ is convex and non-increasing, we obtain for any $h\in \sH$,
\begin{align*}
\sR_{\wt \Phi^{\mathrm{sum}}}(h)
&=\sC_{\wt \Phi^{\mathrm{sum}}}(h,x)\\
&=\frac12 \Phi\paren*{\inf_{x':\norm*{x-x'}_p\leq\gamma}\paren*{h(x', y_1)-h(x',y_2)}} + \frac12 \Phi\paren*{\sup_{x':\norm*{x-x'}_p\leq\gamma}\paren*{h(x',y_1)-h(x', y_2)}}\\
& \geq \Phi\paren*{\frac12 \inf_{x':\norm*{x-x'}_p\leq\gamma}\paren*{h(x', y_1)-h(x',y_2)}- \frac12 \sup_{x':\norm*{x-x'}_p\leq\gamma}\paren*{h(x',y_1)-h(x', y_2)}} \tag{$\Phi$ is convex}\\
& \geq \Phi(0), \tag{$\Phi$ is non-increasing}
\end{align*}
where the equality can be achieved by $h_0$.
Therefore,
\begin{align*}
\sR_{\wt \Phi^{\mathrm{sum}}}^*(\sH)=\sC^*_{\wt \Phi^{\mathrm{sum}}}(\sH, x)=\sR_{\wt \Phi^{\mathrm{sum}}}(h_0)=\Phi(0).
\end{align*}
If \eqref{eq:bound_convex_adv} holds for some non-decreasing function $f$ and $\wt \ell = \wt \Phi^{\mathrm{sum}}$, then, we obtain for any $h\in \sH$,
\begin{align*}
\sR_{\ell_{\gamma}}(h)-\frac12\leq  f\paren*{\sR_{\wt \Phi^{\mathrm{sum}}}(h) - \Phi(0)}.
\end{align*}
Let $h=h_0$, then $f(0)\geq 1/2$. Since $f$ is non-decreasing, for any $t\geq 0$, $f(t)\geq 1/2$.

For the adversarial constrained loss with non-increasing $\Phi$, using the fact that $h(x,y_1)+h(x,y_2)=0$, the conditional $\wt \Phi^{\mathrm{cstnd}}$-risk can be expressed as follows:
\begin{align*}
\sC_{\wt \Phi^{\mathrm{cstnd}}}(h,x) &=  \sum_{y\in \sY} \sfp(y \!\mid\! x) \sup_{x':\norm*{x-x'}_p\leq\gamma}\sum_{y'\neq y}\Phi\paren*{-h(x',y')}\\
& =\frac12 \Phi\paren*{\inf_{x':\norm*{x-x'}_p\leq\gamma}\paren*{-h(x',y_2)}} + \frac12 \Phi\paren*{\inf_{x':\norm*{x-x'}_p\leq\gamma}\paren*{-h(x', y_1)}}\\
& =\frac12 \Phi\paren*{\inf_{x':\norm*{x-x'}_p\leq\gamma}h(x', y_1)} + \frac12 \Phi\paren*{-\sup_{x':\norm*{x-x'}_p\leq\gamma}h(x',y_1)}
\end{align*}
If $\Phi$ is convex and non-increasing, we obtain for any $h\in \sH$,
\begin{align*}
\sR_{\wt \Phi^{\mathrm{cstnd}}}(h)
&=\sC_{\wt \Phi^{\mathrm{cstnd}}}(h,x)\\
&=\frac12 \Phi\paren*{\inf_{x':\norm*{x-x'}_p\leq\gamma}h(x', y_1)} + \frac12 \Phi\paren*{-\sup_{x':\norm*{x-x'}_p\leq\gamma}h(x',y_1)}\\
& \geq \Phi\paren*{\frac12 \inf_{x':\norm*{x-x'}_p\leq\gamma}h(x', y_1)- \frac12 \sup_{x':\norm*{x-x'}_p\leq\gamma}h(x',y_1)} \tag{$\Phi$ is convex}\\
& \geq \Phi(0), \tag{$\Phi$ is non-increasing}
\end{align*}
where the equality can be achieved by $h_0$.
Therefore,
\begin{align*}
\sR_{\wt \Phi^{\mathrm{cstnd}}}^*(\sH)=\sC^*_{\wt \Phi^{\mathrm{cstnd}}}(\sH, x)=\sR_{\wt \Phi^{\mathrm{cstnd}}}(h_0)=\Phi(0).
\end{align*}
If \eqref{eq:bound_convex_adv} holds for some non-decreasing function $f$ and $\wt \ell = \wt \Phi^{\mathrm{cstnd}}$, then, we obtain for any $h\in \sH$,
\begin{align*}
\sR_{\ell_{\gamma}}(h)-\frac12\leq  f\paren*{\sR_{\wt \Phi^{\mathrm{cstnd}}}(h) - \Phi(0)}.
\end{align*}
Let $h=h_0$, then $f(0)\geq 1/2$. Since $f$ is non-decreasing, for any $t\geq 0$, $f(t)\geq 1/2$.

\end{proof}

\section{Proof of \texorpdfstring{$\sH$}{H}-consistency bounds for adversarial max losses \texorpdfstring{$ \wt \Phi^{\mathrm{max}}$}{maxadv}}
\label{app:deferred_proofs_adv_max}

\BoundMaxRhoAdv*
\begin{proof}
By the definition, the conditional $\wt{\Phi}_{\rho}^{\mathrm{max}}$-risk can be expressed as follows:
\begin{equation}
\label{eq:cond_adv_rho_max}
\begin{aligned}
& \sC_{\wt{\Phi}_{\rho}^{\mathrm{max}}}(h,x) =  \sum_{y\in \sY} \sfp(y \!\mid\! x) \sup_{x':\norm*{x-x'}_p\leq \gamma}\Phi_{\rho}\paren*{\rho_h(x', y)}\\
&=\begin{cases}
1-\sfp(\hh(x) \!\mid\! x)+\max\curl*{0,1-\frac{\inf_{x':\norm*{x-x'}_p\leq\gamma}\rho_h(x',\hh(x))}{\rho}}\,\sfp(\hh(x) \!\mid\! x) & h\in \sH_{\gamma}(x) \\
1 & \text{otherwise.}
\end{cases}\\
&=\begin{cases}
1-\min\curl*{1,\frac{\inf_{x':\norm*{x-x'}_p\leq\gamma}\rho_h(x',\hh(x))}{\rho}}\,\sfp(\hh(x) \!\mid\! x) & h\in \sH_{\gamma}(x) \\
1 & \text{otherwise.}
\end{cases}
\end{aligned}   
\end{equation}
Since $\sH$ is symmetric, for any $x\in \sX$, either for any $y\in \sY$, 
\begin{align*}
\sup_{h\in \curl*{h\in\sH_{\gamma}(x): \hh(x)=y}}\inf_{x':\norm*{x-x'}_p\leq\gamma}\rho_h(x',\hh(x))=\sup_{h\in \sH_{\gamma}(x)}\inf_{x':\norm*{x-x'}_p\leq\gamma}\rho_h(x',\hh(x))   
\end{align*}
or $\sH_{\gamma}(x)=\emptyset$. When $\sH_{\gamma}(x)=\emptyset$, \eqref{eq:cond_adv_rho_max} implies that $\sC^*_{\wt{\Phi}_{\rho}^{\mathrm{max}}}(\sH, x)=1$. When $\sH_{\gamma}(x)\neq\emptyset$, 
\begin{align*}
\sC^*_{\wt{\Phi}_{\rho}^{\mathrm{max}}}(\sH, x)=1-\min\curl*{1,\frac{\sup_{h\in \sH_{\gamma}(x)}\inf_{x':\norm*{x-x'}_p\leq\gamma}\rho_h(x',\hh(x))}{\rho}}\max_{y\in \sY}\sfp(y \!\mid\! x).
\end{align*}
Therefore, the minimal conditional $\wt{\Phi}_{\rho}^{\mathrm{max}}$-risk can be expressed as follows:
\begin{align*}
\sC^*_{\wt{\Phi}_{\rho}^{\mathrm{max}}}(\sH, x)=1-\min\curl*{1,\frac{\sup_{h\in \sH_{\gamma}(x)}\inf_{x':\norm*{x-x'}_p\leq\gamma}\rho_h(x',\hh(x))}{\rho}}\max_{y\in \sY}\sfp(y \!\mid\! x)\mathds{1}_{\sH_{\gamma}(x)\neq\emptyset}
\end{align*}
When $\sH_{\gamma}(x)=\emptyset$, $\sC_{\wt{\Phi}_{\rho}^{\mathrm{max}}}(h,x)\equiv 1$, which implies that $\Delta\sC_{\wt{\Phi}_{\rho}^{\mathrm{max}},\sH}(h,x) \equiv 0$. When $\sH_{\gamma}(x)\neq\emptyset$, using the fact that $\mathsf H_{\gamma}(x) =\sY \iff \sH_{\gamma}(x)\neq\emptyset$ when $\sH$ is symmetric,
\begin{align*}
&\Delta\sC_{\wt{\Phi}_{\rho}^{\mathrm{max}},\sH}(h,x)\\
&=\min\curl*{1,\frac{\sup_{h\in \sH_{\gamma}(x)}\inf_{x':\norm*{x-x'}_p\leq\gamma}\rho_h(x',\hh(x))}{\rho}}\max_{y\in \sY}\sfp(y \!\mid\! x)\\
&-\min\curl*{1,\frac{\inf_{x':\norm*{x-x'}_p\leq\gamma}\rho_h(x',\hh(x))}{\rho}}\sfp(\hh(x) \!\mid\! x)\mathds{1}_{h\in \sH_{\gamma}(x)}\\
&\geq  \min\curl*{1,\frac{\sup_{h\in \sH_{\gamma}(x)}\inf_{x':\norm*{x-x'}_p\leq\gamma}\rho_h(x',\hh(x))}{\rho}}\paren*{\max_{y\in \sY} \sfp(y \!\mid\! x)-\sfp(\hh(x) \!\mid\! x)\mathds{1}_{h\in \sH_{\gamma}(x)}}\\
& = \min\curl*{1,\frac{\sup_{h\in \sH_{\gamma}(x)}\inf_{x':\norm*{x-x'}_p\leq\gamma}\rho_h(x',\hh(x))}{\rho}}\Delta\sC_{\ell_{\gamma},\sH}(h,x)\\
& \geq \min\curl*{1,\frac{\sup_{h\in \sH_{\gamma}(x)}\inf_{x':\norm*{x-x'}_p\leq\gamma}\rho_h(x',\hh(x))}{\rho}}\bracket*{\Delta\sC_{\ell_{\gamma},\sH}(h,x)}_{\e}\\
& \geq \min\curl*{1,\frac{\inf_{x\in \curl*{x\in\sX:\sH_{\gamma}(x)\neq \emptyset}}\sup_{h\in \sH_{\gamma}(x)}\inf_{x':\norm*{x-x'}_p\leq\gamma}\rho_h(x',\hh(x))}{\rho}}\bracket*{\Delta\sC_{\ell_{\gamma},\sH}(h,x)}_{\e}
\end{align*}
for any $\e\geq 0$. Therefore, taking $\sP$ be the set of all distributions, $\sH$ be the symmetric hypothesis set, $\e=0$ and
\begin{align*}
\Psi(t)=\min\curl*{1,\frac{\inf_{x\in \curl*{x\in\sX:\sH_{\gamma}(x)\neq \emptyset}}\sup_{h\in \sH_{\gamma}(x)}\inf_{x':\norm*{x-x'}_p\leq\gamma}\rho_h(x',\hh(x))}{\rho}}\,t
\end{align*}
in Theorem~\ref{Thm:excess_bounds_Psi_adv_general}, or, equivalently, $\Gamma(t) = \Psi^{-1}(t)$ in Theorem~\ref{Thm:excess_bounds_Gamma_adv_general}, we obtain for any hypothesis $h\in\sH$ and any distribution,
\begin{align*}
\sR_{\ell_{\gamma}}(h)- \sR_{\ell_{\gamma}}^*(\sH) \leq \frac{  \sR_{\wt{\Phi}_{\rho}^{\mathrm{max}}}(h)-\sR_{\wt{\Phi}_{\rho}^{\mathrm{max}}}^*(\sH)+\sM_{\wt{\Phi}_{\rho}^{\mathrm{max}}}(\sH)}{\min\curl*{1,\frac{\inf_{x\in \curl*{x\in\sX:\sH_{\gamma}(x)\neq \emptyset}}\sup_{h\in \sH_{\gamma}(x)}\inf_{x':\norm*{x-x'}_p\leq\gamma}\rho_h(x',\hh(x))}{\rho}}}-\sM_{\ell_{\gamma}}(\sH).
\end{align*}
\end{proof}

\section{Proof of \texorpdfstring{$\sH$}{H}-consistency bounds for adversarial sum losses \texorpdfstring{$ \wt \Phi^{\mathrm{sum}}$}{sumadv}}
\label{app:deferred_proofs_adv_sum}
\BoundSumRhoAdv*
\begin{proof}
For any $x\in \sX$, we define \[p_{\bracket*{1}}(x), p_{\bracket*{2}}(x), \ldots, p_{\bracket*{c}}(x)\] by sorting the probabilities  $\curl*{\sfp(y \!\mid\! x):y\in \sY}$ in increasing order. Similarly,
for any $x\in \sX$ and $h\in \sH$, we define $h\paren*{x, \curl*{1}_x}, h\paren*{x, \curl*{2}_x},\ldots, h\paren*{x, \curl*{c}_x}$ by sorting the scores $\curl*{h(x,y):y\in \sY}$ in increasing order. In particular, we have
\begin{align*}
h\paren*{x, \curl*{1}_x}=\min_{y\in \sY}h(x,y), \quad  h\paren*{x, \curl*{c}_x}=\max_{y\in \sY}h(x,y), \quad h\paren*{x, \curl*{i}_x}\leq h\paren*{x, \curl*{j}_j},\,\forall i\leq j.
\end{align*}
If there is a tie for the maximum, we pick the label with the highest index under the natural ordering of labels, i.e. $\curl*{c}_x=\hh(x)$.
By the definition, the conditional $\wt{\Phi}_{\rho}^{\mathrm{sum}}$-risk can be expressed as follows:
\begin{equation}
\label{eq:cond_adv_sum_rho}
\begin{aligned}
& \sC_{\wt{\Phi}_{\rho}^{\mathrm{sum}}}(h,x) =  \sum_{y\in \sY} \sfp(y \!\mid\! x) \sup_{x':\norm*{x-x'}_p\leq \gamma}\sum_{y'\neq y} \Phi_{\rho}\paren*{h(x',y)-h(x',y')}\\
& = \sum_{y\in \sY}  \sup_{x':\norm*{x-x'}_p\leq \gamma}\sfp(y \!\mid\! x) \sum_{y'\neq y} \Phi_{\rho}\paren*{h(x',y)-h(x',y')}\\
& =
\sum_{i=1}^c \sup_{x':\norm*{x-x'}_p\leq \gamma} \sfp(\curl*{i}_{x'} \!\mid\! x) \bracket[\Bigg]{\sum_{j=1}^{i-1}\Phi_{\rho}\paren*{h(x',\curl*{i}_{x'})-h(x',\curl*{j}_{x'})} \\ & \qquad + \sum_{j=i+1}^{c}\Phi_{\rho}\paren*{h(x',\curl*{i}_{x'})-h(x',\curl*{j}_{x'})}}\\
& = \sum_{i=1}^c \sup_{x':\norm*{x-x'}_p\leq \gamma} \sfp(\curl*{i}_{x'} \!\mid\! x) \bracket*{\sum_{j=1}^{i-1}\Phi_{\rho}\paren*{h(x',\curl*{i}_{x'})-h(x',\curl*{j}_{x'})} + c-i}
\end{aligned}   
\end{equation}
By the assumption, there exists a hypothesis $h \in \sH$
inducing the same ordering of the labels for any $x'\in \curl*{x'\colon \norm*{x - x'}_p\leq \gamma}$
and such that
$\inf_{x'\colon \norm*{x - x'}_p\leq \gamma}\abs*{h(x', i) - h(x', j)}\geq \rho$ for
any $i\neq j \in \sY$, i.e. $\curl*{k}_{x'}=\curl*{k}_{x}$ for any $k\in \sY$ and $x'\in \curl*{x':\norm*{x-x'}_p\leq \gamma}$. Since $\sH$ is symmetric, we can always choose $h^*$ among these hypotheses such that $h^*$ and $p(x)$ induce the same ordering of the labels, i.e. $\sfp(\curl*{k}_{x} \!\mid\! x)=p_{\bracket*{k}}(x)$ for any $k\in \sY$. Then, by \eqref{eq:cond_adv_sum_rho}, we have
\begin{align*}
\sC^*_{\wt{\Phi}_{\rho}^{\mathrm{sum}}}(\sH, x)
& \leq \sC_{\wt{\Phi}_{\rho}^{\mathrm{sum}}}(h^*, x)\\
& =\sum_{i=1}^c \sup_{x':\norm*{x-x'}_p\leq \gamma} \sfp(\curl*{i}_{x'} \!\mid\! x) \bracket*{\sum_{j=1}^{i-1}\Phi_{\rho}\paren*{h^*(x',\curl*{i}_{x'})-h^*(x',\curl*{j}_{x'})} + c-i} \tag{by $\eqref{eq:cond_adv_sum_rho}$}\\
& = \sum_{i=1}^c \sup_{x':\norm*{x-x'}_p\leq \gamma} \sfp(\curl*{i}_{x'} \!\mid\! x) (c-i) \tag{$\inf_{x':\norm*{x-x'}_p\leq \gamma}\abs*{h^*(x',i)-h^*(x',j)}\geq \rho$ for any $i\neq j$ and $\Phi_{\rho}(t)=0,\,\forall t\geq \rho$}\\
& =  \sum_{i=1}^c \sfp(\curl*{i}_{x} \!\mid\! x) (c-i) \tag{$h^*$ induces the same ordering of the labels for any $x'\in \curl*{x':\norm*{x-x'}_p\leq \gamma}$}\\
& = \sum_{i=1}^c p_{\bracket*{i}}(x) (c-i) \tag{$h^*$ and $p(x)$ induce the same ordering of the labels}\\
& = c - \sum_{i=1}^c i\, p_{\bracket*{i}}(x) \tag{$\sum_{i=1}^c p_{\bracket*{i}}(x)=1$}
\end{align*}
By the assumption, $\sH_{\gamma}(x)\neq\emptyset$ and $\mathsf H_{\gamma}(x)=\sY$ since $\sH$ is symmetric. Thus, for any $h\in\sH$,
\begin{align*}
& \Delta\sC_{\wt{\Phi}_{\rho}^{\mathrm{sum}},\sH}(h,x)\\
& =  \sC_{\wt{\Phi}_{\rho}^{\mathrm{sum}}}(h,x)-\sC^*_{\wt{\Phi}_{\rho}^{\mathrm{sum}}}(\sH, x)\\
& =  \sum_{i=1}^c \sup_{x':\norm*{x-x'}_p\leq \gamma} \sfp(\curl*{i}_{x'} \!\mid\! x) \bracket*{\sum_{j=1}^{i-1}\Phi_{\rho}\paren*{h(x',\curl*{i}_{x'})-h(x',\curl*{j}_{x'})} + c-i} - \paren*{c - \sum_{i=1}^c i\, p_{\bracket*{i}}(x)}\tag{$\Phi_{\rho}(t)=1$ for $t\leq 0$}\\
&\geq \sfp(\hh(x) \!\mid\! x)\mathds{1}_{h\not \in \sH_{\gamma}(x)} +  \sum_{i=1}^c \sup_{x':\norm*{x-x'}_p\leq \gamma} \sfp(\curl*{i}_{x'} \!\mid\! x) \paren*{c-i} - \paren*{c - \sum_{i=1}^c i\, p_{\bracket*{i}}(x)} \tag{$\Phi_{\rho}\geq 0$}\\
& \geq \sfp(\hh(x) \!\mid\! x)\mathds{1}_{h\not \in \sH_{\gamma}(x)} + \sum_{i=1}^c \sfp(\curl*{i}_{x} \!\mid\! x) \paren*{c-i} - \paren*{c - \sum_{i=1}^c i\, p_{\bracket*{i}}(x)} \tag{lower bound the supremum}\\
& = \sfp(\hh(x) \!\mid\! x)\mathds{1}_{h\not \in \sH_{\gamma}(x)} + \sum_{i=1}^c i\, p_{\bracket*{i}}(x) - \sum_{i=1}^c i\, \sfp(\curl*{i}_{x} \!\mid\! x) \tag{$\sum_{i=1}^c \sfp(\curl*{i}\!\mid\! x)=1$}\\
& = \sfp(\hh(x) \!\mid\! x)\mathds{1}_{h\not \in \sH_{\gamma}(x)} + \max_{y\in \sY}\sfp(y \!\mid\! x)-\sfp(\hh(x) \!\mid\! x) +
\\ & \qquad \begin{bmatrix}
c-1\\
c-1\\
c-2\\
\vdots\\
1
\end{bmatrix}
\cdot
\begin{bmatrix}
p_{\bracket*{c}}(x)\\
p_{\bracket*{c-1}}(x)\\
p_{\bracket*{c-2}}(x)\\
\vdots\\
p_{\bracket*{1}}(x)\\
\end{bmatrix}
-
\begin{bmatrix}
c-1\\
c-1\\
c-2\\
\vdots\\
1
\end{bmatrix}
\cdot
\begin{bmatrix}
\sfp(\curl*{c}_{x} \!\mid\! x)\\
\sfp(\curl*{c-1}_{x} \!\mid\! x)\\
\sfp(\curl*{c-2}_{x} \!\mid\! x)\\
\vdots\\
\sfp(\curl*{1}_{x} \!\mid\! x)\\
\end{bmatrix}
\tag{$p_{[c]}(x)=\max_{y\in \sY}\sfp(y \!\mid\! x)$ and $\curl*{c}_x = \hh(x)$}\\
& \geq \sfp(\hh(x) \!\mid\! x)\mathds{1}_{h\not \in \sH_{\gamma}(x)} + \max_{y\in \sY}\sfp(y \!\mid\! x)-\sfp(\hh(x) \!\mid\! x) \tag{by Lemma~\ref{lemma:sum_auxiliary}}\\
& = \max_{y\in \sY}\sfp(y \!\mid\! x)-\sfp(\hh(x) \!\mid\! x)\mathds{1}_{h\in \sH_{\gamma}(x)}\\
& = \Delta\sC_{\ell_{\gamma},\sH}(h,x)\tag{by Lemma~\ref{lemma:explicit_assumption_adv-mhcb} and $\mathsf H_{\gamma}(x) =\sY$}\\
& \geq \bracket*{\Delta\sC_{\ell_{\gamma},\sH}(h,x)}_{\e} \tag{$\bracket*{t}_{\e}\leq t$}
\end{align*}
for any $\e\geq 0$.
Therefore, taking $\sP$ be the set of all distributions, $\sH$ be the symmetric hypothesis set, $\e=0$ and
$\Psi(t)=t$ in Theorem~\ref{Thm:excess_bounds_Psi_adv_general}, or, equivalently, $\Gamma(t) = t$ in Theorem~\ref{Thm:excess_bounds_Gamma_adv_general}, we obtain for any hypothesis $h\in\sH$ and any distribution,
\begin{align*}
\sR_{\ell_{\gamma}}(h)- \sR_{\ell_{\gamma}}^*(\sH) \leq   \sR_{\wt{\Phi}_{\rho}^{\mathrm{sum}}}(h)-\sR_{\wt{\Phi}_{\rho}^{\mathrm{sum}}}^*(\sH)+\sM_{\wt{\Phi}_{\rho}^{\mathrm{sum}}}(\sH)-\sM_{\ell_{\gamma}}(\sH).
\end{align*}
\end{proof}

\section{Proof of \texorpdfstring{$\sH$}{H}-consistency bounds for adversarial constrained losses \texorpdfstring{$ \wt \Phi^{\mathrm{cstnd}}$}{cstndadv}}
\label{app:deferred_proofs_adv_lee}
\BoundLeeRhoAdv*
\begin{proof}
Define $y_{\max}$ by $y_{\max} = \argmax_{y \in \sY} \sfp(y \!\mid\! x)$. If there is a tie, we pick the label with the highest index under the natural ordering of labels. Since $\sum_{y \in \sY} h(x, y) = 0$, by definition
of $\hh(x)$ as a maximizer, we must thus have $h(x,\hh(x)) \geq 0$.
By the definition, the conditional $\wt{\Phi}_{\rho}^{\mathrm{cstnd}}$-risk can be expressed as follows:
\begin{align*}
& \sC_{\wt{\Phi}_{\rho}^{\mathrm{cstnd}}}(h,x) =  \sum_{y\in \sY} \sfp(y \!\mid\! x) \sup_{x':\norm*{x-x'}_p\leq \gamma}\sum_{y'\neq y}\Phi_{\rho}\paren*{-h(x',y')}\\
& =  \sum_{y\in \sY} \sup_{x':\norm*{x-x'}_p\leq \gamma}\sfp(y \!\mid\! x) \sum_{y'\neq y}\Phi_{\rho}\paren*{-h(x',y')}\\
& =  \sum_{y\in \sY} \sup_{x':\norm*{x-x'}_p\leq \gamma}\sfp(y \!\mid\! x)\bracket*{ \sum_{y'\neq y: h(x',y')> 0}\Phi_{\rho}\paren*{-h(x',y')} + \sum_{y'\neq y: h(x',y')\leq 0}\Phi_{\rho}\paren*{-h(x',y')}} \\
& \geq \sum_{y\neq \hh(x')} \sup_{x':\norm*{x-x'}_p\leq \gamma}\sfp(y \!\mid\! x)\\
& \geq 1-\max_{y\in \sY} \sfp(y \!\mid\! x).
\tag{$\Phi_{\rho}\geq 0$ and $\Phi_{\rho}(t)=1$ for $t\leq 0$}
\end{align*}
By the assumption, the equality can be achieved by some $h_{\rho}^* \in \sH$ with constraint $\sum_{y\in \sY}h(x,y)=0$ such that
 $\sup_{x':\norm*{x-x'}_p\leq \gamma}h_{\rho}^*(x',y) \leq -\rho$ for any $y \neq y_{\max}$ and $h_{\rho}^*(x',y_{\max})=-\sum_{y'\neq y_{\max}}h_{\rho}^*(x',y')$ for any $x'\in \curl*{x':\norm*{x-x'}_p\leq \gamma}$. Therefore, the minimal conditional $\wt \Phi_{\rho}^{\mathrm{cstnd}}$-risk can be expressed as follows:
\begin{align*}
\sC^*_{\wt \Phi_{\rho}^{\mathrm{cstnd}}}(\sH, x)=1-\max_{y\in \sY} \sfp(y \!\mid\! x).
\end{align*}
By the assumption, $\sH_{\gamma}(x)\neq\emptyset$ and $\mathsf H_{\gamma}(x)=\sY$ since $\sH$ is symmetric. Thus, for any $h\in\sH$ with the constraint that $\sum_{y\in \sY}h(x,y)=0$,
\begin{align*}
& \Delta\sC_{\wt \Phi_{\rho}^{\mathrm{cstnd}},\sH}(h,x)  = \sC_{\wt \Phi_{\rho}^{\mathrm{cstnd}}}(h,x) - \sC^*_{\wt \Phi_{\rho}^{\mathrm{cstnd}}}(\sH, x) \\
& = \sum_{y\in \sY} \sup_{x':\norm*{x-x'}_p\leq \gamma}\sfp(y \!\mid\! x) \sum_{y'\neq y}\Phi_{\rho}\paren*{-h(x',y')} - \paren*{1-\max_{y\in \sY} \sfp(y \!\mid\! x)}\\
& \geq \sum_{y\in \sY} \sfp(y \!\mid\! x) \sum_{y'\neq y}\Phi_{\rho}\paren*{-h(x,y')} - \paren*{1-\max_{y\in \sY} \sfp(y \!\mid\! x)} \tag{lower bound the supremum}\\
& =  \sum_{y\in \sY} \paren*{1-\sfp(y \!\mid\! x)}\Phi_{\rho}\paren*{-h(x,y)} - \paren*{1-\max_{y\in \sY} \sfp(y \!\mid\! x)} \tag{swap $y$ and $y'$}\\
& \geq \sfp(\hh(x) \!\mid\! x)\mathds{1}_{h\not \in \sH_{\gamma}(x)} + 1-\sfp(\hh(x) \!\mid\! x) - \paren*{1-\max_{y\in \sY} \sfp(y \!\mid\! x)} \\
& = \max_{y\in \sY}\sfp(y \!\mid\! x) - \sfp(\hh(x) \!\mid\! x)\mathds{1}_{h \in \sH_{\gamma}(x)}\\
& = \Delta\sC_{\ell_{\gamma},\sH}(h,x) \tag{by Lemma~\ref{lemma:explicit_assumption_adv-mhcb} and $\mathsf H_{\gamma}(x) =\sY$}\\
& \geq \bracket*{\Delta\sC_{\ell_{\gamma},\sH}(h,x)}_{\e} \tag{$\bracket*{t}_{\e}\leq t$}\\
\end{align*}
for any $\e\geq 0$. Therefore, taking $\sP$ be the set of all distributions, $\sH$ be the symmetric hypothesis set, $\e=0$ and $\Psi(t)=t$ in Theorem~\ref{Thm:excess_bounds_Psi_adv_general}, or, equivalently, $\Gamma(t) = t$ in Theorem~\ref{Thm:excess_bounds_Gamma_adv_general}, we obtain for any hypothesis $h\in\sH$ and any distribution,
\begin{align*}
\sR_{\ell_{\gamma}}(h)- \sR_{\ell_{\gamma}}^*(\sH) \leq   \sR_{\wt{\Phi}_{\rho}^{\mathrm{cstnd}}}(h)-\sR_{\wt{\Phi}_{\rho}^{\mathrm{cstnd}}}^*(\sH)+\sM_{\wt{\Phi}_{\rho}^{\mathrm{cstnd}}}(\sH)-\sM_{\ell_{\gamma}}(\sH).
\end{align*}
\end{proof}
\restoreatoc

\chapter{Appendix to Chapter~\ref{ch4}}

\disableatoc
\section{Proofs of \texorpdfstring{$\sH$}{H}-consistency bounds
  for comp-sum losses (Theorem~\ref{Thm:bound_comp_sum})
  and tightness (Theorem~\ref{Thm:tightness-comp})}
\label{app:bound_comp-sum}

To begin with the proof, we first introduce some notation. We denote
by $p(x, y) = \sD(Y = y \!\mid\! X = x)$ the conditional probability
of $Y=y$ given $X = x$. The generalization error for a surrogate loss
can be rewritten as $ \sR_{\ell}(h) = \mathbb{E}_{X}
\bracket*{\sC_{\ell}(h, x)} $, where $\sC_{\ell}(h, x)$ is the
conditional $\ell$-risk, defined by
\begin{align*}
\sC_{\ell}(h, x) = \sum_{y\in \sY} p(x, y) \ell(h, x, y).
\end{align*}
We denote by $\sC_{\ell}^*(\sH,x) = \inf_{h\in
  \sH}\sC_{\ell}(h, x)$ the minimal conditional
$\ell$-risk. Then, the minimizability gap can be rewritten as follows:
\begin{align*}
\sM_{\ell}(\sH)
 = \sR^*_{\ell}(\sH) - \mathbb{E}_{X} \bracket* {\sC_{\ell}^*(\sH, x)}.
\end{align*}
We further refer to $\sC_{\ell}(h, x)-\sC_{\ell}^*(\sH,x)$ as
the calibration gap and denote it by $\Delta\sC_{\ell,\sH}(h, x)$. 

For any $h \in \sH$ and $x\in \sX$, by the symmetry and completeness
of $\sH$, we can always find a family of hypotheses $\curl*{\ov
  h_{\mu} \colon \mu \in \mathbb{R}}\subset \sH$ such that
$h_{\mu}(x,\cdot)$ take the following values:
\begin{align}
\label{eq:value-h-mu}
\ov h_{\mu}(x,y) = 
\begin{cases}
  h(x, y) & \text{if $y \not \in \curl*{y_{\max}, \hh(x)}$}\\
  \log\paren*{\exp\bracket*{h(x, y_{\max})} + \mu} & \text{if $y = \hh(x)$}\\
  \log\paren*{\exp\bracket*{h(x,\hh(x))} -\mu} & \text{if $y = y_{\max}$}.
\end{cases} 
\end{align}
Note that the hypotheses $\ov h_{\mu}$ has the following property:
\begin{align}
\label{eq:property-h-mu}
\sum_{y\in \sY}e^{h(x, y)} = \sum_{y\in \sY}e^{\ov h_{\mu}(x, y)},\, \forall \mu \in \mathbb{R}.
\end{align}
\begin{restatable}
  {lemma}{LemmaSup}
\label{lemma:lemma-sup}
Assume that $\sH$ is symmetric and complete. Then, for any $h\in \sX$
and $x\in \sX$, the following equality holds:
\begin{align*}
& \sC_{\ell_{\tau}^{\mathrm{comp}}}(h, x) - \inf_{\mu \in \Rset}\sC_{\ell_{\tau}^{\mathrm{comp}}}(\ov h_{\mu},x)\\
& = \sup_{\mu\in \Rset} \bigg\{p(x,y_{\max})\paren*{ \Phi^{\tau}\paren*{\frac{\sum_{y'\in \sY} e^{h(x, y')}}{e^{h(x, y_{\max})}}-1}-\Phi^{\tau}\paren*{\frac{\sum_{y'\in \sY} e^{h(x, y')}}{e^{h(x, \hh(x))-\mu}}-1}}\\
& \qquad +  p(x,\hh(x))\paren*{ \Phi^{\tau}\paren*{\frac{\sum_{y'\in \sY} e^{h(x, y')}}{e^{h(x, \hh(x))}}-1}-\Phi^{\tau}\paren*{\frac{\sum_{y'\in \sY} e^{h(x, y')}}{e^{h(x, y_{\max})+\mu}}-1}} \bigg\}\\
& = \begin{cases}
\frac{1}{\tau-1}\bracket*{\sum_{y'\in \sY}e^{h(x,y')}}^{1-\tau}\bracket*{\frac{\paren*{p(x,y_{\max})^{\frac1{2 - \tau }}+p(x,\hh(x))^{\frac1{2 - \tau }}}^{2-\tau}}{\paren*{e^{h(x,y_{\max})} + e^{h(x,\hh(x))}}^{1 - \tau }}-\frac{p(x,y_{\max})}{\paren*{e^{h(x,y_{\max})}}^{1 - \tau }} - \frac{p(x,\hh(x))}{\paren*{e^{h(x,\hh(x))}}^{1 - \tau }}} \\ \qquad \tau \in [0,2)/\curl*{1}\\
p(x,y_{\max})\log\bracket*{\frac{\paren*{e^{h(x,y_{\max})} + e^{h(x,\hh(x))}}p(x,y_{\max})}{e^{h(x,y_{\max})}\paren*{p(x,y_{\max})+p(x,\hh(x))}}}+p(x,\hh(x))\log\bracket*{\frac{\paren*{e^{h(x,y_{\max})} +e^{h(x,\hh(x))}}p(x,\hh(x))}{ e^{h(x,\hh(x))}\paren*{p(x,y_{\max})+p(x,\hh(x))}}} \\ \qquad \tau =1 \\
\frac{1}{\tau-1}\bracket*{\sum_{y'\in \sY}e^{h(x,y')}}^{1-\tau}\bracket*{\frac{p(x,y_{\max})}{\paren*{e^{h(x,y_{\max})} + e^{h(x,\hh(x))}}^{1 - \tau }}-\frac{p(x,y_{\max})}{\paren*{e^{h(x,y_{\max})}}^{1 - \tau }} - \frac{p(x,\hh(x))}{\paren*{e^{h(x,\hh(x))}}^{1 - \tau }} } \\ \qquad \tau \in [2,+ \infty).
\end{cases}
\end{align*}
\end{restatable}
\begin{proof}
For the comp-sum loss $\ell_{\tau}^{\rm{comp}}$, the conditional $\ell_{\tau}^{\rm{comp}}$-risk can be expressed as follows:
\begin{align*}
\sC_{\ell_{\tau}^{\rm{comp}}}(h, x)
& = \sum_{y\in \sY} p(x,y) \ell_{\tau}^{\rm{comp}}(h, x, y)\\
 & = \sum_{y\in \sY} p(x,y) \Phi^{\tau}\paren*{\sum_{y'\in \sY} e^{h(x, y') - h(x, y)}-1}\\
& = p(x,y_{\max}) \Phi^{\tau}\paren*{\sum_{y'\in \sY} e^{h(x, y') - h(x, y_{\max})}-1}+  p(x,\hh(x)) \Phi^{\tau}\paren*{\sum_{y'\in \sY} e^{h(x, y') - h(x, \hh(x))}-1}\\
& +\sum_{y\notin \curl*{y_{\max},\hh(x)}}p(x,y) \Phi^{\tau}\paren*{\sum_{y'\in \sY} e^{h(x, y') - h(x, y)}-1}.
\end{align*}
Therefore, by \eqref{eq:value-h-mu} and \eqref{eq:property-h-mu}, we
obtain the first equality. The second equality can be obtained by
taking the derivative with respect to $\mu$.
\end{proof}

\begin{restatable}
  {lemma}{LemmaInf}
\label{lemma:lemma-inf}
Assume that $\sH$ is symmetric and complete. Then, for any $h\in \sX$
and $x\in \sX$, the following equality holds
\begin{align*}
& \inf_{h\in \sH}\paren*{\sC_{\ell_{\tau}^{\mathrm{comp}}}(h, x) - \inf_{\mu \in \Rset}\sC_{\ell_{\tau}^{\mathrm{comp}}}(\ov h_{\mu},x)}\\
& = 
\begin{cases}
\frac{2^{2-\tau}}{1-\tau}\bracket*{\frac{p(x,y_{\max})+p(x,\hh(x))}{2}-\bracket*{\frac{p(x,y_{\max})^{\frac1{2 - \tau }}+p(x,\hh(x))^{\frac1{2 - \tau }}}{2}}^{2 - \tau }} & \tau \in [0,1)\\
p(x,y_{\max})\log\bracket*{\frac{2p(x,y_{\max})}{p(x,y_{\max})+p(x,\hh(x))}} + p(x,\hh(x))\log\bracket*{\frac{2p(x,\hh(x))}{p(x,y_{\max})+p(x,\hh(x))}} & \tau =1 \\
\frac{2}{(\tau-1)n^{\tau-1}}\paren*{\bracket*{\frac{p(x,y_{\max})^{\frac1{2-\tau}}+p(x,\hh(x))^{\frac1{2-\tau}}}{2}}^{2-\tau}-\frac{p(x,y_{\max})+p(x,\hh(x))}{2}} & \tau \in (1,2)\\
\frac{1}{(\tau-1)n^{\tau-1}}
\paren*{p(x,y_{\max}) - p(x,\hh(x))} & \tau \in [2,+ \infty).
\end{cases}
\end{align*}
\end{restatable}
\begin{proof}
By using Lemma~\ref{lemma:lemma-sup} and taking infimum with respect to $e^{h(x,1)},\ldots, e^{h(x,n)}$, the equality is proved directly.
\end{proof}
Let $\alpha=p(x,y_{\max})+p(x,\hh(x))\in [0,1]$ and
$\beta=p(x,y_{\max}-p(x,\hh(x)))\in [0,1]$. Then, using the fact that
$p(x,y_{\max})= \frac{\alpha+\beta}{2}$ and
$p(x,\hh(x))= \frac{\alpha-\beta}{2}$, we have
\begin{align}
\label{eq:Psi-tau}
\inf_{h\in \sH}\paren*{\sC_{\ell_{\tau}^{\mathrm{comp}}}(h, x) - \inf_{\mu \in \Rset}\sC_{\ell_{\tau}^{\mathrm{comp}}}(\ov h_{\mu},x)} &= \Psi_{\tau}(\alpha,\beta)\\
&= 
\begin{cases}
\frac{2^{1-\tau}}{1-\tau}\bracket*{\alpha-\bracket*{\frac{\paren*{\alpha+\beta}^{\frac1{2 - \tau }}+\paren*{\alpha-\beta}^{\frac1{2 - \tau }}}{2}}^{2 - \tau }} & \tau \in [0,1)\\
\frac{\alpha+\beta}{2}\log\bracket*{\frac{\alpha+\beta}{\alpha}} + \frac{\alpha-\beta}{2}\log\bracket*{\frac{\alpha-\beta}{\alpha}} & \tau =1 \\
\frac{1}{(\tau-1)n^{\tau-1}}\paren*{\bracket*{\frac{\paren*{\alpha+\beta}^{\frac1{2 - \tau }}+\paren*{\alpha-\beta}^{\frac1{2 - \tau }}}{2}}^{2 - \tau}-\alpha} & \tau \in (1,2)\\
\frac{1}{(\tau-1)n^{\tau-1}}\,
\beta & \tau \in [2,+ \infty).
\end{cases}
\end{align}
By taking the partial derivative of $\Psi_{\tau}(\alpha,\cdot)$ with respect to $\alpha$ and analyzing the minima, we obtain the following result.
\begin{restatable}
  {lemma}{LemmaPsiTau}
\label{lemma:Psi-tau}
For any $\tau \in [0,+\infty)$ and $\alpha\in [0,1]$, the following inequality holds for any $\beta\in [0,1]$,
\begin{align*}
\Psi_{\tau}(\alpha,\beta)\geq \Psi_{\tau}(1,\beta)= \sT_{\tau}(\beta)= \begin{cases}
\frac{2^{1-\tau}}{1-\tau}\bracket*{1 -\bracket*{\frac{\paren*{1 + \beta}^{\frac1{2 - \tau }} +  \paren*{1 - \beta}^{\frac1{2 - \tau }}}{2}}^{2 - \tau }} & \tau \in [0,1)\\
\frac{1+\beta}{2}\log\bracket*{1+\beta} + \frac{1-\beta}{2}\log\bracket*{1-\beta} & \tau =1 \\
\frac{1}{(\tau-1)n^{\tau-1}}\bracket*{\bracket*{\frac{\paren*{1 + \beta}^{\frac1{2 - \tau }} +  \paren*{1 - \beta}^{\frac1{2 - \tau }}}{2}}^{2 - \tau }-1} & \tau \in (1,2)\\
\frac{1}{(\tau-1)n^{\tau-1}}\,
\beta & \tau \in [2,+ \infty).
\end{cases}
\end{align*}
\end{restatable}
We denote by $\sT_{\tau}(\beta)= \Psi_{\tau}(1,\beta)$ and call it the
$\sH$-consistency comp-sum transformation, and denote by
$\Gamma_{\tau}$ the inverse of $\sT_{\tau}$: $\Gamma_{\tau}(t) =
\sT_{\tau}^{-1}(t)$. We then present the proofs of
Theorem~\ref{Thm:bound_comp_sum} and Theorem~\ref{Thm:tightness-comp} in
the below.  \ignore{By using standard inequalities, $\Gamma_{\tau}$
  can be upper-bounded as follows
\begin{align*}
\Gamma_{\tau}(t)\leq \begin{cases}
\sqrt{2^{\frac{\tau}{2-\tau}}(2-\tau) t} & \tau\in [0,1)\\
\sqrt{2^{2-\tau}(2-\tau)n^{\frac{(\tau-1)(4-\tau)}{2-\tau}} t } & \tau\in [1,2) \\
\ignore{\sqrt{\frac{2n^{\frac{2\tau-2}{2-\tau}}}{2-\tau} t } & \tau\in [1,2) \\}
(\tau - 1) n^{\tau - 1} t & \tau \in [2,+ \infty).
\end{cases}
\end{align*}}

\BoundCompSum*
\begin{proof}
Using previous lemmas, we can lower bound the calibration gap of
comp-sum losses as follows, for any $h\in \sH$,
\begin{align*}
& \sC_{\ell_{\tau}^{\mathrm{comp}}}(h, x) - \sC^*_{\ell_{\tau}^{\mathrm{comp}}}\paren*{\sH,x} \\
& \geq \sC_{\ell_{\tau}^{\mathrm{comp}}}(h, x) - \inf_{\mu \in \Rset}\sC_{\ell_{\tau}^{\mathrm{comp}}}(\ov h_{\mu},x) \\
& \geq \inf_{h\in \sH}\paren*{\sC_{\ell_{\tau}^{\mathrm{comp}}}(h, x) - \inf_{\mu \in \Rset}\sC_{\ell_{\tau}^{\mathrm{comp}}}(\ov h_{\mu},x)}\\
& =
\begin{cases}
\frac{2^{2-\tau}}{1-\tau}\bracket*{\frac{p(x,y_{\max})+p(x,\hh(x))}{2}-\bracket*{\frac{p(x,y_{\max})^{\frac1{2 - \tau }}+p(x,\hh(x))^{\frac1{2 - \tau }}}{2}}^{2 - \tau }} & \tau \in [0,1)\\
p(x,y_{\max})\log\bracket*{\frac{2p(x,y_{\max})}{p(x,y_{\max})+p(x,\hh(x))}} + p(x,\hh(x))\log\bracket*{\frac{2p(x,\hh(x))}{p(x,y_{\max})+p(x,\hh(x))}} & \tau =1 \\
\frac{2}{(\tau-1)n^{\tau-1}}\paren*{\bracket*{\frac{p(x,y_{\max})^{\frac1{2-\tau}}+p(x,\hh(x))^{\frac1{2-\tau}}}{2}}^{2-\tau}-\frac{p(x,y_{\max})+p(x,\hh(x))}{2}} & \tau \in (1,2)\\
\frac{1}{(\tau-1)n^{\tau-1}}
\paren*{p(x,y_{\max}) - p(x,\hh(x))} & \tau \in [2,+ \infty)
\end{cases}
\tag{By Lemma~\ref{lemma:lemma-inf}}\\
& \geq \sT_{\tau}\paren*{p(x,y_{\max}) - p(x,\hh(x))} \tag{By \eqref{eq:Psi-tau} and Lemma~\ref{lemma:Psi-tau}}\\
& = \sT_{\tau}\paren*{\sC_{\ell_{0-1}}(h, x) - \sC^*_{\ell_{0-1}}\paren*{\sH,x}} \tag{by \citep[Lemma~ 3]{awasthi2022multi}}
\end{align*}
Therefore, taking $\sP$ be the set of all distributions, $\sH$ be the
symmetric and complete hypothesis set, $\e=0$ and
$\Psi(\beta)= \sT_{\tau}(\beta)$ in \citep[Theorem~
  4]{awasthi2022multi}, or, equivalently, $\Gamma(t) =
\Gamma_{\tau}(t)$ in \citep[Theorem~
  5]{awasthi2022multi}, we obtain for any hypothesis
$h\in\sH$ and any distribution,
\begin{align*}
\sR_{\ell_{0-1}}(h)-\sR^*_{\ell_{0-1}}(\sH)\leq \Gamma_{\tau}\paren*{\sR_{\ell_{\tau}^{\mathrm{comp}}}(h)-\sR^*_{\ell_{\tau}^{\mathrm{comp}}}(\sH)+\sM_{\ell_{\tau}^{\mathrm{comp}}}(\sH)}-\sM_{\ell_{0-1}}(\sH).
\end{align*}
\end{proof}

\TightnessComp*
\begin{proof}
For any $\beta\in [0,1]$, we consider the distribution that
concentrates on a singleton $\curl*{x_0}$ and satisfies
$p(x_0,1)= \frac{1+\beta}{2}$, $p(x_0,2)= \frac{1-\beta}{2}$,
$p(x_0,y)=0,\,3\leq y\leq n$. We take $h_{\tau}\in \sH$ such that
$e^{h_{\tau}(x,1)}=e^{h_{\tau}(x,2)}$, $e^{h_{\tau}(x,y)}=0,\, 3\leq
y\leq n$.  \ignore{We take the hypotheses $\curl*{h_{\tau}:\tau
    \in[0,\infty)}\in \sH$ using the following criteria, when $\tau\in
       [0,1]$, take $h_{\tau}\in \sH$ such that
       $e^{h_{\tau}(x,1)}=e^{h_{\tau}(x,2)}$, $e^{h_{\tau}(x,y)}=0,\,
       3\leq y\leq n$; when $\tau\in (1,+\infty)$, take
       $h_{\tau}\in \sH$ such that
       $e^{h_{\tau}(x,1)}=e^{h_{\tau}(x,2)}= \cdots=e^{h_{\tau}(x,n-1)}=e^{h_{\tau}(x,n)}$.}
  Then,
\begin{align*}
\sR_{\ell_{0-1}}(h_{\tau})- \sR_{\ell_{0-1},\sH}^*+\sM_{\ell_{0-1},\sH} & = \sR_{\ell_{0-1}}(h_{\tau}) - \mathbb{E}_{X} \bracket* {\sC^*_{\ell_{0-1}}(\sH,x)}= \sC_{\ell_{0-1}}(h_{\tau},x_0) - \sC^*_{\ell_{0-1}}\paren*{\sH,x_0}= \beta
\end{align*}
and for any $\tau\in[0,1]$,
\begin{align*}
& \sR_{\ell_{\tau}^{\rm{comp}}}(h_{\tau}) - \sR_{\ell_{\tau}^{\rm{comp}}}^*(\sH)+ \sM_{\ell_{\tau}^{\rm{comp}}}(\sH)\\ & = \sR_{\ell_{\tau}^{\rm{comp}}}(h_{\tau}) - \mathbb{E}_{X} \bracket* {\sC^*_{\ell_{\tau}^{\rm{comp}}}(\sH,x)}\\
& = \sC_{\ell_{\tau}^{\mathrm{comp}}}(h_{\tau},x_0) - \sC^*_{\ell_{\tau}^{\mathrm{comp}}}\paren*{\sH,x_0}\\
& =p(x_0,1) \ell_{\tau}^{\rm{comp}}(h_{\tau}, x_0, 1) + p(x_0,2) \ell_{\tau}^{\rm{comp}}(h_{\tau}, x_0, 2)\\
&\qquad - \inf_{h\in \sH} \bracket*{p(x_0,1) \ell_{\tau}^{\rm{comp}}(h, x_0, 1) + p(x_0,2) \ell_{\tau}^{\rm{comp}}(h, x_0, 2)}\\
& = \sT_{\tau}(\beta), \tag{By \eqref{eq:comp-loss} and \eqref{eq:comp-sum-Cstar}}
\end{align*}
which completes the proof.
\end{proof}

\section{Approximations of \texorpdfstring{$\sT_{\tau}$}{T} and \texorpdfstring{$\Gamma_{\tau}$}{Gamma}}
\label{app:Gamma-upper-bound}
In this section, we show how $\sT_{\tau}$ can be lower bounded by its polynomial approximation $\wt \sT_{\tau}$, and accordingly, $\Gamma_{\tau}$ can then be upper-bounded by $\wt \Gamma_{\tau}= \wt \sT_{\tau}^{-1}$. By analyzing the Taylor expansion, we obtain for any $\beta\in [-1,1]$,
\begin{equation}
\label{eq:inequality-auxiliary-1}
\begin{aligned}
\paren*{\frac{(1+\beta)^r+(1-\beta)^r}{2}}^{\frac1r} & \geq 1+\frac{\beta^2}{2}\paren*{1-\frac1r}, \text{ for all } r\geq 1\\
\paren*{\frac{(1+\beta)^r+(1-\beta)^r}{2}}^{\frac1r} & \leq 1 - \frac{\beta^2}{2}\paren*{1-r}, \text{ for all } \frac12\leq r\leq 1.
\end{aligned}
\end{equation}
and 
\begin{align}
\label{eq:inequality-auxiliary-2}
\frac{1+\beta}{2}\log\bracket*{1+\beta} + \frac{1-\beta}{2}\log\bracket*{1-\beta}\geq \frac{\beta^2}{2}.
\end{align}
For $\tau \in [0,1)$, we have
\begin{align*}
\sT_{\tau}(\beta) 
& = \frac{2^{1-\tau}}{1-\tau}\bracket*{1-\bracket*{\frac{\paren*{1+\beta}^{\frac1{2 - \tau }}+\paren*{1-\beta}^{\frac1{2 - \tau }}}{2}}^{2 - \tau }}\\
& \geq \frac{2^{1-\tau}}{1-\tau}\bracket*{1-\bracket*{1 - \frac{\beta^2}{2}\frac{1-\tau}{2-\tau}}} \tag{using \eqref{eq:inequality-auxiliary-1} with $r= \frac{1}{2-\tau}\in\bracket*{\frac12,1}$}\\
& = \frac{\beta^2}{2^{\tau}(2-\tau)}\\
& = \wt \sT_{\tau}(\beta).
\end{align*}
For $\tau\in (1,2)$, we have
\begin{align*}
\sT_{\tau}(\beta) 
& = \frac{1}{(\tau-1)n^{\tau-1}}\bracket*{\bracket*{\frac{\paren*{1+\beta}^{\frac1{2 - \tau }}+\paren*{1-\beta}^{\frac1{2 - \tau }}}{2}}^{2 - \tau }-1}\\
& \geq \frac{1}{(\tau-1)n^{\tau-1}}\bracket*{\bracket*{1 + \frac{\beta^2}{2}(\tau-1)}-1} \tag{using \eqref{eq:inequality-auxiliary-1} with $r= \frac{1}{2-\tau}\geq 1$}\\
& = \frac{\beta^2}{2n^{\tau-1}}\\
& = \wt \sT_{\tau}(\beta).
\end{align*}
For $\tau=1$, we have
\begin{align*}
\sT_{\tau}(\beta) & = \frac{1+\beta}{2}\log\bracket*{1+\beta} + \frac{1-\beta}{2}\log\bracket*{1-\beta}\\
&\geq \frac{\beta^2}{2} \tag{using \eqref{eq:inequality-auxiliary-2}}\\
& = \wt \sT_{\tau}(\beta).
\end{align*}
For $\tau\geq 2$,  $\sT_{\tau}(\beta)= \frac{\beta}{(\tau-1)n^{\tau-1}}= \wt \sT_{\tau}(\beta)$.
Therefore, for any $\tau\in [0,+\infty)$, 
\begin{align*}
\sT_{\tau}(\beta)\geq \wt\sT_{\tau}(\beta)=
\begin{cases}
\frac{\beta^2}{2^{\tau}(2-\tau)} & \tau \in [0,1)\\
\frac{\beta^2}{2n^{\tau-1}} & \tau \in [1,2)\\
\frac{\beta}{(\tau-1)n^{\tau-1}}& \tau \in [2,+ \infty).
\end{cases}
\end{align*}
Furthermore, by using Taylor expansion, we have
\begin{align*}
\lim_{\beta \to 0^{+}}\frac{\wt\sT_{\tau}(\beta)}{\sT_{\tau}(\beta)} = c>0 \text{ for some constant $c>0$}.
\end{align*}
Thus, the order of polynomials $\wt\sT_{\tau}(\beta)$ is tightest.
Since
$\Gamma_{\tau}= \sT_{\tau}^{-1} $ and $\wt \Gamma_{\tau}=
\wt \sT_{\tau}^{-1}$, we also obtain for any $\tau\in [0,+\infty)$, 
\begin{align*}
\Gamma_{\tau}(t)\leq  \wt \Gamma_{\tau}(t)= \wt\sT_{\tau}^{-1}(t)= \begin{cases}
\sqrt{2^{\tau}(2-\tau) t} & \tau\in [0,1)\\
\sqrt{2n^{\tau-1} t } & \tau\in [1,2) \\
(\tau - 1) n^{\tau - 1} t & \tau \in [2,+ \infty).
\end{cases}
\end{align*}

\section{Characterization of minimizability gaps
  (proofs of Theorem~\ref{Thm:gap-upper-bound} and
  Theorem~\ref{Thm:gap-upper-bound-determi})}
\label{app:gap-upper-bound}
\GapUpperBound*
\begin{proof}
Using the fact that $\Phi^{\tau}$ is concave and non-decreasing, by
\eqref{eq:concave-Phi1}, we can then upper-bound the minimizability
gaps for different $\tau \geq 0$ as follows,
\begin{align*}
\sM_{\ell_{\tau}^{\rm{comp}}}(\sH)
\leq \Phi_{\tau}\paren*{\sR^*_{\ell_{\tau=0}^{\rm{comp}}}(\sH)} - \E_x[\sC^*_{\ell_{\tau}^{\rm{comp}}}(\sH, x)].
\end{align*}
By definition, the conditional $\ell_{\tau}^{\rm{comp}}$-risk can be expressed as follows: 
\begin{align}
\label{eq:cond-comp-sum}
\sC_{\ell_{\tau}^{\rm{comp}}}(h, x)  =  \sum_{y\in \sY} p(x,y) \Phi_{\tau}\paren*{\sum_{y'\neq y}\exp\paren*{h(x,y')-h(x,y)}}.
\end{align}
Note that $\sC_{\ell_{\tau}^{\rm{comp}}}(h, x)$ is convex and differentiable with respect to $h(x,y)$s, by taking the partial derivative and using the derivative of $\Phi_{\tau}$ given in \eqref{eq:Phi1-derivative}, we obtain
\begin{equation}
\label{eq:cond-comp-sum-deriv}
\begin{aligned}
&\frac{\partial \sC_{\ell_{\tau}^{\rm{comp}}}(h, x)}{\partial h(x,y)}\\
& =p(x,y)\frac{\partial \Phi_{\tau}}{\partial u}\paren*{\sum_{y'\neq y}\exp\paren*{h(x,y')-h(x,y)}}\paren*{-\sum_{y'\neq y}\exp\paren*{h(x,y')-h(x,y)}}\\
& + \sum_{y'\neq y} p(x,y')\frac{\partial \Phi_{\tau}}{\partial u}\paren*{\sum_{y''\neq y'}\exp\paren*{h(x,y'')-h(x,y')}}\paren*{\exp\paren*{h(x,y)-h(x,y')}}\\
& =p(x,y)\frac{-\sum_{y'\neq y}\exp\paren*{h(x,y')-h(x,y)}}{\bracket*{\sum_{y'\in \sY}\exp\paren*{h(x,y')-h(x,y)}}^{\tau}} + \sum_{y'\neq y} p(x,y')\frac{\exp\paren*{h(x,y)-h(x,y')}}{\bracket*{\sum_{y''\in \sY}\exp\paren*{h(x,y'')-h(x,y')}}^{\tau}}
\end{aligned}
\end{equation}
Let $\sS(x,y) = \sum_{y'\in \sY}\exp\paren*{h(x,y')-h(x,y)}$. Then,
$\exp\paren*{h(x,y)-h(x,y')} = \frac{\sS(x,y')}{\sS(x,y)}$ and thus
\eqref{eq:cond-comp-sum-deriv} can be written as
\begin{align}
\label{eq:cond-comp-sum-deriv-S}
\frac{\partial \sC_{\ell_{\tau}^{\rm{comp}}}(h, x)}{\partial h(x,y)} = p(x,y)\frac{-\sS(x,y)+1}{\sS(x,y)^{\tau}} + \sum_{y'\neq y} p(x,y') \frac{1}{\sS(x,y')^{\tau-1}\sS(x,y)}
\end{align}
It is straightforward to verify that 
\begin{align}
\label{eq:solution}
\sS^*(x,y) =
\begin{cases}
\frac{\sum_{y'\in \sY}p(x,y')^{\frac{1}{2-\tau}}}{p(x,y)^{\frac{1}{2-\tau}}} & \tau \neq 2\\
\frac{1}{\mathds{1}_{y= \argmax_{y'\in \sY}p(x,y')}} & \tau =2
\end{cases}
\end{align}
satisfy
\begin{align*}
\frac{\partial \sC_{\ell_{\tau}^{\rm{comp}}}(h, x)}{\partial h(x,y)} = 0, \forall y\in \sY.
\end{align*}
When $\sH$ is symmetric and complete, \eqref{eq:solution} can be
attained by some $h^*\in \sH$.  Since
$\sC_{\ell_{\tau}^{\rm{comp}}}(h, x)$ is convex and differentiable with
respect to $h(x,y)$s, we know that $h^*$ achieves the minimum of
$\sC_{\ell_{\tau}^{\rm{comp}}}(h, x)$ within $\sH$.  Then,
\begin{align*}
\sC^*_{\ell_{\tau}^{\rm{comp}}}(\sH, x) 
& =  \sC_{\ell_{\tau}^{\rm{comp}}}(h^*, x)\\
& = \sum_{y\in \sY} p(x,y) \Phi_{\tau}\paren*{\sum_{y'\neq y}\exp\paren*{h^*(x,y')-h^*(x,y)}}\\
& = \sum_{y\in \sY} p(x,y) \Phi_{\tau}\paren*{\sS^*(x,y)-1}\tag{by the def. of $\sS(x,y)$.}\\
& = \begin{cases}
\sum_{y\in \sY} p(x,y) \frac{1}{1 - \tau} \paren*{\sS^*(x,y)^{1 - \tau} - 1} & \tau \geq 0, \tau\neq1\\
\sum_{y\in \sY} p(x,y) \log\bracket*{\sS^*(x,y) } & \tau=1\\
\end{cases} \tag{by \eqref{eq:comp-loss}.}\\
&  = \begin{cases}
\frac{1}{1 - \tau} \paren*{\bracket*{\sum_{y\in \sY}p(x,y)^{\frac{1}{2-\tau}}}^{2 - \tau} - 1} & \tau\geq 0, \tau\neq1, \tau \neq 2\\
-\sum_{y\in \sY} p(x,y) \log\bracket*{p(x,y) } & \tau=1\\
1 - \max_{y\in \sY}p(x,y) & \tau =2.
\end{cases}
\tag{by \eqref{eq:solution}}
\end{align*}
\ignore{ Since $\Phi_{\tau}= \int_{0}^u \frac{1}{(1+t)^{\tau}}\,dt$ is
  decreasing with respect to $\tau\geq 0$, for any $h\in \sH$ and
  $x\in \sX$, $\sC_{\ell_{\tau}^{\rm{comp}}}(h, x)$ expressed in
  \eqref{eq:cond-comp-sum} is a decreasing function of $\tau \geq 0$,
  we obtain that $\sC^*_{\ell_{\tau}^{\rm{comp}}}(\sH, x)= \inf_{h\in
    \sH}\sC_{\ell_{\tau}^{\rm{comp}}}(h, x)$ is also a decreasing
  function of $\tau \geq 0$, which concludes the proof.}
\end{proof}

\GapUpperBoundDetermi*
\begin{proof}
Using the fact that $\Phi_{\tau}$ is concave and non-decreasing, by
\eqref{eq:concave-Phi1}, we can then upper-bound the minimizability
gaps for different $\tau \geq 0$ as follows,
\begin{align*}
\sM_{\ell_{\tau}^{\rm{comp}}}(\sH)
\leq \Phi_{\tau}\paren*{\sR^*_{\ell_{\tau=0}^{\rm{comp}}}(\sH)} - \E_x[\sC^*_{\ell_{\tau}^{\rm{comp}}}(\sH, x)].
\end{align*}
Let $y_{\max} = \argmax p(x,y)$. By definition, for any deterministic distribution, the conditional $\ell_{\tau}^{\rm{comp}}$-risk can be expressed as follows:
\begin{equation}
\label{eq:cond-comp-sum-determi}
\begin{aligned}
\sC_{\ell_{\tau}^{\rm{comp}}}(h, x)  & =  \Phi_{\tau}\paren*{\sum_{y'\neq y_{\max}}\exp\paren*{h(x,y')-h(x,y_{\max})}}\\
& = 
\begin{cases}
\frac{1}{1 - \tau} \paren*{\paren*{1 + \frac{\sum_{y'\neq y_{\max}}\exp\paren*{h(x,y')}}{\exp(h(x,y_{\max}))}}^{1 - \tau} - 1} \\
\log\paren*{1 + \frac{\sum_{y'\neq y_{\max}}\exp\paren*{h(x,y')}}{\exp(h(x,y_{\max}))}}.
\end{cases}
\end{aligned}
\end{equation}
Since for any $\tau>0$, $\Phi_{\tau}$ is increasing, under the
assumption of $\sH$, $h^*$ that satisfies
\begin{align}
\label{eq:solution-determi}
h^*(x,y) = 
\begin{cases}
\Lambda & y = y_{\max}\\
-\Lambda & \text{otherwise}
\end{cases}
\end{align}
achieves the minimum of $\sC_{\ell_{\tau}^{\rm{comp}}}(h, x)$ within $\sH$.
Then, 
\begin{align*}
\sC^*_{\ell_{\tau}^{\rm{comp}}}(\sH, x) 
& =  \sC_{\ell_{\tau}^{\rm{comp}}}(h^*, x)\\
& = 
\begin{cases}
\frac{1}{1 - \tau} \paren*{\paren*{1 + \frac{\sum_{y'\neq y_{\max}}\exp\paren*{h^*(x,y')}}{\exp(h^*(x,y_{\max}))}}^{1 - \tau} - 1} \\
\log\paren*{1 + \frac{\sum_{y'\neq y_{\max}}\exp\paren*{h^*(x,y')}}{\exp(h^*(x,y_{\max}))}}.
\end{cases}\\
& = \begin{cases}
\frac{1}{1 - \tau} \paren*{\bracket*{1 + e^{-2 \Lambda}(n - 1)}^{1 - \tau} - 1} & \tau\geq 0, \tau\neq1\\
\log\bracket*{1 + e^{-2 \Lambda}(n - 1) } & \tau=1.
\end{cases}
\tag{by \eqref{eq:solution-determi}}
\end{align*}
Since $\sC^*_{\ell_{\tau}^{\rm{comp}}}(\sH, x) $ is independent of $x$, we have $\E_x[\sC^*_{\ell_{\tau}^{\rm{comp}}}(\sH, x)]= \sC^*_{\ell_{\tau}^{\rm{comp}}}(\sH, x)$ and thus
\begin{align*}
\sM_{\ell_{\tau}^{\rm{comp}}}(\sH)
\leq \Phi_{\tau}\paren*{\sR^*_{\ell_{\tau=0}^{\rm{comp}}}(\sH)} - \sC^*_{\ell_{\tau}^{\rm{comp}}}(\sH, x),
\end{align*}
which concludes the proof.
\end{proof}

\section{Proof of Lemma~\ref{lemma:lemma-compare}}
\label{app:lemma-compare}
\LemmaCompare*
\begin{proof}
For any $u_1 \geq u_2\geq 0$ and $\tau \neq 1$, we have
\begin{align*}
& \frac{\partial \paren*{\Phi_{\tau}(u_1)-\Phi_{\tau}(u_2)}}{\partial \tau}\\
& = \frac{\paren*{(1 + u_1)^{1 - \tau} - (1 + u_2)^{1 - \tau}}}{(1-\tau)^2} + \frac{1}{1-\tau}\paren*{\paren*{1 + u_2}^{1-\tau}\log(1+u_2)-\paren*{1 + u_1}^{1-\tau}\log(1+u_1)}\\
& = \frac{g(u_1,\tau) - g(u_2,\tau)}{(1-\tau)^2}
\end{align*}
where $g(t,\tau) =(1 + t)^{1 - \tau} - (1-\tau)\paren*{1 +
  t}^{1-\tau}\log(1+t)$. By taking the partial derivative, we obtain
for any $\tau \neq 1$ and $t\geq 0$,
\begin{align*}
\frac{\partial g}{\partial t} 
 = -(1-\tau)^2(1+t)^{\tau}\log (1+t)\leq  0
\end{align*}
Therefore, for any $u_1 \geq u_2\geq 0$ and $\tau \neq 1$,
$g(u_1,\tau) \leq  g(u_2,\tau)$ and
\begin{align*}
\frac{\partial \paren*{\Phi_{\tau}(u_1)-\Phi_{\tau}(u_2)}}{\partial \tau} \leq 0,
\end{align*}
which implies that for any $u_1 \geq u_2\geq 0$ and $\tau\neq 1$,
$\Phi_{\tau}(u_1)-\Phi_{\tau}(u_2)$ is a non-increasing function of
$\tau$.  Moreover, since for $x\geq 1$, $\frac{1}{\tau-1}
\paren*{x^{\tau-1} - 1} \to \log(x)$ as $\tau \to 1$, we know that for
any $u_1 \geq u_2\geq 0$, $\Phi_{\tau}(u_1)-\Phi_{\tau}(u_2)$ is
continuous with respect to $\tau=1$. Therefore, we conclude that for
any $u_1 \geq u_2\geq 0$, $\Phi_{\tau}(u_1)-\Phi_{\tau}(u_2)$ is
non-increasing with respect to $\tau$.
\end{proof}

\section{Proof of adversarial \texorpdfstring{$\sH$}{H}-consistency bound for adversarial comp-sum losses (Theorem~\ref{Thm:bound_comp_rho_adv})}
\label{app:deferred_proofs_adv_comp}

\BoundCompRhoAdv*
\begin{proof}
Let $\ov \sH_\gamma(x) = \curl*{h\in\sH:\inf_{x':\norm*{x-x'}\leq
    \gamma}\rho_h(x', \hh(x)) > 0}$ and $p(x)=(p(x, 1), \ldots, p(x,
c))$.  For any $x\in \sX$ and $h\in \sH$, we define $h\paren*{x,
  \curl*{1}^h_x}, h\paren*{x, \curl*{2}^h_x},\ldots, h\paren*{x,
  \curl*{c}^h_x}$ by sorting the scores $\curl*{h(x, y):y\in \sY}$ in
increasing order, and $p_{\bracket*{1}}(x), p_{\bracket*{2}}(x),
\ldots, p_{\bracket*{c}}(x)$ by sorting the probabilities $\curl*{p(x,
  y):y\in \sY}$ in increasing order.  Note
$\curl*{c}^h_x= \hh(x)$. Since $\sH$ is symmetric and locally
$\rho$-consistent, for any $x\in \sX$, there exists a hypothesis $h^*
\in \sH$ such that
\begin{align*}
& \inf_{x'\colon \norm*{x - x'}\leq \gamma}\abs*{h^*(x', i) - h^*(x', j)}  \geq \rho, \forall i\neq j \in \sY\\
& p(x,\curl*{k}^{h^*}_{x'})  =p_{\bracket*{k}}(x), \forall x'\in \curl*{x'\colon \norm*{x - x'}\leq \gamma}, \forall k\in
\sY.
\end{align*}
Then, we have
\begin{align*}
&\sC^*_{\wt \ell^{\mathrm{comp}}_{\tau,\rho}}(\sH,x)\\
& \leq \sC_{\wt \ell^{\mathrm{comp}}_{\tau,\rho}}(h^*,x)\nonumber\\
& = \sum_{y\in \sY}  \sup_{x':\norm*{x-x'}\leq \gamma}p(x, y)\Phi^{\tau}\paren*{ \sum_{y'\neq y} \Phi_{\rho}\paren*{h^*(x',y)-h^*(x',y')}}\nonumber\\
& =
\sum_{i=1}^c \sup_{x':\norm*{x-x'}\leq \gamma} p(x,\curl*{i}^{h^*}_{x'})\Phi^{\tau} \bracket[\Bigg]{\sum_{j=1}^{i-1}\Phi_{\rho}\paren*{h^*(x',\curl*{i}^{h^*}_{x'})-h^*(x',\curl*{j}^{h^*}_{x'})}\\
& \qquad + \sum_{j=i+1}^{c}\Phi_{\rho}\paren*{h^*(x',\curl*{i}^{h^*}_{x'})-h^*(x',\curl*{j}^{h^*}_{x'})}}\\
& = \sum_{i=1}^c \sup_{x':\norm*{x-x'}\leq \gamma} p(x,\curl*{i}^{h^*}_{x'})\Phi^{\tau} \bracket*{\sum_{j=1}^{i-1}\Phi_{\rho}\paren*{h^*(x',\curl*{i}^{h^*}_{x'})-h^*(x',\curl*{j}^{h^*}_{x'})} + c-i} \tag{$\Phi_{\rho}(t)=1,\,\forall t\leq 0$}\\
& = \sum_{i=1}^c \sup_{x':\norm*{x-x'}\leq \gamma} p(x,\curl*{i}^{h^*}_{x'})\Phi^{\tau} (c-i) \tag{$\inf_{x':\norm*{x-x'}\leq \gamma}\abs*{h^*(x',i)-h^*(x',j)}\geq \rho$ for any $i\neq j$ and $\Phi_{\rho}(t)=0,\,\forall t\geq \rho$}\\
& = \sum_{i=1}^c p_{\bracket*{i}}(x) \Phi^{\tau}(c-i) \tag{$p(x,\curl*{k}^{h^*}_{x'})  =p_{\bracket*{k}}(x), \forall x'\in \curl*{x'\colon \norm*{x - x'}\leq \gamma}, \forall k\in
\sY$}.
\end{align*}

Note $\ov \sH_{\gamma}(x)\neq\emptyset$ under the assumption. Then, use the derivation above, we obtain
\begin{align*}
& \Delta\sC_{\wt \ell^{\mathrm{comp}}_{\tau,\rho},\sH}(h, x)\\
& =  \sum_{i=1}^c \sup_{x':\norm*{x-x'}\leq \gamma} p(x,\curl*{i}^{h}_{x'})\Phi^{\tau} \bracket*{\sum_{j=1}^{i-1}\Phi_{\rho}\paren*{h(x',\curl*{i}^{h}_{x'})-h(x',\curl*{j}^{h}_{x'})} + c-i} -  \sum_{i=1}^c p_{\bracket*{i}}(x) \Phi^{\tau}(c-i)\\
&\geq \Phi^{\tau}(1)\, p(x, \hh(x))\mathds{1}_{h\not \in \ov \sH_{\gamma}(x)} +  \sum_{i=1}^c \sup_{x':\norm*{x-x'}\leq \gamma} p(x,\curl*{i}^{h}_{x'}) \Phi^{\tau}\paren*{c-i} -  \sum_{i=1}^c p_{\bracket*{i}}(x) \Phi^{\tau}(c-i) \tag{$\Phi_{\rho}$ is non-negative and $\Phi^{\tau}$ is non-decreasing}\\
& \geq \Phi^{\tau}(1)\,p(x, \hh(x))\mathds{1}_{h\not \in \ov \sH_{\gamma}(x)} + \sum_{i=1}^c p(x,\curl*{i}^{h}_{x}) \Phi^{\tau}(c-i) - \sum_{i=1}^c p_{\bracket*{i}}(x) \Phi^{\tau}(c-i) \tag{$\sup_{x':\norm*{x-x'}\leq \gamma} p(x,\curl*{i}^{h}_{x'})\geq p(x,\curl*{i}^{h}_{x} $}\\
& = \Phi^{\tau}(1)\,p(x, \hh(x))\mathds{1}_{h\not \in \ov \sH_{\gamma}(x)} +\Phi^{\tau}(1) \paren*{\max_{y\in \sY}p(x, y) - p(x, \hh(x))} \\
&\qquad+ 
\begin{bmatrix}
\Phi^{\tau}(1)\\
\Phi^{\tau}(1)\\
\Phi^{\tau}(2)\\
\vdots\\
\Phi^{\tau}(c-1)
\end{bmatrix}
\cdot
\begin{bmatrix}
p(x,\curl*{c}^{h}_{x})\\
p(x,\curl*{c-1}^{h}_{x})\\
p(x,\curl*{c-2}^{h}_{x})\\
\vdots\\
p(x,\curl*{1}^{h}_{x})\\
\end{bmatrix}
-\begin{bmatrix}
\Phi^{\tau}(1)\\
\Phi^{\tau}(1)\\
\Phi^{\tau}(2)\\
\vdots\\
\Phi^{\tau}(c-1)
\end{bmatrix}
\cdot
\begin{bmatrix}
p_{\bracket*{c}}(x)\\
p_{\bracket*{c-1}}(x)\\
p_{\bracket*{c-2}}(x)\\
\vdots\\
p_{\bracket*{1}}(x)\\
\end{bmatrix}
\tag{$p_{[c]}(x)= \max_{y\in \sY}p(x, y)$, $\curl*{c}^{h}_x = \hh(x)$ and $\Phi^{\tau}(0)=0$}\\
& \geq \Phi^{\tau}(1)\, p(x, \hh(x))\mathds{1}_{h\not \in \ov \sH_{\gamma}(x)} + \Phi^{\tau}(1) \paren*{\max_{y\in \sY}p(x, y) - p(x, \hh(x))} \tag{ rearrangement inequality for $\Phi^{\tau}(1)\leq \Phi^{\tau}(1) \leq \Phi^{\tau}(2)\leq \cdots\leq \Phi^{\tau}(c-1)$ and $p_{\bracket*{c}}(x)\geq \cdots \geq  p_{\bracket*{1}}(x)$}\\
& = \Phi^{\tau}(1)\paren*{\max_{y\in \sY}p(x, y) - p(x, \hh(x))\mathds{1}_{h\in \ov \sH_{\gamma}(x)}}
\end{align*}
for any $h\in\sH$. Since $\sH$ is symmetric and $\ov \sH_{\gamma}(x)\neq\emptyset$, we have 
\begin{align*}
\Delta\sC_{\ell_{\gamma},\sH}(h, x)
& = \sC_{\ell_{\gamma}}(h, x)-\sC^*_{\ell_{\gamma}}(\sH,x)\\
& = \sum_{y\in \sY} p(x,y) \sup_{x':\norm*{x-x'}\leq \gamma}\mathds{1}_{\rho_h(x', y) \leq 0}-\inf_{h\in \sH}\sum_{y\in \sY} p(x,y) \sup_{x':\norm*{x-x'}\leq \gamma}\mathds{1}_{\rho_h(x', y) \leq 0}\\
& = \paren*{1-p(x,\hh(x))}\mathds{1}_{h\in \ov \sH_{\gamma}(x)}+\mathds{1}_{h\not \in \ov \sH_{\gamma}(x)} - \inf_{h\in \sH}\bracket*{\paren*{1-p(x,\hh(x))}\mathds{1}_{h\in \ov \sH_{\gamma}(x)}+ \mathds{1}_{h\not \in \ov \sH_{\gamma}(x)}}\\
& = \paren*{1-p(x,\hh(x))}\mathds{1}_{h\in \ov \sH_{\gamma}(x)} +\mathds{1}_{h\not \in \ov \sH_{\gamma}(x)} - \paren*{1-\max_{y\in \sY} p(x,y)} \tag{$\sH$ is symmetric and $\ov \sH_{\gamma}(x)\neq\emptyset$}\\
& = \max_{y\in \sY}p(x, y) - p(x, \hh(x))\mathds{1}_{h\in \ov \sH_{\gamma}(x)}.
\end{align*}
Therefore, by the definition, we obtain
\begin{align*}
\sR_{\ell_{\gamma}}(h)- \sR^*_{\ell_{\gamma}}(\sH)+\sM_{\ell_{\gamma}}(\sH)
& =
\mathbb{E}_{X}\bracket*{\Delta\sC_{\ell_{\gamma}}(h, x)}\\
& = \mathbb{E}_{X}\bracket*{\max_{y\in \sY}p(x, y) - p(x, \hh(x))\mathds{1}_{h\in \ov \sH_{\gamma}(x)}}\\
&\leq
\Phi^{\tau}(1)\,\mathbb{E}_{X}\bracket*{\Delta\sC_{\wt \ell^{\mathrm{comp}}_{\tau,\rho},\sH}(h, x)}\\
& = \Phi^{\tau}(1) \paren*{\sR_{\wt \ell^{\mathrm{comp}}_{\tau,\rho}}(h)-\sR^*_{\wt \ell^{\mathrm{comp}}_{\tau,\rho}}(\sH) + \sM_{\wt \ell^{\mathrm{comp}}_{\tau,\rho}}(\sH)},
\end{align*}
which implies that
\begin{align*}
\sR_{\ell_{\gamma}}(h)- \sR^*_{\ell_{\gamma}}(\sH) \leq \Phi^{\tau}(1)\paren*{\sR_{\wt \ell^{\mathrm{comp}}_{\tau,\rho}}(h)-\sR^*_{\wt \ell^{\mathrm{comp}}_{\tau,\rho}}(\sH) + \sM_{\wt \ell^{\mathrm{comp}}_{\tau,\rho}}(\sH)} - \sM_{\ell_{\gamma}}(\sH).
\end{align*}
\end{proof}


\section{Learning bounds (proof of Theorem~\ref{th:genbound})}
\label{app:genbound}

\GenBound*
\begin{proof}
  By the standard Rademacher complexity bounds \citep{MohriRostamizadehTalwalkar2018}, the following holds
  with probability at least $1 - \delta$ for all $h \in \sH$:
\[
\abs*{\sR_{\ell_{\tau}^{\rm{comp}}}(h) - \h \sR_{\ell_{\tau}^{\rm{comp}}, S}(h)}
\leq 2 \Rad_m^\tau(\sH) +
B_\tau \sqrt{\tfrac{\log (2/\delta)}{2m}}.
\]
Fix $\e > 0$. By the definition of the infimum, there exists $h^* \in
\sH$ such that $\sR_{\ell_{\tau}^{\rm{comp}}}(h^*) \leq
\sR_{\ell_{\tau}^{\rm{comp}}}^*(\sH) + \e$. By definition of
$\h h_S$, we have
\begin{align*}
  & \sR_{\ell_{\tau}^{\rm{comp}}}(\h h_S) - \sR_{\ell_{\tau}^{\rm{comp}}}^*(\sH)\\
  & = \sR_{\ell_{\tau}^{\rm{comp}}}(\h h_S) - \h \sR_{\ell_{\tau}^{\rm{comp}}, S}(\h h_S) + \h \sR_{\ell_{\tau}^{\rm{comp}}, S}(\h h_S) - \sR_{\ell_{\tau}^{\rm{comp}}}^*(\sH)\\
  & \leq \sR_{\ell_{\tau}^{\rm{comp}}}(\h h_S) - \h \sR_{\ell_{\tau}^{\rm{comp}}, S}(\h h_S) + \h \sR_{\ell_{\tau}^{\rm{comp}}, S}(h^*) - \sR_{\ell_{\tau}^{\rm{comp}}}^*(\sH)\\
  & \leq \sR_{\ell_{\tau}^{\rm{comp}}}(\h h_S) - \h \sR_{\ell_{\tau}^{\rm{comp}}, S}(\h h_S) + \h \sR_{\ell_{\tau}^{\rm{comp}}, S}(h^*) - \sR_{\ell_{\tau}^{\rm{comp}}}^*(h^*) + \e\\
  & \leq
  2 \bracket*{2 \Rad_m^\tau(\sH) +
B_\tau \sqrt{\tfrac{\log (2/\delta)}{2m}}} + \e.
\end{align*}
Since the inequality holds for all $\e > 0$, it implies:
\[
\sR_{\ell_{\tau}^{\rm{comp}}}(\h h_S) - \sR_{\ell_{\tau}^{\rm{comp}}}^*(\sH)
\leq 
4 \Rad_m^\tau(\sH) +
2 B_\tau \sqrt{\tfrac{\log (2/\delta)}{2m}}.
\]
Plugging in this inequality in the bound of
Theorem~\ref{Thm:bound_comp_sum} completes the proof.
\end{proof}
\restoreatoc

\chapter{Appendix to Chapter~\ref{ch5}}

\disableatoc
\section{Minimizability gap}
\label{app:minimizability-gap}

This is a brief discussion of minimizability gaps and their properties.
By definition, for any loss function $\ell$, the minimizability gap is
defined by
\[
\sM_\ell(\sH)
= \inf_{h \in \sH} \curl*{\E_{(x, y) \sim \sD}[\ell(h, x, y)] }
- \E_x \bracket*{\inf_{h \in \sH} \E_{y | x} \bracket*{\ell(h, x, y)} }
= \sR^*_\ell(\sH) - \E_x \bracket*{\sC^*_\ell(\sH, x)}.
\]

\subsection{Zero minimizability}

\begin{lemma}
\label{lemma:zero-minimizability}
  Let $\ell$ be a surrogate loss such that for $(x, y) \in \sX \times
  \sY$ and any measurable function $h \in \sH_{\rm{all}}$, the loss
  $\ell(h, x, y)$ only depends on $h(x)$ and $y$ (thus we can write
  $\ell(h, x, y) = \ov\ell(h(x), y)$ for some function $\ov\ell$). Then, the
  minimizabiliy gap vanishes: $\sM_\ell(\sH_{\rm{all}}) = 0$.
\end{lemma}

\begin{proof}
  Fix $\e > 0$. Then, by definition of the infimum, for any $x \in \sX$,
  there exists $h_x \in \sH_{\rm{all}}$ such that
  \[
\E_{y | x}\bracket*{\ell(h_x, x, y)} \leq \inf_{h \in \sH_{\rm{all}}} \E_{y | x}\bracket*{\ell(h, x, y)} + \e.
\]
Now, define the function $h$ by $h(x) = h_x(x)$, for all $x \in \sX$.
$h$ can be shown to be measurable, for example, when $\sX$ admits
a countable dense subset.
Then,
\begin{align*}
  \E_{(x, y) \sim \sD}[\ell(h, x, y)]
  =  \E_{(x, y) \sim \sD}[\ov\ell(h(x), y)]
  & =  \E_{(x, y) \sim \sD}[\ov\ell(h_x(x), y)]\\
  & = \E_{(x, y) \sim \sD}[\ell(h_x, x, y)]\\
  & \leq \E_x \bracket*{\inf_{h \in \sH_{\rm{all}}} \E_{y | x} \bracket*{\ell(h, x, y)} + \e}\\
  & = \E_x \bracket*{\sC_\ell^*(\sH_{\rm{all}}, x)} + \e.
\end{align*}
Thus, we have
\[
\inf_{h \in \sH_{\rm{all}}} \E_{(x, y) \sim \sD}[\ell(h, x, y)]
\leq \E_x \bracket*{\sC_\ell^*(\sH_{\rm{all}}, x)} + \e.
\]
Since the inequality holds for any $\e > 0$, we have
$\sR^*_\ell(\sH_{\rm{all}}) = \inf_{h \in \sH_{\rm{all}}} \E_{(x, y)
  \sim \sD}[\ell(h, x, y)] \leq \E_x
\bracket*{\sC_\ell^*(\sH_{\rm{all}}, x)}$. This implies equality since
the inequality $\sR^*_\ell(\sH) \geq \E_x \bracket*{\sC_\ell^*(\sH, x)}$
    holds for any $\sH$.  
\end{proof}

\subsection{Relationship with approximation error}

Let $\sA_{\ell}$ denote the approximation error of a loss function
$\ell$ and a hypothesis set $\sH$: $\sA_{\ell}(\sH) =
\sR^*_{\ell}(\sH)-\sR^*_{\ell}(\sH_{\rm{all}})$. We will denote by
$I_{\ell}(\sH)$ the difference of pointwise infima \[I_{\ell}(\sH) =
\mathbb{E}_{x}\left[\sC_{\ell}^*(\sH,x)-\sC_{\ell}^*(\sH_{\rm{all}},x)\right],\]
which is non-negative.  The minimizability gap can be decomposed as
follows in terms of the approximation error and the difference of
pointwise infima:
\begin{align*}
  \sM_{\ell}(\sH)
  & =\sR^*_{\ell}(\sH) - \mathbb{E}_{x}\left[\sC_{\ell}^*(\sH,x)\right]\\
  & = \sR^*_{\ell}(\sH) - \sR^*_{\ell}(\sH_{\rm{all}}) + \sR^*_{\ell}(\sH_{\rm{all}}) - \mathbb{E}_{x}\left[\sC_{\ell}^*(\sH,x)\right]\\
  & = \sA_{\ell}(\sH) + \sR^*_{\ell}(\sH_{\rm{all}}) - \mathbb{E}_{x}\left[\sC_{\ell}^*(\sH,x)\right]\\
  & = \sA_{\ell}(\sH) + \mathbb{E}_{x}\left[\sC_{\ell}^*(\sH_{\rm{all}},x)-\sC_{\ell}^*(\sH,x)\right] \tag{By Lemma~\ref{lemma:zero-minimizability}}\\
  & = \sA_{\ell}(\sH) - I_{\ell}(\sH).
\end{align*}
The decomposition immediately implies the
following result.

\begin{lemma}
 Let $\ell$ be a surrogate loss such that for $(x, y) \in \sX \times
 \sY$ and any measurable function $h \in \sH_{\rm{all}}$, the loss
 $\ell(h, x, y)$ only depends on $h(x)$ and $y$ (thus we can write
 $\ell(h, x, y) = \ov\ell(h(x), y)$ for some function $\ov\ell$).
 Then, for any loss function $\ell$ and hypothesis set $\sH$, we have:
 $\sM_{\ell}(\sH) \leq \sA_{\ell}(\sH)$.
\end{lemma}

By Lemma~\ref{lemma:explicit_assumption_01-chcb}, when $\ell$ is the
zero-one loss, $I_{\ell}(\sH)=0$ when the hypothesis set generates
labels that cover all possible outcomes for each input. However, for a
surrogate loss function, $I_{\ell}(\sH)$ is non-negative, and is
generally non-zero.

Take the example of binary classification and denote the conditional
distribution as $\eta(x)=D(Y=1 | X=x)$.  Let $\sH$ be a family of
functions $h$ with $|h(x)| \leq \Lambda$ for all $x \in \sX$ and such
that all values in $[-\Lambda, +\Lambda]$ can be reached.  Consider
for example the exponential-based margin loss: $\ell(h, x, y) =
e^{-yh(x)}$.  Then, $\sC_{\ell}(h,x) = \eta(x) e^{-h(x)} + (1 -
\eta(x)) e^{h(x)}$. Upon observing this, it becomes apparent that the
infimum over all measurable functions can be expressed in the
following way, for all $x$:
\begin{equation*}
\sC_{\ell}^*(\sH_{\rm{all}},x) =
2\sqrt{\eta(x)(1-\eta(x))},
\end{equation*}
while the infimum over $\sH$, $\sC_{\ell}^*(\sH,x)$, depends on
$\Lambda$ and can be expressed as
  \begin{equation*}
   \sC_{\ell}^*(\sH,x)=\begin{cases}
   \max\curl*{\eta(x),1 - \eta(x)}
e^{-\Lambda} + \min\curl*{\eta(x),1 - \eta(x)} e^{\Lambda} & \Lambda<\frac{1}{2} \abs*{\log \frac{\eta(x)}{1
    -\eta(x)}}\\
    2\sqrt{\eta(x)(1-\eta(x))} & \text{otherwise}.
   \end{cases} 
  \end{equation*}
Thus, in the deterministic scenario, 
\begin{equation*}
I_{\ell}(\sH) = \mathbb{E}_{x}\bracket*{\sC_{\ell}^*(\sH,x)-\sC_{\ell}^*(\sH_{\rm{all}},x)}= e^{-\Lambda}.
\end{equation*}

\subsection{Significance of \texorpdfstring{$\sH$}{H}-consistency bounds}

As shown in the previous section, for target loss $\ell_{0-1}$, the
minimizability gap coincides with the approximation error
$\sM_{\ell_{0-1}}(\sH)=\sA_{\ell_{0-1}}(\sH)$ when the hypothesis set
generates labels that cover all possible outcomes for each
input. However, for a surrogate loss $\ell$, the minimizability gap is
generally strictly less than the approximation error
$\sM_{\ell}(\sH)<\sA_{\ell}(\sH)$.  Thus, an $\sH$-consistency bound,
expressed as follows for some increasing function $\Gamma$:
  \begin{equation*}
   \sR_{\ell_{0-1}}(h)-\sR^*_{\ell_{0-1}}(\sH)+\sM_{\ell_{0-1}}(\sH)\leq \Gamma\paren*{\sR_{\ell}(h)-\sR^*_{\ell}(\sH)+\sM_{\ell}(\sH)}. 
  \end{equation*}
is more favorable than an excess error bound expressed in terms of
approximation errors:
\begin{equation*}
\sR_{\ell_{0-1}}(h) - \sR^*_{\ell_{0-1}}(\sH) +
\sA_{\ell_{0-1}}(\sH) \leq \Gamma\paren*{\sR_{\ell}(h) -
  \sR^*_{\ell}(\sH) + \sA_{\ell}(\sH)}.
\end{equation*}
Here, $\Gamma$ is typically
linear or the square-root function modulo constants. When $\sH =
\sH_{\rm{all}}$, the family of all measurable functions, by Lemma~\ref{lemma:zero-minimizability}, the
$\sH$-consistency bound coincides with the excess error bound and
implies Bayes-consistency by taking the limit. It is therefore
a stronger guarantee than an excess error bound
and Bayes-consistency.

\section{Proofs for comp-sum losses}
\label{app:comp-sum}

Let $y_{\max}=\argmax_{y\in \sY} \sfp(y \!\mid\! x)$ and $\hh(x) = \argmax_{y\in
  \sY}h(x, y)$, where we choose the label with the highest index under
the natural ordering of labels as the tie-breaking strategy.

\subsection{Proof of \texorpdfstring{$\sH$}{H}-consistency bounds
  with \texorpdfstring{$\sT^{\rm{comp}}$}{T} (Theorem~\ref{Thm:bound_comp})}
\label{app:bound_comp}

\BoundComp*
\begin{proof}
For the comp-sum loss $\ell^{\mathrm{comp}}$, the conditional $\ell^{\mathrm{comp}}$-risk can be expressed as follows:
\begin{align*}
\sC_{\ell^{\mathrm{comp}}}(h,x)
& = \sum_{y\in \sY} \sfp(y \!\mid\! x) \ell^{\mathrm{comp}}(h, x, y)\\
&= \sum_{y\in \sY} \sfp(y \!\mid\! x) \Phi\paren*{\frac{e^{h(x,y)}}{\sum_{y'\in \sY}e^{h(x,y')}}}\\
&= \sum_{y\in \sY} \sfp(y \!\mid\! x) \Phi\paren*{S_h(x,y)}\\
& = \sfp(y_{\max} \!\mid\! x) \Phi\paren*{S_h(x,y_{\max})}+\sfp(\hh(x) \!\mid\! x) \Phi\paren*{S_h(x,\hh(x))}\\
& \qquad + \sum_{y\notin \curl*{y_{\max},\hh(x)}}\sfp(y \!\mid\! x) \Phi\paren*{S_h(x,y)}.
\end{align*}
where we let $S_h(x,y)=\frac{e^{h(x,y)}}{\sum_{y'\in \sY}e^{h(x,y')}}$
for any $y\in \sY$ with the constraint that $\sum_{y\in
  \sY}S_h(x,y)=1$.  For any $h \in \sH$ such that $\hh(x) \neq
y_{\max}$ and $x\in \sX$, we can always find a family of hypotheses
$\curl*{h_{\mu}}\subset \sH$ such that
$S_{h,\mu}(x,\cdot)=\frac{e^{h_{\mu}(x,\cdot)}}{\sum_{y'\in
    \sY}e^{h_{\mu}(x,y')}}$ take the following values:
\begin{align*}
S_{h,\mu}(x,y) = 
\begin{cases}
  S_h(x, y) & \text{if $y \not \in \curl*{y_{\max}, \hh(x)}$}\\
 S_h(x, y_{\max}) + \mu & \text{if $y = \hh(x)$}\\
  S_h(x,\hh(x)) -\mu & \text{if $y = y_{\max}$}.
\end{cases} 
\end{align*}
Note that $S_{h,\mu}$ satisfies the constraint:
\begin{align*}
 \sum_{y\in \sY}S_{h,\mu}(x, y)=\sum_{y\in \sY}
 S_h(x, y)=1.
\end{align*}
Let $p_1=\sfp(y_{\max} \!\mid\! x)$, $p_2=\sfp(\hh(x) \!\mid\! x)$, $\tau_1=S_h(x,\hh(x))$ and
$\tau_2=S_h(x,y_{\max})$ to simplify the notation.  Then, by the
definition of $S_{h,\mu}$, we have for any $h\in \sH$ and $x\in \sX$,
\begin{align*}
&\sC_{\ell^{\mathrm{comp}}}(h,x) - \inf_{\mu \in [-\tau_2,\tau_1]}\sC_{\ell^{\mathrm{comp}}}(h_{\mu},x)\\
& = \sup_{\mu\in [-\tau_2,\tau_1]} \bigg\{p_1\bracket*{\Phi\paren*{\tau_2}-\Phi\paren*{\tau_1-\mu}}+  p_2\bracket*{ \Phi\paren*{\tau_1}-\Phi\paren*{\tau_2+\mu}} \bigg\}\\
& = \sup_{\mu\in [-\tau_2,\tau_1]} \bigg\{\frac{\psum+p_1-p_2}{2}\bracket*{\Phi\paren*{\tau_2}-\Phi\paren*{\tau_1-\mu}}+  \frac{\psum-p_1+p_2}{2}\bracket*{ \Phi\paren*{\tau_1}-\Phi\paren*{\tau_2+\mu}} \bigg\}\tag{$\psum=p_1+p_2\in \bracket*{\frac{1}{n-1}\vee p_1-p_2,1}$}\\
& \leq \inf_{\psum\in \bracket*{\frac{1}{n-1}\vee p_1-p_2,1}}\inf\limits_{\substack{ \tau_1\geq \max(\tau_2,1/n)\\ \tau_1+\tau_2\leq 1,\tau_2\geq 0}}\sup_{\mu\in [-\tau_2,\tau_1]} \bigg\{\frac{\psum+p_1-p_2}{2}\bracket*{\Phi\paren*{\tau_2}-\Phi\paren*{\tau_1-\mu}} \\
& \qquad +  \frac{\psum-p_1+p_2}{2}\bracket*{ \Phi\paren*{\tau_1}-\Phi\paren*{\tau_2+\mu}} \bigg\}\tag{$\tau_1\geq \max(\tau_2,1/n), \tau_1+\tau_2\leq 1,\tau_2\geq 0$}\\
& = \sT^{\rm{comp}}\paren*{p_1-p_2}\\
& = \sT^{\rm{comp}}\paren*{\Delta\sC_{\ell_{0-1},\sH}(h,x)} \tag{by Lemma~\ref{lemma:explicit_assumption_01-chcb}},
\end{align*}
where for $n=2$, an additional constraint $\tau_1+\tau_2=1$ is imposed and the expression of $\sT^{\rm{comp}}$ is simplified.
Since $\sT^{\rm{comp}}$ is convex, by Jensen's inequality, we obtain for any hypothesis $h \in \sH$ and any distribution,
\begin{align*}
&\sT^{\rm{comp}}\paren*{\sR_{\ell_{0-1}}(h)-\sR_{\ell_{0-1}}^*(\sH)+\sM_{\ell_{0-1}}(\sH)}\\
&=\sT^{\rm{comp}}\paren*{\E_{X}\bracket*{\Delta\sC_{\ell_{0-1},\sH}(h,x)}}\\
&\leq \E_{X}\bracket*{\sT^{\rm{comp}}\paren*{\Delta\sC_{\ell_{0-1},\sH}(h,x)}}\\
& \leq \E_{X}\bracket*{\Delta\sC_{\ell^{\mathrm{comp}},\sH}(h,x)}\\
& =\sR_{\ell^{\mathrm{comp}}}(h)-\sR_{\ell^{\mathrm{comp}}}^*(\sH)+\sM_{\ell^{\mathrm{comp}}}(\sH).
\end{align*}
For the second part, we first consider $n=2$. For any $t\in [0,1]$, we
consider the distribution that concentrates on a singleton $\curl*{x}$
and satisfies $\sfp(1 \!\mid\! x)=\frac{1+t}{2}$, $\sfp(2 \!\mid\! x)=\frac{1-t}{2}$. For any
$\e>0$, by the definition of infimum, we can take $h\in \sH$ such that
$S_h(x,1)=\tau_{\e}\in \bracket*{0,\frac12}$ and satisfies
\begin{align*}
\sup\limits_{\mu\in [-\tau_{\e},1-\tau_{\e}]} \curl*{\frac{1 +
    t}{2}\bracket*{\Phi\paren*{\tau_{\e}}-\Phi\paren*{1-\tau_{\e}-\mu}}
  + \frac{1 - t}{2}\bracket*{
    \Phi\paren*{1-\tau_{\e}}-\Phi\paren*{\tau_{\e}+\mu}}}<\sT^{\rm{comp}}(t)+\e.
\end{align*}
Then,
\begin{align*}
\sR_{\ell_{0-1}}(h)-
\sR^*_{\ell_{0-1}}(\sH)+\sM_{\ell_{0-1}}(\sH)
&=  \sR_{\ell_{0-1}}(h) - \mathbb{E}_{X} \bracket* {\sC^*_{\ell_{0-1}}(\sH,x)}\\
&=\sC_{\ell_{0-1}}(h,x) - \sC^*_{\ell_{0-1}}\paren*{\sH,x}\\
&=t
\end{align*}
and
\begin{align*}
\sT^{\rm{comp}}(t)
&\leq \sR_{\ell^{\mathrm{comp}}}(h) - \sR_{\ell^{\mathrm{comp}}}^*(\sH) +
\sM_{\ell^{\mathrm{comp}}}(\sH)\\
&=  \sR_{\ell^{\mathrm{comp}}}(h) - \mathbb{E}_{X} \bracket* {\sC^*_{\ell^{\mathrm{comp}}}(\sH,x)}\\
&=\sC_{\ell^{\mathrm{comp}}}(h,x)-\sC^*_{\ell^{\mathrm{comp}}}(\sH,x)\\
&=\sup\limits_{\mu\in [-\tau_{\e},1-\tau_{\e}]} \curl*{\frac{1 + t}{2}\bracket*{\Phi\paren*{\tau_{\e}}-\Phi\paren*{1-\tau_{\e}-\mu}} + \frac{1 - t}{2}\bracket*{ \Phi\paren*{1-\tau_{\e}}-\Phi\paren*{\tau_{\e}+\mu}}}\\
&<\sT^{\rm{comp}}(t)+\e.
\end{align*}
By letting $\e\to 0$, we prove the tightness for $n = 2$. The proof
for $n > 2$ directly extends from the case when $n = 2$. Indeed, for
any $t\in [0,1]$, we consider the distribution that concentrates on a
singleton $\curl*{x}$ and satisfies $\sfp(1 \!\mid\! x)=\frac{1+t}{2}$,
$\sfp(2 \!\mid\! x)=\frac{1-t}{2}$, $\sfp(y \!\mid\! x)=0,\,3\leq y\leq n$. For any $\e>0$, by
the definition of infimum, we can take $h\in \sH$ such that $S_h(x,1)
= \tau_{1, \e}$, $S_h(x,2) = \tau_{2, \e}$, $S_h(x,y) = 0, \,3\leq
y\leq n$ and satisfies $ \tau_{1, \e} + \tau_{2, \e} = 1$, and
\begin{align*}
& \inf\limits_{\psum\in \bracket*{\frac{1}{n-1}\vee t,1}}\sup\limits_{\mu\in [-\tau_{2, \e}, \tau_{1, \e}]} \curl*{\frac{\psum + t}{2}\bracket*{\Phi\paren*{\tau_{2, \e}}-\Phi\paren*{\tau_{1, \e}-\mu}} + \frac{\psum-t}{2}\bracket*{ \Phi\paren*{\tau_{1, \e}}-\Phi\paren*{\tau_{2, \e}+\mu}}}\\
& = \sup\limits_{\mu\in [-\tau_{2, \e}, \tau_{1, \e}]} \curl*{\frac{1 + t}{2}\bracket*{\Phi\paren*{\tau_{2, \e}}-\Phi\paren*{\tau_{1, \e}-\mu}} + \frac{1-t}{2}\bracket*{ \Phi\paren*{\tau_{1, \e}}-\Phi\paren*{\tau_{2, \e}+\mu}}}\\
& < \sT^{\rm{comp}}(t)+\e.
\end{align*}
Then,
\begin{align*}
\sR_{\ell_{0-1}}(h)-
\sR^*_{\ell_{0-1}}(\sH)+\sM_{\ell_{0-1}}(\sH)=t
\end{align*}
and
\begin{align*}
\sT^{\rm{comp}}(t) <\sT^{\rm{comp}}(t)+\e.
\end{align*}
By letting $\e\to 0$, we prove the tightness for $ n > 2$.
\end{proof}

\subsection{Characterization of \texorpdfstring{$\sT^{\rm{comp}}$}{T} (Theorem~\ref{Thm:char_comp})}
\label{app:char_comp}

\CharComp*
\begin{proof}
For $n=2$, we have
\begin{align*}
\sT^{\rm{comp}}(t)
&=\inf\limits_{\tau\in \bracket*{0,\frac12}}\sup\limits_{\mu\in [-\tau,1-\tau]} \curl*{\frac{1 + t}{2}\bracket*{\Phi\paren*{\tau}-\Phi\paren*{1-\tau-\mu}} + \frac{1 - t}{2}\bracket*{ \Phi\paren*{1-\tau}-\Phi\paren*{\tau+\mu}}}\\
&= \inf_{\tau\in \bracket*{0,\frac12}}\paren*{\frac{1 + t}{2} \Phi\paren*{\tau}+ \frac{1 - t}{2}\bracket*{\Phi\paren*{1-\tau}}-\inf_{\mu\in [-\tau,1-\tau]}\curl*{\frac{1 + t}{2} \Phi\paren*{1-\tau-\mu} + \frac{1-t}{2}\Phi\paren*{\tau+\mu}}}\\
&= \inf_{\tau\in \bracket*{0,\frac12}}\paren*{\frac{1 + t}{2} \Phi\paren*{\tau}+ \frac{1 - t}{2}\bracket*{\Phi\paren*{1-\tau}}}-\inf_{\mu\in\bracket*{-\frac12,\frac12}}\curl*{\frac{1 - t}{2} \Phi\paren*{\frac12+\mu} + \frac{1 + t}{2}\Phi\paren*{\frac12-\mu}}\\
&\geq \inf_{\tau\in \bracket*{0,\frac12}}\paren*{\Phi\paren*{\frac12}+\Phi'\paren*{\frac12}t\paren*{\tau-\frac12}}-\inf_{\mu\in\bracket*{-\frac12,\frac12}}\curl*{\frac{1 - t}{2} \Phi\paren*{\frac12+\mu} + \frac{1 + t}{2}\Phi\paren*{\frac12-\mu}}
\tag{$\Phi$ is convex}\\
&= \Phi\paren*{\frac12}-\inf_{\mu\in\bracket*{-\frac12,\frac12}}\curl*{\frac{1 - t}{2} \Phi\paren*{\frac12+\mu} + \frac{1 + t}{2}\Phi\paren*{\frac12-\mu}}
\tag{$\Phi'\paren*{\frac12}<0$, $t\paren*{\tau-\frac12}\leq 0$}
\end{align*}
where the equality can be achieved by $\tau=\frac12$.

For $n>2$, we have
\begin{align*}
&\sT^{\rm{comp}}(t)=
\inf\limits_{\psum\in \bracket*{\frac{1}{n-1},1}}\inf\limits_{\substack{ \tau_1\geq \max(\tau_2,1/n)\\ \tau_1+\tau_2\leq 1,\tau_2\geq 0}}\sup\limits_{\mu\in [-\tau_2,\tau_1]} F\paren*{\psum,\tau_1,\tau_2,\mu}
\end{align*}
where we let $F\paren*{\psum,\tau_1,\tau_2,\mu}=\frac{\psum +
  t}{2}\bracket*{\Phi\paren*{\tau_2}-\Phi\paren*{\tau_1-\mu}} +
\frac{\psum-t}{2}\bracket*{
  \Phi\paren*{\tau_1}-\Phi\paren*{\tau_2+\mu}}$. For simplicity, we
assume that $\Phi$ is differentiable. For general convex $\Phi$, we
can proceed by using left and right derivatives, which are
non-decreasing. Differentiate $F$ with respect to $\mu$, we have
\begin{equation*}
\frac{\partial F}{\partial \mu}=\frac{\psum + t}{2}\Phi'(\tau_1-\mu)+\frac{t-\psum}{2}\Phi'(\tau_2+\mu).
\end{equation*}
Using the fact that $\psum\in \bracket*{\frac{1}{n-1}\vee
  t,1},t\in[0,1]$ and $\Phi'$ is non-decreasing, we obtain that
$\frac{\partial F}{\partial \mu}$ is non-increasing. Furthermore,
$\Phi'$ is non-decreasing and non-positive, $\Phi$ is non-negative, we
obtain that $\Phi'(+ \infty)=0$. This implies that $\frac{\partial
  F}{\partial \mu}(+ \infty)\leq 0$ and $\frac{\partial
  F}{\partial \mu}(- \infty)\geq 0$. Therefore, there exists
$\mu_0\in \Rset$ such that
\begin{equation*}
\frac{\partial F}{\partial \mu}(\mu_0)=\frac{\psum + t}{2}\Phi'(\tau_1-\mu_0)+\frac{t-\psum}{2}\Phi'(\tau_2+\mu_0)=0  
\end{equation*}
By taking $\mu=\tau_1-\tau_2$ and using the fact that $\tau_2\leq \frac12$, $\Phi'\paren*{\frac12}<0$, we have
\begin{equation*}
\frac{\partial F}{\partial \mu}(\tau_1-\tau_2)=\frac{\psum + t}{2}\Phi'(\tau_2)+\frac{t-\psum}{2}\Phi'(\tau_1)< 0.
\end{equation*}
Thus, since $\frac{\partial F}{\partial \mu}$ is non-increasing, we obtain $\mu_0< \tau_1-\tau_2$.
Differentiate $F$ with respect to $\tau_2$ at $\mu_0$, we have
\begin{equation*}
\frac{\partial F}{\partial \tau_2}=\frac{\psum + t }{2}\Phi'(\tau_2)+\frac{t-\psum}{2}\Phi'(\tau_2+\mu_0).
\end{equation*}
Since $\Phi'$ is non-decreasing, we obtain
\begin{equation*}
\frac{\partial F}{\partial \tau_2}\leq \frac{\psum + t }{2}\Phi'(\tau_2)+\frac{t-\psum}{2}\Phi'(\tau_2+\tau_1-\tau_2)=\frac{\partial F}{\partial \mu}(\tau_1-\tau_2)<0,
\end{equation*}
which implies that the infimum $\inf_{\tau_1\geq \max\curl*{\tau_2,\frac1n}}$ is achieved when $\tau_2=\tau_1$.
Differentiate $F$ with respect to $\psum$ at $\mu_0$ and $\tau_1=\tau_2$, by the convexity of $\Phi$, we obtain
\begin{equation*}
\frac{\partial F}{\partial \psum}=\Phi(\tau_1)-\Phi(\tau_1-\mu_0)+\Phi(\tau_1)-\Phi(\tau_1+\mu_0)\leq 0,
\end{equation*}
which implies that the infimum $\inf_{\psum\in \bracket*{\frac{1}{n-1},1}}$ is achieved when $\psum=1$. Above all, we obtain
\begin{align*}
\sT^{\rm{comp}}(t)
&=\inf_{\tau\in\bracket*{\frac1n,\frac12}}\sup_{\mu\in [-\tau,\tau]}F\paren*{1,\tau,\tau,\mu}\\
&=\inf_{\tau\in\bracket*{\frac1n,\frac12}}\curl*{\Phi(\tau) -\inf_{\mu\in [-\tau,\tau]}\curl*{\frac{1 + t}{2}\Phi(\tau-\mu)+\frac{1 - t}{2}\Phi\paren*{\tau+\mu}}}.
\end{align*}
\end{proof}

\subsection{Computation of examples}
\label{app:examples-comp}

\paragraph{Example: $\Phi(t)=-\log(t)$.}
For $n = 2$, plugging in $\Phi(t)=-\log(t)$ in
Theorem~\ref{Thm:char_comp}, gives
\begin{align*}
\sT^{\rm{comp}}
& =\log 2-\inf_{\mu\in\bracket*{-\frac12,\frac12}}\curl*{-\frac{1 - t}{2}\log\paren*{\frac12+\mu}-\frac{1 + t}{2}\log\paren*{\frac12-\mu}}\\
& =\frac{1 + t}{2}\log(1 + t)+\frac{1 - t}{2}\log(1 - t) \tag{minimum achieved at $\mu = -\frac{t}{2}$}.
\end{align*}
Similarly, for $n > 2$, plugging in $\Phi(t)=-\log(t)$ in Theorem~\ref{Thm:char_comp} yields
\begin{align*}
\sT^{\rm{comp}}
&=\inf_{\tau\in\bracket*{\frac1n,\frac12}}\curl*{-\log \tau -\inf_{\mu\in [-\tau,\tau]}\curl*{-\frac{1 - t}{2}\log\paren*{\tau+\mu}-\frac{1 + t}{2}\log\paren*{\tau-\mu}}}\\
&=\frac{1 + t}{2}\log(1 + t)+\frac{1 - t}{2}\log(1 - t)\tag{minimum achieved at $\mu = -\tau t$}.
\end{align*}

\paragraph{Example: $\Phi(t)=\frac{1}{t}-1$.}
For $n = 2$, plugging in $\Phi(t)=\frac{1}{t}-1$ in Theorem~\ref{Thm:char_comp}, gives
\begin{align*}
\sT^{\rm{comp}}
& = 2-\inf_{\mu\in\bracket*{-\frac12,\frac12}}\curl*{\frac{1 - t}{2}\frac{1}{\frac12+\mu}+\frac{1 + t}{2}\frac{1}{\frac12-\mu}}\\
& = 1-\sqrt{1 - t^2} \tag{minimum achieved at $\mu = \frac{(1-t)^{\frac12}-(1+t)^{\frac12}}{2\paren*{(1+t)^{\frac12}+(1-t)^{\frac12}}}$}.
\end{align*}
Similarly, for $n > 2$, plugging in $\Phi(t)=\frac{1}{t}-1$ in Theorem~\ref{Thm:char_comp} yields
\begin{align*}
\sT^{\rm{comp}}
&=\inf_{\tau\in\bracket*{\frac1n,\frac12}}\curl*{\frac{1}{\tau} -\inf_{\mu\in [-\tau,\tau]}\curl*{\frac{1 + t}{2}\frac{1}{\tau-\mu}+\frac{1 + t}{2}\frac{1}{\tau+\mu}}}\\
&=\inf_{\tau\in\bracket*{\frac1n,\frac12}} \frac{1}{2\tau}\paren*{1-\sqrt{1 - t^2}}\tag{minimum achieved at $\mu = \frac{(1-t)^{\frac12}-(1+t)^{\frac12}}{(1+t)^{\frac12}+(1-t)^{\frac12}}\tau$}\\
&=1-\sqrt{1 - t^2}.\tag{minimum achieved at $\tau = \frac12$}
\end{align*}

\paragraph{Example: $\Phi(t)=\frac{1}{q}\paren*{1 - t^{q}},q\in (0,1)$.}

For $n = 2$, plugging in $\Phi(t)=\frac{1}{q}\paren*{1 - t^{q}}$ in Theorem~\ref{Thm:char_comp}, gives
\begin{align*}
\sT^{\rm{comp}}
& = - \frac{1}{q2^{q}}-\inf_{\mu\in\bracket*{-\frac12,\frac12}}\curl*{-\frac{1 - t}{2q}\paren*{\frac12+\mu}^{q}-\frac{1 + t}{2q}\paren*{\frac12-\mu}^{q}}\\
&=\frac{1}{q 2^{q}}\paren*{\frac{\paren*{1 + t}^{\frac1{1-q }} +  \paren*{1 - t}^{\frac1{1-q }}}{2}}^{1-q }-\frac{1}{q 2^{q}} \tag{minimum achieved at $\mu = \frac{(1-t)^{\frac{1}{1-q}}-(1+t)^{\frac{1}{1-q}}}{2\paren*{(1+t)^{\frac{1}{1-q}}+(1-t)^{\frac{1}{1-q}}}}$}.
\end{align*}
Similarly, for $n > 2$, plugging in $\Phi(t)=\frac{1}{q}\paren*{1 - t^{q}}$ in Theorem~\ref{Thm:char_comp} yields
\begin{align*}
\sT^{\rm{comp}}
&=\inf_{\tau\in\bracket*{\frac1n,\frac12}}\curl*{-\frac{\tau^q}{q} -\inf_{\mu\in [-\tau,\tau]}\curl*{-\frac{1 + t}{2q}\paren*{\tau-\mu}^{q}-\frac{1 - t}{2q}\paren*{\tau+\mu}^{q}}}\\
&=\inf_{\tau\in\bracket*{\frac1n,\frac12}} \frac{\tau^{q}}{q}\paren*{\frac{\paren*{1 + t}^{\frac1{1-q }} +  \paren*{1 - t}^{\frac1{1-q }}}{2}}^{1-q }-\frac{\tau^{q}}{q}\tag{minimum achieved at $\mu = \frac{(1-t)^{\frac{1}{1-q}}-(1+t)^{\frac{1}{1-q}}}{(1+t)^{\frac{1}{1-q}}+(1-t)^{\frac{1}{1-q}}}\tau$}\\
&=\frac{1}{q n^{q}}\paren*{\frac{\paren*{1 + t}^{\frac1{1-q }} +  \paren*{1 - t}^{\frac1{1-q }}}{2}}^{1-q }-\frac{1}{q n^{q}}.\tag{minimum achieved at $\tau = \frac1n$}
\end{align*}

\paragraph{Example: $\Phi(t)=1 - t$.} For $n = 2$, plugging in $\Phi(t)=1 - t$ in Theorem~\ref{Thm:char_comp}, gives
\begin{align*}
\sT^{\rm{comp}}
& = \frac{1}{2}-\inf_{\mu\in\bracket*{-\frac12,\frac12}}\curl*{\frac{1 - t}{2}\paren*{\frac12-\mu}+\frac{1 + t}{2}\paren*{\frac12+\mu}}=\frac{1}{2}-\frac{1 - t}{2}=\frac{t}{2}.
\end{align*}
Similarly, for $n > 2$, plugging in $\Phi(t)=1-t$ in Theorem~\ref{Thm:char_comp} yields
\begin{align*}
\sT^{\rm{comp}}
&=\inf_{\tau\in\bracket*{\frac1n,\frac12}}\curl*{(1-\tau) -\inf_{\mu\in [-\tau,\tau]}\curl*{\frac{1 + t}{2}(1-\tau+\mu)+\frac{1 - t}{2}\paren*{1-\tau-\mu}}}\\
&=\inf_{\tau\in\bracket*{\frac1n,\frac12}} \tau \,t\tag{minimum achieved at $\mu = -\tau$}\\
&=\frac{t}{n}.\tag{minimum achieved at $\tau = \frac1n$}
\end{align*}

\paragraph{Example: $\Phi(t)=(1-t)^2$.} For $n = 2$, plugging in $\Phi(t)=(1 - t)^2$ in Theorem~\ref{Thm:char_comp}, gives
\begin{align*}
\sT^{\rm{comp}}
& = \frac{1}{4}-\inf_{\mu\in\bracket*{-\frac12,\frac12}}\curl*{\frac{1 - t}{2}\paren*{\frac12-\mu}^2+\frac{1 + t}{2}\paren*{\frac12+\mu}^2}=\frac{1}{4}-\frac{1-t^2}{4}=\frac{t^2}{4}.
\end{align*}
Similarly, for $n > 2$, plugging in $\Phi(t)=(1-t)^2$ in Theorem~\ref{Thm:char_comp} yields
\begin{align*}
\sT^{\rm{comp}}
&=\inf_{\tau\in\bracket*{\frac1n,\frac12}}\curl*{(1-\tau)^2 -\inf_{\mu\in [-\tau,\tau]}\curl*{\frac{1 + t}{2}(1-\tau+\mu)^2+\frac{1 - t}{2}\paren*{1-\tau-\mu}^2}}\\
&=\inf_{\tau\in\bracket*{\frac1n,\frac12}}\curl*{(1-\tau)^2t^2}\tag{minimum achieved at $\mu = t(\tau - 1)$}\\
&=\frac{t^2}{4}.\tag{minimum achieved at $\tau = \frac12$}
\end{align*}

\section{Proofs for constrained losses}
\label{app:constrained_losses}

Let $y_{\max}=\argmax_{y\in \sY} \sfp(y \!\mid\! x)$ and $\hh(x) = \argmax_{y\in
  \sY}h(x, y)$, where we choose the label with the highest index under
the natural ordering of labels as the tie-breaking strategy.

\subsection{Proof of \texorpdfstring{$\sH$}{H}-consistency bounds
  with \texorpdfstring{$\sT^{\rm{cstnd}}$}{T} (Theorem~\ref{Thm:bound_cstnd})}
\label{app:bound_cstnd}

\BoundCstnd*
\begin{proof}
For the constrained loss $\ell^{\mathrm{cstnd}}$, the conditional $\ell^{\mathrm{cstnd}}$-risk can be expressed as follows:
\begin{align*}
\sC_{\ell^{\mathrm{cstnd}}}(h,x)
& = \sum_{y\in \sY} \sfp(y \!\mid\! x) \ell^{\mathrm{cstnd}}(h, x, y)\\
&= \sum_{y\in \sY} \sfp(y \!\mid\! x) \sum_{y'\neq y}\Phi\paren*{-h(x, y')}\\
&= \sum_{y\in \sY} \Phi\paren*{-h(x, y)} \sum_{y'\neq y}\sfp(y' \!\mid\! x)\\
&= \sum_{y\in \sY} \Phi\paren*{-h(x, y)} \paren*{1-\sfp(y \!\mid\! x)}\\
& = \Phi\paren*{-h(x, y_{\max})} \paren*{1-\sfp(y_{\max} \!\mid\! x)} + \Phi\paren*{-h(x, \hh(x))} \paren*{1-\sfp(\hh(x) \!\mid\! x)}\\
& \qquad + \sum_{y\notin \curl*{y_{\max},\hh(x)}}\Phi\paren*{-h(x, y)} \paren*{1-\sfp(y \!\mid\! x)}.
\end{align*}
For any $h \in \sH$ and $x\in \sX$, by the symmetry and completeness of $\sH$, we can always find a family of hypotheses $\curl*{h_{\mu}:\mu \in \mathbb{R}}\subset \sH$ such that $h_{\mu}(x,\cdot)$ take the following values: 
\begin{align*}
h_{\mu}(x,y) = 
\begin{cases}
  h(x, y) & \text{if $y \not \in \curl*{y_{\max}, \hh(x)}$}\\
 h(x, y_{\max}) + \mu & \text{if $y = \hh(x)$}\\
  h(x,\hh(x)) -\mu & \text{if $y = y_{\max}$}.
\end{cases} 
\end{align*}
Note that the hypotheses $h_{\mu}$ satisfies the constraint:
\begin{align*}
 \sum_{y\in \sY} h_{\mu}(x, y)=\sum_{y\in \sY}h(x, y)=0,\, \forall \mu \in \mathbb{R}.
\end{align*}
Let $p_1=\sfp(y_{\max} \!\mid\! x)$, $p_2=\sfp(\hh(x) \!\mid\! x)$, $\tau_1=h(x,\hh(x))$ and $\tau_2=h(x,y_{\max})$ to simplify the notation. 
Then, by the definition of $h_{\mu}$, we have for any $h\in \sH$ and $x\in \sX$,
\begin{align*}
&\sC_{\ell^{\mathrm{cstnd}}}(h,x) - \inf_{\mu \in \Rset}\sC_{\ell^{\mathrm{cstnd}}}(h_{\mu},x)\\
& = \sup_{\mu\in \Rset} \bigg\{\paren*{1-p_1}\bracket*{ \Phi\paren*{-\tau_2}-\Phi\paren*{-\tau_1+\mu}}+  \paren*{1-p_2}\bracket*{ \Phi\paren*{-\tau_1}-\Phi\paren*{-\tau_2-\mu}} \bigg\}\\
& = \sup_{\mu\in \Rset} \bigg\{\frac{2-\psum-p_1+p_2}{2}\bracket*{ \Phi\paren*{-\tau_2}-\Phi\paren*{-\tau_1+\mu}}+  \frac{2-\psum+p_1-p_2}{2}\bracket*{ \Phi\paren*{-\tau_1}-\Phi\paren*{-\tau_2-\mu}} \bigg\}\tag{$\psum=p_1+p_2\in \bracket*{\frac{1}{n-1},1}$}\\
& = \inf_{\psum\in \bracket*{\frac{1}{n-1},1}}\inf_{\substack{\tau_1\geq \max\curl*{\tau_2,0}}}\sup_{\mu\in \Rset} \bigg\{\frac{2-\psum-p_1+p_2}{2}\bracket*{ \Phi\paren*{-\tau_2}-\Phi\paren*{-\tau_1+\mu}} \\
& \qquad +  \frac{2-\psum+p_1-p_2}{2}\bracket*{ \Phi\paren*{-\tau_1}-\Phi\paren*{-\tau_2-\mu}} \bigg\}\tag{$\tau_1\geq 0$, $\tau_2\leq \tau_1$}\\
& = \sT^{\rm{cstnd}}\paren*{p_1-p_2}\\
& = \sT^{\rm{cstnd}}\paren*{\Delta\sC_{\ell_{0-1},\sH}(h,x)} \tag{by Lemma~\ref{lemma:explicit_assumption_01-chcb}}.
\end{align*}
where for $n=2$, an additional constraint $\tau_1+\tau_2=0$ is imposed and the expression of $\sT^{\rm{comp}}$ is simplified.
Since $\sT^{\rm{cstnd}}$ is convex, by Jensen's inequality, we obtain
for any hypothesis $h \in \sH$ and any distribution,
\begin{align*}
&\sT^{\rm{cstnd}}\paren*{\sR_{\ell_{0-1}}(h)-\sR_{\ell_{0-1}}^*(\sH)+\sM_{\ell_{0-1}}(\sH)}\\
&=\sT^{\rm{cstnd}}\paren*{\E_{X}\bracket*{\Delta\sC_{\ell_{0-1},\sH}(h,x)}}\\
&\leq \E_{X}\bracket*{\sT^{\rm{cstnd}}\paren*{\Delta\sC_{\ell_{0-1},\sH}(h,x)}}\\
& \leq \E_{X}\bracket*{\Delta\sC_{\ell^{\mathrm{cstnd}},\sH}(h,x)}\\
& =\sR_{\ell^{\mathrm{cstnd}}}(h)-\sR_{\ell^{\mathrm{cstnd}}}^*(\sH)+\sM_{\ell^{\mathrm{cstnd}}}(\sH).
\end{align*}
For the second part, we first consider $n=2$. For any $t\in [0,1]$, we consider the distribution that concentrates on a singleton $\curl*{x}$ and satisfies $\sfp(1 \!\mid\! x)=\frac{1+t}{2}$, $\sfp(2 \!\mid\! x)=\frac{1-t}{2}$. For any $\e>0$, by the definition of infimum, we can take $h\in \sH$ such that $h(x,2)=\tau_{\e}\geq 0$ and satisfies 
\begin{align*}
\sup\limits_{\mu\in \Rset} \curl*{\frac{1 - t}{2}\bracket*{ \Phi\paren*{\tau_{\e}}-\Phi\paren*{-\tau_{\e}+\mu}} +  \frac{1 + t}{2}\bracket*{ \Phi\paren*{-\tau_{\e}}-\Phi\paren*{\tau_{\e}-\mu}}}<\sT^{\rm{cstnd}}(t)+\e.
\end{align*}
Then,
\begin{align*}
\sR_{\ell_{0-1}}(h)-
\sR^*_{\ell_{0-1}}(\sH)+\sM_{\ell_{0-1}}(\sH)
&=  \sR_{\ell_{0-1}}(h) - \mathbb{E}_{X} \bracket* {\sC^*_{\ell_{0-1}}(\sH,x)}\\
&=\sC_{\ell_{0-1}}(h,x) - \sC^*_{\ell_{0-1}}\paren*{\sH,x}\\
&=t
\end{align*}
and
\begin{align*}
\sT^{\rm{cstnd}}(t)
&\leq \sR_{\ell^{\mathrm{cstnd}}}(h) - \sR_{\ell^{\mathrm{cstnd}}}^*(\sH) +
\sM_{\ell^{\mathrm{cstnd}}}(\sH)\\
&=  \sR_{\ell^{\mathrm{cstnd}}}(h) - \mathbb{E}_{X} \bracket* {\sC^*_{\ell^{\mathrm{cstnd}}}(\sH,x)}\\
&=\sC_{\ell^{\mathrm{cstnd}}}(h,x)-\sC^*_{\ell^{\mathrm{cstnd}}}(\sH,x)\\
&=\sup\limits_{\mu\in \Rset} \curl*{\frac{1 - t}{2}\bracket*{ \Phi\paren*{\tau_{\e}}-\Phi\paren*{-\tau_{\e}+\mu}} +  \frac{1 + t}{2}\bracket*{ \Phi\paren*{-\tau_{\e}}-\Phi\paren*{\tau_{\e}-\mu}}}\\
&<\sT^{\rm{cstnd}}(t)+\e.
\end{align*}
By letting $\e\to 0$, we conclude the proof.
The proof for $n > 2$ directly extends from the case when $n = 2$. Indeed, For any $t\in [0,1]$, we consider the distribution that concentrates on a singleton $\curl*{x}$ and satisfies $\sfp(1 \!\mid\! x)=\frac{1+t}{2}$, $\sfp(2 \!\mid\! x)=\frac{1-t}{2}$, $\sfp(y \!\mid\! x)=0, \, 3\leq y\leq n$. For any $\e>0$, by the definition of infimum, we can take $h\in \sH$ such that $h(x,1)=\tau_{1,\e}$, $h(x,2)=\tau_{2,\e}$, $h(x,y)=0, \, 3\leq y\leq n$ and satisfies $\tau_{1,\e} + \tau_{2,\e} = 0$, and
\begin{align*}
&\inf\limits_{\psum\in \bracket*{\frac{1}{n-1},1}}\sup\limits_{\mu\in \Rset} \curl*{\frac{2-\psum-t}{2}\bracket*{ \Phi\paren*{-\tau_{2,\e}}-\Phi\paren*{-\tau_{1,\e}+\mu}} +  \frac{2-\psum + t}{2}\bracket*{ \Phi\paren*{-\tau_{1,\e}}-\Phi\paren*{-\tau_{2,\e}-\mu}}}\\
&=\sup\limits_{\mu\in \Rset} \curl*{\frac{1-t}{2}\bracket*{ \Phi\paren*{-\tau_{2,\e}}-\Phi\paren*{-\tau_{1,\e}+\mu}} +  \frac{1 + t}{2}\bracket*{ \Phi\paren*{-\tau_{1,\e}}-\Phi\paren*{-\tau_{2,\e}-\mu}}}\\
&<\sT^{\rm{cstnd}}(t)+\e.
\end{align*}
Then,
\begin{align*}
\sR_{\ell_{0-1}}(h)-
\sR^*_{\ell_{0-1}}(\sH)+\sM_{\ell_{0-1}}(\sH)=t
\end{align*}
and
\begin{align*}
\sT^{\rm{cstnd}}(t)
&\leq \sR_{\ell^{\mathrm{cstnd}}}(h) - \sR_{\ell^{\mathrm{cstnd}}}^*(\sH) +
\sM_{\ell^{\mathrm{cstnd}}}(\sH)<\sT^{\rm{cstnd}}(t)+\e.
\end{align*}
\end{proof}

\subsection{Characterization of
  \texorpdfstring{$\sT^{\rm{cstnd}}$}{T} (Theorem~\ref{Thm:char_cstnd})}
\label{app:char_cstnd}

\CharCstnd*
\begin{proof}
For $n=2$, we have
\begin{align*}
\sT^{\rm{cstnd}}(t)
&=\inf_{\tau\geq 0}\sup_{\mu\in \Rset} \curl*{\frac{1 - t}{2}\bracket*{ \Phi\paren*{\tau}-\Phi\paren*{-\tau+\mu}} +  \frac{1 + t}{2}\bracket*{ \Phi\paren*{-\tau}-\Phi\paren*{\tau-\mu}}}\\
&= \inf_{\tau\geq 0}\paren*{\frac{1 - t}{2} \Phi\paren*{\tau}+ \frac{1 + t}{2}\bracket*{\Phi\paren*{-\tau}}}-\inf_{\mu\in \Rset}\curl*{\frac{1 - t}{2} \Phi\paren*{-\tau+\mu} + \frac{1 + t}{2}\Phi\paren*{\tau-\mu}}\\
&= \inf_{\tau\geq 0}\paren*{\frac{1 - t}{2} \Phi\paren*{\tau}+ \frac{1 + t}{2}\bracket*{\Phi\paren*{-\tau}}}-\inf_{\mu\in \Rset}\curl*{\frac{1 - t}{2} \Phi\paren*{\mu} + \frac{1 + t}{2}\Phi\paren*{-\mu}}\\
&\geq \inf_{\tau\geq 0}\paren*{\Phi(0)-\Phi'(0)t\tau}-\inf_{\mu\in \Rset}\curl*{\frac{1 - t}{2} \Phi\paren*{\mu} + \frac{1 + t}{2}\Phi\paren*{-\mu}}
\tag{$\Phi$ is convex}\\
&= \Phi(0)-\inf_{\mu\in \Rset}\curl*{\frac{1 - t}{2} \Phi\paren*{\mu} + \frac{1 + t}{2}\Phi\paren*{-\mu}}
\tag{$\Phi'(0)<0$, $t\tau\geq 0$}
\end{align*}
where the equality can be achieved by $\tau=0$.

For $n>2$, we have
\begin{align*}
&\sT^{\rm{cstnd}}(t)=\inf_{\psum\in \bracket*{\frac{1}{n-1},1}}\inf_{\tau_1\geq \max\curl*{\tau_2,0}}\sup_{\mu\in \Rset} F\paren*{\psum,\tau_1,\tau_2,\mu}
\end{align*}
where we let
$F\paren*{\psum,\tau_1,\tau_2,\mu}=\frac{2-\psum-t}{2}\bracket*{
  \Phi\paren*{-\tau_2}-\Phi\paren*{-\tau_1+\mu}} + \frac{2-\psum +
  t}{2}\bracket*{ \Phi\paren*{-\tau_1}-\Phi\paren*{-\tau_2-\mu}}$. For
simplicity, we assume that $\Phi$ is differentiable. For general
convex $\Phi$, we can proceed by using left and right derivatives,
which are non-decreasing. Differentiate $F$ with respect to $\mu$, we
have
\begin{equation*}
\frac{\partial F}{\partial \mu}=\frac{\psum + t - 2}{2}\Phi'(-\tau_1+\mu)+\frac{2-\psum + t}{2}\Phi'(-\tau_2-\mu).
\end{equation*}
Using the fact that $\psum\in \bracket*{\frac{1}{n-1},1},t\in[0,1]$
and $\Phi'$ is non-decreasing, we obtain that $\frac{\partial
  F}{\partial \mu}$ is non-increasing. Furthermore, $\Phi'$ is
non-decreasing and non-positive, $\Phi$ is non-negative, we obtain
that $\Phi'(+ \infty)=0$. This implies that $\frac{\partial
  F}{\partial \mu}(+ \infty)\leq 0$ and $\frac{\partial
  F}{\partial \mu}(- \infty)\geq 0$. Therefore, there exists
$\mu_0\in \Rset$ such that
\begin{equation*}
\frac{\partial F}{\partial \mu}(\mu_0)=\frac{\psum + t - 2}{2}\Phi'(-\tau_1+\mu_0)+\frac{2-\psum + t}{2}\Phi'(-\tau_2-\mu_0)=0  
\end{equation*}
By taking $\mu=\tau_1-\tau_2$ and using the fact that $\Phi'(0)<0$, we have
\begin{equation*}
\frac{\partial F}{\partial \mu}(\tau_1-\tau_2)=\frac{\psum + t - 2}{2}\Phi'(-\tau_2)+\frac{2-\psum + t}{2}\Phi'(-\tau_1)< 0.
\end{equation*}
Thus, since $\frac{\partial F}{\partial \mu}$ is non-increasing, we obtain $\mu_0< \tau_1-\tau_2$.
Differentiate $F$ with respect to $\tau_2$ at $\mu_0$, we have
\begin{equation*}
\frac{\partial F}{\partial \tau_2}=\frac{\psum + t - 2}{2}\Phi'(-\tau_2)+\frac{2-\psum + t}{2}\Phi'(-\tau_2-\mu_0).
\end{equation*}
Since $\Phi'$ is non-decreasing, we obtain
\begin{equation*}
\frac{\partial F}{\partial \tau_2}\leq \frac{\psum + t - 2}{2}\Phi'(-\tau_2)+\frac{2-\psum + t}{2}\Phi'(-\tau_2-\tau_1+\tau_2)=\frac{\partial F}{\partial \mu}(\tau_1-\tau_2)<0,
\end{equation*}
which implies that the infimum $\inf_{\tau_1\geq \max\curl*{\tau_2,0}}$ is achieved when $\tau_2=\tau_1$.
Differentiate $F$ with respect to $\psum$ at $\mu_0$ and $\tau_1=\tau_2$, by the convexity of $\Phi$, we obtain
\begin{equation*}
\frac{\partial F}{\partial \psum}=\Phi(-\tau_1+\mu_0)-\Phi(-\tau_1)-\Phi(-\tau_1)+\Phi(-\tau_1-\mu_0)\geq 0,
\end{equation*}
which implies that the infimum $\inf_{\psum\in \bracket*{\frac{1}{n-1},1}}$ is achieved when $\psum=\frac{1}{n-1}$. Above all, we obtain
\begin{align*}
\sT^{\rm{cstnd}}(t)
&=\inf_{\tau\geq 0}\sup_{\mu\in \Rset}F\paren*{\frac{1}{n-1},\tau,\tau,\mu}\\
&=\inf_{\tau\geq 0}\curl*{\paren*{2-\frac{1}{n-1}}\Phi(-\tau) -\inf_{\mu\in \mathbb{R}}\curl*{\frac{2-t-\frac{1}{n-1}}{2}\Phi(-\tau+\mu)+\frac{2+t-\frac{1}{n-1}}{2}\Phi(-\tau-\mu)}}\\
&\geq \inf_{\tau\geq 0}\sup_{\mu\in \Rset}F\paren*{0,\tau,\tau,\mu}\\
&=\inf_{\tau\geq 0}\curl*{2\Phi(-\tau) -\inf_{\mu\in \mathbb{R}}\curl*{\frac{2-t}{2}\Phi(-\tau+\mu)+\frac{2+t}{2}\Phi(-\tau-\mu)}}.
\end{align*}
\end{proof}

\subsection{Computation of examples}
\label{app:examples-cstnd}

\paragraph{Example: $\Phi(t)=\Phi_{\rm{exp}}(t)=e^{-t}$.}
For $n = 2$, plugging in $\Phi(t)=e^{-t}$ in
Theorem~\ref{Thm:char_cstnd}, gives
\begin{align*}
\sT^{\rm{comp}}
&=1 -\inf_{\mu\in \mathbb{R}}\curl*{\frac{1 - t}{2}e^{-\mu}+\frac{1 + t}{2}e^{\mu}}\\
&=1-\sqrt{1-t^2} \tag{minimum achieved at $\mu = \frac{1}{2}\log\frac{1-t}{1+t}$}.
\end{align*}
For $n > 2$, plugging in $\Phi(t)=e^{-t}$ in Theorem~\ref{Thm:char_cstnd} yields
\begin{align*}
\sT^{\rm{comp}}
&\geq \inf_{\tau\geq 0}\curl*{2e^{\tau} -\inf_{\mu\in \mathbb{R}}\curl*{\frac{2-t}{2}e^{\tau-\mu}+\frac{2+t}{2}e^{\tau+\mu}}}\\
&\geq 2 -\inf_{\mu\in \mathbb{R}}\curl*{\frac{2-t}{2}e^{-\mu}+\frac{2+t}{2}e^{\mu}} \tag{minimum achieved at $\tau=0$}\\
&=2-\sqrt{4-t^2} \tag{minimum achieved at $\mu = \frac{1}{2}\log\frac{2-t}{2+t}$}.
\end{align*}

\paragraph{Example: $\Phi(t)=\Phi_{\rm{hinge}}(t)=\max\curl*{0,1 - t}$.}
For $n = 2$, plugging in $\Phi(t)=\max\curl*{0,1 - t}$ in Theorem~\ref{Thm:char_cstnd}, gives
\begin{align*}
\sT^{\rm{comp}}
&=1 -\inf_{\mu\in \mathbb{R}}\curl*{\frac{1 - t}{2}\max\curl*{0,1 - \mu}+\frac{1 + t}{2}\max\curl*{0,1 +\mu}}\\
&=t \tag{minimum achieved at $\mu = -1$}.
\end{align*}
For $n > 2$, plugging in $\Phi(t)=\max\curl*{0,1 - t}$ in Theorem~\ref{Thm:char_cstnd} yields
\begin{align*}
\sT^{\rm{comp}}
&\geq \inf_{\tau\geq 0}\curl*{2\max\curl*{0,1 +\tau} -\inf_{\mu\in \mathbb{R}}\curl*{\frac{2-t}{2}\max\curl*{0,1 +\tau-\mu}+\frac{2+t}{2}\max\curl*{0,1 +\tau+\mu}}}\\
&=2 -\inf_{\mu\in \mathbb{R}}\curl*{\frac{2-t}{2}\max\curl*{0,1-\mu}+\frac{2+t}{2}\max\curl*{0,1 +\mu}}\tag{minimum achieved at $\tau=0$}\\
&=t \tag{minimum achieved at $\mu = -1$}.
\end{align*}

\paragraph{Example: $\Phi(t)=\Phi_{\rm{sq-hinge}}(t)=(1 - t)^2 \mathds{1}_{t\leq 1}$.}
For $n = 2$, plugging in $\Phi(t)=(1 - t)^2 \mathds{1}_{t\leq 1}$ in Theorem~\ref{Thm:char_cstnd}, gives
\begin{align*}
\sT^{\rm{comp}}
&=1 -\inf_{\mu\in \mathbb{R}}\curl*{\frac{1 - t}{2}(1 - \mu)^2 \mathds{1}_{\mu\leq 1}+\frac{1 + t}{2}(1 +\mu)^2 \mathds{1}_{\mu\geq -1}}\\
&=t^2  \tag{minimum achieved at $\mu = -t$}.
\end{align*}
For $n > 2$, plugging in $\Phi(t)=(1 - t)^2 \mathds{1}_{t\leq 1}$ in Theorem~\ref{Thm:char_cstnd} yields
\begin{align*}
\sT^{\rm{comp}}
&\geq \inf_{\tau\geq 0}\curl*{2(1 +\tau)^2 \mathds{1}_{\tau\geq -1} -\inf_{\mu\in \mathbb{R}}\curl*{\frac{2-t}{2}(1 +\tau-\mu)^2 \mathds{1}_{-\tau+\mu\leq 1}+\frac{2+t}{2}(1 +\tau+\mu)^2 \mathds{1}_{\tau+\mu\geq -1}}}\\
&\geq 2 -\inf_{\mu\in \mathbb{R}}\curl*{\frac{2-t}{2}(1 -\mu)^2 \mathds{1}_{\mu\leq 1}+\frac{2+t}{2}(1+\mu)^2 \mathds{1}_{\mu\geq -1}} \tag{minimum achieved at $\tau=0$}\\
&=\frac{t^2}{2} \tag{minimum achieved at $\mu = -\frac{t}{2}$}.
\end{align*}

\paragraph{Example: $\Phi(t)=\Phi_{\rm{sq}}(t)=(1-t)^2$.}
For $n = 2$, plugging in $\Phi(t)=(1-t)^2$ in Theorem~\ref{Thm:char_cstnd}, gives
\begin{align*}
\sT^{\rm{comp}}
&=1 -\inf_{\mu\in \mathbb{R}}\curl*{\frac{1 - t}{2}(1-\mu)^2+\frac{1 + t}{2}(1+\mu)^2}\\
&=t^2 \tag{minimum achieved at $\mu = -t$}.
\end{align*}
For $n > 2$, plugging in $\Phi(t)=(1-t)^2$ in Theorem~\ref{Thm:char_cstnd} yields
\begin{align*}
\sT^{\rm{comp}}
&\geq \inf_{\tau\geq 0}\curl*{2(1+\tau)^2 -\inf_{\mu\in \mathbb{R}}\curl*{\frac{2-t}{2}(1+\tau-\mu)^2+\frac{2+t}{2}(1+\tau+\mu)^2}}\\
&\geq 2 -\inf_{\mu\in \mathbb{R}}\curl*{\frac{2-t}{2}(1-\mu)^2+\frac{2+t}{2}(1+\mu)^2}\tag{minimum achieved at $\tau=0$}\\
&=\frac{t^2}{2} \tag{minimum achieved at $\mu = -\frac{t}{2}$}.
\end{align*}

\section{Extensions of comp-sum losses}
\label{app:extension-comp}

\subsection{Proof of \texorpdfstring{$\ov\sH$}{H}-consistency bounds
  with \texorpdfstring{$\ov\sT^{\rm{comp}}$}{T} (Theorem~\ref{Thm:bound_comp_BD})}
\label{app:bound_comp_BD}

\BoundCompBD*
\begin{proof}
For the comp-sum loss $\ell^{\mathrm{comp}}$, the conditional
$\ell^{\mathrm{comp}}$-risk can be expressed as follows:
\begin{align*}
&\sC_{\ell^{\mathrm{comp}}}(h,x)\\
& = \sum_{y\in \sY} \sfp(y \!\mid\! x) \ell^{\mathrm{comp}}(h, x, y)\\
&= \sum_{y\in \sY} \sfp(y \!\mid\! x) \Phi\paren*{\frac{e^{h(x,y)}}{\sum_{y'\in \sY}e^{h(x,y')}}}\\
&= \sum_{y\in \sY} \sfp(y \!\mid\! x) \Phi\paren*{S_h(x,y)}\\
& = \sfp(y_{\max} \!\mid\! x) \Phi\paren*{S_h(x,y_{\max})}+\sfp(\hh(x) \!\mid\! x) \Phi\paren*{S_h(x,\hh(x))} + \sum_{y\notin \curl*{y_{\max},\hh(x)}}\sfp(y \!\mid\! x) \Phi\paren*{S_h(x,y)}
\end{align*}
where we let $S_h(x,y)=\frac{e^{h(x,y)}}{\sum_{y'\in \sY}e^{h(x,y')}}$
for any $y\in \sY$ with the constraint that $\sum_{y\in
  \sY}S_h(x,y)=1$. Note that for any $h\in \sH$, \begin{align*}
  \frac{1}{1+(n-1)e^{2\Lambda(x)}}=\frac{e^{-\Lambda(x)}}{e^{-\Lambda(x)}+(n-1)e^{\Lambda(x)}}\leq
  S_h(x,y)\leq
  \frac{e^{\Lambda(x)}}{e^{\Lambda(x)}+(n-1)e^{-\Lambda(x)}}=\frac{1}{1+(n-1)e^{-2\Lambda(x)}}
\end{align*}
Therefore for any $(x,y)\in \sX\times \sY$, $S_h(x,y)\in
\bracket*{S_{\min},S_{\max}}$, where we let
$S_{\max}=\frac{1}{1+(n-1)e^{-2\Lambda(x)}}$ and
$S_{\min}=\frac{1}{1+(n-1)e^{2\Lambda(x)}}$. Furthermore, all values
in $\bracket*{S_{\min},S_{\max}}$ of $S_h$ can be reached for some
$h\in \sH$. Observe that $0\leq S_{\max}+S_{\min}\leq 1$.  Let
$y_{\max}=\argmax_{y\in \sY}\sfp(y \!\mid\! x)$, where we choose the label with
the highest index under the natural ordering of labels as the
tie-breaking strategy.  For any $h \in \sH$ such that $\hh(x) \neq
y_{\max}$ and $x\in \sX$, we can always find a family of hypotheses
$\curl*{h_{\mu}}\subset \sH$ such that
$S_{h,\mu}(x,\cdot)=\frac{e^{h_{\mu}(x,\cdot)}}{\sum_{y'\in
    \sY}e^{h_{\mu}(x,y')}}$ take the following values:
\begin{align*}
S_{h,\mu}(x,y) = 
\begin{cases}
  S_h(x, y) & \text{if $y \not \in \curl*{y_{\max}, \hh(x)}$}\\
 S_h(x, y_{\max}) + \mu & \text{if $y = \hh(x)$}\\
  S_h(x,\hh(x)) -\mu & \text{if $y = y_{\max}$}.
\end{cases} 
\end{align*}
Note that $S_{h,\mu}$ satisfies the constraint:
\begin{align*}
 \sum_{y\in \sY}S_{h,\mu}(x, y)=\sum_{y\in \sY}
 S_h(x, y)=1.
\end{align*}
Since $S_{h,\mu}(x,y)\in \bracket*{S_{\min},S_{\max}}$, we have the following constraints on $\mu$:
\begin{equation}
\label{eq:cons-mu}
\begin{aligned}
&S_{\min}-S_h(x, y_{\max})\leq \mu\leq S_{\max}-S_h(x, y_{\max})\\
&S_h(x,\hh(x))- S_{\max}\leq \mu\leq S_h(x,\hh(x))-S_{\min}.
\end{aligned}
\end{equation}
Let $p_1=\sfp(y_{\max} \!\mid\! x)$, $p_2=\sfp(\hh(x) \!\mid\! x)$, $\tau_1=S_h(x,\hh(x))$ and
$\tau_2=S_h(x,y_{\max})$ to simplify the notation.
Let $\ov C=\curl*{\mu \in \Rset: \mu \text{ verify constraint \eqref{eq:cons-mu}}}$. Since $S_h(x,\hh(x))- S_{\max}\leq S_{\max}-S_h(x, y_{\max})$ and $S_{\min}-S_h(x, y_{\max})\leq S_h(x,\hh(x))-S_{\min}$, $\ov C$ is not an empty set and can be expressed as $\ov C=\bracket*{\max\curl*{S_{\min}-\tau_2,\tau_1- S_{\max}},\min\curl*{S_{\max}-\tau_2,\tau_1- S_{\min}}}$.

Then, by the definition of $S_{h,\mu}$, we have for any $h\in \sH$ and
$x\in \sX$,
\begin{align*}
&\sC_{\ell^{\mathrm{comp}}}(h,x) - \inf_{\mu\in \ov C}\sC_{\ell^{\mathrm{comp}}}(h_{\mu},x)\\
& = \sup_{\mu\in \ov C} \bigg\{p_1\bracket*{\Phi\paren*{\tau_2}-\Phi\paren*{\tau_1-\mu}}+  p_2\bracket*{ \Phi\paren*{\tau_1}-\Phi\paren*{\tau_2+\mu}} \bigg\}\\
& = \sup_{\mu\in \ov C} \bigg\{\frac{\psum+p_1-p_2}{2}\bracket*{\Phi\paren*{\tau_2}-\Phi\paren*{\tau_1-\mu}}+  \frac{\psum-p_1+p_2}{2}\bracket*{ \Phi\paren*{\tau_1}-\Phi\paren*{\tau_2+\mu}} \bigg\}\tag{$\psum=p_1+p_2\in \bracket*{\frac{1}{n-1}\vee p_1-p_2,1}$}\\
& \geq \inf_{\psum\in \bracket*{\frac{1}{n-1}\vee p_1-p_2,1}}\inf_{\substack{S_{\min}\leq \tau_2\leq \tau_1\leq S_{\max}\\ \tau_1+\tau_2\leq 1}}\sup_{\mu\in \ov C} \bigg\{\frac{\psum+p_1-p_2}{2}\bracket*{\Phi\paren*{\tau_2}-\Phi\paren*{\tau_1-\mu}}\\
&\qquad +  \frac{\psum-p_1+p_2}{2}\bracket*{ \Phi\paren*{\tau_1}-\Phi\paren*{\tau_2+\mu}} \bigg\}\tag{$S_{\min}\leq \tau_2\leq \tau_1\leq S_{\max}$, $\tau_1+\tau_2\leq 1$}\\
&\geq \inf_{\psum\in \bracket*{\frac{1}{n-1}\vee p_1-p_2,1}}\inf_{\substack{S_{\min}\leq \tau_2\leq \tau_1\leq S_{\max}\\ \tau_1+\tau_2\leq 1}}\sup_{\mu\in C} \bigg\{\frac{\psum+p_1-p_2}{2}\bracket*{\Phi\paren*{\tau_2}-\Phi\paren*{\tau_1-\mu}}\\
&\qquad +  \frac{\psum-p_1+p_2}{2}\bracket*{ \Phi\paren*{\tau_1}-\Phi\paren*{\tau_2+\mu}} \bigg\}\tag{$S_{\min}\leq s_{\min}\leq s_{\max}\leq S_{\max}$}\\
& = \sT^{\rm{comp}}\paren*{p_1-p_2}\\
& = \sT^{\rm{comp}}\paren*{\Delta\sC_{\ell_{0-1},\sH}(h,x)}, \tag{by Lemma~\ref{lemma:explicit_assumption_01-chcb}}
\end{align*}
where $\C=\bracket*{\max\curl*{s_{\min}-\tau_2,\tau_1- s_{\max}},\min\curl*{s_{\max}-\tau_2,\tau_1- s_{\min}}}\subset \ov \C$, $s_{\max}=\frac{1}{1+(n-1)e^{-2\inf_{x}\Lambda(x)}}$ and  $s_{\min}=\frac{1}{1+(n-1)e^{2\inf_{x}\Lambda(x)}}$.
Note that for $n=2$, an additional constraint $\tau_1+\tau_2=1$ is imposed and the expression can be simplified as
\begin{align*}
&\sC_{\ell^{\mathrm{comp}}}(h,x) - \inf_{\mu\in \ov C}\sC_{\ell^{\mathrm{comp}}}(h_{\mu},x) \\
&\geq\inf\limits_{\tau\in \bracket*{0,\frac12}}\sup\limits_{\mu\in \bracket*{s_{\min}-\tau,1-\tau-s_{\min}}} \curl*{\frac{1 + p_1-p_2}{2}\bracket*{\Phi\paren*{\tau}-\Phi\paren*{1-\tau-\mu}} + \frac{1 - p_1+p_2}{2}\bracket*{ \Phi\paren*{1-\tau}-\Phi\paren*{\tau+\mu}}}\\
& = \sT^{\rm{comp}}\paren*{p_1-p_2}\\
& = \sT^{\rm{comp}}\paren*{\Delta\sC_{\ell_{0-1},\sH}(h,x)}, \tag{by Lemma~\ref{lemma:explicit_assumption_01-chcb}}
\end{align*}
where we use the fact that $s_{\max}+s_{\min}=1$ and $P=1$ when $n=2$.
Since $\sT^{\rm{comp}}$ is convex, by Jensen's inequality, we obtain for any hypothesis $h \in \sH$ and any distribution,
\begin{align*}
&\sT^{\rm{comp}}\paren*{\sR_{\ell_{0-1}}(h)-\sR_{\ell_{0-1}}^*(\sH)+\sM_{\ell_{0-1}}(\sH)}\\
&=\sT^{\rm{comp}}\paren*{\E_{X}\bracket*{\Delta\sC_{\ell_{0-1},\sH}(h,x)}}\\
&\leq \E_{X}\bracket*{\sT^{\rm{comp}}\paren*{\Delta\sC_{\ell_{0-1},\sH}(h,x)}}\\
& \leq \E_{X}\bracket*{\Delta\sC_{\ell^{\mathrm{comp}},\sH}(h,x)}\\
& =\sR_{\ell^{\mathrm{comp}}}(h)-\sR_{\ell^{\mathrm{comp}}}^*(\sH)+\sM_{\ell^{\mathrm{comp}}}(\sH).
\end{align*}
For the second part, we first consider $n=2$. For any $t\in [0,1]$, we
consider the distribution that concentrates on a singleton $\curl*{x}$
and satisfies $\sfp(1 \!\mid\! x)=\frac{1+t}{2}$, $\sfp(2 \!\mid\! x)=\frac{1-t}{2}$. For any
$\e>0$, by the definition of infimum, we can take $h\in \sH$ such that
$S_h(x,1)=\tau_{\e}\in \bracket*{0,\frac12}$ and satisfies
\begin{align*}
\sup\limits_{\mu\in \bracket*{s_{\min}-\tau_{\e},1-\tau_{\e}-s_{\min}}} \curl*{\frac{1 + t}{2}\bracket*{\Phi\paren*{\tau_{\e}}-\Phi\paren*{1-\tau_{\e}-\mu}} + \frac{1 - t}{2}\bracket*{ \Phi\paren*{1-\tau_{\e}}-\Phi\paren*{\tau_{\e}+\mu}}}<\sT^{\rm{comp}}(t)+\e.
\end{align*}
Then,
\begin{align*}
\sR_{\ell_{0-1}}(h)-
\sR^*_{\ell_{0-1}}(\sH)+\sM_{\ell_{0-1}}(\sH)
&=  \sR_{\ell_{0-1}}(h) - \mathbb{E}_{X} \bracket* {\sC^*_{\ell_{0-1}}(\sH,x)}\\
&=\sC_{\ell_{0-1}}(h,x) - \sC^*_{\ell_{0-1}}\paren*{\sH,x}\\
&=t
\end{align*}
and
\begin{align*}
\sT^{\rm{comp}}(t)
&\leq \sR_{\ell^{\mathrm{comp}}}(h) - \sR_{\ell^{\mathrm{comp}}}^*(\sH) +
\sM_{\ell^{\mathrm{comp}}}(\sH)\\
&=  \sR_{\ell^{\mathrm{comp}}}(h) - \mathbb{E}_{X} \bracket* {\sC^*_{\ell^{\mathrm{comp}}}(\sH,x)}\\
&=\sC_{\ell^{\mathrm{comp}}}(h,x)-\sC^*_{\ell^{\mathrm{comp}}}(\sH,x)\\
&=\sup\limits_{\mu\in \bracket*{s_{\min}-\tau_{\e},1-\tau_{\e}-s_{\min}}} \curl*{\frac{1 + t}{2}\bracket*{\Phi\paren*{\tau_{\e}}-\Phi\paren*{1-\tau_{\e}-\mu}} + \frac{1 - t}{2}\bracket*{ \Phi\paren*{1-\tau_{\e}}-\Phi\paren*{\tau_{\e}+\mu}}}\\
&<\sT^{\rm{comp}}(t)+\e.
\end{align*}
By letting $\e\to 0$, we conclude the proof.
The proof for $n > 2$ directly extends from the case when $n = 2$. Indeed, For any $t\in [0,1]$, we consider the distribution that concentrates on a singleton $\curl*{x}$ and satisfies $\sfp(1 \!\mid\! x)=\frac{1+t}{2}$, $\sfp(2 \!\mid\! x)=\frac{1-t}{2}$, $\sfp(y \!\mid\! x)=0,\,3\leq y\leq n$. For any $\e>0$, by the definition of infimum, we can take $h\in \sH$ such that $S_h(x,1)=\tau_{1,\e}$, $S_h(x,2)=\tau_{2,\e}$ and $S_h(x,y)=0,\,3\leq y\leq n$ and satisfies $\tau_{1,\e} + \tau_{2,\e} = 1$, and
\begin{align*}
&\inf\limits_{\psum\in \bracket*{\frac{1}{n-1}\vee t,1}}\sup\limits_{\mu\in \C} \curl*{\frac{\psum + t}{2}\bracket*{\Phi\paren*{\tau_{2,\e}}-\Phi\paren*{\tau_{1,\e}-\mu}} + \frac{\psum-t}{2}\bracket*{ \Phi\paren*{\tau_{1,\e}}-\Phi\paren*{\tau_{2,\e}+\mu}}}\\
&=\sup\limits_{\mu\in \C} \curl*{\frac{1 + t}{2}\bracket*{\Phi\paren*{\tau_{2,\e}}-\Phi\paren*{\tau_{1,\e}-\mu}} + \frac{1-t}{2}\bracket*{ \Phi\paren*{\tau_{1,\e}}-\Phi\paren*{\tau_{2,\e}+\mu}}}\\
&<\sT^{\rm{comp}}(t)+\e.
\end{align*}
Then,
\begin{align*}
\sR_{\ell_{0-1}}(h)-
\sR^*_{\ell_{0-1}}(\sH)+\sM_{\ell_{0-1}}(\sH)
=t
\end{align*}
and
\begin{align*}
\sT^{\rm{comp}}(t)
\leq \sR_{\ell^{\mathrm{comp}}}(h) - \sR_{\ell^{\mathrm{comp}}}^*(\sH) +
\sM_{\ell^{\mathrm{comp}}}(\sH)
<\sT^{\rm{comp}}(t)+\e.
\end{align*}
By letting $\e\to 0$, we conclude the proof.
\end{proof}

\subsection{Logistic loss}
\label{app:bound-logistic}

\Logistic*
\begin{proof}
For the multinomial logistic loss $\ell_{\rm{log}}$, plugging in $\Phi(t)=-\log(t)$ in Theorem~\ref{Thm:bound_comp_BD}, gives $\ov\sT^{\rm{comp}}$
\begin{equation*}
\begin{aligned}
\geq \inf\limits_{\psum\in \bracket*{\frac{1}{n-1}\vee t,1}}\inf\limits_{\substack{S_{\min}\leq \tau_2\leq \tau_1\leq S_{\max}\\ \tau_1+\tau_2\leq 1}}\sup\limits_{\mu\in \C} \curl*{\frac{\psum + t}{2}\bracket*{-\log\paren*{\tau_2}+\log\paren*{\tau_1-\mu}} + \frac{\psum-t}{2}\bracket*{ -\log\paren*{\tau_1}+\log\paren*{\tau_2+\mu}}}
\end{aligned}
\end{equation*}
where $\C=\bracket*{\max\curl*{s_{\min}-\tau_2,\tau_1-
    s_{\max}},\min\curl*{s_{\max}-\tau_2,\tau_1- s_{\min}}}$.  Here,
we only compute the expression for $n>2$. The expression for $n=2$
will lead to the same result since it can be viewed as a special case
of the expression for $n>2$.  By differentiating with respect to
$\tau_2$ and $\psum$, we can see that the infimum is achieved when
$\tau_1=\tau_2=\frac{s_{\min}+s_{\max}}{2}$ and $\psum=1$ modulo some
elementary analysis. Thus, $\ov\sT^{\rm{comp}}$ can be reformulated as
\begin{align*}
\ov\sT^{\rm{comp}}&=\sup\limits_{\mu\in \C} \bigg\{\frac{1 + t}{2}\bracket*{-\log\paren*{\frac{s_{\min}+s_{\max}}{2}}+\log\paren*{\frac{s_{\min}+s_{\max}}{2}-\mu}}\\
&\qquad+ \frac{1-t}{2}\bracket*{ -\log\paren*{\frac{s_{\min}+s_{\max}}{2}}+\log\paren*{\frac{s_{\min}+s_{\max}}{2}+\mu}}\bigg\}\\
& = -\log\paren*{\frac{s_{\min}+s_{\max}}{2}}+\sup\limits_{\mu\in \C}g(\mu)
\end{align*}
where
$\C=\bracket*{\frac{s_{\min}-s_{\max}}{2},\frac{s_{\max}-s_{\min}}{2}}$
and $g(\mu)=\frac{1 +
  t}{2}\log\paren*{\frac{s_{\min}+s_{\max}}{2}-\mu}+\frac{1-t}{2}\log\paren*{\frac{s_{\min}+s_{\max}}{2}+\mu}$. Since
$g$ is continuous, it attains its supremum over a compact set.  Note
that $g$ is concave and differentiable.  In view of that, the maximum
over the open set $(-\infty, + \infty)$ can be obtained by setting
its gradient to zero.  Differentiate $g(\mu)$ to optimize, we obtain
\begin{align*}
g(\mu^*)=0, \quad \mu^*=-\frac{t\paren*{s_{\min}+s_{\max}}}{2}.   
\end{align*}
Moreover, by the concavity, $g(\mu)$ is non-increasing when $\mu\geq \mu^*$.
Since $s_{\max}-s_{\min}\geq 0$, we have
\begin{align*}
  \mu^*\leq 0\leq \frac{s_{\max}-s_{\min}}{2}
\end{align*}
 In view of the constraint $C$, if $\mu^*\geq
 \frac{s_{\min}-s_{\max}}{2}$, the maximum is achieved by
 $\mu=\mu^*$. Otherwise, if $\mu^*< \frac{s_{\min}-s_{\max}}{2}$,
 since $g(\mu)$ is non-increasing when $\mu\geq \mu^*$, the maximum is
 achieved by $\mu=\frac{s_{\min}-s_{\max}}{2}$.  Since $\mu^*\geq
 \frac{s_{\min}-s_{\max}}{2}$ is equivalent to $t\leq
 \frac{s_{\max}-s_{\min}}{s_{\min}+s_{\max}}$, the maximum can be
 expressed as
\begin{equation*}
\begin{aligned}
\max_{\mu \in C} g(\mu)
&=\begin{cases}
g(\mu^*) &   t\leq \frac{s_{\max}-s_{\min}}{s_{\min}+s_{\max}}\\
g\paren*{\frac{s_{\min}-s_{\max}}{2}} & \text{otherwise}
\end{cases}
\end{aligned}
\end{equation*}
Computing the value of $g$ at these points yields:
\begin{equation*}
\begin{aligned}
g(\mu^*) &= \frac{1+t}{2}\log\frac{(1+t)(s_{\min}+s_{\max})}{2}+\frac{1-t}{2}\log\frac{(1-t)(s_{\min}+s_{\max})}{2}\\
g\paren*{\frac{s_{\min}-s_{\max}}{2}} &= \frac{1+t}{2}\log(s_{\max})+\frac{1-t}{2}\log(s_{\min})
\end{aligned}
\end{equation*}
Then, if $t\leq \frac{s_{\max}-s_{\min}}{s_{\min}+s_{\max}}$, we obtain
\begin{align*}
\ov\sT^{\rm{comp}}&= -\log\paren*{\frac{s_{\min}+s_{\max}}{2}}+\frac{1+t}{2}\log\frac{(1+t)(s_{\min}+s_{\max})}{2}+\frac{1-t}{2}\log\frac{(1-t)(s_{\min}+s_{\max})}{2}\\
&=\frac{1+t}{2}\log\paren*{1+t}+\frac{1-t}{2}\log\paren*{1-t}.
\end{align*}
Otherwise, we obtain
\begin{align*}
\ov\sT^{\rm{comp}}&=-\log\paren*{\frac{s_{\min}+s_{\max}}{2}}+\frac{1+t}{2}\log(s_{\max})+\frac{1-t}{2}\log(s_{\min})\\
&= \frac{t}{2}\log\paren*{\frac{s_{\max}}{s_{\min}}}+\log\paren*{\frac{2\sqrt{s_{\max}s_{\min}}}{s_{\max}+s_{\min}}}.
\end{align*}
Since $\ov\sT^{\rm{comp}}$ is convex, by Theorem~\ref{Thm:bound_comp_BD}, for any $h\in \ov\sH$ and any distribution,
\begin{align*}
\sR_{\ell_{0-1}}\paren*{h}-\sR_{\ell_{0-1}}^*\paren*{\ov \sH}+\sM_{\ell_{0-1}}\paren*{\ov \sH}
& \leq \Psi^{-1}
\paren*{\sR_{\ell_{\rm{log}}}\paren*{h}
  - \sR_{\ell_{\rm{log}}}^*\paren*{\ov \sH}+\sM_{\ell_{\rm{log}}}\paren*{\ov \sH}},
\end{align*}
where
\begin{align*}
\Psi(t)=\begin{cases}
 \frac{1 + t}{2}\log(1 + t)+\frac{1 - t}{2}\log(1 - t)& t\leq \frac{s_{\max}-s_{\min}}{s_{\min}+s_{\max}}\\
 \frac{t}{2}\log\paren*{\frac{s_{\max}}{s_{\min}}}+\log\paren*{\frac{2\sqrt{s_{\max}s_{\min}}}{s_{\max}+s_{\min}}}& \mathrm{otherwise}.
\end{cases}
\end{align*}
\end{proof}

\subsection{Sum exponential loss}
\label{app:bound-exponential}

\Exponential*
\begin{proof}
For the sum exponential loss $\ell_{\rm{exp}}$, plugging in
$\Phi(t)=\frac{1}{t}-1$ in Theorem~\ref{Thm:bound_comp_BD}, gives
$\ov\sT^{\rm{comp}}$
\begin{equation*}
\begin{aligned}
\geq \inf\limits_{\psum\in \bracket*{\frac{1}{n-1}\vee t,1}}\inf\limits_{\substack{S_{\min}\leq \tau_2\leq \tau_1\leq S_{\max}\\ \tau_1+\tau_2\leq 1}}\sup\limits_{\mu\in \C} \curl*{\frac{\psum + t}{2}\bracket*{\frac{1}{\tau_2}-\frac{1}{\tau_1-\mu}} + \frac{\psum-t}{2}\bracket*{\frac{1}{\tau_1}-\frac{1}{\tau_2+\mu}}}
\end{aligned}
\end{equation*}
where $\C=\bracket*{\max\curl*{s_{\min}-\tau_2,\tau_1-
    s_{\max}},\min\curl*{s_{\max}-\tau_2,\tau_1- s_{\min}}}$. Here, we
only compute the expression for $n>2$. The expression for $n=2$ will
lead to the same result since it can be viewed as a special case of
the expression for $n>2$.  By differentiating with respect to $\tau_2$
and $\psum$, we can see that the infimum is achieved when
$\tau_1=\tau_2=\frac{s_{\min}+s_{\max}}{2}$ and $\psum=1$ modulo some
elementary analysis. Thus, $\ov\sT^{\rm{comp}}$ can be reformulated as
\begin{align*}
\ov\sT^{\rm{comp}}&=\sup\limits_{\mu\in \C} \bigg\{\frac{1 + t}{2}\bracket*{\frac{2}{s_{\min}+s_{\max}}-\frac{2}{s_{\min}+s_{\max}-2\mu}}\\
&\qquad+ \frac{1-t}{2}\bracket*{ \frac{2}{s_{\min}+s_{\max}}-\frac{2}{s_{\min}+s_{\max}+2\mu}}\bigg\}\\
& =\frac{2}{s_{\min}+s_{\max}}+\sup\limits_{\mu\in \C}g(\mu)
\end{align*}
where
$\C=\bracket*{\frac{s_{\min}-s_{\max}}{2},\frac{s_{\max}-s_{\min}}{2}}$
and $g(\mu)=-\frac{1 +
  t}{s_{\min}+s_{\max}-2\mu}-\frac{1-t}{s_{\min}+s_{\max}+2\mu}$.
Since $g$ is continuous, it attains its supremum over a compact set.
Note that $g$ is concave and differentiable.  In view of that, the
maximum over the open set $(-\infty, + \infty)$ can be obtained by
setting its gradient to zero.  Differentiate $g(\mu)$ to optimize, we
obtain
\begin{align*}
g(\mu^*)=0, \quad \mu^*=\frac{s_{\min}+s_{\max}}{2} \frac{\sqrt{1-t}-\sqrt{1+t}}{\sqrt{1+t}+ \sqrt{1-t}}
\end{align*}
Moreover, by the concavity, $g(\mu)$ is non-increasing when $\mu\geq \mu^*$.
Since $s_{\max}-s_{\min}\geq 0$, we have
\begin{align*}
  \mu^*\leq 0\leq \frac{s_{\max}-s_{\min}}{2}
\end{align*}
In view of the constraint $C$, if $\mu^*\geq
\frac{s_{\min}-s_{\max}}{2}$, the maximum is achieved by
$\mu=\mu^*$. Otherwise, if $\mu^*< \frac{s_{\min}-s_{\max}}{2}$, since
$g(\mu)$ is non-increasing when $\mu\geq \mu^*$, the maximum is
achieved by $\mu=\frac{s_{\min}-s_{\max}}{2}$.  Since $\mu^*\geq
\frac{s_{\min}-s_{\max}}{2}$ is equivalent to $t\leq
\frac{s_{\max}^2-s_{\min}^2}{s_{\min}^2+s_{\max}^2}$, the maximum can
be expressed as
\begin{equation*}
\begin{aligned}
\max_{\mu \in C} g(\mu)
&=\begin{cases}
g(\mu^*) &   t\leq \frac{s_{\max}^2-s_{\min}^2}{s_{\min}^2+s_{\max}^2}\\
g\paren*{\frac{s_{\min}-s_{\max}}{2}} & \text{otherwise}
\end{cases}
\end{aligned}
\end{equation*}
Computing the value of $g$ at these points yields:
\begin{equation*}
\begin{aligned}
g(\mu^*) &= 1-\sqrt{1-t^2}-\frac{2}{s_{\min}+s_{\max}}\\
g\paren*{\frac{s_{\min}-s_{\max}}{2}} &= -\frac{1+t}{2s_{\max}}
-\frac{1-t}{2s_{\min}}
\end{aligned}
\end{equation*}
Then, if $t\leq \frac{s_{\max}^2-s_{\min}^2}{s_{\min}^2+s_{\max}^2}$, we obtain
\begin{align*}
\ov\sT^{\rm{comp}}&= \frac{2}{s_{\min}+s_{\max}}+1-\sqrt{1-t^2}-\frac{2}{s_{\min}+s_{\max}}\\
&=1-\sqrt{1-t^2}.
\end{align*}
Otherwise, we obtain
\begin{align*}
\ov\sT^{\rm{comp}}&=\frac{2}{s_{\min}+s_{\max}}-\frac{1+t}{2s_{\max}}-\frac{1-t}{2s_{\min}}\\
&= \frac{s_{\max}-s_{\min}}{2s_{\max}s_{\min}}t-\frac{\paren*{s_{\max}-s_{\min}}^2}{2s_{\max}s_{\min}\paren*{s_{\max}+s_{\min}}}.
\end{align*}
Since $\ov\sT^{\rm{comp}}$ is convex, by Theorem~\ref{Thm:bound_comp_BD}, for any $h\in \ov\sH$ and any distribution,
\begin{align*}
\sR_{\ell_{0-1}}\paren*{h}-\sR_{\ell_{0-1}}^*\paren*{\ov \sH}+\sM_{\ell_{0-1}}\paren*{\ov \sH}
& \leq \Psi^{-1}
\paren*{\sR_{\ell_{\rm{exp}}}\paren*{h}-\sR_{\ell_{\rm{exp}}}^*\paren*{\ov \sH}+\sM_{\ell_{\rm{exp}}}\paren*{\ov \sH}},
\end{align*}
where
\begin{align*}
\Psi(t)=\begin{cases}
 1-\sqrt{1 - t^2} & t\leq \frac{s_{\max}^2-s_{\min}^2}{s_{\min}^2+s_{\max}^2}\\
 \frac{s_{\max}-s_{\min}}{2s_{\max}s_{\min}}t-\frac{\paren*{s_{\max}-s_{\min}}^2}{2s_{\max}s_{\min}\paren*{s_{\max}+s_{\min}}}& \mathrm{otherwise}.
\end{cases}
\end{align*}
\end{proof}

\subsection{Generalized cross-entropy loss}
\label{app:bound-gce}

\begin{restatable}[\textbf{$\ov \sH$-consistency bounds for generalized cross-entropy loss}]{theorem}{GCE}
\label{thm:bound-gce}
For any $h\in \ov \sH$ and any distribution, we have
\begin{align*}
\sR_{\ell_{0-1}}\paren*{h}-\sR_{\ell_{0-1}}^*\paren*{\ov \sH}+\sM_{\ell_{0-1}}\paren*{\ov \sH}
& \leq \Psi^{-1}
\paren*{\sR_{\ell_{\rm{gce}}}\paren*{h}
  - \sR_{\ell_{\rm{gce}}}^*\paren*{\ov \sH}+\sM_{\ell_{\rm{gce}}}\paren*{\ov \sH}},
\end{align*}
where $\ell_{\rm{gce}}
= \frac{1}{q}\bracket*{1-\paren*{\frac{e^{h(x,y)}}{\sum_{y'\in \sY}e^{h(x,y')}}}^q}$ and
\[\Psi(t)=\begin{cases}
 \frac{1}{q}\paren*{\frac{s_{\min}+s_{\max}}{2}}^q\bracket*{\paren*{\frac{\paren*{1 + t}^{\frac1{1-q }} + \paren*{1 - t}^{\frac1{1-q }}}{2}}^{1-q}-1}& t\leq \frac{s_{\max}^{1-q}-s_{\min}^{1-q}}{s_{\min}^{1-q}+s_{\max}^{1-q}}\\
\frac{t}{2q}\paren*{s_{\max}^q-s_{\min}^q}+\frac1q\paren*{\frac{s_{\min}^q+s_{\max}^q}{2}-\paren*{\frac{s_{\min}+s_{\max}}{2}}^q}& \mathrm{otherwise}.
\end{cases}.\]
\end{restatable}
\begin{proof}
For generalized cross-entropy loss $\ell_{\rm{gce}}$, plugging $\Phi(t)=\frac{1}{q}\paren*{1 - t^{q}}$ in Theorem~\ref{Thm:bound_comp_BD}, gives $\ov\sT^{\rm{comp}}$
\begin{equation*}
\begin{aligned}
\geq \inf\limits_{\psum\in \bracket*{\frac{1}{n-1}\vee t,1}}\inf\limits_{\substack{S_{\min}\leq \tau_2\leq \tau_1\leq S_{\max}\\ \tau_1+\tau_2\leq 1}}\sup\limits_{\mu\in \C} \curl*{\frac{\psum + t}{2}\bracket*{-\frac1q\paren*{\tau_2}^q+\frac1q\paren*{\tau_1-\mu}^q} + \frac{\psum-t}{2}\bracket*{ -\frac1q\paren*{\tau_1}^q+\frac1q\paren*{\tau_2+\mu}^q}}
\end{aligned}
\end{equation*}
where $\C=\bracket*{\max\curl*{s_{\min}-\tau_2,\tau_1- s_{\max}},\min\curl*{s_{\max}-\tau_2,\tau_1- s_{\min}}}$.
Here, we only compute the expression for $n>2$. The expression for $n=2$ will lead to the same result since it can be viewed as a special case of the expression for $n>2$.
By differentiating with respect to $\tau_2$ and $\psum$, we can see that the infimum is achieved when $\tau_1=\tau_2=\frac{s_{\min}+s_{\max}}{2}$ and $\psum=1$ modulo some elementary analysis. Thus, $\ov\sT^{\rm{comp}}$ can be reformulated as 
\begin{align*}
\ov\sT^{\rm{comp}}&=\sup\limits_{\mu\in \C} \bigg\{\frac{1 + t}{2q}\bracket*{-\paren*{\frac{s_{\min}+s_{\max}}{2}}^q+\paren*{\frac{s_{\min}+s_{\max}}{2}-\mu}^q}\\
&\qquad+ \frac{1-t}{2q}\bracket*{ -\paren*{\frac{s_{\min}+s_{\max}}{2}}^q+\paren*{\frac{s_{\min}+s_{\max}}{2}+\mu}^q}\bigg\}\\
& = -\frac{1}{q}\paren*{\frac{s_{\min}+s_{\max}}{2}}^q+\sup\limits_{\mu\in \C}g(\mu)
\end{align*}
where $\C=\bracket*{\frac{s_{\min}-s_{\max}}{2},\frac{s_{\max}-s_{\min}}{2}}$ and $g(\mu)=\frac{1 + t}{2q}\paren*{\frac{s_{\min}+s_{\max}}{2}-\mu}^q+\frac{1-t}{2q}\paren*{\frac{s_{\min}+s_{\max}}{2}+\mu}^q$. Since $g$ is continuous, it attains its supremum over a compact set.
Note that $g$ is concave and
differentiable.
In view of that, the maximum over
the open set $(-\infty, + \infty)$ can be obtained by
setting its gradient to zero. 
Differentiate $g(\mu)$ to optimize, we obtain
\begin{align*}
g(\mu^*)=0, \quad \mu^*=\frac{(1-t)^{\frac{1}{1-q}}-(1+t)^{\frac{1}{1-q}}}{(1+t)^{\frac{1}{1-q}}+(1-t)^{\frac{1}{1-q}}}\frac{s_{\min}+s_{\max}}{2}.   
\end{align*}
Moreover, by the concavity, $g(\mu)$ is non-increasing when $\mu\geq \mu^*$.
Since $s_{\max}-s_{\min}\geq 0$, we have
\begin{align*}
  \mu^*\leq 0\leq \frac{s_{\max}-s_{\min}}{2}
\end{align*}
In view of the constraint $C$, if $\mu^*\geq
\frac{s_{\min}-s_{\max}}{2}$, the maximum is achieved by
$\mu=\mu^*$. Otherwise, if $\mu^*< \frac{s_{\min}-s_{\max}}{2}$, since
$g(\mu)$ is non-increasing when $\mu\geq \mu^*$, the maximum is
achieved by $\mu=\frac{s_{\min}-s_{\max}}{2}$.  Since $\mu^*\geq
\frac{s_{\min}-s_{\max}}{2}$ is equivalent to $t\leq
\frac{s_{\max}^{1-q}-s_{\min}^{1-q}}{s_{\min}^{1-q}+s_{\max}^{1-q}}$,
the maximum can be expressed as
\begin{equation*}
\begin{aligned}
\max_{\mu \in C} g(\mu)
&=\begin{cases}
g(\mu^*) &   t\leq \frac{s_{\max}^{1-q}-s_{\min}^{1-q}}{s_{\min}^{1-q}+s_{\max}^{1-q}}\\
g\paren*{\frac{s_{\min}-s_{\max}}{2}} & \text{otherwise}
\end{cases}
\end{aligned}
\end{equation*}
Computing the value of $g$ at these points yields:
\begin{equation*}
\begin{aligned}
g(\mu^*) &= \frac{1}{q}\paren*{\frac{s_{\min}+s_{\max}}{2}}^q\paren*{\frac{\paren*{1 + t}^{\frac1{1-q }} +  \paren*{1 - t}^{\frac1{1-q }}}{2}}^{1-q }\\
g\paren*{\frac{s_{\min}-s_{\max}}{2}} &= \frac{1+t}{2q}(s_{\max})^q+\frac{1-t}{2q}(s_{\min})^q
\end{aligned}
\end{equation*}
Then, if $t\leq \frac{s_{\max}^{1-q}-s_{\min}^{1-q}}{s_{\min}^{1-q}+s_{\max}^{1-q}}$, we obtain
\begin{align*}
\ov\sT^{\rm{comp}}&= \frac{1}{q}\paren*{\frac{s_{\min}+s_{\max}}{2}}^q\paren*{\frac{\paren*{1 + t}^{\frac1{1-q }} + \paren*{1 - t}^{\frac1{1-q }}}{2}}^{1-q }-\frac{1}{q}\paren*{\frac{s_{\min}+s_{\max}}{2}}^q\\
&=\frac{1}{q}\paren*{\frac{s_{\min}+s_{\max}}{2}}^q\bracket*{\paren*{\frac{\paren*{1 + t}^{\frac1{1-q }} + \paren*{1 - t}^{\frac1{1-q }}}{2}}^{1-q}-1}
\end{align*}
Otherwise, we obtain
\begin{align*}
\ov\sT^{\rm{comp}}&=-\frac{1}{q}\paren*{\frac{s_{\min}+s_{\max}}{2}}^q+\frac{1+t}{2q}(s_{\max})^q+\frac{1-t}{2q}(s_{\min})^q\\
&= \frac{t}{2q}\paren*{s_{\max}^q-s_{\min}^q}+\frac1q\paren*{\frac{s_{\min}^q+s_{\max}^q}{2}-\paren*{\frac{s_{\min}+s_{\max}}{2}}^q}
\end{align*}
Since $\ov\sT^{\rm{comp}}$ is convex, by Theorem~\ref{Thm:bound_comp_BD}, for any $h\in \ov\sH$ and any distribution,
\begin{align*}
\sR_{\ell_{0-1}}\paren*{h}-\sR_{\ell_{0-1}}^*\paren*{\ov \sH}+\sM_{\ell_{0-1}}\paren*{\ov \sH}
& \leq \Psi^{-1}
\paren*{\sR_{\ell_{\rm{gce}}}\paren*{h}
  - \sR_{\ell_{\rm{gce}}}^*\paren*{\ov \sH}+\sM_{\ell_{\rm{gce}}}\paren*{\ov \sH}},
\end{align*}
where
\begin{align*}
\Psi(t)=\begin{cases}
 \frac{1}{q}\paren*{\frac{s_{\min}+s_{\max}}{2}}^q\bracket*{\paren*{\frac{\paren*{1 + t}^{\frac1{1-q }} + \paren*{1 - t}^{\frac1{1-q }}}{2}}^{1-q}-1}& t\leq \frac{s_{\max}^{1-q}-s_{\min}^{1-q}}{s_{\min}^{1-q}+s_{\max}^{1-q}}\\
\frac{t}{2q}\paren*{s_{\max}^q-s_{\min}^q}+\frac1q\paren*{\frac{s_{\min}^q+s_{\max}^q}{2}-\paren*{\frac{s_{\min}+s_{\max}}{2}}^q}& \mathrm{otherwise}.
\end{cases}
\end{align*}
\end{proof}

\subsection{Mean absolute error loss}
\label{app:bound-mae}

\begin{restatable}[\textbf{$\ov \sH$-consistency bounds for mean absolute error loss}]{theorem}{MAE}
\label{thm:bound-mae}
For any $h\in \ov \sH$ and any distribution, we have
\begin{align*}
\sR_{\ell_{0-1}}\paren*{h}-\sR_{\ell_{0-1}}^*\paren*{\ov \sH}+\sM_{\ell_{0-1}}\paren*{\ov \sH}
& \leq \frac{2\paren*{\sR_{\ell_{\rm{mae}}}\paren*{h}
    - \sR_{\ell_{\rm{mae}}}^*\paren*{\ov \sH}
    + \sM_{\ell_{\rm{mae}}}\paren*{\ov \sH}}}{s_{\max} - s_{\min}}.
\end{align*}
\end{restatable}
\begin{proof}
For mean absolute error loss $\ell_{\rm{mae}}$, plugging $\Phi(t)=1 - t$ in Theorem~\ref{Thm:bound_comp_BD}, gives $\ov\sT^{\rm{comp}}$
\begin{equation*}
\begin{aligned}
\geq \inf\limits_{\psum\in \bracket*{\frac{1}{n-1}\vee t,1}}\inf\limits_{\substack{S_{\min}\leq \tau_2\leq \tau_1\leq S_{\max}\\ \tau_1+\tau_2\leq 1}}\sup\limits_{\mu\in \C} \curl*{\frac{\psum + t}{2}\bracket*{-\paren*{\tau_2}+\paren*{\tau_1-\mu}} + \frac{\psum-t}{2}\bracket*{ -\paren*{\tau_1}+\paren*{\tau_2+\mu}}}
\end{aligned}
\end{equation*}
where $\C=\bracket*{\max\curl*{s_{\min}-\tau_2,\tau_1-
    s_{\max}},\min\curl*{s_{\max}-\tau_2,\tau_1- s_{\min}}}$.  Here,
we only compute the expression for $n>2$. The expression for $n=2$
will lead to the same result since it can be viewed as a special case
of the expression for $n>2$.  By differentiating with respect to
$\tau_2$ and $\psum$, we can see that the infimum is achieved when
$\tau_1=\tau_2=\frac{s_{\min}+s_{\max}}{2}$ and $\psum=1$ modulo some
elementary analysis. Thus, $\ov\sT^{\rm{comp}}$ can be reformulated as
\begin{align*}
\ov\sT^{\rm{comp}}&=\sup\limits_{\mu\in \C} \bigg\{\frac{1 + t}{2}\bracket*{-\paren*{\frac{s_{\min}+s_{\max}}{2}}+\paren*{\frac{s_{\min}+s_{\max}}{2}-\mu}}\\
&\qquad+ \frac{1-t}{2}\bracket*{ -\paren*{\frac{s_{\min}+s_{\max}}{2}}+\paren*{\frac{s_{\min}+s_{\max}}{2}+\mu}}\bigg\}\\
& =\sup\limits_{\mu\in \C}-t\mu
\end{align*}
where $\C=\bracket*{\frac{s_{\min}-s_{\max}}{2},\frac{s_{\max}-s_{\min}}{2}}$. Since $-t\mu$ is monotonically non-increasing, the maximum over
$C$ can be achieved by
\begin{align*}
\mu^*=\frac{s_{\min}-s_{\max}}{2}, \quad \ov\sT^{\rm{comp}}=\frac{s_{\max}-s_{\min}}{2}\,t.
\end{align*}
Since $\ov\sT^{\rm{comp}}$ is convex, by Theorem~\ref{Thm:bound_comp_BD}, for any $h\in \ov\sH$ and any distribution,
\begin{align*}
\sR_{\ell_{0-1}}\paren*{h}-\sR_{\ell_{0-1}}^*\paren*{\ov \sH}+\sM_{\ell_{0-1}}\paren*{\ov \sH}
& \leq \frac{2\paren*{\sR_{\ell_{\rm{mae}}}\paren*{h}
  - \sR_{\ell_{\rm{mae}}}^*\paren*{\ov \sH}+\sM_{\ell_{\rm{mae}}}\paren*{\ov \sH}}}{s_{\max}-s_{\min}}.
\end{align*}
\end{proof}

\section{Extensions of constrained losses}
\label{label:extension-cstnd}

\subsection{Proof of \texorpdfstring{$\ov\sH$}{H}-consistency bound
  with \texorpdfstring{$\ov \sT^{\rm{cstnd}}$}{T} (Theorem~\ref{Thm:bound_cstnd_BD})}
\label{app:bound_cstnd_BD}

\BoundCstndBD*
\begin{proof}
For the constrained loss $\ell^{\mathrm{cstnd}}$, the conditional
$\ell^{\mathrm{cstnd}}$-risk can be expressed as follows:
\begin{align*}
\sC_{\ell^{\mathrm{cstnd}}}(h,x)
& = \sum_{y\in \sY} \sfp(y \!\mid\! x) \ell^{\mathrm{cstnd}}(h, x, y)\\
&= \sum_{y\in \sY} \sfp(y \!\mid\! x) \sum_{y'\neq y}\Phi\paren*{-h(x, y')}\\
&= \sum_{y\in \sY} \Phi\paren*{-h(x, y)} \sum_{y'\neq y}\sfp(y' \!\mid\! x)\\
&= \sum_{y\in \sY} \Phi\paren*{-h(x, y)} \paren*{1-\sfp(y \!\mid\! x)}\\
& = \Phi\paren*{-h(x, y_{\max})} \paren*{1-\sfp(y_{\max} \!\mid\! x)} + \Phi\paren*{-h(x, \hh(x))} \paren*{1-\sfp(\hh(x) \!\mid\! x)}\\
& \qquad + \sum_{y\notin \curl*{y_{\max},\hh(x)}}\Phi\paren*{-h(x, y)} \paren*{1-\sfp(y \!\mid\! x)}.
\end{align*}
For any $h \in \ov\sH$ and $x\in \sX$, by the definition of $\ov\sH$, we can always find a family of hypotheses $\curl*{h_{\mu}}\subset \sH$ such that $h_{\mu}(x,\cdot)$ take the following values: 
\begin{align*}
h_{\mu}(x,y) = 
\begin{cases}
  h(x, y) & \text{if $y \not \in \curl*{y_{\max}, \hh(x)}$}\\
 h(x, y_{\max}) + \mu & \text{if $y = \hh(x)$}\\
  h(x,\hh(x)) -\mu & \text{if $y = y_{\max}$}.
\end{cases} 
\end{align*}
Note that the hypotheses $h_{\mu}$ satisfies the constraint:
\begin{align*}
 \sum_{y\in \sY} h_{\mu}(x, y)=\sum_{y\in \sY}h(x, y)=0,\, \forall \mu \in \mathbb{R}.
\end{align*}
Since $h_{\mu}(x,y)\in \bracket*{-\Lambda(x),\Lambda(x)}$, we have the following constraints on $\mu$:
\begin{equation*}
\begin{aligned}
&-\Lambda(x)-h(x, y_{\max})\leq \mu\leq \Lambda(x)-h(x, y_{\max})\\
&- \Lambda(x)+h(x,\hh(x))\leq \mu\leq \Lambda(x)+ h(x,\hh(x).
\end{aligned}
\end{equation*}
Let $p_1=\sfp(y_{\max} \!\mid\! x)$, $p_2=\sfp(\hh(x) \!\mid\! x)$, $\tau_1=h(x,\hh(x))$ and $\tau_2=h(x,y_{\max})$ to simplify the notation. 
Then, the constraint on $\mu$ can be expressed as 
\begin{equation*}
\mu \in \ov \C, \quad \ov \C=\bracket*{\max\curl*{\tau_1,-\tau_2}-\Lambda(x),\min\curl*{\tau_1,-\tau_2}+\Lambda(x)}
\end{equation*}
Since $\max\curl*{\tau_1,-\tau_2}-\min\curl*{\tau_1,-\tau_2}=\abs*{\tau_1+\tau_2}\leq \abs*{\tau_1}+\abs*{\tau_2}\leq 2\Lambda(x)$, $C$ is not an empty set. By the definition of $h_{\mu}$, we have for any $h\in \sH$ and $x\in \sX$,
\begin{align*}
&\sC_{\ell^{\mathrm{cstnd}}}(h,x) - \inf_{\mu \in \ov \C}\sC_{\ell^{\mathrm{cstnd}}}(h_{\mu},x)\\
& = \sup_{\mu\in \ov \C} \bigg\{\paren*{1-p_1}\bracket*{ \Phi\paren*{-\tau_2}-\Phi\paren*{-\tau_1+\mu}}+  \paren*{1-p_2}\bracket*{ \Phi\paren*{-\tau_1}-\Phi\paren*{-\tau_2-\mu}} \bigg\}\\
& = \sup_{\mu\in \ov \C} \bigg\{\frac{2-\psum-p_1+p_2}{2}\bracket*{ \Phi\paren*{-\tau_2}-\Phi\paren*{-\tau_1+\mu}}+  \frac{2-\psum+p_1-p_2}{2}\bracket*{ \Phi\paren*{-\tau_1}-\Phi\paren*{-\tau_2-\mu}} \bigg\}\tag{$\psum=p_1+p_2\in \bracket*{\frac{1}{n-1},1}$}\\
& = \inf_{\psum\in \bracket*{\frac{1}{n-1},1}}\inf_{\substack{\tau_1\geq \max\curl*{\tau_2,0}}}\sup_{\mu\in \ov \C} \bigg\{\frac{2-\psum-p_1+p_2}{2}\bracket*{ \Phi\paren*{-\tau_2}-\Phi\paren*{-\tau_1+\mu}} \\
& \qquad +  \frac{2-\psum+p_1-p_2}{2}\bracket*{ \Phi\paren*{-\tau_1}-\Phi\paren*{-\tau_2-\mu}} \bigg\}\tag{$\tau_1\geq 0$, $\tau_2\leq \tau_1$}\\
& \geq \inf_{\psum\in \bracket*{\frac{1}{n-1},1}}\inf_{\substack{\tau_1\geq \max\curl*{\tau_2,0}}}\sup_{\mu\in \C} \bigg\{\frac{2-\psum-p_1+p_2}{2}\bracket*{ \Phi\paren*{-\tau_2}-\Phi\paren*{-\tau_1+\mu}} \\
& \qquad +  \frac{2-\psum+p_1-p_2}{2}\bracket*{ \Phi\paren*{-\tau_1}-\Phi\paren*{-\tau_2-\mu}} \bigg\}\tag{$\C=\bracket*{\max\curl*{\tau_1,-\tau_2}-\Lambda_{\rm{min}},\min\curl*{\tau_1,-\tau_2}+\Lambda_{\rm{min}}}\subset \ov\C$ since $\Lambda_{\rm{min}}\leq \Lambda(x)$}\\
& = \inf_{\psum\in \bracket*{\frac{1}{n-1},1}}\inf_{\substack{\tau_1\geq \max\curl*{\tau_2,0}}} \bigg\{\frac{2-\psum-p_1+p_2}{2} \Phi\paren*{-\tau_2}+\frac{2-\psum+p_1-p_2}{2} \Phi\paren*{-\tau_1} \\
& \qquad -\inf_{\mu\in \C}\curl*{\frac{2-\psum-p_1+p_2}{2}\Phi\paren*{-\tau_1+\mu}+\frac{2-\psum+p_1-p_2}{2} \Phi\paren*{-\tau_2-\mu}}  \bigg\}\\
& = \sT^{\rm{cstnd}}\paren*{p_1-p_2}\\
& = \sT^{\rm{cstnd}}\paren*{\Delta\sC_{\ell_{0-1},\sH}(h,x)} \tag{by Lemma~\ref{lemma:explicit_assumption_01-chcb}}.
\end{align*}
Note that for $n=2$, an additional constraint $\tau_1+\tau_2=1$ is imposed and the expression can be simplified as
\begin{align*}
&\sC_{\ell^{\mathrm{cstnd}}}(h,x) - \inf_{\mu \in \ov \C}\sC_{\ell^{\mathrm{cstnd}}}(h_{\mu},x) \\
&\geq\inf\limits_{\tau\geq 0}\sup\limits_{\mu\in \bracket*{\tau-\Lambda_{\rm{min}},\tau+\Lambda_{\rm{min}}}} \curl*{\frac{1 - p_1+p_2}{2}\bracket*{ \Phi\paren*{\tau}-\Phi\paren*{-\tau+\mu}} +  \frac{1 + p_1-p_2}{2}\bracket*{ \Phi\paren*{-\tau}-\Phi\paren*{\tau-\mu}}}\\
& = \sT^{\rm{cstnd}}\paren*{p_1-p_2}\\
& = \sT^{\rm{cstnd}}\paren*{\Delta\sC_{\ell_{0-1},\sH}(h,x)} \tag{by Lemma~\ref{lemma:explicit_assumption_01-chcb}}.
\end{align*}
Since $\sT^{\rm{cstnd}}$ is convex, by Jensen's inequality, we obtain for any hypothesis $h \in \sH$ and any distribution,
\begin{align*}
&\sT^{\rm{cstnd}}\paren*{\sR_{\ell_{0-1}}(h)-\sR_{\ell_{0-1}}^*(\sH)+\sM_{\ell_{0-1}}(\sH)}\\
&=\sT^{\rm{cstnd}}\paren*{\E_{X}\bracket*{\Delta\sC_{\ell_{0-1},\sH}(h,x)}}\\
&\leq \E_{X}\bracket*{\sT^{\rm{cstnd}}\paren*{\Delta\sC_{\ell_{0-1},\sH}(h,x)}}\\
& \leq \E_{X}\bracket*{\Delta\sC_{\ell^{\mathrm{cstnd}},\sH}(h,x)}\\
& =\sR_{\ell^{\mathrm{cstnd}}}(h)-\sR_{\ell^{\mathrm{cstnd}}}^*(\sH)+\sM_{\ell^{\mathrm{cstnd}}}(\sH).
\end{align*}
Let $n=2$. For any $t\in [0,1]$, we consider the distribution that
concentrates on a singleton $\curl*{x}$ and satisfies
$\sfp(1 \!\mid\! x)=\frac{1+t}{2}$, $\sfp(2 \!\mid\! x)=\frac{1-t}{2}$. For any $\e>0$, by the
definition of infimum, we can take $h\in \sH$ such that
$h(x,2)=\tau_{\e}\geq 0$ and satisfies
\begin{align*}
\sup\limits_{\mu\in \bracket*{\tau_{\e}-\Lambda_{\rm{min}},\tau_{\e}+\Lambda_{\rm{min}}}} \curl*{\frac{1 - t}{2}\bracket*{ \Phi\paren*{\tau_{\e}}-\Phi\paren*{-\tau_{\e}+\mu}} +  \frac{1 + t}{2}\bracket*{ \Phi\paren*{-\tau_{\e}}-\Phi\paren*{\tau_{\e}-\mu}}}<\sT^{\rm{cstnd}}(t)+\e.
\end{align*}
Then,
\begin{align*}
\sR_{\ell_{0-1}}(h)-
\sR^*_{\ell_{0-1}}(\sH)+\sM_{\ell_{0-1}}(\sH)
&=  \sR_{\ell_{0-1}}(h) - \mathbb{E}_{X} \bracket* {\sC^*_{\ell_{0-1}}(\sH,x)}\\
&=\sC_{\ell_{0-1}}(h,x) - \sC^*_{\ell_{0-1}}\paren*{\sH,x}\\
&=t
\end{align*}
and
\begin{align*}
\sT^{\rm{cstnd}}(t)
&\leq \sR_{\ell^{\mathrm{cstnd}}}(h) - \sR_{\ell^{\mathrm{cstnd}}}^*(\sH) +
\sM_{\ell^{\mathrm{cstnd}}}(\sH)\\
&=  \sR_{\ell^{\mathrm{cstnd}}}(h) - \mathbb{E}_{X} \bracket* {\sC^*_{\ell^{\mathrm{cstnd}}}(\sH,x)}\\
&=\sC_{\ell^{\mathrm{cstnd}}}(h,x)-\sC^*_{\ell^{\mathrm{cstnd}}}(\sH,x)\\
&=\sup\limits_{\mu\in \bracket*{\tau_{\e}-\Lambda_{\rm{min}},\tau_{\e}+\Lambda_{\rm{min}}}} \curl*{\frac{1 - t}{2}\bracket*{ \Phi\paren*{\tau_{\e}}-\Phi\paren*{-\tau_{\e}+\mu}} +  \frac{1 + t}{2}\bracket*{ \Phi\paren*{-\tau_{\e}}-\Phi\paren*{\tau_{\e}-\mu}}}\\
&<\sT^{\rm{cstnd}}(t)+\e.
\end{align*}
By letting $\e\to 0$, we conclude the proof.  The proof for $n > 2$
directly extends from the case when $n = 2$. Indeed, for any $t\in
[0,1]$, we consider the distribution that concentrates on a singleton
$\curl*{x}$ and satisfies $\sfp(1 \!\mid\! x)=\frac{1+t}{2}$,
$\sfp(2 \!\mid\! x)=\frac{1-t}{2}$, $\sfp(y \!\mid\! x)=0,3\leq y\leq n$. For any $\e>0$, by
the definition of infimum, we can take $h\in \sH$ such that
$h(x,1)=\tau_{1,\e}$, $h(x,2)=\tau_{2,\e}$, $h(x,3)=0, \, 3\leq y\leq
n$ and satisfies $\tau_{1,\e} + \tau_{2,\e} = 0$, and
\begin{align*}
&\inf\limits_{\psum\in \bracket*{\frac{1}{n-1},1}}\sup\limits_{\mu\in \C} \curl*{\frac{2-\psum-t}{2}\bracket*{ \Phi\paren*{-\tau_{2,\e}}-\Phi\paren*{-\tau_{1,\e}+\mu}} +  \frac{2-\psum + t}{2}\bracket*{ \Phi\paren*{-\tau_{1,\e}}-\Phi\paren*{-\tau_{2,\e}-\mu}}}\\
&=\sup\limits_{\mu\in \C} \curl*{\frac{1-t}{2}\bracket*{ \Phi\paren*{-\tau_{2,\e}}-\Phi\paren*{-\tau_{1,\e}+\mu}} +  \frac{1 + t}{2}\bracket*{ \Phi\paren*{-\tau_{1,\e}}-\Phi\paren*{-\tau_{2,\e}-\mu}}}\\
&<\sT^{\rm{cstnd}}(t)+\e.
\end{align*}
Then,
\begin{align*}
\sR_{\ell_{0-1}}(h)-
\sR^*_{\ell_{0-1}}(\sH)+\sM_{\ell_{0-1}}(\sH)=t
\end{align*}
and
\begin{align*}
\sT^{\rm{cstnd}}(t)
\leq \sR_{\ell^{\mathrm{cstnd}}}(h) - \sR_{\ell^{\mathrm{cstnd}}}^*(\sH) +
\sM_{\ell^{\mathrm{cstnd}}}(\sH)<\sT^{\rm{cstnd}}(t)+\e.
\end{align*}
By letting $\e\to 0$, we conclude the proof.
\end{proof}

\subsection{Constrained exponential loss}
\label{app:bound-exponential-cstnd}

\ExponentialCstnd*
\begin{proof}
For $n = 2$,  plugging in $\Phi(t)=e^{-t}$ in Theorem~\ref{Thm:bound_cstnd_BD}, gives
\begin{align*}
\ov\sT^{\rm{cstnd}}(t)
&= \inf\limits_{\tau\geq 0}\sup\limits_{\mu\in \bracket*{\tau-\Lambda_{\rm{min}},\tau+\Lambda_{\rm{min}}}} \curl*{\frac{1 - t}{2}\bracket*{ e^{-\tau}-e^{\tau-\mu}} +  \frac{1 + t}{2}\bracket*{ e^{\tau}-e^{-\tau+\mu}}}.
\end{align*}
By differentiating with respect to $\tau$, we can see that the infimum is achieved when $\tau=0$ modulo some elementary analysis.
Thus, $\ov\sT^{\rm{cstnd}}$ can be reformulated as 
\begin{align*}
\ov\sT^{\rm{cstnd}}&=\sup\limits_{\mu\in \bracket*{-\Lambda_{\rm{min}},\Lambda_{\rm{min}}}} \curl*{\frac{1 - t}{2}\bracket*{ 1-e^{-\mu}} +  \frac{1 + t}{2}\bracket*{ 1-e^{\mu}}}\\
&=1+\sup\limits_{\mu\in \bracket*{-\Lambda_{\rm{min}},\Lambda_{\rm{min}}}}g(\mu).
\end{align*}
where $g(\mu)=-\frac{1 - t}{2}e^{-\mu}- \frac{1 + t}{2}e^{\mu}$. Since $g$ is continuous, it attains its supremum over a compact set.
Note that $g$ is concave and
differentiable.
In view of that, the maximum over
the open set $(-\infty, + \infty)$ can be obtained by
setting its gradient to zero. 
Differentiate $g(\mu)$ to optimize, we obtain
\begin{align*}
g(\mu^*)=0, \quad \mu^*=\frac{1}{2}\log\frac{1-t}{1+t}
\end{align*}
Moreover, by the concavity, $g(\mu)$ is non-increasing when $\mu\geq \mu^*$.
Since $\mu^*\leq 0$ and $\Lambda_{\rm{min}}\geq 0$, we have
\begin{align*}
  \mu^*\leq 0\leq \Lambda_{\rm{min}}
\end{align*}
In view of the constraint, if $\mu^*\geq -\Lambda_{\rm{min}}$, the maximum is achieved by $\mu=\mu^*$. Otherwise, if $\mu^*< -\Lambda_{\rm{min}}$, since $g(\mu)$ is non-increasing when $\mu\geq \mu^*$, the maximum is achieved by $\mu=-\Lambda_{\rm{min}}$.
Since $\mu^*\geq -\Lambda_{\rm{min}}$ is equivalent to $t\leq \frac{e^{2\Lambda_{\rm{min}}}-1}{e^{2\Lambda_{\rm{min}}}+1}$, the maximum can be expressed as
\begin{equation*}
\begin{aligned}
\max_{\mu\in \bracket*{-\Lambda_{\rm{min}},\Lambda_{\rm{min}}}} g(\mu)
&=\begin{cases}
g(\mu^*) &  t\leq \frac{e^{2\Lambda_{\rm{min}}}-1}{e^{2\Lambda_{\rm{min}}}+1}\\
g\paren*{-\Lambda_{\rm{min}}} & \text{otherwise}
\end{cases}
\end{aligned}
\end{equation*}
Computing the value of $g$ at these points yields:
\begin{equation*}
\begin{aligned}
g(\mu^*) &= -\sqrt{1-t^2}\\
g\paren*{-\Lambda_{\rm{min}}} &= -\frac{1 - t}{2}e^{\Lambda_{\rm{min}}}- \frac{1 + t}{2}e^{-\Lambda_{\rm{min}}}.
\end{aligned}
\end{equation*}
Then, if $t\leq \frac{e^{2\Lambda_{\rm{min}}}-1}{e^{2\Lambda_{\rm{min}}}+1}$, we obtain
\begin{align*}
\ov\sT^{\rm{cstnd}}&= 1-\sqrt{1-t^2}.
\end{align*}
Otherwise, we obtain
\begin{align*}
\ov\sT^{\rm{cstnd}}&=1-\frac{1 - t}{2}e^{\Lambda_{\rm{min}}}- \frac{1 + t}{2}e^{-\Lambda_{\rm{min}}}\\
&= \frac{t}{2}\paren*{e^{\Lambda_{\rm{min}}}-e^{-\Lambda_{\rm{min}}}}+\frac{2-e^{\Lambda_{\rm{min}}}-e^{-\Lambda_{\rm{min}}}}{2}.
\end{align*}
For $n > 2$,  plugging in $\Phi(t)=e^{-t}$ in Theorem~\ref{Thm:bound_cstnd_BD}, gives
\begin{align*}
\ov\sT^{\rm{cstnd}}(t)
&= \inf\limits_{\psum\in \bracket*{\frac{1}{n-1},1}}\inf\limits_{\tau_1\geq \max\curl*{\tau_2,0}}\sup\limits_{\mu\in \C} \curl*{\frac{2-\psum-t}{2}\bracket*{ e^{\tau_2}-e^{\tau_1-\mu}} +  \frac{2-\psum + t}{2}\bracket*{ e^{\tau_1}-e^{\tau_2+\mu}}}.
\end{align*}
where $\C=\bracket*{\max\curl*{\tau_1,-\tau_2}-\Lambda_{\rm{min}},\min\curl*{\tau_1,-\tau_2}+\Lambda_{\rm{min}}}$.
By differentiating with respect to $\tau_2$ and $P$, we can see that the infimum is achieved when $\tau_2=\tau_1=0$ and $P=1$ modulo some elementary analysis.
Thus, $\ov\sT^{\rm{cstnd}}$ can be reformulated as 
\begin{align*}
\ov\sT^{\rm{cstnd}}&=\sup\limits_{\mu\in \C} \curl*{\frac{1-t}{2}\bracket*{ 1-e^{-\mu}} +  \frac{1 + t}{2}\bracket*{ 1-e^{\mu}}}\\
&=1+\sup\limits_{\mu\in \C}g(\mu).
\end{align*}
where $\C=\bracket*{-\Lambda_{\rm{min}},\Lambda_{\rm{min}}}$ and
$g(\mu)=-\frac{1 - t}{2}e^{-\mu}- \frac{1 + t}{2}e^{\mu}$. Since $g$
is continuous, it attains its supremum over a compact set.  Note that
$g$ is concave and differentiable.  In view of that, the maximum over
the open set $(-\infty, + \infty)$ can be obtained by setting its
gradient to zero.  Differentiate $g(\mu)$ to optimize, we obtain
\begin{align*}
g(\mu^*)=0, \quad \mu^*=\frac{1}{2}\log\frac{1-t}{1+t}
\end{align*}
Moreover, by the concavity, $g(\mu)$ is non-increasing when $\mu\geq \mu^*$.
Since $\mu^*\leq 0$ and $\Lambda_{\rm{min}}\geq 0$, we have
\begin{align*}
  \mu^*\leq 0\leq \Lambda_{\rm{min}}
\end{align*}
In view of the constraint, if $\mu^*\geq -\Lambda_{\rm{min}}$, the
maximum is achieved by $\mu=\mu^*$. Otherwise, if $\mu^*<
-\Lambda_{\rm{min}}$, since $g(\mu)$ is non-increasing when $\mu\geq
\mu^*$, the maximum is achieved by $\mu=-\Lambda_{\rm{min}}$.  Since
$\mu^*\geq -\Lambda_{\rm{min}}$ is equivalent to $t\leq
\frac{e^{2\Lambda_{\rm{min}}}-1}{e^{2\Lambda_{\rm{min}}}+1}$, the
maximum can be expressed as
\begin{equation*}
\begin{aligned}
\max_{\mu\in \bracket*{-\Lambda_{\rm{min}},\Lambda_{\rm{min}}}} g(\mu)
&=\begin{cases}
g(\mu^*) &  t\leq \frac{e^{2\Lambda_{\rm{min}}}-1}{e^{2\Lambda_{\rm{min}}}+1}\\
g\paren*{-\Lambda_{\rm{min}}} & \text{otherwise}
\end{cases}
\end{aligned}
\end{equation*}
Computing the value of $g$ at these points yields:
\begin{equation*}
\begin{aligned}
g(\mu^*) &= -\sqrt{1-t^2}\\
g\paren*{-\Lambda_{\rm{min}}} &= -\frac{1 - t}{2}e^{\Lambda_{\rm{min}}}- \frac{1 + t}{2}e^{-\Lambda_{\rm{min}}}.
\end{aligned}
\end{equation*}
Then, if $t\leq \frac{e^{2\Lambda_{\rm{min}}}-1}{e^{2\Lambda_{\rm{min}}}+1}$, we obtain
\begin{align*}
\ov\sT^{\rm{cstnd}}&= 1-\sqrt{1-t^2}.
\end{align*}
Otherwise, we obtain
\begin{align*}
\ov\sT^{\rm{cstnd}}&=1-\frac{1 - t}{2}e^{\Lambda_{\rm{min}}}- \frac{1 + t}{2}e^{-\Lambda_{\rm{min}}}\\
&= \frac{t}{2}\paren*{e^{\Lambda_{\rm{min}}}-e^{-\Lambda_{\rm{min}}}}+\frac{2-e^{\Lambda_{\rm{min}}}-e^{-\Lambda_{\rm{min}}}}{2}.
\end{align*}
Since $\ov\sT^{\rm{cstnd}}$ is convex, by Theorem~\ref{Thm:bound_cstnd_BD}, for any $h\in \ov\sH$ and any distribution,
\begin{align*}
\sR_{\ell_{0-1}}\paren*{h}-\sR_{\ell_{0-1}}^*\paren*{\ov \sH}+\sM_{\ell_{0-1}}\paren*{\ov \sH}
& \leq \Psi^{-1}
\paren*{\sR_{\ell^{\rm{cstnd}}}\paren*{h}-\sR_{\ell^{\rm{cstnd}}}^*\paren*{\ov \sH}+\sM_{\ell^{\rm{cstnd}}}\paren*{\ov \sH}}
\end{align*}
where
\begin{align*}
\Psi(t)=\begin{cases}
 1-\sqrt{1-t^2} & t\leq \frac{e^{2\Lambda_{\rm{min}}}-1}{e^{2\Lambda_{\rm{min}}}+1}\\
\frac{t}{2}\paren*{e^{\Lambda_{\rm{min}}}-e^{-\Lambda_{\rm{min}}}}+\frac{2-e^{\Lambda_{\rm{min}}}-e^{-\Lambda_{\rm{min}}}}{2}& \mathrm{otherwise}.
\end{cases}
\end{align*}
\end{proof}
\restoreatoc

\chapter{Appendix to Chapter~\ref{ch6}}

\disableatoc
\section{Pointwise loss functions -
  Proof of Lemma~\ref{lemma:approximation-error}}
\label{app:lemma}

\ApproximationError*
\begin{proof}
By definition, for a pointwise loss function $\ell$, there exists a
measurable function $\hat \ell \colon \Rset^n \times \sY \to
\Rset_{+}$ such that $\ell(h, x, y) = \hat \ell(h(x), y)$, where $h(x)
= \bracket*{h(x, 1), \ldots, h(x, n)}$ is the score vector of the
predictor $h$. Thus, the following inequality holds:
\begin{equation*}
\sC^*_{\ell}\paren*{\sH_{\rm{all}}, x} = \inf_{h \in \sH} \E_y \bracket*{\ell(h, x, y) \mid x} = \inf_{\alpha \in \Rset^n}  \E_{y} \bracket*{\hat \ell(\alpha, y) \mid x}.
\end{equation*}
Since $\hat \ell \colon (\alpha, y) \mapsto \Rset_{+}$ is measurable, the function $(\alpha, x) \mapsto \E_{y} \bracket*{\hat \ell(\alpha, y) \mid x}$ is also measurable. 
We now show that the function $x \mapsto \sC^*_{\ell}\paren*{\sH_{\rm{all}}, x} = \inf_{\alpha \in \Rset}  \E_{y} \bracket*{\hat \ell(\alpha, y) \mid x}$ is also measurable. 

To do this, we consider for any $\beta > 0$, the set $\curl*{x \colon \inf_{\alpha \in \Rset}  \E_{y} \bracket*{\hat \ell(\alpha, y) \mid x} < \beta}$ which can be expressed as 
\begin{align*}
 \curl*{x \colon \inf_{\alpha \in \Rset^n}  \E_{y} \bracket*{\hat \ell(\alpha, y) \mid x} < \beta} 
 & = \curl*{x \colon \exists \alpha \in \Rset^n \text{ such that } \E_{y} \bracket*{\hat \ell(\alpha, y) \mid x} < \beta }\\
 & = \Pi_{\sX} \curl*{(\alpha, x)\colon \E_{y} \bracket*{\hat \ell(\alpha, y) \mid x} < \beta},
\end{align*}
where $\Pi_{\sX}$ is projection onto $\sX$. By measurable
projection theorem, $x \mapsto \inf_{\alpha \in \Rset^n} \E_{y}
\bracket*{\hat \ell(\alpha, y) \mid x}$ is measurable. Then, since a
pointwise difference of measurable functions is measurable, for all $n
\in \Nset$, the set $\curl*{(\alpha, x) \colon \E_{y} \bracket*{\hat
    \ell(\alpha, y) \mid x} < \inf_{\alpha \in \Rset^n} \E_{y}
  \bracket*{\hat \ell(\alpha, y) \mid x} + \frac1n}$ is
measurable. Thus, by \citet{KuratowskiRyllNardzewski1965}'s measurable
selection theorem, for all $n \in \Nset$, there exists a measurable
function $h_{n} \colon x \mapsto \alpha \in \Rset^n$ such that the
following holds:
\begin{equation*}
\sC_{\ell}\paren*{h_n, x}  = \E_{y} \bracket*{\hat \ell(\alpha, y) \mid x} < \inf_{\alpha \in \Rset^n}  \E_{y} \bracket*{\hat \ell(\alpha, y) \mid x} + \frac1n =  \sC^*_{\ell}\paren*{\sH_{\rm{all}}, x} + \frac1n.
\end{equation*}
Therefore, we have
\begin{equation*}
\sE^*_{\ell}(\sH_{\rm{all}}) \leq \E_{x}\bracket*{\sC_{\ell}\paren*{h_n, x}}\leq \E_{x} \bracket*{\sC^*_{\ell}\paren*{\sH_{\rm{all}}, x}} + \frac1n \leq \sE^*_{\ell}(\sH_{\rm{all}}) + \frac1n.
\end{equation*}
By taking the limit $n \to + \infty$, we obtain $\sE^*_{\ell}(\sH_{\rm{all}})  = \E_{x} \bracket*{\sC^*_{\ell}\paren*{\sH_{\rm{all}}, x}}$. By definition, $\sA_{\ell}(\sH) = \sE_{\ell}^*(\sH) -
  \sE_{\ell}^*\paren*{\sH_{\mathrm{all}}} = \sE_{\ell}^*(\sH) - \E_{x} \bracket*{\sC^*_{\ell}\paren*{\sH_{\rm{all}}, x}}$. This completes the proof.
\end{proof}

\section{General form of \texorpdfstring{$\sH$}{H}-consistency bounds}
\label{app:explicit-form}

Fix a target loss function $\ell_2$ and a surrogate loss $\ell_1$.
Given a hypothesis set $\sH$, a bound relating the
estimation errors of these loss functions admits the following form:
\begin{align}
\label{eq:bound}
\forall h \in \sH, \quad
	\sE_{\ell_2}(h) - \sE^*_{\ell_2}(\sH)
    	\leq \Gamma_{\sD} \paren*{\sE_{\ell_1}(h) - \sE^*_{\ell_1}(\sH)},
\end{align}
where, for any distribution $\sD$, $\Gamma_{\sD}\colon \Rset_+ \to
\Rset_+$ is a non-decreasing function on $\Rset_+$. We will assume
that $\Gamma_{\sD}$ is concave.
In particular, the bound should hold for any point mass distribution
$\delta_x$, $x \in \sX$. We will operate under the assumption that the
same bound holds uniformly over $\sX$ and thus that there exists a
fixed concave function $\Gamma$ such that $\Gamma_{\delta_x} = \Gamma$ for
all $x$.

Observe that for any point mass distribution $\delta_x$, the
conditional loss and the expected loss coincide and therefore that we
have $\sE_{\ell_2}(h) - \sE^*_{\ell_2}(\sH) = \Delta \sC_{\ell_2, \sH}(\sH,
x)$, and similarly with $\ell_1$. Thus, we can write:
\[
\forall h \in \sH, \forall x \in \sX, \quad
	\Delta \sC_{\ell_2, \sH}(h, x)
	\leq \Gamma\paren*{\Delta \sC_{\ell_1, \sH}(h, x)}.
\]
Therefore, by Jensen's inequality, for any distribution $\sD$, we have
\[
\forall h \in \sH, \forall x \in \sX, \quad
	\E_{x}\bracket*{\Delta \sC_{\ell_2, \sH}(h, x)}
	\leq \E_{x}\bracket*{\Gamma\paren*{\Delta	\sC_{\ell_1, \sH}(h, x)}}
	\leq \Gamma\paren*{\E_{x}\bracket*{\Delta	\sC_{\ell_1, \sH}(h, x)}}.
\]
Since $\E_{x}\bracket*{\Delta \sC_{\ell_2, \sH}(h, x)} = \sE_{\ell_2}(h) -
\sE^*_{\ell_2}(\sH) + \sM_{\ell_2}(\sH)$ and similarly with $\ell_1$,
we obtain the following bound for all distributions $\sD$:
\begin{equation}
\label{eq:H-consistency-bound-srd}
\forall h \in \sH, \quad
\sE_{\ell_2}(h) - \sE^*_{\ell_2}(\sH) + \sM_{\ell_2}(\sH)
\leq \Gamma\paren*{\sE_{\ell_1}(h)-\sE^*_{\ell_1}(\sH) + \sM_{\ell_1}(\sH)}.
\end{equation}
This leads to the general form of $\sH$-consistency bounds that we will be 
considering, which includes the key role of the minimizability gaps.

\section{Properties of minimizability gaps}
\label{app:properties}

By Lemma~\ref{lemma:approximation-error}, for a pointwise loss
function, we have $\sE^*_\ell\paren*{\sH_{\rm{all}}} =
\E_{x}\bracket*{\sC_{\ell}^*(\sH_{\rm{all}},x)}$, thus the
minimizability gap vanishes for the family of all measurable
functions.

\begin{lemma}
\label{lemma:zero-minimizability-srd}
  Let $\ell$ be a pointwise loss function. Then, we have
  $\sM_\ell(\sH_{\rm{all}}) = 0$.
\end{lemma}
\ignore{
\begin{proof}
By definition, $\sM_{\ell}(\sH_{\rm{all}}) =
\sE^*_\ell\paren*{\sH_{\rm{all}}} -
\E_{x}\bracket*{\sC_{\ell}^*(\sH_{\rm{all}},x)}$. By
lemma~\ref{lemma:approximation-error},
$\sE^*_\ell\paren*{\sH_{\rm{all}}} =
\E_{x}\bracket*{\sC_{\ell}^*(\sH_{\rm{all}},x)}$, which
implies that $\sM_{\ell}(\sH_{\rm{all}})=0$.
\end{proof}
}
Thus, in that case, \eqref{eq:H-consistency-bound-srd} takes the following
simpler form:
\begin{equation}
\forall h \in \sH, \quad
\sE_{\ell_2}(h) - \sE^*_{\ell_2}(\sH_{\rm{all}})
\leq \Gamma \paren*{\sE_{\ell_1}(h) - \sE^*_{\ell_1}(\sH_{\rm{all}})}.
\end{equation}
In general, however, the minimizabiliy gap is non-zero for a
restricted hypothesis set $\sH$ and is therefore important to analyze.
Let $\sI_\ell(\sH)$ be the difference of pointwise infima
$\sI_\ell(\sH) = \E_x \bracket[big]{\sC^*_\ell(\sH, x) -
  \sC^*_\ell(\sH_{\rm{all}}, x)}$, which is non-negative. Note that,
for a pointwise loss function, the minimizability gap can be
decomposed as follows in terms of the approximation error and the
difference of pointwise infima:
\begin{align*}
  \sM_\ell(\sH)
  & = \sE^*_\ell(\sH) - \sE^*_\ell\paren*{\sH_{\rm{all}}}
  + \sE^*_\ell\paren*{\sH_{\rm{all}}} - \E_x \bracket*{\sC^*_\ell(\sH, x)}\\
  & = \sA_\ell(\sH) + \sE^*_\ell\paren*{\sH_{\rm{all}}}
  - \E_x \bracket*{\sC^*_\ell(\sH, x)}\\
  & = \sA_\ell(\sH) - \sI_\ell(\sH)
  \leq \sA_\ell(\sH).
\end{align*}
Thus, the minimizabiliy gap can be upper-bounded by the approximation
error. It is however a finer quantity than the approximation error and
can lead to more favorable guarantees. When the difference of
pointwise infima can be evaluated or bounded, this decomposition can
provide a convenient way to analyze the minimizability gap in terms of
the approximation error.

Note that $\sI_\ell(\sH)$ can be non-zero for families of bounded
functions.  Let $\sY = \curl*{-1, +1}$ and $\sH$ be a family of functions $h$ with
$\abs*{h(x)} \leq \Lambda$ for all $x \in \sX$ and such that all
values in $[-\Lambda, +\Lambda]$ can be reached.  Consider for example
the exponential-based margin loss: $\ell(h, x, y) = e^{-yh(x)}$. Let
$\eta(x) = \sfp(+1 \!\mid\! x) = \sD(Y = + 1\!\mid\! X = x)$.  Thus,
$\sC_{\ell}(h, x) = \eta(x) e^{-h(x)} + (1 - \eta(x)) e^{h(x)}$. Then,
it is not hard to see that $\sC^*_{\ell}(\sH_{\rm{all}}, x) =
2\sqrt{\eta(x)(1 - \eta(x))}$ for all $x$ but $\sC^*_{\ell}(\sH, x)$
depends on $\Lambda$ with the minimizing value for $h(x)$ being: $\min
\curl*{\frac{1}{2} \log \frac{\eta(x)}{1 - \eta(x)}, \Lambda}$ if
$\eta(x) \geq 1/2$, $\max \curl*{\frac{1}{2} \log \frac{\eta(x)}{1 -
    \eta(x)}, -\Lambda}$ otherwise. Thus, in the deterministic case,
$\sI_\ell(\sH) = e^{-\Lambda}$.

When the best-in-class error coincides with the Bayes error,
$\sE^*_\ell\paren*{\sH} = \sE^*_\ell\paren*{\sH_{\rm{all}}}$, both the
approximation error and minimizability gaps vanish.
\begin{lemma}
\label{lemma:vanish-minimizability}
For any loss function $\ell$ such that $\sE^*_\ell\paren*{\sH} =
\sE^*_\ell\paren*{\sH_{\rm{all}}} =
\E_{x}\bracket*{\sC_{\ell}^*(\sH_{\rm{all}},x)}$, we have
$\sM_\ell(\sH) = \sA_\ell(\sH) = 0$.
\end{lemma}
\begin{proof}
By definition, $\sA_{\ell}(\sH) = \sE^*_\ell(\sH) -
\sE^*_\ell\paren*{\sH_{\rm{all}}} = 0$. Since we have $\sM_\ell(\sH) \leq
\sA_\ell(\sH)$, this implies  $\sM_{\ell}(\sH)=0$.
\end{proof}

\section{Examples of \texorpdfstring{$\sH$}{H}-consistency bounds}
\label{app:bounds-example}

Here, we compile some common examples of $\sH$-consistency bounds
for both binary and multi-class classification.
Table~\ref{tab:bounds-binary}, \ref{tab:bounds-comp} and
\ref{tab:bounds-cstnd} include the examples of $\sH$-consistency
bounds for binary margin-based losses, comp-sum losses and constrained
losses, respectively.

These bounds are due to previous work by \citet{awasthi2022Hconsistency} for
binary margin-based losses, by \citet{mao2023cross} for multi-class
comp-sum losses, and by \citet{awasthi2022multi} and
\citet{MaoMohriZhong2023characterization} for multi-class constrained
losses, respectively. We consider complete hypothesis sets for binary
classification (see Section~\ref{sec:binary}), and symmetric and
complete hypothesis sets for multi-class classification (see
Section~\ref{sec:multi}).

\begin{table}[t]
  \centering
   \resizebox{\textwidth}{!}{
  \begin{tabular}{@{\hspace{0cm}}lll@{\hspace{0cm}}}
    \toprule
    $\Phi(u)$  & margin-based losses $\ell$   & $\sH$-Consistency bounds\\
    \midrule
    $e^{u}$ & $e^{-yh(x)}$  & $\sE_{\ell_{0-1}}(h) - \sE^*_{\ell_{0-1}}(\sH) + \sM_{\ell_{0-1}}(\sH) \leq \sqrt{2}\paren*{\sE_{\ell}(h)
-\sE^*_{\ell}(\sH) + \sM_{\ell}(\sH)}^{\frac12}$ \\
   $ \log(1 + e^{u})$   & $ \log(1 + e^{-yh(x)})$ & $\sE_{\ell_{0-1}}(h) - \sE^*_{\ell_{0-1}}(\sH) + \sM_{\ell_{0-1}}(\sH) \leq \sqrt{2}\paren*{\sE_{\ell}(h)
-\sE^*_{\ell}(\sH) + \sM_{\ell}(\sH)}^{\frac12}$    \\
  $\max\curl*{0, 1 + u}^2$   & $\max\curl*{0, 1 - yh(x)}^2$ & $\sE_{\ell_{0-1}}(h) - \sE^*_{\ell_{0-1}}(\sH) + \sM_{\ell_{0-1}}(\sH) \leq \paren*{\sE_{\ell}(h)
-\sE^*_{\ell}(\sH) + \sM_{\ell}(\sH)}^{\frac12}$    \\
    $\max\curl*{0, 1 + u}$   & $\max\curl*{0, 1 - yh(x)}$ & $\sE_{\ell_{0-1}}(h) - \sE^*_{\ell_{0-1}}(\sH) + \sM_{\ell_{0-1}}(\sH) \leq \sE_{\ell}(h)
-\sE^*_{\ell}(\sH) + \sM_{\ell}(\sH)$    \\

    \bottomrule
  \end{tabular}
  }
  \caption{Examples of $\sH$-consistency bounds for binary margin-based losses.}
\label{tab:bounds-binary}
\end{table}

\begin{table}[t]
  \centering
   \resizebox{\textwidth}{!}{
  \begin{tabular}{@{\hspace{0cm}}lll@{\hspace{0cm}}}
    \toprule
    $\Phi(u)$  & Comp-sum losses $\ell$   & $\sH$-Consistency bounds\\
    \midrule
    $\frac{1 - u}{u}$ & $\sum_{y'\neq y} e^{h(x, y') - h(x, y)}$  & $\sE_{\ell_{0-1}}(h) - \sE^*_{\ell_{0-1}}(\sH) + \sM_{\ell_{0-1}}(\sH) \leq \sqrt{2}\paren*{\sE_{\ell}(h)
-\sE^*_{\ell}(\sH) + \sM_{\ell}(\sH)}^{\frac12}$ \\
    $-\log(u)$   & $-\log\paren*{\frac{e^{h(x, y)}}{\sum_{y'\in \sY}e^{h(x, y')}}}$ & $\sE_{\ell_{0-1}}(h) - \sE^*_{\ell_{0-1}}(\sH) + \sM_{\ell_{0-1}}(\sH) \leq \sqrt{2}\paren*{\sE_{\ell}(h)
-\sE^*_{\ell}(\sH) + \sM_{\ell}(\sH)}^{\frac12}$    \\
    $\frac{1}{\alpha}\bracket*{1 - u^{\alpha}}$    & $\frac{1}{\alpha}\bracket*{1 - \bracket*{\frac{e^{h(x, y)}}
    {\sum_{y'\in  \sY} e^{h(x, y')}}}^{\alpha}}$  &  $\sE_{\ell_{0-1}}(h) - \sE^*_{\ell_{0-1}}(\sH) + \sM_{\ell_{0-1}}(\sH) \leq \sqrt{2n^{\alpha}}\paren*{\sE_{\ell}(h)
-\sE^*_{\ell}(\sH) + \sM_{\ell}(\sH)}^{\frac12}$  \\
    $1 - u$ & $ 1 - \frac{e^{h(x, y)}}{\sum_{y'\in \sY} e^{h(x, y')}}$ & $\sE_{\ell_{0-1}}(h) - \sE^*_{\ell_{0-1}}(\sH) + \sM_{\ell_{0-1}}(\sH) \leq n \paren*{\sE_{\ell}(h)
-\sE^*_{\ell}(\sH) + \sM_{\ell}(\sH)}$    \\
    \bottomrule
  \end{tabular}
  }
  \caption{Examples of $\sH$-consistency bounds for comp-sum losses.}
\label{tab:bounds-comp}
\end{table}

\begin{table}[t]
  \centering
   \resizebox{\textwidth}{!}{
  \begin{tabular}{@{\hspace{0cm}}lll@{\hspace{0cm}}}
    \toprule
    $\Phi(u)$  & Constrained losses $\ell$   & $\sH$-Consistency bounds\\
    \midrule
    $e^{u}$ & $\sum_{y'\neq y}e^{h(x, y')}$  & $\sE_{\ell_{0-1}}(h) - \sE^*_{\ell_{0-1}}(\sH) + \sM_{\ell_{0-1}}(\sH) \leq \sqrt{2}\paren*{\sE_{\ell}(h)
-\sE^*_{\ell}(\sH) + \sM_{\ell}(\sH)}^{\frac12}$ \\
   $\max\curl*{0, 1 + u}^2$   & $\sum_{y'\neq y}\max \curl*{0, 1 + h(x, y')}^2$ & $\sE_{\ell_{0-1}}(h) - \sE^*_{\ell_{0-1}}(\sH) + \sM_{\ell_{0-1}}(\sH) \leq \paren*{\sE_{\ell}(h)
-\sE^*_{\ell}(\sH) + \sM_{\ell}(\sH)}^{\frac12}$    \\
  $(1 + u)^2$   & $\sum_{y'\neq y} \paren*{1 + h(x, y')}^2$ & $\sE_{\ell_{0-1}}(h) - \sE^*_{\ell_{0-1}}(\sH) + \sM_{\ell_{0-1}}(\sH) \leq \paren*{\sE_{\ell}(h)
-\sE^*_{\ell}(\sH) + \sM_{\ell}(\sH)}^{\frac12}$    \\
    $\max\curl*{0, 1 + u}$   & $\sum_{y'\neq y}\max \curl*{0, 1 + h(x, y')}$ & $\sE_{\ell_{0-1}}(h) - \sE^*_{\ell_{0-1}}(\sH) + \sM_{\ell_{0-1}}(\sH) \leq \sE_{\ell}(h)
-\sE^*_{\ell}(\sH) + \sM_{\ell}(\sH)$    \\

    \bottomrule
  \end{tabular}
  }
  \caption{Examples of $\sH$-consistency bounds for constrained losses with
$\sum_{y\in \sY} h(x, y) = 0$.}
\label{tab:bounds-cstnd}
\end{table}

\section{Comparison with excess error bounds}
\label{app:excess-bounds}

Excess error bounds can be used to derive
bounds for a hypothesis set $\sH$ expressed in terms of the
approximation error. Here, we show, however, that, the resulting
bounds are looser than $\sH$-consistency bounds.

Fix a target loss function $\ell_2$ and a surrogate loss
$\ell_1$. Excess error bounds, also known as \emph{surrogate regret bounds},
are bounds relating the excess errors of these loss functions of the
following form:
\begin{equation}
\label{eq:excess-error-bound}
\forall h \in \sH_{\rm{all}}, \quad
	\psi\paren*{\sE_{\ell_2}(h) - \sE^*_{\ell_2}(\sH_{\rm{all}})}
    	\leq \sE_{\ell_1}(h) - \sE^*_{\ell_1}(\sH_{\rm{all}}),
\end{equation}
where $\psi\colon \Rset_+ \to \Rset_+$ is a non-decreasing and convex
function on $\Rset_+$. Recall that as shown in
\eqref{eq:excess-error-decomp}, the excess error can be written as the
sum of the estimation error and the approximation error. Thus, the
excess error bound can be equivalently expressed as follows:
\begin{equation}
\label{eq:excess-error-bound-equiv}
\forall h \in \sH_{\rm{all}}, \quad
	\psi\paren*{\sE_{\ell_2}(h) - \sE^*_{\ell_2}(\sH) + \sA_{\ell_2}(\sH)}
    	\leq \sE_{\ell_1}(h) - \sE^*_{\ell_1}(\sH) + \sA_{\ell_1}(\sH).
\end{equation}
In Section~\ref{sec:min}, we have shown that the minimizabiliy gap can
be upper-bounded by the approximation error $\sM_{\ell}(\sH)\leq
\sA(\sH)$ and is in general a finer quantity for a surrogate loss
$\ell_1$. However, we will show that for a target loss $\ell_2$ that
is \emph{discrete}, the minimizabiliy gap in general coincides with
the approximation error.
\begin{definition}
We say that a target loss $\ell_2$ is \emph{discrete} if we can write
$\ell_2(h, x, y) = \sfL(\hh(x), y)$ for some binary function
$\sfL\colon \sY\times\sY \to \Rset_{+}$.
\end{definition}
In other words, a discrete target loss $ \ell_2 $ is explicitly a
function of both the prediction $ \hh(x) $ and the true label $ y $,
where both belong to the label space $ \sY $. Consequently, it can
assume at most $ n^2 $ distinct discrete values.

Next, we demonstrate that for such discrete target loss functions, if
for any instance, the set of predictions generated by the hypothesis
set completely spans the label space, then the minimizability gap is
precisely equal to the approximation error. For convenience, we denote
by $\sfH(x)$ the set of predictions generated by the hypothesis set on
input $x \in \sX$, defined as $ \sfH(x) = \curl*{\hh(x)\colon h \in
  \sH}$
\begin{theorem}
\label{thm:min-discrete}
Given a discrete target loss function $\ell_2$. Assume that the
hypothesis set $\sH$ satisfies, for any $x \in \sX$, $\sfH(x) =
\sY$. Then, we have $\sI_{\ell_2}(\sH) = 0$ and $\sM_{\ell_2}(\sH) =
\sA_{\ell_2}(\sH)$.
\end{theorem}
\begin{proof}
As shown in Section~\ref{sec:min}, the minimizability gap can be
decomposed in terms of the approximation error and the difference of
pointwise infima:
\begin{align*}
  \sM_{\ell_{2}}(\sH)
  & = \sA_{\ell_{2}}(\sH) - \sI_{\ell_{2}}(\sH)\\
  & = \sA_{\ell_{2}}(\sH) - \E_x \bracket[\Big]{\sC^*_{\ell_{2}}(\sH, x) - \sC^*_{\ell_{2}}(\sH_{\rm{all}}, x)}.
\end{align*}
By definition and the fact that $\ell_2$ is discrete, the conditional error can be written as
\begin{align*}
\sC_{\ell_2}(h,x) = \sum_{y\in \sY} \sfp(y \!\mid\! x) \ell_2(h, x, y) = \sum_{y\in \sY} \sfp(y \!\mid\! x) \sfL(\hh(x), y).
\end{align*}
Thus, for any $x\in \sX$, the best-in-class conditional error can be expressed as 
\begin{equation*}
\sC_{\ell_2}^*(\sH,x) = \inf_{h\in \sH} \sum_{y\in \sY} \sfp(y \!\mid\! x) \sfL(\hh(x), y) = \inf_{y' \in \sfH(x)} \sum_{y\in \sY} \sfp(y \!\mid\! x) \sfL(y', y).
\end{equation*}
By the assumption that $\sfH(x) = \sY$, we obtain
\begin{equation*}
\forall x\in \sX, \quad \sC_{\ell_2}^*(\sH,x) = \inf_{y' \in \sfH(x)} \sum_{y\in \sY} \sfp(y \!\mid\! x) \sfL(y', y) = \inf_{y' \in \sY} \sum_{y\in \sY} \sfp(y \!\mid\! x) \sfL(y', y) = \sC_{\ell_2}^*(\sH_{\rm{all}},x).
\end{equation*}
Therefore, $\sI_{\ell_{2}}(\sH) = \E_x \bracket[\Big]{\sC^*_{\ell_{2}}(\sH, x) - \sC^*_{\ell_{2}}(\sH_{\rm{all}}, x)} = 0$ and $\sM_{\ell_{2}}(\sH) = \sA_{\ell_{2}}(\sH)$.
\end{proof}
By Theorem~\ref{thm:min-discrete}, for a target loss $\ell_2$ that is
discrete and hypothesis sets $\sH$ modulo mild assumptions, the
minimizabiliy gap coincides with the approximation error. In such
cases, by comparing an excess error bound
\eqref{eq:excess-error-bound-equiv} with the $\sH$-consistency bound
\eqref{eq:H-consistency-bound-srd}:
\begin{align*}
&\text{Excess error bound:} \quad 
	\psi\paren*{\sE_{\ell_2}(h) - \sE^*_{\ell_2}(\sH) + \sA_{\ell_2}(\sH)}
    	\leq \sE_{\ell_1}(h) - \sE^*_{\ell_1}(\sH) + \sA_{\ell_1}(\sH)\\
&\text{$\sH$-consistency bound:} \quad
\psi \paren*{\sE_{\ell_2}(h) - \sE^*_{\ell_2}(\sH) + \sM_{\ell_2}(\sH)} \leq \sE_{\ell_1}(h) - \sE^*_{\ell_1}(\sH) + \sM_{\ell_1}(\sH),
\end{align*}
we obtain that the left-hand side of both bounds are equal (since
$\sM_{\ell_2}(\sH) = \sA_{\ell_2}(\sH) $), while the right-hand side
of the $\sH$-consistency bound is always upper-bounded by and can be
finer than the right-hand side of the excess error bound (since
$\sM_{\ell_1}(\sH) \leq \sA_{\ell_1}(\sH) $), which implies that
excess error bounds (or surrogate regret bounds) are in general
inferior to $\sH$-consistency bounds.

\section{Polyhedral losses versus smooth losses}
\label{app:poly-smooth}

Since $\sH$-consistency bounds subsume excess error bounds as a
special case (Appendix~\ref{app:excess-bounds}), the linear growth
rate of polyhedral loss excess error bounds
(\citet{finocchiaro2019embedding}) also dictates a linear growth rate
for polyhedral $\sH$-consistency bounds, if they exist. This is
illustrated by the hinge loss or $\rho$-margin loss which have been
shown to benefit from $\sH$-consistency bounds
\citep{awasthi2022Hconsistency}.

\ignore{
As mentioned in Appendix~\ref{app:excess-bounds}, $\sH$-consistency bounds include excess error bounds as a special case when $\sH = \sH_{\rm{all}}$. Therefore, the linear growth rate of the excess error bound of polyhedral losses shown by \citet{finocchiaro2019embedding} also implies that the growth rate for the $\sH$-consistency bound in the polyhedral case is linear. 
}

Here, we compare in more detail polyhedral losses and the smooth losses. Assume that a hypothesis set $\sH$ is complete and thus $\sfH(x) =
\sY$ for any $x \in \sX$. By Theorem~\ref{thm:min-discrete}, we have $\sA_{\ell_{0-1}}(\sH) = \sM_{\ell_{0-1}}(\sH)$. As shown by \citet[Theorem~3]{frongillo2021surrogate}, a Bayes-consistent polyhedral loss $\Phi_{\rm{poly}}$ admits the following linear excess error bound, for some $\beta_1 > 0$,
\begin{equation}
\label{eq:bound-poly}
\forall h\in \sH,\, \beta_1 \paren*{\sE_{\ell_{0-1}}(h) - \sE^*_{\ell_{0-1}}(\sH) + \sM_{\ell_{0-1}}(\sH)}
    	\leq \sE_{\Phi_{\rm{poly}}}(h) - \sE^*_{\Phi_{\rm{poly}}}(\sH) + \sA_{\Phi_{\rm{poly}}}(\sH).
\end{equation}
However, for a smooth loss $\Phi_{\rm{smooth}}$, if it satisfies the condition of Theorem~\ref{thm:binary-lower},  $\Phi_{\rm{smooth}}$ admits the following $\sH$-consistency bound:
\begin{equation}
\label{eq:bound-smooth}
\forall h\in \sH,\, \sT\paren*{\sE_{\ell_{0-1}}(h) - \sE^*_{\ell_{0-1}}(\sH) + \sM_{\ell_{0-1}}(\sH)}
    	\leq \sE_{\Phi_{\rm{smooth}}}(h) - \sE^*_{\Phi_{\rm{smooth}}}(\sH) + \sM_{\Phi_{\rm{smooth}}}(\sH).
\end{equation}
where $\sT(t) = \Theta(t^2)$. Therefore, our theory offers a principled basis for comparing polyhedral losses \eqref{eq:bound-poly} and smooth losses \eqref{eq:bound-smooth}, which depends on the following factors:
\begin{itemize}
    \item The growth rate: linear for polyhedral losses, while square-root for smooth losses.
    
    \item The optimization property: smooth losses are more favorable for optimization compared to polyhedral losses, in particular with deep neural networks.
    
    \item The approximation theory: the approximation error $ \sA_{\Phi_{\rm{poly}}}(\sH)$ appears on the right-hand side of the bound for polyhedral losses, whereas a finer quantity, the minimizability gap $\sM_{\Phi_{\rm{smooth}}}(\sH)$, is present on the right-hand side of the bound for smooth losses.
\end{itemize}

\section{Comparison of minimizability gaps across comp-sum losses}

\label{app:M-gaps-comp}
For $\ell_{\tau}^{\rm{comp}}$ loss functions, $\tau \in [0, 2)$, we
  can characterize minimizability gaps as follows.

\begin{restatable}
  {theorem}{GapUpperBoundDetermiSrd}
\label{Thm:gap-upper-bound-determi-srd}
Assume that for any $x \in \sX$, we have $\curl*{\paren*{h(x, 1),
    \ldots, h(x, n)}\colon h \in \sH}$ = $[-\Lambda,
  +\Lambda]^n$. Then, for comp-sum losses $\ell_{\tau}^{\rm{comp}}$
and any deterministic distribution, the minimizability gaps can be
expressed as follows:
\begin{align}
\sM_{\ell_{\tau}^{\rm{comp}}}(\sH)
\leq \wt \sM_{\ell_{\tau}^{\rm{comp}}}(\sH) = f_{\tau} \paren*{\sR^*_{\ell_{\tau = 0}^{\rm{comp}}}(\sH)} - f_{\tau} \paren*{\sC^*_{\ell_{\tau = 0}^{\rm{comp}}}(\sH, x)},
\end{align}
where $f_{\tau}(u) = \log(1 + u) 1_{\tau = 1} + \frac{1}{1 - \tau}
\paren*{(1 + u)^{1 - \tau} - 1} 1_{\tau \neq 1}$ and
$\sC^*_{\ell_{\tau = 0}^{\rm{comp}}}(\sH, x) = e^{-2 \Lambda}(n -
1)$. Moreover, $\wt \sM_{\ell_{\tau}^{\rm{comp}}}(\sH)$ is a
non-increasing function of $\tau$.
\end{restatable}

\begin{proof}
Since $f_{\tau}$ is concave and non-decreasing, and the equality $\ell_{\tau} = f_{\tau} \paren*{\ell_{\tau = 0}}$ holds, the minimizability
gaps can be upper-bounded as follows, for any $\tau \geq $0,
\begin{align*}
\sM_{\ell_{\tau}^{\rm{comp}}}(\sH)
\leq f_{\tau}\paren*{\sR^*_{\ell_{\tau=0}^{\rm{comp}}}(\sH)} - \E_x[\sC^*_{\ell_{\tau}^{\rm{comp}}}(\sH, x)].
\end{align*}
Since the distribution is deterministic, the conditional error can be expressed as follows:
\begin{equation}
\label{eq:cond-comp-sum-determi-srd}
\begin{aligned}
\sC_{\ell_{\tau}^{\rm{comp}}}(h, x)  =  & f_{\tau}\paren*{\sum_{y'\neq y_{\max}}\exp\paren*{h(x,y')-h(x,y_{\max})}}
\end{aligned}
\end{equation}
where $y_{\max} = \argmax \sfp(y \!\mid\! x)$.
Using the fact that $f_{\tau}$ is increasing for any $\tau>0$, the hypothesis
$
h^*\colon (x, y) \mapsto \Lambda 1_{y = y_{\max}} - \Lambda 1_{y \neq y_{\max}}
$
achieves the best-in-class conditional error.
Thus, 
\begin{align*}
\sC^*_{\ell_{\tau}^{\rm{comp}}}(\sH, x) 
=  \sC_{\ell_{\tau}^{\rm{comp}}}(h^*, x) = f_{\tau}\paren*{\sC^*_{\ell_{\tau = 0}^{\rm{comp}}}(\sH, x)}
\end{align*}
where $\sC^*_{\ell_{\tau = 0}^{\rm{comp}}}(\sH, x) = e^{-2\Lambda}(n - 1)$. Therefore,
\begin{align*}
\sM_{\ell_{\tau}^{\rm{comp}}}(\sH)
\leq f_{\tau}\paren*{\sR^*_{\ell_{\tau = 0}^{\rm{comp}}}(\sH)} - f_{\tau}\paren*{\sC^*_{\ell_{\tau = 0}^{\rm{comp}}}(\sH, x)}.
\end{align*}
This completes the first part of the proof.
Using the fact that $\tau \mapsto f_{\tau} (u_1) - f_{\tau}(u_2)$ is a non-increasing function of $\tau$ for any $u_1 \geq u_2 \geq 0$, the second proof is completed as well.
\end{proof}

The theorem shows that for comp-sum
loss functions $\ell_{\tau}^{\rm{comp}}$, the minimizability gaps are
non-increasing with respect to $\tau$. Note that $\Phi^{\tau}$
satisfies the conditions of Theorem~\ref{thm:comp-lower} for any $\tau
\in [0, 2)$. Therefore, focusing on behavior near zero (ignoring
  constants), the theorem provides a principled comparison of
  minimizability gaps and $\sH$-consistency bounds across different
  comp-sum losses.

\section{Small surrogate minimizability gaps: proof for binary classification}
\label{app:small-M-gaps}

\ZeroMinGap*

\begin{proof}
By definition of $h^*$, using the shorthand $p = \P[y = +1]$, we can write
  \begin{align*}  
  \inf_{h \in \sH} \E[\ell(h(x), y)]
  & = \E[\ell(h^*(x), y)]\\
  & = p \E[\ell(h^*(x), +1) \mid y = +1]
  + (1 - p) \E[\ell(h^*(x), -1) \mid y = -1].
\end{align*}
  Since the distribution is deterministic, the expected pointwise infimum
  can be rewritten as follows:
\begin{align*}  
  \E_{x}\bracket*{\inf_{h \in \sH} \E_{y}[\ell(h(x), y) \mid x ]}
  =  \E_{x}\bracket*{\inf_{\alpha \in A} \E_{y}[\ell(\alpha, y) \mid x]}
  & = p \inf_{\alpha \in A} \ell(\alpha, +1)
  + (1 - p) \inf_{\alpha \in A} \ell(\alpha, -1)\\
  & = p \ell_+
  + (1 - p) \ell_-,
\end{align*}
where $\ell_+ = \inf_{\alpha \in A} \ell(\alpha, +1)$
and $\ell_- = \inf_{\alpha \in A} \ell(\alpha, -1)$.
Thus, we have
\begin{align*}
  \sM(\sH)
  & = p \E\bracket*{\ell(h^*(x), +1) - \ell_+ \mid y = +1}
  + (1 - p) \E\bracket*{\ell(h^*(x), -1) - \ell_- \mid y = -1}.
\end{align*}
In view of that, since, by definition of $\ell_+$ and $\ell_-$, the
expressions within the conditional expectations are non-negative, the
equality $\sM(\sH) = 0$ holds iff $\ell(h^*(x), +1) - \ell_+ = 0$
almost surely for any $x$ in $\sX_+$ and $\ell(h^*(x), -1) - \ell_- =
0$ almost surely for any $x$ in $\sX_-$. This completes the first part of the proof. 
Furthermore,
$\sM(\sH) \leq \e$ is equivalent to
\[
p \E\bracket*{\ell(h^*(x), +1) - \ell_+ \mid y = +1}
+ (1 - p) \E\bracket*{\ell(h^*(x), -1) - \ell_- \mid y = -1} \leq \e
\]
that is
\[
p \paren*{\E \bracket*{\ell(h^*(x), +1) \mid y = +1} - \ell_+ }
+ (1 - p) \paren*{\E\bracket*{\ell(h^*(x), -1) \mid y = -1} - \ell_-} \leq \e.
\]
In light of the non-negativity of the expressions, this implies in
particular:
\begin{align*}
  & \E \bracket*{\ell(h^*(x), +1) \mid y = +1} - \ell_+ \leq \frac{\e}{p}
  \quad \text{and} \quad
  \E \bracket*{\ell(h^*(x), -1) \mid y = -1} - \ell_- \leq \frac{\e}{1 - p}.
\end{align*}
This completes the second part of the proof.
\end{proof}

\ZeroMinGapStochastic*

\begin{proof}
Assume that the best-in-class
  error is achieved by some $h^*\in \sH$.
Then, we can write
  \begin{align*}  
  \inf_{h \in \sH} \E[\ell(h(x), y)]
   = \E[\ell(h^*(x), y)]=\E_{x}\bracket*{\E_{y}[\ell(h^*(x), y) \mid x]}.
\end{align*}
The expected pointwise infimum
  can be rewritten as follows:
\begin{align*}  
  \E_{x}\bracket*{\inf_{h \in \sH} \E_{y}[\ell(h(x), y) \mid x]}
  =  \E_{x}\bracket*{\inf_{\alpha \in A} \E_{y}[\ell(\alpha, y) \mid x]}.
\end{align*}
Thus, we have
\begin{align*}
  \sM(\sH)
  & = \E_{x}\bracket*{\E_{y}[\ell(h^*(x), y) \mid x]
    - \inf_{\alpha \in A} \E_{y}[\ell(\alpha, y) \mid x]}.
\end{align*}
In view of that, since, by the definition of infimum, the
expressions within the marginal expectations are non-negative, the condition that $\sM(\sH) = 0$ implies that  \begin{align*}
  \E_{y}[\ell(h^*(x), y) \mid x] = \inf_{\alpha \in A} \E_{y}[\ell(\alpha, y) \mid x] \text{ a.s.\ over $\sX$}.
  \end{align*}
On the other hand, if there exists $h^*\in \sH$ such that the
condition \eqref{eq:cond-zero-stochastic} holds, then,
\begin{align*}
 \sM(\sH)
  & =  \inf_{h \in \sH} \E[\ell(h(x), y)] -  \E_{x}\bracket*{\inf_{\alpha \in A} \E_{y}[\ell(\alpha, y) \mid x]}\\
  & \leq \E_{x}\bracket*{\E_{y}[\ell(h^*(x), y) \mid x]-\inf_{\alpha \in A} \E_{y}[\ell(\alpha, y) \mid x]} = 0.
\end{align*}
Since $\sM(\sH)$ is non-negative, the inequality is achieved. Thus, we
have
\begin{align*}
  \sM(\sH) = 0 \text{ and } \inf_{h \in \sH} \E[\ell(h(x), y)] = \E[\ell(h^*(x), y)].
\end{align*}
If there exists $h^*\in \sH$ such that the
condition~\eqref{eq:cond-epsilon-stochastic} holds, then,
\begin{align*}
 \sM(\sH)
  & =  \inf_{h \in \sH} \E[\ell(h(x), y)] -  \E_{x}\bracket*{\inf_{\alpha \in A} \E_{y}[\ell(\alpha, y) \mid x]}\\
  & \leq \E_{x}\bracket*{\E_{y}[\ell(h^*(x), y) \mid x]-\inf_{\alpha \in A} \E_{y}[\ell(\alpha, y) \mid x]} = \e.
\end{align*}
On the other hand, since we have
\begin{align*}
  \sM(\sH)
  & = \E_{x}\bracket*{\E_{y}[\ell(h^*(x), y) \mid x] - \inf_{\alpha \in A} \E_{y}[\ell(\alpha, y) \mid x]},
\end{align*}
$\sM(\sH) \leq \e$ implies that
\[
\E_{x}\bracket*{\E_{y}[\ell(h^*(x), y) \mid x] - \inf_{\alpha \in A} \E_{y}[\ell(\alpha, y) \mid x]} \leq \e.
\]
This completes the proof.
\end{proof}

\section{Small surrogate minimizability gaps: multi-class classification}
\label{app:small-M-gaps-multi}

We consider the multi-class setting with label space $[n] = \curl*{1,
  2, \ldots, n}$. In this setting, the surrogate loss incurred by a
predictor $h$ at a labeled point $(x, y)$ can be expressed by
$\ell(h(x), y)$, where $h(x) = \bracket*{h(x, 1), \ldots, h(x, n)}$ is
the score vector of the predictor $h$.
We denote by $A$ the set of values in $\Rset^n$ taken by the score
vector of predictors in $\sH$ at $x$, which we assume to be
independent of $x$: $A = \curl*{h(x) \colon h \in \sH}$, for all $x
\in \sX$.

\subsection{Deterministic scenario}
We first consider the deterministic scenario, where the conditional
probability $\sfp(y \!\mid\! x)$ is either zero or one. For a deterministic distribution, we denote by $\sX_{k}$ the subset of
$\sX$ over which the label is $k$. For convenience, let  $\ell_k = \inf_{\alpha \in A} \ell(\alpha, k)$, for any $k\in [n]$.

\begin{restatable}{theorem}{ZeroMinGapMulti}
\label{th:ZeroMinGapMulti}
  Assume that $\sD$ is deterministic and that the best-in-class error
  is achieved by some $h^* \in \sH$. Then, the minimizability gap is
  null, $\sM(\sH) = 0$, iff
  \begin{align*}
  \forall k \in [n],\, \ell(h^*(x), k) & = \ell_k  \text{ a.s.\ over $\sX_{k}$}.
  \end{align*}
  If further $\alpha \mapsto \ell(\alpha, k)$ is injective and
  $\ell_k = \ell(\alpha_{k}, k)$ for all $k \in [n]$,
  then, the condition is equivalent to
  $
    \forall k \in [n],\, h^*(x) =
    \alpha_{k}  \text{ a.s.\ for $x \in \sX_{k}$}.
  $ Furthermore, the minimizability
  gap is bounded by $\e$, $\sM(\sH) \leq \e$, iff
\[
\sum_{k \in [n]}  p_k \paren*{\E \bracket*{\ell(h^*(x), k) \mid y = k} - \ell_k }\leq \e.
\]
In particular, the condition implies:
\begin{align*}
  \E \bracket*{\ell(h^*(x), k) \mid y = k} - \ell_k \leq \frac{\e}{p_k},\, \forall k\in [n],
\end{align*}
\end{restatable}

\begin{proof}
By definition of $h^*$, using the shorthand $p_{k} = \P[y = k]$ for any $k\in [n]$, we can write
  \begin{align*}  
  \inf_{h \in \sH} \E[\ell(h(x), y)] = \E[\ell(h^*(x), y)]= \sum_{k \in [n]}  p_k \E[\ell(h^*(x), k) \mid y = k].
\end{align*}
  Since the distribution is deterministic, the expected pointwise
  infimum can be rewritten as follows:
\begin{align*}  
  \E_{x}\bracket*{\inf_{h \in \sH} \E_{y}[\ell(h(x), y) \mid x]}
  =  \E_{x}\bracket*{\inf_{\alpha \in A} \E_{y}[\ell(\alpha, y) \mid x]} = \sum_{k \in [n]}  p_k \inf_{\alpha \in A} \ell(\alpha, k) = \sum_{k \in [n]}  p_k \ell_k,
\end{align*}
where $\ell_k = \inf_{\alpha \in A} \ell(\alpha, k)$, for any $k\in [n]$.
Thus, we have
\begin{align*}
  \sM(\sH)
  & = \sum_{k \in [n]}  p_k \E\bracket*{\ell(h^*(x), k) - \ell_k \mid y = k}.
\end{align*}
In view of that, since, by definition of $\ell_k$, the expressions
within the conditional expectations are non-negative, the equality
$\sM(\sH) = 0$ holds iff $\ell(h^*(x), k) - \ell_k = 0$ almost surely
for any $x$ in $\sX_k$, $\forall k \in [n]$. Furthermore,
$\sM(\sH) \leq \e$ is equivalent to
\[
\sum_{k \in [n]}  p_k \E\bracket*{\ell(h^*(x), k) - \ell_k \mid y = k} \leq \e
\]
that is
\[
\sum_{k \in [n]}  p_k \paren*{\E \bracket*{\ell(h^*(x), k) \mid y = k} - \ell_k }\leq \e.
\]
In light of the non-negativity of the expressions, this implies in
particular:
\begin{align*}
 \E \bracket*{\ell(h^*(x), k) \mid y = k} - \ell_k \leq \frac{\e}{p_k}, \, \forall k\in [n].
\end{align*}
This completes the proof.
\end{proof}

The theorem suggests that, under those assumptions, for the surrogate
minimizability gap to be zero, the score vector of best-in-class
hypothesis must be piecewise constant with specific values on
$\sX_k$. The existence of such a hypothesis in $\sH$ depends both on
the complexity of the decision surface separating $\sX_k$ and on that
of the hypothesis set $\sH$. The theorem also suggests that when the score vector of best-in-class classifier
$\e$-approximates $\alpha_k$ over $\sX_k$ for any $k\in [n]$,
then the minimizability gap is bounded by $\e$. The existence of such
a hypothesis in $\sH$ depends on the complexity of the decision
surface.

\subsection{Stochastic scenario}
In the previous sections, we analyze instances featuring small
minimizability gaps in a deterministic setting. Moving forward, we aim
to extend this analysis to the stochastic scenario. We first provide
two general results, which are the direct extensions of that in the
deterministic scenario. The following result shows that the
minimizability gap is zero when there exists $h^*\in \sH$ that matches
$\alpha^*(x)$ for all $x$, where $\alpha^*(x)$ is the minimizer of the
conditional error.  It also shows that the
minimizability gap is bounded by $\e$ when there exists $h^*\in \sH$
whose conditional error $\e$-approximates best-in-class conditional
error for all $x$.

\begin{restatable}{theorem}{ZeroMinGapStochasticMulti}
\label{th:ZeroMinGapStochasticMulti}
  The best-in-class
  error is achieved by some $h^*\in \sH$ and the minimizability
  gap is null, $\sM(\sH) = 0$, iff there exists $h^*\in \sH$ such that
  \begin{align}
  \label{eq:cond-zero-stochastic-Multi}
  \E_{y}[\ell(h^*(x), y) \mid x] = \inf_{\alpha \in A} \E_{y}[\ell(\alpha, y) \mid x] \text{ a.s.\ over $\sX$}.
  \end{align}
  If further $\alpha \mapsto \E_{y}[\ell(\alpha, y) \mid x]$ is injective
  and $\inf_{\alpha \in A} \E_{y}[\ell(\alpha, y) \mid x] = \E_{y}[\ell(\alpha^*(x), y) \mid x]$, then, the condition is equivalent to
  $h^*(x) =
    \alpha^*(x)  \text{ a.s.\ for $x \in \sX$}$. Furthermore, the minimizability
  gap is bounded by $\e$, $\sM(\sH) \leq \e$, iff there exists $h^*\in \sH$ such that
\begin{align}
\label{eq:cond-epsilon-stochastic-Multi}
\E_{x}\bracket*{\E_{y}[\ell(h^*(x), y) \mid x] - \inf_{\alpha \in A} \E_{y}[\ell(\alpha, y) \mid x]} \leq \e.
\end{align}
\end{restatable}
\begin{proof}
Assume that the best-in-class
  error is achieved by some $h^*\in \sH$.
Then, we can write
  \begin{align*}  
  \inf_{h \in \sH} \E[\ell(h(x), y)]
   = \E[\ell(h^*(x), y)]=\E_{x}\bracket*{\E_{y}[\ell(h^*(x), y) \mid x]}.
\end{align*}
The expected pointwise infimum
  can be rewritten as follows:
\begin{align*}  
  \E_{x}\bracket*{\inf_{h \in \sH} \E_{y}[\ell(h(x), y) \mid x]}
  =  \E_{x}\bracket*{\inf_{\alpha \in A} \E_{y}[\ell(\alpha, y) \mid x]}.
\end{align*}
Thus, we have
\begin{align*}
  \sM(\sH)
  & = \E_{x}\bracket*{\E_{y}[\ell(h^*(x), y) \mid x]
    - \inf_{\alpha \in A} \E_{y}[\ell(\alpha, y) \mid x]}.
\end{align*}
In view of that, since, by the definition of infimum, the
expressions within the marginal expectations are non-negative, the condition that $\sM(\sH) = 0$ implies that  \begin{align*}
  \E_{y}[\ell(h^*(x), y) \mid x] = \inf_{\alpha \in A} \E_{y}[\ell(\alpha, y) \mid x] \text{ a.s.\ over $\sX$}.
  \end{align*}
On the other hand, if there exists $h^*\in \sH$ such that the
condition \eqref{eq:cond-zero-stochastic} holds, then,
\begin{align*}
 \sM(\sH)
  & =  \inf_{h \in \sH} \E[\ell(h(x), y)] -  \E_{x}\bracket*{\inf_{\alpha \in A} \E_{y}[\ell(\alpha, y) \mid x]}\\
  & \leq \E_{x}\bracket*{\E_{y}[\ell(h^*(x), y) \mid x]-\inf_{\alpha \in A} \E_{y}[\ell(\alpha, y) \mid x]} = 0.
\end{align*}
Since $\sM(\sH)$ is non-negative, the inequality is achieved. Thus, we
have
\begin{align*}
  \sM(\sH) = 0 \text{ and } \inf_{h \in \sH} \E[\ell(h(x), y)] = \E[\ell(h^*(x), y)].
\end{align*}
If there exists $h^*\in \sH$ such that the
condition~\eqref{eq:cond-epsilon-stochastic} holds, then,
\begin{align*}
 \sM(\sH)
  & =  \inf_{h \in \sH} \E[\ell(h(x), y)] -  \E_{x}\bracket*{\inf_{\alpha \in A} \E_{y}[\ell(\alpha, y) \mid x]}\\
  & \leq \E_{x}\bracket*{\E_{y}[\ell(h^*(x), y) \mid x]-\inf_{\alpha \in A} \E_{y}[\ell(\alpha, y) \mid x]} = \e.
\end{align*}
On the other hand, since we have
\begin{align*}
  \sM(\sH)
  & = \E_{x}\bracket*{\E_{y}[\ell(h^*(x), y) \mid x] - \inf_{\alpha \in A} \E_{y}[\ell(\alpha, y) \mid x]},
\end{align*}
$\sM(\sH) \leq \e$ implies that
\[
\E_{x}\bracket*{\E_{y}[\ell(h^*(x), y) \mid x] - \inf_{\alpha \in A} \E_{y}[\ell(\alpha, y) \mid x]} \leq \e.
\]
This completes the proof.
\end{proof}

\subsection{Examples}
\label{app:examples}

Note that when the distribution is assumed to be deterministic, the
condition~\eqref{eq:cond-epsilon-stochastic} and condition~\eqref{eq:cond-epsilon-stochastic-Multi} are reduced to the
condition of Theorem~\ref{th:ZeroMinGap} in binary classification
and that of Theorem~\ref{th:ZeroMinGapMulti} in multi-class
classification, respectively.  In the stochastic scenario, the
existence of such a hypothesis not only depends on the complexity of
the decision surface, but also depends on the distributional
assumption on the conditional distribution $p(x) = \paren*{\sfp(y \!\mid\! x)}_{y\in \sY}$, where $\sfp(y \!\mid\! x) = \sD(Y = y \!\mid\! X = x)$ is the
conditional probability of $Y = y$ given $X = x$. In the binary
classification, we have $p(x) = \paren*{\sfp(+1 \!\mid\! x), \sfp(-1 \!\mid\! x)}$, where $\sfp(+1 \!\mid\! x) + \sfp(-1 \!\mid\! x) = 1$. For simplicity, we
use the notation $\eta(x)$ and $1 - \eta(x)$ to represent $\sfp(+1 \!\mid\! x)$ and $\sfp(-1 \!\mid\! x)$ respectively. In the multi-class
classification with $\sY = \curl*{1, \ldots, n}$, we have $p(x) =
\paren*{\sfp(1 \!\mid\! x), \sfp(2 \!\mid\! x), \ldots, \sfp(n \!\mid\! x)}$ where $n$ is the number of
classes. As examples, here too, we examine exponential loss and
logistic loss in binary classification and multi-class logistic loss
in multi-class classification.

\textbf{A. Example: binary classification.} 
Let $\e\in [0,\frac12]$.  We denote by $\sX_+$ the subset of $\sX$
over which $\eta(x)=1-\e$ and by $\sX_-$ the subset of $\sX$ over
which $\eta(x) = \e$. Let $\sH$ be a family of functions $h$ with
$\abs*{h(x)} \leq \Lambda$ for all $x \in \sX$ and such that all
values in $[-\Lambda, +\Lambda]$ can be reached. Thus, $A =
\bracket*{-\Lambda, \Lambda}$ for any $x\in \sX$. Consider the
exponential loss: $\ell(h, x, y) = e^{-yh(x)}$. Then, for any $x\in
\sX$ and $\alpha\in A$, we have
\begin{align*}
\E_{y}[\ell(\alpha, y) \mid x] = \begin{cases}
(1-\e)e^{-\alpha} + \e e^{\alpha} & x\in \sX_{+}\\
\e e^{-\alpha} + (1-\e) e^{\alpha} & x\in \sX_{-}.
\end{cases}
\end{align*}
Thus, it is not hard to see that for any $\e \leq \frac{1}{e^{2\Lambda}+1}$, the infimum $\inf_{\alpha \in A} \E_{y}[\ell(\alpha, y) \mid x]$ can be achieved by $\alpha^*(x) = \begin{cases}
\Lambda & x\in \sX_{+}\\
-\Lambda & x\in \sX_{-}
\end{cases}\in A$. Similarly, for the logistic loss $\ell(h, x, y) = \log\paren*{1 + e^{-yh(x)}}$, we have that
\begin{align*}
\E_{y}[\ell(\alpha, y) \mid x] = \begin{cases}
(1 - \e)\log\paren*{1 + e^{-\alpha}} + \e \log\paren*{1 + e^{\alpha}} & x\in \sX_{+}\\
\e \log\paren*{1 + e^{-\alpha}} + (1 - \e) \log\paren*{1 + e^{\alpha}} & x\in \sX_{-}
\end{cases}
\end{align*}
and for $\e \leq \frac{1}{e^{\Lambda}+1}$, the infimum $\inf_{\alpha \in A} \E_{y}[\ell(\alpha, y) \mid x]$ can be achieved by $\alpha^*(x) = \begin{cases}
\Lambda & x\in \sX_{+}\\
-\Lambda & x\in \sX_{-}
\end{cases}$. Therefore, by Theorem~\ref{th:ZeroMinGapStochastic},
for these distributions and loss functions, when the best-in-class
classifier $h^*$ $\e$-approximates $\alpha_{+} = \Lambda$ over $\sX_+$
and $\alpha_{-} = -\Lambda$ over $\sX_-$, then the minimizability gap
is bounded by $\e$. The existence of such a hypothesis in $\sH$
depends on the complexity of the decision surface. For example, as
previously noted, when the decision surface is characterized by a
hyperplane, a hypothesis set of linear functions, coupled with a
sigmoid activation function, can offer a highly effective
approximation (see Figure~\ref{fig:illustration} for illustration).

\textbf{B. Example: multi-class classification.} 
Let $\e\in [0,\frac12]$. We denote by $\sX_{k}$ the subset of $\sX$
over which $\sfp(k \!\mid\! x) = 1 - \e$ and $\sfp(j \!\mid\! x) = \frac{\e}{n-1}$ for $j\neq
k$. Let $\sH$ be a family of functions $h$ with $\abs*{h(x, \cdot)}
\leq \Lambda$ for all $x \in \sX$ and such that all values in
$[-\Lambda, +\Lambda]$ can be reached. Thus, $A = \bracket*{-\Lambda,
  \Lambda}^n$ for any $x\in \sX$. Consider the multi-class logistic
loss: $\ell(h, x, y) = - \log \bracket*{\frac{e^{h(x,y)}}{\sum_{y' \in
      \sY} e^{h(x,y')}}}$. For any $\alpha = \bracket*{\alpha^1,
  \ldots, \alpha^n} \in A$, we denote by $S_k =
\frac{e^{\alpha^k}}{\sum_{k' \in [n]} e^{\alpha^{k'}}}$.  Then, for
any $x\in \sX$ and $\alpha\in A$,
\begin{align*}
\E_{y}[\ell(\alpha, y) \mid x] = - (1 - \e) \log \paren*{S_{k}} -
\frac{\e}{n - 1} \sum_{k'\neq k} \log \paren*{S_{k'}} \text{ if } x\in
\sX_k.
\end{align*}
Thus, it is not hard to see that for any $\e \leq \frac{n -
  1}{e^{2\lambda} + n - 1}$, the infimum $\inf_{\alpha \in A} \E_{y}[\ell(\alpha, y) \mid x]$ can be achieved by $\alpha^*(x) =
\bracket*{-\Lambda, \ldots, \Lambda, \ldots, -\Lambda}$, where
$\Lambda$ occupies the $k$-th position for $x\in \sX_{k}$.  Therefore,
by Theorem~\ref{th:ZeroMinGapStochastic}, for these distributions
and loss functions, when the best-in-class classifier $h^*$
$\e$-approximates $\alpha_{k} = \bracket*{-\Lambda, \ldots,
  \underset{k-\text{th}}{\Lambda}, \ldots, -\Lambda}$ over $\sX_{k}$,
then the minimizability gap is bounded by $\e$. The existence of such
a hypothesis in $\sH$ depends on the complexity of the decision
surface.

\section{Proof for binary margin-based losses (Theorem~\ref{thm:binary-lower})}
\label{app:binary-lower}

\BinaryLower*
\begin{proof}
Since $\Phi$ is convex and in $C^2$, $f_t$ is also convex and differentiable with respect to $u$. For any $t \in [0, 1]$, differentiate $f_t$ with respect to $u$, we have
\begin{equation*}
f'_t(u) = \frac{1 - t}{2} \Phi'(u) - \frac{1 + t}{2} \Phi'(-u).
\end{equation*}
Consider the function $F$ defined over $\Rset^2$ by $F(t, a) = \frac{1 - t}{2} \Phi'(a) - \frac{1+ t}{2} \Phi'(-a)$. Observe that $F(0, 0) = 0$ and that the partial derivative of $F$ with respect to $a$ at $(0, 0)$ is $\Phi''(0) > 0$:
\begin{equation*}
\frac{\partial F}{\partial a}(t, a) = \frac{1 - t}{2} \Phi''(a) + \frac{1 + t}{2} \Phi''(-a), \quad \frac{\partial F}{\partial a}(0, 0) = \Phi''(0) > 0.
\end{equation*}
Consequently, by the implicit function theorem, there exists a continuously differentiable function $\ov a$ such that $F (t, \ov a(t)) = 0$ in a neighborhood $[-\e, \e]$ around zero. Thus, by the convexity of $f_t$ and the definition of $F$, for $t \in [0, \e]$, $\inf_{u \in \Rset} f_t(u)$ is reached by $\ov a(t)$ and we can denote it by $a^*_t$. Then, $a^*_t$ is continuously differentiable over $[0, \epsilon]$. The minimizer $a^*_t$ satisfies the following equality:
\begin{equation}
\label{eq:a_t}
f'_t(a^*_t) = \frac{1 - t}{2} \Phi'(a^*_t) - \frac{1 + t}{2} \Phi'(-a^*_t) = 0.
\end{equation}
Specifically, at $t = 0$, we have $\Phi'(a^*_0) = \Phi'(-a^*_0)$.
Since $\Phi$ is convex, its derivative $\Phi'$ is non-decreasing.
Therefore, if $a^*_0$ were non-zero, then $\Phi'$ would be constant
over segment $[-\abs*{a^*_0}, \abs*{a^*_0}]$. This would contradict the
condition $\Phi''(0) > 0$, as a constant function cannot have a
positive second derivative at any point. Thus, we must have $a^*_0 =
0$ and since $\Phi'$ is non-decreasing and $\Phi''(0) > 0$, we have $a^*_t > 0$ for all $t \in (0, \e]$. By Theorem~\ref{thm:binary-char} and Taylor's theorem with an integral remainder, 
$\sT$ can be expressed as follows: for any $t \in [0, \e]$,
\begin{align}
\sT(t) 
& = f_t(0) -\inf_{u \in \Rset} f_t(u) \nonumber\\
& = f_t(0) - f_t(a^*_t) \nonumber\\
& = f'_t(a^*_t) (0 - a^*_t) + \int_{a^*_t}^0 (0 - u) f_t''(u) \, du
\tag{$f'_t(a^*_t) = 0$} \nonumber\\
& = \int_0^{a^*_t} u f''_t(u) \, du \nonumber\\
& = \int_0^{a^*_t} u \bracket*{\frac{1 - t}{2} \Phi''(u) + \frac{1 + t}{2} \Phi''(-u)} \, du.
\label{eq:T}
\end{align}
Since $a^*_t$ is a function of class $C^1$, we can differentiate \eqref{eq:a_t} with respect to $t$, which gives the following equality for any $t$ in $(0, \epsilon]$:
\begin{equation*}
-\frac12 \Phi'(a_t^*) + \frac{1 - t}{2 }\Phi''(a_t^*) \frac{d a_t^*}{d t}(t) - \frac12 \Phi'(-a_t^*) + \frac{1 + t}{2} \Phi''(-a_t^*)
\frac{d a_t^*}{d t}(t) = 0.
\end{equation*}
Taking the limit $t \to 0$ yields
\begin{equation*}
-\frac12 \Phi'(0) + \frac12 \Phi''(0) \frac{d a_t^*}{d t}(0) - \frac12 \Phi'(0) + \frac12 \Phi''(0) \frac{d a_t^*}{d t}(0) = 0.
\end{equation*}
This implies that
\begin{equation*}
\frac{d a_t^*}{d t}(0) = \frac{\Phi'(0)}{\Phi''(0)} > 0.
\end{equation*}
Since $\lim_{t \to 0} \frac{a_t^*}{t} = \frac{d a_t^*}{d t}(0) = \frac{\Phi'(0)}{\Phi''(0)} > 0$, we have $a^*_t = \Theta(t)$. 

Since $\Phi''(0) > 0$ and $\Phi''$ is continuous, there is a non-empty interval $[- \alpha, + \alpha]$ over which $\Phi''$ is positive. Since $a^*_0 = 0$ and $a^*_t$ is continuous, there exists a sub-interval $[0, \epsilon'] \subseteq [0, \epsilon]$ over which $a^*_t \leq \alpha$. Since $\Phi''$ is continuous, it admits a minimum and a maximum over any compact set and we can define $c = \min_{u \in [-\alpha, \alpha]} \Phi''(u)$ and $C = \max_{u \in [-\alpha, \alpha]} \Phi''(u)$. $c$ and $C$ are both positive since we have $\Phi''(0) > 0$. Thus, for $t$ in $[0, \epsilon']$, by \eqref{eq:T}, the following inequality holds:
\begin{align*}
C \frac{(a^*_t)^2}{2} = \int_0^{a^*_t}  u C \, du \geq \sT(t) &= \int_0^{a^*_t} u \bracket*{\frac{1 - t}{2} \Phi''(u) + \frac{1 + t}{2} \Phi''(-u)} \, du
\geq \int_0^{a^*_t}  u C \, du
= c \frac{(a^*_t)^2}{2}.
\end{align*}
This implies that $\sT(t) = \Theta(t^2)$.
\end{proof}

\section{Proof for comp-sum losses (Theorem~\ref{thm:comp-lower})}
\label{app:comp-lower}
\CompLower*

\begin{proof}
\ignore{
Observe that, for any $t \in [0, 1]$ and 
$\tau \geq 0$, we can write:
\begin{align*}
  \sup_{|u| \leq \tau} \curl*{
    \frac{1 - t}{2} \Phi(\tau + u) + \frac{1 + t}{2} \Phi\paren*{\tau - u}}
  & =  \sup_{|u| \leq 1} \curl*{
    \frac{1 - t}{2} \Phi\paren*{(1 + u) \tau} + \frac{1 + t}{2} \Phi\paren*{(1 - u)\tau}}.
\end{align*}
In light of this identity, for any $t \in [0, 1]$, $\tau \mapsto \sup_{|u| \leq \tau} \curl*{
  \frac{1 - t}{2} \Phi(\tau + u) + \frac{1 + t}{2} \Phi\paren*{\tau - u}}$ is continuous over $\Rset_+$, since the supremum over a fixed compact
set of a family of continuous functions is continuous.
  Thus, the infimum over the compact set $\bracket*{\frac1n, \frac12}$ is reached at some $\tau^* \in [\frac1n, \frac12]$, and the function $\sT$ can be rewritten as follows, for any $t \in [0, 1]$,
  \begin{equation}
  \label{eq:T-psi}
 \begin{aligned}
  \sT(t) 
  & =  \Phi(\tau^*) - \inf_{|u| \leq \tau^*}\curl*{ \frac{1 - t}{2}\Phi(\tau^* + u) + \frac{1 + t}{2} \Phi \paren*{\tau^* - u}}\\
  & = \Psi(0) - \inf_{|u| \leq \tau^*}\curl*{ \frac{1 - t}{2}\Psi(u) + \frac{1 + t}{2} \Psi \paren*{- u}},
 \end{aligned}
 \end{equation}
where $ u \mapsto \Psi(u) = \Phi(\tau^* + u)$ is convex, twice continuously differentiable, and satisfies the properties $\Psi'(0) < 0$ and $\Psi''(0) > 0$. Let $f_t(u) = \frac{1 - t}{2}\Psi(u) + \frac{1 + t}{2} \Psi \paren*{- u}$. Since $\Psi$ is convex and in $C^2$, $f_t$ is also convex and differentiable with respect to $u$. For any $t \in [0, 1]$, differentiate $f_t$ with respect to $u$, we have
\begin{equation*}
f'_t(u) = \frac{1 - t}{2} \Psi'(u) - \frac{1 + t}{2} \Psi'(-u).
\end{equation*}
Since $\Psi$ is convex, $\Psi'$ is non-decreasing, for any $t \in [0, 1]$, $f_t'$ is non-decreasing with respect to $u$. Plugging $u = -\tau^*$ and $u = \tau^*$ into the expression, we have
\begin{align*}
f'_t(-\tau^*) = \frac{1 - t}{2} \Psi'(-\tau^*) - \frac{1 + t}{2} \Psi'(\tau^*)\\
f'_t(\tau^*) = \frac{1 - t}{2} \Psi'(\tau^*) - \frac{1 + t}{2} \Psi'(-\tau^*).
\end{align*}
Since $\Psi'$ is non-decreasing and $\Psi''(0) > 0$, $\Psi'(\tau^*) > \Psi'(-\tau^*)$. Thus, we have 
\begin{align*}
f'_0(-\tau^*) &= \frac{1}{2} \Psi'(-\tau^*) - \frac{1}{2} \Psi'(\tau^*) < 0\\
f'_0(\tau^*) &= \frac{1}{2} \Psi'(\tau^*) - \frac{1}{2} \Psi'(-\tau^*) > 0.
\end{align*}
Since $f'_t(-\tau^*)$ and $f'_t(\tau^*)$ is continuous with respect to $t$, there exists $T > 0$ such that for any $0 \leq t < T$, $f'_t(-\tau^*) < 0$ and $f'_t(\tau^*) > 0$. Since $f_t'$ is non-decreasing with respect to $u$, for any $0 \leq t < T$, there exists $a^*_t \in (-\tau^*, \tau^*)$ such that $f_t(a^*_t) = 0$. Since $\Psi$ is convex, this implies that the infimum of $f_t$ over $u \in \Rset$ is reached within $(-\tau^*, \tau^*)$ and $a^*_t$ is the global minimizer. The minimizer $a^*_t$ satisfies the following equality:
\begin{equation}
\label{eq:a_t_comp}
f'_t(a^*_t) = \frac{1 - t}{2} \Psi'(a^*_t) - \frac{1 + t}{2} \Psi'(-a^*_t) = 0.
\end{equation}
Specifically, at $t = 0$, we have $\Psi'(a^*_0) = \Psi'(-a^*_0)$.
Since $\Psi$ is convex, its derivative $\Psi'$ is non-decreasing.
Therefore, if $a^*_0$ were non-zero, then $\Psi'$ would be constant
over the segment $[-\abs*{a^*_0}, \abs*{a^*_0}]$. This would contradict the
condition $\Psi''(0) > 0$, as a constant function cannot have a
positive second derivative at any point. Thus, we must have $a^*_0 =
0$ and since $\Psi'$ is non-decreasing and negative over $\bracket*{0, \frac12}$, and $\Psi''(0) > 0$, we have $a^*_t < 0$ for all $t \in (0, T)$. By \eqref{eq:T-psi} and Taylor's theorem with an integral remainder, 
$\sT$ can be expressed as follows: for any $t \in [0, T)$,
\begin{equation}
\label{eq:T-comp}
\begin{aligned}
\sT(t) 
& = f_t(0) - f_t(a^*_t)\\
& = f'_t(a^*_t) (0 - a^*_t) + \int_{a^*_t}^0 (0 - u) f_t''(u) \, du\\
& = \int_{a^*_t}^0 (0 - u) f''_t(u) \, du\\
& = \int_0^{-a^*_t} u f''_t(-u) \, du\\
& = \int_0^{-a^*_t} u \bracket*{\frac{1 - t}{2} \Psi''(-u) + \frac{1 + t}{2} \Psi''(u)} \, du.
\end{aligned}
\end{equation}
Consider the function $F$ defined over $\Rset^2$ by $F(t, a) = \frac{1 - t}{2} \Psi'(a) - \frac{1+ t}{2} \Psi'(-a)$. Observe that $F(0, 0) = 0$ and that the partial derivative of $F$ with respect to $a$ at $(0, 0)$ is $\Psi''(0) > 0$:
\begin{equation*}
\frac{\partial F}{\partial a}(t, a) = \frac{1 - t}{2} \Psi''(a) + \frac{1 + t}{2} \Psi''(-a), \quad \frac{\partial F}{\partial a}(0, 0) = \Psi''(0) > 0.
\end{equation*}
Consequently, by the implicit function theorem, there exists a continuously differentiable function $\ov a$ such that $F (t, \ov a(t)) = 0$ in a neighborhood $[-\e, \e]$ around zero. Thus, by the convexity of $f_t$ and the definition of $F$, for $t \in [0, \e] \subset [0, T)$, we have $\ov a(t) = \argmin_{u \in \Rset} f_t(u)$. Choose it as $a^*_t$ when the minimizer is not unique. Then, $a^*_t$ is continuously differentiable over $[0, \epsilon]$.

Since $a^*_t$ is in $C^1$, we can differentiate \eqref{eq:a_t_comp} with respect to $t$, which gives the following equality for any $t$ in $(0, \epsilon]$:
\begin{equation*}
-\frac12 \Psi'(a_t^*) + \frac{1 - t}{2 }\Psi''(a_t^*) \frac{d a_t^*}{d t}(t) - \frac12 \Psi'(-a_t^*) + \frac{1 + t}{2} \Psi''(-a_t^*)
\frac{d a_t^*}{d t}(t) = 0
\end{equation*}
By letting $t \to 0$, we have
\begin{equation*}
-\frac12 \Psi'(0) + \frac12 \Psi''(0) \frac{d a_t^*}{d t}(0) - \frac12 \Psi'(0) + \frac12 \Psi''(0) \frac{d a_t^*}{d t}(0) = 0.
\end{equation*}
This implies that
\begin{equation*}
\frac{d a_t^*}{d t}(0) = \frac{\Psi'(0)}{\Psi''(0)} < 0.
\end{equation*}
Since $\lim_{t \to 0} \frac{-a_t^*}{t} = -\frac{d a_t^*}{d t}(0) = -\frac{\Psi'(0)}{\Psi''(0)} > 0$, we have $-a^*_t = \Omega(t)$. 

Since $\Psi''(0) > 0$ and $\Psi''$ is continuous, there is a non-empty interval $[- \alpha, + \alpha]$ over which $\Psi''$ is positive. Since $a^*_0 = 0$ and $a^*_t$ is continuous, there exists a sub-interval $[0, \epsilon'] \subseteq [0, \epsilon]$ over which $a^*_t \leq \alpha$. Since $\Psi''$ is continuous, it admits a minimum over any compact set and we can define $C = \min_{u \in [-\e', \e']} \Psi''(u)$. $C$ is positive since we have $\Psi''(0) > 0$. Thus, for $t$ in $[0, \epsilon']$, by \eqref{eq:T-comp}, the following inequality holds:
\begin{align*}
\sT(t) &= \int_0^{-a^*_t} u \bracket*{\frac{1 - t}{2} \Psi''(-u) + \frac{1 + t}{2} \Psi''(u)} \, du
\geq \int_0^{-a^*_t}  u C \, du
= C \frac{(a^*_t)^2}{2}
= \Omega(t^2).
\end{align*}
This completes the proof.}
For any $\tau \in \bracket*{\frac1n, \frac12}$, define the function $\sT_\tau$ by
\begin{align*}
  \forall t \in [0, 1], \quad  \sT_\tau(t)
  & = \sup_{|u| \leq \tau} \curl*{\Phi(\tau) - \frac{1 - t}{2} \Phi(\tau + u) - \frac{1 + t}{2}\Phi(\tau - u)}\\
  & = f_{t, \tau}(0) - \inf_{|u| \leq \tau} f_{t, \tau}(u),
\end{align*}
where
\[
f_{t, \tau}(u)
= \frac{1 - t}{2} \Phi_{\tau}(u) + \frac{1 + t}{2} \Phi_{\tau}(-u)
\quad \text{and} \quad
\Phi_{\tau}(u) = \Phi(\tau + u).
\]
We aim to establish a lower bound for $\inf_{\tau \in \bracket*{\frac1n, \frac12}}
\sT_\tau(t)$.  For any fixed $\tau \in \bracket*{\frac1n, \frac12}$, this situation is
parallel to that of binary classification (Theorem~\ref{thm:binary-char}
and Theorem~\ref{thm:binary-lower}), since we have $\Phi'_{\tau}(0) =
\Phi'(\tau) < 0$ and $\Phi''_{\tau}(0) = \Phi''(\tau) > 0$.  Let $a^*_{t, \tau}$ denotes the minimizer of $f_{t, \tau}$ over $\Rset$. By applying Theorem~\ref{thm:a_implicit} to the function $F\colon (t, u, \tau) \mapsto f'_{t, \tau}(u) = \frac{1 - t}{2} \Phi'_{\tau}(u) - \frac{1 + t}{2} \Phi'_{\tau}(-u)$ and the convexity of $f_{t, \tau}$ with respect to $u$, $a^*_{t, \tau}$ exists, is unique and is
continuously differentiable over $[0, t'_0] \times \bracket*{\frac1n, \frac12}$, for some $t'_0 > 0$.
Moreover, by using the fact that $f'_{0, \tau}(\tau) > 0$ and $f'_{0, \tau}(-\tau) < 0$, and the convexity of $f_{0, \tau}$ with respect to $u$, we have $\abs*{a^*_{0, \tau}} \leq \tau$, $\forall \tau \in \bracket*{\frac1n, \frac12}$.  By the continuity of $a^*_{t, \tau}$, we have $\abs*{a^*_{t, \tau}} \leq \tau$ over $[0, t_0] \times \bracket*{\frac1n, \frac12}$, for some $t_0 > 0$ and $t_0 \leq t'_0$. 

Next, we will
leverage the proof of Theorem~\ref{thm:binary-lower}. Adopting a similar
notation, while incorporating the $\tau$ subscript to distinguish
different functions $\Phi_\tau$ and $f_{t, \tau}$, we can write
\[
\forall t \in [0, t_0], \quad
\sT_\tau(t) = \int_0^{-a^*_{t, \tau}} u \bracket*{\frac{1 - t}{2} \Phi''_{\tau}(-u) + \frac{1 + t}{2} \Phi''_{\tau}(u)} \, du.
\]
where $a^*_{t, \tau}$  verifies
\begin{equation}
\label{eq:DerivativeAtZero-comp}
a_{0, \tau}^* = 0 \quad \text{and} \quad
\frac{\partial a_{t, \tau}^*}{\partial t}(0) = \frac{\Phi'_{\tau}(0)}{\Phi''_{\tau}(0)} = c_\tau < 0.
\end{equation} 
We first show the lower bound $\inf_{\tau \in \bracket*{\frac1n, \frac12}} -a_{t,
  \tau}^* = \Omega(t)$.
Given the equalities \eqref{eq:DerivativeAtZero-comp}, it follows that for
any $\tau$, the following holds: $\lim_{t \to 0} \paren*{-a_{t, \tau}^*
  + c_\tau t} = 0$. For any $\tau \in \bracket*{\frac1n, \frac12}$, $t \mapsto
\paren*{-a_{t, \tau}^* + c_\tau t}$ is a continuous function over $[0,
  t_0]$ since $a_{t, \tau}^*$ is a function of class $C^1$. Since the
infimum over a fixed compact set of a family of continuous functions is
continuous, $t \mapsto \inf_{\tau \in \bracket*{\frac1n, \frac12}}\curl*{-a_{t, \tau}^* +
  c_\tau t}$ is continuous.  Thus, for any $\e > 0$, there exists $t_1
> 0$, $t_1 \leq t_0$, such that for any $t \in [0, t_1]$,
\[
\abs*{\inf_{\tau \in \bracket*{\frac1n, \frac12}} \curl*{-a_{t, \tau}^* + c_\tau t}} \leq \e,
\]
which implies
\[
\forall \tau \in \bracket*{\frac1n, \frac12}, \quad
-a_{t, \tau}^*
\geq -c_\tau t - \e
\geq c t - \e,
\]
where $c = \inf_{\tau \in \bracket*{\frac1n, \frac12}} -c_\tau$. Since $\Phi'_{\tau}(0)$ and
$\Phi''_{\tau}(0)$ are positive and continuous functions of $\tau$,
this infimum is attained over the compact set $\bracket*{\frac1n, \frac12}$, leading to $c
> 0$. Since the lower bound holds uniformly over $\tau$, this shows
that for $t \in [0, t_1]$, we have $\inf_{\tau \in \bracket*{\frac1n, \frac12}} -a_{t,
  \tau}^* = \Omega(t)$.

Now, since for any $\tau \in \bracket*{\frac1n, \frac12}$, $-a_{t, \tau}^*$ is a function of
class $C^1$ and thus continuous, its supremum over a compact set,
$\sup_{\tau \in \bracket*{\frac1n, \frac12}} -a_{t, \tau}^*$, is also continuous and is
bounded over $[0, t_1]$ by some $a > 0$. For $|u| \leq a$ and $\tau
\in \bracket*{\frac1n, \frac12}$, we have $\frac12 - a \leq \tau + u \leq \frac12 + a$ and $\frac12 - a \leq
\tau - u \leq \frac12 + a$. Since $\Phi''$ is positive and continuous, it
reaches its minimum $C > 0$ over the compact set $\bracket*{\frac12 - a, \frac12 + a}$.
Thus, we can write 
\begin{align*}
\forall t \in [0, t_1], \forall \tau \in \bracket*{\frac1n, \frac12}, \quad
   \sT_\tau(t)
  & = \int_0^{-a^*_{t, \tau}} u \bracket*{\frac{1 - t}{2} \Phi''_{\tau}(-u) + \frac{1 + t}{2} \Phi''_{\tau}(u)} \, du\\
  & \geq \int_0^{-a^*_{t, \tau}} u \bracket*{\frac{1 - t}{2} C + \frac{1 + t}{2} C} \, du\\
  & = \int_0^{-a^*_{t, \tau}} C u  \, du
  = C \frac{(-a^{*}_{t, \tau})^2}{2}.
\end{align*}
Thus, for $t \leq t_1$, we have
\[
\inf_{\tau \in \bracket*{\frac1n, \frac12}} \sT_\tau(t) \geq C
\frac{(\inf_{\tau \in \bracket*{\frac1n, \frac12}} -a^{*}_{t, \tau})^2}{2} \geq \Omega(t^2).
\]
Similarly, we aim to establish an upper bound for $\inf_{\tau \in \bracket*{\frac1n, \frac12}}
\sT_\tau(t)$.  We first show the upper bound $\sup_{\tau \in \bracket*{\frac1n, \frac12}} -a_{t,
  \tau}^* = O(t)$.
Given the equalities \eqref{eq:DerivativeAtZero-comp}, it follows that for
any $\tau$, the following holds: $\lim_{t \to 0} \paren*{-a_{t, \tau}^*
  + c_\tau t} = 0$. For any $\tau \in \bracket*{\frac1n, \frac12}$, $t \mapsto
\paren*{-a_{t, \tau}^* + c_\tau t}$ is a continuous function over $[0,
  t_0]$ since $a_{t, \tau}^*$ is a function of class $C^1$. Since the
supremum over a fixed compact set of a family of continuous functions is
continuous, $t \mapsto \sup_{\tau \in \bracket*{\frac1n, \frac12}}\curl*{-a_{t, \tau}^* +
  c_\tau t}$ is continuous.  Thus, for any $\e > 0$, there exists $t_1
> 0$, $t_1 \leq t_0$, such that for any $t \in [0, t_1]$,
\[
\abs[\Big]{\sup_{\tau \in \bracket*{\frac1n, \frac12}} \curl*{-a_{t, \tau}^* + c_\tau t}} \leq \e,
\]
which implies
\[
\forall \tau \in \bracket*{\frac1n, \frac12}, \quad
-a_{t, \tau}^*
\leq -c_\tau t + \e
\leq c t + \e,
\]
where $c = \sup_{\tau \in \bracket*{\frac1n, \frac12}} -c_\tau$. Since $\Phi'_{\tau}(0)$ and
$\Phi''_{\tau}(0)$ are positive and continuous functions of $\tau$,
this supremum is attained over the compact set $\bracket*{\frac1n, \frac12}$, leading to $c
> 0$. Since the upper bound holds uniformly over $\tau$, this shows
that for $t \in [0, t_1]$, we have $\sup_{\tau \in \bracket*{\frac1n, \frac12}} -a_{t,
  \tau}^* = O(t)$.

Now, since for any $\tau \in \bracket*{\frac1n, \frac12}$, $-a_{t, \tau}^*$ is a function of
class $C^1$ and thus continuous, its supremum over a compact set,
$\sup_{\tau \in \bracket*{\frac1n, \frac12}} -a_{t, \tau}^*$, is also continuous and is
bounded over $[0, t_1]$ by some $a > 0$. For $|u| \leq a$ and $\tau
\in \bracket*{\frac1n, \frac12}$, we have $\frac12 - a \leq \tau + u \leq \frac12 + a$ and $\frac12 - a \leq
\tau - u \leq \frac12 + a$. Since $\Phi''$ is positive and continuous, it
reaches its maximum $C > 0$ over the compact set $\bracket*{\frac12 - a, \frac12 + a}$.
Thus, we can write
\begin{align*}
  \forall t \in [0, t_1], \forall \tau \in \bracket*{\frac1n, \frac12}, \quad
  \sT_\tau(t)
  & = \int_0^{-a^*_{t, \tau}} u \bracket*{\frac{1 - t}{2} \Phi''_{\tau}(-u) + \frac{1 + t}{2} \Phi''_{\tau}(u)} \, du\\
  & \leq \int_0^{-a^*_{t, \tau}} u \bracket*{\frac{1 - t}{2} C + \frac{1 + t}{2} C} \, du\\
  & = \int_0^{-a^*_{t, \tau}} C u  \, du
  = C \frac{(-a^{*}_{t, \tau})^2}{2}.
\end{align*}
Thus, for $t \leq t_1$, we have
\[
\inf_{\tau \in \bracket*{\frac1n, \frac12}} \sT_\tau(t) \leq C
\frac{(\sup_{\tau \in \bracket*{\frac1n, \frac12}} -a^{*}_{t, \tau})^2}{2} \leq O(t^2).
\]
This completes the proof.
\end{proof}

\section{Proof for constrained losses (Theorem~\ref{thm:cstnd-lower})}
\label{app:cstnd-lower}
\CstndLower*
\begin{proof}
\ignore{
Define the function $G$ as $G \colon (t, \tau, u) \mapsto \Phi(\tau) -
\frac{1 - t}{2} \Phi(\tau + u) - \frac{1 + t}{2}\Phi(\tau - u)$. Since
$\Phi$ is convex and differentiable, for any $(t, \tau) \in [0, 1]
\times \Rset_{+}$, $u \mapsto G(t, \tau, u)$ is concave and
differentiable. For any $(t, \tau) \in [0, 1] \times \Rset_{+}$,
differentiating $G$ with respect to $u$, we have
\begin{equation*}
\frac{\partial G}{\partial u}(t, \tau, u) = - \frac{1 - t}{2} \Phi'(\tau + u) + \frac{1 + t}{2} \Phi'(\tau - u).
\end{equation*}
Since $\Phi$ is convex, $\Phi'$ is non-decreasing, thus, for any $(t,
\tau) \in [0, 1] \times \Rset_{+}$, $u \mapsto \frac{\partial
  G}{\partial u}(t, \tau, u)$ is non-increasing. Fix $a > 0$. Plugging
$u = -a$ and $u = a$ into the expression gives
\begin{align*}
\frac{\partial G}{\partial u}(t, \tau, -a) &= - \frac{1 - t}{2} \Phi'(\tau - a) + \frac{1 + t}{2} \Phi'(\tau + a)\\
\frac{\partial G}{\partial u}(t, \tau, a) &= - \frac{1 - t}{2} \Phi'(\tau + a) + \frac{1 + t}{2} \Phi'(\tau - a).
\end{align*}
Thus, since $\Phi'$ is strictly increasing over $\Rset_{+}$ ($\Phi''(t) > 0$ for any $t \geq 0$) and $\tau + a > \max \curl*{0, \tau - a}$, we have $\Phi'(\tau + a) > \max \curl*{0, \Phi'(\tau - a)}$. Therefore, for any $\tau \geq 0$,
\begin{align*}
\forall t \in [0, 1],\, \frac{\partial G}{\partial u}(t, \tau, -a) &\geq \max \curl*{t \Phi'(\tau + a), \frac{1 - t}{2} \paren*{\Phi'(\tau + a) - \Phi'(\tau -a)}} > 0\\
\frac{\partial G}{\partial u}(0, \tau, a) &= \frac{1}{2} \paren*{\Phi'(\tau - a) - \Phi'(\tau + a)} < 0.
\end{align*}
Since for any $\tau \geq 0$, $\frac{\partial G}{\partial u}(0, \tau,
a) < 0$ and $t \mapsto \frac{\partial G}{\partial u}(t, \tau, a)$ is
continuous, there exists $t_0 > 0$ such that for any $0 \leq t < t_0$,
$\frac{\partial G}{\partial u}(t, \tau, a) < 0$. Since for any $\tau
\geq 0$ and $0 \leq t < t_0$, $u \mapsto \frac{\partial G}{\partial
  u}(t, \tau, u)$ is non-increasing, there exists some $u_0 \in [-a,
  a]$ satisfying $\frac{\partial G}{\partial u}(t, \tau, u_0) =
0$. Since for any $\tau \geq 0$ and $0 \leq t < t_0$, $u \mapsto G(t,
\tau, u)$ is concave, $u \mapsto G(t, \tau, u)$ achieves the supremum
within $[-a, a]$. Therefore, there exist positive constants $t_0 > 0$
and $a > 0$, such that for any $\tau \geq 0$ and $0 \leq t < t_0$,
\begin{equation*}
\sup_{u \in \Rset} \curl*{ \Phi(\tau) - \frac{1 - t}{2} \Phi(\tau + u) - \frac{1 + t}{2}\Phi(\tau - u) } = \sup_{u \in \Rset} G(t, \tau, u) = \sup_{|u| \leq a} G(t, \tau, u).
\end{equation*}
Consider the function $t \in [0, t_0) \mapsto \inf_{\tau \in [0, A]}
  \sup_{|u| \leq a} G(t, \tau, u)$. For any $ t \in [0, t_0)$, $(\tau,
    u) \mapsto G(t, \tau, u)$ is a continuous function. Since the
    supremum over a fixed compact set of a jointly continuous function
    is continuous, for any $t \in [0, t_0)$, $\tau \mapsto \sup_{|u|
        \leq a} G(t, \tau, u)$ is well defined and is a continuous
      function. Since $\tau \mapsto \sup_{|u| \leq a} G(t, \tau, u)$
      is continuous over $[0, +\infty)$, its infimum is reached over
        $[0, A]$, at some $\tau^*_t \in [0, A]$.
Thus, the function $\sU\colon t \in [0, t_0) \mapsto \inf_{\tau \in [0, A]}
  \sup_{u \in \Rset} G(t, \tau, u)$ can be equivalently expressed
  as follows.
}
For any $\tau \in [0, A]$, define the function $\sT_\tau$ by
\begin{align*}
  \forall t \in [0, 1], \quad  \sT_\tau(t)
  & = \sup_{u \in \Rset} \curl*{\Phi(\tau) - \frac{1 - t}{2} \Phi(\tau + u) - \frac{1 + t}{2}\Phi(\tau - u)}\\
  & = f_{t, \tau}(0) - \inf_{u \in \Rset} f_{t, \tau}(u),
\end{align*}
where
\[
f_{t, \tau}(u)
= \frac{1 - t}{2} \Phi_{\tau}(u) + \frac{1 + t}{2} \Phi_{\tau}(-u)
\quad \text{and} \quad
\Phi_{\tau}(u) = \Phi(\tau + u).
\]
We aim to establish a lower bound for $\inf_{\tau \in [0, A]}
\sT_\tau(t)$.  For any fixed $\tau \in [0, A]$, this situation is
parallel to that of binary classification (Theorem~\ref{thm:binary-char}
and Theorem~\ref{thm:binary-lower}), since we also have $\Phi'_{\tau}(0) =
\Phi'(\tau) > 0$ and $\Phi''_{\tau}(0) = \Phi''(\tau) > 0$.  Let $a^*_{t, \tau}$ denotes the minimizer of $f_{t, \tau}$ over $\Rset$. By applying Theorem~\ref{thm:a_implicit} to the function $F\colon (t, u, \tau) \mapsto f'_{t, \tau}(u) = \frac{1 - t}{2} \Phi'_{\tau}(u) - \frac{1 + t}{2} \Phi'_{\tau}(-u)$ and the convexity of $f_{t, \tau}$ with respect to $u$, $a^*_{t, \tau}$ exists, is unique and is
continuously differentiable over $[0, t_0] \times [0, A]$, for some $t_0 > 0$. 

Next, we will
leverage the proof of Theorem~\ref{thm:binary-lower}. Adopting a similar
notation, while incorporating the $\tau$ subscript to distinguish
different functions $\Phi_\tau$ and $f_{t, \tau}$, we can write
\[
\forall t \in [0, t_0], \quad
\sT_\tau(t) = \int_0^{a^*_{t, \tau}} u \bracket*{\frac{1 - t}{2} \Phi''_{\tau}(u) + \frac{1 + t}{2} \Phi''_{\tau}(-u)} \, du.
\]
where $a^*_{t, \tau}$  verifies
\begin{equation}
\label{eq:DerivativeAtZero}
a_{0, \tau}^* = 0 \quad \text{and} \quad
\frac{\partial a_{t, \tau}^*}{\partial t}(0) = \frac{\Phi'_{\tau}(0)}{\Phi''_{\tau}(0)} = c_\tau > 0.
\end{equation} 
We first show the lower bound $\inf_{\tau \in [0, A]} a_{t,
  \tau}^* = \Omega(t)$.
Given the equalities \eqref{eq:DerivativeAtZero}, it follows that for
any $\tau$, the following holds: $\lim_{t \to 0} \paren*{a_{t, \tau}^*
  - c_\tau t} = 0$. For any $\tau \in [0, A]$, $t \mapsto
\paren*{a_{t, \tau}^* - c_\tau t}$ is a continuous function over $[0,
  t_0]$ since $a_{t, \tau}^*$ is a function of class $C^1$. Since the
infimum over a fixed compact set of a family of continuous functions is
continuous, $t \mapsto \inf_{\tau \in [0, A]}\curl*{a_{t, \tau}^* -
  c_\tau t}$ is continuous.  Thus, for any $\e > 0$, there exists $t_1
> 0$, $t_1 \leq t_0$, such that for any $t \in [0, t_1]$,
\[
\abs[\Big]{\inf_{\tau \in [0, A]} \curl*{a_{t, \tau}^* - c_\tau t}} \leq \e,
\]
which implies
\[
\forall \tau \in [0, A], \quad
a_{t, \tau}^*
\geq c_\tau t - \e
\geq c t - \e,
\]
where $c = \inf_{\tau \in [0, A]} c_\tau$. Since $\Phi'_{\tau}(0)$ and
$\Phi''_{\tau}(0)$ are positive and continuous functions of $\tau$,
this infimum is attained over the compact set $[0, A]$, leading to $c
> 0$. Since the lower bound holds uniformly over $\tau$, this shows
that for $t \in [0, t_1]$, we have $\inf_{\tau \in [0, A]} a_{t,
  \tau}^* = \Omega(t)$.

Now, since for any $\tau \in [0, A]$, $a_{t, \tau}^*$ is a function of
class $C^1$ and thus continuous, its supremum over a compact set,
$\sup_{\tau \in [0, A]} a_{t, \tau}^*$, is also continuous and is
bounded over $[0, t_1]$ by some $a > 0$. For $|u| \leq a$ and $\tau
\in [0, A]$, we have $A - a \leq \tau + u \leq A + a$ and $A - a \leq
\tau - u \leq A + a$. Since $\Phi''$ is positive and continuous, it
reaches its minimum $C > 0$ over the compact set $[A - a, A + a]$.
Thus, we can write
\begin{align*}
  \forall t \in [0, t_1], \forall \tau \in [0, A], \quad
  \sT_\tau(t)
  & = \int_0^{a^*_{t, \tau}} u \bracket*{\frac{1 - t}{2} \Phi''_{\tau}(u) + \frac{1 + t}{2} \Phi''_{\tau}(-u)} \, du\\
  & \geq \int_0^{a^*_{t, \tau}} u \bracket*{\frac{1 - t}{2} C + \frac{1 + t}{2} C} \, du\\
  & = \int_0^{a^*_{t, \tau}} C u  \, du
  = C \frac{(a^{*}_{t, \tau})^2}{2}.
\end{align*}
Thus, for $t \leq t_1$, we have
\[
\inf_{\tau \in [0, A]} \sT_\tau(t) \geq C
\frac{(\inf_{\tau \in [0, A]} a^{*}_{t, \tau})^2}{2} \geq \Omega(t^2).
\]
Similarly, we aim to establish an upper bound for $\inf_{\tau \in [0, A]}
\sT_\tau(t)$. We first show the upper bound $\sup_{\tau \in [0, A]} a_{t,
  \tau}^* = O(t)$.
Given the equalities \eqref{eq:DerivativeAtZero}, it follows that for
any $\tau$, the following holds: $\lim_{t \to 0} \paren*{a_{t, \tau}^*
  - c_\tau t} = 0$. For any $\tau \in [0, A]$, $t \mapsto
\paren*{a_{t, \tau}^* - c_\tau t}$ is a continuous function over $[0,
  t_0]$ since $a_{t, \tau}^*$ is a function of class $C^1$. Since the
supremum over a fixed compact set of a family of continuous functions is
continuous, $t \mapsto \sup_{\tau \in [0, A]}\curl*{a_{t, \tau}^* -
  c_\tau t}$ is continuous.  Thus, for any $\e > 0$, there exists $t_1
> 0$, $t_1 \leq t_0$, such that for any $t \in [0, t_1]$,
\[
\abs[\Big]{\sup_{\tau \in [0, A]} \curl*{a_{t, \tau}^* - c_\tau t}} \leq \e,
\]
which implies
\[
\forall \tau \in [0, A], \quad
a_{t, \tau}^*
\leq c_\tau t + \e
\leq c t + \e,
\]
where $c = \sup_{\tau \in [0, A]} c_\tau$. Since $\Phi'_{\tau}(0)$ and
$\Phi''_{\tau}(0)$ are positive and continuous functions of $\tau$,
this supremum is attained over the compact set $[0, A]$, leading to $c
> 0$. Since the upper bound holds uniformly over $\tau$, this shows
that for $t \in [0, t_1]$, we have $\sup_{\tau \in [0, A]} a_{t,
  \tau}^* = O(t)$.

Now, since for any $\tau \in [0, A]$, $a_{t, \tau}^*$ is a function of
class $C^1$ and thus continuous, its supremum over a compact set,
$\sup_{\tau \in [0, A]} a_{t, \tau}^*$, is also continuous and is
bounded over $[0, t_1]$ by some $a > 0$. For $|u| \leq a$ and $\tau
\in [0, A]$, we have $A - a \leq \tau + u \leq A + a$ and $A - a \leq
\tau - u \leq A + a$. Since $\Phi''$ is positive and continuous, it
reaches its maximum $C > 0$ over the compact set $[A - a, A + a]$.
Thus, we can write
\begin{align*}
  \forall t \in [0, t_1], \forall \tau \in [0, A], \quad
  \sT_\tau(t)
  & = \int_0^{a^*_{t, \tau}} u \bracket*{\frac{1 - t}{2} \Phi''_{\tau}(u) + \frac{1 + t}{2} \Phi''_{\tau}(-u)} \, du\\
  & \leq \int_0^{a^*_{t, \tau}} u \bracket*{\frac{1 - t}{2} C + \frac{1 + t}{2} C} \, du\\
  & = \int_0^{a^*_{t, \tau}} C u  \, du
  = C \frac{(a^{*}_{t, \tau})^2}{2}.
\end{align*}
Thus, for $t \leq t_1$, we have
\[
\inf_{\tau \in [0, A]} \sT_\tau(t) \leq C
\frac{(\sup_{\tau \in [0, A]} a^{*}_{t, \tau})^2}{2} \leq O(t^2).
\]
This completes the proof.
\end{proof}

\section{Analysis of the function of \texorpdfstring{$\tau$}{tau}}
\label{app:analysis}

Let $F$ be the function defined by
\[
\forall t \in \bracket*{0, \tfrac{1}{2}}, \tau \in \Rset, \quad
F(t, \tau)
= \sup_{u \in \Rset} \curl*{\Phi(\tau) - \frac{1 - t}{2} \Phi(\tau + u) - \frac{1 + t}{2} \Phi(\tau - u)},
\]
where $\Phi$ is a convex function in $C^2$ with $\Phi', \Phi'' >
0$. In light of the analysis of the previous sections, for any $(\tau,
t)$, there exists a unique function $a_{t, \tau}$ solution of the
maximization (supremum in $F$), a $C^1$ function over a neighborhood $U$
of $(\tau, 0)$ with $a_{0, \tau} = 0$, $a_{t, \tau} > 0$ for $t > 0$, and
$\frac{\partial a_{t, \tau}}{\partial t}(0, \tau)
= \frac{\Phi'(\tau)}{\Phi''(\tau)} = c_\tau$. Thus, we have $\lim_{t \to 0} \frac{a_{t, \tau}}{t c_\tau} = 1$. The optimality of
$a_{t, \tau}$ implies
\[
\frac{1 - t}{2} \Phi'(\tau + a_{t, \tau}) = \frac{1 + t}{2} \Phi'(\tau - a_{t, \tau}).
\]
Thus, the partial derivative of $F$ over the appropriate neighborhood $U$
is given by
\begin{align*}
  \frac{\partial F}{\partial \tau}(t, \tau)
  & = \Phi'(\tau) - \frac{1 - t}{2} \Phi'(\tau + a_{t, \tau}) \paren*{\frac{\partial a_{t, \tau}}{\partial \tau}(t, \tau) + 1} - \frac{1 + t}{2} \Phi'(\tau - a_{t, \tau}) \paren*{-\frac{\partial a_{t, \tau}}{\partial \tau}(t, \tau) + 1}\\
  & = \Phi'(\tau) - \frac{1 - t}{2} \Phi'(\tau + a_{t, \tau}) \paren*{\frac{\partial a_{t, \tau}}{\partial \tau}(t, \tau) + 1 - \frac{\partial a_{t, \tau}}{\partial \tau}(t, \tau) + 1} \\
  & = \Phi'(\tau) - (1 - t) \Phi'(\tau + a_{t, \tau}).
\end{align*}
Since $\Phi'$ is continuous, by the mean value theorem, there exists
$\xi \in (\tau, \tau + a_{t, \tau})$ such that $\Phi'(\tau + a_{t,
  \tau}) - \Phi'(\tau) = a_{t, \tau} \Phi''(\xi)$. Thus, we can write
\begin{align*}
  \frac{\partial F}{\partial \tau}(t, \tau)
  & = \Phi'(\tau) - (1 - t) \Phi'(\tau) - (1 - t)  a_{t, \tau} \Phi''(\xi)\\
  & = t \Phi'(\tau) - (1 - t)  a_{t, \tau} \Phi''(\xi)\\
  & = t \Phi'(\tau) \bracket*{1 - (1 - t) \frac{a_{t, \tau}}{t c_\tau} \frac{\Phi''(\xi)}{ \Phi''(\tau)}}.
\end{align*}
Note that if $\Phi''$ is locally non-increasing, then we have $\Phi''(\xi) \leq \Phi''(\tau)$ and for $t$ sufficiently small, since $\Phi'$ is increasing and $\frac{a_{t, \tau}}{t c_\tau} \sim 1$:
\begin{align}
  \frac{\partial F}{\partial \tau}(t, \tau)
  & \geq t \Phi'(\tau) \bracket*{1 - (1 - t) \frac{a_{t, \tau}}{t c_\tau} }
  \geq 0.
\end{align}
In that case, for any $A > 0$, we can find a neighborhood $\sO$ of $t$
around zero over which $\frac{\partial F}{\partial \tau}(t, \tau)$ is
defined for all $(t, \tau) \in \sO \times [0, A]$ and $\frac{\partial
  F}{\partial \tau}(t, \tau) \geq 0$. From this, we can conclude that
the infimum of $F$ over $\tau \in [0, A]$ is reached at zero for $t$ sufficiently
small ($t \in \sO$).

\section{Generalization bounds}
\label{app:generalization-bound}

Let $S = \paren*{(x_1, y_1), \ldots, (x_m, y_m)}$ be a sample drawn
from $\sD^m$. Denote by $\h h_S$ an empirical minimizer within $\sH$
for the surrogate loss $\ell$: $ \h h_S \in \argmin_{h\in \sH}
\frac{1}{m}\sum_{i = 1}^m \ell (h, x_i, y_i)$. Let
$\sH_{\ell}$ denote the hypothesis set $\curl*{(x, y) \mapsto \ell(h, x, y)
  \colon h \in \sH}$ and $\Rad_m^{\ell}(\sH)$ its Rademacher
complexity. We also write $B_{\ell}$ to denote an upper bound for
$\ell$. Then, given the following $\sH$-consistency bound:
\begin{equation}
\label{eq:H-consistency-bounds}
\forall h \in \sH, \quad
\sE_{\ell_{0-1}}(h) - \sE^*_{\ell_{0-1}}(\sH) + \sM_{\ell_{0-1}}(\sH)
\leq \Gamma\paren*{\sE_{\ell}(h)-\sE^*_{\ell}(\sH) + \sM_{\ell}(\sH)},
\end{equation}
for any
$\delta > 0$, with probability at least $1 - \delta$ over the draw of an
i.i.d.\ sample $S$ of size $m$, the following estimation
bound holds for $\h h_S$:
\begin{equation*}
\forall h \in \sH, \quad
\sE_{\ell_{0-1}}(h) - \sE^*_{\ell_{0-1}}(\sH) 
\leq \Gamma\paren[\bigg]{4
    \Rad_m^{\sfL}(\sH) + 2 B_{\sfL} \sqrt{\tfrac{\log
        \frac{2}{\delta}}{2m}} + \sM_{\ell}(\sH)} - \sM_{\ell_{0-1}}(\sH).
\end{equation*}
\begin{proof}
  By the standard Rademacher complexity bounds
  \citep{MohriRostamizadehTalwalkar2018}, for any $\delta > 0$, with
  probability at least $1 - \delta$, the following holds for all $h
  \in \sH$:
\[
\abs*{\sE_{\ell}(h) - \h\sE_{\ell,S}(h)}
\leq 2 \Rad_m^{\ell}(\sH) +
B_{\ell} \sqrt{\tfrac{\log (2/\delta)}{2m}}.
\]
For any $\e > 0$, by definition of the infimum, there exists $h^* \in
\sH$ such that $\sE_{\ell}(h^*) \leq
\sE_{\ell}^*(\sH) + \e$. By the definition of
$\h h_S$, we obtain
\begin{align*}
  \sE_{\ell}(\h h_S) - \sE_{\ell}^*(\sH)
  & = \sE_{\ell}(\h h_S) - \h\sE_{\ell,S}(\h h_S) + \h\sE_{\ell,S}(\h h_S) - \sE_{\ell}^*(\sH)\\
  & \leq \sE_{\ell}(\h h_S) - \h\sE_{\ell,S}(\h h_S) + \h\sE_{\ell,S}(h^*) - \sE_{\ell}^*(\sH)\\
  & \leq \sE_{\ell}(\h h_S) - \h\sE_{\ell,S}(\h h_S) + \h\sE_{\ell,S}(h^*) - \sE_{\ell}^*(h^*) + \e\\
  & \leq
  2 \bracket*{2 \Rad_m^{\ell}(\sH) +
B_{\ell} \sqrt{\tfrac{\log (2/\delta)}{2m}}} + \e.
\end{align*}
Since the inequality holds for all $\e > 0$, it implies the following:
\[
\sE_{\ell}(\h h_S) - \sE_{\ell}^*(\sH)
\leq 
4 \Rad_m^{\ell}(\sH) +
2 B_{\ell} \sqrt{\tfrac{\log (2/\delta)}{2m}}.
\]
Plugging in this inequality in the $\sH$-consistency bound
\eqref{eq:H-consistency-bounds} completes the proof.
\end{proof}
These bounds for surrogate loss minimizers, expressed in terms of
minimizability gaps, offer more detailed and informative insights
compared to existing bounds based solely on approximation errors. Our
analysis of growth rates suggests that for commonly used smooth loss
functions, $\Gamma$ varies near zero with a square-root dependency.
Furthermore, this dependency cannot be generally improved for
arbitrary distributions.
\restoreatoc



\cleardoublepage
\phantomsection


 
\addtocontents{toc}{\protect\setcounter{tocdepth}{0}}

\disableatoc
\addcontentsline{toc}{chapter}{Bibliography}
\restoreatoc

\printbibliography

\addtocontents{toc}{\protect\setcounter{tocdepth}{2}}
\end{document}